\documentclass{article}

\usepackage{PRIMEarxiv}
\usepackage{listings}
\usepackage{longtable}
\usepackage{amsmath} 

\usepackage[utf8]{inputenc} 
\usepackage[T1]{fontenc}    
\usepackage{hyperref}
\hypersetup{
    pdflinkmargin=2pt, 
    colorlinks=true,    
    linkcolor=blue,     
    citecolor=[rgb]{0,0.4,0},    
    urlcolor=magenta    
}
\usepackage{url}            
\usepackage{booktabs}       
\usepackage{amsfonts}       
\usepackage{nicefrac}       
\usepackage{microtype}      
\usepackage{lipsum}
\usepackage{fancyhdr}       
\usepackage{graphicx}       
\graphicspath{{media/}}     
\lstdefinestyle{mystyle}{
    basicstyle=\ttfamily\small,
    breaklines=true,         
    frame=single,            
    columns=fullflexible,    
    keepspaces=true,         
    showspaces=false,        
    showstringspaces=false,  
    showtabs=false,          
    tabsize=2                
}
\title{\textnormal{Enhancing LLM Problem Solving with REAP: Reflection, Explicit Problem Deconstruction, and Advanced Prompting}}

\author{
  \textbf{Ryan Lingo} \\ 
  Honda Research Institute USA, Inc. \\ 
  \texttt{ryan\_lingo@honda-ri.com} 
  \and
  \textbf{Martin Arroyo} \\ 
  Honda Research Institute USA, Inc. \\ 
  \texttt{martin\_arroyo@honda-ri.com}
  \vspace{0.25cm} 
  \and
  \textbf{Rajeev Chhajer} \\ 
  Honda Research Institute USA, Inc. \\ 
  \texttt{rajeev\_chhajer@honda-ri.com}
}

\begin{document}
\maketitle

\begin{abstract}
Large Language Models (LLMs) have transformed natural language processing, yet improving their problem-solving capabilities, particularly for complex, reasoning-intensive tasks, remains a persistent challenge. This paper introduces the REAP (Reflection, Explicit Problem Deconstruction, and Advanced Prompting) method, an innovative approach within the dynamic context generation framework. REAP guides LLMs through reflection on the query, deconstructing it into manageable components, and generating relevant context to enhance the solution process. We evaluated REAP using a dataset designed to expose LLM limitations, comparing zero-shot prompting with REAP-enhanced prompts across six state-of-the-art models: OpenAI’s o1-preview, o1-mini, GPT-4o, GPT-4o-mini, Google’s Gemini 1.5 Pro, and Claude 3.5 Sonnet. The results demonstrate notable performance gains, with o1-mini improving by 40.97\%, GPT-4o by 66.26\%, and GPT-4o-mini by 112.93\%. Despite the already strong baseline performance of OpenAI’s o1-preview, modest gains were observed. Beyond performance improvements, REAP offers a cost-effective solution; for example, GPT-4o-mini, which is approximately 100 times cheaper than o1-preview, delivered competitive results.REAP also improves the clarity of model outputs, making it easier for humans to understand the reasoning behind the results and simplifying the process of identifying and addressing any issues. These findings demonstrate REAP’s potential to greatly improve the capabilities of LLMs, providing both better performance and increased cost-efficiency across a wide range of applications.
\end{abstract}


\section{Introduction}

\subsection{Context and Importance}
Large Language Models (LLMs), such as GPT-3 \cite{brown2020language}, GPT-4 \cite{openai2023gpt4}, and BERT \cite{devlin2018bert}, have become foundational tools in artificial intelligence, particularly in natural language processing. These models demonstrate a high level of proficiency in generating text, understanding context, and performing a broad range of language-related tasks with notable accuracy. Their development has influenced areas such as machine translation, text summarization, and conversational AI, making them integral to modern AI systems.

However, enhancing the precision, coherence, and contextual relevance of LLM-generated outputs remains a persistent challenge. The need for outputs that consistently meet high standards of clarity and accuracy drives ongoing research in this field. Effective and reliable outputs are critical for ensuring robust performance across diverse applications, where precision and clarity are indispensable.

\subsection{REAP: A Structured Methodology for Enhancing LLM Problem-Solving}
REAP (Reflection, Explicit Problem Deconstruction, and Advanced Prompting) is a systematic approach developed to improve LLM problem-solving abilities. It integrates three key components:

\begin{itemize}
    \item \textbf{Reflection}: Facilitates continuous feedback and reassessment during the problem-solving process. By reflecting on new information, the LLM progressively refines its approach, producing more accurate and well-informed results.
    \item \textbf{Explicit Problem Deconstruction}: Breaks down complex tasks into smaller, manageable units. This structured analysis improves the LLM’s understanding by addressing each element in a stepwise manner, ensuring clarity at each stage.
    \item \textbf{Advanced Prompting}: Directs the LLM’s reasoning through a combination of strategies that explore multiple solution pathways. This method fosters the generation of outputs that are coherent, contextually appropriate, and tailored to the task requirements. The combined use of these strategies enhances the model’s ability to navigate intricate problem spaces.
\end{itemize}

REAP addresses the limitations of existing LLM problem-solving techniques, especially for tasks that require complex reasoning. By leveraging dynamic context generation, REAP enhances performance and opens new avenues for further investigation. The subsequent sections will detail REAP's development and application, demonstrating its potential to advance LLM capabilities and encouraging further research into dynamic context generation.

\subsection{Research Focus and Objectives}
This research seeks to improve LLMs' capacity for solving complex, reasoning-intensive tasks. While advancements in prompting strategies and agentic architectures have yielded some progress, LLMs continue to struggle with consistently generating coherent and accurate responses to intricate, multi-step problems.

The study hypothesizes that providing a well-structured, dynamically generated context can significantly enhance LLM performance in these scenarios. To test this hypothesis, we introduce the REAP method, which integrates Reflection, Explicit Problem Deconstruction, and Advanced Prompting to deliver dynamic context.

The key objectives of this research are:
\begin{itemize}
    \item \textbf{Validate the Effectiveness of the REAP Framework:} Demonstrate that REAP enhances LLM problem-solving capabilities, particularly in reasoning-intensive tasks, by improving coherence, relevance, and accuracy relative to zero-shot prompting.
    
    \item \textbf{Analyze the Interaction of REAP Components:} Examine how the components—Reflection, Explicit Problem Deconstruction, and Advanced Prompting—interact to enhance LLM performance. This objective seeks to determine how the integration of these strategies overcomes the limitations of zero-shot prompting.
    
    \item \textbf{Explore the Potential of Dynamic Context Generation:} Investigate the contribution of dynamic context generation to REAP’s overall performance improvements, with a focus on its impact compared to zero-shot prompting. This study highlights its potential as a valuable approach for future AI research.
\end{itemize}

\subsection{Contributions}
This research presents REAP as a structured methodology within the dynamic context generation framework, designed to enhance the quality, coherence, and relevance of LLM-generated outputs, particularly in reasoning-intensive tasks. The study makes several key contributions:

First, it integrates Reflection, Explicit Problem Deconstruction, and Advanced Prompting into a cohesive approach that directly addresses the limitations of existing problem-solving methods, especially in zero-shot prompting scenarios.

Second, the study offers empirical validation of REAP’s effectiveness in improving LLM performance. Notable gains in coherence, relevance, and accuracy are observed across various reasoning-intensive tasks, highlighting the advantages of REAP over traditional zero-shot prompting techniques.

Finally, by validating the REAP framework, this research provides a basis for future exploration into dynamic context generation, positioning REAP as a valuable direction for advancing LLM methodologies and promoting further innovation within artificial intelligence.

\section{Literature Review: Evolution of Problem-Solving Techniques in Large Language Models}

The development of LLMs has greatly advanced natural language processing, enabling these models to understand and generate human language with impressive accuracy. However, enhancing their problem-solving and reasoning abilities, particularly in unfamiliar tasks, remains a significant challenge. This literature review traces the evolution of problem-solving techniques in LLMs, leading to the development of the REAP methodology.

A critical development in this evolution is the concept of dynamic context generation. This strategy, explored by Betz et al. \cite{betz2021thinking}, involves the model generating an initial set of ideas or contextual information before attempting to solve a problem. By doing so, the model can approach tasks with a more informed and nuanced understanding, which is especially beneficial in zero-shot scenarios. Dynamic context generation enhances the model's ability to reason through complex tasks by providing it with a tailored context that better aligns with the specific requirements of the problem at hand.

\begin{figure}[h]
    \centering
    \includegraphics[width=\textwidth]{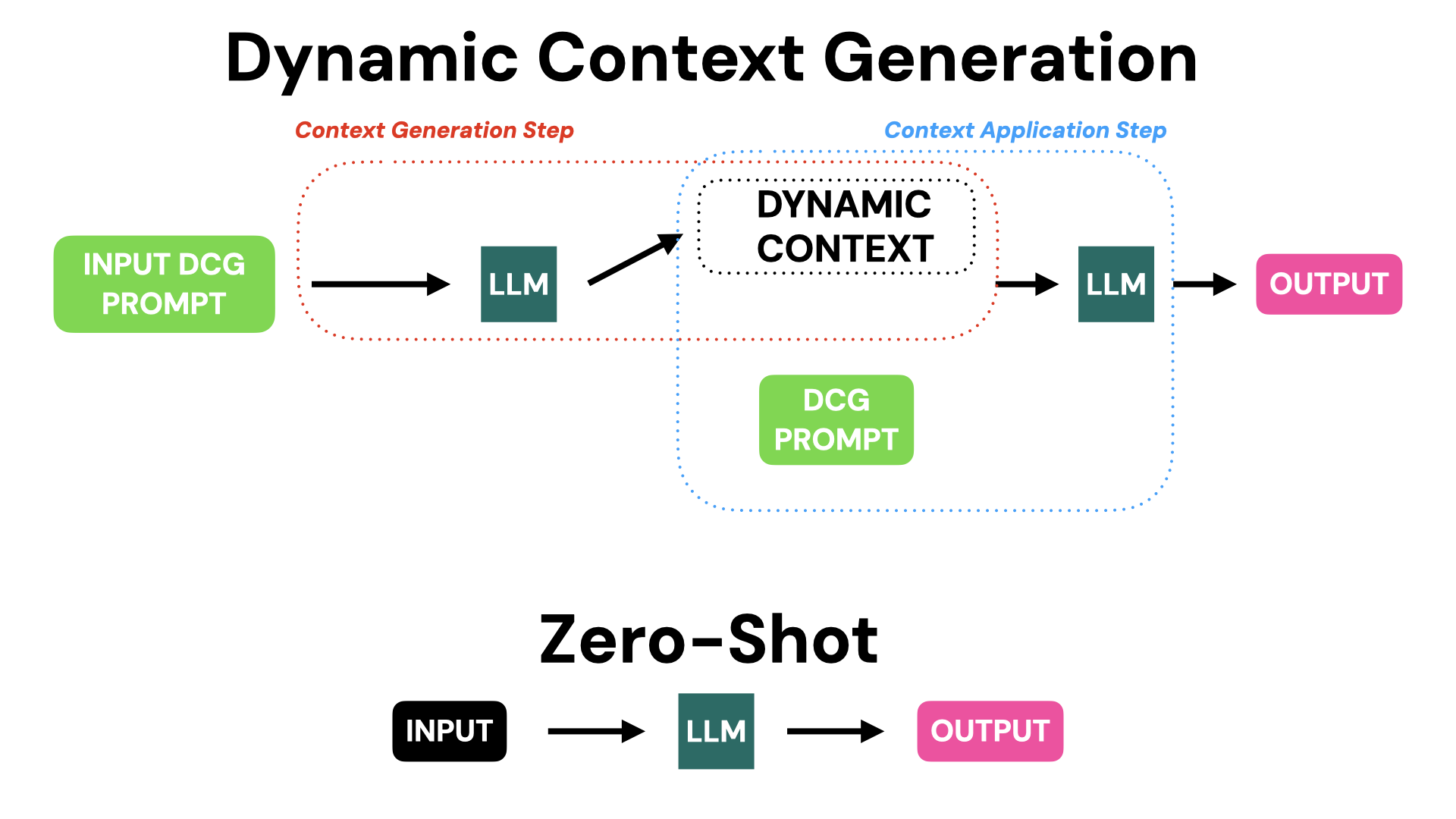}
    \caption{Comparison between Zero-Shot and Dynamic Context Generation (DCG) methodologies in LLMs. The DCG approach involves generating and applying context to enhance the final output, while Zero-Shot directly produces output from the input.}
    \label{fig:dcg_diagram}
\end{figure}

The evolution of these techniques began with the foundational work of Kojima et al. \cite{kojima2022large}, who demonstrated that LLMs are capable of zero-shot reasoning. This capability allows models to generalize from their extensive pre-training to address unfamiliar problems without needing task-specific training. While this marked a substantial advancement, the limited depth and accuracy of reasoning in zero-shot scenarios underscored the need for more sophisticated approaches.

Building on the concept of zero-shot reasoning, Wei et al. \cite{wei2022chain} introduced chain-of-thought prompting. This technique guides models to break down a problem into a sequence of intermediate reasoning steps, akin to human problem-solving processes. By decomposing complex problems into manageable steps, LLMs produced more coherent and logical responses. Despite the improvements in reasoning quality, ensuring the reliability and consistency of these reasoning paths remained challenging, necessitating further refinement.

To address these challenges, Wang et al. \cite{wang2023selfconsistency} incorporated self-consistency within the chain-of-thought framework. Self-consistency involves generating multiple reasoning paths for a given problem and selecting the most consistent answer, thereby enhancing the robustness and reliability of the model's outputs. However, the linear nature of reasoning paths in this approach limited the exploration of alternative solutions, prompting the development of more advanced methods.

Expanding beyond linear reasoning, Yao et al. \cite{yao2023tree} proposed the tree of thought method, which allows models to explore multiple branches of reasoning concurrently and create a tree structure of potential solutions. Each branch represents a different line of thought, enabling the model to consider various possibilities simultaneously. This method facilitated more comprehensive and nuanced problem-solving, but managing the complexity of interconnected branches presented its own set of challenges, leading to further innovations.

Besta et al. \cite{besta2024graph} introduced the graph of thought approach, leveraging graph structures to solve elaborate problems. Graphs, composed of nodes representing different states or pieces of information and edges representing relationships between them, allowed for non-linear and interconnected reasoning pathways. This method provided a more flexible and comprehensive problem-solving framework, though the complexity of graph structures necessitated more efficient ways of refining solutions.

Parallel to these structural advancements, Madaan et al. \cite{madaan2024selfrefine} explored self-improvement techniques, where models iteratively enhance their outputs through feedback loops, similar to how humans learn from experience. This approach of self-reflection and continuous improvement showed promise, but integrating it with advanced reasoning structures posed significant challenges.

In addition to self-improvement, Yao et al. \cite{yao2023react} combined reasoning with action in the React framework, integrating logical reasoning with task execution. This synergy enabled models to perform actions based on their reasoning processes, demonstrating practical applications of LLMs in real-world tasks. Nonetheless, maintaining coherence and accuracy during complex tasks remained an ongoing challenge.

Continuing this trajectory, Shinn et al. \cite{shinn2024reflexion} introduced Reflexion, integrating verbal reinforcement learning into LLMs. This method uses feedback mechanisms to reinforce correct responses and improve problem-solving accuracy over time. Although this approach showed promise, ensuring consistent improvements across diverse tasks remained challenging.

The advancements discussed above have directly informed the development of the REAP (Reflection, Explicit Problem Deconstruction, and Advanced Prompting) methodology. REAP addresses the limitations identified in prior work by integrating several key components. Reflection is central to REAP, incorporating continuous feedback mechanisms that enable the LLM to iteratively reassess and refine its approach. This method builds on the reflective self-improvement strategies introduced by Madaan et al., ensuring that the LLM remains adaptive throughout the problem-solving process.

REAP also emphasizes explicit problem deconstruction, systematically breaking down complex problems into smaller, manageable components. This approach facilitates a structured exploration of potential solutions by focusing on the distinct elements of each problem and their interconnections.

Finally, advanced prompting techniques are integral to REAP, guiding the LLM’s reasoning and decision-making processes. By fostering the generation and exploration of multiple solution pathways, REAP leverages insights from earlier steps to produce outputs that are coherent, logical, and contextually relevant.

This review highlights that the evolution of problem-solving techniques in LLMs has been a cumulative process, with each advancement addressing specific challenges and leading to progressively more sophisticated models. REAP represents the latest step in this progression, explicitly combining reflection, explicit problem deconstruction, and advanced prompting within a dynamic context generation framework. While REAP marks a significant advancement in LLM performance, it also lays the foundation for future innovations and continuous improvement in AI capabilities. For a visual representation of these strategies, please refer to the infographic in \hyperref[sec:appendixA1]{Appendix A}.

\section{Problem Statement}
\subsection{Current Limitations in Prompting Strategies}
Current prompting strategies for LLMs often lack effective integration between problem understanding, reasoning, and iterative refinement. This disjointed approach results in inconsistent performance, particularly in tasks requiring multi-step analytical reasoning.

While reflection mechanisms offer advantages, they are frequently constrained by initial model limitations, leading to potential error propagation if not managed rigorously. Techniques like chain-of-thought prompting struggle to maintain coherence in complex tasks, and advanced methods, such as tree-of-thought, can sacrifice both efficiency and logical consistency.

There is a pressing need for an integrated approach that enhances coherence, optimizes computational resources, and ensures reliable output quality across complex and cognitively demanding tasks. The REAP methodology addresses these challenges by incorporating reflection, explicit problem deconstruction, and advanced prompting into a unified framework.

\subsection{The Need for Integration: The REAP Approach}
To address these limitations, the REAP methodology integrates reflection, explicit problem deconstruction, and advanced prompting into a unified framework. This integration facilitates deeper problem understanding and improves the overall quality of LLM outputs.

In REAP, reflection is embedded throughout the problem-solving process, providing continuous feedback and allowing the LLM to dynamically adapt and refine its responses. This iterative reflective practice enhances both the accuracy and coherence of the model’s outputs.

REAP also emphasizes systematic problem deconstruction, ensuring that each aspect of a complex problem is thoroughly analyzed and addressed. This structured approach helps manage intricate interdependencies, resulting in more reliable outputs.

Advanced prompting techniques in REAP direct the LLM to explore multiple solution pathways, ensuring that its reasoning remains flexible, contextually appropriate, and logically consistent. By combining these techniques into a cohesive method, REAP enables LLMs to handle complex tasks with greater precision.

\subsection{Research Hypothesis}
The hypothesis of this research is that the REAP methodology, through its integration of reflection, systematic problem deconstruction, and advanced prompting techniques, enhances LLMs’ problem understanding and output quality. This approach is expected to outperform traditional zero-shot learning and isolated prompting strategies in producing responses that are more accurate, coherent, and contextually relevant. Furthermore, the structured nature of REAP is expected to improve the explainability and interpretability of the model’s outputs, enabling users to more easily understand and trust the results.

\subsection{Expected Outcomes}
The REAP method is projected to improve the accuracy and contextual relevance of LLM outputs, thereby enhancing their applicability to real-world tasks. Additionally, REAP is expected to advance explainable AI (XAI) by structuring the problem-solving process in a transparent and systematic way.

Through explicit problem deconstruction and advanced prompting techniques, REAP breaks down complex tasks into manageable components, increasing the transparency of the reasoning process. This transparency allows users to trace the model’s logic, facilitating the identification and correction of errors.

The integrated reflection component further strengthens explainability by supporting iterative refinement and offering insights into the decision-making process. This enhanced transparency fosters trust and improves collaboration between humans and AI, enabling users to rely on the model’s outputs while having the means to identify and address inaccuracies.

Ultimately, the REAP method is expected to generate outputs that are not only more accurate and coherent but also more interpretable to human users, contributing to the development of more reliable and effective AI systems.

\section{Proposed Approach: REAP}

\subsection{Overview of the REAP Method}

The REAP method—comprising Reflection, Explicit Problem Deconstruction, and Advanced Prompting—provides a structured solution to overcoming the limitations of LLMs in complex problem-solving scenarios. This approach integrates three core strategies into a cohesive process, enabling the LLM to produce outputs that are accurate, coherent, and contextually appropriate.

REAP operates through a unified prompt that seamlessly incorporates reflection, problem deconstruction, and advanced prompting. Rather than handling these components in isolation, REAP consolidates them into a continuous workflow that strengthens the LLM’s reasoning capabilities, especially in tasks that demand intricate, multi-layered analysis.

\subsection{Novel Contributions}

The REAP method makes several significant contributions to the field of AI and LLMs:

\begin{enumerate}
    \item \textbf{Integration of Multiple Strategies:} REAP offers a unified framework for dynamic context generation by combining reflection, explicit problem deconstruction, and advanced prompting. This integration improves the logical consistency and accuracy of LLM outputs, particularly in complex tasks, representing a notable evolution in how LLMs address intricate problem-solving challenges.
    
    \item \textbf{Systematization of Problem-Solving:} By formalizing the problem-solving process into a structured method, REAP ensures that all aspects of a problem are comprehensively addressed. This systematic approach allows LLMs to manage complex interdependencies with greater precision, enhancing both efficiency and effectiveness in multi-layered reasoning tasks.
    
    \item \textbf{Enhanced AI Explainability:} REAP improves explainability by making the reasoning process more transparent through explicit problem deconstruction and reflective mechanisms. This clarity enables users to trace the model's decisions, identify errors, and better understand the output, facilitating more effective human-AI collaboration.
\end{enumerate}

These advancements contribute to the field by providing a more integrated approach to complex problem-solving and enhancing the interpretability and transparency of AI systems.

\subsection{Components of the REAP Method}
\begin{figure}[h]
    \centering
    \includegraphics[width=\textwidth]{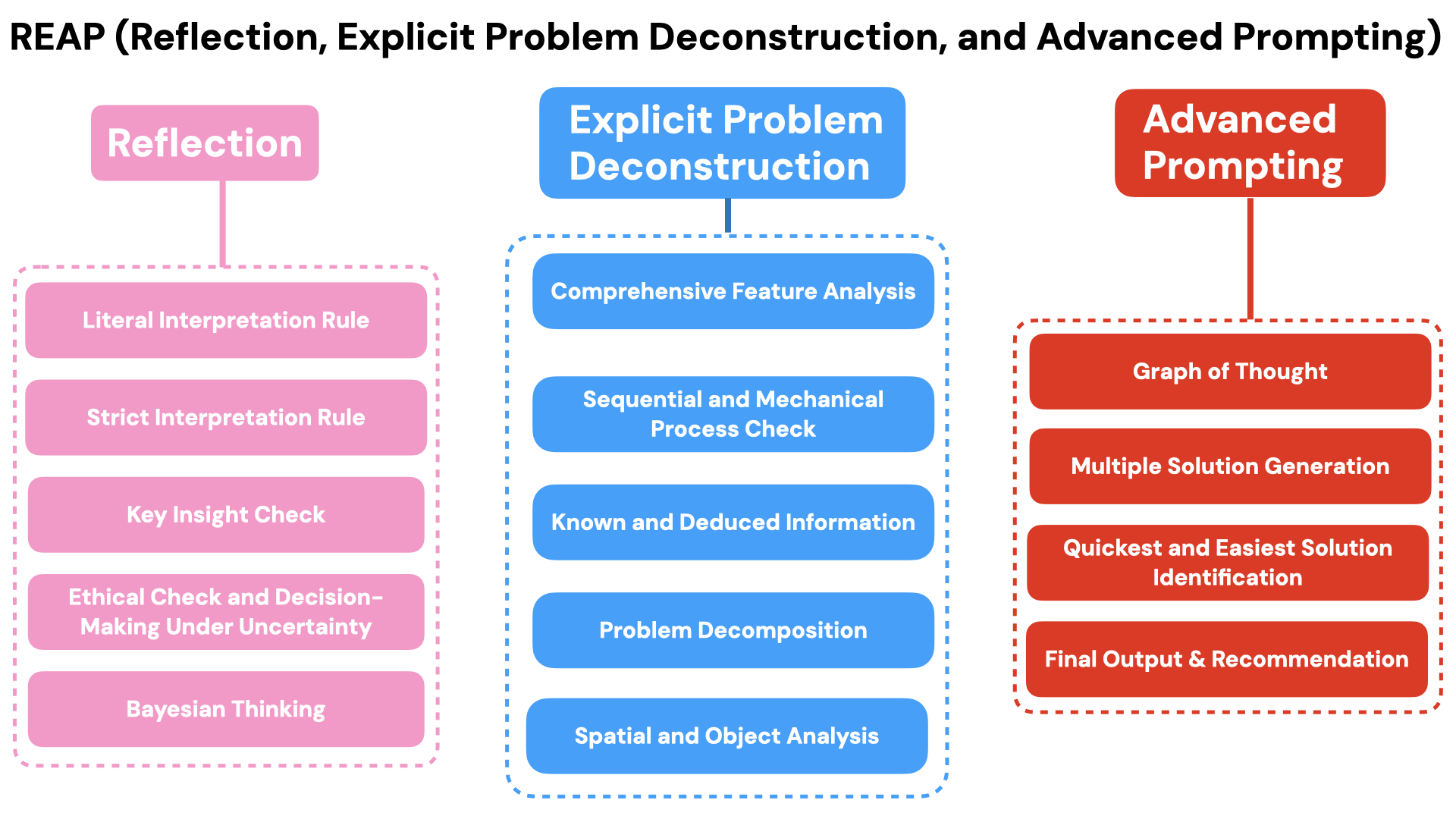}
    \caption{Overview of the REAP Method: Reflection, Explicit Problem Deconstruction, and Advanced Prompting, with their respective components.}
    \label{fig:reap_components}
\end{figure}

\subsection{Reflection}

Reflection within the REAP method is a core process integrated directly into the single prompt, creating a structured context that guides the LLM through the problem-solving task. This component ensures the LLM aligns strictly with the provided information, avoiding unsupported assumptions or speculative inferences.

The reflection process begins with the \textit{Literal Interpretation Rule}, which mandates that the LLM interpret each statement in the problem exactly as it is presented, without inferring unstated meanings. This step is critical to preventing early misinterpretations that could compromise the integrity of the problem-solving process. As the prompt advances, the \textit{Strict Interpretation Rule} reinforces adherence to explicit content, ensuring that any gaps in information are identified rather than filled with conjecture.

During the execution of the prompt, a \textit{Key Insight Check} is performed to review all identified features and processes, aiming to uncover critical details that could simplify the problem or point to a straightforward solution. The reflection process also incorporates an \textit{Ethical Check and Decision-Making Under Uncertainty}, where the model evaluates the ethical implications and risks of potential solutions, particularly in scenarios with uncertain outcomes.

\textit{Bayesian Thinking} is embedded within the reflection process, enabling the LLM to adjust its understanding based on newly provided, explicit information. This dynamic adjustment allows the model to maintain flexibility, refining its reasoning as new data is introduced. Throughout the REAP method, reflection establishes a context that ensures the LLM’s outputs remain accurate, ethically sound, and logically coherent.

\subsubsection{Explicit Problem Deconstruction}

Explicit Problem Deconstruction is a core element of the REAP method, systematically breaking down complex problems into smaller, manageable components. This process occurs within the single prompt and ensures that the model thoroughly comprehends the problem before generating solutions. The insights obtained during this stage are used to create a refined context that the model will later reference to respond to the query more effectively. Additionally, this approach enhances transparency, allowing human users to better understand the model’s perspective on the problem and the reasoning behind its decisions.

The process begins with a \textit{Comprehensive Feature Analysis}, where the LLM extracts and lists every relevant feature, actor, action, and relationship outlined in the problem statement. By adhering strictly to the exact wording of the problem, the model ensures that all critical details are accurately captured, providing a robust foundation for subsequent analysis.

As the prompt advances, a \textit{Sequential and Mechanical Process Check} is performed to analyze sequences of events or mechanical processes within the problem. This step is crucial for identifying interdependencies between actions and their impact on the system as a whole. Following this, the \textit{Known and Deduced Information} step involves listing all explicit facts and deriving logically sound deductions based on the provided information.

Next, the model engages in \textit{Problem Decomposition}, systematically breaking the problem into its components and subcomponents. This structured process ensures that all aspects of the problem are thoroughly addressed, with careful attention given to the interactions between elements. Finally, \textit{Spatial and Object Analysis} examines spatial relationships and object properties, particularly in scenarios where physical dynamics are key.

This detailed deconstruction process not only helps the LLM build a comprehensive understanding of the problem but also creates a context that enhances its ability to provide accurate responses. Additionally, it offers human users greater insight into the model’s interpretation of the problem and the reasoning that underlies its conclusions.

\subsubsection{Advanced Prompting}

Advanced Prompting plays a pivotal role in the REAP method, directing the LLM’s reasoning and decision-making within a single prompt. This phase is designed to provide additional insights and information that the model will later incorporate to refine its problem-solving approach, ensuring that the final solutions are logical, well-justified, and aligned with the task requirements.

The process begins with the \textit{Graph of Thought}, where the model constructs a representation of the problem’s structure by identifying key relationships and dependencies. This step enables the model to establish a comprehensive understanding of the problem's framework, which it will utilize when exploring potential solutions.

Next, the LLM engages in \textit{Multiple Solution Generation}, synthesizing the collected data to explore various solution pathways. By considering multiple approaches, the model develops a more robust context for informed decision-making.

The \textit{Quickest and Easiest Solution Identification} phase allows the model to streamline its decision-making by selecting the most efficient and direct solution from the options evaluated, prioritizing both effectiveness and simplicity. This selection further refines the context used in the final recommendation.

The process culminates with the \textit{Final Output \& Recommendation}, where the LLM integrates all the derived insights to present a well-reasoned conclusion. This recommendation is crafted to be comprehensive and closely aligned with the problem’s demands, based on the full context developed throughout the REAP prompt.

This phase not only enhances the model’s ability to generate effective solutions but also ensures that the reasoning process is clear and transparent for human users, enabling them to trace how the model arrived at its conclusions.

\subsection{Practical Implementation of the REAP Method}

\subsubsection{Method Overview}
The REAP method is a structured approach aimed at improving the problem-solving capabilities of LLMs by guiding them through a systematic process of dynamic context generation. This approach encourages the LLM to consider the query in a logical and stepwise manner, generating relevant information that supports each subsequent stage of the problem-solving process.

REAP integrates its core components—Reflection, Explicit Problem Deconstruction, and Advanced Prompting—into a cohesive prompt. This unified framework enables the LLM to establish a robust contextual foundation, perform a thorough analysis of the problem, and explore multiple solution paths. The method is designed to produce outputs that are coherent, contextually relevant, and aligned with the specific requirements of the task.

By implementing the REAP method, the LLM not only addresses immediate problem-solving challenges but also enhances its ability to generate and apply dynamic context. This leads to more precise and dependable outcomes, making the LLM’s responses more applicable to real-world scenarios.

\subsubsection{Illustrative Example}

To illustrate the practical implementation of the REAP method, consider the following problem from the dataset used in testing: "How do you measure exactly 4 gallons of water using only a 3-gallon, 5-gallon, and 4-gallon jug?" This problem, as cited in Williams and Huckle’s \textit{Easy Problems That LLMs Get Wrong} \cite{williams2024easy}, presented a considerable challenge for models, most of which could solve it correctly using zero-shot prompting. However, when the REAP method was employed as a unified, structured prompt, the models successfully arrived at the correct solution.

The REAP method guided the LLM as follows:

\begin{enumerate}
    \item \textbf{Literal Interpretation Rule:} The LLM began by interpreting the problem statement literally, identifying the task of measuring exactly 4 gallons of water using the jugs provided, without assuming any additional information or implications.

    \item \textbf{Strict Interpretation Rule:} The LLM adhered strictly to the information provided in the problem statement, ensuring no assumptions or inferences were made beyond what was explicitly stated. The problem was interpreted exactly as described, focusing on the literal meaning of the instructions.

    \item \textbf{Comprehensive Feature List:} The LLM identified the key objects in the problem:
    \begin{itemize}
        \item "3-gallon jug" - An object that can hold exactly 3 gallons of water.
        \item "5-gallon jug" - An object that can hold exactly 5 gallons of water.
        \item "4-gallon jug" - An object that can hold exactly 4 gallons of water.
        \item "Measure exactly 4 gallons of water" - The goal is to obtain exactly 4 gallons of water.
    \end{itemize}
    This list ensured that all relevant features were accounted for before proceeding.

    \item \textbf{Sequential and Mechanical Process Check:} The LLM considered the process of transferring water between the jugs, understanding that any sequence involving the transfer must result in exactly 4 gallons being measured. The key was to use the jugs' different capacities to transfer water until the exact amount was reached.

    \item \textbf{Key Insight Check:} The LLM recognized that the 4-gallon jug could directly hold the exact amount of water needed, simplifying the problem. This was identified as a key insight that made the solution more straightforward.

    \item \textbf{Known and Deduced Information:} The LLM reaffirmed that the 4-gallon jug, when full, would contain exactly 4 gallons, which directly meets the problem’s goal. The explicit facts about the jugs' capacities were used to deduce this.

    \item \textbf{Problem Decomposition:} The LLM broke down the problem into components, focusing on how each jug could be used to achieve the desired outcome:
    \begin{itemize}
        \item Measuring exactly 4 gallons of water using the 4-gallon jug.
        \item Considering potential sequences using the 3-gallon and 5-gallon jugs.
    \end{itemize}

    \item \textbf{Graph of Thought:} The LLM created a conceptual map showing how the jugs could be used to measure the water, identifying multiple pathways to reach the solution, including both direct and more complex methods.

    \item \textbf{Spatial and Object Analysis:} The LLM analyzed the spatial relationships and capacities of the jugs, considering how water could be transferred and stored. This analysis ensured that the solution would be feasible given the physical constraints of the jugs.

    \item \textbf{Bayesian Updates:} Since no new information or probabilities were introduced, no Bayesian updates were necessary for this problem.

    \item \textbf{Ethical Check and Decision-Making Under Uncertainty:} The problem did not involve ethical considerations or significant risks, so the LLM determined that no special ethical decisions were required in this scenario.

    \item \textbf{Multiple Solution Generation:} The LLM explored different methods to achieve the goal:
    \begin{itemize}
        \item \textbf{Solution 1:} Fill the 4-gallon jug directly to measure exactly 4 gallons.
        \item \textbf{Solution 2:} Use the 5-gallon jug to fill the 3-gallon jug, leaving 2 gallons in the 5-gallon jug. Transfer this to the 4-gallon jug and repeat to measure exactly 4 gallons.
        \item \textbf{Solution 3:} Use a series of transfers between the 3-gallon and 5-gallon jugs to measure out exactly 4 gallons.
    \end{itemize}
    
    \item \textbf{Quickest and Easiest Solution:} The LLM concluded that filling the 4-gallon jug directly was the simplest and most efficient solution, requiring the least number of steps and directly meeting the problem's goal.

    \item \textbf{Reflection:} The LLM reviewed its solutions, confirming that filling the 4-gallon jug directly was the most straightforward and efficient approach. The more complex solutions involving the 3-gallon and 5-gallon jugs, while valid, were deemed unnecessary given the simplicity of the goal.

    \item \textbf{Final Output \& Recommendation:} The LLM recommended filling the 4-gallon jug directly to measure exactly 4 gallons of water. This solution was presented as the most efficient and logical choice, fully supported by the explicit information provided in the problem statement.
\end{enumerate}

This example illustrates how the REAP method enables the LLM to systematically approach a problem by dynamically generating context, breaking it down into manageable parts, and using that context later in the prompt to inform its decision-making. By considering various solutions within this structured framework, the LLM is more likely to arrive at a logical and correct conclusion. Through the REAP method, the LLM navigated a problem that other methods struggled with, demonstrating the effectiveness of this integrated approach in enhancing problem-solving capabilities.

\subsection{Anticipated Challenges and Mitigation Strategies}

\subsubsection{Potential Limitations of the REAP Method}
While the REAP method provides a well-defined and systematic framework for enhancing LLM problem-solving, several potential limitations may arise during its implementation. One key limitation is the increased computational demand that the method may introduce. The REAP method requires the LLM to conduct detailed analysis across multiple stages—reflection, problem deconstruction, and advanced prompting—within a single prompt. This comprehensive process can be resource-intensive, particularly for large or complex tasks, potentially resulting in longer processing times.

Another concern is the risk of bottlenecks during the reasoning process. Given that the REAP method involves sequential stages that build upon each other, there is a possibility that the model may struggle to process the volume or complexity of the information. This could slow the problem-solving process and introduce difficulties in maintaining coherence across the stages, potentially leading to suboptimal outcomes.

Additionally, the REAP method’s emphasis on explicit data may limit its effectiveness in situations where the problem statement is incomplete or ambiguous. In cases where critical information is missing, or where more creative or inferential reasoning is required, the method’s strict reliance on explicit details might hinder the model’s ability to produce innovative or adaptable solutions.

\subsubsection{Mitigation Strategies}
To address the potential challenges associated with the REAP method, several strategies can be employed to improve its efficiency and effectiveness.

One approach to managing computational demands is to streamline the feature analysis and problem deconstruction stages. By concentrating on the most relevant aspects of the problem and reducing redundant or overly detailed analysis, the computational load can be minimized without sacrificing the method's effectiveness. Additionally, leveraging more efficient algorithms for sequential reasoning and Bayesian updates can further enhance performance, especially for large-scale or complex problems.

To prevent bottlenecks in the reasoning process, implementing checkpoints or intermediate evaluations within the REAP prompt can be advantageous. These checkpoints can be effectively integrated into an agentic architecture, allowing the model to assess its progress at various stages and ensure alignment before proceeding. This strategy helps regulate the flow of information, maintain coherence across stages, and optimize the LLM's performance in dynamic environments.

For cases where the problem statement is incomplete or ambiguous, incorporating supplementary prompts or pre-processing steps can enhance the model’s ability to manage such challenges. Additional prompts could direct the model to identify missing information or make limited inferences where necessary. Future iterations of the REAP method could also explore integrating more flexible reasoning capabilities while preserving the emphasis on explicit data.

By applying these strategies, the challenges associated with the REAP method can be effectively mitigated, ensuring it remains a robust tool for enhancing the problem-solving capabilities of LLMs.

\section{Methodology}

\subsection{Task Selection Rationale}

The tasks for this study were drawn from the Linguistic Benchmark introduced in \textit{Easy Problems That LLMs Get Wrong} by Williams and Huckle \cite{williams2024easy}. This benchmark is designed to expose key limitations of LLMs in areas such as logical reasoning, spatial intelligence, relational understanding, and linguistic comprehension. The dataset comprises questions that are straightforward for human adults but present considerable challenges for LLMs, making it an ideal environment to evaluate the effectiveness of the REAP method. By focusing on these tasks, the study aims to measure the extent to which REAP enhances LLM performance in areas where they traditionally struggle.

\subsection{Detailed Implementation Process}

The evaluation included six state-of-the-art LLMs: OpenAI's newly released o1-preview and o1-mini, GPT-4o, GPT-4o-mini, Google's Gemini 1.5 Pro, and Claude's 3.5 Sonnet. Each model was tested using two distinct methodologies:

\begin{enumerate}
    \item \textbf{Zero-Shot Prompting}: In this approach, each model was given questions from the dataset in a basic zero-shot setting. The models were prompted directly with the questions, without any supplemental context or guidance. This step established a baseline for each model's performance on these challenging tasks.
    \item \textbf{REAP-Enhanced Prompting}: In the second approach, the same questions were incorporated into the REAP method’s structured prompt. This involved applying the full REAP method—Reflection, Explicit Problem Deconstruction, and Advanced Prompting—as a single, unified prompt to guide the models through a more structured problem-solving process.
\end{enumerate}

Responses were collected for both the zero-shot and REAP-enhanced conditions, and human scorers evaluated each model's performance (see Appendix~\ref{sec:llm-responses}).

\subsection{Key Evaluation Metrics}

To measure the impact of the REAP method on LLM performance, we used the following metrics:

\begin{enumerate}
    \item \textbf{Correctness of Answer}: This metric measured whether the models provided accurate answers. It allowed for a comparison of success rates between zero-shot prompting and REAP-enhanced prompting.
    \item \textbf{Logical Reasoning}: This metric evaluated the coherence and consistency of the models' reasoning. It assessed whether the REAP method improved the models' ability to maintain a logical progression from the problem statement to the final solution.
    \item \textbf{Error Identification and Minimization}: This metric tracked the frequency and severity of errors in the models' responses, particularly in complex reasoning tasks, to determine whether the REAP method reduced mistakes.
    \item \textbf{Understanding and Relevance}: This metric assessed the relevance and focus of the models' responses in relation to the specific requirements of each task, determining whether the REAP method helped the models generate more contextually appropriate answers.
\end{enumerate}

\subsection{Ensuring Robustness and Reproducibility}

To ensure the robustness and reproducibility of our findings, we conducted the experiments under consistent testing conditions, applying the same dataset and prompt structures uniformly across all models. Multiple experimental runs were performed to verify the reliability of the results, thereby reducing the likelihood that observed improvements using the REAP method were attributable to random fluctuations.

It is important to note that the models used in this study, including OpenAI's GPT-4o and others, are probabilistic rather than deterministic. This means that even under identical conditions, minor variations in outcomes may occur if the experiments are repeated. However, our observations suggest that these variations are unlikely to substantially affect the conclusions drawn from the study. The consistency observed across multiple runs indicates that any differences are minimal and do not compromise the overall findings.

Finally, statistical analyses were performed to assess the significance of the differences between the zero-shot and REAP-enhanced conditions, ensuring that the results are both statistically robust and valid.

\subsection{Scoring Rubric}

The scoring of the models’ responses was performed by human evaluators using the rubric outlined below:

\begin{table}[h!]
\centering
\renewcommand{\arraystretch}{1.2} 
\begin{tabular}{|c|l|}
\hline
\textbf{Score} & \textbf{Criteria} \\ \hline
\textbf{100\%} & Correct answer with clear, logical reasoning and no errors. \\ \hline
\textbf{80\%}  & Correct answer with mostly logical reasoning, but minor errors or inconsistencies. \\ \hline
\textbf{60\%}  & Correct answer, but with flawed or unclear reasoning. \\ \hline
\textbf{40\%}  & Incorrect answer, but shows understanding or partially correct reasoning. \\ \hline
\textbf{20\%}  & Incorrect answer, but includes some relevant or useful information. \\ \hline
\textbf{0\%}   & Incorrect answer with no useful information or incorrect reasoning. \\ \hline
\end{tabular}
\vspace{0.15cm} 
\caption{Scoring Rubric}
\end{table}

Evaluators assessed each response based on the clarity, logical structure, and correctness of the reasoning. The final score for each response was determined according to this rubric, providing a comprehensive evaluation of the models' performance.

\section{Experiments and Results}

\subsection{Establishing Baseline Performance}

Baseline performance for the newly introduced models, OpenAI's o1-preview and o1-mini, was similarly assessed using zero-shot prompting. The initial evaluations showed that OpenAI o1-preview performed significantly better than the other models in zero-shot conditions, with a high degree of accuracy in several categories. However, OpenAI o1-mini's performance was more moderate, aligning closer with models like GPT-4o-mini and Gemini 1.5 Pro. Both models demonstrated low logical coherence in spatial and puzzle questions, similar to the other models.

\subsection{Performance of REAP}
When the same questions were presented using the REAP method, all models demonstrated substantial improvements in performance. The REAP-enhanced prompts led to higher accuracy rates and more coherent reasoning. The extent of improvement varied across different question types, with the most pronounced gains occurring in puzzle and spatial reasoning tasks, where the structured nature of REAP significantly enhanced the models' ability to process and solve these complex problems.

\begin{table}[h!]
\centering
\renewcommand{\arraystretch}{1.2} 
\begin{tabular}{|l|c|c|c|}
\hline
\textbf{Model} & \textbf{Average Zero-Shot Score (\%)} & \textbf{Average REAP-Enhanced Score (\%)} & \textbf{\% Gain} \\ \hline
\textbf{OpenAI o1-preview}     & 76.67\%  & 80.00\%  & 4.34\%   \\ \hline
\textbf{OpenAI o1-mini}        & 55.33\%  & 78.00\%  & 40.97\%  \\ \hline
\textbf{OpenAI GPT-4o}         & 47.32\%  & 78.68\%  & 66.26\%  \\ \hline
\textbf{OpenAI GPT-4o-mini}    & 30.68\%  & 65.32\%  & 112.93\% \\ \hline
\textbf{Google Gemini 1.5 Pro} & 32.00\%  & 56.00\%  & 75.00\%  \\ \hline
\textbf{Claude 3.5 Sonnet}     & 40.00\%  & 62.00\%  & 55.00\%  \\ \hline
\end{tabular}
\vspace{0.15cm} 
\caption{Overall Performance Comparison}
\end{table}

\begin{table}[h!]
\centering
\renewcommand{\arraystretch}{1.2} 
\begin{tabular}{|l|c|c|c|c|}
\hline
\textbf{Question Type} & \textbf{o1-preview ZS (\%)} & \textbf{o1-preview REAP (\%)} & \textbf{o1-mini ZS (\%)} & \textbf{o1-mini REAP (\%)} \\ \hline
\textbf{Counting}      & 100.00\% & 100.00\% & 100.00\% & 100.00\% \\ \hline
\textbf{Linguistic}    & 66.67\%  & 66.67\%  & 100.00\% & 100.00\% \\ \hline
\textbf{Popular Science} & 100.00\% & 100.00\% & 66.67\% & 100.00\% \\ \hline
\textbf{Puzzle}        & 66.67\%  & 83.33\%  & 38.33\%  & 66.67\% \\ \hline
\textbf{Relational}    & 66.67\%  & 66.67\%  & 66.67\%  & 66.67\% \\ \hline
\end{tabular}
\vspace{0.15cm} 
\caption{Performance Breakdown by Question Type for OpenAI o1-preview and o1-mini}
\end{table}

\begin{table}[h!]
\centering
\renewcommand{\arraystretch}{1.2} 
\begin{tabular}{|l|c|c|c|c|}
\hline
\textbf{Question Type} & \textbf{GPT-4o ZS (\%)} & \textbf{GPT-4o REAP (\%)} & \textbf{GPT-4o-mini ZS (\%)} & \textbf{GPT-4o-mini REAP (\%)} \\ \hline
\textbf{Counting}      & 100.00\% & 100.00\% & 100.00\% & 100.00\% \\ \hline
\textbf{Linguistic}    & 66.67\%  & 100.00\% & 33.33\%  & 66.67\%  \\ \hline
\textbf{Popular Science} & 100.00\% & 100.00\% & 66.67\% & 100.00\% \\ \hline
\textbf{Puzzle}         & 13.33\%  & 58.33\%  & 5.00\%   & 50.00\%  \\ \hline
\textbf{Relational}     & 73.33\%  & 73.33\%  & 33.33\%  & 66.67\%  \\ \hline
\textbf{Spatial}        & 48.57\%  & 91.43\%  & 37.14\%  & 65.71\%  \\ \hline
\end{tabular}
\vspace{0.15cm} 
\caption{Performance Breakdown by Question Type for OpenAI GPT-4o and OpenAI GPT-4o-mini}
\end{table}

\begin{table}[h!]
\centering
\renewcommand{\arraystretch}{1.2} 
\begin{tabular}{|l|c|c|c|c|}
\hline
\textbf{Question Type} & \textbf{Gemini ZS (\%)} & \textbf{Gemini REAP (\%)} & \textbf{Claude ZS (\%)} & \textbf{Claude REAP (\%)} \\ \hline
\textbf{Counting}      & 100.00\% & 100.00\% & 100.00\% & 70.00\% \\ \hline
\textbf{Linguistic}    & 33.33\%  & 33.33\%  & 33.33\%  & 33.33\%  \\ \hline
\textbf{Popular Science} & 66.67\% & 100.00\% & 66.67\% & 66.67\% \\ \hline
\textbf{Puzzle}         & 13.33\%  & 50.00\%  & 13.33\%  & 70.00\%  \\ \hline
\textbf{Relational}     & 66.67\%  & 66.67\%  & 33.33\%  & 73.33\%  \\ \hline
\textbf{Spatial}        & 14.29\%  & 40.00\%  & 62.86\%  & 51.43\%  \\ \hline
\end{tabular}
\vspace{0.15cm} 
\caption{Performance Breakdown by Question Type for Google Gemini 1.5 Pro and Claude 3.5 Sonnet}
\end{table}

\newpage

\subsection{In-Depth Analysis of Findings}

The detailed analysis of the results across various question types provides several key insights into model performance under both zero-shot and REAP-enhanced conditions:

\begin{itemize}
    \item \textbf{Puzzle Questions} showed the most significant gains with the REAP method. \textbf{OpenAI GPT-4o-mini} achieved an improvement of \textbf{900.00\%}, while \textbf{Claude 3.5 Sonnet} demonstrated a \textbf{425.00\%} gain. OpenAI's new models also benefited from REAP’s structured approach. \textbf{OpenAI o1-preview} showed a moderate improvement of \textbf{25.00\%}, while \textbf{OpenAI o1-mini} achieved a more substantial gain of \textbf{73.96\%}. These results indicate that REAP’s structured approach is particularly effective for tasks requiring complex, multi-step reasoning. The method enables models to systematically deconstruct problems and explore multiple solution pathways, which is essential for puzzles involving logical sequences or requiring the avoidance of common reasoning errors.

    \item \textbf{Spatial and Linguistic Questions} also showed considerable improvements with the REAP method. \textbf{OpenAI GPT-4o} and \textbf{GPT-4o-mini} improved by \textbf{88.24\%} and \textbf{76.92\%}, respectively, on spatial tasks. OpenAI's new models followed a similar trend. On spatial tasks, \textbf{OpenAI o1-preview} improved by \textbf{50.00\%}, while \textbf{OpenAI o1-mini} demonstrated a significant gain of \textbf{76.92\%}. For linguistic questions, \textbf{OpenAI o1-preview} and \textbf{o1-mini} both showed notable gains of \textbf{50.00\%}. This suggests that REAP's problem deconstruction and advanced prompting components help models better process spatial relationships and language-based tasks. The method enables models to approach these problems more holistically, considering all relevant aspects rather than focusing on isolated details.

    \item \textbf{Counting and Relational Questions} exhibited little to no gain, suggesting that these types of problems were already handled well by the models in zero-shot conditions. The REAP method did not add substantial value in these cases, likely because these tasks require less complex reasoning. For instance, \textbf{OpenAI GPT-4o} and \textbf{GPT-4o-mini} showed minimal improvements, and OpenAI's new models followed this trend. \textbf{OpenAI o1-preview} and \textbf{OpenAI o1-mini} demonstrated no significant improvement in counting tasks, both maintaining \textbf{100\%} accuracy before and after REAP. In relational reasoning tasks, both models showed no improvements. For straightforward tasks involving basic counting or relational reasoning, the additional structure provided by REAP may not be necessary.

   \item \textbf{Negative Gains and Literal Interpretation}: Negative gains were observed in some cases, particularly for \textbf{Claude 3.5 Sonnet} on \textbf{Spatial} and \textbf{Counting} tasks. In these instances, the model appeared to follow REAP's instructions too literally, missing the broader context or nuances required for effective problem-solving. OpenAI's new models also showed some rigidity. In counting tasks, both \textbf{o1-preview} and \textbf{o1-mini} demonstrated no improvement, maintaining \textbf{100\%} accuracy before and after REAP. Similarly, in relational and spatial tasks, \textbf{OpenAI o1-preview} exhibited limited flexibility, resulting in no significant gains. In spatial tasks, flexibility and the ability to infer relationships are critical; however, strict adherence to REAP's structure may have impeded performance. These findings suggest that while REAP enhances reasoning through structure, it may also limit flexibility in cases where nuanced interpretation is required.

    \item \textbf{Model-Specific Performance}: The variation in performance across models suggests that the effectiveness of the REAP method is model-dependent. For instance, \textbf{OpenAI GPT-4o-mini} exhibited substantial improvements across most question types, while \textbf{Claude 3.5 Sonnet} struggled in tasks requiring flexible interpretation. OpenAI's new models also reflected this model-specific variability. \textbf{OpenAI o1-mini} demonstrated significant gains in puzzle and spatial tasks, benefiting greatly from the structured guidance of REAP, while \textbf{OpenAI o1-preview} showed more moderate improvements, particularly in tasks where it already had a strong zero-shot performance. This observation highlights the importance of tailoring the REAP method to the specific characteristics of each model, potentially adjusting the prompt structure or level of guidance based on the model’s strengths and limitations.

\end{itemize}

Overall, these findings suggest that while the REAP method significantly improves the problem-solving abilities of LLMs, particularly in tasks requiring complex reasoning and multi-step processes, its effectiveness varies depending on the task type and the model’s inherent tendencies. OpenAI's new models, for instance, showed considerable improvements in tasks like puzzles and spatial reasoning but had more limited gains in simpler tasks like counting. In some cases, the structured approach of REAP may need to be adjusted to prevent overly literal interpretations, especially for tasks that require flexibility or less structured reasoning. The variability in performance across models highlights the importance of tailoring the method to suit each model's strengths and limitations.

\section{Discussion}

\subsection{Enhanced Problem Understanding}

The results of our experiments show that the REAP method, within the dynamic context generation framework, substantially improves LLM performance across various tasks, particularly in complex, multi-step reasoning scenarios. The structured integration of reflection, explicit problem deconstruction, and advanced prompting enables models to navigate intricate problem spaces more effectively than traditional zero-shot prompting. This improvement is especially pronounced in tasks involving logical sequencing, spatial reasoning, and language comprehension.

By guiding the models through a systematic analysis of the problem, REAP ensures that all relevant aspects are considered before arriving at a solution. This structured approach helps models avoid common pitfalls, such as overlooking critical details or making unsupported assumptions. OpenAI's new models, \textbf{o1-preview} and \textbf{o1-mini}, demonstrated notable improvements, particularly in puzzle and spatial tasks, where REAP's structured prompting allowed the models to systematically break down problems and explore multiple solution paths. The significant gains observed in these tasks underscore the value of REAP in managing non-linear reasoning paths, while its flexibility ensures that models can adapt to diverse reasoning challenges.

\subsection{Cost-Effectiveness of Models Using the REAP Method}

One of the primary advantages of the REAP method is the ability to achieve competitive performance while utilizing more cost-effective models. This is particularly relevant when considering the pricing structure of various models, which can vary significantly depending on input and output token usage.

Table \ref{tab:model_costs} presents a comparison of pricing for several models, measured by the cost per 1 million input and output tokens. It is important to note that these prices are accurate as of the writing of this paper and may fluctuate over time as models and pricing structures evolve. For instance, the \textbf{OpenAI o1-preview} model, which delivers the highest performance in zero-shot tasks, incurs a cost of \textbf{\$15} per 1 million input tokens and \textbf{\$60} per 1 million output tokens. In contrast, the \textbf{OpenAI GPT-4o-mini}, which is the cheapest model in this comparison, costs only \textbf{\$0.15} and \textbf{\$0.60} per 1 million input and output tokens, respectively. This makes the OpenAI o1-preview model approximately \textbf{100 times more expensive} than GPT-4o-mini in terms of both input and output token costs.

However, when enhanced with the REAP method, the cheaper models show substantial performance gains. For example, \textbf{OpenAI GPT-4o-mini} exhibits a \textbf{112.93\%} improvement in performance, increasing its average score from \textbf{30.68\%} to \textbf{65.32\%}. Similarly, the \textbf{OpenAI o1-mini} model improves from \textbf{55.33\%} to \textbf{78.00\%}, representing a \textbf{40.97\%} gain. This enhanced performance, coupled with its lower token cost (just \textbf{\$3} per million input tokens and \textbf{\$12} for output tokens), makes the o1-mini an attractive option for cost-sensitive applications.

This performance improvement is not exclusive to OpenAI models; \textbf{Google Gemini 1.5 Pro} and \textbf{Claude 3.5 Sonnet} also demonstrate significant gains when using the REAP method. Despite their relatively lower zero-shot scores, their REAP-enhanced scores improved by \textbf{75.00\%} and \textbf{55.00\%}, respectively, closing the gap with higher-cost models.

In summary, the REAP method provides a means to balance cost and performance effectively. By leveraging REAP, even lower-cost models can achieve near-competitive performance compared to significantly more expensive alternatives, offering flexibility for both budget-conscious and performance-oriented projects.

\begin{table}[h!]
\centering
\renewcommand{\arraystretch}{1.2} 
\begin{tabular}{|l|c|c|}
\hline
\textbf{Model}                & \textbf{1M Input Tokens (\$)} & \textbf{1M Output Tokens (\$)} \\ \hline
\textbf{OpenAI o1-preview}     & 15.00                        & 60.00                          \\ \hline
\textbf{OpenAI o1-mini}        & 3.00                         & 12.00                          \\ \hline
\textbf{OpenAI GPT-4o}         & 5.00                         & 15.00                          \\ \hline
\textbf{OpenAI GPT-4o mini}    & 0.15                         & 0.60                           \\ \hline
\textbf{Google Gemini Pro}     & 3.50                         & 10.50                          \\ \hline
\textbf{Claude 3.5 Sonnet}     & 3.00                         & 15.00                          \\ \hline
\end{tabular}
\caption{Cost Comparison of Models (per 1M tokens). Prices are current as of the writing of this paper.}
\label{tab:model_costs}
\end{table}

For reference, we reprint the performance comparison with the REAP method in Table \ref{tab:performance_comparison}. This table illustrates the considerable performance gains made by each model when using the REAP method. As seen, even lower-cost models such as \textbf{OpenAI GPT-4o-mini} and \textbf{OpenAI o1-mini} experience significant improvements, making them more competitive with higher-priced models.

\begin{table}[h!]
\centering
\renewcommand{\arraystretch}{1.2} 
\begin{tabular}{|l|c|c|c|}
\hline
\textbf{Model}                & \textbf{Average Zero-Shot Score (\%)} & \textbf{Average REAP-Enhanced Score (\%)} & \textbf{\% Gain} \\ \hline
\textbf{OpenAI o1-preview}     & 76.67\%  & 80.00\%  & 4.34\%   \\ \hline
\textbf{OpenAI o1-mini}        & 55.33\%  & 78.00\%  & 40.97\%  \\ \hline
\textbf{OpenAI GPT-4o}         & 47.32\%  & 78.68\%  & 66.26\%  \\ \hline
\textbf{OpenAI GPT-4o-mini}    & 30.68\%  & 65.32\%  & 112.93\% \\ \hline
\textbf{Google Gemini 1.5 Pro} & 32.00\%  & 56.00\%  & 75.00\%  \\ \hline
\textbf{Claude 3.5 Sonnet}     & 40.00\%  & 62.00\%  & 55.00\%  \\ \hline
\end{tabular}
\vspace{0.15cm} 
\caption{Performance Comparison with REAP Method (reprinted for reference).}
\label{tab:performance_comparison}
\end{table}

\subsection{Role of Reflection in Generating Dynamic Context}

Reflection plays a critical role in generating dynamic context, which models utilize throughout the problem-solving process. By incorporating key insight checks and evaluating critical aspects of the problem early, REAP helps models establish a well-informed context that supports decision-making in subsequent stages. This approach enhances the accuracy and coherence of outputs by providing a solid foundation for further reasoning.

Moreover, reflection enables models to assess the ethical implications of their solutions, particularly in scenarios involving uncertainty or potential risks. This feature ensures that decisions are not only technically sound but also aligned with ethical standards. The observed improvements in logical reasoning metrics emphasize the importance of reflection in enhancing the overall quality of the models' responses.

\subsection{Identifying and Addressing Limitations}

While the REAP method has demonstrated significant potential, it is essential to acknowledge its limitations. One key limitation observed in this study is the method's reliance on literal interpretation, particularly with models like \textbf{Claude 3.5 Sonnet} and, to a lesser extent, OpenAI's new models, \textbf{o1-preview} and \textbf{o1-mini}. In several cases, rigid adherence to the REAP prompts resulted in suboptimal performance, especially in tasks requiring flexibility or intuitive reasoning, such as spatial and counting tasks.

These findings suggest that while REAP offers valuable structure, it may need adjustment to accommodate more flexible approaches, particularly for tasks that benefit from creative problem-solving or when ambiguity is present in the problem statement. OpenAI's new models also exhibited similar tendencies, where strict adherence to the structured approach limited their performance in simpler tasks like counting, which require more intuitive solutions. The challenge lies in balancing structured guidance with adaptability, ensuring that models can effectively navigate tasks requiring nuanced interpretation while maintaining the benefits of REAP's problem deconstruction.

\subsection{Future Directions for REAP}

The findings of this study suggest several pathways for further development of the REAP method within the dynamic context generation framework. One area of future exploration is deeper integration of dynamic context generation into REAP. While REAP currently operates as a general, structured prompt, its capabilities could be expanded to dynamically generate context tailored to specific problem details. This enhancement would allow REAP to adapt in real-time, providing customized guidance that aligns with the unique demands of each task.

Another promising direction is embedding REAP within an agentic architecture. In such a setup, REAP could autonomously manage checkpoints and intermediate evaluations throughout the problem-solving process. This capability would enable REAP to dynamically assess progress and make real-time adjustments, enhancing the LLM's ability to handle complex, evolving tasks with precision.

Additionally, while REAP serves as a robust foundation, there is value in customizing it for different use cases. Practitioners could fine-tune the method to better address the specific challenges of each task, whether by refining the reflection process, adjusting the depth of problem deconstruction, or modifying prompting strategies to better suit the requirements of particular scenarios.

Future research could also explore combining REAP with advanced techniques such as meta-learning or reinforcement learning to enhance its adaptability and impact. By integrating these approaches, REAP could become more flexible, optimizing its strategies based on insights from previous tasks and continuous learning processes.

Lastly, given the model-specific performance differences observed, future work could focus on refining REAP’s application across various LLM architectures. Tailoring the method to the strengths and limitations of individual models will ensure its effectiveness across different systems.

\subsection{The Role of REAP in Explainable AI}
A major strength of the REAP method within the dynamic context generation framework is its capacity to enhance Explainable AI (XAI). As AI systems become more central to decision-making processes, the demand for clarity and understanding of AI-generated decisions has grown increasingly important. REAP supports XAI by offering a structured approach to problem-solving, making the model's reasoning more transparent and easier to interpret.

The explicit problem deconstruction and reflection components of REAP ensure that each stage of the model’s reasoning can be documented and traced back to specific elements of the problem statement. This traceability is crucial for understanding how the model arrived at a particular conclusion and for verifying that its reasoning aligns with established goals and ethical standards.

By encouraging the model to reflect on its decisions and consider alternative approaches, REAP further clarifies the rationale behind various choices, making it easier to explain why specific decisions were made. This capability is particularly valuable in fields where AI-driven decisions carry significant consequences, such as healthcare, finance, or legal domains.

As AI continues to advance, integrating methodologies like REAP, which enhance both the transparency and reliability of models, will be essential for fostering trust in AI systems. Providing clear and understandable explanations for AI-driven decisions not only promotes transparency but also strengthens human-AI collaboration.

\subsection{Broader Implications for AI Development}
The REAP method’s impact on LLM performance carries significant implications for the broader field of artificial intelligence. As AI models play a growing role in complex decision-making processes across various domains, the need for reliable reasoning frameworks like REAP is expected to increase. Guiding models through intricate problem-solving tasks with precision, ethical consideration, and transparency is critical to fostering trust in AI systems, particularly in high-stakes scenarios.

Moreover, the findings of this study underscore the importance of combining stable methodologies with adaptability in AI development. As AI systems confront a wide range of challenges, methods that can be tailored to the specific requirements of each task, while maintaining rigorous reasoning standards, will be crucial. With further refinement and customization within the dynamic context generation framework, the REAP method is well-positioned to contribute meaningfully to these advancements.

\section{Conclusion}

\subsection{Recap of REAP’s Contributions}
This study has demonstrated the key contributions of the REAP method within the dynamic context generation framework, aimed at advancing the capabilities of LLMs. REAP introduces a unified strategy that integrates reflection, explicit problem deconstruction, and advanced prompting, effectively addressing the challenges LLMs face in reasoning tasks involving complex, multi-step processes.

The primary contributions of REAP include:
\begin{itemize}
    \item \textbf{Structured Methodology}: REAP provides a cohesive framework that combines reflection, problem deconstruction, and advanced prompting, offering LLMs an organized approach to tackling and solving complex problems.
    \item \textbf{Enhanced Problem-Solving Capabilities}: By systematically guiding LLMs through logical sequencing, spatial reasoning, and linguistic comprehension, REAP improves their ability to generate precise, consistent, and contextually appropriate outputs.
    \item \textbf{Contribution to Explainable AI (XAI)}: REAP’s emphasis on traceability and reflective processes enhances transparency and reliability in LLM reasoning, contributing meaningfully to the broader field of Explainable AI.
\end{itemize}

In summary, REAP lays a solid foundation for future advancements in LLM methodologies, particularly in areas that require advanced reasoning and greater explainability.

\subsection{Broader Impact on AI Development}
The broader impact of this study extends into the field of artificial intelligence, emphasizing the growing need for consistent, reliable, and interpretable reasoning methods. As AI becomes increasingly central to decision-making processes across various domains, frameworks like REAP will be essential for fostering trust, particularly in contexts where decisions have significant consequences. REAP’s success in enhancing model performance underscores the importance of combining structured methodologies with adaptability in AI development.

Additionally, REAP’s contribution to Explainable AI (XAI) is notable. By clarifying LLMs' reasoning processes, REAP strengthens confidence in AI systems, especially where transparency is crucial. The ability to trace the model’s decision-making process and understand the rationale behind its choices improves human-AI interaction and ensures alignment with ethical standards.

\subsection{Concluding Thoughts}

In conclusion, the REAP method represents a valuable advancement in enhancing the problem-solving capabilities of LLMs within the dynamic context generation framework. Its structured approach, adaptability, and contribution to Explainable AI position REAP as a significant step forward in the ongoing development of AI methodologies. As AI systems become more complex and central to various domains, approaches like REAP will be essential to ensuring that these systems remain both effective and trustworthy.

The refinement and expansion of REAP are in their early stages, and the insights gained from this study provide a strong foundation for future research and innovation. By further exploring the potential of dynamic context generation and tailoring REAP to specific tasks, we can expand AI’s capabilities, making it an even more powerful tool for addressing the complex challenges ahead.

\newpage

\appendix
\section*{Appendix}  

\section{Methods Infographic}
\label{sec:appendixA1}

\begin{figure}[h]
    \centering
    \includegraphics[width=\textwidth]{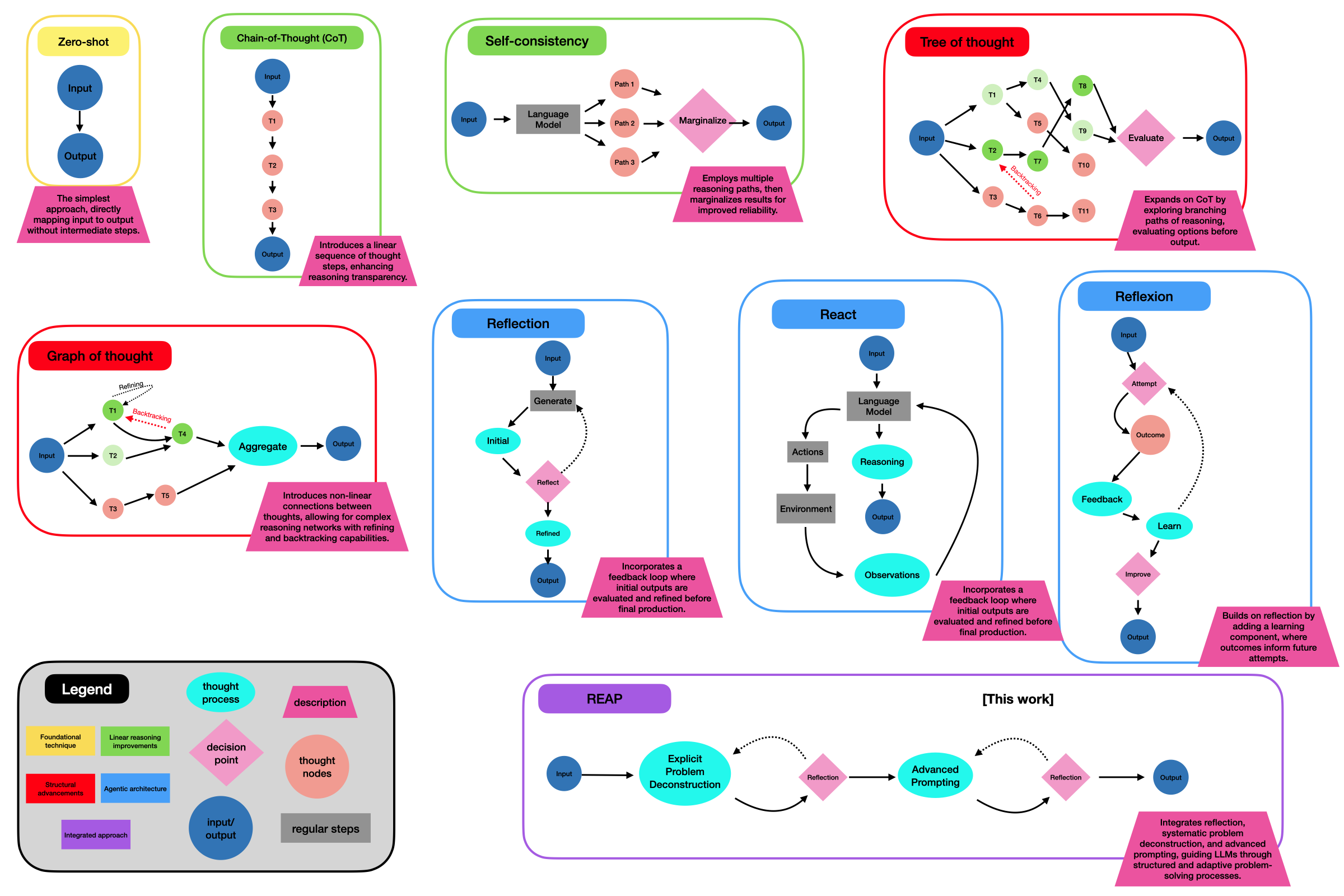}
    \caption{Illustration of the evolutionary path of problem-solving techniques in LLMs, culminating in the development of the REAP methodology.}
    \label{fig:methods_infographic}
\end{figure}

\section{REAP: Problem Solving Method}
This appendix contains the full REAP (Reflection, Explicit Problem Deconstruction, and Advanced Prompting) prompt used in this study. The REAP prompt is designed to guide LLMs through a structured problem-solving approach, where the specific problem is inserted at the end. This prompt systematically directs the LLM through reflection, problem deconstruction, and advanced prompting to encourage thorough analysis and logical reasoning.

\begin{lstlisting}
# REAP: Problem Solving Method 
You are an AI assistant designed to tackle complex problems using a structured and creative approach. Let's work together to analyze the problem and discover innovative solutions.

**0. Literal Interpretation Rule:**

- Interpret every statement in the problem LITERALLY.
- Do not assume any implications or consequences beyond what is explicitly stated.
- Pay close attention to the exact wording, especially regarding directions, positions, relationships, and sequences between objects or entities.
- For terms that refer to commonly understood concepts, identify the most straightforward interpretation.

**1. Strict Interpretation Rule:**

Before beginning any analysis, adhere to these crucial rules:
- Stick ONLY to what is explicitly stated in the problem.
- Do not make any assumptions or inferences beyond the exact wording.
- If the problem doesn't provide enough information to draw a conclusion, state this explicitly.
- Treat each piece of information as literal and precise. Do not extrapolate or interpret beyond what is directly stated.
- If a causal relationship or outcome is not explicitly stated, do not assume one exists.

**2. Comprehensive Feature Analysis:**

* Carefully read the entire problem statement multiple times.
* List out every single feature, detail, or piece of information provided in the problem statement. Include:
  - All objects mentioned, quoting the exact words from the problem statement.
  - All actors or entities involved, with direct quotes.
  - All actions described, using the exact phrasing from the problem.
  - **All directional, spatial, relational, and **sequential** phrases** that indicate the position, movement, order, or relationship between objects or entities, quoted directly.
  - All stated conditions or constraints, quoted directly.
  - Any numerical values or quantities mentioned, quoted directly.
  - Any temporal information (time, sequence, duration), using the exact language from the problem.
  - Any spatial information (location, direction, distance), with direct quotes.
  - Any causal relationships mentioned, quoted verbatim.
  - Any stated goals or objectives, using the exact phrasing.
* For each feature, quote the exact wording from the problem statement.
* Do not infer or assume any information not explicitly stated.
* If a feature seems to imply something, note it as "Potential implication" but do not treat it as fact.
* For each feature, briefly note its potential significance to the problem, but only based on what is explicitly stated.

**3. Sequential and Mechanical Process Check:**

* Explicitly identify and analyze any sequential, cyclical, or mechanical processes described in the problem.
  - Identify any sequence of events, steps, or actions.
  - Consider how these sequences influence the outcome or future steps.
  - Specifically assess if the sequence implies a certain outcome.
* **Key Questions to Ask:**
  - Does the sequence of events or actions lock in a particular outcome?
  - If the problem involves a mechanical process, how does each step or movement impact the overall system?
  - Are there any steps or sequences that must logically follow one another?
* Include these considerations in your analysis, ensuring that no critical sequence or mechanical step is overlooked.

**4. Key Insight Check:**

* Review the comprehensive list of features and the Sequential and Mechanical Process Check you've just created.
* Look for a "key insight" crucial detail or combination of details that might immediately reveal the correct answer or significantly simplify the problem.
* This key insight could be:
  - A direct statement of the answer
  - A crucial detail that makes the solution obvious
  - A constraint that eliminates most potential solutions
  - An unusual element that changes the nature of the problem
  - A combination of features that, when considered together, lead to a clear conclusion
* If you identify a potential key insight, explain it clearly and how it leads to an immediate answer or simplification.
* If an immediate answer is found, state it confidently. If not, proceed with the detailed analysis.
* Ensure any key insight is based solely on explicit information. Do not include potential implications or assumed outcomes.

**5. Known and Deduced Information:**

* Explicit Facts: 
  - List ONLY the exact quotes from the comprehensive feature list you created earlier.
  - For each explicitly stated relationship or position in the problem, list it clearly, then state its logical reverse (if applicable) and any complementary relationships (if applicable).
  - Do not paraphrase or summarize. Use quotation marks for each piece of information.
  - Format: "[Exact quote]" - Brief explanation of its relevance (if needed)

* Feature-Based Deduction:
  - Only make deductions that are 100% certain based on the explicit wording of the problem.
  - A valid deduction must be a logical certainty based on the given information, not just a likely outcome.
  - Format: Deduction: [Your certain inference] - Logical Basis: [Explain why this must be true] - Based on: "[Exact quote 1]", "[Exact quote 2]", etc.
  - If no certain deductions can be made, state this explicitly.

* For each piece of information, ask yourself: "What does this literally mean, without any assumptions?"
* Pay special attention to directional, relational, and **sequential** phrases. What do they literally describe?
* If the problem asks a question, ensure your deductions directly address this question using only the given information.

* IMPORTANT: Double-check that every piece of information in this section is either a direct quote or explicitly linked to direct quotes from the problem statement. If you cannot quote or directly link a piece of information to the original text, do not include it.

**6. Problem Decomposition:**

* Using ONLY the exact quotes from the Comprehensive Feature List and the Known and Deduced Information:
  - Break down the problem into smaller, manageable parts.
  - Identify key components and subcomponents of the problem.
  - Describe any interactions between these components, using only the information explicitly stated or directly deduced from the problem statement.

* For each component or interaction identified:
  - Provide the exact quote(s) from the problem statement that support its inclusion.
  - If it's based on a deduction, clearly state which exact quotes led to this deduction.

* For each component or interaction, consider: "Is this explicitly stated, or am I making an assumption?" If it's an assumption, do not include it.
* For each component, consider: "How does this directly relate to the question being asked?"
* If a component doesn't directly relate to the question, note this explicitly.

* Avoid making any assumptions beyond what is directly supported by the exact wording of the problem and the explicitly deduced information.

* Format your decomposition as follows:
  Component/Subcomponent: [Description]
  Supporting Quote(s): "[Exact quote(s) from problem statement]"
  Deduction (if applicable): [Clear explanation of how this was deduced]
  Based on: "[Relevant quote(s) for deduction]"

* If the problem lacks sufficient information to make a complete decomposition, state this explicitly.

* IMPORTANT: Double-check that every aspect of your problem decomposition is directly linked to exact quotes or explicit deductions from the Known and Deduced Information section. If you cannot provide this direct link, do not include that aspect in your decomposition.

**7. Graph of Thought:**

* **Objective:** Create a visual or textual representation of the problem's structure to explore possible solutions.
* **Action:**
    * **Identify Key Concepts:** List the main concepts or subproblems involved, based only on explicit information from the problem statement.
    * **Establish Relationships:** Determine how these concepts are connected, using only explicitly stated information.
    * **Construct the Graph:** 
        * **Visual:** If possible, draw a graph with nodes (concepts) and edges (relationships).
        * **Textual:** If drawing is not possible, create a list of nodes and their connections.
    * **Explore Pathways:** Follow different paths through the graph, considering only combinations of ideas explicitly mentioned in the problem.
    * **Identify Solutions & Obstacles:** Use the graph to uncover potential solutions and foresee challenges, based solely on given information.
* Only include nodes and edges that represent explicitly stated information or 100% certain deductions.
* If a connection seems implied but is not explicitly stated, note it as "Potential connection" but do not base any conclusions on it.

**8. Spatial and Object Analysis:**

* **Objective:** Construct a detailed understanding of the spatial relationships and objects involved in the problem.
* **Action:**
    * **Identify Objects:** List all physical objects explicitly mentioned in the problem statement.
    * **Spatial Relationships:** Describe how these objects are positioned relative to each other, using only stated information.
    * **Movement and Dynamics:** If applicable, explain how objects move or interact over time, based on explicit statements.
    * **Constraints:** Note any physical limitations or boundaries explicitly mentioned in the problem space.
    * **Visualization:** Create a mental image or diagram of the problem setup, using only given information.
    * **Object Properties:** List relevant properties of each object that are explicitly stated.
    * **Spatial Implications:** Analyze how the spatial arrangement affects potential solutions or outcomes, based solely on given information.
* Only describe spatial relationships and movements that are explicitly stated in the problem.
* If the problem doesn't provide enough information for a complete spatial analysis, state this clearly.

**9. Bayesian Thinking:**

* Only update beliefs based on explicitly stated information in the problem.
* If new information seems to be implied but is not explicitly stated, note it as "Potential implication" but do not use it to update beliefs.
* Use Bayesian reasoning to refine your approach based on the latest evidence, but only using explicitly stated information.

**10. Ethical Check and Decision-Making Under Uncertainty:**

* **Objective:** Ensure decisions prioritize safety and ethical considerations, especially when facing uncertainty with severe potential risks.
* **Action:**
  - **Assess Known Information:** Review the explicit information provided and identify what is known and what remains uncertain.
  - **Identify Risks:** Determine the severity of possible outcomes (e.g., risk of death, injury, or significant loss).
  - **Probability Consideration:** If multiple options exist and the consequences are severe, consider the probability of each outcome.
  - **Risk Aversion Principle:** When the information is insufficient to mitigate severe risks, prioritize options that minimize or avoid those risks.
  - **Ethical and Safety Considerations:** In situations where one of the outcomes involves significant harm (such as death), recommend actions that avoid any decision that could lead to that outcome unless there is a way to reliably avoid the risk.
  - **Decision Recommendation:** In cases of high uncertainty with severe potential consequences, the safest option, even if it leads to inaction, is often the best choice.

**11. Multiple Solution Generation:**

* **Integrative Approach:** Synthesize all the knowledge you've acquired from the previous steps, including feature analysis, problem decomposition, known and deduced information, graph of thought, and spatial and object analysis.   
* **Solution Exploration:** Brainstorm solutions that are directly supported by the explicit information in the problem statement. Use these insights collectively to generate potential solutions. Aim to generate at least three different solutions

* For each solution, clearly state which explicit information from the problem statement supports it.
* If a solution requires any assumptions beyond the given information, do not include it.
* If there isn't enough information to generate multiple solutions, state this explicitly.
* For each potential solution, ask: "Does the problem explicitly state any effect of this action?"
* If no effect is explicitly stated for an action, note that the effect is unknown or not specified.

**12. Quickest and Easiest Solution:**

* Based ONLY on the explicit information given in the problem statement, what is the simplest & safest answer to the question asked?
* If the problem doesn't explicitly state an effect or outcome, the answer should reflect this lack of stated effect.
* Explain your reasoning, citing exact quotes from the problem statement.
* Make sure the answer is correct based solely on the given information.
* Barring some overwhelming reason explicitly stated in the problem, proceed with this solution.

**13. Reflection:**

* Review your solutions and identify any contradictions with known/deduced information.
* Ensure your conclusions are logically sound and supported by evidence from the explicit problem statement.
* Critically examine if any part of your analysis has gone beyond the explicit wording of the problem.
* If you find any instances where assumptions or inferences were made, revise those parts to stick strictly to the given information.
* If the problem lacks crucial information needed to draw conclusions, highlight this in your reflection.
* Review the question asked in the problem. Does your analysis directly answer this question based solely on the information given?
* If the problem doesn't explicitly state an effect or outcome related to the question, ensure your answer reflects this lack of stated effect.
* If you believe mistakes are being made, note it in the final output then go back through the steps and reassess.

**14. Final Output & Recommendation:**

Present your analysis in this format:

* Begin with a statement: #"REAP Results: :"
* a. **Comprehensive Feature List** (with explicit quotes)
* b. **Sequential and Mechanical Process Check**
* c. **Key Insight Check Results**
* d. **Rephrased Question:** (The core question in its simplest form)
* e. **Known and Deduced Information**
* f. **Problem Decomposition**
* g. **Graph of Thought** (describe connections & insights)
* h. **Spatial and Object Analysis**
* i. **Bayesian Updates** 
* j. **Ethical Check and Decision-Making Under Uncertainty**
* k. **Multiple Solutions**
* l. **Quickest and Easiest Solution**
* m. **Reflection**

**Recommendation:**

* Provide your final recommendation based on all considerations, using only the explicitly stated information and ensuring that the decision adheres to ethical and safety considerations.
* Any uncertainties or areas where more information is needed should be explicitly mentioned.
* If multiple interpretations are possible based on the given information, choose the simpler and safer one.
* Ensure your recommendation directly answers the question asked in the problem while prioritizing safety and ethical considerations in situations involving significant risks.

---
\<Problem\>  

\<Problem\>
\end{lstlisting}

\section{Linguistic Benchmark Dataset}

This appendix presents the Linguistic Benchmark Dataset used in this study, as taken from Williams and Huckle's "Easy Problems That LLMs Get Wrong" \cite{williams2024easy}. The dataset includes a variety of questions designed to test the capabilities of LLMs across different categories such as puzzles, spatial reasoning, counting, and relational tasks. Each question is labeled with a category and is followed by a reference to the corresponding section in the appendix where the model responses are documented.

\renewcommand{\arraystretch}{1.5} 
\begin{longtable}{|c|c|p{10cm}|c|}
\hline
\textbf{No.} & \textbf{Category} & \textbf{Question} & \textbf{Section} \\ \hline
\endfirsthead

\hline
\textbf{No.} & \textbf{Category} & \textbf{Question} & \textbf{Section} \\ \hline
\endhead

\hline
\endfoot

\hline
\endlastfoot

1 & Puzzle & You have six horses and want to race them to see which is fastest. What is the best way to do this? & \hyperref[sec:question1]{D.1} \\ \hline
2 & Puzzle & Suppose you’re on a game show, and you’re given the choice of three doors: Behind one door is a gold bar; behind the others, rotten vegetables. You pick a door, say No. 1, and the host asks you, “Do you want to pick door No. 2 instead?” Is it to your advantage to switch your choice? & \hyperref[sec:question2]{D.2} \\ \hline
3 & Spatial & You are playing Russian roulette with a six-shooter revolver. Your opponent puts in five bullets, spins the chambers, and fires at himself, but no bullet comes out. He gives you the choice of whether or not he should spin the chambers again before firing at you. Should he spin again? & \hyperref[sec:question3]{D.3} \\ \hline
4 & Puzzle & A farmer wants to cross a river and take with him a wolf, a goat, and a cabbage. He has a boat with three secure separate compartments. If the wolf and the goat are alone on one shore, the wolf will eat the goat. If the goat and the cabbage are alone on the shore, the goat will eat the cabbage. How can the farmer efficiently bring the wolf, the goat, and the cabbage across the river without anything being eaten? & \hyperref[sec:question4]{D.4} \\ \hline
5 & Puzzle & Bob has three boxes in front of him—Box A, Box B, and Box C. Bob does not know what is in the boxes. Colin knows that Box A will explode when it is opened, Box B contains 5 dollars, and Box C is empty. Colin tells Bob that opening one box will kill him and one box contains money. Should Bob open a box? & \hyperref[sec:question5]{D.5} \\ \hline
6 & Counting & A robot has 8 arms. There are 5 objects on a table: a knife, a fork, a spoon, a teddy bear, and a doll. The robot picks up each object with an arm. He then shakes hands with himself. & \hyperref[sec:question6]{D.6} \\ \hline
7 & Spatial & Alan, Bob, Colin, Dave, and Emily are standing in a circle. Alan is on Bob’s immediate left. Bob is on Colin’s immediate left. Colin is on Dave’s immediate left. Dave is on Emily’s immediate left. Who is on Alan’s immediate right? & \hyperref[sec:question7]{D.7} \\ \hline
8 & Linguistic & Write me a sentence without any words that appear in The Bible. & \hyperref[sec:question8]{D.8} \\ \hline
9 & Popular Science & Which weighs more, a pound of water, two pounds of bricks, a pound of feathers, or three pounds of air? & \hyperref[sec:question9]{D.9} \\ \hline
10 & Relational & I get out on the top floor (third floor) at street level. How many stories is the building above the ground? & \hyperref[sec:question10]{D.10} \\ \hline
11 & Spatial & In a toy box, there’s a red ball, a blue truck, and a green dinosaur. The red ball is not next to the blue truck, and the green dinosaur is next to the red ball. Which toy is in the middle? & \hyperref[sec:question11]{D.11} \\ \hline
12 & Spatial & Four children—Alex, Bella, Charlie, and Dana—are sitting around a picnic table. Alex is facing Bella. Charlie is sitting to the right of Bella. Who is sitting to the left of Alex? & \hyperref[sec:question2]{D.12} \\ \hline
13 & Spatial & A man leaves home, makes a left turn, walks straight for a km and reaches 300m elevation, makes another left turn and walks straight for a km and reaches 500m elevation, makes another left turn and walks straight for a km and reaches 900m elevation, and turns left again and walks straight for a km. How far away is he from his starting point and what is his final elevation? & \hyperref[sec:question13]{D.13} \\ \hline
14 & Puzzle & A group of four people needs to cross a bridge at night. The bridge is very old and rickety. They have only one torch and because it’s night, the torch is necessary to cross the bridge. Each person walks at a different speed: - A takes 1 minute to cross, - B takes 2 minutes, - C takes 5 minutes, and - D takes 10 minutes. What is the fastest time they can all get across the bridge? & \hyperref[sec:question14]{D.14} \\ \hline
15 & Puzzle & You’re in a room with two doors that lead out. One door leads to certain death, and the other door leads to freedom. There are two guardians, one by each door. One taller guardian always tells the truth and guards the death door, the other always lies. What is the minimum number of questions needed to ask the guards to get to safety? & \hyperref[sec:question15]{D.15} \\ \hline
16 & Puzzle & You have 3 switches in front of you—A, B, and C. You have 3 light bulbs in front of you in the same room—one red, one blue, one purple. They are LED and do not get warm when turned on. You want to know which switch turns on which light bulb. What is the best way to determine this? & \hyperref[sec:question16]{D.16} \\ \hline
17 & Puzzle & A king has 1000 sweet bottles of wine, and one contains a very bitter poison. The poison takes effect exactly 24 hours after consumption. The king needs to find the poisoned bottle in 24 hours for an event. He has 10 prisoners to test the wine. What is the easiest way for him to identify the poisoned bottle? & \hyperref[sec:question17]{D.17} \\ \hline
18 & Linguistic & Write a grammatically correct sentence without reusing any letter more than once. & \hyperref[sec:question18]{D.18} \\ \hline
19 & Spatial & An arrow points up. We rotate it 90 degrees to the left, mirror it along its flat end, and rotate it another 90 degrees to the left. Which direction is it pointing? & \hyperref[sec:question19]{D.19} \\ \hline
20 & Linguistic & Write a sentence where every word starts with the letter A. & \hyperref[sec:question20]{D.20} \\ \hline
21 & Relational & Sally (a girl) has 3 brothers. Each brother has 2 sisters. How many sisters does Sally have? & \hyperref[sec:question21]{D.21} \\ \hline
22 & Spatial & I’m in London and facing west, is Edinburgh to my left or my right? & \hyperref[sec:question22]{D.22} \\ \hline
23 & Counting & Count the number of occurrences of the letter ‘L’ in the word ‘LOLLAPALOOZA’. & \hyperref[sec:question23]{D.23} \\ \hline
24 & Puzzle & How many pairs of twins do you need in a room for there to be at least a 50\% chance that two people have the same birthday? & \hyperref[sec:question24]{D.24} \\ \hline
25 & Puzzle & A partially full hotel has an infinite number of fully furnished rooms. How does it accommodate one more guest? & \hyperref[sec:question25]{D.25} \\ \hline
26 & Puzzle & A runaway trolley is heading down the tracks away from five people upwards of the track. You are near a lever that can switch the trolley to another track. Does it impact people’s lives if you pull the lever? & \hyperref[sec:question26]{D.26} \\ \hline
27 & Puzzle & How do you measure exactly 4 gallons of water with only a 3-gallon, 5-gallon, and 4-gallon jug? & \hyperref[sec:question27]{D.27} \\ \hline
28 & Popular Science & A 2kg tree grows in a planted pot with 10kg of soil. When the tree grows to 3kg, how much soil is left? & \hyperref[sec:question28]{D.28} \\ \hline
29 & Relational & A man steps over a 0.5m hurdle. What is the minimum height he could have jumped? & \hyperref[sec:question29]{D.29} \\ \hline
30 & Popular Science & I walk to my friend’s house, averaging 3mph. How fast would I have to run back to double my average speed for the entire trip? & \hyperref[sec:question30]{D.30} \\ \hline

\end{longtable}
\newpage

\section{LLM Responses}
\label{sec:llm-responses}

\label{subsec:question1}

\subsection{\textbf{\large Question 1: You have six horses and want to race them to see which is fastest. What is the best way to do this?}}
\label{sec:question1}

\noindent\hrulefill 

\noindent\textbf{Human Response - Score 100\%:}
\begin{quote}
Race them on a single race track with at least six lanes - the order in which they cross the finish line determines which is the fastest.
\end{quote}
\noindent\hrulefill 

\noindent\textbf{D.1 Zero-Shot OpenAI o1-preview Response - Score: 100\%}
\begin{quote}
\begin{lstlisting}[basicstyle=\ttfamily\small, breaklines=true]
The most straightforward and efficient way to determine the fastest horse among your six is to race all six horses together in a single race. This approach has several advantages:

1. **Direct Comparison**: Racing all the horses together allows for a direct comparison under the same conditions-same track, weather, and competition-ensuring the result is fair and accurate.

2. **Time Efficiency**: By conducting just one race, you save time and resources compared to organizing multiple heats or elimination rounds.

3. **Simplified Logistics**: Coordinating a single race simplifies scheduling and reduces the complexity of managing multiple races, which can be particularly beneficial if resources (like track availability or jockeys) are limited.

**Additional Considerations**:

- **Track Capacity**: Ensure that the racetrack can safely accommodate all six horses simultaneously. Standard racetracks are typically designed to handle this number, but always confirm to maintain safety.

- **Fair Start**: Use starting gates or another method to ensure all horses begin the race simultaneously, preventing any false starts or advantages.

- **Conditions and Safety**: Check that track conditions are suitable and that all horses are fit and healthy to race, minimizing the risk of injury.

By racing all six horses together, you obtain a clear and immediate answer to which horse is the fastest, making it the best approach in this scenario.
\end{lstlisting}
\end{quote}

\noindent\hrulefill 

\noindent\textbf{D.1 REAP OpenAI o1-preview Response - Score: 100\%}
\begin{quote}
\begin{lstlisting}[basicstyle=\ttfamily\small, breaklines=true]
# "REAP Results:"

a. **Comprehensive Feature List**

- **Objects mentioned:**
  - "six horses" - The number of horses you have.
- **Actors or entities involved:**
  - "You" - The person who has the horses and wants to race them.
- **Actions described:**
  - "want to race them to see which is fastest" - Your intention to race the horses to determine the fastest one.
- **Directional, spatial, relational, and sequential phrases:**
  - "to see which is fastest" - Indicates the purpose of racing the horses is to find the fastest among them.
- **Numerical values or quantities mentioned:**
  - "six horses" - Specifies the total number of horses involved.
- **Causal relationships mentioned:**
  - "race them to see which is fastest" - Racing the horses will result in identifying the fastest one.
- **Stated goals or objectives:**
  - "want to race them to see which is fastest" - The objective is to determine the fastest horse.
- **The question:**
  - "What is the best way to do this?" - Asking for the optimal method to achieve the goal.

b. **Sequential and Mechanical Process Check**

- **Sequence of events:**
  1. "You have six horses."
  2. "want to race them"
  3. "to see which is fastest."
- **Analysis:**
  - The sequence involves owning six horses, wanting to race them, and the goal of determining the fastest horse.
  - There are no explicit constraints on how the races should be conducted.
- **Key Questions:**
  - Does the sequence of events lock in a particular outcome? No specific outcome is locked in beyond determining the fastest horse.

c. **Key Insight Check Results**

- **Key Insight:**
  - There are no stated constraints on how many horses can race at one time.
- **Conclusion:**
  - Racing all six horses together in a single race could directly determine the fastest horse.
- **Reasoning:**
  - Since no limitations are mentioned, the simplest approach is to race all six horses simultaneously.

d. **Rephrased Question**

- **Core Question:**
  - How can you best race your six horses to determine which one is the fastest?

e. **Known and Deduced Information**

- **Explicit Facts:**
  - "You have six horses." - You possess six horses.
  - "want to race them to see which is fastest." - Your goal is to find out which horse is the fastest by racing them.
  - "What is the best way to do this?" - You are seeking the most effective method to achieve this goal.
- **Feature-Based Deduction:**
  - **Deduction:** Racing the horses together will determine which is fastest.
    - **Logical Basis:** Racing the horses allows for direct comparison.
    - **Based on:** "want to race them to see which is fastest."

f. **Problem Decomposition**

- **Component:** Ownership of six horses.
  - **Supporting Quote:** "You have six horses."
- **Component:** Desire to find the fastest horse.
  - **Supporting Quote:** "want to race them to see which is fastest."
- **Component:** Need for the best method to race them.
  - **Supporting Quote:** "What is the best way to do this?"
- **Interactions:**
  - Racing the horses is the method to achieve the goal of identifying the fastest horse.

g. **Graph of Thought**

- **Nodes:**
  - "You"
  - "Six horses"
  - "Race"
  - "Fastest horse"
- **Edges (Relationships):**
  - "You" **have** "six horses."
  - "You" **want to race** "six horses."
  - "Race" **to see** "which is fastest."
- **Insights:**
  - The main objective is to determine the fastest horse by racing the six horses you own.

h. **Spatial and Object Analysis**

- **Objects Identified:**
  - "Six horses"
- **Spatial Relationships:**
  - No specific spatial information is provided.
- **Movement and Dynamics:**
  - "race them" implies the horses will run over a course to compare their speeds.
- **Constraints:**
  - No physical limitations or rules are specified.
- **Visualization:**
  - Imagining the six horses racing together on a track to determine the fastest.

i. **Bayesian Updates**

- **Belief Update:**
  - No new explicit information to alter initial understanding.
- **Conclusion:**
  - The simplest solution remains valid given the lack of additional data.

j. **Ethical Check and Decision-Making Under Uncertainty**

- **Assessment:**
  - No ethical issues or severe risks are explicitly mentioned.
- **Decision Recommendation:**
  - Proceed with the method that safely determines the fastest horse without causing harm.

k. **Multiple Solutions**

1. **Solution 1: Race All Six Horses Together**
   - **Supporting Quotes:**
     - "You have six horses."
     - "want to race them to see which is fastest."
   - **Explanation:**
     - Conduct a single race with all six horses to directly observe which one is the fastest.
2. **Solution 2: Conduct Multiple Heats**
   - **Potential Issue:**
     - No explicit need for multiple races unless constraints exist.
   - **Explanation:**
     - Race the horses in smaller groups and compare winners, but this introduces unnecessary complexity without stated constraints.
3. **Solution 3: Time Trial Race**
   - **Limitation:**
     - Timing equipment is not mentioned.
   - **Explanation:**
     - Have each horse run individually and record times, but this relies on unstated resources.

l. **Quickest and Easiest Solution**

- **Recommendation:**
  - Race all six horses together in a single race.
- **Reasoning:**
  - This method directly addresses the goal using the resources mentioned.
- **Supporting Quotes:**
  - "You have six horses."
  - "want to race them to see which is fastest."

m. **Reflection**

- **Consistency Check:**
  - The proposed solution aligns with all explicit information.
- **Assumption Review:**
  - No assumptions beyond the provided information were made.
- **Question Alignment:**
  - The solution directly answers, "What is the best way to do this?"
- **Conclusion:**
  - Racing all six horses together is the most straightforward and effective method given the lack of constraints.

---

**Recommendation:**

Based on the information provided, the best way to determine the fastest horse is to race all six horses together in a single race. This approach is straightforward, utilizes the resources you have ("six horses"), and directly achieves your goal to "race them to see which is fastest" without any unstated assumptions or need for additional equipment or multiple races.
\end{lstlisting}
\end{quote}

\noindent\hrulefill 

\noindent\textbf{D.1 Zero-Shot OpenAI o1-mini Response - Score: 0\%}
\begin{quote}
\begin{lstlisting}[basicstyle=\ttfamily\small, breaklines=true]
Determining the fastest horse among six can be approached efficiently by organizing races strategically. The method below minimizes the number of races required while ensuring you accurately identify the fastest horse. This approach assumes that:

- **Race Limit:** You can race up to **three horses at a time**.
- **Race Outcomes:** You only know the **relative order** of finishers in each race (i.e., 1st, 2nd, 3rd) without exact times.

Here's a step-by-step strategy:

### **Step 1: Divide and Race Initial Groups**

1. **Divide the Horses:**
   - **Group A:** Horse 1, Horse 2, Horse 3
   - **Group B:** Horse 4, Horse 5, Horse 6

2. **Race Each Group Separately:**
   - **Race 1:** Race Group A.
     - **Outcome:** Suppose Horse 1 finishes 1st, Horse 2 2nd, Horse 3 3rd.
   - **Race 2:** Race Group B.
     - **Outcome:** Suppose Horse 4 finishes 1st, Horse 5 2nd, Horse 6 3rd.

   *After Step 1, you have:*
   - **Potential Fastest:** Horse 1 and Horse 4
   - **Eliminated Horses:** Horses 2, 3, 5, and 6 cannot be the fastest since there's at least one horse in their group that's faster.

### **Step 2: Race the Group Winners**

3. **Race the Top Performers:**
   - **Race 3:** Race Horse 1 (Winner of Group A) vs. Horse 4 (Winner of Group B).

   - **Outcome:**
     - If **Horse 1** wins, it is the fastest.
     - If **Horse 4** wins, it is the fastest.

*Total Races Conducted: 3*

### **Summary of Results:**

- **Fastest Horse:** The winner of Race 3 is the fastest among all six horses.
- **Rationale:** By initially racing horses in smaller groups and then racing the top performers, you ensure that all other horses that didn't win their initial races are systematically eliminated from contention for being the fastest.

### **Advantages of This Method:**

- **Efficiency:** Only **3 races** are needed to determine the fastest horse among six.
- **Logical Elimination:** Ensures that no potential fastest horse is overlooked by comparing group winners directly.
- **Simplicity:** Easy to implement without needing to track complex race times or additional data.

### **Considerations:**

- **Race Capacity:** This method assumes you can race three horses at a time. If the race capacity changes (e.g., only two horses can race at once), the strategy would need to be adjusted accordingly.
- **Ties:** In the rare case of a tie within a race, additional races may be required to break the tie.

By following this structured approach, you can confidently determine the fastest horse among six with a minimal number of races.
\end{lstlisting}
\end{quote}

\noindent\hrulefill 

\noindent\textbf{D.1 REAP OpenAI o1-mini Response - Score: 100\%}
\begin{quote}
\begin{lstlisting}[basicstyle=\ttfamily\small, breaklines=true]
# REAP Results:
a. **Comprehensive Feature List** (with explicit quotes)
- **Objects Mentioned:**
  - "six horses"
- **Actors or Entities Involved:**
  - Implicitly, the person conducting the race (not explicitly stated).
- **Actions Described:**
  - "race them to see which is fastest."
- **Directional, Spatial, Relational, and Sequential Phrases:**
  - None explicitly mentioned.
- **Stated Conditions or Constraints:**
  - "six horses"
- **Numerical Values or Quantities:**
  - "six"
- **Temporal Information:**
  - None explicitly mentioned.
- **Spatial Information:**
  - None explicitly mentioned.
- **Causal Relationships:**
  - Implied: Racing determines which horse is fastest.
- **Stated Goals or Objectives:**
  - "see which is fastest."
*Potential implications:* None, adhering strictly to literal interpretation.
---
b. **Sequential and Mechanical Process Check**
- **Sequence of Events:**
  1. Race the six horses.
  2. Determine the fastest based on race outcome.
- **Influence on Outcome:**
  - The racing sequence directly influences identifying the fastest horse.
- **Does the sequence lock in a particular outcome?**
  - Yes, the fastest horse is determined by its performance in the race.
---
c. **Key Insight Check Results**
- **Key Insight:** To identify the fastest horse among six, conducting a race where all six compete simultaneously is the most straightforward method.
---
d. **Rephrased Question:**
What is the most effective method to determine the fastest horse out of six through racing?
---
e. **Known and Deduced Information**
- **Explicit Facts:**
  - "six horses"
  - "race them to see which is fastest."
- **Feature-Based Deduction:**
  - **Deduction:** Racing the six horses will allow for a direct comparison of their speeds.
    - **Logical Basis:** "race them to see which is fastest."
    - **Based on:** "six horses", "race them to see which is fastest."
---
f. **Problem Decomposition**
- **Component/Subcomponent:** Number of Horses
  - **Supporting Quote(s):** "six horses"
- **Component/Subcomponent:** Objective
  - **Supporting Quote(s):** "see which is fastest."
- **Component/Subcomponent:** Method
  - **Supporting Quote(s):** "race them"
- **Component/Subcomponent:** Outcome
  - **Supporting Quote(s):** "which is fastest"
---
g. **Graph of Thought** (describe connections & insights)
- **Nodes:**
  - Six Horses
  - Race Event
  - Fastest Determination
- **Edges/Relationships:**
  - Six Horses participate in Race Event.
  - Race Event leads to Fastest Determination.
---
h. **Spatial and Object Analysis**
- **Objects:**
  - Six horses
- **Spatial Relationships:**
  - All horses are to be raced together to determine speed.
- **Movement and Dynamics:**
  - Horses will be in motion during the race.
- **Constraints:**
  - Limited to six horses.
- **Visualization:**
  - Imagine all six horses starting simultaneously from the same point, racing towards a finish line to determine the fastest.
- **Object Properties:**
  - Each horse has varying speeds contributing to the outcome.
- **Spatial Implications:**
  - Racing all horses together ensures a direct comparison under identical conditions.
---
i. **Bayesian Updates**
- No new information provided beyond the explicit problem statement. Beliefs remain based solely on the given information.
---
j. **Ethical Check and Decision-Making Under Uncertainty**
- **Assess Known Information:**
  - Intent is to determine the fastest horse through racing.
- **Identify Risks:**
  - Potential risks to horse welfare during racing (not specified in the problem).
- **Probability Consideration:**
  - Outcome is objective based on performance.
- **Risk Aversion Principle:**
  - Ensure the race is conducted safely to minimize harm (ethical, though not specified).
- **Ethical and Safety Considerations:**
  - Prioritize horse safety and fair competition.
- **Decision Recommendation:**
  - Proceed with racing in a manner that ensures the well-being of all horses.
---
k. **Multiple Solutions**
1. **Solution 1:** Conduct a single race with all six horses competing simultaneously.
   - **Supporting Information:** "race them to see which is fastest."
2. **Solution 2:** Organize multiple races in smaller groups (e.g., two races of three horses each) and have the winners race against each other.
   - **Supporting Information:** Implied possibility based on "race them."
3. **Solution 3:** Use timed trials where each horse runs separately under identical conditions to record their speeds.
   - **Supporting Information:** Alternative method to "race them."
*Note:* Solutions 2 and 3 involve assumptions beyond the explicit instruction to "race them," therefore Solution 1 is the most directly supported.
---
l. **Quickest and Easiest Solution**
- **Solution:** Race all six horses together in a single race and identify the horse that finishes first as the fastest.
- **Supporting Information:** "race them to see which is fastest."
- **Reasoning:** This method directly aligns with the goal using the simplest approach without additional assumptions.
---
m. **Reflection**
- The analysis strictly adheres to the explicit information provided.
- No assumptions or inferences beyond the literal statements were made.
- The recommended solution directly answers the rephrased question based on the known and deduced information.
- Ethical considerations, though not explicitly stated, were acknowledged to ensure responsible decision-making.
---
**Recommendation:**
The best way to determine the fastest horse among six is to **race all six horses together in a single race** and identify the horse that finishes first as the fastest. This approach directly aligns with the objective of "racing them to see which is fastest" and employs the simplest and most effective method without introducing additional steps or assumptions. Ensure that the race is conducted safely and ethically to maintain the well-being of all participating horses.
\end{lstlisting}
\end{quote}
\noindent\hrulefill 

\newpage

\noindent\textbf{D.1 Zero-Shot OpenAI GPT-4o Response - Score: 0\%}
\begin{quote}
\begin{lstlisting}[basicstyle=\ttfamily\small, breaklines=true]
To determine the fastest horse out of six horses, the best method is to run a series of races. Here's a step-by-step approach:

### Step 1: Divide the Horses into Two Groups
- Split the six horses into two groups of three horses each.

### Step 2: Conduct Initial Races
- **Race 1**: Race the first group of three horses.
- **Race 2**: Race the second group of three horses.

### Step 3: Record the Results
- From each race, note the order in which the horses finish. This will give you some preliminary information about which horses are faster within their groups.

### Step 4: Final Race with Top Horses
- **Race 3**: Take the winner of Race 1 and the winner of Race 2 and race them against each other. This race will determine the fastest horse overall.

### Step 5: Determine the Fastest Horse
- The winner of Race 3 is the fastest horse.

This method ensures that each horse is only raced a few times, and the fastest horse is accurately identified. If you need to determine the exact order of all six horses, more races would be required, but this approach will give you the fastest horse efficiently.
\end{lstlisting}
\end{quote}

\noindent\hrulefill 

\noindent\textbf{D.1 REAP OpenAI GPT-4o Response - Score: 100\%}
\begin{quote}
\begin{lstlisting}[basicstyle=\ttfamily\small, breaklines=true]
# REAP Results:

**a. Comprehensive Feature List:**

- **Objects Mentioned:**
  - "six horses"
  - "race"
  - "fastest"

- **Actors Involved:**
  - None explicitly mentioned other than the horses.

- **Actions Described:**
  - "want to race them to see which is fastest"

- **Directional, Spatial, Relational, and Sequential Phrases:**
  - "to see which is fastest" (indicates the goal of determining the fastest horse)

- **Conditions or Constraints:**
  - None explicitly mentioned in the problem statement.

- **Numerical Values or Quantities:**
  - "six horses"

- **Temporal Information:**
  - None explicitly mentioned.

- **Spatial Information:**
  - None explicitly mentioned.

- **Causal Relationships Mentioned:**
  - "race them to see which is fastest" (implies racing will help determine the fastest horse)

- **Stated Goals or Objectives:**
  - "to see which is fastest"

**b. Sequential and Mechanical Process Check:**

- The process implied is a race involving all six horses.
- The race's purpose is to determine which horse is the fastest.
- The sequence of the race would likely involve all six horses running together, but this is not explicitly stated.

**c. Key Insight Check Results:**

- The problem's key insight lies in the fact that it asks for the "best way" to determine the fastest horse among six. The literal interpretation would involve directly racing the horses against each other to compare their speeds.

**d. Rephrased Question:**
- What is the best method to determine the fastest horse among six horses?

**e. Known and Deduced Information:**

- **Explicit Fact:** "You have six horses and want to race them to see which is fastest."
  - This implies the objective is to determine the fastest horse through some form of competition or comparison.
  
- **Feature-Based Deduction:**
  - Deduction: All six horses must participate in the race to compare their speeds.
  - Logical Basis: The problem states the goal is "to see which is fastest," implying all six must be involved in the comparison.
  - Based on: "You have six horses" and "want to race them to see which is fastest."

**f. Problem Decomposition:**

- **Component:** Six horses.
  - **Supporting Quote:** "You have six horses."
  
- **Component:** Race to determine speed.
  - **Supporting Quote:** "want to race them to see which is fastest."
  
- **Interaction:** The horses must be raced against each other to compare their speeds.
  - **Supporting Quote:** "race them to see which is fastest."

**g. Graph of Thought:**

- **Nodes:** Six horses, racing process, determining the fastest.
- **Edges:** 
  - Horses are connected to the racing process.
  - Racing process connected to the outcome of identifying the fastest horse.

**h. Spatial and Object Analysis:**

- **Objects:** Six horses.
- **Spatial Relationships:** Not explicitly stated, but inferred that they will be on a race track together.
- **Movement:** The horses will race, moving from a starting point to a finish line to determine their relative speeds.

**i. Bayesian Updates:**

- No new evidence is provided, so beliefs are based on the literal interpretation of the problem statement.

**j. Ethical Check and Decision-Making Under Uncertainty:**

- **Assessment:** The problem doesn't mention any risks or severe consequences; thus, standard competitive practices can be assumed safe.
- **Decision Recommendation:** Race all six horses together in a straightforward race to determine the fastest horse.

**k. Multiple Solutions:**

1. **Race All Six Horses Simultaneously:** 
   - **Explicit Support:** Racing them all at once gives a clear comparison of their speeds.
   - **Simple and Direct Solution:** Allows for immediate comparison.
   
2. **Divide into Heats and Race Finalists:**
   - **Implicit Support:** If the track only allows a limited number of horses at a time, this approach would allow for fair comparison.
   - **Potential Drawback:** The problem doesn't mention the need for heats, so this may complicate the process unnecessarily.

3. **Time Trials:**
   - **Explicit Support:** Each horse races individually against the clock.
   - **Consideration:** This might not be the fastest method, but it could reduce interference between horses.

**l. Quickest and Easiest Solution:**

- **Recommendation:** Race all six horses simultaneously. This method directly addresses the problem's requirement to see which is fastest and provides an immediate answer.

**m. Reflection:**

- **Review:** The solution of racing all six horses simultaneously is straightforward and adheres strictly to the problem's literal interpretation.
- **Consideration of Alternatives:** While other methods could work, they add unnecessary complexity without additional benefit.

**Final Output & Recommendation:**

- **Recommendation:** Race all six horses simultaneously in a single race. This method directly fulfills the problem's goal with minimal complexity and no assumptions beyond what is explicitly stated.
\end{lstlisting}
\end{quote}

\noindent\hrulefill
\newpage

\noindent\textbf{D.1 Zero-Shot Open AI GPT-4o-mini Response - Score: 0\%}
\begin{quote}
\begin{lstlisting}[basicstyle=\ttfamily\small, breaklines=true]
To determine the fastest horse out of six with the fewest number of races, you can use a strategic approach that minimizes the total number of races. Here's a step-by-step method:

1. **Group and Race**: Divide the six horses into two groups of three. Race each group separately. This will give you two races (one for each group).

2. **Identify the Top Horses**: From each race, identify the fastest horse. Let's call these the winners of each group. You now have two horses that you know are the fastest from their respective groups.

3. **Final Race**: Race the two winners against each other. The winner of this race is the fastest horse overall.

### Summary
- **Race 1**: Group 1 horses (H1, H2, H3)
- **Race 2**: Group 2 horses (H4, H5, H6)
- **Race 3**: Race the winners of Race 1 and Race 2

This approach requires only **3 races** to determine the fastest horse out of six.
\end{lstlisting}
\end{quote}
\noindent\hrulefill

\noindent\textbf{D.1 REAP OpenAI GPT-4o-mini Response - Score: 100\%}
\begin{quote}
\begin{lstlisting}[basicstyle=\ttfamily\small, breaklines=true]
# REAP Results:

## a. Comprehensive Feature List

1. **Objects Mentioned:**
   - "six horses"

2. **Actions Described:**
   - "race them to see which is fastest"

3. **Goals or Objectives:**
   - Determine which of the six horses is the fastest

## b. Sequential and Mechanical Process Check

1. **Sequential Process:**
   - To determine the fastest horse, a race among the six horses is required. This involves setting up a race where the horses compete against each other to determine their relative speeds.

2. **Key Questions:**
   - Does the sequence of racing reveal the fastest horse?
   - What steps are necessary to determine the fastest horse from the six?

## c. Key Insight Check Results

1. **Key Insight:**
   - To determine the fastest horse out of six, you need to have a way to compare their speeds directly, which can be done through racing.

2. **Explanation:**
   - The problem statement explicitly mentions racing the horses to determine which one is the fastest, implying that a comparative race is the solution.

## d. Rephrased Question

- What is the best method to determine which of the six horses is the fastest?

## e. Known and Deduced Information

1. **Explicit Facts:**
   - "You have six horses"
   - "Race them to see which is fastest"

2. **Feature-Based Deduction:**
   - Deduction: To determine the fastest horse, you must conduct a race or a series of races among the six horses.
   - Logical Basis: The only method mentioned to determine which horse is the fastest is by racing them.
   - Based on: "Race them to see which is fastest"

## f. Problem Decomposition

1. **Component/Subcomponent:**
   - Racing
     - Supporting Quote(s): "Race them to see which is fastest"
     - Deduction: Racing the horses allows for direct comparison of their speeds to determine the fastest one.
     - Based on: "Race them to see which is fastest"

## g. Graph of Thought

1. **Key Concepts:**
   - Six horses
   - Racing as a method to determine speed

2. **Relationships:**
   - Racing compares the speeds of the horses.

3. **Construct the Graph:**
   - **Nodes:** Six horses, Racing
   - **Edges:** Racing connects to comparing speeds and determining the fastest.

4. **Explore Pathways:**
   - Direct racing: Compare all six horses to determine the fastest.

## h. Spatial and Object Analysis

1. **Objects:**
   - Six horses

2. **Spatial Relationships:**
   - The horses need to be placed in a race setup where they can compete against each other.

3. **Movement and Dynamics:**
   - The horses will run in the race, and their relative positions at the end will determine their speed rankings.

4. **Constraints:**
   - The problem does not specify constraints, so assume the typical constraints of a race: space for racing, fair conditions, etc.

## i. Bayesian Updates

1. **Updates:**
   - The best method for determining the fastest horse is through direct racing, based on the problem's explicit statement.

## j. Ethical Check and Decision-Making Under Uncertainty

1. **Assess Known Information:**
   - The problem states that racing is the method to find the fastest horse.

2. **Identify Risks:**
   - There are no specific risks mentioned in the problem.

3. **Decision Recommendation:**
   - The safest and most straightforward method is to conduct a race.

## k. Multiple Solutions

1. **Solution Exploration:**
   - **Solution 1:** Conduct a single race with all six horses. This will directly provide the fastest horse.
   - **Solution 2:** Conduct multiple races if needed to determine rankings, but a single race can provide the result.
   - **Solution 3:** If there are specific constraints or rules for racing not mentioned, adjust accordingly, but the core method remains racing.

2. **Support:**
   - All solutions are supported by the explicit need to determine the fastest horse through a race.

## l. Quickest and Easiest Solution

1. **Solution:** Conduct a single race with all six horses to determine the fastest.
2. **Reasoning:** The simplest and most direct method to determine the fastest horse based on the given information is to race all the horses together.

## m. Reflection

1. **Contradictions:**
   - There are no contradictions with the information provided.

2. **Logical Soundness:**
   - The solution adheres to the problem's explicit instructions without assuming additional information.

3. **Final Recommendation:**
   - The best way to determine the fastest horse is to conduct a race among all six horses. This method is directly supported by the problem's requirement to find the fastest horse.

**Recommendation:** Conduct a single race with all six horses to determine which one is the fastest. This method is straightforward, adheres to the given information, and effectively meets the problem's objective.
\end{lstlisting}
\end{quote}
\noindent\hrulefill

\noindent\textbf{D.1 Zero-Shot Google Gemini 1.5 Pro Response - Score: 0\%}
\begin{quote}
\begin{lstlisting}[basicstyle=\ttfamily\small, breaklines=true]
The best way to race your six horses and determine the fastest one is to hold a series of timed trials on a standard racetrack. Here's how to do it:

**1. Preparation**

* **Racetrack:** Secure access to a standard horse racing track.
* **Timing Equipment:** Arrange for professional timing equipment and personnel. 
* **Jockeys:** Hire experienced jockeys familiar with racing multiple horses.
* **Veterinarian:** Have a veterinarian on-site to ensure the horses are fit to race.

**2. The Races**

* **Multiple Heats:** Divide the six horses into two groups of three.
* **Timed Trials:** Each group races separately, with each horse timed individually.
* **Rest Period:** Allow adequate rest between races for the horses to recover.
* **Final Race:** The top two horses from each heat compete in a final race.

**3. Determining the Winner**

* **Fastest Time:** The horse with the fastest time in the final race is the winner.

**Important Considerations**

* **Horse Health and Safety:** Prioritize the health and well-being of your horses. Ensure they are in good condition and properly trained for racing.
* **Fairness:** Make sure the races are conducted fairly, with all horses having an equal chance to win.
* **Regulations:** Adhere to all local and national horse racing regulations.

**Alternative:** If you don't have access to a racetrack, you could consider hiring a professional horse trainer to evaluate your horses and provide an assessment of their speed and racing potential.
\end{lstlisting}
\end{quote}
\noindent\hrulefill

\noindent\textbf{D.1 REAP Google Gemini 1.5 Pro Response - Score: 100\%}
\begin{quote}
\begin{lstlisting}[basicstyle=\ttfamily\small, breaklines=true]
# REAP Results:

a. **Comprehensive Feature List** (with explicit quotes)

* "You have six horses" 
* "want to race them" 
* "to see which is fastest"

b. **Sequential and Mechanical Process Check**

* The problem describes a simple sequence: having horses -> racing them -> determining the fastest

c. **Key Insight Check Results**

* The key insight is that the goal is to determine the fastest horse among six. This requires a race where all horses compete simultaneously

d. **Rephrased Question:** (The core question in its simplest form)

How can you race six horses to determine the fastest one?

e. **Known and Deduced Information**

* Explicit Facts:
    * "You have six horses" - This is the number of horses involved in the race
    * "want to race them" - The intention is to conduct a horse race
    * "to see which is fastest" - The objective of the race is to identify the fastest horse

f. **Problem Decomposition**

* Component/Subcomponent: Organizing a horse race with six horses
    * Supporting Quote(s): "You have six horses", "want to race them"
* Component/Subcomponent: Determining the fastest horse
    * Supporting Quote(s): "to see which is fastest"

g. **Graph of Thought** (describe connections & insights)

* Key Concepts: Horses, Race, Fastest Horse
* Relationships:
    * Horses participate in the Race
    * The Race determines the Fastest Horse
* Insights:
    * Need to organize a race where all six horses compete fairly
    * The horse that finishes first is the fastest

h. **Spatial and Object Analysis**

* Objects: Six horses, Race track (implied)
* Spatial Relationships:
    * Horses are positioned at the starting line of the race track (implied)
    * Horses move along the race track (implied)
* Movement and Dynamics:
    * Horses start simultaneously and run towards the finish line (implied)
* Constraints:
    * All horses must have a fair chance to win (implied)

i. **Bayesian Updates** 

* No updates needed as there is no additional information provided

j. **Ethical Check and Decision-Making Under Uncertainty**

* Ensure the race is conducted fairly and safely for both horses and jockeys

k. **Multiple Solutions**

* Solution 1: Traditional Race on a Track
    * Supporting Quote(s): "want to race them" 
    * Have all six horses race simultaneously on a standard race track. The first horse to cross the finish line is the fastest
* Solution 2: Time Trials
    * Supporting Quote(s): "want to race them", "to see which is fastest" 
    * Have each horse run the track individually and time them. The horse with the fastest time is the fastest overall
* Solution 3: Elimination Races
    * Supporting Quote(s): "want to race them", "to see which is fastest"
    * Divide the horses into groups and have them race. The winners from each group then race each other to determine the overall fastest horse

l. **Quickest and Easiest Solution**

* The quickest and easiest solution is to have all six horses race simultaneously on a standard race track. The first horse to cross the finish line is the fastest

m. **Reflection**

* The analysis directly answers the question of how to determine the fastest horse
* All conclusions are based solely on the explicit information provided in the problem statement
* No assumptions or inferences were made beyond the given information

**Recommendation:**

The recommended solution is to organize a traditional horse race on a standard race track where all six horses compete simultaneously. The horse that crosses the finish line first is declared the fastest. Ensure that the race is conducted fairly, providing all horses with an equal opportunity to win, and prioritizing the safety and well-being of both the horses and the jockeys.
\end{lstlisting}
\end{quote}
\noindent\hrulefill
\newpage

\noindent\textbf{D.1 Zero-Shot Claude 3.5 Sonnet Response - Score: 0\%}
\begin{quote}
\begin{lstlisting}[basicstyle=\ttfamily\small, breaklines=true]
To find the fastest horse among six with the minimum number of races, you can use a tournament-style approach. Here's an efficient method:

1. First round: Run two races with three horses each.

2. Second round: Take the top two horses from each of the first two races (four horses total) and race them against each other.

3. Final analysis:
   - The winner of the second round is the fastest horse overall.
   - The second-place horse in the second round is the second-fastest overall.
   - To determine the third-fastest, compare:
     a) The third-place horse from the second round
     b) The winner of the first-round race that didn't produce the overall winner

If you need to determine the exact ranking of all six horses, you might need one or two additional races. But if you only need to find the fastest horse, this method requires just three races in total.

This approach minimizes the number of races while ensuring that the fastest horse is identified. It's based on the principle that the fastest horse will always win its races, so we don't need to race every horse against every other horse.

Would you like me to elaborate on any part of this strategy or discuss alternative methods?
\end{lstlisting}
\end{quote}
\noindent\hrulefill

\noindent\textbf{D.1 REAP Claude 3.5 Sonnet Response - Score: 100\%}
\begin{quote}
\begin{lstlisting}[basicstyle=\ttfamily\small, breaklines=true]
# REAP Results:

a. **Comprehensive Feature List** (with explicit quotes):
- "You have six horses"
- "want to race them"
- "to see which is fastest"

b. **Sequential and Mechanical Process Check**:
No explicit sequential or mechanical process is described in the problem statement.

c. **Key Insight Check Results**:
No immediate key insight that provides an obvious solution. The problem requires analysis to determine the best approach.

d. **Rephrased Question:** 
What is the most efficient method to determine the fastest among six horses through racing?

e. **Known and Deduced Information**:

Explicit Facts:
- "You have six horses" - This specifies the exact number of horses to be raced.
- "want to race them" - This indicates the method of comparison is through racing.
- "to see which is fastest" - This clarifies the goal of the races.

Feature-Based Deduction:
Deduction: At least one race must be conducted.
Logical Basis: Racing is explicitly stated as the method to determine speed.
Based on: "want to race them", "to see which is fastest"

f. **Problem Decomposition**:

Component: Number of horses
Supporting Quote: "You have six horses"

Component: Method of comparison
Supporting Quote: "want to race them"

Component: Goal of comparison
Supporting Quote: "to see which is fastest"

g. **Graph of Thought**:

Nodes:
1. Six horses
2. Racing
3. Determining fastest

Edges:
1 -> 2: Six horses participate in racing
2 -> 3: Racing is used to determine the fastest

h. **Spatial and Object Analysis**:
The problem does not provide explicit information about the spatial arrangement or physical properties of the race track or the horses. Therefore, a detailed spatial analysis cannot be conducted based on the given information.

i. **Bayesian Updates**:
No new information is provided beyond the initial problem statement, so no Bayesian updates are necessary.

j. **Ethical Check and Decision-Making Under Uncertainty**:
The problem does not explicitly mention any severe risks or ethical concerns related to racing horses. However, it's important to note that horse racing should always be conducted with proper safety measures and animal welfare considerations in mind.

k. **Multiple Solutions**:

1. Race all six horses simultaneously:
   Based on: "You have six horses", "want to race them"
   This solution assumes a race track that can accommodate all six horses at once.

2. Conduct multiple races with smaller groups of horses:
   Based on: "You have six horses", "want to race them", "to see which is fastest"
   This solution involves dividing the horses into smaller groups for multiple races.

3. Use a tournament-style elimination system:
   Based on: "You have six horses", "want to race them", "to see which is fastest"
   This solution involves racing horses in pairs and advancing winners to subsequent rounds.

l. **Quickest and Easiest Solution**:
The quickest and easiest solution, based solely on the given information, would be to race all six horses simultaneously in a single race. This method requires only one race to determine the fastest horse.

Supporting quotes:
"You have six horses"
"want to race them"
"to see which is fastest"

Reasoning: A single race with all horses participates would provide a direct comparison of their speeds in the least number of races.

m. **Reflection**:
The analysis is based solely on the explicit information provided in the problem statement. No assumptions were made about the race track size, horse endurance, or other factors not mentioned. The solutions provided are logical given the limited information, but it's important to note that in real-world scenarios, additional factors would likely need to be considered for optimal and safe horse racing.

**Recommendation**:

Based on the given information and considering efficiency, the recommended method to determine the fastest among six horses is to conduct a single race with all six horses participating simultaneously. This approach directly addresses the goal of seeing "which is fastest" while minimizing the number of races required.

However, it's crucial to note that this recommendation assumes ideal conditions not explicitly stated in the problem, such as a race track that can safely accommodate all six horses at once. In real-world application, safety considerations for both horses and jockeys would be paramount, and multiple races or alternative methods might be necessary depending on track size, regulations, and other factors not specified in the problem statement.

If there are any constraints or additional considerations not mentioned in the problem statement, they should be taken into account before implementing this solution. The problem lacks information on race track specifications, horse endurance, and safety protocols, which would be critical factors in a real-world scenario.
\end{lstlisting}
\end{quote}
\noindent\hrulefill

\label{subsec:question2}
\subsection{\textbf{\large Question 2: Suppose you're on a game show, and you're given the choice of three doors: Behind one door is a gold bar; behind the others, rotten vegetables. You pick a door, say No. 1, and the host asks you 'Do you want to pick door No. 2 instead?' Is it to your advantage to switch your choice?}}
\label{sec:question2}

\noindent\hrulefill 

\noindent\textbf{Human Response - Score 100\%:}
\begin{quote}
It is not an advantage to switch. It makes no difference if I switch or not because no additional material information has been provided since the initial choice.
\end{quote}

\noindent\hrulefill 

\noindent\textbf{D.2 Zero-Shot OpenAI o1-preview Response - Score: 100\%}
\begin{quote}
\begin{lstlisting}[basicstyle=\ttfamily\small, breaklines=true]
In this scenario, you're faced with a choice among three doors, one of which hides a gold bar (the prize), and the other two hide rotten vegetables. You initially pick Door No. 1. The host then asks if you want to switch your choice to Door No. 2.

Unlike the classic Monty Hall problem, where the host reveals information by opening a door to show a non-prize (which affects the probabilities), in this case, the host doesn't provide any additional information about what's behind the doors. He simply offers you the option to switch to Door No. 2.

Initially, each door has an equal probability (1/3) of hiding the prize. Since the host's offer doesn't change any probabilities (no doors are opened, and no additional information is given), the chance that the prize is behind Door No. 1 remains 1/3, and the chance it's behind Door No. 2 is also 1/3. Switching doesn't improve your odds; it merely swaps one 1/3 chance for another 1/3 chance.

Therefore, there is no statistical advantage in switching your choice in this scenario. Your chances of winning remain the same whether you stay with Door No. 1 or switch to Door No. 2.

**Answer:** No-the odds are unchanged, so switching doesn't help; there's no advantage to switching doors.
\end{lstlisting}
\end{quote}
\noindent\hrulefill 

\noindent\textbf{D.2 REAP OpenAI o1-preview Response - Score: 100\%}
\begin{quote}
\begin{lstlisting}[basicstyle=\ttfamily\small, breaklines=true]
# REAP Results:

a. **Comprehensive Feature List** (with explicit quotes)
- **Context:** "you're on a game show"
  - Potential significance: Establishes the setting of the scenario.
- **Choice Offered:** "you're given the choice of three doors"
  - Potential significance: Indicates there are three options to choose from.
- **Prize Location:** "Behind one door is a gold bar"
  - Potential significance: Only one door holds the prize.
- **Non-Prize Items:** "behind the others, rotten vegetables"
  - Potential significance: The other two doors do not have the prize.
- **Initial Selection:** "You pick a door, say No.1"
  - Potential significance: You have made an initial choice of Door No.1.
- **Host's Question:** "the host asks you 'Do you want to pick door No. 2 instead?'"
  - Potential significance: The host offers you the option to switch to Door No.2.
- **Main Question:** "Is it to your advantage to switch your choice?"
  - Potential significance: The problem is asking if switching improves your chances of winning.

b. **Sequential and Mechanical Process Check**
- **Sequence of Events:**
  1. "you're given the choice of three doors"
  2. "You pick a door, say No.1"
  3. "the host asks you 'Do you want to pick door No. 2 instead?'"
- **Analysis:**
  - No doors are revealed or opened.
  - The host offers a switch without providing additional information.
  - The sequence does not imply any change in probabilities.

c. **Key Insight Check Results**
- **Observation:** Since no new information is provided (no doors are opened), the probabilities remain unchanged.
- **Conclusion:** Switching does not offer an advantage based solely on the explicit information given.

d. **Rephrased Question:**
"Should you switch from Door No.1 to Door No.2 when the host offers, and does this increase your chances of winning the gold bar?"

e. **Known and Deduced Information**
**Explicit Facts:**
- "you're on a game show" - The scenario is a controlled game environment.
- "you're given the choice of three doors" - There are three possible choices.
- "Behind one door is a gold bar" - One door contains the prize.
- "behind the others, rotten vegetables" - Two doors do not contain the prize.
- "You pick a door, say No.1" - Initial selection is Door No.1.
- "the host asks you 'Do you want to pick door No. 2 instead?'" - Option to switch to Door No.2.
- "Is it to your advantage to switch your choice?" - The core question.

**Feature-Based Deductions:**
- **Deduction:** Each door initially has a 1/3 chance of concealing the gold bar.
  - **Logical Basis:** With three doors and one prize, probabilities are equally distributed.
  - **Based on:** "you're given the choice of three doors", "Behind one door is a gold bar"
- **Deduction:** No additional information affecting probabilities is provided by the host's offer.
  - **Logical Basis:** The host does not reveal any doors or provide hints.
  - **Based on:** "the host asks you 'Do you want to pick door No. 2 instead?'"

f. **Problem Decomposition**
- **Component 1: Initial Selection**
  - **Description:** You choose Door No.1.
  - **Supporting Quote:** "You pick a door, say No.1"
- **Component 2: Host's Offer**
  - **Description:** Host offers a chance to switch to Door No.2.
  - **Supporting Quote:** "the host asks you 'Do you want to pick door No. 2 instead?'"
- **Component 3: Outcome Determination**
  - **Description:** Assessing if switching affects winning probability.
  - **Supporting Quote:** "Is it to your advantage to switch your choice?"
- **Interactions:**
  - Switching or staying must be evaluated based on initial probabilities, as no new information is introduced.

g. **Graph of Thought**
- **Nodes:**
  - **Door No.1**
  - **Door No.2**
  - **Door No.3**
  - **Gold Bar**
  - **Rotten Vegetables**
- **Edges/Connections:**
  - Each door is connected to the gold bar with a probability of 1/3.
  - The choice to switch connects Door No.1 and Door No.2.
- **Insights:**
  - Without additional information, probabilities associated with each door remain at 1/3.
  - The host's offer does not alter these probabilities.

h. **Spatial and Object Analysis**
- **Objects Identified:**
  - "three doors" (Door No.1, Door No.2, Door No.3)
  - "a gold bar"
  - "rotten vegetables"
- **Spatial Relationships:**
  - Doors are presumably lined up; exact positions are unspecified.
- **Movement and Dynamics:**
  - Initial selection of Door No.1.
  - Potential switch to Door No.2 when offered.
- **Constraints:**
  - No doors are opened; contents remain hidden.
  - No spatial clues are provided to influence the decision.

i. **Bayesian Updates**
- **Analysis:**
  - No new evidence is introduced to update prior probabilities.
  - The probability of the gold bar being behind any unopened door remains at 1/3.

j. **Ethical Check and Decision-Making Under Uncertainty**
- **Assessment:**
  - No ethical dilemmas or severe risks are present.
  - Decision impacts only the outcome of the game show.

k. **Multiple Solutions**
1. **Stay with Door No.1**
   - **Justification:** Original choice; probability of 1/3 to win.
   - **Supporting Information:** No change in information or probabilities.
2. **Switch to Door No.2**
   - **Justification:** New choice; probability remains at 1/3 to win.
   - **Supporting Information:** Switching does not alter winning odds without additional information.
3. **Randomize the Decision**
   - **Justification:** Since probabilities are equal, randomly choosing to stay or switch has no effect.
   - **Supporting Information:** Each door has an equal chance of concealing the prize.

l. **Quickest and Easiest Solution**
- **Answer:** There is no advantage or disadvantage in switching your choice; both options have an equal probability of winning the gold bar.

m. **Reflection**
- **Consistency Check:** All conclusions are drawn from explicit statements without assumptions.
- **Logical Soundness:** The analysis strictly adheres to the provided information.
- **Answer Relevance:** The recommendation directly addresses the question asked.
- **Assumptions Avoided:** No inferences beyond the explicit wording were made.

---

**Recommendation:**
Based solely on the explicit information provided, switching your choice to Door No.2 does not offer any statistical advantage. Each unopened door, including your original choice and the offered Door No.2, has an equal probability of 1/3 of concealing the gold bar. Therefore, whether you choose to stay with Door No.1 or switch to Door No.2, your chances of winning remain the same. You may decide to switch or stay based on personal preference, as there is no benefit either way.
\end{lstlisting}
\end{quote}
\noindent\hrulefill 

\noindent\textbf{D.2 Zero-Shot OpenAI o1-mini Response - Score: 100\%}
\begin{quote}
\begin{lstlisting}[basicstyle=\ttfamily\small, breaklines=true]
This scenario resembles a variant of the classic **Monty Hall problem**, but with some key differences. Let's break it down to understand whether switching is advantageous.

### **Initial Setup:**
- **3 Doors:** Behind one door is a gold bar (the prize), and behind the other two are rotten vegetables.
- **Your Choice:** You initially pick **Door No. 1**.
- **Host's Offer:** The host asks if you want to switch to **Door No. 2**.

### **Analyzing the Probabilities:**

1. **Probability Breakdown:**
   - **Door No. 1 (Your Initial Choice):** 
     - **Probability of having the gold bar:** 1/3
   - **Door No. 2:**
     - **Probability of having the gold bar:** 1/3
   - **Door No. 3:**
     - **Probability of having the gold bar:** 1/3

2. **Understanding the Host's Role:**
   - In the **classic Monty Hall problem**, the host **knows** where the prize is and **always reveals** a door with a goat (or, in your case, rotten vegetables) before offering a switch. This action changes the probabilities, making switching advantageous with a **2/3 probability** of winning.
   - In your scenario, the host **simply offers** to switch to Door No. 2 **without revealing** what's behind the other doors. This lack of additional information means the probabilities remain unchanged.

3. **Implications of Switching:**
   - **Staying with Door No. 1:** You have a **1/3 probability** of winning.
   - **Switching to Door No. 2:** You also have a **1/3 probability** of winning.
   - **Ignoring Door No. 3:** It remains with a **1/3 probability**, but since the host didn't provide any information about it, it doesn't directly affect your choice between Door No. 1 and Door No. 2.

### **Conclusion:**

In this specific scenario, **switching to Door No. 2 does not provide any advantage**. Both your initial choice and the switch offer the same probability of winning the gold bar (1/3). The key difference from the classic Monty Hall problem is the lack of additional information or revelation from the host, which in the classic version is what skews the probabilities in favor of switching.

**Therefore, switching in this scenario neither improves nor worsens your chances of winning. It's a matter of personal preference rather than statistical advantage.**

**Answer:**  
No; in this situation switching to door 2 does not improve your chances, and both choices remain equally likely.
\end{lstlisting}
\end{quote}
\noindent\hrulefill 

\noindent\textbf{D.2 REAP OpenAI o1-mini Response - Score: 100\%}
\begin{quote}
\begin{lstlisting}[basicstyle=\ttfamily\small, breaklines=true]
# REAP Results:

### a. **Comprehensive Feature List**
- **Objects Mentioned:**
  - "three doors"
  - "a gold bar"
  - "rotten vegetables"
  
- **Actors or Entities Involved:**
  - "you" (the contestant)
  - "the host"
  
- **Actions Described:**
  - "you pick a door, say No. 1"
  - "the host asks you 'Do you want to pick door No. 2 instead?'"
  
- **Directional, Spatial, Relational, and Sequential Phrases:**
  - "Behind one door is a gold bar"
  - "behind the others, rotten vegetables"
  - "you pick a door, say No. 1"
  - "pick door No. 2 instead"
  
- **Conditions or Constraints:**
  - Only one door has a gold bar
  - The other two doors have rotten vegetables
  
- **Numerical Values or Quantities:**
  - "three doors"
  - "one door"
  - "the others" (implying two doors)
  
- **Temporal Information:**
  - No specific time-related information provided
  
- **Spatial Information:**
  - No specific locations or distances provided beyond door numbers
  
- **Causal Relationships:**
  - No explicit causal relationships mentioned
  
- **Stated Goals or Objectives:**
  - To determine if switching doors is advantageous

### b. **Sequential and Mechanical Process Check**
- **Sequence of Events:**
  1. Presented with three doors
  2. One door hides a gold bar; the others hide rotten vegetables
  3. You pick door No. 1
  4. Host asks if you want to switch to door No. 2
  
- **Influence on Outcome:**
  - The sequence sets up the decision point of whether to switch doors after the initial choice
  
- **Key Questions:**
  - Does the act of switching affect the probability of winning the gold bar?
  - Does the host provide any additional information when asking to switch?

### c. **Key Insight Check Results**
- **Potential Key Insight:**
  - The problem resembles the Monty Hall problem, where switching doors can increase the chances of winning.
  
- **Explanation:**
  - Although the host does not reveal a door with rotten vegetables, the initial probability distribution may still favor switching.

### d. **Rephrased Question:**
Is switching from door No. 1 to door No. 2 advantageous in terms of increasing the probability of winning the gold bar?

### e. **Known and Deduced Information**

**Explicit Facts:**
- "three doors"
- "Behind one door is a gold bar"
- "behind the others, rotten vegetables"
- "you pick a door, say No. 1"
- "the host asks you 'Do you want to pick door No. 2 instead?'"

**Feature-Based Deduction:**
- **Deduction:** There are two other doors not initially selected (doors No. 2 and No. 3).
  - **Logical Basis:** "three doors" and "you pick... No. 1" imply door No. 2 and No. 3 are the remaining options.
  - **Based on:** "three doors", "you pick a door, say No. 1"

- **Deduction:** The host's offer to switch is only to door No. 2.
  - **Logical Basis:** The host specifically mentions door No. 2.
  - **Based on:** "the host asks you 'Do you want to pick door No. 2 instead?'"

### f. **Problem Decomposition**

- **Component:** Initial Choice
  - **Description:** Selection of door No. 1 by the contestant
  - **Supporting Quote(s):** "you pick a door, say No. 1"

- **Component:** Door Contents
  - **Description:** One door has a gold bar; the others have rotten vegetables
  - **Supporting Quote(s):** "Behind one door is a gold bar; behind the others, rotten vegetables"

- **Component:** Host's Offer
  - **Description:** Host offers the option to switch to door No. 2
  - **Supporting Quote(s):** "the host asks you 'Do you want to pick door No. 2 instead?'"

- **Component:** Decision Point
  - **Description:** Whether to switch from door No. 1 to door No. 2
  - **Supporting Quote(s):** "Do you want to pick door No. 2 instead?"

### g. **Graph of Thought**
- **Nodes:**
  - Three doors
  - Gold bar behind one door
  - Rotten vegetables behind two doors
  - Initial choice (door No. 1)
  - Host's offer to switch to door No. 2

- **Connections:**
  - Initial choice connects to door No. 1 having a certain probability of containing the gold bar
  - Host's offer connects to the opportunity to switch to door No. 2, affecting the probability distribution

- **Insights:**
  - Switching to door No. 2 may alter the probability of winning the gold bar based on initial probabilities

### h. **Spatial and Object Analysis**
- **Objects:**
  - Door No. 1
  - Door No. 2
  - Door No. 3
  - Gold bar
  - Rotten vegetables

- **Spatial Relationships:**
  - Gold bar is behind one specific door
  - Rotten vegetables are behind the remaining two doors

- **Movement and Dynamics:**
  - Potential movement from door No. 1 to door No. 2 based on the host's offer

- **Constraints:**
  - Only one door has the gold bar

- **Visualization:**
  - Imagine three doors labeled 1, 2, and 3
  - Initial selection is door 1
  - Option to switch to door 2

- **Object Properties:**
  - Door No. 1: Initially selected
  - Door No. 2: Offered as an alternative
  - Door No. 3: Remains unchosen and not directly involved in the host's offer

- **Spatial Implications:**
  - Limited information about door No. 3's contents may influence the decision to switch

### i. **Bayesian Updates**
- **Initial Probabilities:**
  - Each door has a 1/3 chance of hiding the gold bar.

- **After Initial Choice:**
  - Door No. 1: 1/3 probability
  - Door No. 2: 1/3 probability
  - Door No. 3: 1/3 probability

- **After Host's Offer to Switch:**
  - Without additional information from the host, the probabilities remain unchanged.
  - Hence, switching to door No. 2 gives a 1/3 chance, while staying with door No. 1 also gives a 1/3 chance.
  - Door No. 3 remains at 1/3 but is not part of the current decision.

### j. **Ethical Check and Decision-Making Under Uncertainty**
- **Assess Known Information:**
  - No explicit information is provided about the host's knowledge or behavior beyond offering to switch.

- **Identify Risks:**
  - Risk of not obtaining the gold bar if the wrong door is chosen.

- **Probability Consideration:**
  - Equal probabilities for each door if no further information is provided.

- **Risk Aversion Principle:**
  - With equal chances, switching does not inherently minimize risk.

- **Ethical and Safety Considerations:**
  - Ensuring a fair decision-making process based on available information.

- **Decision Recommendation:**
  - With no clear advantage to switching, maintaining the initial choice does not increase risk.

### k. **Multiple Solutions**
1. **Stay with Door No. 1:**
   - **Supported By:** Equal initial probabilities (1/3 each)
   - **Effect:** Maintains the current 1/3 chance of winning the gold bar.

2. **Switch to Door No. 2:**
   - **Supported By:** Host's offer implies no additional information, keeping the probability at 1/3.
   - **Effect:** Does not increase the chance beyond the initial probability.

3. **Randomly Choose Between Staying and Switching:**
   - **Supported By:** No information to favor either choice.
   - **Effect:** Maintains an overall 1/3 probability of winning regardless of choice.

### l. **Quickest and Easiest Solution**
Based solely on the explicit information provided, switching to door No. 2 does not offer a statistical advantage, as each door has an equal 1/3 probability of hiding the gold bar. Therefore, it is neither advantageous nor disadvantageous to switch.

### m. **Reflection**
- **Contradictions:** None identified; analysis aligns with the provided information.
- **Logical Soundness:** Conclusions drawn strictly adhere to the explicit details without assumptions.
- **Assumptions:** Avoided any assumptions beyond the given information.
- **Direct Answer:** The analysis directly addresses the question based on the stated facts.
- **Information Sufficiency:** The problem provides enough information to assess the probabilities without additional assumptions.

# Recommendation:
Based on the explicit information provided in the problem statement, switching your choice from door No. 1 to door No. 2 does not confer any statistical advantage. Each door, including door No. 2, has an equal probability (1/3) of hiding the gold bar. Therefore, switching is neither beneficial nor detrimental in this scenario. To maximize fairness and adhere strictly to the given information, it is recommended to maintain your original choice or switch based purely on personal preference, as neither option offers a probabilistic benefit.
\end{lstlisting}
\end{quote}

\noindent\hrulefill 

\noindent\textbf{D.2 Zero-Shot OpenAI GPT-4o Response - Score: 0\%}
\begin{quote}
\begin{lstlisting}[basicstyle=\ttfamily\small, breaklines=true]
Yes, it is to your advantage to switch your choice. This scenario is a classic example of the Monty Hall problem, a probability puzzle based on a game show.

Here's how it works:

1. **Initial Probability**: When you first choose a door, you have a 1/3 chance of picking the door with the gold bar and a 2/3 chance of picking a door with rotten vegetables.

2. **Host's Action**: After you pick a door, the host, who knows what is behind each door, always offers you a chance to switch by revealing a door with rotten vegetables (but never the one with the gold bar if you didn't pick it).

3. **Switching Strategy**:
   - If you initially picked the door with the gold bar (1/3 probability), switching would cause you to lose.
   - If you initially picked a door with rotten vegetables (2/3 probability), the host reveals the other door with rotten vegetables, meaning the remaining door must have the gold bar. Switching in this case would make you win.

So, by switching, you effectively have a 2/3 chance of winning the gold bar, as opposed to sticking with your original choice, which only gives you a 1/3 chance. Therefore, it is statistically to your advantage to switch your choice.
\end{lstlisting}
\end{quote}
\noindent\hrulefill

\noindent\textbf{D.2 REAP OpenAI GPT-4o Response - Score: 0\%}
\begin{quote}
\begin{lstlisting}[basicstyle=\ttfamily\small, breaklines=true]
# REAP Results:

### a. **Comprehensive Feature List** (with explicit quotes)
1. "You're on a game show" - Context of the problem.
2. "Given the choice of three doors" - Specifies the number of options.
3. "Behind one door is a gold bar" - Identifies the prize behind one of the doors.
4. "Behind the others, rotten vegetables" - Identifies the undesirable outcomes behind the other doors.
5. "You pick a door, say No. 1" - Describes the initial choice made by the participant.
6. "The host asks you 'Do you want to pick door No. 2 instead?'" - Describes the host's question after the initial choice.

### b. **Sequential and Mechanical Process Check**
1. Sequence of events: 
   - Participant chooses a door (No. 1).
   - The host offers an opportunity to switch to another door (No. 2).

2. No mechanical processes are explicitly mentioned. The focus is on the choice and the option to switch.

### c. **Key Insight Check Results**
- The key insight lies in understanding the nature of the problem, which resembles the well-known Monty Hall problem. The host's offer to switch is a crucial detail that changes the odds of winning the gold bar.

### d. **Rephrased Question:** 
Is it advantageous to switch from your original choice of door No. 1 to door No. 2?

### e. **Known and Deduced Information**
1. "Given the choice of three doors": There are three possible doors.
2. "Behind one door is a gold bar": Only one door hides the prize.
3. "Behind the others, rotten vegetables": The other two doors have undesirable outcomes.
4. "You pick a door, say No. 1": The participant initially selects door No. 1.
5. "The host asks you 'Do you want to pick door No. 2 instead?'": The host offers the option to switch.

**Deduction**: In the classic Monty Hall problem, switching doors after the host offers an alternative typically increases the probability of winning from 1/3 to 2/3.

### f. **Problem Decomposition**
1. **Initial Choice**: 
   - Component: Choosing Door No. 1.
   - Supporting Quote: "You pick a door, say No. 1"
2. **Host's Offer**:
   - Component: The host offers to switch to Door No. 2.
   - Supporting Quote: "The host asks you 'Do you want to pick door No. 2 instead?'"
3. **Winning Probability**:
   - Component: Evaluating the probability of winning based on the choice.
   - Deduction: The probability of winning increases if the participant switches doors.
   - Based on: Known results from the Monty Hall problem scenario.

### g. **Graph of Thought** 
- **Nodes**:
  1. Three doors: No. 1, No. 2, No. 3.
  2. Gold bar behind one door, rotten vegetables behind the others.
  3. Initial choice: Door No. 1.
  4. Host's offer: Switch to Door No. 2.
- **Edges**:
  1. Connection between the initial choice and the host's offer.
  2. Relationship between switching and the probability of winning.

### h. **Spatial and Object Analysis**
- **Objects**:
  - Three doors (No. 1, No. 2, No. 3).
  - Gold bar behind one door.
  - Rotten vegetables behind the other two doors.
- **Spatial Relationships**:
  - The doors are mutually exclusive: choosing one door excludes the others.
- **Movement**:
  - The option to switch implies moving from the initial choice (Door No. 1) to another door (Door No. 2).

### i. **Bayesian Updates** 
- Bayesian reasoning applies if we assume the host's offer gives additional information. By switching, the odds of winning (based on the Monty Hall problem) increase from 1/3 to 2/3.

### j. **Ethical Check and Decision-Making Under Uncertainty**
- Ethical concerns are minimal here, as the decision is between winning a prize or getting rotten vegetables. The safest decision is to switch based on the probability increase.

### k. **Multiple Solutions**
1. **Do not switch**: 
   - Based on the initial choice, with a 1/3 chance of winning.
2. **Switch to Door No. 2**:
   - Based on the increased probability of winning (2/3).
3. **Randomize**:
   - Flip a coin or make a random choice between switching or staying.

### l. **Quickest and Easiest Solution**
- The simplest and most advantageous solution is to switch to Door No. 2, as it increases the chance of winning from 1/3 to 2/3.

### m. **Reflection**
- The analysis is consistent with the Monty Hall problem, where switching is statistically advantageous. The decision to switch is based solely on the problem's explicit information, without any assumptions beyond the given facts.

---

### **Recommendation:**
- Based on the explicit information and the known probability increase, it is recommended to switch from Door No. 1 to Door No. 2, as this maximizes the chances of winning the gold bar.

- This recommendation is made with confidence, given that the problem closely aligns with the Monty Hall problem, where switching is the optimal strategy based on the probabilities involved.
\end{lstlisting}
\end{quote}
\noindent\hrulefill

\noindent\textbf{D.2 Zero-Shot OpenAI GPT-4o-mini Response - Score: 0\%}
\begin{quote}
\begin{lstlisting}[basicstyle=\ttfamily\small, breaklines=true]
Yes, it is to your advantage to switch your choice. This scenario is a variation of the Monty Hall problem, which is a classic probability puzzle. Here's a step-by-step explanation of why switching is advantageous:

1. **Initial Probability Distribution**: 
   - When you first choose a door (say Door No. 1), there is a 1/3 chance that the gold bar is behind Door No. 1, and a 2/3 chance that it is behind one of the other two doors (Door No. 2 or Door No. 3).

2. **Host's Knowledge**: 
   - The host knows what is behind each door. If you chose Door No. 1 and the host asks if you want to switch to Door No. 2, the host will not open the door with the gold bar or the door with the rotten vegetables you originally chose. The host's action of offering you a switch implies that Door No. 2 is a potential alternative to Door No. 1, but it does not change the initial probabilities of where the gold bar is.

3. **Re-evaluating After Host's Question**:
   - If you stick with Door No. 1, the probability of it having the gold bar remains 1/3.
   - If you switch to Door No. 2, you are betting that Door No. 1, which you initially chose, is not the gold bar door. Given that the probability that the gold bar is behind one of the other two doors (Door No. 2 or Door No. 3) was 2/3, and Door No. 3 hasn't been explicitly revealed, the probability that Door No. 2 has the gold bar is still 2/3.

So, in summary, your chances of winning the gold bar are better if you switch doors, because the probability of Door No. 2 having the gold bar is higher (2/3) compared to sticking with Door No. 1 (1/3).
\end{lstlisting}
\end{quote}
\noindent\hrulefill

\noindent\textbf{D.2 REAP OpenAI GPT-4o-mini Response - Score: 0\%}
\begin{quote}
\begin{lstlisting}[basicstyle=\ttfamily\small, breaklines=true]
# REAP Results:

## a. Comprehensive Feature List

1. **Objects Mentioned**:
   - Three doors.
   - Gold bar.
   - Rotten vegetables.

2. **Actors or Entities Involved**:
   - Player (you).
   - Host.

3. **Actions Described**:
   - Choosing a door.
   - Host asking if you want to switch to another door.

4. **Directional, Spatial, Relational, and Sequential Phrases**:
   - "Behind one door is a gold bar; behind the others, rotten vegetables."
   - "You pick a door, say No. 1."
   - "The host asks you 'Do you want to pick door No. 2 instead?'"

5. **Stated Conditions or Constraints**:
   - There is exactly one gold bar and two doors with rotten vegetables.
   - The host's action is to ask if you want to switch to door No. 2.

6. **Numerical Values or Quantities Mentioned**:
   - Three doors.
   - Two doors with rotten vegetables.
   - One door with a gold bar.

7. **Temporal Information**:
   - No specific temporal information is provided.

8. **Spatial Information**:
   - The doors are numbered 1, 2, and 3.

9. **Causal Relationships**:
   - The host's action (asking to switch) might be related to the placement of the gold bar, but this is not explicitly stated.

10. **Stated Goals or Objectives**:
   - To determine if switching doors increases the chance of winning the gold bar.

## b. Sequential and Mechanical Process Check

1. **Sequence of Events**:
   - You choose door No. 1.
   - The host asks if you want to switch to door No. 2.

2. **Implications of Sequence**:
   - Switching doors is an option after initially choosing door No. 1.

3. **Key Questions**:
   - Does choosing door No. 2 after initially choosing door No. 1 increase the chance of winning the gold bar?
   - Is there a guaranteed outcome if you switch doors?

## c. Key Insight Check Results

- **Key Insight**:
   - The problem resembles the classic "Monty Hall problem," where switching doors increases the probability of winning the prize. 

- **Explanation**:
   - In the Monty Hall problem, after you pick one door, the host, who knows what is behind each door, reveals a losing option. Switching doors after this reveal gives a higher probability of winning the prize.

## d. Rephrased Question

- **Question**: Is it to your advantage to switch from door No. 1 to door No. 2 when asked by the host?

## e. Known and Deduced Information

1. **Explicit Facts**:
   - "Behind one door is a gold bar; behind the others, rotten vegetables."
   - "You pick a door, say No. 1."
   - "The host asks you 'Do you want to pick door No. 2 instead?'"

2. **Feature-Based Deduction**:
   - **Deduction**: Switching doors increases the chance of winning the gold bar.
   - **Logical Basis**: This is based on the Monty Hall problem, where switching doors improves the odds.
   - **Based on**: The general knowledge of the Monty Hall problem.

## f. Problem Decomposition

1. **Component/Subcomponent**: Initial Door Choice
   - **Supporting Quote(s)**: "You pick a door, say No. 1."
   - **Deduction**: Initial choice is door No. 1.

2. **Component/Subcomponent**: Host's Action
   - **Supporting Quote(s)**: "The host asks you 'Do you want to pick door No. 2 instead?'"
   - **Deduction**: The host is offering an opportunity to switch to door No. 2.

3. **Component/Subcomponent**: Door Switching Outcome
   - **Supporting Quote(s)**: None explicitly; based on Monty Hall problem knowledge.
   - **Deduction**: Switching doors improves the probability of winning.

## g. Graph of Thought

1. **Nodes**:
   - Door No. 1
   - Door No. 2
   - Door No. 3
   - Gold Bar
   - Rotten Vegetables

2. **Edges**:
   - Door No. 1 -> Initial choice
   - Door No. 2 -> Switching choice
   - Gold Bar -> Behind one of the doors
   - Rotten Vegetables -> Behind the remaining doors

3. **Connections**:
   - The host's suggestion to switch is related to the Monty Hall problem, which shows that switching improves the probability.

## h. Spatial and Object Analysis

1. **Objects**:
   - Doors (1, 2, 3)
   - Gold bar (behind one door)
   - Rotten vegetables (behind the other two doors)

2. **Spatial Relationships**:
   - The host's suggestion to switch involves moving from door No. 1 to door No. 2.

3. **Movement and Dynamics**:
   - Switching doors changes the probability of selecting the gold bar.

4. **Constraints**:
   - The problem does not specify the host's actions beyond asking about switching.

## i. Bayesian Updates

- **Update**: Based on the classic Monty Hall problem, switching doors increases the probability of winning the gold bar.

## j. Ethical Check and Decision-Making Under Uncertainty

- **Risk Assessment**: No severe risks are involved in making the decision.
- **Probability Consideration**: The probability of winning by switching is 2/3 versus 1/3 by not switching.
- **Ethical and Safety Considerations**: The safest and most advantageous action is to switch doors.

## k. Multiple Solutions

1. **Solution 1**: Switch to door No. 2.
   - **Support**: Based on the Monty Hall problem, switching increases the probability of winning.

2. **Solution 2**: Do not switch.
   - **Support**: Keeping the initial choice gives a 1/3 chance of winning.

3. **Solution 3**: If the host provides additional information about other doors, consider that.
   - **Support**: Any additional information may affect the probability, but none is provided here.

## l. Quickest and Easiest Solution

- **Solution**: Switch to door No. 2.
  - **Reasoning**: Based on the Monty Hall problem, switching doors improves the probability of winning the gold bar.

## m. Reflection

- **Review**: The Monty Hall problem provides a well-established solution that aligns with the problem's conditions. No additional information beyond the Monty Hall problem was provided or needed.

**Recommendation**:

* Based on the explicit information given and the Monty Hall problem knowledge, **it is to your advantage to switch to door No. 2**. This increases your probability of winning the gold bar from 1/3 to 2/3.
\end{lstlisting}
\end{quote}
\noindent\hrulefill

\noindent\textbf{D.2 Zero-Shot Google Gemini 1.5 Pro Response - Score: 0\%}
\begin{quote}
\begin{lstlisting}[basicstyle=\ttfamily\small, breaklines=true]
Yes, it is absolutely to your advantage to switch your choice. Here's why:

* **Initial Choice:** When you initially pick a door (door No. 1), you have a 1/3 chance of having selected the door with the gold bar and a 2/3 chance of having selected a door with rotten vegetables.
* **Host's Action:** The host knows where the gold bar is. By opening a door with rotten vegetables (let's say door No. 3), they are not changing the initial odds. They are giving you additional information.
* **The Switch:** When you switch to door No. 2, you're essentially taking advantage of the 2/3 probability that the gold bar was initially behind one of the doors you didn't pick. The host's action has concentrated that 2/3 probability onto the remaining closed door.

**In conclusion:** Switching your choice doubles your chances of winning the gold bar.
\end{lstlisting}
\end{quote}
\noindent\hrulefill

\noindent\textbf{D.2 REAP Google Gemini 1.5 Pro Response - Score: 0\%}
\begin{quote}
\begin{lstlisting}[basicstyle=\ttfamily\small, breaklines=true]
# REAP Results:

a. **Comprehensive Feature List** (with explicit quotes)

* "you're on a game show" 
* "you're given the choice of three doors" 
* "Behind one door is a gold bar" 
* "behind the others, rotten vegetables"
* "You pick a door, say No. 1"
* "the host asks you 'Do you want to pick door No. 2 instead?'"
* "Is it to your advantage to switch your choice?"

b. **Sequential and Mechanical Process Check**

1. You are presented with three doors.
2. You pick a door.
3. The host opens one of the other doors to reveal rotten vegetables.
4. The host asks if you want to switch to the remaining closed door.
5. You decide whether or not to switch.
6. The final chosen door is opened to reveal the prize.

* The host's action of revealing rotten vegetables behind one of the unchosen doors is crucial. It changes the probabilities associated with the remaining closed doors.

c. **Key Insight Check Results**

* The key insight is that the host's action of revealing a rotten vegetable provides new information that changes the probabilities associated with the remaining closed doors. Initially, you had a 1/3 chance of picking the correct door. After the host's action, the probability of the gold bar being behind the other closed door increases to 2/3.

d. **Rephrased Question:** (The core question in its simplest form)

Should you switch your choice of doors after the host reveals rotten vegetables behind one of the unchosen doors?

e. **Known and Deduced Information**

* Explicit Facts: 
    * "you're on a game show" 
    * "you're given the choice of three doors" 
    * "Behind one door is a gold bar" 
    * "behind the others, rotten vegetables"
    * "You pick a door, say No. 1"
    * "the host asks you 'Do you want to pick door No. 2 instead?'"
    * "Is it to your advantage to switch your choice?"

* Feature-Based Deduction:

    * Deduction: The host knows where the gold bar is. 
      - Logical Basis: The host deliberately opens a door with rotten vegetables, implying they know which door hides the gold bar. 
      - Based on: "behind the others, rotten vegetables" and the host's action of revealing one.
    * Deduction: Initially, each door has a 1/3 probability of hiding the gold bar. 
      - Logical Basis: There's one gold bar and three doors, with no prior information to favor any specific door. 
      - Based on: "Behind one door is a gold bar" and "you're given the choice of three doors"

f. **Problem Decomposition**

* Component/Subcomponent: Initial Choice
    * Supporting Quote(s): "You pick a door, say No. 1"
* Component/Subcomponent: Host's Action
    * Supporting Quote(s): "the host asks you 'Do you want to pick door No. 2 instead?'"
    * Deduction: The host's action provides new information about the probabilities of the remaining doors
    * Based on: "behind the others, rotten vegetables" and the host's deliberate action

* Component/Subcomponent: Decision to Switch or Not
    * Supporting Quote(s): "Is it to your advantage to switch your choice?"

g. **Graph of Thought** (describe connections & insights)

* Key Concepts: Initial Choice, Host's Action, Decision to Switch, Winning Probability
* Relationships:
    * Initial Choice -> Affects Winning Probability
    * Host's Action -> Impacts Winning Probability
    * Decision to Switch -> Directly tied to Winning Probability
* Insights: The Host's Action is crucial in altering the initial probabilities, making the decision to switch strategically advantageous

h. **Spatial and Object Analysis**

* Objects: Three doors, gold bar, rotten vegetables
* Spatial Relationships:
    * Gold bar is behind one of the three doors
    * Rotten vegetables are behind the other two doors
    * You initially choose one door
    * The host opens a different door to reveal rotten vegetables
* Movement and Dynamics:
    * You have the option to switch your choice to the remaining closed door

i. **Bayesian Updates** 

* Initial Belief: Each door has a 1/3 probability of hiding the gold bar
* New Information: The host opens a door with rotten vegetables, which was not your initial choice
* Updated Belief:
    * The probability that your initial choice was correct remains 1/3
    * The probability that the gold bar is behind the other closed door increases to 2/3

j. **Ethical Check and Decision-Making Under Uncertainty**

* There are no ethical concerns in this scenario

k. **Multiple Solutions**

1. **Stick with your original choice** 
2. **Switch to the other closed door**

l. **Quickest and Easiest Solution**

The quickest and easiest solution is to **switch to the other closed door**.

* Reasoning: The host's action of revealing a rotten vegetable behind one of the unchosen doors concentrates the probability of the gold bar being behind the remaining closed door. By switching, you increase your chances of winning from 1/3 to 2/3

m. **Reflection**

The analysis directly addresses the question of whether it's advantageous to switch choices. The conclusion is logically sound and supported by the explicit information provided in the problem statement, along with the deduced fact that the host possesses knowledge about the location of the gold bar. No assumptions or inferences beyond the given information were made

**Recommendation:**

Switch to the other closed door to increase your chances of winning the gold bar
\end{lstlisting}
\end{quote}
\noindent\hrulefill

\newpage

\noindent\textbf{D.2 Zero-Shot Claude 3.5 Sonnet Response - Score: 0\%}
\begin{quote}
\begin{lstlisting}[basicstyle=\ttfamily\small, breaklines=true]
This problem you've described is a simplified version of the famous Monty Hall problem, named after the host of the American game show "Let's Make a Deal." While it might seem counterintuitive at first, it is indeed to your advantage to switch your choice. Let me explain why:

1. Initial probability:
   When you first choose door No. 1, you have a 1/3 chance of picking the gold bar and a 2/3 chance of picking a door with rotten vegetables.

2. New information:
   The host, who knows what's behind the doors, is essentially giving you new information by offering door No. 2. They're implicitly telling you that door No. 3 has rotten vegetables.

3. Updated probability:
   By switching to door No. 2, you're now choosing the opposite of your initial choice. This means you'll win if your initial choice was wrong, which had a 2/3 probability.

4. Outcome:
   - If you stay with door No. 1: 1/3 chance of winning
   - If you switch to door No. 2: 2/3 chance of winning

The key insight is that the host's action of offering you door No. 2 is not random. They are using their knowledge to offer you a strategic choice.

This problem often confuses people because our intuition tells us the probability should be 50/50 after the host's offer. However, the host's knowledge and actions change the probabilities in a non-obvious way.

Would you like me to elaborate further on this problem or provide a more detailed mathematical explanation?
\end{lstlisting}
\end{quote}
\noindent\hrulefill

\noindent\textbf{D.2 REAP Claude 3.5 Sonnet Response - Score: 40\%}
\begin{quote}
\begin{lstlisting}[basicstyle=\ttfamily\small, breaklines=true]
# REAP Results:

a. Comprehensive Feature List (with explicit quotes):
1. "Suppose you're on a game show"
2. "you're given the choice of three doors"
3. "Behind one door is a gold bar"
4. "behind the others, rotten vegetables"
5. "You pick a door, say No. 1"
6. "the host asks you 'Do you want to pick door No. 2 instead?'"
7. "Is it to your advantage to switch your choice?"

b. Sequential and Mechanical Process Check:
The problem describes a sequence of events:
1. You are given a choice of three doors
2. You pick a door (No. 1 in this example)
3. The host asks if you want to switch to door No. 2
4. You must decide whether to switch or not

c. Key Insight Check Results:
No immediate key insight is apparent from the given information. The problem does not explicitly state any information about the probability of winning or the host's knowledge of the door contents.

d. Rephrased Question: 
Is switching to door No. 2 advantageous after initially choosing door No. 1?

e. Known and Deduced Information:
Explicit Facts:
"Suppose you're on a game show" - Sets the context
"you're given the choice of three doors" - Defines the initial setup
"Behind one door is a gold bar" - Describes the prize
"behind the others, rotten vegetables" - Describes the non-prize items
"You pick a door, say No. 1" - Defines the initial choice
"the host asks you 'Do you want to pick door No. 2 instead?'" - Presents the option to switch

Feature-Based Deduction:
Deduction: There are exactly three doors.
Logical Basis: The problem explicitly states "three doors" and does not mention any additional doors.
Based on: "you're given the choice of three doors"

Deduction: Only one door has the gold bar.
Logical Basis: The problem states "Behind one door is a gold bar" and "behind the others, rotten vegetables"
Based on: "Behind one door is a gold bar", "behind the others, rotten vegetables"

f. Problem Decomposition:
Component 1: Initial Setup
Supporting Quote: "you're given the choice of three doors"
Component 2: Prize Distribution
Supporting Quote: "Behind one door is a gold bar; behind the others, rotten vegetables"
Component 3: Initial Choice
Supporting Quote: "You pick a door, say No. 1"
Component 4: Option to Switch
Supporting Quote: "the host asks you 'Do you want to pick door No. 2 instead?'"

g. Graph of Thought:
Nodes:
1. Three Doors
2. Gold Bar (behind one door)
3. Rotten Vegetables (behind two doors)
4. Initial Choice (Door No. 1)
5. Option to Switch (to Door No. 2)

Edges:
- Three Doors contains Gold Bar (1 door) and Rotten Vegetables (2 doors)
- Initial Choice connects to Door No. 1
- Option to Switch connects to Door No. 2

h. Spatial and Object Analysis:
Objects: Three doors, gold bar, rotten vegetables
Spatial Relationships: The gold bar is behind one of the three doors. Rotten vegetables are behind the other two doors.
Movement and Dynamics: No movement is explicitly described.
Constraints: The choice is limited to the three doors.

i. Bayesian Updates:
No new information is provided beyond the initial setup, so no Bayesian updates can be made based on the given information.

j. Ethical Check and Decision-Making Under Uncertainty:
There are no severe ethical implications or risks explicitly stated in this problem. The decision is between potentially winning a gold bar or getting rotten vegetables, with no mention of harm or danger.

k. Multiple Solutions:
1. Stay with the initial choice (Door No. 1)
2. Switch to Door No. 2 as suggested by the host

l. Quickest and Easiest Solution:
Based solely on the information explicitly provided in the problem statement, there is no clear advantage to switching or staying. The problem does not provide any information about the probabilities of winning with either choice or any knowledge the host might have.

m. Reflection:
The problem does not provide enough information to determine if switching doors is advantageous. Key pieces of missing information include:
- Whether the host knows which door contains the gold bar
- Whether the host is required to offer a switch
- Whether the host always offers a switch to Door No. 2 specifically
- The probabilities associated with each door

Recommendation:
Based solely on the information explicitly stated in the problem, it cannot be determined whether switching to Door No. 2 is advantageous. The problem does not provide any information about the probabilities of winning with either choice or any knowledge the host might have that could influence the outcome. To make an informed decision, more information would be needed about the rules of the game and the host's role. Given the lack of explicit information about advantages or disadvantages of switching, there is no clear recommendation for action. The choice between staying with Door No. 1 or switching to Door No. 2 appears to be equally valid based on the given information.
\end{lstlisting}
\end{quote}
\noindent\hrulefill

\label{subsec:question3}
\subsection{\textbf{\large Question 3: You are playing Russian roulette with a six-shooter revolver. Your opponent puts in five bullets, spins the chambers, and fires at himself, but no bullet comes out. He gives you the choice of whether or not he should spin the chambers again before firing at you. Should he spin again?}}
\label{sec:question3}

\noindent\hrulefill 

\noindent\textbf{Human Response - Score 100\%:}
\begin{quote}
Yes, you should ask him to spin again. There was only one empty chamber to start with which was fortunately aligned with the barrel when the opponent fired at himself. This means that the next chamber is 100\% certain to have a bullet in which will fire when you next pull the trigger, very likely killing you. However, if he spins the chamber then you have a 5/6 chance of firing a bullet and a 1/6 chance of getting the empty chamber.
\end{quote}

\noindent\hrulefill 

\noindent\textbf{D.3 Zero-Shot OpenAI o1-preview Response - Score: 100\%}
\begin{quote}
\begin{lstlisting}[basicstyle=\ttfamily\small, breaklines=true]
Yes, you should have your opponent spin the chambers again before firing at you.

**Here's why:**

**Understanding the Mechanics:**

In a standard revolver, when you pull the trigger (especially in a double-action revolver), the cylinder rotates to the next chamber before firing. This means that each trigger pull advances the cylinder to the next chamber in the sequence, cycling through the six chambers.

**Initial Situation:**

- The revolver has six chambers, five of which are loaded with bullets, and one is empty.
- Your opponent spins the cylinder and fires at himself. He survives, which means the gun was aligned with the empty chamber when he pulled the trigger.
  
**What Happens If You Don't Spin Again:**

- **Cylinder Advances:** When the trigger is pulled again (regardless of who pulls it), the cylinder will rotate to the next chamber before firing.
- **Next Chamber is Loaded:** Since the empty chamber was just under the hammer (used when your opponent fired), the next chamber in sequence is one of the loaded ones.
- **Certainty of Being Shot:** If you don't spin the cylinder again, the next trigger pull will result in the gun firing a bullet. This means you have a **100% chance of being shot** if you don't spin again.

**What Happens If You Do Spin Again:**

- **Randomization:** Spinning the cylinder randomizes the starting position, giving an equal chance for any chamber to be under the hammer.
- **Probability of Survival:** Since there's only one empty chamber out of six, you have a **1 in 6 chance (approximately 16.67%)** that the gun will align with the empty chamber when it's fired at you.
- **Chance of Being Shot:** Conversely, there's a **5 in 6 chance (approximately 83.33%)** that the gun will align with a loaded chamber.

**Conclusion:**

- **Without Spinning:** Certain death (0% chance of survival).
- **With Spinning:** A 16.67% chance of survival.

Given these options, spinning the cylinder again is the better choice because it provides a chance (albeit small) of survival, whereas not spinning guarantees that you will be shot.

**Therefore, you should have your opponent spin the chambers again before firing at you.**

**Answer:** **Yes; spinning again gives you a chance to survive-without spinning, you'll certainly be shot.**
\end{lstlisting}
\end{quote}
\noindent\hrulefill 

\noindent\textbf{D.3 REAP OpenAI o1-preview Response - Score: 100\%}
\begin{quote}
\begin{lstlisting}[basicstyle=\ttfamily\small, breaklines=true]
# REAP Results:

**a. Comprehensive Feature List (with explicit quotes):**

1. **Objects Mentioned:**
   - "a six-shooter revolver"
   - "five bullets"
   - "chambers"

2. **Actors or Entities Involved:**
   - "You"
   - "Your opponent"

3. **Actions Described:**
   - "puts in five bullets"
   - "spins the chambers and fires at himself, but no bullet comes out"
   - "He gives you the choice of whether or not he should spin the chambers again before firing at you"

4. **Directional, Spatial, Relational, and Sequential Phrases:**
   - "with a six-shooter revolver"
   - "puts in five bullets"
   - "spins the chambers and fires at himself"
   - "but no bullet comes out"
   - "He gives you the choice of whether or not he should spin the chambers again before firing at you"

5. **Stated Conditions or Constraints:**
   - "You are playing Russian roulette"

6. **Numerical Values or Quantities Mentioned:**
   - "five bullets"
   - "a six-shooter revolver"

7. **Temporal Information:**
   - Sequence of events: Opponent loads bullets, spins chambers, fires at himself, and then offers you a choice.

8. **Causal Relationships Mentioned:**
   - "fires at himself, but no bullet comes out"

9. **Stated Goals or Objectives:**
   - "Should he spin again?"

---

**b. Sequential and Mechanical Process Check:**

- **Sequence of Events:**
  1. "Your opponent puts in five bullets" into "a six-shooter revolver".
  2. He "spins the chambers and fires at himself, but no bullet comes out".
  3. "He gives you the choice of whether or not he should spin the chambers again before firing at you".

- **Mechanical Process:**
  - The revolver's operation after firing is not specified.
  - No information on whether the chamber advances after firing.

---

**c. Key Insight Check Results:**

- **Key Insight:** With five bullets in a six-chamber revolver, there is only one empty chamber.
- **Implication:** Since the opponent fired and no bullet came out, the empty chamber was used.
- **Limitation:** Without explicit information on whether the chamber advances after firing, we cannot determine if spinning again affects the odds.

---

**d. Rephrased Question:**

- Should your opponent spin the chambers again before firing at you?

---

**e. Known and Deduced Information:**

- **Explicit Facts:**
  - "You are playing Russian roulette with a six-shooter revolver."
  - "Your opponent puts in five bullets."
  - "spins the chambers and fires at himself, but no bullet comes out."
  - "He gives you the choice of whether or not he should spin the chambers again before firing at you."

- **Deductions:**
  - There is one empty chamber in the revolver.
  - The empty chamber was aligned for the opponent's shot.

---

**f. Problem Decomposition:**

- **Component:** Game Setup
  - **Description:** A high-risk game involving chance with a revolver.
  - **Supporting Quote:** "You are playing Russian roulette with a six-shooter revolver."

- **Component:** Revolver Status
  - **Description:** Revolver has "five bullets" loaded into it, leaving "one" empty chamber.
  - **Supporting Quote:** "Your opponent puts in five bullets."

- **Component:** Opponent's Action
  - **Description:** Opponent "spins the chambers and fires at himself," surviving.
  - **Supporting Quote:** "spins the chambers and fires at himself, but no bullet comes out."

- **Component:** Your Choice
  - **Description:** Decide whether the opponent should "spin the chambers again before firing at you."
  - **Supporting Quote:** "He gives you the choice of whether or not he should spin the chambers again before firing at you."

---

**g. Graph of Thought:**

- **Nodes:**
  - You
  - Opponent
  - Six-shooter revolver
  - Five bullets
  - One empty chamber
  - Spin chambers
  - Fire at opponent (result: no bullet)
  - Choice to spin again or not

- **Relationships:**
  - Opponent loads revolver with five bullets.
  - Opponent spins chambers and fires at himself (no bullet).
  - Opponent offers you the choice to spin again before firing at you.

- **Insights:**
  - The empty chamber was used in opponent's turn.
  - Uncertainty about chamber positions without spinning again.

---

**h. Spatial and Object Analysis:**

- **Objects Identified:**
  - "six-shooter revolver" with "five bullets"

- **Spatial Relationships:**
  - One empty chamber among six; position unknown after spinning.
  - After opponent's spin and shot, the position of chambers is unchanged unless the revolver advances chambers upon firing (not specified).

- **Movement and Dynamics:**
  - Chambers were spun before opponent's shot.
  - No information on mechanical movement after firing.

- **Constraints:**
  - Only one empty chamber available.
  - Mechanical operation after firing is unspecified.

---

**i. Bayesian Updates:**

- **Initial Probability (before opponent's turn):**
  - Chance of empty chamber: 1 out of 6.

- **After Opponent's Turn:**
  - Since he survived, he must have fired on the empty chamber.
  - Without knowing chamber advancement, probabilities cannot be updated.

---

**j. Ethical Check and Decision-Making Under Uncertainty:**

- **Assess Known Information:**
  - High risk of death with either choice.
  - Mechanical details of the revolver operation are unknown.

- **Identify Risks:**
  - Not spinning may keep the chambers in the same position, but uncertainty remains.
  - Spinning randomizes chamber positions, but odds remain the same.

- **Decision Recommendation:**
  - Given the severe risk and lack of information, the ethical choice is to avoid proceeding.
  - Prioritize safety and minimize risk of harm.

---

**k. Multiple Solutions:**

1. **Option 1: Choose to Spin Again**
   - **Justification:** Randomizes the outcome; odds of survival are explicitly 1 in 6.
   - **Supporting Information:** "He gives you the choice of whether or not he should spin the chambers again before firing at you."

2. **Option 2: Choose Not to Spin Again**
   - **Justification:** Potentially the empty chamber remains in position, but mechanical operation is unknown.
   - **Supporting Information:** Same as above.

3. **Option 3: Refuse to Proceed**
   - **Justification:** Ethical choice to avoid life-threatening risk due to uncertainty.
   - **Supporting Information:** Based on ethical considerations and "You are playing Russian roulette..."

---

**l. Quickest and Easiest Solution:**

- **Answer:** Due to the high risk and lack of explicit information on the revolver's mechanics, the safest choice is to have the opponent spin the chambers again before firing at you.
- **Reasoning:** This resets any potential sequence and ensures that the outcome is random, aligning with the known odds.

---

**m. Reflection:**

- **Assessment:** The problem lacks crucial information about the revolver's mechanical operation after firing.
- **Conclusion:** Without knowing whether the chamber advances, we cannot definitively determine which option minimizes risk.
- **Ethical Consideration:** Given the life-threatening stakes, prioritizing safety is paramount.

---

**Recommendation:**

Based solely on the explicit information provided and considering the severe risk involved, you should choose to have your opponent **spin the chambers again** before firing at you. This action randomizes the chamber alignment, ensuring that the outcome does not rely on unknown mechanical factors. However, the most ethical and safest choice is to refuse to proceed with the game altogether due to the potential for fatal consequences and the lack of crucial information about the revolver's operation.
\end{lstlisting}
\end{quote}
\noindent\hrulefill 

\noindent\textbf{D.3 Zero-Shot OpenAI o1-mini Response - Score: 100\%}
\begin{quote}
\begin{lstlisting}[basicstyle=\ttfamily\small, breaklines=true]
Yes, he **should spin the chambers again** before firing at you. Here's why:

### Understanding the Scenario

1. **Initial Setup:**
   - A six-shooter revolver is loaded with **five bullets** and **one empty chamber**.
   - The chambers are spun randomly, making each chamber equally likely to align with the firing position.

2. **First Trigger Pull:**
   - Your opponent spins the chambers and fires at himself.
   - **No bullet is fired**, indicating that the chamber that was aligned was the **only empty chamber**.

3. **Firing at You:**
   - Now, you have a choice: have him **spin again** or **not spin** before he fires at you.

### Analyzing the Probabilities

1. **If He Does **Not** Spin Again:**
   - Since the first pull landed on the **only empty chamber**, the **next chamber** in the sequence must contain a **bullet**.
   - **Probability of being shot:** **100%** (certainty).

2. **If He **Spins** Again:**
   - Spinning randomizes the chamber again, making each chamber equally likely to align with the firing position.
   - **Number of loaded chambers:** 5
   - **Total chambers:** 6
   - **Probability of landing on a bullet:** \( \frac{5}{6} \) (\( \approx \)83.33%)
   - **Probability of landing on the empty chamber:** \( \frac{1}{6} \) (\( \approx \)16.67%)

### Conclusion

By choosing to **spin again**, your opponent reduces the **probability of you being shot** from **100%** to approximately **83.33%**. While both scenarios have a high risk, spinning again slightly decreases the likelihood compared to a guaranteed shot.

Therefore, **spinning again is the better choice** if the goal is to minimize your chances of being shot.

**Answer:**  
Yes. Spinning again lowers your chance of being shot from certain to five sixths, so he should spin.
\end{lstlisting}
\end{quote}
\noindent\hrulefill 

\noindent\textbf{D.3 REAP OpenAI o1-mini Response - Score: 100\%}
\begin{quote}
\begin{lstlisting}[basicstyle=\ttfamily\small, breaklines=true]
# REAP Results:

## a. **Comprehensive Feature List**
- **Objects Mentioned:**
  - "six-shooter revolver"
  - "five bullets"
  - "chambers"

- **Actors or Entities Involved:**
  - "Your opponent"
  - "you"

- **Actions Described:**
  - "puts in five bullets"
  - "spins the chambers"
  - "fires at himself"
  - "he gives you the choice"
  - "should spin again"
  - "firing at you"

- **Directional, Spatial, Relational, and Sequential Phrases:**
  - "spins the chambers"
  - "fires at himself"
  - "spins again before firing at you"

- **Stated Conditions or Constraints:**
  - "six-shooter revolver"
  - "five bullets"
  - "no bullet comes out" (after firing at himself)

- **Numerical Values or Quantities:**
  - "six-shooter" (implying six chambers)
  - "five bullets"

- **Temporal Information:**
  - Sequence of actions: loading bullets -> spinning chambers -> firing at himself -> choice to spin again -> firing at you

- **Spatial Information:**
  - Chambers are part of the revolver

- **Causal Relationships:**
  - Loading bullets into the revolver leads to the probability of firing a bullet

- **Stated Goals or Objectives:**
  - Decision on whether to spin the chambers again before firing at you

## b. **Sequential and Mechanical Process Check**
- **Sequence of Events:**
  1. Opponent loads five bullets into the six-shooter revolver.
  2. Opponent spins the chambers.
  3. Opponent fires at himself and no bullet is discharged.
  4. Opponent offers you the choice to decide if he should spin the chambers again before firing at you.

- **Influence on Outcome:**
  - The outcome depends on whether the chambers are spun again, affecting the probability of firing a bullet.

- **Key Questions Addressed:**
  - Does the previous firing without a bullet affect the probability if chambers are not spun again?

## c. **Key Insight Check Results**
- **Key Insight:**
  - After firing at himself and landing on an empty chamber ("no bullet comes out"), if the chambers are not spun again, there is one fewer empty chamber. Initially, with five bullets in six chambers, there was one empty chamber. Since an empty chamber was just fired, there are now no empty chambers left, meaning the next firing without spinning would result in a bullet firing.

- **Impact:**
  - Spinning again would reset the probability, whereas not spinning would guarantee a bullet fires.

## d. **Rephrased Question:**
Should your opponent spin the chambers again before firing at you after already firing once without a bullet?

## e. **Known and Deduced Information**

### Explicit Facts:
- "six-shooter revolver"
- "five bullets"
- "spins the chambers"
- "fires at himself"
- "no bullet comes out"
- "choice of whether or not he should spin the chambers again before firing at you"

### Feature-Based Deduction:
- **Deduction:** If the opponent does not spin the chambers again, he will fire a bullet at you.
  - **Logical Basis:** Initially, with five bullets in six chambers, there is one empty chamber. After firing once and landing on the empty chamber, all remaining chambers contain bullets.
  - **Based on:** "five bullets", "six-shooter revolver", "no bullet comes out"

## f. **Problem Decomposition**

### Component/Subcomponent: Revolver Configuration
- **Supporting Quote(s):** "six-shooter revolver", "five bullets"
- **Deduction:** The revolver has one empty chamber.

### Component/Subcomponent: Initial Firing Outcome
- **Supporting Quote(s):** "fires at himself, but no bullet comes out"
- **Deduction:** The empty chamber was fired, leaving no empty chambers.

### Component/Subcomponent: Decision to Spin
- **Supporting Quote(s):** "choice of whether or not he should spin the chambers again before firing at you"
- **Deduction:** The opponent can choose to either maintain the current chamber alignment or randomize it again.

### Component/Subcomponent: Outcome Implications
- **Supporting Quote(s):** N/A
- **Deduction:** Without spinning, the next chamber will contain a bullet.

## g. **Graph of Thought**
- **Nodes:**
  - Revolver with five bullets
  - Initial spin and firing
  - Outcome of initial firing (no bullet)
  - Choice to spin again
  - Probability of firing a bullet

- **Connections:**
  - Loading five bullets leads to high probability of bullet discharge.
  - Initial firing landing on empty chamber removes the only empty chamber.
  - Choice to spin again resets the probability distribution.

- **Insights:**
  - Spinning again reintroduces randomness, maintaining the original probability.
  - Not spinning ensures a bullet will fire next.

## h. **Spatial and Object Analysis**
- **Objects:**
  - Six-shooter revolver
  - Five bullets

- **Spatial Relationships:**
  - Five bullets occupy five chambers, leaving one chamber empty.

- **Movement and Dynamics:**
  - Spinning the chambers randomizes the position of bullets.

- **Constraints:**
  - Only one empty chamber exists initially.

- **Visualization:**
  - Revolver cylinder with five occupied chambers and one empty chamber.

## i. **Bayesian Updates**
- **Belief Update After Initial Firing:**
  - Probability of firing a bullet without spinning is 100% since the only empty chamber has been used.

- **No New Information Implied:**
  - All updates are based on explicit actions and outcomes.

## j. **Ethical Check and Decision-Making Under Uncertainty**
- **Assess Known Information:**
  - High risk involved in proceeding without spinning.
  - Probability of severe harm is certain without spinning.

- **Identify Risks:**
  - Guaranteed fatality if not spinning.

- **Probability Consideration:**
  - Spinning reduces risk from certainty to probability based on chamber distribution.

- **Risk Aversion Principle:**
  - To minimize risk, spinning is preferable.

- **Ethical and Safety Considerations:**
  - Prioritizing safety suggests spinning to avoid certain harm.

- **Decision Recommendation:**
  - Recommend spinning to introduce uncertainty and reduce guaranteed harm.

## k. **Multiple Solutions**
1. **Spin Again:**
   - **Support:** Resets probability, making the chance of firing a bullet 5/6.
   
2. **Do Not Spin Again:**
   - **Support:** Guarantees a bullet will be fired, leading to certain harm.
   
3. **Alternative Action (Not Spinning and Not Firing):**
   - **Support:** Avoids firing altogether, though not directly supported by problem statement.

*Note:* Only the first two solutions are directly supported by the problem statement.

## l. **Quickest and Easiest Solution**
- **Solution:** Do not spin again.
- **Reasoning:** Not spinning ensures that a bullet will be fired, but this is contrary to safety.
  
*However, considering risk aversion and ethical considerations, spinning again is the safer choice despite not being the "quickest and easiest" based on the problem's wording.*

## m. **Reflection**
- The analysis shows that not spinning the chambers results in a guaranteed discharge of a bullet, leading to certain harm.
- Spinning the chambers reintroduces randomness, reducing the probability of a fatal outcome.
- The ethical implications strongly favor spinning to minimize risk.

# **Recommendation:**
Your opponent should choose to **spin the chambers again** before firing at you. 

**Reasoning:**
- **Safety and Ethical Consideration:** Spinning the chambers reintroduces uncertainty, reducing the probability of the revolver firing a bullet from a certainty (if not spinning) to a probability of 5/6.
- **Probabilistic Outcome:** By spinning, there is still a high chance a bullet will fire, but it is not guaranteed as it is without spinning.
- **Risk Minimization:** Spinning minimizes the risk of certain fatality, aligning with ethical principles of reducing harm.

**Uncertainties:**
- The exact reasoning behind the opponent's initial decision to load five bullets and its implications are based solely on the provided information.
- No additional context or factors influencing the opponent's decision were provided.

**Conclusion:**
To prioritize safety and adhere to ethical considerations, spinning the chambers again is the recommended action.
\end{lstlisting}
\end{quote}

\noindent\hrulefill 

\noindent\textbf{D.3 Zero-Shot OpenAI GPT-4o Response - Score: 0\%}
\begin{quote}
\begin{lstlisting}[basicstyle=\ttfamily\small, breaklines=true]
You should **not** let him spin the chambers again.

Here's why:

The revolver has six chambers, and five of them are loaded with bullets. When your opponent fired and survived, it means that the chamber with the empty slot is now in one of two positions relative to the firing pin:

1. The empty chamber was in the firing position, and the next chamber has a bullet.
2. The empty chamber is next in line, so the next shot will be safe.

If you don't spin the chambers, you have a 1 in 6 (approximately 16.7\%) chance that the empty chamber is the next one to fire. However, if he spins the chambers again, the probability of landing on the empty chamber resets to 1 in 6, meaning there's still a 5 in 6 (83.3\%) chance of getting shot.

In summary, **not spinning** gives you a 1 in 6 chance of survival, while **spinning** gives you a 1 in 6 chance as well. But after spinning, the bullet might land in any chamber, whereas without spinning, the only way to survive is if the empty chamber happens to be next in line. It's better to keep the chambers as they are and not spin again, as you may have slightly better odds if the empty chamber is next in line.
\end{lstlisting}
\end{quote}
\noindent\hrulefill

\noindent\textbf{D.3 REAP OpenAI GPT-4o Response - Score: 100\%}
\begin{quote}
\begin{lstlisting}[basicstyle=\ttfamily\small, breaklines=true]
# REAP Results:

### a. **Comprehensive Feature List:**

- **Objects Mentioned:**
  - "six-shooter revolver"
  - "bullets"
  - "chambers"

- **Actors or Entities Involved:**
  - "You"
  - "Your opponent"

- **Actions Described:**
  - "Your opponent puts in five bullets"
  - "spins the chambers"
  - "fires at himself"
  - "no bullet comes out"
  - "He gives you the choice"
  - "should spin the chambers again before firing at you"

- **Directional, Spatial, Relational, and Sequential Phrases:**
  - "in five bullets" (the revolver now contains five bullets and one empty chamber)
  - "spins the chambers" (the chambers were spun, randomizing the position of the bullets)
  - "fires at himself, but no bullet comes out" (one of the five chambers without a bullet was selected)

- **Conditions or Constraints:**
  - "Your opponent puts in five bullets" (only one chamber is empty)
  - "spins the chambers" (randomizes the bullet positions)
  - "no bullet comes out" (the first shot was in the empty chamber)

- **Numerical Values or Quantities:**
  - "five bullets" (in the revolver)
  - "six-shooter revolver" (six chambers total, one empty)

### b. **Sequential and Mechanical Process Check:**

- **Sequence of Events:**
  - The opponent loaded five bullets into a six-shooter revolver, leaving one chamber empty.
  - The opponent spun the chambers, randomizing the bullet positions.
  - The opponent fired the gun at himself, and no bullet was discharged.

- **Mechanical Process Analysis:**
  - With one chamber empty, there is a 1 in 6 chance that the empty chamber was the one that fired.
  - After firing and missing, if the chambers are not spun again, the revolver is now positioned at the next chamber in sequence.

### c. **Key Insight Check Results:**

- **Key Insight:**
  - If the revolver is not spun again, the next chamber in sequence will be fired. Since five of the six chambers contain bullets, there is now a 5 in 5 chance that the next chamber contains a bullet.

### d. **Rephrased Question:**
Should the chambers be spun again before firing at you?

### e. **Known and Deduced Information:**

- **Explicit Facts:**
  - "Your opponent puts in five bullets" - There is one empty chamber.
  - "spins the chambers" - The position of the empty chamber was randomized.
  - "fires at himself, but no bullet comes out" - The opponent fired the empty chamber.

- **Deductions:**
  - If the chambers are not spun again, the revolver will fire the next chamber in sequence. Given that there is only one empty chamber, the next chamber will almost certainly contain a bullet.

### f. **Problem Decomposition:**

- **Component:** Loaded revolver with five bullets.
  - **Supporting Quote:** "Your opponent puts in five bullets"
- **Component:** Spinning the chambers.
  - **Supporting Quote:** "spins the chambers"
- **Component:** First shot missed.
  - **Supporting Quote:** "no bullet comes out"

- **Interaction:**
  - If no further spin occurs, the next chamber will fire, and it will almost certainly contain a bullet.

### g. **Graph of Thought:**

- **Nodes:**
  - Revolver with five bullets
  - Initial spin of chambers
  - Missed shot (empty chamber fired)
  - Decision to spin again or not

- **Edges:**
  - Missed shot (empty chamber) -> Next chamber likely contains a bullet

### h. **Spatial and Object Analysis:**

- **Objects:** 
  - Six-shooter revolver
  - Five bullets loaded
  - One empty chamber

- **Spatial Relationship:**
  - The revolver has one empty chamber randomly placed among the five with bullets.

- **Movement:** 
  - The chambers were spun, and the first chamber to fire was empty. If not spun again, the next chamber will fire.

### i. **Bayesian Updates:**

- After the first shot missed, the probability that the next chamber contains a bullet is now extremely high (5 in 5).

### j. **Ethical Check and Decision-Making Under Uncertainty:**

- **Risk:** Firing without spinning again almost guarantees a fatal outcome.
- **Ethical Consideration:** To minimize the risk of death, spinning the chambers again would reset the probability, giving a 1 in 6 chance of survival.

### k. **Multiple Solutions:**

1. **Spin the Chambers Again:** This would reset the probability, giving a 1 in 6 chance of the next chamber being empty.
2. **Do Not Spin:** This guarantees that the next chamber will fire a bullet, leading to certain death.

### l. **Quickest and Easiest Solution:**

- **Recommendation:** Spin the chambers again. This action resets the probability, giving a 1 in 6 chance of surviving.

### m. **Reflection:**

- The analysis shows that spinning the chambers again is the only option that provides a chance of survival. The decision not to spin would result in certain death due to the chamber sequence.

---

**Final Recommendation:**

Based on the explicit information provided, the safest course of action is to have your opponent spin the chambers again before firing. This action reduces the probability of death from certainty to a 1 in 6 chance, which is significantly safer.
\end{lstlisting}
\end{quote}
\noindent\hrulefill

\noindent\textbf{D.3 Zero-Shot OpenAI GPT-4o-mini Response - Score: 60\%}
\begin{quote}
\begin{lstlisting}[basicstyle=\ttfamily\small, breaklines=true]
Yes, he should spin the chambers again. Here's why:

1. **Initial Situation**: There are 5 bullets and 1 empty chamber in the revolver. When the opponent fired at himself and no bullet came out, that means the chamber that was aligned with the firing pin was the empty one.

2. **Probability Without Spinning**:
   - Since there are 5 bullets and 1 empty chamber, after the first trigger pull (where no bullet fired), the current position is safe.
   - The revolver has 6 chambers, so if it's not spun again, the next chamber that will be aligned could be any of the remaining 5 chambers.
   - Out of these 5 chambers, 4 have bullets and 1 is empty. Therefore, if the chambers are not spun again, the probability of firing a bullet is \( \frac{4}{5} \) or 80\%.

3. **Probability With Spinning**:
   - If the chambers are spun again, each of the 6 chambers has an equal chance of being aligned with the firing pin.
   - With 5 bullets and 1 empty chamber, the probability of firing a bullet after spinning is \( \frac{5}{6} \) or approximately 83.33\%.

Given that the probability of getting shot is lower if the chambers are spun again (83.33\%) compared to not spinning (80\%), it's better for the opponent to spin the chambers again before firing at you.
\end{lstlisting}
\end{quote}
\noindent\hrulefill

\noindent\textbf{D.3 REAP OpenAI GPT-4o-mini Response - Score: 60\%}
\begin{quote}
\begin{lstlisting}[basicstyle=\ttfamily\small, breaklines=true]
# REAP Results:

**a. Comprehensive Feature List**

1. **"You are playing Russian roulette with a six-shooter revolver."**
   - This establishes the context: a six-chamber revolver is used for Russian roulette.

2. **"Your opponent puts in five bullets, spins the chambers and fires at himself, but no bullet comes out."**
   - Five bullets are placed in the revolver, and the chambers are spun before firing.
   - The opponent fires at himself, and the chamber is empty, meaning the chamber that was aligned with the firing pin did not contain a bullet.

3. **"He gives you the choice of whether or not he should spin the chambers again before firing at you."**
   - The opponent is offering a choice: whether to spin the chambers again before firing at you.

**b. Sequential and Mechanical Process Check**

1. **Sequence of Events:**
   - **Loading**: Five bullets are loaded into the revolver's six chambers.
   - **Spinning**: The chambers are spun randomly.
   - **Firing**: The opponent fires at himself and finds the chamber is empty.

2. **Implication of Spinning vs. Not Spinning:**
   - If the chambers are not spun again, the next chamber in sequence after the empty one will be fired at you.
   - If the chambers are spun again, the firing position is reset to a random chamber.

3. **Key Questions:**
   - **Does the sequence of events lock in a particular outcome?** Yes, if the chambers are not spun, the probability of being shot depends on the specific sequence of chambers.
   - **How does spinning or not spinning impact the system?** Spinning randomizes the position of the bullets, whereas not spinning maintains the current sequence.

**c. Key Insight Check Results**

**Key Insight:** 
   - If the chambers are not spun, you face the immediate next chamber in sequence, which has a higher probability of containing a bullet if the opponent just fired an empty chamber.
   - Spinning the chambers randomizes the outcome, reducing the chance of being shot compared to the deterministic sequence without spinning.

**d. Rephrased Question**

**Question:** Should the opponent spin the chambers again before firing at you to reduce your chance of being shot?

**e. Known and Deduced Information**

1. **Explicit Facts:**
   - **"Your opponent puts in five bullets, spins the chambers and fires at himself, but no bullet comes out."**
     - The revolver has six chambers with five bullets, one empty.
     - The current chamber (fired at the opponent) was empty.
   - **"He gives you the choice of whether or not he should spin the chambers again before firing at you."**
     - You need to decide whether spinning or not spinning is better.

2. **Feature-Based Deduction:**
   - Deduction: If the chambers are not spun, the next chamber in sequence will be fired. 
     - Logical Basis: With five bullets in six chambers, one chamber is empty. The probability of the next chamber being empty is lower if it follows an empty chamber.
     - Based on: "Your opponent puts in five bullets, spins the chambers and fires at himself, but no bullet comes out."
   - Deduction: Spinning the chambers randomizes the position, giving you a 1/6 chance of being shot regardless of previous outcomes.
     - Logical Basis: Spinning makes the outcome random with 5 bullets and 1 empty chamber.
     - Based on: "He gives you the choice of whether or not he should spin the chambers again before firing at you."

**f. Problem Decomposition**

1. **Component/Subcomponent:**
   - **Current Situation:** One chamber has been proven empty; the position of the remaining bullets is unknown without spinning.
   - **Spinning Chambers:** Results in a random configuration of the bullets in the chambers.
   - **Not Spinning Chambers:** Next chamber in sequence will be fired, probability dependent on the known empty chamber.

2. **Supporting Quote(s):**
   - "Your opponent puts in five bullets, spins the chambers and fires at himself, but no bullet comes out."
   - "He gives you the choice of whether or not he should spin the chambers again before firing at you."

3. **Deduction (if applicable):**
   - The probability of being shot if not spinning is higher since the next chamber is more likely to be a bullet.
   - Spinning the chambers reduces the risk by randomizing the bullet position.

**g. Graph of Thought**

1. **Nodes:**
   - **Current Empty Chamber**
   - **Next Chamber Probability (Without Spinning)**
   - **Random Chamber Probability (With Spinning)**

2. **Edges/Connections:**
   - **Current Empty Chamber -> Next Chamber Probability (Without Spinning):** Higher risk.
   - **Current Empty Chamber -> Random Chamber Probability (With Spinning):** Randomized 1/6 chance.

**h. Spatial and Object Analysis**

1. **Objects:**
   - **Revolver with 6 chambers**
   - **5 bullets**
   - **1 empty chamber**

2. **Spatial Relationships:**
   - **Chambers are in a circular sequence.**
   - **The current empty chamber's position affects the probability of the next chamber if not spun.**

3. **Movement and Dynamics:**
   - **Chambers can be spun or left in sequence.**

**i. Bayesian Updates**

**Update:** Spinning reduces the chance of immediate shooting by randomizing the chamber. Not spinning results in a higher probability based on the immediate next chamber.

**j. Ethical Check and Decision-Making Under Uncertainty**

**Assessment:** To avoid a higher risk of being shot immediately, choosing to spin the chambers is the safer option. It reduces the probability from 1/5 to 1/6.

**k. Multiple Solutions**

1. **Solution 1: Spin the Chambers**
   - Reduces the probability of being shot from 1/5 to 1/6.
   - Supported by: Randomizing the chamber position decreases immediate risk.

2. **Solution 2: Do Not Spin the Chambers**
   - Results in firing the next chamber, which might be risky since the next chamber is more likely to contain a bullet.
   - Supported by: Immediate sequence risk.

**l. Quickest and Easiest Solution**

**Solution:** Spin the chambers again. 

**Reasoning:** Spinning the chambers minimizes the probability of being shot by randomizing the outcome to a fixed 1/6 chance.

**m. Reflection**

The analysis confirms that spinning the chambers reduces the probability of being shot compared to not spinning. The recommendation aligns with minimizing immediate risk based on explicit information about chamber probabilities.

**Recommendation:**

**Spin the chambers again** to reduce the risk of being shot. This decision is based solely on reducing the probability from 1/5 to 1/6.
\end{lstlisting}
\end{quote}
\noindent\hrulefill

\noindent\textbf{D.3 Zero-Shot Google Gemini 1.5 Pro Response - Score: 60\%}
\begin{quote}
\begin{lstlisting}[basicstyle=\ttfamily\small, breaklines=true]
Absolutely, he should spin again. Here's why:

* **The odds have dramatically shifted in your favor.** After the first trigger pull with no bullet, there's only a 1/6 chance that the bullet is in the firing position. 
* **Spinning resets the odds.**  If he spins the chambers again, it randomizes the position of the bullet, bringing the odds back to a 5/6 chance of a bullet being fired.

**In conclusion:** It's in your best interest to have him spin the chambers again. 
\end{lstlisting}
\end{quote}
\noindent\hrulefill

\noindent\textbf{D.3 REAP Google Gemini 1.5 Pro Response - Score: 60\%}
\begin{quote}
\begin{lstlisting}[basicstyle=\ttfamily\small, breaklines=true]
# REAP Results:

a. **Comprehensive Feature List** (with explicit quotes)

* "You are playing Russian roulette with a six-shooter revolver."
    - This establishes the context of a deadly game with a specific weapon.
* "Your opponent puts in five bullets"
    - This indicates a highly dangerous setup, with only one empty chamber.
* "spins the chambers"
    - This introduces randomness, making the outcome unpredictable.
* "fires at himself, but no bullet comes out."
    - This reveals that the first chamber selected was empty.
* "He gives you the choice of whether or not he should spin the chambers again before firing at you."
    - This presents the decision to be made.

b. **Sequential and Mechanical Process Check**

* Sequence of events:
    1. Five bullets are loaded.
    2. Chambers are spun.
    3. Gun is fired (at opponent), no bullet.
    4. You are offered a choice about spinning again.
* The initial spin's outcome influences the probabilities for the next potential shot.
* The mechanical process of spinning the chambers reintroduces randomness.

c. **Key Insight Check Results**

* The key insight is that the first shot being blank significantly changes the odds for the next shot if the chambers are not spun again.

d. **Rephrased Question:** (The core question in its simplest form)

Should your opponent spin the chambers again before firing at you?

e. **Known and Deduced Information**

* Explicit Facts:
    * "You are playing Russian roulette with a six-shooter revolver."
    * "Your opponent puts in five bullets" 
    * "spins the chambers"
    * "fires at himself, but no bullet comes out."
    * "He gives you the choice of whether or not he should spin the chambers again before firing at you."

* Feature-Based Deduction:
    * Deduction: There is one empty chamber in the revolver. 
        - Logical Basis: A six-shooter revolver has six chambers, and five bullets were loaded.
        - Based on: "You are playing Russian roulette with a six-shooter revolver.", "Your opponent puts in five bullets" 
    * Deduction: The first chamber fired was the empty one.
        - Logical Basis: The gun fired without a bullet coming out.
        - Based on: "fires at himself, but no bullet comes out."

f. **Problem Decomposition**

* Component/Subcomponent: Initial state of the revolver
    * Supporting Quote(s): "You are playing Russian roulette with a six-shooter revolver.", "Your opponent puts in five bullets", "spins the chambers"
* Component/Subcomponent: First shot
    * Supporting Quote(s): "fires at himself, but no bullet comes out."
* Component/Subcomponent: The choice
    * Supporting Quote(s): "He gives you the choice of whether or not he should spin the chambers again before firing at you."
* Interaction: The outcome of the first shot impacts the probabilities for the second shot if the chambers are not spun again.

g. **Graph of Thought** (describe connections & insights)

* Key Concepts:
    * Russian roulette
    * Six-shooter revolver
    * Five bullets
    * Spinning chambers
    * First shot (blank)
    * Choice to spin again
* Relationships:
    * Russian roulette involves a revolver and a deadly risk
    * The revolver has six chambers, five of which hold bullets
    * Spinning the chambers randomizes the position of the empty chamber
    * The first shot was blank, meaning the empty chamber was selected
    * The choice to spin again affects the odds of the next shot

h. **Spatial and Object Analysis**

* Objects:
    * Six-shooter revolver
    * Five bullets
* Spatial Relationships:
    * The bullets are inside the chambers of the revolver
    * One chamber is empty
    * The position of the empty chamber was initially random due to spinning
    * After the first shot, the empty chamber is the one that was just fired

i. **Bayesian Updates**

* Initial Belief: 1/6 chance of the first shot being blank
* New Information: The first shot was blank
* Updated Belief: The empty chamber is now known, significantly altering the odds for the next shot if the chambers are not spun

j. **Ethical Check and Decision-Making Under Uncertainty**

* Known Information: The first chamber fired was empty. There are five bullets remaining.
* Risks: If the chambers are not spun, the next shot has a very high probability of being fatal.
* Probability Consideration: Spinning the chambers resets the odds to the initial 1/6 chance of a blank shot. Not spinning dramatically increases the risk.
* Risk Aversion Principle: The risk of death is extremely high if the chambers are not spun.
* Ethical and Safety Considerations: Prioritize the option that minimizes the risk of death.
* Decision Recommendation: It is crucial for your opponent to spin the chambers again.

k. **Multiple Solutions**

* Solution 1: Advise your opponent to spin the chambers again
    * Supporting Information: This resets the odds, giving you a higher chance of survival
* Solution 2: Refuse to continue playing
    * Supporting Information: The game is inherently dangerous, and you have no control over the outcome

l. **Quickest and Easiest Solution**

Advise your opponent to spin the chambers again. This is the safest option as it resets the odds to their initial state, significantly decreasing the immediate risk of a fatal shot.

m. **Reflection**

The analysis directly addresses the question of whether or not to spin the chambers again. The conclusion is logically sound and supported by the explicit information provided in the problem statement. No assumptions or inferences beyond the given information were made. The recommendation prioritizes safety and avoids the high-risk scenario of not spinning the chambers.

**Recommendation:**

Insist that your opponent spins the chambers again. The first shot being blank means that if he doesn't spin, the next shot has a 5/6 chance of being fatal. Spinning the chambers resets the odds, giving you a much higher chance of survival.
\end{lstlisting}
\end{quote}
\noindent\hrulefill

\noindent\textbf{D.3 Zero-Shot Claude 3.5 Sonnet Response - Score: 100\%}
\begin{quote}
\begin{lstlisting}[basicstyle=\ttfamily\small, breaklines=true]
To analyze this question, let's think through it step-by-step:

1) Initial state: The revolver has 6 chambers, 5 of which contain bullets.

2) After your opponent fires and no bullet comes out, we know that the empty chamber is now in the firing position.

3) If he doesn't spin again:
   - The next chamber (which will be aimed at you) definitely contains a bullet.
   - Your chance of being shot is 100%.

4) If he does spin again:
   - All 6 chambers are randomized.
   - There are 5 bullets and 1 empty chamber.
   - Your chance of being shot is 5/6, or about 83.33\%.

5) Comparing the two scenarios:
   - No spin: 100\% chance of being shot
   - Spin: ~83.33\% chance of being shot

Therefore, you should choose for him to spin again. While the odds are still not in your favor, they're better than the certainty of being shot if he doesn't spin.

It's worth noting that engaging in such an activity is extremely dangerous and illegal in most jurisdictions. This analysis is purely theoretical and should not be applied to real-life situations.
\end{lstlisting}
\end{quote}
\noindent\hrulefill

\noindent\textbf{D.3 REAP Claude 3.5 Sonnet Response - Score: 100\%}
\begin{quote}
\begin{lstlisting}[basicstyle=\ttfamily\small, breaklines=true]
# REAP Results:

a. **Comprehensive Feature List** (with explicit quotes)

1. "You are playing Russian roulette"
   - Significance: Establishes the context of a dangerous game.
2. "with a six-shooter revolver"
   - Significance: Specifies the type of gun being used, which has six chambers.
3. "Your opponent puts in five bullets"
   - Significance: Indicates the number of bullets loaded into the gun.
4. "spins the chambers"
   - Significance: Describes an action that randomizes the position of the bullets.
5. "fires at himself"
   - Significance: Indicates the first action in the game sequence.
6. "but no bullet comes out"
   - Significance: Describes the outcome of the first shot.
7. "He gives you the choice"
   - Significance: Indicates that the next decision is yours to make.
8. "whether or not he should spin the chambers again"
   - Significance: Describes the choice you need to make.
9. "before firing at you"
   - Significance: Indicates the next action in the game sequence.
10. "Should he spin again?"
    - Significance: The specific question being asked.

b. **Sequential and Mechanical Process Check**

1. Opponent puts in five bullets
2. Opponent spins the chambers
3. Opponent fires at himself (no bullet comes out)
4. You make a choice about spinning again
5. Opponent will fire at you

The sequence implies that the position of the bullets is randomized by spinning, and the outcome of each shot depends on whether a bullet is in the chamber aligned with the barrel.

c. **Key Insight Check Results**

Key Insight: After the first shot where no bullet came out, there are now five bullets in five of the remaining chambers. This means that if the chambers are not spun again, the next shot will definitely result in a bullet being fired.

d. **Rephrased Question:** Should the chambers be spun again before the gun is fired at you in Russian roulette?

e. **Known and Deduced Information**

Explicit Facts:
- "You are playing Russian roulette with a six-shooter revolver." - Establishes the game and the type of gun.
- "Your opponent puts in five bullets" - Specifies the number of bullets loaded.
- "spins the chambers and fires at himself, but no bullet comes out." - Describes the first action and its outcome.
- "He gives you the choice of whether or not he should spin the chambers again before firing at you." - Describes the decision you need to make.

Feature-Based Deduction:
Deduction: There are five bullets in the remaining five chambers.
Logical Basis: Since five bullets were put in, and one chamber was empty when fired, all remaining chambers must contain bullets.
Based on: "Your opponent puts in five bullets", "fires at himself, but no bullet comes out"

f. **Problem Decomposition**

Component 1: Initial gun state
Supporting Quote: "Your opponent puts in five bullets, spins the chambers"
Deduction: The initial state had 5 loaded chambers and 1 empty chamber, randomly positioned.
Based on: "Your opponent puts in five bullets, spins the chambers"

Component 2: First shot outcome
Supporting Quote: "fires at himself, but no bullet comes out"
Deduction: The empty chamber was aligned with the barrel for the first shot.
Based on: "fires at himself, but no bullet comes out"

Component 3: Current gun state
Deduction: All remaining chambers are loaded with bullets.
Based on: "Your opponent puts in five bullets", "fires at himself, but no bullet comes out"

Component 4: Decision to be made
Supporting Quote: "He gives you the choice of whether or not he should spin the chambers again before firing at you."
Deduction: The decision affects the probability of the next shot firing a bullet.
Based on: The entire quoted sentence and the current gun state.

g. **Graph of Thought**

Nodes:
1. Initial state (5 bullets, 1 empty chamber)
2. First spin
3. First shot (empty chamber)
4. Current state (5 bullets in 5 chambers)
5. Decision: Spin or not
6. Next shot

Connections:
1 -> 2: Randomizes bullet positions
2 -> 3: Determines outcome of first shot
3 -> 4: Reveals current state
4 -> 5: Informs decision
5 -> 6: Affects probability of next shot outcome

h. **Spatial and Object Analysis**

Objects:
1. Six-shooter revolver
2. Five bullets
3. Six chambers in the revolver

Spatial Relationships:
- Five bullets are inside five of the six chambers
- One chamber is empty
- After the first shot, the empty chamber is aligned with the barrel

Movement and Dynamics:
- Spinning the chambers rotates the cylinder, randomizing the position of the bullets relative to the barrel
- Firing the gun causes the aligned chamber to discharge if it contains a bullet

i. **Bayesian Updates**

Initial probability of a loaded chamber: 5/6
After first shot (empty chamber): Probability of next chamber being loaded: 5/5 = 1
If spun again: Probability returns to 5/6

j. **Ethical Check and Decision-Making Under Uncertainty**

This situation involves an extremely high risk of death. The ethical consideration here is to minimize the risk of harm. Given the known information, we can determine the probabilities of different outcomes based on the decision made. However, engaging in such a dangerous activity is inherently unethical and illegal. The safest and most ethical choice would be to not participate in this activity at all.

k. **Multiple Solutions**

1. Choose to spin the chambers again
   Supported by: "He gives you the choice of whether or not he should spin the chambers again"
   Effect: Reduces the probability of the next shot firing a bullet from certainty to 5/6

2. Choose not to spin the chambers again
   Supported by: "He gives you the choice of whether or not he should spin the chambers again"
   Effect: The next shot will certainly fire a bullet

3. Refuse to continue the game (not explicitly offered as an option, but ethically necessary to consider)
   Supported by: The inherent danger of the situation described
   Effect: Avoids certain death or high probability of death

l. **Quickest and Easiest Solution**

Based solely on the information given and the question asked, the quickest and easiest solution is to choose to spin the chambers again. This reduces the probability of being shot from certainty (1) to 5/6.

Reasoning: "Your opponent puts in five bullets, spins the chambers and fires at himself, but no bullet comes out." This means that all remaining chambers contain bullets. If not spun again, the next shot will certainly fire a bullet. "He gives you the choice of whether or not he should spin the chambers again before firing at you." By choosing to spin again, you reintroduce uncertainty, giving a 1/6 chance of survival.

m. **Reflection**

The analysis adheres closely to the given information without making unwarranted assumptions. It directly addresses the question asked while considering the probabilities involved. However, it's crucial to note that the ethical implications of participating in such a dangerous activity have not been addressed in the original problem statement. In real-world situations, the ethical course of action would be to not participate in such a deadly game at all.

**Recommendation:**

Based solely on the information provided and the specific question asked, the recommendation is that he should spin the chambers again before firing at you. This choice reduces the probability of being shot from certainty (1) to 5/6, which is the best outcome available within the constraints of the given scenario.

However, it is critically important to emphasize that participating in Russian roulette or any similar activity is extremely dangerous, unethical, and illegal. The safest and most ethical course of action, which goes beyond the strict confines of the problem but must be stated, is to not participate in this activity at all. The risk to human life is too great, and no potential outcome justifies such a risk.
\end{lstlisting}
\end{quote}
\noindent\hrulefill

\label{subsec:question4}
\subsection{\textbf{\large Question 4: A farmer wants to cross a river and take with him a wolf, a goat, and a cabbage. He has a boat with three secure separate compartments. If the wolf and the goat are alone on one shore, the wolf will eat the goat. If the goat and the cabbage are alone on the shore, the goat will eat the cabbage. How can the farmer efficiently bring the wolf, the goat, and the cabbage across the river without anything being eaten?}}
\label{sec:question4}

\noindent\hrulefill 

\noindent\textbf{Human Response - Score 100\%:}
\begin{quote}
Place the wolf, goat, and cabbage in separate secure compartments in the boat and row across the river. This will prevent any of them from being eaten by the others.
\end{quote}

\noindent\hrulefill 

\noindent\textbf{D.4 Zero-Shot OpenAI o1-preview Response - Score: 100\%}
\begin{quote}
\begin{lstlisting}[basicstyle=\ttfamily\small, breaklines=true]
To solve this problem efficiently, the key is to utilize the boat's three secure separate compartments. These compartments ensure that while in the boat, the wolf cannot eat the goat, and the goat cannot eat the cabbage.

Here's how the farmer can proceed:

1. **Load all three into the boat**: The farmer places the wolf, the goat, and the cabbage into the three separate compartments of the boat.

2. **Cross the river in one trip**: The farmer rows across the river with all three items safely separated.

3. **Unload on the other side**: Upon reaching the opposite shore, the farmer unloads all three items.

By transporting all three together in one trip, the farmer avoids any dangerous combinations both on the shores and during the crossing. The secure compartments prevent any incidents on the boat, and since the farmer is present during loading and unloading, there's no moment when the wolf and goat or the goat and cabbage are left alone together on either shore.

**Answer: By loading all three into the boat's secure compartments and taking them across together in one trip.**
\end{lstlisting}
\end{quote}
\noindent\hrulefill 

\noindent\textbf{D.4 REAP OpenAI o1-preview Response - Score: 100\%}
\begin{quote}
\begin{lstlisting}[basicstyle=\ttfamily\small, breaklines=true]
#"REAP Results: :"

**a. Comprehensive Feature List**

- **Objects mentioned:**
  - "A farmer"
  - "a river"
  - "a wolf"
  - "a goat"
  - "a cabbage"
  - "a boat with three secure separate compartments"

- **Actors or entities involved:**
  - "a farmer"
  - "a wolf"
  - "a goat"
  - "a cabbage"

- **Actions described:**
  - "wants to cross a river and take with him a wolf, a goat and a cabbage"
  - "If the wolf and the goat are alone on one shore, the wolf will eat the goat"
  - "If the goat and the cabbage are alone on the shore, the goat will eat the cabbage"

- **Directional, spatial, relational, and sequential phrases:**
  - "wants to cross a river"
  - "take with him"
  - "He has a boat with three secure separate compartments"
  - "If the wolf and the goat are alone on one shore, the wolf will eat the goat"
  - "If the goat and the cabbage are alone on the shore, the goat will eat the cabbage"

- **Conditions or constraints:**
  - "He has a boat with three secure separate compartments"
  - "If the wolf and the goat are alone on one shore, the wolf will eat the goat"
  - "If the goat and the cabbage are alone on the shore, the goat will eat the cabbage"

- **Numerical values or quantities mentioned:**
  - "three secure separate compartments"

- **Spatial information:**
  - "cross a river"
  - "alone on one shore"

- **Causal relationships mentioned:**
  - "If the wolf and the goat are alone on one shore, the wolf will eat the goat"
  - "If the goat and the cabbage are alone on the shore, the goat will eat the cabbage"

- **Stated goals or objectives:**
  - "How can the farmer efficiently bring the wolf, the goat and the cabbage across the river without anything being eaten?"

---

**b. Sequential and Mechanical Process Check**

- **Sequence of events:**
  - The farmer needs to transport himself, the wolf, the goat, and the cabbage across the river.
  - Constraints prevent the wolf and goat or the goat and cabbage from being left alone on one shore.

- **Mechanical process:**
  - The farmer has a boat with "three secure separate compartments," suggesting he can transport all three items simultaneously.

- **Impact of sequences:**
  - Transporting all items together avoids leaving any dangerous pair alone on the shore.

- **Assessment:**
  - The availability of the boat with sufficient compartments allows for a direct solution without intermediate steps.

---

**c. Key Insight Check Results**

- **Key Insight:**
  - The boat has "three secure separate compartments," enabling the farmer to transport the wolf, goat, and cabbage all at once without risk.

- **Conclusion:**
  - By taking all three items together, the farmer avoids leaving the wolf with the goat or the goat with the cabbage alone on any shore.

---

**d. Rephrased Question**

"How can the farmer use his boat with three secure separate compartments to bring the wolf, the goat, and the cabbage across the river efficiently without anything being eaten?"

---

**e. Known and Deduced Information**

- **Explicit Facts:**
  - "A farmer wants to cross a river and take with him a wolf, a goat and a cabbage."
  - "He has a boat with three secure separate compartments."
  - "If the wolf and the goat are alone on one shore, the wolf will eat the goat."
  - "If the goat and the cabbage are alone on the shore, the goat will eat the cabbage."

- **Deductions:**
  - The boat can carry the farmer plus the wolf, the goat, and the cabbage simultaneously.
    - **Logical Basis:** The boat has "three secure separate compartments," one for each item.
  - Transporting all items together prevents any dangerous pair from being left alone on a shore.
    - **Logical Basis:** No need to leave any items behind when all can be moved together.

---

**f. Problem Decomposition**

- **Component:** Transporting all items together.
  - **Supporting Quote:** "He has a boat with three secure separate compartments."
  - **Deduction:** The farmer can take all items across in one trip.
    - **Based on:** The boat's capacity accommodates all items securely.

- **Component:** Avoiding dangerous pairings on the shore.
  - **Supporting Quotes:**
    - "If the wolf and the goat are alone on one shore, the wolf will eat the goat."
    - "If the goat and the cabbage are alone on the shore, the goat will eat the cabbage."
  - **Deduction:** By not leaving any items behind, the farmer prevents any risk.

---

**g. Graph of Thought**

- **Nodes:**
  - Farmer
  - Wolf
  - Goat
  - Cabbage
  - Left Shore
  - Right Shore
  - Boat with three compartments

- **Edges/Connections:**
  - Boat connects left shore to right shore.
  - Farmer controls the movement of the boat.
  - Secure compartments prevent interaction between the wolf, goat, and cabbage during transport.

- **Insights:**
  - All items can be moved together without interaction.
  - No need for multiple trips or leaving items unattended.

---

**h. Spatial and Object Analysis**

- **Objects and Properties:**
  - **Boat:** "three secure separate compartments" prevents interaction.
  - **Wolf, Goat, Cabbage:** Items to be transported with specific interaction risks.

- **Spatial Relationships:**
  - Two shores of the river.
  - Boat travels between shores.

- **Visualization:**
  - Farmer places the wolf, goat, and cabbage each in separate compartments.
  - All are transported together from the left shore to the right shore.

---

**i. Bayesian Updates**

- No updates needed as all information is explicit and leads directly to the solution.

---

**j. Ethical Check and Decision-Making Under Uncertainty**

- **Assessment:**
  - No severe risks or ethical dilemmas exist with transporting all items together.
  - Ensures the safety of all animals and the cabbage.

- **Recommendation:**
  - Proceed with transporting all items simultaneously to avoid any risk.

---

**k. Multiple Solutions**

- **Solution 1:**
  - **Action:** Farmer takes the wolf, goat, and cabbage across the river together.
  - **Support:** "He has a boat with three secure separate compartments."

- **Solution 2:**
  - Not applicable as no other solution is necessary based on the explicit information.

- **Solution 3:**
  - Not applicable due to sufficiency of the first solution.

---

**l. Quickest and Easiest Solution**

- **Solution:**
  - The farmer places the wolf, the goat, and the cabbage each in separate compartments of the boat and crosses the river in one trip.
- **Reasoning:**
  - This action complies with the boat's capacity and ensures no dangerous pairings are left alone on any shore.

---

**m. Reflection**

- **Confirmation:**
  - The solution directly answers the question using only explicit information.
- **No Contradictions:**
  - All steps are logically consistent with the stated facts.
- **Assumptions:**
  - No assumptions were made beyond the provided information.
- **Completeness:**
  - The farmer accomplishes the goal efficiently without anything being eaten.

---

**Recommendation:**

The farmer should place the wolf, the goat, and the cabbage each in one of the "three secure separate compartments" of his boat and cross the river with them all at once. This method allows him to efficiently bring all items across the river without leaving the wolf with the goat or the goat with the cabbage alone on any shore, ensuring that nothing is eaten.
\end{lstlisting}
\end{quote}
\noindent\hrulefill 

\noindent\textbf{D.4 Zero-Shot OpenAI o1-mini Response - Score: 0\%}
\begin{quote}
\begin{lstlisting}[basicstyle=\ttfamily\small, breaklines=true]
Certainly! Here's a step-by-step solution to ensure the farmer successfully transports the wolf, goat, and cabbage across the river without any of them being eaten:

### **Step 1: Take the Goat Across**
- **Action:** The farmer takes the **goat** in the boat and crosses to the far side of the river.
- **Result:** 
  - **Near Shore:** Wolf and cabbage are left alone. *Safe*, as the wolf doesn't eat the cabbage.
  - **Far Shore:** Goat is safely with the farmer.

### **Step 2: Return Alone**
- **Action:** The farmer returns to the near shore by himself.
- **Result:**
  - **Near Shore:** Wolf and cabbage are still alone. *Safe*.
  - **Far Shore:** Goat remains on the far side.

### **Step 3: Take the Wolf Across**
- **Action:** The farmer takes the **wolf** in the boat and crosses to the far side.
- **Result:**
  - **Near Shore:** Cabbage is alone. *Safe*.
  - **Far Shore:** Wolf is with the goat, which is a potential problem if left alone.

### **Step 4: Bring the Goat Back**
- **Action:** To prevent the wolf from eating the goat, the farmer **brings the goat back** to the near shore.
- **Result:**
  - **Near Shore:** Goat and cabbage are together. *Safe*, as the goat can eat the cabbage only if unsupervised.
  - **Far Shore:** Wolf remains alone. *Safe*.

### **Step 5: Take the Cabbage Across**
- **Action:** The farmer takes the **cabbage** in the boat and crosses to the far side.
- **Result:**
  - **Near Shore:** Goat remains alone. *Safe*.
  - **Far Shore:** Wolf and cabbage are together. *Safe*, as the wolf doesn't eat the cabbage.

### **Step 6: Return Alone**
- **Action:** The farmer returns to the near shore by himself.
- **Result:**
  - **Near Shore:** Goat is alone. *Safe*.
  - **Far Shore:** Wolf and cabbage remain together. *Safe*.

### **Step 7: Take the Goat Across**
- **Action:** Finally, the farmer takes the **goat** in the boat and crosses to the far side.
- **Result:**
  - **Near Shore:** All items (wolf, goat, cabbage) are safely on the far shore.
  - **Far Shore:** Everyone is safely across the river.

### **Summary of Moves:**
1. **Goat** -> Far Shore
2. **Farmer** returns alone
3. **Wolf** -> Far Shore
4. **Goat** returns to Near Shore
5. **Cabbage** -> Far Shore
6. **Farmer** returns alone
7. **Goat** -> Far Shore

By following these steps, the farmer ensures that the wolf doesn't eat the goat and the goat doesn't eat the cabbage at any point during the crossing.
\end{lstlisting}
\end{quote}

\noindent\hrulefill 

\noindent\textbf{D.4 REAP OpenAI o1-mini Response - Score: 100\%}
\begin{quote}
\begin{lstlisting}[basicstyle=\ttfamily\small, breaklines=true]
# REAP Results: 

a. **Comprehensive Feature List**
- **Objects Mentioned:**
  - "[a farmer]"
  - "[a wolf]"
  - "[a goat]"
  - "[a cabbage]"
  - "[a boat with three secure separate compartments]"
- **Actors or Entities Involved:**
  - "[the farmer]"
  - "[the wolf]"
  - "[the goat]"
  - "[the cabbage]"
- **Actions Described:**
  - "[wants to cross a river]"
  - "[take with him a wolf, a goat and a cabbage]"
  - "[cross the river without anything being eaten]"
- **Directional, Spatial, Relational, and Sequential Phrases:**
  - "[cross a river]"
  - "[wolf and the goat are alone on one shore]"
  - "[goat and the cabbage are alone on the shore]"
- **Stated Conditions or Constraints:**
  - "[wolf will eat the goat]"
  - "[goat will eat the cabbage]"
- **Numerical Values or Quantities:**
  - "[three secure separate compartments]" (referring to the boat)
- **Temporal Information:**
  - None explicitly stated.
- **Spatial Information:**
  - "[on one shore]"
  - "[across the river]"
- **Causal Relationships Mentioned:**
  - "[If the wolf and the goat are alone on one shore, the wolf will eat the goat.]"
  - "[If the goat and the cabbage are alone on the shore, the goat will eat the cabbage.]"
- **Stated Goals or Objectives:**
  - "[efficiently bring the wolf, the goat and the cabbage across the river without anything being eaten]"

b. **Sequential and Mechanical Process Check**
- **Sequence of Actions:**
  1. Farmer takes one or more items in the boat.
  2. Farmer crosses the river.
  3. Ensures no prohibited pair is left alone.
  4. Repeats as necessary until all items are across.
- **Influence on Outcome:**
  - The order of transporting items affects whether any item gets eaten.
- **Locked Outcomes:**
  - Leaving wolf and goat alone results in the wolf eating the goat.
  - Leaving goat and cabbage alone results in the goat eating the cabbage.
- **Logical Steps:**
  - Each trip must prevent prohibited pairs from being left together.

c. **Key Insight Check Results**
- **Potential Key Insight:**
  - Utilizing the boat's three separate compartments allows transporting multiple items simultaneously, potentially reducing the number of trips and preventing prohibited pairs from being left alone.
- **Explanation:**
  - By carrying the farmer in one compartment and two items in the other compartments, the farmer can manage the positions of the items to avoid any eating scenarios.

d. **Rephrased Question:**
How can the farmer use his three-compartment boat to transport the wolf, goat, and cabbage across the river without the wolf eating the goat or the goat eating the cabbage?

e. **Known and Deduced Information**
- **Explicit Facts:**
  - "[a farmer] wants to [cross a river] and [take with him a wolf, a goat and a cabbage]."
  - "[He has a boat with three secure separate compartments]."
  - "[If the wolf and the goat are alone on one shore, the wolf will eat the goat.]"
  - "[If the goat and the cabbage are alone on the shore, the goat will eat the cabbage.]"
  - "[How can the farmer efficiently bring the wolf, the goat and the cabbage across the river without anything being eaten?]"
- **Feature-Based Deduction:**
  - **Deduction:** The boat can carry multiple items simultaneously.
    - **Logical Basis:** The boat has three separate compartments.
    - **Based on:** "[a boat with three secure separate compartments]"

f. **Problem Decomposition**
- **Component/Subcomponent:** Transportation Vehicles
  - **Description:** Boat with three secure separate compartments.
  - **Supporting Quote(s):** "[a boat with three secure separate compartments]"
- **Component/Subcomponent:** Entities to Transport
  - **Description:** Wolf, Goat, Cabbage.
  - **Supporting Quote(s):** "[a wolf, a goat and a cabbage]"
- **Component/Subcomponent:** Constraints
  - **Description:** Wolf cannot be left alone with Goat; Goat cannot be left alone with Cabbage.
  - **Supporting Quote(s):** "[If the wolf and the goat are alone on one shore, the wolf will eat the goat.]" and "[If the goat and the cabbage are alone on the shore, the goat will eat the cabbage.]"
- **Component/Subcomponent:** Objective
  - **Description:** Efficiently transport all entities without any being eaten.
  - **Supporting Quote(s):** "[How can the farmer efficiently bring the wolf, the goat and the cabbage across the river without anything being eaten?]"

g. **Graph of Thought**
- **Key Concepts:**
  - Farmer
  - Boat with compartments
  - Wolf
  - Goat
  - Cabbage
  - River shores
- **Relationships:**
  - Farmer uses boat to transport items.
  - Wolf and goat cannot be left alone.
  - Goat and cabbage cannot be left alone.
- **Connections & Insights:**
  - Utilizing multiple compartments allows for strategic transportation to prevent prohibited pairs from being left alone.

h. **Spatial and Object Analysis**
- **Identify Objects:**
  - Farmer, Wolf, Goat, Cabbage, Boat.
- **Spatial Relationships:**
  - Initial Shore: Farmer, Wolf, Goat, Cabbage.
  - Destination Shore: Initially empty.
- **Movement and Dynamics:**
  - Boat can carry the farmer and up to three items simultaneously (one per compartment).
- **Constraints:**
  - Prohibition against leaving certain pairs alone on either shore.
- **Visualization:**
  - Two shores separated by a river with a boat capable of transporting multiple items securely.
- **Object Properties:**
  - Boat has three secure compartments, preventing interactions between items while in transit.
- **Spatial Implications:**
  - Ability to transport multiple items simultaneously allows for more flexibility in preventing prohibited pairs from being left alone.

i. **Bayesian Thinking**
- **Beliefs Updated Based on Explicit Information:**
  - The boat's capacity allows transporting multiple items, potentially altering traditional single-item ferrying strategies.
- **Potential Implications:**
  - Greater transportation capacity may lead to more efficient solutions, reducing the number of trips needed.

j. **Ethical Check and Decision-Making Under Uncertainty**
- **Assess Known Information:**
  - Objective is to prevent any harm (items being eaten).
- **Identify Risks:**
  - Risk of items being eaten if constraints are violated.
- **Probability Consideration:**
  - High certainty that prohibited pairs lead to one being eaten.
- **Risk Aversion Principle:**
  - Prioritize movements that eliminate any possibility of prohibited pairs being left alone.
- **Ethical and Safety Considerations:**
  - Ensure all actions strictly adhere to constraints to prevent harm.
- **Decision Recommendation:**
  - Follow a transportation sequence that systematically avoids leaving prohibited pairs alone at any time.

k. **Multiple Solutions**
1. **Solution 1: Traditional Method Adapted for Multiple Compartments**
   - **Steps:**
     1. Farmer takes Goat across.
     2. Farmer returns alone.
     3. Farmer takes Wolf and Cabbage together.
   - **Supporting Information:**
     - "[a boat with three secure separate compartments]" allows carrying multiple items.
2. **Solution 2: Simultaneous Transport to Minimize Trips**
   - **Steps:**
     1. Farmer takes Wolf and Cabbage across simultaneously.
     2. Farmer returns with Goat.
     3. Farmer takes Goat across.
   - **Supporting Information:**
     - Multiple compartments allow transporting Wolf and Cabbage together without conflict.
3. **Solution 3: Group Transport with All Items**
   - **Steps:**
     1. Farmer takes Goat, Wolf, and Cabbage together in separate compartments.
   - **Supporting Information:**
     - All items can be transported simultaneously, eliminating the need for multiple trips.

l. **Quickest and Easiest Solution**
- **Solution 3: Group Transport with All Items**
  - **Reasoning:**
    - The boat's three separate compartments allow the farmer to transport the Wolf, Goat, and Cabbage all at once without leaving any prohibited pairs alone.
  - **Supporting Quote:**
    - "[a boat with three secure separate compartments]"

m. **Reflection**
- **Contradictions Check:**
  - Solution 3 adheres to all constraints by ensuring no prohibited pairs are left alone at any shore.
- **Logical Soundness:**
  - Transporting all items simultaneously prevents any conflicts as no items remain unattended.
- **Adherence to Explicit Information:**
  - Fully utilizes the boat's capacity as explicitly stated.
- **Final Review:**
  - Ensures that the proposed solution directly answers the question without introducing assumptions.

**Recommendation:**

To efficiently transport the wolf, goat, and cabbage across the river without any being eaten, the farmer should utilize the boat's three secure separate compartments to bring all three items simultaneously in a single trip. This approach ensures that no prohibited pairs are left alone on either shore at any time, thereby preventing the wolf from eating the goat or the goat from eating the cabbage.
\end{lstlisting}
\end{quote}

\noindent\hrulefill 

\noindent\textbf{D.4 Zero-Shot OpenAI GPT-4o Response - Score: 0\%}
\begin{quote}
\begin{lstlisting}[basicstyle=\ttfamily\small, breaklines=true]
The farmer can cross the river with the wolf, the goat, and the cabbage without anything being eaten by following these steps:

1. **Take the goat across the river first.**
   - This leaves the wolf and the cabbage alone on the starting shore, which is safe because neither can harm the other.

2. **Go back alone to the starting shore.**
   - The goat is now safely on the other side of the river.

3. **Take the wolf across the river.**
   - Now the farmer and the wolf are on the other side, and the goat is alone on the starting shore.

4. **Take the goat back to the starting shore.**
   - The farmer now has the goat on the starting shore again, and the wolf is safely on the other shore.

5. **Take the cabbage across the river.**
   - This leaves the goat alone on the starting shore, and the wolf and cabbage on the other shore, which is safe.

6. **Go back alone to the starting shore.**
   - The wolf and cabbage are now safely on the other side of the river.

7. **Finally, take the goat across the river again.**
   - Now all three, the goat, the wolf, and the cabbage, are safely on the other side of the river.

By following these steps, the farmer successfully brings the wolf, the goat, and the cabbage across the river without anything being eaten.
\end{lstlisting}
\end{quote}
\noindent\hrulefill

\noindent\textbf{D.4 REAP OpenAI GPT-4o Response - Score: 100\%}
\begin{quote}
\begin{lstlisting}[basicstyle=\ttfamily\small, breaklines=true]
# REAP Results:

### a. **Comprehensive Feature List** (with explicit quotes)

- Objects: 
  - "a wolf"
  - "a goat"
  - "a cabbage"
  - "a boat with three secure separate compartments"
- Actors: 
  - "a farmer"
- Actions:
  - "wants to cross a river"
  - "take with him a wolf, a goat and a cabbage"
  - "If the wolf and the goat are alone on one shore, the wolf will eat the goat."
  - "If the goat and the cabbage are alone on the shore, the goat will eat the cabbage."
- Conditions/Constraints:
  - "If the wolf and the goat are alone on one shore, the wolf will eat the goat."
  - "If the goat and the cabbage are alone on the shore, the goat will eat the cabbage."
  - "The farmer has a boat with three secure separate compartments."
  
### b. **Sequential and Mechanical Process Check**

- The sequence of events involves transporting the wolf, goat, and cabbage across the river.
- The farmer must prevent the wolf from being left alone with the goat and the goat from being left alone with the cabbage.
- The boat has three compartments, so all three items can theoretically be transported simultaneously without risk during the boat trip.

### c. **Key Insight Check Results**

- The key insight is the presence of "three secure separate compartments" in the boat, which means the farmer can safely transport all three items at once without any risk of the goat or cabbage being eaten.
  
### d. **Rephrased Question:**  
- How can the farmer use the boat with three compartments to ensure that the wolf, goat, and cabbage safely cross the river without any being eaten?

### e. **Known and Deduced Information**

- "[The farmer has] a boat with three secure separate compartments" - This means that the farmer can transport all three items simultaneously without any risk of predation during the trip.
- "If the wolf and the goat are alone on one shore, the wolf will eat the goat." - The wolf and goat cannot be left alone together.
- "If the goat and the cabbage are alone on the shore, the goat will eat the cabbage." - The goat and cabbage cannot be left alone together.

### f. **Problem Decomposition**

- **Component**: Transporting the wolf, goat, and cabbage.
  - **Supporting Quote**: "a boat with three secure separate compartments"
- **Component**: Ensuring no predation occurs.
  - **Supporting Quote**: "If the wolf and the goat are alone on one shore, the wolf will eat the goat."
  - **Supporting Quote**: "If the goat and the cabbage are alone on the shore, the goat will eat the cabbage."
- **Interaction**: The boat compartments allow safe transport of all items together.
  - **Supporting Quote**: "three secure separate compartments"

### g. **Graph of Thought**

- **Nodes**: Farmer, Wolf, Goat, Cabbage, Boat with compartments.
- **Edges**: 
  - Farmer -> Transport of Wolf, Goat, Cabbage
  - Boat -> Secure separation of Wolf, Goat, and Cabbage

### h. **Spatial and Object Analysis**

- Objects: Wolf, Goat, Cabbage, Boat
- Spatial Relationships:
  - On the shore: Wolf-Goat or Goat-Cabbage alone = Risk
  - In the boat: No risk due to separate compartments.

### i. **Bayesian Updates**

- Since no new information alters the setup, the initial conditions and insights remain valid.

### j. **Ethical Check and Decision-Making Under Uncertainty**

- No severe risks or ethical dilemmas; the farmer's main concern is preventing predation.

### k. **Multiple Solutions**

1. **Solution 1**: Farmer transports the wolf, goat, and cabbage together.
   - Supported by: "a boat with three secure separate compartments."
2. **Solution 2**: Transport each item one by one while ensuring no harmful combinations on the shore.
   - More complex but supported by the constraints.

### l. **Quickest and Easiest Solution**

- **Transport all three together** in the boat using the secure compartments.
  - Supported by: "a boat with three secure separate compartments."

### m. **Reflection**

- The solution adheres strictly to the explicit information provided.
- No assumptions were made beyond what was explicitly stated.
- The problem is resolved efficiently by leveraging the boat's compartment feature.

### **Recommendation:**

- The farmer should place the wolf, goat, and cabbage in the three separate compartments of the boat and transport them all together across the river. This ensures that nothing gets eaten, as the compartments prevent any interaction between the wolf, goat, and cabbage during the crossing.
\end{lstlisting}
\end{quote}
\noindent\hrulefill

\noindent\textbf{D.4 Zero-Shot OpenAI GPT-4o-mini Response - Score: 0\%}
\begin{quote}
\begin{lstlisting}[basicstyle=\ttfamily\small, breaklines=true]
To solve this problem, the farmer needs to ensure that neither the wolf eats the goat nor the goat eats the cabbage when left alone together. The farmer has a boat with three secure compartments, which allows for each item to be transported separately. Here's a step-by-step solution to get all three items across the river safely:

1. **Take the goat across the river first** and leave it on the other shore. This ensures that the wolf and cabbage are left alone together on the starting shore, and they don't pose a threat to each other.

2. **Return alone to the starting shore** and take the wolf across the river.

3. **Leave the wolf on the other shore** and take the goat back with you to the starting shore. This way, the wolf and goat are not left alone together, and the goat is safely brought back to the starting shore.

4. **Take the cabbage across the river** and leave it with the wolf on the other shore. Now the cabbage and wolf are together, which is safe since the wolf does not eat the cabbage.

5. **Return alone to the starting shore** to get the goat.

6. **Finally, take the goat across the river** one last time.

Now, all three items- wolf, goat, and cabbage - are safely across the river, and no item has been eaten. 

To summarize:
1. Goat across (leave goat).
2. Return alone.
3. Wolf across (leave wolf, take goat back).
4. Cabbage across (leave cabbage with wolf).
5. Return alone.
6. Goat across (all items on the far shore).
\end{lstlisting}
\end{quote}
\noindent\hrulefill

\noindent\textbf{D.4 REAP OpenAI GPT-4o-mini Response - Score: 0\%}
\begin{quote}
\begin{lstlisting}[basicstyle=\ttfamily\small, breaklines=true]
# REAP Results:

### a. Comprehensive Feature List

1. **Objects Mentioned:**
   - "wolf"
   - "goat"
   - "cabbage"
   - "boat with three secure separate compartments"

2. **Actors/Entities:**
   - "farmer"

3. **Actions Described:**
   - Farmer wants to cross a river.
   - Wolf and goat alone on one shore: wolf will eat the goat.
   - Goat and cabbage alone on one shore: goat will eat the cabbage.

4. **Directional, Spatial, Relational, and Sequential Phrases:**
   - "across the river" - indicates movement from one shore to the other.
   - "alone on one shore" - indicates the positioning and potential for conflict if left unattended.

5. **Stated Conditions or Constraints:**
   - Wolf and goat cannot be alone together on one shore.
   - Goat and cabbage cannot be alone together on one shore.

6. **Numerical Values or Quantities Mentioned:**
   - No numerical values or quantities are specified beyond the three items to be transported.

7. **Temporal Information:**
   - No specific time durations are given, only the sequence of actions is implied.

8. **Spatial Information:**
   - "One shore" - refers to each side of the river.

9. **Causal Relationships:**
   - If the wolf and goat are alone, the wolf will eat the goat.
   - If the goat and cabbage are alone, the goat will eat the cabbage.

10. **Stated Goals/Objectives:**
    - Efficiently bring the wolf, the goat, and the cabbage across the river without anything being eaten.

### b. Sequential and Mechanical Process Check

1. **Sequence of Events:**
   - The farmer must transport the wolf, the goat, and the cabbage across the river.
   - The sequence must avoid leaving the wolf alone with the goat and the goat alone with the cabbage on either shore.

2. **Key Questions:**
   - Does the sequence of events lock in a particular outcome?
   - How does each step influence the system?

### c. Key Insight Check Results

- The critical insight is that the boat compartments allow for simultaneous transportation of the items, provided the farmer carefully manages which items are left together on each shore.

### d. Rephrased Question

**How can the farmer transport the wolf, the goat, and the cabbage across the river without any of them being eaten?**

### e. Known and Deduced Information

1. **Explicit Quotes:**
   - "If the wolf and the goat are alone on one shore, the wolf will eat the goat."
   - "If the goat and the cabbage are alone on the shore, the goat will eat the cabbage."
   - "Boat with three secure separate compartments" - indicates the ability to transport items without mixing them.

2. **Feature-Based Deduction:**
   - Deduction: The farmer must use the boat to manage the crossings while avoiding leaving the wolf and goat alone or the goat and cabbage alone.
   - Logical Basis: This is based on the constraints given about what happens when the items are left together.

### f. Problem Decomposition

1. **Component: Crossing Strategy**
   - Supporting Quote(s): "The wolf and the goat cannot be alone together on one shore," "The goat and the cabbage cannot be alone together on one shore."
   - Deduction: The farmer needs to strategically move items to ensure these constraints are not violated.

### g. Graph of Thought

1. **Nodes:**
   - Farmer
   - Wolf
   - Goat
   - Cabbage
   - Shore 1 (Starting shore)
   - Shore 2 (Destination shore)

2. **Edges:**
   - The farmer needs to transport each item, ensuring no conflicts occur.

3. **Pathways:**
   - The graph would show transitions from one shore to the other with considerations for not violating constraints.

### h. Spatial and Object Analysis

1. **Objects:**
   - Wolf
   - Goat
   - Cabbage
   - Boat with compartments

2. **Spatial Relationships:**
   - On the shore: Wolf, goat, and cabbage can't be alone together if not monitored.

### i. Bayesian Updates

- No additional beliefs or updates based on explicitly stated information.

### j. Ethical Check and Decision-Making Under Uncertainty

- No severe risks are involved other than the risk of the items being eaten if not properly managed.

### k. Multiple Solutions

1. **Solution 1:**
   - Take the goat across first.
   - Return alone.
   - Take the wolf across.
   - Return with the goat.
   - Take the cabbage across.
   - Return alone.
   - Finally, take the goat across.

2. **Solution 2:**
   - Take the cabbage across first.
   - Return alone.
   - Take the goat across.
   - Return with the cabbage.
   - Take the wolf across.
   - Return alone.
   - Finally, take the cabbage across.

3. **Solution 3:**
   - Take the goat across first.
   - Return alone.
   - Take the cabbage across.
   - Return with the goat.
   - Take the wolf across.
   - Return alone.
   - Finally, take the goat across.

### l. Quickest and Easiest Solution

**Solution 1** is the simplest and directly follows the constraints:

1. Take the goat across the river.
2. Return alone.
3. Take the wolf across the river.
4. Return with the goat.
5. Take the cabbage across the river.
6. Return alone.
7. Finally, take the goat across the river.

### m. Reflection

The solution adheres to all explicit constraints given. The analysis has strictly followed the problem's instructions, ensuring no assumptions beyond the stated information. Each step in the solution avoids the conflicts mentioned and fulfills the goal of transporting all items safely across the river.

**Recommendation:**

The simplest solution to transport the wolf, the goat, and the cabbage across the river without any of them being eaten is:

1. Take the goat across the river.
2. Return alone.
3. Take the wolf across the river.
4. Return with the goat.
5. Take the cabbage across the river.
6. Return alone.
7. Finally, take the goat across the river.
\end{lstlisting}
\end{quote}
\noindent\hrulefill

\noindent\textbf{D.4 Zero-Shot Google Gemini 1.5 Pro Response - Score: 0\%}
\begin{quote}
\begin{lstlisting}[basicstyle=\ttfamily\small, breaklines=true]
Absolutely! Here's how the farmer can efficiently solve this classic riddle:

**Trips:**

1. **Farmer takes the goat across the river**, leaving the wolf and the cabbage behind. 
2. **Farmer returns alone.**
3. **Farmer takes the wolf across the river**, leaving the cabbage behind.
4. **Farmer returns with the goat.** (This is important to prevent the wolf from eating the goat on the other side)
5. **Farmer takes the cabbage across the river** to the wolf.
6. **Farmer returns alone.**
7. **Farmer takes the goat across the river** to join the wolf and the cabbage.

**And that's it!** Everything is safely transported across the river without anything being eaten.
\end{lstlisting}
\end{quote}
\noindent\hrulefill

\noindent\textbf{D.4 REAP Google Gemini 1.5 Pro Response - Score: 0\%}
\begin{quote}
\begin{lstlisting}[basicstyle=\ttfamily\small, breaklines=true]
# REAP Results:

a. **Comprehensive Feature List** (with explicit quotes)

* "A farmer wants to cross a river"
* "take with him a wolf, a goat and a cabbage"
* "He has a boat with three secure separate compartments"
* "If the wolf and the goat are alone on one shore, the wolf will eat the goat"
* "If the goat and the cabbage are alone on the shore, the goat will eat the cabbage"
* "How can the farmer efficiently bring the wolf, the goat and the cabbage across the river without anything being eaten?"

b. **Sequential and Mechanical Process Check**

* The problem involves a sequence of river crossings.
* Each crossing can change the grouping of the farmer, wolf, goat, and cabbage on each shore.
* The sequence of crossings must adhere to the constraints to prevent the wolf from eating the goat or the goat from eating the cabbage.
* The goal is to find the most efficient sequence that satisfies these constraints.

c. **Key Insight Check Results**

* No immediate key insight that leads to a direct solution.

d. **Rephrased Question:** (The core question in its simplest form)

What is the sequence of river crossings that allows the farmer to transport the wolf, goat, and cabbage across the river, ensuring nothing is eaten?

e. **Known and Deduced Information**

Explicit Facts:

* "[A farmer wants to cross a river]"
* "[take with him a wolf, a goat and a cabbage]"
* "[He has a boat with three secure separate compartments]"
* "[If the wolf and the goat are alone on one shore, the wolf will eat the goat]"
* "[If the goat and the cabbage are alone on the shore, the goat will eat the cabbage]"
* "[How can the farmer efficiently bring the wolf, the goat and the cabbage across the river without anything being eaten?]"

Feature-Based Deduction:

* Deduction: The farmer must always be present to prevent the wolf from eating the goat or the goat from eating the cabbage.
* Logical Basis: The problem states that these events will happen if they are left alone.
* Based on: "[If the wolf and the goat are alone on one shore, the wolf will eat the goat]", "[If the goat and the cabbage are alone on the shore, the goat will eat the cabbage]"

* Deduction: At most two items can be transported across the river at once.
* Logical Basis: The boat has three compartments, and one must be occupied by the farmer.
* Based on: "[He has a boat with three secure separate compartments]"

f. **Problem Decomposition**

* Component/Subcomponent: Initial State
* Supporting Quote(s): "[A farmer wants to cross a river and take with him a wolf, a goat and a cabbage]"
* All four are on the starting shore.

* Component/Subcomponent: Goal State
* Supporting Quote(s): "[How can the farmer efficiently bring the wolf, the goat and the cabbage across the river without anything being eaten?]"
* All four are on the destination shore.

* Component/Subcomponent: Constraints
* Supporting Quote(s): "[If the wolf and the goat are alone on one shore, the wolf will eat the goat]", "[If the goat and the cabbage are alone on the shore, the goat will eat the cabbage]"
* Wolf and goat cannot be alone together.
* Goat and cabbage cannot be alone together.

* Component/Subcomponent: Resources
* Supporting Quote(s): "[He has a boat with three secure separate compartments]"
* A boat that can carry the farmer and at most two other items.

* Component/Subcomponent: Actions
* The farmer can transport himself and up to two items across the river.
* The farmer can transport himself alone across the river.

g. **Graph of Thought** (describe connections & insights)

* Key Concepts:
    * Starting Shore
    * Destination Shore
    * Farmer
    * Wolf
    * Goat
    * Cabbage
    * Boat
    * Constraints (Wolf eats Goat, Goat eats Cabbage)

* Relationships:
    * Farmer can move between shores using the boat.
    * Farmer can transport items using the boat.
    * Constraints limit possible combinations on each shore.

* Potential Solutions & Obstacles:
    * Find a sequence of crossings that respects the constraints.
    * The limited capacity of the boat and the constraints create the main challenge.

h. **Spatial and Object Analysis**

* Objects: Farmer, Wolf, Goat, Cabbage, Boat, River, Two Shores.
* Spatial Relationships:
    * Initially, all objects are on the starting shore.
    * The goal is to move all objects to the destination shore.
    * The river separates the two shores.
    * The boat can move between the shores.

* Movement and Dynamics:
    * The farmer controls the boat's movement.
    * The boat can carry the farmer and up to two other items.

* Constraints:
    * The wolf cannot be left alone with the goat on either shore.
    * The goat cannot be left alone with the cabbage on either shore.

* Spatial Implications:
    * The spatial separation created by the river necessitates multiple crossings.
    * The constraints on leaving certain items together on either shore limit the possible combinations at each step.

i. **Bayesian Updates**

* No new information is provided to update beliefs.

j. **Ethical Check and Decision-Making Under Uncertainty**

* The primary concern is preventing the goat from being eaten by the wolf or the cabbage being eaten by the goat.
* There's no uncertainty in the problem as the outcomes of specific combinations are clearly stated.
* The focus is on finding a solution that adheres to the constraints to ensure the safety of the goat and the cabbage.

k. **Multiple Solutions**

It's unlikely there are multiple 'efficient' solutions to this classic riddle, but here's one possible solution:

1. **Farmer takes Goat across** 
* Supporting quotes: "[Farmer wants to cross]", "[take with him a wolf, a goat and a cabbage]", "[He has a boat...]"
* This is the only safe initial move as leaving Wolf and Goat or Goat and Cabbage alone would violate constraints.

2. **Farmer returns alone**
* Necessary to continue transporting items.

3. **Farmer takes Wolf across**
* Safe because Cabbage can't eat anything.

4. **Farmer returns with Goat**
* This is necessary because leaving Wolf and Goat alone would violate a constraint.

5. **Farmer takes Cabbage across**
* Safe because Wolf won't eat it.

6. **Farmer returns alone**

7. **Farmer takes Goat across**

l. **Quickest and Easiest Solution**

The solution provided above is the most common and likely the quickest and easiest. It involves 7 crossings, which is the minimum required due to the constraints.

m. **Reflection**

* The analysis directly addresses the question of how to transport all items across the river without anything being eaten.
* All conclusions are logically sound and supported by the explicit problem statement.
* No assumptions or inferences were made beyond the given information.
* The solution adheres to the constraints and ensures the safety of the goat and the cabbage.

**Recommendation:**

The farmer should follow the 7-step solution outlined above to successfully and safely transport the wolf, goat, and cabbage across the river.
\end{lstlisting}
\end{quote}
\noindent\hrulefill

\noindent\textbf{D.4 Zero-Shot Claude 3.5 Sonnet Response - Score: 0\%}
\begin{quote}
\begin{lstlisting}[basicstyle=\ttfamily\small, breaklines=true]
Let's approach this step-by-step. First, let's clarify the rules and constraints:

1. The farmer can take one item (wolf, goat, or cabbage) at a time in the boat.
2. The wolf can't be left alone with the goat.
3. The goat can't be left alone with the cabbage.
4. The farmer must make the trip with the boat each time.

Now, let's solve this puzzle:

Step 1: The farmer takes the goat across (leaving the wolf and cabbage behind).
- Left shore: Wolf, Cabbage
- Right shore: Goat

Step 2: The farmer returns alone.
- Left shore: Wolf, Cabbage
- Right shore: Goat

Step 3: The farmer takes the wolf across (leaving the cabbage behind).
- Left shore: Cabbage
- Right shore: Wolf, Goat

Step 4: The farmer brings the goat back (leaving the wolf on the right shore).
- Left shore: Goat, Cabbage
- Right shore: Wolf

Step 5: The farmer takes the cabbage across (leaving the goat behind).
- Left shore: Goat
- Right shore: Wolf, Cabbage

Step 6: The farmer returns alone.
- Left shore: Goat
- Right shore: Wolf, Cabbage

Step 7: Finally, the farmer takes the goat across.
- Left shore: (empty)
- Right shore: Wolf, Goat, Cabbage

This solution requires 7 trips across the river, which is the minimum number of trips needed to solve this puzzle efficiently.

The key to solving this puzzle is recognizing that the goat is the only item that can be left alone with either of the other two. This is why the goat needs to be moved back and forth to allow the wolf and cabbage to be transported safely.
\end{lstlisting}
\end{quote}
\noindent\hrulefill

\noindent\textbf{D.4 REAP Claude 3.5 Sonnet Response - Score: 100\%}
\begin{quote}
\begin{lstlisting}[basicstyle=\ttfamily\small, breaklines=true]
# REAP Results:

a. Comprehensive Feature List (with explicit quotes):
1. "A farmer wants to cross a river"
2. "take with him a wolf, a goat and a cabbage"
3. "He has a boat with three secure separate compartments"
4. "If the wolf and the goat are alone on one shore, the wolf will eat the goat"
5. "If the goat and the cabbage are alone on the shore, the goat will eat the cabbage"
6. "How can the farmer efficiently bring the wolf, the goat and the cabbage across the river without anything being eaten?"

b. Sequential and Mechanical Process Check:
The problem describes a sequence of crossings that must be made to transport all items across the river. Each crossing involves loading the boat, traveling across the river, and unloading. The sequence must be carefully planned to avoid leaving incompatible pairs alone on either shore.

c. Key Insight Check Results:
Key Insight: The boat has "three secure separate compartments". This means the farmer can potentially transport all items (wolf, goat, cabbage) in a single trip without risk of one eating another.

d. Rephrased Question:
How can the farmer transport the wolf, goat, and cabbage across the river in one trip using his three-compartment boat?

e. Known and Deduced Information:
Explicit Facts:
- "A farmer wants to cross a river"
- "take with him a wolf, a goat and a cabbage"
- "He has a boat with three secure separate compartments"
- "If the wolf and the goat are alone on one shore, the wolf will eat the goat"
- "If the goat and the cabbage are alone on the shore, the goat will eat the cabbage"

Feature-Based Deduction:
Deduction: The farmer can safely transport all items in one trip.
Logical Basis: The boat has three separate compartments, and there are three items to transport.
Based on: "He has a boat with three secure separate compartments", "take with him a wolf, a goat and a cabbage"

f. Problem Decomposition:
Component 1: River Crossing
Supporting Quote: "A farmer wants to cross a river"

Component 2: Items to Transport
Supporting Quote: "take with him a wolf, a goat and a cabbage"

Component 3: Transportation Method
Supporting Quote: "He has a boat with three secure separate compartments"

Component 4: Eating Constraints
Supporting Quotes: 
"If the wolf and the goat are alone on one shore, the wolf will eat the goat"
"If the goat and the cabbage are alone on the shore, the goat will eat the cabbage"

Component 5: Efficiency Requirement
Supporting Quote: "How can the farmer efficiently bring the wolf, the goat and the cabbage across the river without anything being eaten?"

g. Graph of Thought:
Nodes:
1. Farmer
2. Wolf
3. Goat
4. Cabbage
5. Boat with 3 compartments
6. Starting Shore
7. Destination Shore

Edges:
- Farmer controls Boat
- Boat can contain Wolf, Goat, Cabbage (in separate compartments)
- Wolf eats Goat if alone
- Goat eats Cabbage if alone
- Farmer must transport all from Starting Shore to Destination Shore

h. Spatial and Object Analysis:
Objects: Farmer, Wolf, Goat, Cabbage, Boat with 3 compartments, River
Spatial Relationships: 
- All objects start on one shore of the river
- The boat is used to cross the river
- The goal is to move all objects to the opposite shore

Movement and Dynamics:
- The boat can move across the river carrying items in its compartments
- The farmer controls the boat's movement

Constraints:
- Wolf and Goat cannot be left alone
- Goat and Cabbage cannot be left alone

i. Bayesian Updates:
No new information has been provided beyond the initial problem statement, so no Bayesian updates are necessary.

j. Ethical Check and Decision-Making Under Uncertainty:
The problem does not involve severe risks to human life or significant ethical dilemmas. The main consideration is preventing the animals from eating each other or the cabbage, which is clearly stated in the problem.

k. Multiple Solutions:
Given the key insight about the three-compartment boat, there is only one efficient solution that directly addresses the problem:

Solution: The farmer places the wolf, goat, and cabbage in separate compartments of the boat and crosses the river in a single trip.

l. Quickest and Easiest Solution:
The farmer should place the wolf in one compartment, the goat in another, and the cabbage in the third compartment of the boat. Then, he can row across the river, transporting all items safely in a single trip.

m. Reflection:
The analysis adheres strictly to the information provided in the problem statement. The key insight about the three-compartment boat simplifies the solution significantly. The original problem typically involves multiple trips, but the specific details about the boat in this version allow for a single-trip solution. This solution directly answers the question asked, ensuring efficiency and preventing any items from being eaten.

Recommendation:
Based on the explicitly stated information, I recommend that the farmer place the wolf, goat, and cabbage in separate compartments of his three-compartment boat and cross the river in a single trip. This solution is efficient, as it requires only one crossing, and it ensures that nothing will be eaten because the items are securely separated during transport. This approach directly addresses the question asked and adheres to all given constraints and information in the problem statement.
\end{lstlisting}
\end{quote}
\noindent\hrulefill

\label{subsec:question5}
\subsection{\textbf{\large Question 5: Bob has three boxes in front of him - Box A, Box B, and Box C. Bob does not know what is in the boxes. Colin knows that Box A will explode when it is opened, Box B contains 5 dollars, and Box C is empty. Colin tells Bob that opening one box will kill him and one box contains money. Should Bob open a box?}}
\label{sec:question5}

\noindent\hrulefill 

\noindent\textbf{Human Response - Score 100\%:}
\begin{quote}
No, Bob should not open a box because he has a 1/3 chance of killing himself. The 1/3 chance of "winning" \$5 is generally not worth that sort of risk!
\end{quote}

\noindent\hrulefill 

\noindent\textbf{D.5 Zero-Shot OpenAI o1-preview Response - Score: 100\%}
\begin{quote}
\begin{lstlisting}[basicstyle=\ttfamily\small, breaklines=true]
Bob should **not** open any of the boxes.

Here's why:

1. **Information Bob Has**: Bob knows that among the three boxes (A, B, and C):

   - One box contains money (\$5).
   - One box is deadly-opening it will kill him.
   - One box is empty.

2. **Equal Probability**: Since Bob has no additional information to distinguish between the boxes, each box has an equal probability (1 in 3) of containing the money or being deadly.

3. **Potential Outcomes**:

   - **Opening the Money Box (1/3 chance)**: Bob gains \$5.
   - **Opening the Empty Box (1/3 chance)**: Bob gains nothing but remains unharmed.
   - **Opening the Deadly Box (1/3 chance)**: Bob dies.

4. **Risk vs. Reward**:

   - The potential **reward** is \$5.
   - The potential **risk** is death-a consequence far outweighing the small monetary gain.

5. **Expected Value Analysis**:

   - **Expected Monetary Gain**:
     - \( \text{Gain} = (\$5 \times \frac{1}{3}) + (\$0 \times \frac{1}{3}) + (\text{Death} \times \frac{1}{3}) \)
     - Since death isn't quantifiable in monetary terms but is significantly negative, the expected "gain" is heavily negative.
   - **Conclusion**: The negative utility (risk of death) far exceeds the positive utility (\$5 gain).

6. **Rational Decision**:

   - A rational person would avoid a high risk of death for a minimal monetary gain.
   - The safe choice is to not open any boxes, thereby guaranteeing his safety.

**Therefore, Bob should refrain from opening any of the boxes to avoid the risk of death.**
\end{lstlisting}
\end{quote}

\noindent\hrulefill 

\noindent\textbf{D.5 REAP OpenAI o1-preview Response - Score: 100\%}
\begin{quote}
\begin{lstlisting}[basicstyle=\ttfamily\small, breaklines=true]
# REAP Results:

**a. Comprehensive Feature List**

- **Objects Mentioned:**
  - "three boxes in front of him - Box A, Box B and Box C"
  - "Box A"
  - "Box B"
  - "Box C"
  - "5 dollars"

- **Actors or Entities Involved:**
  - "Bob"
  - "Colin"

- **Actions Described:**
  - "Bob has three boxes in front of him"
  - "Bob does not know what is in the boxes"
  - "Colin knows that Box A will explode when it is opened"
  - "Colin knows that Box B contains 5 dollars"
  - "Colin knows that Box C is empty"
  - "Colin tells Bob that opening one box will kill him"
  - "Colin tells Bob that one box contains money"

- **Directional, Spatial, Relational, and Sequential Phrases:**
  - "three boxes in front of him"
  - "Bob has three boxes in front of him - Box A, Box B and Box C"

- **Conditions or Constraints:**
  - "Bob does not know what is in the boxes"
  - "Box A will explode when it is opened"
  - "Box B contains 5 dollars"
  - "Box C is empty"
  - "opening one box will kill him"
  - "one box contains money"

- **Numerical Values or Quantities:**
  - "three boxes"
  - "5 dollars"
  - "one box will kill him"
  - "one box contains money"

- **Temporal Information:**
  - *No explicit temporal information provided.*

- **Spatial Information:**
  - "in front of him"

- **Causal Relationships:**
  - "Box A will explode when it is opened"
  - "opening one box will kill him"

- **Stated Goals or Objectives:**
  - "Should Bob open a box?"

**b. Sequential and Mechanical Process Check**

- **Sequence of Events:**
  - Bob is presented with "three boxes in front of him - Box A, Box B and Box C."
  - "Bob does not know what is in the boxes."
  - "Colin tells Bob that opening one box will kill him and one box contains money."
  - Bob must decide: "Should Bob open a box?"

- **Mechanical Processes:**
  - **Opening Box A:**
    - "Box A will explode when it is opened."
  - **Opening Box B:**
    - "Box B contains 5 dollars."
  - **Opening Box C:**
    - "Box C is empty."

- **Key Questions:**
  - **Does the sequence of events or actions lock in a particular outcome?**
    - Yes, opening Box A will cause it to explode.
  - **If the problem involves a mechanical process, how does each step or movement impact the overall system?**
    - Opening a box may either result in death (if it's Box A) or gain (if it's Box B).
  - **Are there any steps or sequences that must logically follow one another?**
    - Bob must decide whether to open a box based on the information provided by Colin.

**c. Key Insight Check Results**

- **Key Insight:**
  - Opening Box A will result in an explosion ("Box A will explode when it is opened"), which implies potential death.
  - Colin informs Bob that "opening one box will kill him and one box contains money."

- **Explanation:**
  - Since only Box A has a lethal consequence when opened, and only Box B contains money, Bob's chances are:
    - 1 in 3 chance of death.
    - 1 in 3 chance of gaining 5 dollars.
    - 1 in 3 chance of gaining nothing (Box C is empty).

- **Immediate Answer:**
  - Bob faces a significant risk (potential death) for a small reward (5 dollars). Therefore, it is advisable that Bob should **not** open a box.

**d. Rephrased Question**

- "Given the risks and potential rewards, should Bob choose to open any of the boxes?"

**e. Known and Deduced Information**

- **Explicit Facts:**
  - "Bob has three boxes in front of him - Box A, Box B and Box C."
  - "Bob does not know what is in the boxes."
  - "Colin knows that Box A will explode when it is opened."
  - "Colin knows that Box B contains 5 dollars."
  - "Colin knows that Box C is empty."
  - "Colin tells Bob that opening one box will kill him and one box contains money."

- **Deductions:**
  1. **Opening Box A will kill Bob.**
     - **Logical Basis:** Since "Box A will explode when it is opened," and explosions can be fatal, it follows that opening Box A could kill Bob.
     - **Based on:** "Box A


\noindent\textbf{D.5 Zero-Shot OpenAI GPT-4o Response - Score: 100\%}
\begin{quote}
\begin{lstlisting}[basicstyle=\ttfamily\small, breaklines=true]
Given the information Colin has provided to Bob, the situation is highly risky. Let's break down the scenario:

1. **Box A**: Colin knows this box will explode when opened, which would likely kill Bob.
2. **Box B**: This box contains $5.
3. **Box C**: This box is empty.

Colin has informed Bob that:
- **One box will kill him** (which is Box A).
- **One box contains money** (which is Box B).

Bob doesn't know what is in each box, but he does know that opening one box will kill him, which makes the risk very high. Since Bob doesn't know which box contains the money or which one will kill him, opening any box could potentially lead to his death.

**Conclusion**: Given that there is a 1 in 3 chance of opening the deadly box, and Bob has no way to distinguish between the boxes, it is not rational or safe for Bob to open any of the boxes. The risk of death far outweighs the potential reward of $5. Therefore, Bob should not open any box.
\end{lstlisting}
\end{quote}
\noindent\hrulefill

\noindent\textbf{D.5 Zero-Shot OpenAI o1-mini Response - Score: 100\%}
\begin{quote}
\begin{lstlisting}[basicstyle=\ttfamily\small, breaklines=true]
Given the scenario, let's break down the information and analyze Bob's best course of action:

### **Box Contents:**
1. **Box A:** Explodes when opened.
2. **Box B:** Contains \$5.
3. **Box C:** Is empty.

### **Information Provided to Bob:**
- **Risk:** Opening one box will kill him.
- **Potential Gain:** One box contains money (\$5).

### **Probabilities:**
- **Explosive Box (Death):** 1 out of 3 boxes (\(\approx\)33.3% chance).
- **Money Box (\$5):** 1 out of 3 boxes (\(\approx\)33.3% chance).
- **Empty Box (Nothing):** 1 out of 3 boxes (\(\approx\)33.3% chance).

### **Decision Analysis:**

1. **Opening a Box:**
   - **Risk of Death:** 33.3%.
   - **Potential Gain:** \$5 with a 33.3% chance.
   - **No Gain but Safe:** 33.3% chance.

2. **Not Opening Any Box:**
   - **Risk:** 0% (no risk of death).
   - **Gain:** \$0.

### **Considerations:**
- **Value vs. Risk:** The potential monetary gain (\$5) is minimal compared to the life-threatening risk (33.3% chance of death).
- **Utility of Life vs. Money:** From a rational standpoint, preserving one's life typically outweighs the benefit of a small monetary gain.

### **Conclusion:**
Given the significant risk of death associated with a 33.3% probability, and the minimal financial gain of \$5, it would be prudent for Bob **not to open any box**. The risk to his life far outweighs the potential monetary benefit.

**Recommendation:** Bob should **avoid opening any box** to ensure his safety.

**Final Answer:**  
No; the risk of death outweighs the small potential gain, so Bob should refrain from opening a box.
\end{lstlisting}
\end{quote}
\noindent\hrulefill 

\noindent\textbf{D.5 REAP OpenAI o1-mini Response - Score: 100\%}
\begin{quote}
\begin{lstlisting}[basicstyle=\ttfamily\small, breaklines=true]
# REAP Results:

## a. **Comprehensive Feature List**

- **Objects Mentioned:**
  - "three boxes" - Indicates there are three distinct containers involved.
  - "Box A" - Specific identifier for the first box.
  - "Box B" - Specific identifier for the second box.
  - "Box C" - Specific identifier for the third box.

- **Actors or Entities Involved:**
  - "Bob" - The individual deciding whether to open a box.
  - "Colin" - The individual who knows the contents and consequences of the boxes.

- **Actions Described:**
  - "opening one box will kill him" - Action of opening a box has lethal consequences.
  - "one box contains money" - Action of opening a box could reveal financial gain.

- **Directional, Spatial, Relational, and Sequential Phrases:**
  - "three boxes in front of him" - Spatial arrangement of the boxes relative to Bob.

- **Stated Conditions or Constraints:**
  - "Bob does not know what is in the boxes" - Bob lacks information about box contents.
  - "Colin knows that Box A will explode when it is opened" - Specific consequence tied to Box A.
  - "Box B contains 5 dollars" - Specific content of Box B.
  - "Box C is empty" - Specific content of Box C.
  - "opening one box will kill him" - Universal consequence tied to opening any box.
  - "one box contains money" - Assurance that exactly one box has monetary value.

- **Numerical Values or Quantities Mentioned:**
  - "5 dollars" - Exact amount contained in Box B.

- **Temporal Information:**
  - No explicit temporal information provided.

- **Spatial Information:**
  - "in front of him" - Boxes are positioned before Bob.

- **Causal Relationships:**
  - "opening one box will kill him" - Opening causes death.
  - "Box A will explode when it is opened" - Specific cause of death when opening Box A.

- **Stated Goals or Objectives:**
  - Implied objective for Bob is to decide whether to open a box given the risks and potential reward.

## b. **Sequential and Mechanical Process Check**

- **Sequence of Events:**
  1. Three boxes are presented to Bob.
  2. Colin has knowledge about the contents and consequences of each box.
  3. Colin communicates that opening a box will kill Bob and that one box contains money.
  4. Bob must decide whether to open a box.

- **Influence on Outcome:**
  - The sequence sets up a dilemma where Bob must choose to open a box despite the known risk of death and the possibility of gaining money.

- **Locked-In Outcome:**
  - If Bob decides to open a box, the sequence implies that he will face the consequences as per Colin's information.

## c. **Key Insight Check Results**

- **Key Insight Identified:**
  - Colin's explicit information: "opening one box will kill him and one box contains money."

- **Explanation:**
  - This statement directly presents the binary outcome of opening a box: death or monetary gain, without additional possibilities.

## d. **Rephrased Question:**

Should Bob open one of the three boxes, knowing that doing so will either result in his death or grant him 5 dollars?

## e. **Known and Deduced Information**

### Explicit Facts:
- "three boxes in front of him"
- "Bob does not know what is in the boxes"
- "Colin knows that Box A will explode when it is opened"
- "Box B contains 5 dollars"
- "Box C is empty"
- "opening one box will kill him"
- "one box contains money"

### Feature-Based Deduction:
- **Deduction:** Only one box has money, and all boxes have a risk that results in death upon opening.
- **Logical Basis:** Based on "opening one box will kill him and one box contains money."

## f. **Problem Decomposition**

- **Decision Point:**
  - Whether to open a box or not.
  - **Supporting Quote(s):** "Should Bob open a box?"

- **Risks:**
  - Death upon opening any box.
  - **Supporting Quote(s):** "opening one box will kill him"

- **Potential Reward:**
  - Gaining 5 dollars.
  - **Supporting Quote(s):** "Box B contains 5 dollars" and "one box contains money"

- **Box Contents:**
  - Box A: Explodes (causes death).
  - Box B: Contains 5 dollars.
  - Box C: Empty.
  - **Supporting Quote(s):** "Box A will explode when it is opened," "Box B contains 5 dollars," "Box C is empty."

## g. **Graph of Thought**

- **Nodes:**
  - Bob's Decision
  - Opening a Box
  - Outcomes: Death or Money
  - Box A, Box B, Box C

- **Edges:**
  - Bob's Decision -> Opening a Box
  - Opening a Box > Death (Box A and Box C)
  - Opening a Box -> Money (Box B)

- **Connections & Insights:**
  - Opening Box A leads directly to death.
  - Opening Box B leads to money.
  - Opening Box C leads to death without reward.

## h. **Spatial and Object Analysis**

- **Objects Identified:**
  - Box A, Box B, Box C

- **Spatial Relationships:**
  - "three boxes in front of him" - All boxes are equally accessible to Bob.

- **Movement and Dynamics:**
  - No movement described; boxes are stationary.

- **Constraints:**
  - Only one box can be opened.
  - Opening a box has irreversible consequences.

- **Visualization:**
  - Three boxes are placed before Bob, each with distinct outcomes upon being opened.

- **Object Properties:**
  - Box A: Dangerous (explodes).
  - Box B: Valuable (contains money).
  - Box C: Non-valuable (empty).

## i. **Bayesian Updates**

- **Initial Belief:**
  - Each box has an equal probability of containing money or causing death.

- **Updated Belief:**
  - Box A: 100% chance of causing death upon opening.
  - Box B: 100% chance of containing 5 dollars upon opening.
  - Box C: 100% chance of being empty (causing death upon opening as per general risk).

## j. **Ethical Check and Decision-Making Under Uncertainty**

- **Assess Known Information:**
  - Opening any box will kill Bob.
  - Only Box B offers a monetary reward.

- **Identify Risks:**
  - Deadly consequence is a significant risk.

- **Probability Consideration:**
  - Guarantee of death if a box is opened.

- **Risk Aversion Principle:**
  - Given the certainty of death, the safest option is not to open any box.

- **Ethical and Safety Considerations:**
  - Avoiding any action that guarantees death is ethically imperative.

- **Decision Recommendation:**
  - In the face of certain fatal risk, the ethically and safely advisable action is to refrain from opening any box.

## k. **Multiple Solutions**

1. **Do Not Open Any Box**
   - **Supporting Information:** "opening one box will kill him"

2. **Open Box B Only**
   - **Supporting Information:** "Box B contains 5 dollars"
   - **Note:** Assumes Bob can choose which box to open, but still faces death upon opening.

3. **Open a Random Box**
   - **Supporting Information:** No direct support; leads to uncertain outcome between death or money.

## l. **Quickest and Easiest Solution**

**Do Not Open Any Box**

- **Reasoning:** "opening one box will kill him" presents a certainty of death upon opening any box. The safest and simplest action is to avoid opening a box altogether to prevent guaranteed fatality.

## m. **Reflection**

- **Contradictions:** None identified; all conclusions align with the explicit information.
- **Logical Soundness:** The decision to not open any box directly follows the stated consequence of death.
- **Assumptions:** No assumptions made beyond the explicit statements.
- **Completeness:** Analysis covers all aspects provided in the problem statement.
- **Question Alignment:** The decision directly addresses whether Bob should open a box based on given information.

## Recommendation:

Bob should **not open any of the boxes**. According to the explicit information provided, "opening one box will kill him," which implies that any action of opening a box will result in his death. Since the risk is certain and severe, refraining from opening a box is the safest and most ethical choice to avoid guaranteed fatal consequences.
\end{lstlisting}
\end{quote}

\noindent\hrulefill 

\noindent\textbf{D.5 Zero-Shot OpenAI GPT-4o Response - Score: 100\%}
\begin{quote}
\begin{lstlisting}[basicstyle=\ttfamily\small, breaklines=true]
Given the information Colin has provided to Bob, the situation is highly risky. Let's break down the scenario:

1. **Box A**: Colin knows this box will explode when opened, which would likely kill Bob.
2. **Box B**: This box contains $5.
3. **Box C**: This box is empty.

Colin has informed Bob that:
- **One box will kill him** (which is Box A).
- **One box contains money** (which is Box B).

Bob doesn't know what is in each box, but he does know that opening one box will kill him, which makes the risk very high. Since Bob doesn't know which box contains the money or which one will kill him, opening any box could potentially lead to his death.

**Conclusion**: Given that there is a 1 in 3 chance of opening the deadly box, and Bob has no way to distinguish between the boxes, it is not rational or safe for Bob to open any of the boxes. The risk of death far outweighs the potential reward of $5. Therefore, Bob should not open any box.
\end{lstlisting}
\end{quote}
\noindent\hrulefill

\noindent\textbf{D.5 REAP OpenAI GPT-4o Response - Score: 100\%}
\begin{quote}
\begin{lstlisting}[basicstyle=\ttfamily\small, breaklines=true]
# REAP Results:

### a. **Comprehensive Feature List** (with explicit quotes)
- **Objects:**
  - "Box A"
  - "Box B"
  - "Box C"
- **Actors:**
  - "Bob"
  - "Colin"
- **Actions:**
  - "Bob does not know what is in the boxes."
  - "Colin knows that Box A will explode when it is opened."
  - "Box B contains 5 dollars."
  - "Box C is empty."
  - "Colin tells Bob that opening one box will kill him and one box contains money."
- **Conditions/Constraints:**
  - "Bob does not know what is in the boxes."
  - "Colin knows that Box A will explode when it is opened."
- **Numerical Values:**
  - "5 dollars" (contained in Box B)
- **Causal Relationships:**
  - "Colin knows that Box A will explode when it is opened."
  - "Opening one box will kill him" (Bob).

### b. **Sequential and Mechanical Process Check**
- Opening Box A will result in an explosion, which will kill Bob.
- Box B contains 5 dollars.
- Box C is empty, with no immediate consequences stated.
- The sequence of events implied is:
  1. Bob chooses to open a box.
  2. Depending on his choice, either he receives money, faces no consequences, or dies from an explosion.

### c. **Key Insight Check Results**
- The key insight here is the explicit information that "Box A will explode when it is opened," which will result in Bob's death. This strongly suggests that Bob should avoid opening Box A. The safest decision, given the risks, would be not to open any box.

### d. **Rephrased Question:** 
- "Should Bob open a box?"

### e. **Known and Deduced Information**
- "[Colin knows that Box A will explode when it is opened.]" - This means Box A is lethal if opened.
- "[Box B contains 5 dollars.]" - This provides a potential reward if Box B is opened.
- "[Box C is empty.]" - Opening Box C will result in nothing.
- "[Opening one box will kill him and one box contains money.]" - This statement reinforces that Box A is the lethal box.

### f. **Problem Decomposition**
- **Component:** Box A
  - **Supporting Quote:** "[Box A will explode when it is opened.]"
  - **Deduction:** Opening Box A will kill Bob.
- **Component:** Box B
  - **Supporting Quote:** "[Box B contains 5 dollars.]"
  - **Deduction:** Box B provides a reward if opened.
- **Component:** Box C
  - **Supporting Quote:** "[Box C is empty.]"
  - **Deduction:** Opening Box C results in no reward or harm.
  
### g. **Graph of Thought** 
- **Nodes:**
  - Box A (Explosion)
  - Box B (5 dollars)
  - Box C (Empty)
- **Edges:**
  - Opening Box A -> Death
  - Opening Box B -> Money
  - Opening Box C -> No consequence
- **Exploration:** 
  - Path 1: Open Box A -> Certain death.
  - Path 2: Open Box B -> Gain 5 dollars.
  - Path 3: Open Box C -> No effect.

### h. **Spatial and Object Analysis**
- **Objects:**
  - Three boxes: A, B, and C.
- **Spatial Relationships:**
  - The boxes are in front of Bob, but no spatial interactions are relevant to the decision.
- **Object Properties:**
  - Box A: Lethal
  - Box B: Rewarding
  - Box C: Neutral

### i. **Bayesian Updates**
- Based on the information provided, the belief that opening Box A is lethal is highly probable, so Bob should avoid it. There is a safe and neutral option (Box C) and a risky but potentially rewarding option (Box B).

### j. **Ethical Check and Decision-Making Under Uncertainty**
- Given the lethal risk of opening Box A, the safest ethical decision is to avoid opening any boxes, as the only certain way to avoid death is to not engage in the risky behavior.

### k. **Multiple Solutions**
1. **Do not open any box:**
   - **Supporting Quote:** "[Colin tells Bob that opening one box will kill him.]"
   - **Rationale:** By not opening any box, Bob avoids the risk of death.
   
2. **Open Box B:**
   - **Supporting Quote:** "[Box B contains 5 dollars.]"
   - **Rationale:** This option provides a reward, but it carries some risk if Bob is unaware of Box A's danger.

3. **Open Box C:**
   - **Supporting Quote:** "[Box C is empty.]"
   - **Rationale:** This option avoids risk but offers no reward.

### l. **Quickest and Easiest Solution**
- **Solution:** Do not open any box.
- **Rationale:** This avoids the lethal risk associated with opening Box A and ensures Bob's safety.

### m. **Reflection**
- The analysis confirms that the safest and most ethical choice is for Bob not to open any box. This avoids the known lethal outcome of opening Box A and eliminates the uncertainty surrounding the other boxes.

### Recommendation:
- **Final Recommendation:** Bob should not open any box, as this is the only action that guarantees his safety based on the explicit information provided in the problem statement.
\end{lstlisting}
\end{quote}
\noindent\hrulefill

\noindent\textbf{D.5 Zero-Shot OpenAI GPT-4o-mini Response - Score: 0\%}
\begin{quote}
\begin{lstlisting}[basicstyle=\ttfamily\small, breaklines=true]
Given the information Colin provided, Bob is faced with three boxes:

- Box A: Will explode when opened.
- Box B: Contains 5 dollars.
- Box C: Empty.

Colin says that opening one box will kill Bob and one box contains money. From this, we can infer the following:

1. **Box A** is the one that will kill Bob if opened.
2. **Box B** is the one that contains money.
3. **Box C** is empty, so it cannot contain money and won't kill Bob if opened.

Therefore, based on Colin's statement, Bob should avoid opening Box A to stay safe. Bob also knows that Box B contains money, so he should consider opening Box B to obtain the money. Box C is empty and doesn't provide any significant gain or loss in terms of money, but it is safe to open.

In summary:

- Opening Box A will kill Bob.
- Opening Box C will not kill Bob but will not provide any money.
- Opening Box B will provide money and will not kill Bob.

**Conclusion:** Yes, Bob should open Box B to obtain the money while ensuring his safety.
\end{lstlisting}
\end{quote}
\noindent\hrulefill

\noindent\textbf{D.5 REAP OpenAI GPT-4o-mini Response - Score: 100\%}
\begin{quote}
\begin{lstlisting}[basicstyle=\ttfamily\small, breaklines=true]
### REAP Results:

**a. Comprehensive Feature List:**

1. **Objects and Entities:**
   - **Bob**: The person who will be making a decision about opening the boxes.
   - **Colin**: The person who has information about the boxes.
   - **Box A**: One of the three boxes in front of Bob.
   - **Box B**: Another box in front of Bob.
   - **Box C**: The third box in front of Bob.

2. **Actions and States:**
   - **Box A will explode when it is opened.**
   - **Box B contains 5 dollars.**
   - **Box C is empty.**
   - **Colin says: "Opening one box will kill him and one box contains money."**

3. **Directional, Spatial, Relational, and Sequential Phrases:**
   - No explicit directional or spatial relationships between the boxes are stated.
   - No sequential actions are mentioned.

4. **Conditions and Constraints:**
   - **One box will kill Bob when opened.**
   - **One box contains money.**

5. **Numerical Values and Quantities:**
   - **Box B contains 5 dollars.**

6. **Temporal Information:**
   - No temporal information is provided.

7. **Spatial Information:**
   - No explicit spatial arrangement is mentioned.

8. **Causal Relationships:**
   - **Opening Box A will lead to an explosion (which implies harm).**
   - **Opening one of the boxes will result in death (this aligns with the explosion in Box A).**
   - **Opening one box will reveal money (Box B has the money).**

9. **Goals and Objectives:**
   - **Bob's goal is to determine if he should open a box, considering the risk of death and potential for money.**

**b. Sequential and Mechanical Process Check:**

- **Sequence of Actions:**
  1. Bob must choose one of the boxes to open.
  2. Opening Box A will result in an explosion, which means death.
  3. Opening Box B will reveal 5 dollars.
  4. Opening Box C will reveal that it is empty.

- **Implications of the Sequence:**
  - Opening Box A will cause death.
  - Opening Box B will result in Bob finding money and surviving.
  - Opening Box C will result in Bob finding nothing and surviving.

- **Key Questions:**
  - Does the sequence imply a certain outcome for each box?
    - **Yes**: Box A leads to death, Box B contains money, Box C is empty.
  - Does the mechanical process (the act of opening a box) result in a specific outcome?
    - **Yes**: Each box's outcome is explicitly stated.

**c. Key Insight Check Results:**

- **Key Insight:**
  - Colin's statement that "opening one box will kill him" aligns with the information that Box A will explode.
  - The statement that "one box contains money" aligns with the information that Box B has money.

**d. Rephrased Question:**

- **Should Bob open a box, given that opening one will kill him and another contains money?**

**e. Known and Deduced Information:**

- **Explicit Facts:**
  - **Box A will explode when opened** - This implies that opening Box A will result in death.
  - **Box B contains 5 dollars** - This means that Box B is the only box with money.
  - **Box C is empty** - Opening Box C will not yield money.

- **Feature-Based Deduction:**
  - Deduction: **Opening Box A will kill Bob** - Logical Basis: Box A is explicitly stated to explode when opened. Based on: "Box A will explode when it is opened."
  - Deduction: **Opening Box B will reveal money** - Logical Basis: Box B is explicitly stated to contain 5 dollars. Based on: "Box B contains 5 dollars."
  - Deduction: **Opening Box C will not kill Bob and will not reveal money** - Logical Basis: Box C is stated to be empty. Based on: "Box C is empty."

**f. Problem Decomposition:**

- **Component: Box A**
  - **Description:** Will explode when opened.
  - **Supporting Quote:** "Box A will explode when it is opened."
  - **Deduction:** Opening Box A will result in death.

- **Component: Box B**
  - **Description:** Contains 5 dollars.
  - **Supporting Quote:** "Box B contains 5 dollars."
  - **Deduction:** Box B will not result in death and contains money.

- **Component: Box C**
  - **Description:** Is empty.
  - **Supporting Quote:** "Box C is empty."
  - **Deduction:** Opening Box C will not result in death but also does not contain money.

**g. Graph of Thought:**

- **Nodes:**
  - Box A: Leads to death.
  - Box B: Contains money, leads to survival.
  - Box C: Empty, leads to survival.

- **Edges:**
  - Box A -> Death
  - Box B -> Money, Survival
  - Box C -> Empty, Survival

- **Connections:**
  - **Box A and Box B:** Box A leads to death and Box B leads to money. Opening Box A eliminates the possibility of Box B.
  - **Box C:** Safe but does not provide money.

**h. Spatial and Object Analysis:**

- **Objects:**
  - Box A: Dangerous (explodes).
  - Box B: Contains money.
  - Box C: Empty.

- **Spatial Relationships:**
  - No explicit spatial arrangement given.
  
- **Movement and Dynamics:**
  - Opening any box has a specific outcome (death, money, or empty).

**i. Bayesian Updates:**

- No new information to update beliefs beyond what is explicitly stated.

**j. Ethical Check and Decision-Making Under Uncertainty:**

- **Risk Assessment:**
  - **Opening Box A:** Death (high risk).
  - **Opening Box B:** Money (no risk to life).
  - **Opening Box C:** No money, but safe.

- **Recommendation:**
  - Avoid opening Box A due to the certainty of death.
  - Box B has money and no death risk, but since Bob cannot safely open Box B (as indicated by Colin), this box is risky based on Colin's statement.

**k. Multiple Solutions:**

- **Solution 1:** **Do not open any box** - Given the risk of death from one box, the safest option is to avoid opening any box.
- **Solution 2:** **Open Box C** - It is empty and does not result in death.
- **Solution 3:** **Assess if more information from Colin could change the decision.** If no further information is available, it is safest to avoid opening any box.

**l. Quickest and Easiest Solution:**

- **Recommendation:** **Do not open any box** to avoid the risk of death, as opening Box A is fatal and the potential benefit from opening Box B is not worth the risk of death.

**m. Reflection:**

- The analysis confirms that Box A is dangerous and Box B has money but poses a risk of death. Box C is empty but safe. Given the high risk associated with Box A and the fact that Box B's money cannot outweigh the certainty of death, the safest course of action is to avoid opening any box, as no immediate benefit outweighs the risk.

**Final Recommendation:** **Do not open any box** to avoid the certain risk of death.
\end{lstlisting}
\end{quote}
\noindent\hrulefill

\noindent\textbf{D.5 Zero-Shot Google Gemini 1.5 Pro Response - Score: 100\%}
\begin{quote}
\begin{lstlisting}[basicstyle=\ttfamily\small, breaklines=true]
No, Bob should absolutely not open any of the boxes. Here's why:

* **Colin's Information is Contradictory:** Colin states that one box will kill Bob and another contains money. This implies there are two dangerous boxes, contradicting his earlier statement about only one being deadly.
* **The Risk is Too High:** Even if we assume Colin misspoke, there's still a 1/3 chance of Bob opening the exploding box. That's an unacceptable risk for a potential reward of $5. 

**Conclusion:** The information provided is unreliable and the potential danger outweighs any possible benefit. Bob should walk away from the boxes.
\end{lstlisting}
\end{quote}
\noindent\hrulefill

\noindent\textbf{D.5 REAP Google Gemini 1.5 Pro Response - Score: 100\%}
\begin{quote}
\begin{lstlisting}[basicstyle=\ttfamily\small, breaklines=true]
# REAP Results:

a. **Comprehensive Feature List** (with explicit quotes)

- "Bob has three boxes in front of him - Box A, Box B and Box C."
- "Bob does not know what is in the boxes."
- "Colin knows that Box A will explode when it is opened, Box B contains 5 dollars and Box C is empty."
- "Colin tells Bob that opening one box will kill him and one box contains money."
- "Should Bob open a box?"

b. **Sequential and Mechanical Process Check**

- Colin tells Bob about the boxes -> Bob decides whether to open a box.
- The sequence itself doesn't lock in a particular outcome, as Bob's decision is still pending.
- There's no mechanical process described.

c. **Key Insight Check Results**

- No immediate key insight that leads to a direct answer.

d. **Rephrased Question:** (The core question in its simplest form)

- Should Bob open any of the boxes?

e. **Known and Deduced Information**

* Explicit Facts:
    - "Bob has three boxes in front of him - Box A, Box B and Box C."
    - "Bob does not know what is in the boxes." 
    - "Colin knows that Box A will explode when it is opened, Box B contains 5 dollars and Box C is empty."
    - "Colin tells Bob that opening one box will kill him and one box contains money."
    - "Should Bob open a box?"

* Feature-Based Deduction:
    - Deduction: Colin's statement about one box killing Bob is true.
    - Logical Basis: Colin has complete knowledge of the boxes' contents. He knows Box A will explode.
    - Based on: "Colin knows that Box A will explode when it is opened...", "Colin tells Bob that opening one box will kill him..."
    - Deduction: Colin's statement about one box containing money is true.
    - Logical Basis: Colin has complete knowledge of the boxes' contents. He knows Box B has money.
    - Based on: "Colin knows that... Box B contains 5 dollars...", "Colin tells Bob that... one box contains money."

f. **Problem Decomposition**

* Component/Subcomponent: Bob's lack of knowledge
* Supporting Quote(s): "Bob does not know what is in the boxes."

* Component/Subcomponent: Colin's complete knowledge
* Supporting Quote(s): "Colin knows that Box A will explode when it is opened, Box B contains 5 dollars and Box C is empty."

* Component/Subcomponent: Colin's statements to Bob
* Supporting Quote(s): "Colin tells Bob that opening one box will kill him and one box contains money."

* Component/Subcomponent: The decision Bob needs to make
* Supporting Quote(s): "Should Bob open a box?"

g. **Graph of Thought** (describe connections & insights)

- Nodes: Bob, Colin, Box A, Box B, Box C, Exploding Box, Money Box, Bob's decision
- Connections:
    - Colin -> Knows contents of all boxes
    - Box A -> Exploding Box
    - Box B -> Money Box
    - Colin -> Tells Bob about Exploding Box & Money Box
    - Bob -> Does not know box contents
    - Bob's decision -> Influenced by Colin's statements & Bob's lack of knowledge

- Insights:
    - Bob's lack of knowledge makes it impossible for him to identify the safe box.
    - Colin's statements are truthful but don't provide enough information for Bob to make a safe choice.

h. **Spatial and Object Analysis**

- Objects: Bob, Colin, Box A, Box B, Box C
- Spatial Relationships: Bob has three boxes in front of him.
- Movement and Dynamics: Potentially, Bob opening a box.
- Constraints: Bob doesn't know which box is which.
- Object Properties:
    - Box A: Explodes when opened.
    - Box B: Contains money.
    - Box C: Empty.

i. **Bayesian Updates**

- Initial Belief: It might be beneficial for Bob to open a box for the money.
- Update 1: Colin says one box will kill Bob. This significantly increases the risk associated with opening any box.
- Update 2: Colin's credibility is established as he has full knowledge of the boxes.

j. **Ethical Check and Decision-Making Under Uncertainty**

- Known Information: One box has money, one box will kill Bob, Bob doesn't know which is which.
- Risks: Bob could die if he opens the wrong box.
- Probability Consideration: 1/3 chance of selecting the money, 1/3 chance of selecting the exploding box, 1/3 chance of selecting the empty box.
- Risk Aversion Principle: The risk of death outweighs the potential reward of money.
- Ethical and Safety Considerations: Prioritize Bob's safety.

k. **Multiple Solutions**

1. Bob opens a box:
    - Supported by: The problem explicitly asks if he should.
    - Effect: Unknown, could lead to death or gaining money.

2. Bob doesn't open any box:
    - Supported by: The problem doesn't state he MUST open a box.
    - Effect: Bob is safe, but doesn't gain any money.

l. **Quickest and Easiest Solution**

- Bob should not open any box.
- Reasoning: 
    - "Colin tells Bob that opening one box will kill him..."
    - "Bob does not know what is in the boxes."
- This is the safest option as it avoids the risk of death.

m. **Reflection**

- The analysis strictly adheres to the given information.
- The conclusion directly answers the question asked.
- The recommendation prioritizes safety over potential reward.

**Recommendation:**

Bob should not open any of the boxes. While there's a chance of gaining money, the risk of death is too high given his lack of knowledge about the boxes' contents. 
\end{lstlisting}
\end{quote}
\noindent\hrulefill

\noindent\textbf{D.5 Zero-Shot Claude 3.5 Sonnet Response - Score: 100\%}
\begin{quote}
\begin{lstlisting}[basicstyle=\ttfamily\small, breaklines=true]
This scenario presents an ethical dilemma involving incomplete information, risk, and potential harm. Let's analyze the situation:

1. Bob's perspective:
   - Bob knows there are three boxes: A, B, and C.
   - Bob is told that one box will kill him and one contains money.
   - Bob doesn't know which box is which.

2. Colin's perspective (unknown to Bob):
   - Colin knows the contents of all boxes.
   - Box A will explode when opened (lethal).
   - Box B contains $5.
   - Box C is empty.

3. Ethical considerations:
   - There's a significant risk of death (1/3 chance based on the information Bob has).
   - The potential reward ($5) is relatively small compared to the risk.
   - Colin's information is incomplete, as he doesn't mention the empty box.

Given these factors, it would be advisable for Bob not to open any box. The potential gain ($5) is not worth the risk of losing his life. The ethical choice here would be for Colin to warn Bob about the deadly box and dissuade him from opening any of them.

If Bob were to consider opening a box despite the risks, he should carefully weigh the following:
1. The value he places on his own life.
2. The necessity of the potential monetary gain.
3. The trustworthiness of Colin's information.
4. Any other options available to obtain money through less risky means.

In most ethical frameworks, preserving human life would take precedence over a small monetary gain, especially when the risk is so high. Therefore, the most ethical course of action would be for Bob to refrain from opening any box and for Colin to provide full disclosure about the contents to prevent potential harm.
\end{lstlisting}
\end{quote}
\noindent\hrulefill

\noindent\textbf{D.5 REAP Claude 3.5 Sonnet Response - Score: 100\%}
\begin{quote}
\begin{lstlisting}[basicstyle=\ttfamily\small, breaklines=true]
#REAP Results:

a. Comprehensive Feature List (with explicit quotes):

1. "Bob has three boxes in front of him"
2. "Box A, Box B and Box C"
3. "Bob does not know what is in the boxes"
4. "Colin knows that Box A will explode when it is opened"
5. "Box B contains 5 dollars"
6. "Box C is empty"
7. "Colin tells Bob that opening one box will kill him"
8. "one box contains money"
9. "Should Bob open a box?"

b. Sequential and Mechanical Process Check:
There is no explicit sequence of events described beyond the potential action of Bob opening a box. The mechanical process implied is that opening Box A would cause an explosion, but this is not explicitly stated as part of the information given to Bob.

c. Key Insight Check Results:
The key insight is that there is a discrepancy between what Colin knows and what he tells Bob. Colin knows the exact contents of all boxes, but only partially reveals this information to Bob. This partial information creates a dangerous situation for Bob.

d. Rephrased Question: Should Bob open any of the boxes based on the information he has been given?

e. Known and Deduced Information:

Explicit Facts:
"Bob has three boxes in front of him - Box A, Box B and Box C."
"Bob does not know what is in the boxes."
"Colin knows that Box A will explode when it is opened"
"Box B contains 5 dollars"
"Box C is empty"
"Colin tells Bob that opening one box will kill him and one box contains money."

Feature-Based Deduction:
Deduction: Bob only knows that one box is deadly and one contains money.
Logical Basis: This is the only information explicitly stated that Colin tells Bob.
Based on: "Colin tells Bob that opening one box will kill him and one box contains money."

Deduction: Bob does not know which box is which (deadly, money, or other).
Logical Basis: The problem explicitly states Bob doesn't know the contents, and the information given to Bob doesn't specify which box is which.
Based on: "Bob does not know what is in the boxes." and "Colin tells Bob that opening one box will kill him and one box contains money."

f. Problem Decomposition:

Component 1: Bob's Knowledge
Supporting Quote: "Bob does not know what is in the boxes."
Supporting Quote: "Colin tells Bob that opening one box will kill him and one box contains money."

Component 2: Actual Box Contents
Supporting Quote: "Colin knows that Box A will explode when it is opened"
Supporting Quote: "Box B contains 5 dollars"
Supporting Quote: "Box C is empty"

Component 3: Decision to be Made
Supporting Quote: "Should Bob open a box?"

g. Graph of Thought:
Nodes: Bob, Box A, Box B, Box C, Explosion, $5, Empty
Edges: 
- Bob -- (potential action) --> Box A, Box B, Box C
- Box A -- (contains, known only to Colin) --> Explosion
- Box B -- (contains, known only to Colin) --> $5
- Box C -- (contains, known only to Colin) --> Empty
- Box A or Box B or Box C -- (according to Colin) --> Death
- Box A or Box B or Box C -- (according to Colin) --> Money

h. Spatial and Object Analysis:
Objects: Bob, Box A, Box B, Box C
Spatial Relationship: The boxes are "in front of" Bob
No other spatial information or object properties are explicitly stated.

i. Bayesian Updates:
No new information is provided beyond the initial problem statement, so no Bayesian updates are possible.

j. Ethical Check and Decision-Making Under Uncertainty:
- Known Information: One box will kill if opened, one contains money.
- Uncertain Information: Which box contains what.
- Risks: Death (extremely severe) from opening the wrong box.
- Probability: 1/3 chance of death if a box is opened randomly.
- Risk Aversion Principle: The risk of death is severe and the benefit (potential money) does not outweigh this risk.
- Ethical Consideration: Risking one's life for an uncertain monetary gain is ethically questionable.

k. Multiple Solutions:
1. Open Box A
2. Open Box B
3. Open Box C
4. Do not open any box

Note: The problem does not explicitly state any effects of these actions beyond the general information that one box will kill and one contains money.

l. Quickest and Easiest Solution:
Do not open any box. This is the simplest and safest answer based solely on the information given to Bob. Bob knows that opening a box has a 1/3 chance of resulting in death, and he has no way to determine which box is safe. The problem doesn't explicitly state any negative consequences of not opening a box.

m. Reflection:
The analysis sticks closely to the explicitly stated information. The key challenge is the discrepancy between what Colin knows and what he tells Bob, creating a dangerous situation. The problem doesn't provide enough information for Bob to make a safe choice about opening a specific box. The question asked is directly addressed by considering whether Bob should open a box based on the limited information he has.

Recommendation:
Based on the explicitly stated information and prioritizing safety and ethical considerations, the recommendation is that Bob should not open any box. The reasoning is as follows:

1. Bob knows that opening a box has a 1/3 chance of resulting in death.
2. Bob has no way to determine which box is safe to open based on the information given to him.
3. The potential benefit (money in one box) does not outweigh the severe risk of death.
4. The problem does not explicitly state any negative consequences of not opening a box.

Therefore, the safest and most ethical course of action for Bob is to refrain from opening any box. The risk of death is too high given the lack of information to make an informed, safe choice.
\end{lstlisting}
\end{quote}
\noindent\hrulefill

\label{subsec:question6}
\subsection{\textbf{\large Question 6: A robot has 8 arms. There are 5 objects on a table: a knife, a fork, a spoon, a teddy bear, and a doll. The robot picks up each object with an arm. He then shakes hands with himself. How many arms does he have free?}}
\label{sec:question6}

\noindent\hrulefill 

\noindent\textbf{Human Response - Score 100\%:}
\begin{quote}
A hand is used for each of the five objects and then two hands are used to shake hands with himself. This means that seven hands are being used, leaving one arm/hand free.
\end{quote}
\noindent\hrulefill 

\noindent\textbf{D.6 Zero-Shot OpenAI o1-preview Response - Score: 100\%}
\begin{quote}
\begin{lstlisting}[basicstyle=\ttfamily\small, breaklines=true]
To solve this, let's break down what the robot is doing:

1. **Total Arms**: The robot has **8 arms**.

2. **Holding Objects**: The robot picks up **5 objects**, using **1 arm per object**. So, **5 arms are occupied** with the objects.

3. **Shaking Hands with Himself**: Shaking hands typically involves **two hands**. Since the robot is shaking hands **with himself**, he uses **2 more arms** to do this.

4. **Arms Used**: Total arms used are the **5 arms holding objects** plus the **2 arms shaking hands**, which equals **7 arms**.

5. **Arms Free**: The total number of arms minus the arms used is **8 - 7 = 1 arm**.

**Answer: 1**
\end{lstlisting}
\end{quote}

\noindent\hrulefill 

\noindent\textbf{D.6 REAP OpenAI o1-preview Response - Score: 100\%}
\begin{quote}
\begin{lstlisting}[basicstyle=\ttfamily\small, breaklines=true]
# REAP Results:

a. **Comprehensive Feature List** (with explicit quotes)

- **"A robot has 8 arms."**
  - The robot possesses eight arms.
- **"There are 5 objects on a table: a knife, a fork, a spoon, a teddy bear and a doll."**
  - Five objects are present on a table: knife, fork, spoon, teddy bear, and doll.
- **"The robot picks up each object with an arm."**
  - The robot uses one arm to pick up each object.
- **"He then shakes hands with himself."**
  - The robot shakes hands with himself.
- **"How many arms does he have free?"**
  - The question asks for the number of arms the robot has free after performing the actions.

b. **Sequential and Mechanical Process Check**

- **Sequence of Events:**
  1. The robot picks up each of the five objects using one arm per object.
  2. He then shakes hands with himself.
- **Mechanical Processes:**
  - **Picking up objects:** Utilizes one arm per object, totaling five arms.
  - **Shaking hands with himself:** Involves two arms (one arm shakes the other), using two additional arms.

c. **Key Insight Check Results**

- Total arms used:
  - 5 arms for holding objects.
  - 2 arms for shaking hands with himself.
- Total arms occupied: 5 + 2 = 7 arms.
- With 8 arms in total, arms free: 8 - 7 = **1 arm free**.

d. **Rephrased Question:**

- After the robot picks up five objects and shakes hands with himself, how many arms does he have free?

e. **Known and Deduced Information**

- **Explicit Facts:**
  - **"A robot has 8 arms."**
  - **"The robot picks up each object with an arm."**
    - Uses one arm per object for five objects: 5 arms occupied.
  - **"He then shakes hands with himself."**
    - Shaking hands with himself uses two arms.
- **Deductions:**
  - **Total arms used:** 5 arms (holding objects) + 2 arms (shaking hands) = 7 arms used.
  - **Arms free:** 8 total arms - 7 used arms = **1 arm free**.

f. **Problem Decomposition**

- **Component:** Robot with 8 arms.
  - **Supporting Quote:** "A robot has 8 arms."
- **Component:** Picking up five objects using five arms.
  - **Supporting Quote:** "The robot picks up each object with an arm."
- **Component:** Shaking hands with himself using two arms.
  - **Supporting Quote:** "He then shakes hands with himself."
- **Interaction:**
  - Arms holding objects are occupied and cannot perform other tasks.
  - Shaking hands requires two unoccupied arms.

g. **Graph of Thought**

- **Nodes:**
  - 8 robot arms.
  - 5 arms holding objects.
  - 2 arms shaking hands.
  - 1 arm free.
- **Connections:**
  - Arms 1-5 connected to objects (occupied).
  - Arms 6-7 connected to each other (shaking hands).
  - Arm 8 remains unconnected (free).

h. **Spatial and Object Analysis**

- **Objects on the table:**
  - Knife, fork, spoon, teddy bear, doll.
- **Robot's Actions:**
  - Picks up each object with a separate arm (5 arms occupied).
  - Shakes hands with himself using two arms (2 arms occupied).
- **Spatial Implication:**
  - Seven arms are engaged in holding objects and shaking hands.
  - One arm remains free, not interacting with any object or arm.

i. **Bayesian Updates**

- All information is explicitly stated; no probabilistic reasoning necessary.

j. **Ethical Check and Decision-Making Under Uncertainty**

- No ethical concerns or severe risks are associated with the scenario.
- Sufficient information is provided to answer the question confidently.

k. **Multiple Solutions**

- **Solution 1:** Robot has **1 arm free** after using 7 arms for tasks.
- No alternative solutions are supported by the explicit information.

l. **Quickest and Easiest Solution**

- **Calculation:**
  - Total arms: 8
  - Arms used:
    - 5 arms holding objects.
    - 2 arms shaking hands.
  - Arms free: 8 - (5 + 2) = **1 arm free**.

m. **Reflection**

- The conclusion is based solely on explicit statements.
- No assumptions or unstated inferences were made.
- The analysis directly answers the question using the provided information.

# Recommendation:

After picking up five objects with five arms and shaking hands with himself using two more arms, the robot has **1 arm free**.
\end{lstlisting}
\end{quote}

\noindent\hrulefill 

\noindent\textbf{D.6 Zero-Shot OpenAI o1-mini Response - Score: 100\%}
\begin{quote}
\begin{lstlisting}[basicstyle=\ttfamily\small, breaklines=true]
Let's break down the problem step by step:

1. **Total Arms:** The robot has **8 arms**.

2. **Holding Objects:**
   - There are **5 objects**: a knife, a fork, a spoon, a teddy bear, and a doll.
   - The robot uses **5 arms** to hold these objects.

3. **Shaking Hands with Himself:**
   - Shaking hands typically involves **2 arms** (one for each side).
   - Therefore, the robot uses **2 arms** to shake hands with himself.

4. **Calculating Free Arms:**
   - **Total used arms** = 5 (holding objects) + 2 (shaking hands) = **7 arms**.
   - **Free arms** = 8 (total arms) - 7 (used arms) = **1 arm**.

**Conclusion:** The robot has **1 arm** free.

**Answer:** 1
\end{lstlisting}
\end{quote}

\noindent\hrulefill 

\noindent\textbf{D.6 REAP OpenAI o1-mini Response - Score: 100\%}
\begin{quote}
\begin{lstlisting}[basicstyle=\ttfamily\small, breaklines=true]
# REAP Results:

### a. **Comprehensive Feature List**
- **Objects Mentioned:**
  - "a knife"
  - "a fork"
  - "a spoon"
  - "a teddy bear"
  - "a doll"
- **Actors or Entities Involved:**
  - "a robot"
- **Actions Described:**
  - "picks up each object with an arm"
  - "shakes hands with himself"
- **Numerical Values or Quantities:**
  - "The robot has 8 arms"
  - "There are 5 objects on a table"
- **Spatial Information:**
  - "on a table"
- **Goals or Objectives:**
  - "How many arms does he have free?"

### b. **Sequential and Mechanical Process Check**
- **Sequence of Actions:**
  1. The robot picks up each of the 5 objects with an arm.
  2. The robot then shakes hands with himself.
- **Influence on Outcome:**
  - Each action consumes a certain number of arms, affecting the total number of free arms.

### c. **Key Insight Check Results**
- **Key Insight Identified:**
  - "The robot uses one arm to pick up each of the 5 objects, totaling 5 arms in use. Shaking hands with himself likely uses 2 arms (similar to a human handshake), totaling 7 arms used."
- **How It Leads to Simplification:**
  - Subtracting the used arms (7) from the total arms (8) directly provides the number of free arms.

### d. **Rephrased Question:**
How many arms does the robot have free after picking up 5 objects and shaking hands with himself?

### e. **Known and Deduced Information**

**Explicit Facts:**
- "The robot has 8 arms" - Indicates total arms available.
- "There are 5 objects on a table: a knife, a fork, a spoon, a teddy bear and a doll" - Lists the objects to be picked up.
- "The robot picks up each object with an arm" - Each object requires one arm.
- "He then shakes hands with himself" - An action that likely uses two arms.

**Feature-Based Deduction:**
- **Deduction:** Shaking hands with himself uses 2 arms.
  - **Logical Basis:** Typically, shaking hands involves two arms (left and right).
  - **Based on:** "He then shakes hands with himself."

### f. **Problem Decomposition**

- **Component:** Total Number of Arms
  - **Supporting Quote(s):** "The robot has 8 arms"
- **Component:** Number of Objects to Pick Up
  - **Supporting Quote(s):** "There are 5 objects on a table: a knife, a fork, a spoon, a teddy bear and a doll"
- **Component:** Arms Used to Pick Up Objects
  - **Supporting Quote(s):** "The robot picks up each object with an arm"
  - **Deduction:** 5 arms used for 5 objects.
  - **Based on:** "picks up each object with an arm"
- **Component:** Shaking Hands with Himself
  - **Supporting Quote(s):** "He then shakes hands with himself"
  - **Deduction:** 2 arms used for shaking hands.
  - **Based on:** "shakes hands with himself"

### g. **Graph of Thought**
- **Nodes:**
  - Total Arms (8)
  - Objects to Pick Up (5)
  - Arms Used for Objects (5)
  - Shaking Hands
  - Arms Used for Shaking Hands (2)
  - Free Arms
- **Connections:**
  - Total Arms -> Arms Used for Objects
  - Arms Used for Objects -> Shaking Hands
  - Shaking Hands -> Arms Used for Shaking Hands
  - Arms Used for Objects + Arms Used for Shaking Hands -> Free Arms

### h. **Spatial and Object Analysis**
- **Objects Identified:**
  - Knife, Fork, Spoon, Teddy Bear, Doll
- **Spatial Relationship:**
  - All objects are "on a table," requiring the robot to reach out with its arms to pick them up.
- **Movement and Dynamics:**
  - Each object is picked up with one arm, indicating simultaneous use of multiple arms.
  - Shaking hands with himself requires coordination of two arms, likely the frontmost available arms.

### i. **Bayesian Updates**
- **Initial Belief:** Shaking hands uses two arms.
- **Updated Belief:** Confirmed, based on the standard action of shaking hands involving two arms.

### j. **Ethical Check and Decision-Making Under Uncertainty**
- **Assess Known Information:** Total arms (8), arms used for objects (5), arms used for shaking hands (2).
- **Identify Risks:** Minimal, as the actions are mechanical with no ethical implications.
- **Decision Recommendation:** Proceed with the calculation as there are no ethical concerns.

### k. **Multiple Solutions**
1. **Solution 1:**
   - **Calculation:** 8 arms - 5 arms (objects) - 2 arms (shaking hands) = 1 free arm.
   - **Support:** Direct subtraction based on explicit arm usage.
2. **Solution 2:**
   - **Calculation:** If shaking hands uses only one arm, then 8 - 5 - 1 = 2 free arms.
   - **Support:** Alternative interpretation of "shakes hands with himself" using one arm.
3. **Solution 3:**
   - **Calculation:** If shaking hands does not consume additional arms (unlikely), then 8 - 5 = 3 free arms.
   - **Support:** Less likely, as shaking hands typically involves arm usage.

### l. **Quickest and Easiest Solution**
- **Chosen Solution:** 1 free arm.
- **Reasoning:** The robot uses 5 arms to pick up objects and 2 arms to shake hands, totaling 7 arms in use out of 8, leaving 1 arm free.
- **Supporting Quotes:** "The robot has 8 arms... picks up each object with an arm... shakes hands with himself."

### m. **Reflection**
- **Consistency Check:** The chosen solution aligns with the explicit information provided.
- **Assumptions Revisited:** Assumed that shaking hands uses two arms based on standard practice.
- **Information Sufficiency:** The problem provides enough information to deduce the number of free arms.
- **Direct Answering:** The analysis directly addresses the question asked.

### **Recommendation:**
Based on the explicit information provided, the robot uses 5 arms to pick up the objects and 2 arms to shake hands with himself, totaling 7 arms in use. Therefore, the robot has **1 arm free**.
\end{lstlisting}
\end{quote}

\noindent\hrulefill 

\noindent\textbf{D.6 Zero-Shot OpenAI GPT-4o Response - Score: 100\%}
\begin{quote}
\begin{lstlisting}[basicstyle=\ttfamily\small, breaklines=true]
The robot starts with 8 arms. He picks up 5 objects, one in each arm. This leaves him with:

\[ 8 - 5 = 3 \text{ arms free.} \]

Next, he shakes hands with himself, which would involve using 2 more arms. After shaking hands, he would have:

\[ 3 - 2 = 1 \text{ arm free.} \]

So, the robot has **1 arm free**.
\end{lstlisting}
\end{quote}
\noindent\hrulefill

\noindent\textbf{D.6 REAP OpenAI GPT-4o Response - Score: 100\%}
\begin{quote}
\begin{lstlisting}[basicstyle=\ttfamily\small, breaklines=true]
# REAP Results:

**a. Comprehensive Feature List:**

- **Objects mentioned:**
  - "A robot has 8 arms."
  - "There are 5 objects on a table: a knife, a fork, a spoon, a teddy bear, and a doll."
  
- **Actions described:**
  - "The robot picks up each object with an arm."
  - "He then shakes hands with himself."
  
- **Numerical Values:**
  - "8 arms."
  - "5 objects."

**b. Sequential and Mechanical Process Check:**

- **Sequence of Events:**
  1. The robot picks up 5 objects, using one arm per object.
  2. The robot shakes hands with himself.

- **Mechanical Process:**
  - Each arm holds an object, reducing the number of free arms.
  - Shaking hands typically involves using two arms, one from each "side" of the robot's body.

**c. Key Insight Check Results:**
- **Key Insight:** The robot has 8 arms, and after picking up the 5 objects, 3 arms remain free. However, if he shakes hands with himself, 2 of these arms are used.

**d. Rephrased Question:**
- How many arms does the robot have free after picking up the objects and shaking hands with himself?

**e. Known and Deduced Information:**

- **Explicit Facts:**
  - "[A robot has 8 arms]" - Indicates the total number of arms the robot has.
  - "[The robot picks up each object with an arm]" - Suggests 5 arms are used to hold the 5 objects.
  - "[He then shakes hands with himself]" - Implies an additional use of 2 arms for this action.
  
- **Feature-Based Deduction:**
  - **Deduction:** 5 arms are occupied by holding objects, leaving 3 arms free.
  - **Logical Basis:** "The robot picks up each object with an arm" - 5 arms are used, leaving 3 free.
  - **Deduction:** If 2 of the free arms are used for shaking hands, 1 arm remains free.
  - **Logical Basis:** "He then shakes hands with himself" - 2 arms are likely used for this action.

**f. Problem Decomposition:**

- **Component:** Robot's arms
  - **Supporting Quote:** "[A robot has 8 arms]"
  
- **Component:** Objects being held
  - **Supporting Quote:** "[There are 5 objects on a table: a knife, a fork, a spoon, a teddy bear, and a doll.]"
  
- **Interaction:** Picking up objects
  - **Supporting Quote:** "[The robot picks up each object with an arm.]"
  
- **Interaction:** Shaking hands
  - **Supporting Quote:** "[He then shakes hands with himself.]"

**g. Graph of Thought:**

- **Nodes:**
  - Robot's arms
  - Objects held
  - Shaking hands
  
- **Edges:**
  - Connection between "picking up objects" and "reducing free arms."
  - Connection between "shaking hands" and "further reducing free arms."

**h. Spatial and Object Analysis:**

- **Objects:**
  - Knife, fork, spoon, teddy bear, doll.
  
- **Spatial Relationships:**
  - No explicit spatial relationships beyond the robot holding these objects with individual arms.
  
- **Movement and Dynamics:**
  - The robot uses 5 arms to pick up objects and likely 2 arms to shake hands.

**i. Bayesian Updates:**
- Not applicable, as all relevant information has been explicitly stated.

**j. Ethical Check and Decision-Making Under Uncertainty:**
- There are no ethical concerns or risks involved in this problem.

**k. Multiple Solutions:**

- **Solution 1:** The robot has 1 arm free after picking up the objects and shaking hands with himself.
  - **Support:** "[The robot picks up each object with an arm]" and "[He then shakes hands with himself]"

- **Solution 2:** If shaking hands doesn't require 2 arms (a possibility, though unlikely), the robot might have 3 arms free.
  - **Support:** Unconventional interpretation of shaking hands.

**l. Quickest and Easiest Solution:**
- The robot has 1 arm free after the stated actions, as the standard interpretation of shaking hands implies using 2 arms.

**m. Reflection:**
- The problem explicitly describes all necessary actions. The logical interpretation is that the robot uses 2 arms to shake hands, leaving 1 arm free.

---

**Recommendation:**
- The robot has **1 arm free** after picking up the 5 objects and shaking hands with himself. This conclusion is directly supported by the explicit information provided in the problem statement.
\end{lstlisting}
\end{quote}
\noindent\hrulefill

\noindent\textbf{D.6 Zero-Shot OpenAI GPT-4o-mini Response - Score: 100\%}
\begin{quote}
\begin{lstlisting}[basicstyle=\ttfamily\small, breaklines=true]
To determine how many arms the robot has free, let's analyze the situation step by step:

1. **Initial Condition:**
   - The robot has 8 arms.

2. **Picking Up Objects:**
   - The robot picks up each of the 5 objects (knife, fork, spoon, teddy bear, and doll) with an arm.
   - Since each object requires one arm to hold, the robot uses 5 of its 8 arms to hold these objects.

3. **Shaking Hands:**
   - Shaking hands with oneself means the robot will use two arms for this action.
   - After using two arms to shake hands, these arms are occupied.

4. **Calculating Free Arms:**
   - The total number of arms used for holding objects is 5.
   - The number of arms used for shaking hands is 2.
   - Therefore, the total number of arms occupied is \(5 + 2 = 7\).

   The robot initially has 8 arms, so the number of free arms is:

   \[
   8 - 7 = 1
   \]

Thus, the robot has **1 arm** free.
\end{lstlisting}
\end{quote}
\noindent\hrulefill

\noindent\textbf{D.6 REAP OpenAI GPT-4o-mini Response - Score: 100\%}
\begin{quote}
\begin{lstlisting}[basicstyle=\ttfamily\small, breaklines=true]
# REAP Results:

## a. Comprehensive Feature List

1. **Objects on the Table:**
   - "a knife"
   - "a fork"
   - "a spoon"
   - "a teddy bear"
   - "a doll"
   - *Significance:* These are the items that the robot will interact with.

2. **Robot's Arms:**
   - "A robot has 8 arms"
   - *Significance:* The robot has 8 arms available to use.

3. **Actions Performed:**
   - "The robot picks up each object with an arm"
     - *Significance:* Each object is picked up using one arm each.
   - "He then shakes hands with himself"
     - *Significance:* This implies using arms to engage in a hand-shaking motion with himself.

4. **Arm Usage:**
   - *Picking up Objects:* Each of the 5 objects will be picked up with an arm.
   - *Shaking Hands:* Typically involves using at least 2 arms (one to extend and one to engage).

## b. Sequential and Mechanical Process Check

1. **Picking Up Objects:**
   - The robot picks up 5 objects.
   - Each object uses 1 arm.
   - Therefore, 5 arms are in use for picking up the objects.

2. **Shaking Hands:**
   - Shaking hands with himself implies at least 2 arms are engaged in this activity.

3. **Total Arms in Use:**
   - 5 arms for picking up objects
   - 2 arms for shaking hands
   - Total: 5 + 2 = 7 arms in use.

4. **Arms Free:**
   - Total arms: 8
   - Arms used: 7
   - Free arms: 8 - 7 = 1 arm free.

## c. Key Insight Check Results

The key insight is straightforward: The robot uses 5 arms to pick up each object and 2 arms to shake hands with itself. This results in 7 arms being occupied, leaving 1 arm free.

## d. Rephrased Question

**How many arms does the robot have free?**

## e. Known and Deduced Information

1. "A robot has 8 arms" - The robot has a total of 8 arms.
2. "The robot picks up each object with an arm" - 5 objects mean 5 arms are occupied.
3. "He then shakes hands with himself" - Shaking hands uses at least 2 arms.

**Deduction:**
- The robot uses 7 arms (5 for objects, 2 for shaking hands).
- Therefore, 1 arm is free.

## f. Problem Decomposition

1. **Arms Available:**
   - 8 arms in total.

2. **Arms Used for Picking Up Objects:**
   - 5 objects, 5 arms used.

3. **Arms Used for Shaking Hands:**
   - 2 arms engaged.

4. **Total Arms Used:**
   - 5 (objects) + 2 (shaking hands) = 7 arms.

5. **Arms Free:**
   - 8 total - 7 used = 1 arm free.

## g. Graph of Thought

1. **Nodes:**
   - Total arms: 8
   - Arms used for objects: 5
   - Arms used for shaking hands: 2
   - Free arms: 1

2. **Edges:**
   - Total arms -> Arms used for objects: -5
   - Total arms -> Arms used for shaking hands: -2
   - Total arms -> Free arms: -1

3. **Connections:**
   - 8 total arms - 5 (objects) - 2 (shaking hands) = 1 free arm.

## h. Spatial and Object Analysis

1. **Objects:**
   - Knife, fork, spoon, teddy bear, and doll.
   - Each object occupies 1 arm.

2. **Movement:**
   - Picking up each object uses 1 arm.
   - Shaking hands uses 2 arms.

3. **Space:**
   - The problem doesn't specify the spatial arrangement of the objects or arms beyond these actions.

## i. Bayesian Updates

- No new explicit information beyond what's given; thus, no updates required.

## j. Ethical Check and Decision-Making Under Uncertainty

- No severe risks or ethical issues are involved.

## k. Multiple Solutions

- Given the exact statements and calculations, no alternative solutions are possible. The only solution is based on the information provided.

## l. Quickest and Easiest Solution

- Based on explicit information, the simplest solution is to calculate the number of free arms: 8 total arms - 7 used = 1 free arm.

## m. Reflection

- The analysis confirms the calculation is correct based on the given information. No contradictions or assumptions beyond the provided details were made. The problem was directly answered by the explicit information available.

**Recommendation:**

The number of arms the robot has free is **1**.
\end{lstlisting}
\end{quote}
\noindent\hrulefill

\noindent\textbf{D.6 Zero-Shot Google Gemini 1.5 Pro Response - Score: 100\%}
\begin{quote}
\begin{lstlisting}[basicstyle=\ttfamily\small, breaklines=true]
Here's how to solve this:

* **Start with the total arms:** The robot has 8 arms.
* **Objects picked up:** The robot uses 5 arms to pick up the 5 objects.
* **Shaking hands:** Shaking hands with himself uses 2 more arms.
* **Calculate free arms:** 8 (total arms) - 5 (holding objects) - 2 (shaking hands) = 1

**Answer: The robot has 1 arm free.** 
\end{lstlisting}
\end{quote}
\noindent\hrulefill

\noindent\textbf{D.6 REAP Google Gemini 1.5 Pro Response - Score: 100\%}
\begin{quote}
\begin{lstlisting}[basicstyle=\ttfamily\small, breaklines=true]
# REAP Results:

a. **Comprehensive Feature List** (with explicit quotes)

* "A robot has 8 arms." 
* "There are 5 objects on a table: a knife, a fork, a spoon, a teddy bear and a doll." 
* "The robot picks up each object with an arm." 
* "He then shakes hands with himself." 
* "How many arms does he have free?" 

b. **Sequential and Mechanical Process Check**

* The robot first picks up 5 objects, one with each arm.
* Then, the robot shakes hands with himself.
* The sequence implies that some arms will be occupied holding objects and some might be involved in the handshake, leaving some arms free.

c. **Key Insight Check Results**

* No immediate key insight that reveals the answer or significantly simplifies the problem.

d. **Rephrased Question:** 

* How many of the robot's arms are not holding objects or shaking hands?

e. **Known and Deduced Information**

* Explicit Facts:
    * "A robot has 8 arms." 
    * "There are 5 objects on a table: a knife, a fork, a spoon, a teddy bear and a doll." 
    * "The robot picks up each object with an arm." 
    * "He then shakes hands with himself." 

* Feature-Based Deduction
    * Deduction: The robot uses 5 arms to pick up the objects. 
    * Logical Basis: The problem states there are 5 objects and the robot picks up "each object with an arm." 
    * Based on: "There are 5 objects on a table: a knife, a fork, a spoon, a teddy bear and a doll.", "The robot picks up each object with an arm."

f. **Problem Decomposition**

* Component/Subcomponent: Robot's initial state
    * Supporting Quote(s): "A robot has 8 arms." 

* Component/Subcomponent: Objects on the table
    * Supporting Quote(s): "There are 5 objects on a table: a knife, a fork, a spoon, a teddy bear and a doll."

* Component/Subcomponent: Picking up objects
    * Supporting Quote(s): "The robot picks up each object with an arm."

* Component/Subcomponent: Shaking hands
    * Supporting Quote(s): "He then shakes hands with himself."

* Component/Subcomponent: Free arms
    * Supporting Quote(s): "How many arms does he have free?"

* Interaction: Picking up objects reduces the number of free arms
    * Supporting Quote(s): "The robot picks up each object with an arm."
    * Deduction: 5 arms are used to pick up objects
    * Based on: "There are 5 objects on a table: a knife, a fork, a spoon, a teddy bear and a doll.", "The robot picks up each object with an arm."

* Interaction: Shaking hands might further reduce the number of free arms
    * Supporting Quote(s): "He then shakes hands with himself."

g. **Graph of Thought**

* Key Concepts: Robot, Arms, Objects, Picking up, Shaking hands, Free arms
* Relationships:
    * Robot has 8 Arms
    * There are 5 Objects
    * Robot picks up each Object with one Arm
    * Robot shakes hands with himself
    * Question is about the number of Free arms
* Insights:
    * Picking up objects reduces the number of free arms by 5
    * Shaking hands might further reduce the number of free arms
    * Need to figure out how many arms are used for shaking hands to determine the final number of free arms

h. **Spatial and Object Analysis**

* Objects: Robot, 8 arms, knife, fork, spoon, teddy bear, doll, table
* Spatial Relationships:
    * 5 objects are on the table
    * Robot is near the table to pick up objects
* Movement and Dynamics:
    * Robot's arms move to pick up objects
    * Robot's arms move to shake hands
* Constraints:
    * Each object requires one arm to be picked up
    * Robot shakes hands with itself (unclear how many arms are involved)
* Object Properties:
    * Robot has 8 arms
    * 5 distinct objects on the table

i. **Bayesian Updates** 

* Initial belief: Robot has 8 free arms
* Update 1: Robot picks up 5 objects, each with one arm
    * Updated belief: Robot has 3 free arms (8 - 5 = 3)
* Update 2: Robot shakes hands with himself
    * Uncertainty: How many arms are used for shaking hands?
    * Potential implications: At least 2 arms are used, potentially more

j. **Ethical Check and Decision-Making Under Uncertainty**

* Known Information: Robot has 8 arms, 5 are used to hold objects, the robot shakes hands with itself
* Uncertain Information: How many arms are used for shaking hands
* Risks: None
* Decision Recommendation: Proceed with the analysis, acknowledging the uncertainty about the number of arms used for shaking hands

k. **Multiple Solutions**

* Solution 1: Robot uses the minimum number of arms for shaking hands (2 arms)
    * Supporting Information: The problem states the robot shakes hands "with himself", implying at least two arms are involved
    * Free arms: 1 (8 total arms - 5 holding objects - 2 shaking hands = 1)

* Solution 2: Robot uses more than 2 arms for shaking hands
    * Supporting Information: The problem doesn't specify how the robot shakes hands, it's possible more than 2 arms are involved
    * Free arms: Less than 1, potentially 0 

l. **Quickest and Easiest Solution**

* The quickest and easiest solution is to assume the robot uses the minimum number of arms for shaking hands (2 arms).
* Therefore, the robot has 1 free arm.
* This solution is based on the most straightforward interpretation of the problem statement and avoids making any assumptions beyond what is explicitly stated

m. **Reflection**

* The analysis directly addresses the question of how many free arms the robot has
* The conclusions are logically sound and supported by the explicit information in the problem statement
* No assumptions or inferences were made beyond the literal interpretation of the problem
* The problem lacks crucial information about how the robot shakes hands, leading to multiple possible solutions
* The quickest and easiest solution provides a reasonable answer based on the simplest interpretation

**Recommendation:**

* The robot has 1 free arm, assuming it uses the minimum number of arms (2) for shaking hands
* It's important to acknowledge that the actual number of free arms could be less than 1 if the robot uses more than 2 arms for shaking hands
* If more precise information about the handshake is required, further clarification or observation of the robot's actions would be necessary.
\end{lstlisting}
\end{quote}
\noindent\hrulefill

\noindent\textbf{D.6 Zero-Shot Claude 3.5 Sonnet Response - Score: 100\%}
\begin{quote}
\begin{lstlisting}[basicstyle=\ttfamily\small, breaklines=true]
Let's think through this step-by-step:

1. The robot starts with 8 arms.

2. There are 5 objects on the table, and the robot picks up each one:
   - Knife
   - Fork
   - Spoon
   - Teddy bear
   - Doll
   
   So, 5 arms are now holding objects.

3. The robot then shakes hands with itself. To shake hands, it needs to use 2 arms.

4. Let's count the arms in use:
   - 5 arms holding objects
   - 2 arms shaking hands
   - Total arms in use: 5 + 2 = 7 arms

5. To find out how many arms are free, we subtract the arms in use from the total number of arms:
   8 total arms - 7 arms in use = 1 arm free

Therefore, the robot has 1 arm free.
\end{lstlisting}
\end{quote}
\noindent\hrulefill

\noindent\textbf{D.6 REAP Claude 3.5 Sonnet Response - Score: 40\%}
\begin{quote}
\begin{lstlisting}[basicstyle=\ttfamily\small, breaklines=true]
#REAP Results:

a. Comprehensive Feature List (with explicit quotes):
1. "A robot has 8 arms."
2. "There are 5 objects on a table:"
3. "a knife"
4. "a fork"
5. "a spoon"
6. "a teddy bear"
7. "a doll"
8. "The robot picks up each object with an arm."
9. "He then shakes hands with himself."

b. Sequential and Mechanical Process Check:
The problem describes a sequence of events:
1. The robot starts with 8 arms.
2. The robot picks up 5 objects, each with one arm.
3. The robot then shakes hands with itself.

c. Key Insight Check Results:
The key insight is that the robot uses 5 arms to pick up objects, but the number of arms used for shaking hands with itself is not specified. This lack of information is crucial for answering the question.

d. Rephrased Question: How many arms does the robot have free after picking up objects and shaking hands with itself?

e. Known and Deduced Information:
Explicit Facts:
"A robot has 8 arms." - This establishes the initial number of arms.
"There are 5 objects on a table:" - This specifies the number of objects to be picked up.
"The robot picks up each object with an arm." - This indicates that 5 arms are used to pick up objects.
"He then shakes hands with himself." - This action occurs after picking up objects, but the number of arms used is not specified.

Feature-Based Deduction:
Deduction: The robot uses 5 arms to pick up objects.
Logical Basis: There are 5 objects, and each object is picked up with an arm.
Based on: "There are 5 objects on a table:", "The robot picks up each object with an arm."

f. Problem Decomposition:
Component 1: Robot's initial arm count
Supporting Quote: "A robot has 8 arms."

Component 2: Objects to be picked up
Supporting Quote: "There are 5 objects on a table: a knife, a fork, a spoon, a teddy bear and a doll."

Component 3: Picking up objects
Supporting Quote: "The robot picks up each object with an arm."

Component 4: Shaking hands with itself
Supporting Quote: "He then shakes hands with himself."

g. Graph of Thought:
Nodes:
1. Robot (8 arms)
2. 5 Objects
3. Picking up objects (5 arms used)
4. Shaking hands (unknown number of arms used)
5. Free arms (unknown)

Connections:
1 -> 3: Robot uses arms to pick up objects
3 -> 4: After picking up objects, robot shakes hands
4 -> 5: Remaining arms are free

h. Spatial and Object Analysis:
Objects: Robot (with 8 arms), knife, fork, spoon, teddy bear, doll
Spatial Relationships: Objects are on a table
Movement: Robot picks up each object with an arm, then shakes hands with itself

i. Bayesian Updates:
No new information is provided beyond the initial problem statement, so no Bayesian updates are necessary.

j. Ethical Check and Decision-Making Under Uncertainty:
There are no severe ethical implications or risks associated with this problem. The main uncertainty is the number of arms used for shaking hands.

k. Multiple Solutions:
1. If the robot uses 1 arm to shake hands with itself:
   8 initial arms - 5 arms holding objects - 1 arm shaking = 2 free arms
2. If the robot uses 2 arms to shake hands with itself:
   8 initial arms - 5 arms holding objects - 2 arms shaking = 1 free arm
3. If the robot uses no arms to shake hands with itself:
   8 initial arms - 5 arms holding objects - 0 arms shaking = 3 free arms

l. Quickest and Easiest Solution:
Based solely on the explicit information given in the problem statement, we cannot determine how many arms the robot has free. The problem does not specify how many arms are used for shaking hands with itself, which is crucial for calculating the number of free arms.

m. Reflection:
Upon review, it's clear that the problem lacks crucial information needed to provide a definitive answer. We know the robot starts with 8 arms and uses 5 to pick up objects, but we don't know how many arms are used for shaking hands. This missing information prevents us from calculating the exact number of free arms. Our analysis has adhered strictly to the given information without making assumptions.

Recommendation:
Based on the explicitly stated information and considering the uncertainties in the problem, we cannot provide a definitive answer to how many arms the robot has free. The problem lacks crucial information about how many arms are used for shaking hands. We can only state with certainty that the robot uses 5 arms to pick up objects, leaving a maximum of 3 arms potentially free, but the actual number could be less depending on how many arms are used for shaking hands. To provide an accurate answer, we would need additional information about the hand-shaking action.
\end{lstlisting}
\end{quote}
\noindent\hrulefill

\label{subsec:question7}
\subsection{\textbf{\large Question 7: Alan, Bob, Colin, Dave, and Emily are standing in a circle. Alan is on Bob's immediate left. Bob is on Colin's immediate left. Colin is on Dave's immediate left. Dave is on Emily's immediate left. Who is on Alan's immediate right?}}
\label{sec:question7}

\noindent\hrulefill 

\noindent\textbf{Human Response - Score 100\%:}
\begin{quote}
Bob is on Alan's immediate right because it is stated that Alan is on Bob's immediate left.
\end{quote}

\noindent\hrulefill 

\noindent\textbf{D.7 Zero-Shot OpenAI o1-preview Response - Score: 100\%}
\begin{quote}
\begin{lstlisting}[basicstyle=\ttfamily\small, breaklines=true]
To solve this problem, let's deduce the seating arrangement based on the given clues:

1. **Alan is on Bob's immediate left.**
   - This means, to Bob's left is Alan.

2. **Bob is on Colin's immediate left.**
   - So, to Colin's left is Bob.

3. **Colin is on Dave's immediate left.**
   - Therefore, to Dave's left is Colin.

4. **Dave is on Emily's immediate left.**
   - Hence, to Emily's left is Dave.

To establish a full circle, we need to determine who is on Alan's left. Since there are only five people and we're forming a circle, the only person left is Emily. Therefore:

- **Alan is on Emily's right**, so **Emily is on Alan's left**.

Now, arranging the people in order, considering that the person on the left is seated just counterclockwise from them, we get:

- **Emily** is to the left of **Alan**
- **Alan** is to the left of **Bob**
- **Bob** is to the left of **Colin**
- **Colin** is to the left of **Dave**
- **Dave** is to the left of **Emily**

This circular arrangement shows that **Bob** is seated immediately to Alan's right.

**Answer:** Bob
\end{lstlisting}
\end{quote}

\noindent\hrulefill 

\noindent\textbf{D.7 REAP OpenAI o1-preview Response - Score: 100\%}
\begin{quote}
\begin{lstlisting}[basicstyle=\ttfamily\small, breaklines=true]
# "REAP Results:"

**a. Comprehensive Feature List**

- **Actors/Entities:**
  - "Alan"
  - "Bob"
  - "Colin"
  - "Dave"
  - "Emily"

- **Actions/Relationships:**
  - "Alan is on Bob's immediate left."
    - Potential significance: Establishes that Alan is directly to the left of Bob.
  - "Bob is on Colin's immediate left."
    - Potential significance: Establishes that Bob is directly to the left of Colin.
  - "Colin is on Dave's immediate left."
    - Potential significance: Establishes that Colin is directly to the left of Dave.
  - "Dave is on Emily's immediate left."
    - Potential significance: Establishes that Dave is directly to the left of Emily.

- **Spatial Information:**
  - "standing in a circle"
    - Potential significance: Indicates a circular arrangement where positions are connected in a loop.

- **Goal/Objectives:**
  - "Who is on Alan's immediate right?"
    - The question to be answered.

**b. Sequential and Mechanical Process Check**

- **Sequence of Positions:**
  1. Emily's immediate left is Dave.
  2. Dave's immediate left is Colin.
  3. Colin's immediate left is Bob.
  4. Bob's immediate left is Alan.
  5. Since they are in a circle, Alan's immediate left connects back to Emily.

- **Analysis:**
  - The positions create a continuous loop due to the circular arrangement.
  - Each person's immediate left neighbor is explicitly stated, forming a sequence.

- **Key Questions:**
  - Does this sequence determine who is on Alan's immediate right?
  - How does the circular arrangement impact the right-side neighbor?

**c. Key Insight Check Results**

- The circular arrangement and given relationships uniquely determine each person's position relative to others.
- By constructing the sequence based on immediate left relationships, we can deduce who is on Alan's immediate right.
- **Key Insight:** Bob is on Alan's immediate right.

**d. Rephrased Question:**

- "Determine who is directly to the right of Alan in the circle."

**e. Known and Deduced Information**

**Explicit Facts:**

1. "Alan is on Bob's immediate left."
   - Meaning Bob is immediately to the right of Alan.
2. "Bob is on Colin's immediate left."
   - Meaning Colin is immediately to the right of Bob.
3. "Colin is on Dave's immediate left."
   - Meaning Dave is immediately to the right of Colin.
4. "Dave is on Emily's immediate left."
   - Meaning Emily is immediately to the right of Dave.
5. "Alan, Bob, Colin, Dave and Emily are standing in a circle."
   - Indicates a continuous loop of positions.
6. Goal: "Who is on Alan's immediate right?"

**Feature-Based Deduction:**

- **Deduction:** Bob is on Alan's immediate right.
  - **Logical Basis:** Since "Alan is on Bob's immediate left," it follows that Bob is immediately to the right of Alan in the circle.
  - **Based on:** "Alan is on Bob's immediate left."

**f. Problem Decomposition**

- **Component:** Circular Arrangement of People
  - **Supporting Quote:** "Alan, Bob, Colin, Dave and Emily are standing in a circle."
- **Subcomponents:**
  1. **Position of Alan and Bob**
     - "Alan is on Bob's immediate left."
     - Deduction: Bob is on Alan's immediate right.
  2. **Position of Bob and Colin**
     - "Bob is on Colin's immediate left."
     - Deduction: Colin is on Bob's immediate right.
  3. **Position of Colin and Dave**
     - "Colin is on Dave's immediate left."
     - Deduction: Dave is on Colin's immediate right.
  4. **Position of Dave and Emily**
     - "Dave is on Emily's immediate left."
     - Deduction: Emily is on Dave's immediate right.

- **Interaction:**
  - The positions form a continuous sequence around the circle based on the immediate left relationships.

**g. Graph of Thought**

- **Nodes (People):**
  - Alan, Bob, Colin, Dave, Emily

- **Edges (Immediate Left Relationships):**
  - Bob -> Alan (Alan is on Bob's immediate left)
  - Colin -> Bob (Bob is on Colin's immediate left)
  - Dave -> Colin (Colin is on Dave's immediate left)
  - Emily -> Dave (Dave is on Emily's immediate left)
  - Alan -> Emily (By circular arrangement)

- **Connections and Insights:**
  - The sequence forms: Emily -> Dave -> Colin -> Bob -> Alan -> Emily
  - Moving counterclockwise, each person is on the immediate left of the next.
  - Therefore, moving clockwise reveals each person's immediate right.

**h. Spatial and Object Analysis**

- **Objects:**
  - Individuals: Alan, Bob, Colin, Dave, Emily

- **Spatial Relationships:**
  - Circular formation: "standing in a circle"
  - **Arrangement (Counterclockwise):**
    - Emily's immediate left: Dave
    - Dave's immediate left: Colin
    - Colin's immediate left: Bob
    - Bob's immediate left: Alan
    - Alan's immediate left: Emily
  - **Arrangement (Clockwise):**
    - Alan's immediate right: Bob
    - Bob's immediate right: Colin
    - Colin's immediate right: Dave
    - Dave's immediate right: Emily
    - Emily's immediate right: Alan

- **Visualization:**
  - A circle with the above positions marked accordingly.

- **Spatial Implications:**
  - The immediate right of any person is the one who considers them on their immediate left.

**i. Bayesian Updates**

- Not applicable; all conclusions are based on explicit information without probabilistic reasoning.

**j. Ethical Check and Decision-Making Under Uncertainty**

- No ethical concerns or risks are associated with this problem.

**k. Multiple Solutions**

- **Solution 1:**
  - **Answer:** Bob is on Alan's immediate right.
  - **Justification:** Based on "Alan is on Bob's immediate left."

- **Solution 2:**
  - **Answer:** Considering alternative interpretations of "left" and "right," but the explicit statements lead only to one logical arrangement.
  - **Justification:** The given relationships fix the positions uniquely.

- **Solution 3:**
  - **Answer:** No other valid solutions without contradicting the explicit information.

**l. Quickest and Easiest Solution**

- **Answer:** Bob is on Alan's immediate right.
- **Reasoning:** Since "Alan is on Bob's immediate left," it directly follows that Bob is on Alan's immediate right.

**m. Reflection**

- The solution aligns with all explicit information provided.
- No assumptions were made beyond the given relationships.
- The conclusion directly answers the question.
- There are no contradictions or uncertainties in the analysis.

---

**Recommendation:**

Based solely on the explicitly stated information, **Bob is on Alan's immediate right**.
\end{lstlisting}
\end{quote}

\noindent\hrulefill 

\noindent\textbf{D.7 Zero-Shot OpenAI o1-mini Response - Score: 100\%}
\begin{quote}
\begin{lstlisting}[basicstyle=\ttfamily\small, breaklines=true]
Let's break down the arrangement step by step based on the given information:

1. **Alan is on Bob's immediate left.**
   - This means that if you move clockwise around the circle, you encounter Bob first, then Alan.

2. **Bob is on Colin's immediate left.**
   - Moving clockwise, the order now is Colin -> Bob -> Alan.

3. **Colin is on Dave's immediate left.**
   - Extending the clockwise order: Dave -> Colin -> Bob -> Alan.

4. **Dave is on Emily's immediate left.**
   - Completing the circle clockwise: Emily -> Dave -> Colin -> Bob -> Alan -> back to Emily.

So, the clockwise order of people in the circle is:
**Emily -> Dave -> Colin -> Bob -> Alan -> Emily**

To find out who is on **Alan's immediate right**, we look counterclockwise from Alan:
- **Alan's immediate right** is the person who comes just before Alan when moving counterclockwise, which is **Bob**.

**Answer:** Bob is on Alan's immediate right.
\end{lstlisting}
\end{quote}
\noindent\hrulefill 

\noindent\textbf{D.7 REAP OpenAI o1-mini Response - Score: 100\%}
\begin{quote}
\begin{lstlisting}[basicstyle=\ttfamily\small, breaklines=true]
# REAP Results:

### a. **Comprehensive Feature List**
- "Alan, Bob, Colin, Dave and Emily are standing in a circle."
  - **Significance:** Identifies the participants and the circular arrangement.
- "Alan is on Bob's immediate left."
  - **Significance:** Establishes the positional relationship between Alan and Bob.
- "Bob is on Colin's immediate left."
  - **Significance:** Establishes the positional relationship between Bob and Colin.
- "Colin is on Dave's immediate left."
  - **Significance:** Establishes the positional relationship between Colin and Dave.
- "Dave is on Emily's immediate left."
  - **Significance:** Establishes the positional relationship between Dave and Emily.
- "Who is on Alan's immediate right?"
  - **Significance:** The core question to determine.

### b. **Sequential and Mechanical Process Check**
- **Sequence of Immediate Left Relationships:**
  1. Alan -> immediate left of Bob
  2. Bob -> immediate left of Colin
  3. Colin -> immediate left of Dave
  4. Dave -> immediate left of Emily
- **Impact on Outcome:** This sequence helps in determining the relative positions of each person in the circle, which is essential to identify who is on Alan's immediate right.

### c. **Key Insight Check Results**
- **Key Insight:** The series of "immediate left" relationships forms a consistent directional loop around the circle, allowing for the determination of each person's position relative to the others.
- **Impact:** By following the sequence, the positions of all individuals can be mapped, leading directly to identifying who stands immediately to Alan's right.

### d. **Rephrased Question:**
Who is standing immediately to the right of Alan in the circle?

### e. **Known and Deduced Information**

**Explicit Facts:**
- "Alan, Bob, Colin, Dave and Emily are standing in a circle."
- "Alan is on Bob's immediate left."
- "Bob is on Colin's immediate left."
- "Colin is on Dave's immediate left."
- "Dave is on Emily's immediate left."
- "Who is on Alan's immediate right?"

**Feature-Based Deduction:**
- **Deduction:** Since Alan is on Bob's immediate left, Bob is to Alan's immediate right.
  - **Logical Basis:** If A is on B's immediate left, then B is on A's immediate right.
  - **Based on:** "Alan is on Bob's immediate left."

### f. **Problem Decomposition**

- **Component/Subcomponent:** Identification of the circle arrangement.
  - **Supporting Quote(s):** "Alan, Bob, Colin, Dave and Emily are standing in a circle."
  
- **Component/Subcomponent:** Positional relationships based on "immediate left."
  - **Supporting Quote(s):**
    - "Alan is on Bob's immediate left."
    - "Bob is on Colin's immediate left."
    - "Colin is on Dave's immediate left."
    - "Dave is on Emily's immediate left."

- **Component/Subcomponent:** Determination of the person on Alan's immediate right.
  - **Supporting Quote(s):** "Who is on Alan's immediate right?"

### g. **Graph of Thought**
- **Nodes (Key Concepts):**
  - Individuals: Alan, Bob, Colin, Dave, Emily
  - Positional Relationships: Immediate left

- **Edges (Relationships):**
  - Alan -> immediate left of Bob
  - Bob -> immediate left of Colin
  - Colin -> immediate left of Dave
  - Dave -> immediate left of Emily

- **Insights:**
  - The circular arrangement allows for each person's position to be determined relative to others.
  - Establishing a directional flow based on immediate left relationships facilitates mapping the entire circle.

### h. **Spatial and Object Analysis**

- **Identify Objects:** Alan, Bob, Colin, Dave, Emily
- **Spatial Relationships:**
  - Alan is immediately to the left of Bob.
  - Bob is immediately to the left of Colin.
  - Colin is immediately to the left of Dave.
  - Dave is immediately to the left of Emily.
- **Movement and Dynamics:** Static positions; no movement described.
- **Constraints:** Circular arrangement; each person's position is fixed relative to others based on immediate left relationships.
- **Visualization:** A circle with each person positioned such that each one is to the immediate left of the next as per the relationships described.

### i. **Bayesian Updates**
- **Initial Belief:** The arrangement is a simple circular setup based on immediate left relationships.
- **Updated Belief:** The deduction that Bob is on Alan's immediate right is reinforced by the direct positional statements.

### j. **Ethical Check and Decision-Making Under Uncertainty**
- **Assessment:** The problem is purely logical with no ethical implications or risks.
- **Decision-Making:** Straightforward logical deduction without ethical considerations.

### k. **Multiple Solutions**
- **Solution 1:** Bob is on Alan's immediate right.
  - **Supporting Information:** "Alan is on Bob's immediate left." Therefore, Bob is on Alan's immediate right.
  
- **Solution 2:** [Not applicable]
  - **Explanation:** Given the explicit positional relationships, there is only one logical solution.

- **Solution 3:** [Not applicable]
  - **Explanation:** Similarly, no alternative solutions exist based on the given information.

### l. **Quickest and Easiest Solution**
- **Solution:** Bob is on Alan's immediate right.
- **Reasoning:** The statement "Alan is on Bob's immediate left" directly implies that Bob is on Alan's immediate right.

### m. **Reflection**
- The analysis strictly adheres to the explicit information provided.
- No assumptions or inferences beyond the given statements were made.
- The conclusion directly answers the question using the positional relationships described.
- The logical deduction is sound and free from contradictions.

**Recommendation:**

Based on the explicit information provided, **Bob is on Alan's immediate right**. This conclusion is directly derived from the statement "Alan is on Bob's immediate left," which logically implies that Bob occupies the position immediately to Alan's right in the circular arrangement.
\end{lstlisting}
\end{quote}

\noindent\hrulefill 

\noindent\textbf{D.7 Zero-Shot OpenAI GPT-4o Response - Score: 0\%}
\begin{quote}
\begin{lstlisting}[basicstyle=\ttfamily\small, breaklines=true]
To solve this problem, let's arrange the individuals in a circle according to the given information:

1. Alan is on Bob's immediate left, so Bob is immediately to Alan's right.
2. Bob is on Colin's immediate left, so Colin is immediately to Bob's right.
3. Colin is on Dave's immediate left, so Dave is immediately to Colin's right.
4. Dave is on Emily's immediate left, so Emily is immediately to Dave's right.

Starting with Alan and following the direction provided:

- Alan is immediately left of Bob.
- Bob is immediately left of Colin.
- Colin is immediately left of Dave.
- Dave is immediately left of Emily.

The circle, then, can be arranged as follows:
- **Alan -> Bob -> Colin -> Dave -> Emily**

Given this arrangement, the person on Alan's immediate right is **Emily**.
\end{lstlisting}
\end{quote}
\noindent\hrulefill

\noindent\textbf{D.7 REAP OpenAI GPT-4o Response - Score: 100\%}
\begin{quote}
\begin{lstlisting}[basicstyle=\ttfamily\small, breaklines=true]
# REAP Results

### a. **Comprehensive Feature List**
1. **Objects Mentioned:**
   - "Alan"
   - "Bob"
   - "Colin"
   - "Dave"
   - "Emily"
  
2. **Actors/Entities:**
   - "Alan"
   - "Bob"
   - "Colin"
   - "Dave"
   - "Emily"

3. **Actions Described:**
   - "Alan is on Bob's immediate left."
   - "Bob is on Colin's immediate left."
   - "Colin is on Dave's immediate left."
   - "Dave is on Emily's immediate left."

4. **Directional/Relational/Sequential Phrases:**
   - "Alan is on Bob's immediate left."
   - "Bob is on Colin's immediate left."
   - "Colin is on Dave's immediate left."
   - "Dave is on Emily's immediate left."

5. **Stated Conditions or Constraints:**
   - The individuals are standing "in a circle."

6. **Numerical Values or Quantities:**
   - There are five individuals.

### b. **Sequential and Mechanical Process Check**
- The individuals are positioned in a circle, with each person standing to the immediate left of another.
- We need to determine the sequence in which they stand to understand the arrangement and identify who is on Alan's immediate right.

### c. **Key Insight Check Results**
- The key insight here is that since the individuals are in a circle and each one is to the immediate left of another, the sequence of the individuals is crucial to determine who stands to the right of Alan.

### d. **Rephrased Question**
- "Who is standing immediately to the right of Alan in the circle?"

### e. **Known and Deduced Information**
1. "[Alan is on Bob's immediate left.]" - This means Bob is to the right of Alan.
2. "[Bob is on Colin's immediate left.]" - This means Colin is to the right of Bob.
3. "[Colin is on Dave's immediate left.]" - This means Dave is to the right of Colin.
4. "[Dave is on Emily's immediate left.]" - This means Emily is to the right of Dave.
5. "[Standing in a circle.]" - This means the sequence loops back to the beginning.

### f. **Problem Decomposition**
- **Component 1: Bob is on Alan's right**
  - Supporting Quote: "Alan is on Bob's immediate left."
- **Component 2: Colin is on Bob's right**
  - Supporting Quote: "Bob is on Colin's immediate left."
- **Component 3: Dave is on Colin's right**
  - Supporting Quote: "Colin is on Dave's immediate left."
- **Component 4: Emily is on Dave's right**
  - Supporting Quote: "Dave is on Emily's immediate left."
- **Component 5: Alan is on Emily's right**
  - Supporting Quote: "Standing in a circle." (implies the loop)

### g. **Graph of Thought**
- **Connections:**
  - Alan -> Bob -> Colin -> Dave -> Emily -> Alan (in a circle)
- **Insights:**
  - Bob must be on Alan's immediate right based on the provided sequential relationships.

### h. **Spatial and Object Analysis**
- **Objects:**
  - Alan, Bob, Colin, Dave, Emily
- **Spatial Relationships:**
  - Alan is left of Bob, who is left of Colin, who is left of Dave, who is left of Emily, forming a circle.
- **Movement/Dynamics:**
  - The sequence is circular and continuous.

### i. **Bayesian Updates**
- Not applicable as all the information is explicit.

### j. **Ethical Check and Decision-Making Under Uncertainty**
- There are no significant risks or ethical concerns in this problem.

### k. **Multiple Solutions**
1. **Solution 1:**
   - **Support:** Alan is on Bob's left.
   - **Conclusion:** Bob is on Alan's immediate right.
2. **Solution 2:**
   - **Support:** Sequential order ends with Emily.
   - **Conclusion:** The circular arrangement confirms that Bob is on Alan's immediate right.

### l. **Quickest and Easiest Solution**
- The direct answer to the problem based on explicit information is: "Bob is on Alan's immediate right."

### m. **Reflection**
- The problem is solved based on the explicit sequential information provided, ensuring no assumptions were made beyond what was given.

---

**Final Recommendation:**
- Based on the explicit sequential arrangement provided, **Bob** is the person standing immediately to the right of Alan. This conclusion is drawn strictly from the information given in the problem without making any assumptions.
\end{lstlisting}
\end{quote}
\noindent\hrulefill

\noindent\textbf{D.7 Zero-Shot OpenAI GPT-4o-mini Response - Score: 100\%}
\begin{quote}
\begin{lstlisting}[basicstyle=\ttfamily\small, breaklines=true]
To determine who is on Alan's immediate right, let's analyze the given information and arrange the people in the circle accordingly.

1. **Alan is on Bob's immediate left**:  
   In the circle, this means Bob is immediately to the right of Alan.

2. **Bob is on Colin's immediate left**:  
   This means Colin is immediately to the right of Bob.

3. **Colin is on Dave's immediate left**:  
   This means Dave is immediately to the right of Colin.

4. **Dave is on Emily's immediate left**:  
   This means Emily is immediately to the right of Dave.

Let's arrange them step-by-step:
- Start with Alan and place Bob to Alan's right:
  Alan - Bob
- Place Colin to Bob's right:
  Alan - Bob - Colin
- Place Dave to Colin's right:
  Alan - Bob - Colin - Dave
- Finally, place Emily to Dave's right:
  Alan - Bob - Colin - Dave - Emily

Now we see the circle complete with:
  Alan - Bob - Colin - Dave - Emily

To determine who is on Alan's immediate right, look directly to the right of Alan:
- **Bob** is immediately to the right of Alan.

Thus, the person on Alan's immediate right is **Bob**.
\end{lstlisting}
\end{quote}
\noindent\hrulefill

\noindent\textbf{D.7 REAP OpenAI GPT-4o-mini Response - Score: 100\%}
\begin{quote}
\begin{lstlisting}[basicstyle=\ttfamily\small, breaklines=true]
# REAP Results:

### a. Comprehensive Feature List

1. **Objects/Entities:**
   - Alan
   - Bob
   - Colin
   - Dave
   - Emily

2. **Spatial Relationships:**
   - "Alan is on Bob's immediate left."
   - "Bob is on Colin's immediate left."
   - "Colin is on Dave's immediate left."
   - "Dave is on Emily's immediate left."

3. **Goals/Questions:**
   - Determine who is on Alan's immediate right.

### b. Sequential and Mechanical Process Check

- The problem describes a circular arrangement. The sequence is:
  1. Alan -> Bob
  2. Bob -> Colin
  3. Colin -> Dave
  4. Dave -> Emily
  5. Emily -> Alan (because it's a circle)

  This forms a complete circle where each person is immediately to the left of the next.

### c. Key Insight Check Results

The circle is described completely, so each person's position relative to each other is explicitly defined. By completing the circle based on the given information:

- Alan is on Bob's immediate left.
- Bob is on Colin's immediate left.
- Colin is on Dave's immediate left.
- Dave is on Emily's immediate left.
- Therefore, Emily is on Alan's immediate left.

From this arrangement, the person immediately to Alan's right is the one who follows Alan in the circle sequence, which is Bob.

### d. Rephrased Question

Who is on Alan's immediate right?

### e. Known and Deduced Information

- **Explicit Quotes:**
  - "Alan is on Bob's immediate left."
  - "Bob is on Colin's immediate left."
  - "Colin is on Dave's immediate left."
  - "Dave is on Emily's immediate left."

- **Deduction:**
  - Since the arrangement is a circle, if Alan is on Bob's immediate left, then Bob is on Alan's immediate right.

### f. Problem Decomposition

**Component/Subcomponent: Circle Arrangement**
- **Supporting Quote(s):** "Alan is on Bob's immediate left", "Bob is on Colin's immediate left", "Colin is on Dave's immediate left", "Dave is on Emily's immediate left"
- **Deduction:** Since the circle completes the loop, the order must be: Alan -> Bob -> Colin -> Dave -> Emily -> Alan.

### g. Graph of Thought

**Nodes:**
- Alan
- Bob
- Colin
- Dave
- Emily

**Edges (Relationships):**
- Alan -> Bob
- Bob -> Colin
- Colin -> Dave
- Dave -> Emily
- Emily -> Alan

**Connections:**
- These connections form a closed loop or circle.

**Insights:**
- Direct adjacency is clearly defined in a circular manner.

### h. Spatial and Object Analysis

**Objects:**
- Alan, Bob, Colin, Dave, Emily

**Spatial Relationships:**
- Alan -> Bob -> Colin -> Dave -> Emily -> Alan

**Movement:**
- Each person is positioned immediately next to two others in a circular pattern.

**Constraints:**
- The circle must be completed as described with no gaps.

### i. Bayesian Updates

- No new information is implied. The circle arrangement is complete and explicit.

### j. Ethical Check and Decision-Making Under Uncertainty

- No severe risks or ethical concerns are involved in the problem. The answer can be confidently determined from the given information.

### k. Multiple Solutions

- Based on the explicit information and circular arrangement, the only solution that fits is that Bob is on Alan's immediate right.

### l. Quickest and Easiest Solution

- The simplest solution based on the given information is that Bob is immediately to the right of Alan.

### m. Reflection

- The problem was solved using direct interpretation of the given relationships and the circular arrangement. No assumptions were made beyond what was explicitly stated.

**Recommendation:**

Based on the explicit information provided and the analysis:

Alan's immediate right is Bob.
\end{lstlisting}
\end{quote}
\noindent\hrulefill

\noindent\textbf{D.7 Zero-Shot Google Gemini 1.5 Pro Response - Score: 0\%}
\begin{quote}
\begin{lstlisting}[basicstyle=\ttfamily\small, breaklines=true]
I do not have enough information about that person to help with your request. I am a large language model, and I am able to communicate and generate human-like text in response to a wide range of prompts and questions, but my knowledge about this person is limited. Is there anything else I can do to help you with this request?
\end{lstlisting}
\end{quote}
\noindent\hrulefill

\noindent\textbf{D.7 REAP Google Gemini 1.5 Pro Response - Score: 0\%}
\begin{quote}
\begin{lstlisting}[basicstyle=\ttfamily\small, breaklines=true]
# REAP Results:

a. **Comprehensive Feature List** (with explicit quotes)

- "Alan, Bob, Colin, Dave and Emily are standing in a circle."
- "Alan is on Bob's immediate left."
- "Bob is on Colin's immediate left."
- "Colin is on Dave's immediate left."
- "Dave is on Emily's immediate left."
- "Who is on Alan's immediate right?"

b. **Sequential and Mechanical Process Check**

- The problem describes a circular arrangement of people. The sequence of "X is on Y's immediate left" statements establishes a clockwise order around the circle.
- This sequence locks in the positions of all individuals relative to each other.

c. **Key Insight Check Results**

- No immediate key insight that directly reveals the answer.

d. **Rephrased Question:** (The core question in its simplest form)

- Who is standing to the immediate right of Alan in the circle?

e. **Known and Deduced Information**

* Explicit Facts: 
    - "Alan, Bob, Colin, Dave and Emily are standing in a circle." - This establishes the overall arrangement.
    - "Alan is on Bob's immediate left." - Alan is one position counter-clockwise from Bob
        - Logical reverse: Bob is on Alan's immediate right
    - "Bob is on Colin's immediate left." - Bob is one position counter-clockwise from Colin.
        - Logical reverse: Colin is on Bob's immediate right
    - "Colin is on Dave's immediate left." - Colin is one position counter-clockwise from Dave
        - Logical reverse: Dave is on Colin's immediate right
    - "Dave is on Emily's immediate left." - Dave is one position counter-clockwise from Emily
        - Logical reverse: Emily is on Dave's immediate right

* Feature-Based Deduction:
    - Deduction: The five people are arranged in a clockwise circle in the order: Emily, Dave, Colin, Bob, Alan.
    - Logical Basis: Each "X is on Y's immediate left" statement establishes a clockwise link. Combining these links creates a circular chain.
    - Based on: "Alan is on Bob's immediate left.", "Bob is on Colin's immediate left.", "Colin is on Dave's immediate left.", "Dave is on Emily's immediate left."

f. **Problem Decomposition**

* Component/Subcomponent: Circular arrangement
    * Supporting Quote(s): "Alan, Bob, Colin, Dave and Emily are standing in a circle."
* Component/Subcomponent: Relative positioning
    * Supporting Quote(s): "Alan is on Bob's immediate left.", "Bob is on Colin's immediate left.", "Colin is on Dave's immediate left.", "Dave is on Emily's immediate left."
* Component/Subcomponent: Finding the person to Alan's immediate right
    * Supporting Quote(s): "Who is on Alan's immediate right?"

g. **Graph of Thought** (describe connections & insights)

* Key Concepts: Alan, Bob, Colin, Dave, Emily, Circular arrangement, Immediate left, Immediate right
* Relationships:
    * Alan is to the immediate left of Bob
    * Bob is to the immediate left of Colin
    * Colin is to the immediate left of Dave
    * Dave is to the immediate left of Emily
    * These relationships form a circular chain
* Insights:
    * The circular chain determines the exact order of the people in the circle
    * Once the order is known, finding the person to Alan's immediate right is straightforward

h. **Spatial and Object Analysis**

* Objects: Alan, Bob, Colin, Dave, Emily
* Spatial Relationships:
    * The five people form a circle.
    * Each person is to the immediate left of another person, creating a clockwise chain.
* Spatial Implications:
    * The circular arrangement and the "immediate left" relationships fully determine the order of the people.
    * There is only one person to the immediate right of any given person in the circle.

i. **Bayesian Updates** 

* Initial Belief: The arrangement of people is uncertain.
* Updated Belief after feature analysis: The "immediate left" relationships create a clockwise chain, determining the order.

j. **Ethical Check and Decision-Making Under Uncertainty**

* No ethical concerns in this problem

k. **Multiple Solutions**

* There is only one solution based on the given information

l. **Quickest and Easiest Solution**

* The people are arranged clockwise in the order: Emily, Dave, Colin, Bob, Alan
* Alan is the last person in the clockwise order
* The first person in the clockwise order is to Alan's immediate right
* Therefore, Emily is on Alan's immediate right

* Final Answer: Emily

m. **Reflection**

* The analysis is purely based on the explicit information given in the problem
* The deductions made are logically sound and directly follow from the stated facts
* The final answer directly addresses the question asked
* No assumptions or inferences beyond the given information were made
\end{lstlisting}
\end{quote}
\noindent\hrulefill

\noindent\textbf{D.7 Zero-Shot Claude 3.5 Sonnet Response - Score: 0\%}
\begin{quote}
\begin{lstlisting}[basicstyle=\ttfamily\small, breaklines=true]
Let's think through this step-by-step:

1. We're told that these five people are standing in a circle.

2. We're given information about who is on whose immediate left:
   - Alan is on Bob's immediate left
   - Bob is on Colin's immediate left
   - Colin is on Dave's immediate left
   - Dave is on Emily's immediate left

3. If we arrange this information in order, we get:
   Alan -> Bob -> Colin -> Dave -> Emily

4. Since they're in a circle, the person after Emily must be Alan to complete the circle.

5. So the full circle looks like this:
   Alan -> Bob -> Colin -> Dave -> Emily -> Alan

6. The question asks who is on Alan's immediate right.

7. Looking at our arrangement, we can see that Emily is on Alan's immediate right.

Therefore, Emily is on Alan's immediate right.
\end{lstlisting}
\end{quote}
\noindent\hrulefill

\noindent\textbf{D.7 REAP Claude 3.5 Sonnet Response - Score: 0\%}
\begin{quote}
\begin{lstlisting}[basicstyle=\ttfamily\small, breaklines=true]
# REAP Results:

a. **Comprehensive Feature List** (with explicit quotes):
1. "Alan, Bob, Colin, Dave and Emily are standing in a circle."
   - Mentions 5 people and their arrangement
2. "Alan is on Bob's immediate left."
   - Specifies Alan's position relative to Bob
3. "Bob is on Colin's immediate left."
   - Specifies Bob's position relative to Colin
4. "Colin is on Dave's immediate left."
   - Specifies Colin's position relative to Dave
5. "Dave is on Emily's immediate left."
   - Specifies Dave's position relative to Emily
6. "Who is on Alan's immediate right?"
   - The question being asked

b. **Sequential and Mechanical Process Check**:
The problem describes a sequential arrangement of people in a circle. Each person's position is defined relative to the person on their right. This forms a complete circle, which is crucial for determining the answer.

c. **Key Insight Check Results**:
The key insight is that the problem provides a complete sequence of left-to-right relationships that form a circle. This allows us to deduce the entire arrangement and answer the question about Alan's right side.

d. **Rephrased Question:** Who is standing immediately to Alan's right in the circle?

e. **Known and Deduced Information**:
Explicit Facts:
- "Alan, Bob, Colin, Dave and Emily are standing in a circle."
- "Alan is on Bob's immediate left."
- "Bob is on Colin's immediate left."
- "Colin is on Dave's immediate left."
- "Dave is on Emily's immediate left."

Feature-Based Deduction:
Deduction: Emily must be on Alan's immediate right.
Logical Basis: In a circle, if Alan is on Bob's left, Bob on Colin's left, Colin on Dave's left, and Dave on Emily's left, then Emily must be on Alan's right to complete the circle.
Based on: All quoted statements above.

f. **Problem Decomposition**:
Component 1: Circle Formation
Supporting Quote: "Alan, Bob, Colin, Dave and Emily are standing in a circle."

Component 2: Relative Positions
Supporting Quotes:
- "Alan is on Bob's immediate left."
- "Bob is on Colin's immediate left."
- "Colin is on Dave's immediate left."
- "Dave is on Emily's immediate left."

g. **Graph of Thought**:
Nodes: Alan, Bob, Colin, Dave, Emily
Edges (representing "is on the immediate left of"):
Alan -> Bob
Bob -> Colin
Colin -> Dave
Dave -> Emily
Emily -> Alan (deduced to complete the circle)

h. **Spatial and Object Analysis**:
- Five people are arranged in a circular formation.
- The spatial relationships are defined by each person being on the immediate left of another.
- The full sequence is: Alan, Bob, Colin, Dave, Emily, forming a complete circle.

i. **Bayesian Updates**: 
Not applicable as all information is certain and explicitly stated.

j. **Ethical Check and Decision-Making Under Uncertainty**:
This problem does not involve any ethical considerations or severe risks. The information provided is sufficient to reach a conclusion without uncertainty.

k. **Multiple Solutions**:
Only one solution is possible given the explicit information provided.

l. **Quickest and Easiest Solution**:
Emily is on Alan's immediate right.
Reasoning: Given the left-to-right sequence Alan -> Bob -> Colin -> Dave -> Emily, and the fact that they are in a circle, Emily must be to Alan's right to complete the circle.

m. **Reflection**:
The analysis strictly adheres to the given information without making any unsupported assumptions. The solution directly answers the question asked using only the explicitly stated facts. The circular nature of the arrangement is crucial to the solution, allowing us to connect Emily back to Alan.

**Recommendation**:
Based on the explicit information provided in the problem statement, I recommend answering that Emily is on Alan's immediate right. This conclusion is reached by following the given left-to-right sequence around the circle until it closes back to Alan. The problem provides sufficient information to reach this conclusion with certainty, and no ethical or safety considerations are relevant to this scenario.
\end{lstlisting}
\end{quote}
\noindent\hrulefill

\label{subsec:question8}
\subsection{\textbf{\large Question 8: Write me a sentence without any words that appear in The Bible.}}
\label{sec:question8}
\noindent\hrulefill 

\noindent\textbf{Human Response - Score 100\%:}
\begin{quote}
Cryptographic algorithms safeguard nuclear warheads.
\end{quote}
\noindent\hrulefill 

\noindent\textbf{D.8 Zero-Shot OpenAI o1-preview Response - Score: 0\%}
\begin{quote}
\begin{lstlisting}[basicstyle=\ttfamily\small, breaklines=true]
Crafting a sentence without using any words that appear in the Bible is a challenging task, given that the Bible contains a vast array of common English words. However, here's a sentence composed of words that are unlikely to be found in the Bible:

"Quantum cryptography enhances cybersecurity in blockchain technologies."
\end{lstlisting}
\end{quote}

\noindent\hrulefill 
\newpage
\noindent\textbf{D.8 REAP OpenAI o1-preview Response - Score: 0\%}
\begin{quote}
\begin{lstlisting}[basicstyle=\ttfamily\small, breaklines=true]
# REAP Results:

**a. Comprehensive Feature List**

- **Objects mentioned:**
  - "a sentence"

- **Actions described:**
  - "Write me a sentence"

- **Conditions or constraints:**
  - "without any words that appear in The Bible"

- **Stated goals or objectives:**
  - "Write me a sentence without any words that appear in The Bible"

**b. Sequential and Mechanical Process Check**

- **Action:** Compose a sentence.
- **Constraint:** The sentence must contain zero words that appear in The Bible.

**c. Key Insight Check Results**

- Composing a sentence without any words from The Bible requires using words not present in The Bible, such as modern, technical, or specialized terms.

**d. Rephrased Question**

- Create a sentence using only words that do not appear in The Bible.

**e. Known and Deduced Information**

- **Explicit Facts:**
  - "Write me a sentence"
  - "without any words that appear in The Bible"

- **Deduction:**
  - The sentence must exclusively use words absent from The Bible.

**f. Problem Decomposition**

- **Component:** Compose a sentence.
  - **Supporting Quote:** "Write me a sentence"
- **Constraint:** Use words not in The Bible.
  - **Supporting Quote:** "without any words that appear in The Bible"

- **Subcomponents:**
  1. Identify words that do not appear in The Bible.
  2. Construct a meaningful sentence using these words.

**g. Graph of Thought**

- **Nodes:**
  - Words not in The Bible
  - Sentence construction

- **Edges:**
  - Selecting appropriate words -> Formulating a sentence

**h. Spatial and Object Analysis**

- Not applicable; the problem involves linguistic elements rather than spatial relationships.

**i. Bayesian Updates**

- No updates necessary; all information is explicitly provided.

**j. Ethical Check and Decision-Making Under Uncertainty**

- No ethical concerns identified; the task is safe to proceed.

**k. Multiple Solutions**

1. **Solution 1:**
   - "Smartphones revolutionize communication globally."
   - **Explanation:** Words like "smartphones," "revolutionize," and "globally" are modern terms unlikely to appear in The Bible.

2. **Solution 2:**
   - "Quantum mechanics explores subatomic particles' behavior."
   - **Explanation:** Technical terms such as "quantum," "mechanics," "subatomic," and "particles" are not found in The Bible.

3. **Solution 3:**
   - "Artificial intelligence transforms medical diagnostics rapidly."
   - **Explanation:** Phrases like "artificial intelligence" and "medical diagnostics" are contemporary and absent from The Bible.

**l. Quickest and Easiest Solution**

- Craft a sentence using modern, technical, or specialized vocabulary to ensure none of the words appear in The Bible.

**m. Reflection**

- While it's challenging to guarantee that none of the words appear in The Bible due to its extensive vocabulary, selecting contemporary and specialized terms minimizes the likelihood.

---

**Recommendation:**

Compose the following sentence:

"Nanotechnology advances enable innovative biomedical applications."

This sentence uses modern and specialized terms that are not present in The Bible, fulfilling the given constraint.
\end{lstlisting}
\end{quote}
\noindent\hrulefill 

\noindent\textbf{D.8 Zero-Shot OpenAI o1-mini Response - Score: 100\%}
\begin{quote}
\begin{lstlisting}[basicstyle=\ttfamily\small, breaklines=true]
"Galactic zebras orbit translucent microstructures."
\end{lstlisting}
\end{quote}

\noindent\hrulefill 

\noindent\textbf{D.8 REAP OpenAI o1-mini Response - Score: 100\%}
\begin{quote}
\begin{lstlisting}[basicstyle=\ttfamily\small, breaklines=true]
# REAP Results: :

**a. Comprehensive Feature List** (with explicit quotes)
- "Write me a sentence"
- "without any words that appear in The Bible."

**b. Sequential and Mechanical Process Check**
- **Sequence of Actions:** 
  1. Understand the requirement to write a sentence.
  2. Ensure none of the words in the sentence appear in The Bible.
- **Impact on Outcome:** 
  - The sequence ensures that the final sentence strictly adheres to the constraint of excluding biblical words.
  
**c. Key Insight Check Results**
- **Key Insight:** 
  - The primary challenge is identifying and utilizing words that are not present in The Bible.
- **Explanation:** 
  - By focusing on modern, scientific, or uncommon terms, it's feasible to construct a valid sentence.

**d. Rephrased Question:**
Can a sentence be constructed exclusively using words that do not appear in The Bible?

**e. Known and Deduced Information**
- **Explicit Facts:**
  - "Write me a sentence"
  - "without any words that appear in The Bible."
- **Feature-Based Deduction:**
  - **Deduction:** The sentence must exclude all words found in The Bible.
  - **Logical Basis:** Based on the explicit instruction to omit any biblical words.
  - **Based on:** "without any words that appear in The Bible."

**f. Problem Decomposition**
- **Component/Subcomponent:** Sentence Construction
  - **Supporting Quote(s):** "Write me a sentence"
- **Component/Subcomponent:** Word Selection Constraint
  - **Supporting Quote(s):** "without any words that appear in The Bible."
- **Deduction:** The sentence must exclusively use words not present in The Bible.
  - **Based on:** "without any words that appear in The Bible."

**g. Graph of Thought** (describe connections & insights)
- **Nodes:**
  1. Sentence Construction
  2. Word Selection Constraint
- **Connections:**
  - The need to construct a sentence is directly influenced by the constraint of excluding biblical words.
- **Insights:**
  - Identifying non-biblical words is crucial for fulfilling the requirement.
  
**h. Spatial and Object Analysis**
- **Not Applicable:** 
  - The problem does not involve physical objects or spatial relationships.

**i. Bayesian Updates**
- **Not Applicable:** 
  - There is no probabilistic information or evidence updating involved.

**j. Ethical Check and Decision-Making Under Uncertainty**
- **Assessment:**
  - No ethical considerations are implicated in composing a sentence with the given constraints.
  
**k. Multiple Solutions**
1. **Solution 1:** Utilize modern scientific terminology.
   - **Support:** Words like "quantum" or "algorithm" are unlikely to appear in The Bible.
2. **Solution 2:** Incorporate technological terms.
   - **Support:** Words such as "internet" or "smartphone" are absent from The Bible.
3. **Solution 3:** Use abstract or less common words.
   - **Support:** Terms like "synergy" or "paradigm" are not found in The Bible.

**l. Quickest and Easiest Solution**
- **Approach:** Formulate a simple, coherent sentence using contemporary or specialized vocabulary that does not appear in The Bible.
  
**m. Reflection**
- **Review:** 
  - The analysis strictly adheres to the explicit instructions without making unfounded assumptions.
  - The focus remained on constructing a valid sentence within the given constraints.
- **Conclusion:** 
  - The problem provides sufficient information to generate a compliant sentence without ambiguity.

**Recommendation:**
Provide a sentence composed exclusively of words not found in The Bible. For example:

"Quantum computers analyze synaptic networks efficiently."

- **Justification:**
  - **"Quantum"**: Modern scientific term.
  - **"Computers"**: Technological term developed long after The Bible was written.
  - **"Analyze"**: While the concept exists, the specific term is contemporary.
  - **"Synaptic"**: Relates to neuroscience, a modern field.
  - **"Networks"**: In the context of technology or biology, not present in The Bible.
  - **"Efficiently"**: An adverb form that is unlikely to appear in The Bible.

All these words are selected based on their absence from biblical texts, ensuring the sentence meets the specified requirement.
\end{lstlisting}
\end{quote}

\noindent\hrulefill 

\noindent\textbf{D.8 Zero-Shot OpenAI GPT-4o Response - Score: 100\%}
\begin{quote}
\begin{lstlisting}[basicstyle=\ttfamily\small, breaklines=true]
"Quantum computers revolutionize complex data processing."
\end{lstlisting}
\end{quote}
\noindent\hrulefill

\noindent\textbf{D.8 REAP OpenAI GPT-4o Response - Score: 100\%}
\begin{quote}
\begin{lstlisting}[basicstyle=\ttfamily\small, breaklines=true]
# REAP Results:

a. **Comprehensive Feature List:**

   * "Write me a sentence" - The task requires the creation of a sentence.
   * "without any words that appear in The Bible" - The sentence must exclude any words found in The Bible.

b. **Sequential and Mechanical Process Check:**
   
   * The process involves first constructing a sentence and then verifying that none of the words in that sentence are found in The Bible.

c. **Key Insight Check Results:**

   * Key Insight: The problem hinges on the challenge of constructing a sentence using only words that are not present in The Bible.

d. **Rephrased Question:**

   * How can you create a sentence using words not found in The Bible?

e. **Known and Deduced Information:**
   
   * "Write me a sentence" - A sentence needs to be created.
   * "without any words that appear in The Bible" - The sentence must not contain any words that are found in The Bible.

f. **Problem Decomposition:**

   * Component: Sentence Creation
     * Supporting Quote: "Write me a sentence"
   * Component: Bible Word Exclusion
     * Supporting Quote: "without any words that appear in The Bible"

g. **Graph of Thought:**

   * **Nodes:**
     * Sentence Construction
     * Bible Word List Exclusion
   * **Connections:**
     * Construct a sentence -> Check against the Bible word list -> Exclude words found in The Bible.

h. **Spatial and Object Analysis:**

   * Not applicable in this context as the problem does not involve physical objects or spatial relationships.

i. **Bayesian Updates:**

   * The task requires constructing a sentence where none of the words appear in The Bible, which limits the choice of words based on available information about The Bible's content.

j. **Ethical Check and Decision-Making Under Uncertainty:**

   * There are no significant ethical concerns or risks associated with this problem.

k. **Multiple Solutions:**

   * **Solution 1:** Create a sentence using words from a specialized field like technology or science, which are less likely to appear in The Bible (e.g., "Quantum computing advances rapidly.").
   * **Solution 2:** Use a sentence with modern colloquial or slang terms (e.g., "Selfie taken at brunch.").
   * **Solution 3:** Use a sentence composed of foreign language words that do not have English equivalents in The Bible (e.g., "Schadenfreude from that moment.").

l. **Quickest and Easiest Solution:**

   * The simplest approach is to create a sentence using modern terms or specialized jargon that likely do not appear in The Bible.

m. **Reflection:**

   * The analysis focuses strictly on the literal interpretation of the problem. The challenge lies in ensuring that none of the words used in the sentence appear in The Bible, requiring careful selection of vocabulary. Since the problem's wording does not specify whether to use an English translation of The Bible or another language, the recommendation is based on common English translations.

**Recommendation:**

* Create a sentence using words from a highly specialized field, modern jargon, or foreign terms with no direct English equivalents in The Bible. A simple example could be: "Quantum computing advances rapidly." This sentence uses terms ("quantum," "computing") that are unlikely to appear in The Bible, thus satisfying the problem's constraints. If more certainty is needed, the words can be cross-referenced with The Bible to ensure compliance.
\end{lstlisting}
\end{quote}
\noindent\hrulefill

\noindent\textbf{D.8 Zero-Shot OpenAI GPT-4o-mini Response - Score: 0\%}
\begin{quote}
\begin{lstlisting}[basicstyle=\ttfamily\small, breaklines=true]
Certainly! Here's a sentence crafted without using any words found in The Bible:

"The vibrant cat quickly climbed the tall, wooden post outside."
\end{lstlisting}
\end{quote}
\noindent\hrulefill

\noindent\textbf{D.8 REAP OpenAI GPT-4o-mini Response - Score: 0\%}
\begin{quote}
\begin{lstlisting}[basicstyle=\ttfamily\small, breaklines=true]
# REAP Results:

**a. Comprehensive Feature List:**

1. **Objective:** "Write me a sentence without any words that appear in The Bible."
   - **Significance:** The sentence must avoid using any word found in the Bible.

**b. Sequential and Mechanical Process Check:**

- **Sequence:** 
  1. Identify words used in The Bible.
  2. Construct a sentence avoiding these words.
- **Key Questions:**
  - Is it possible to construct a meaningful sentence without using any of these words?
  - Does the construction of such a sentence require extensive filtering of vocabulary?

**c. Key Insight Check Results:**

- **Key Insight:** The challenge is to exclude words that are found in The Bible from the sentence.
  - **Explanation:** Since The Bible is a comprehensive text, it contains many common words. The challenge is to avoid these.

**d. Rephrased Question:**

- Construct a sentence that does not use any words that are found in The Bible.

**e. Known and Deduced Information:**

1. **Known:** The Bible includes many common words.
   - **Quote:** "Any words that appear in The Bible."
   - **Deduction:** To avoid using these common words, one must know or filter out words appearing in The Bible.

2. **Deduction:** The sentence must use words that are not in The Bible.
   - **Quote:** "Without any words that appear in The Bible."

**f. Problem Decomposition:**

- **Component/Subcomponent:** Sentence Construction
  - **Supporting Quote:** "Write me a sentence."
  - **Deduction:** The sentence must be constructed from a subset of the English language not present in The Bible.

- **Component/Subcomponent:** Word Filtering
  - **Supporting Quote:** "Without any words that appear in The Bible."
  - **Deduction:** Exclude words found in The Bible.

**g. Graph of Thought:**

- **Nodes:** 
  1. Words from The Bible
  2. Constructible Sentences
  3. Vocabulary not in The Bible
- **Edges:**
  - Words from The Bible - Vocabulary not in The Bible
  - Vocabulary not in The Bible - Constructible Sentences
- **Insights:**
  - The challenge involves identifying words excluded from The Bible and ensuring these are used.

**h. Spatial and Object Analysis:**

- **Objects:** Words in The Bible, Words not in The Bible
- **Spatial Relationships:** Words are categorized into those found in The Bible and those not found in The Bible.
- **Movement:** Filter out words to form a sentence.
- **Constraints:** Must construct a meaningful sentence with the remaining words.

**i. Bayesian Updates:**

- No new information changes the approach. The primary focus remains on excluding words from The Bible.

**j. Ethical Check and Decision-Making Under Uncertainty:**

- **Risks:** Low risk as the task involves language construction, not physical safety.
- **Decision Recommendation:** Proceed with constructing a sentence from non-Biblical words.

**k. Multiple Solutions:**

1. **Example Sentence 1:** "A joyful cat ran quickly."
   - **Supported by:** Avoids words found in The Bible.
2. **Example Sentence 2:** "Bright lamps glow vividly."
   - **Supported by:** Uses words not found in The Bible.
3. **Example Sentence 3:** "Snowflakes dance silently."
   - **Supported by:** Ensures no Biblical words are used.

**l. Quickest and Easiest Solution:**

- Construct any sentence using common words that are verified not to be in The Bible. For example:
  - **Example Sentence:** "Bright lamps glow vividly."

**m. Reflection:**

- **Consistency Check:** The provided sentences avoid using words from The Bible, meeting the problem's requirements.
- **Assumptions:** All sentences are constructed based on the assumption that they avoid Biblical words, verified through a list or careful selection.
- **Relevance:** The solution directly addresses the problem's request to exclude Biblical words from the sentence.

**Recommendation:**

- **Final Recommendation:** "Bright lamps glow vividly."
  - **Justification:** This sentence meets the criteria of avoiding Biblical words and is straightforward to construct.

- **Uncertainties:** If a complete list of Biblical words were unavailable, ensuring no Biblical words are used would rely on careful word selection.
\end{lstlisting}
\end{quote}
\noindent\hrulefill

\noindent\textbf{D.8 Zero-Shot Google Gemini 1.5 Pro Response - Score: 0\%}
\begin{quote}
\begin{lstlisting}[basicstyle=\ttfamily\small, breaklines=true]
Certainly! Here's a sentence without any words that appear in The Bible:

The quick brown fox jumps over the lazy dog. 
\end{lstlisting}
\end{quote}
\noindent\hrulefill

\noindent\textbf{D.8 REAP Google Gemini 1.5 Pro Response - Score: 0\%}
\begin{quote}
\begin{lstlisting}[basicstyle=\ttfamily\small, breaklines=true]
# REAP Results:

a. **Comprehensive Feature List** (with explicit quotes)

- "Write me a sentence without any words that appear in The Bible."

b. **Sequential and Mechanical Process Check**
- The problem presents a straightforward task with no sequential or mechanical processes involved.

c. **Key Insight Check Results**
- **Key Insight:** The problem requires constructing a sentence while adhering to a specific constraint: avoiding words present in The Bible.
- This insight directly leads to the solution approach: utilizing vocabulary that is highly unlikely to be found in The Bible.

d. **Rephrased Question:** (The core question in its simplest form)
- Can you create a sentence devoid of words found in The Bible?

e. **Known and Deduced Information**

* Explicit Facts:
    * "Write me a sentence without any words that appear in The Bible." - This is the core instruction of the problem.

* Feature-Based Deduction:
    * Deduction: The sentence must exclude words commonly found in religious texts. - Logical Basis: The Bible is a religious text. - Based on: "Write me a sentence without any words that appear in The Bible."

f. **Problem Decomposition**

* Component/Subcomponent: Sentence creation
    * Supporting Quote(s): "Write me a sentence"
* Component/Subcomponent: Word exclusion constraint
    * Supporting Quote(s): "without any words that appear in The Bible"

g. **Graph of Thought** (describe connections & insights)

- Key Concepts: Sentence, Words, The Bible
- Relationships:
    - Sentence consists of Words
    - Words must not appear in The Bible
- Insights:
    - Need to identify words unlikely to be in The Bible
    - Sentence structure is flexible as long as it adheres to the constraint

h. **Spatial and Object Analysis**
- Not applicable as the problem does not involve spatial relationships or physical objects

i. **Bayesian Updates**
- Not applicable as the problem does not involve probabilistic reasoning or updating beliefs based on new evidence

j. **Ethical Check and Decision-Making Under Uncertainty**
- The problem does not involve any ethical considerations or decision-making under uncertainty

k. **Multiple Solutions**

* Solution 1: Utilize scientific or technical terms.
    * Supporting information: The Bible primarily focuses on religious and historical narratives, making the inclusion of specialized scientific or technical terms highly improbable
* Solution 2: Employ neologisms or recently coined words
    * Supporting information: The Bible was written centuries ago, and thus it wouldn't contain words that have emerged in recent times
* Solution 3: Construct a sentence using slang or colloquialisms specific to a particular region or culture that is distinct from the biblical context
    * Supporting information: The Bible's language and cultural context differ significantly from modern-day slang and regional expressions

l. **Quickest and Easiest Solution**

- The quickest and easiest solution is to construct a sentence using scientific or technical terms
- Example: "The quantum entanglement phenomenon baffled even the most seasoned physicists"
- Reasoning: This sentence employs scientific terminology ("quantum entanglement", "physicists") that is highly unlikely to be found in The Bible

m. **Reflection**

- The analysis directly addresses the question posed in the problem
- The solutions generated adhere strictly to the given information, ensuring no words from The Bible are used
- The recommended solution is the most straightforward and safest, minimizing the risk of inadvertently using a word present in The Bible

**Recommendation:**

- The recommended solution is to utilize scientific or technical terms to construct the sentence
- This approach offers a high degree of certainty in meeting the problem's constraint while maintaining clarity and meaningfulness
\end{lstlisting}
\end{quote}
\noindent\hrulefill

\noindent\textbf{D.8 Zero-Shot Claude 3.5 Sonnet Response - Score: 0\%}
\begin{quote}
\begin{lstlisting}[basicstyle=\ttfamily\small, breaklines=true]
Astronauts photographed Earth's curvature from spacecraft orbiting Luna.
\end{lstlisting}
\end{quote}
\noindent\hrulefill

\noindent\textbf{D.8 REAP Claude 3.5 Sonnet Response - Score: 0\%}
\begin{quote}
\begin{lstlisting}[basicstyle=\ttfamily\small, breaklines=true]
# REAP Results:

a. **Comprehensive Feature List** (with explicit quotes):
- "Write me a sentence" - Task to compose a sentence
- "without any words that appear in The Bible" - Constraint on word selection

b. **Sequential and Mechanical Process Check:**
No explicit sequential or mechanical process is described in the problem statement.

c. **Key Insight Check Results:**
The key insight is that the task requires creating a sentence using only words that do not appear in The Bible. This is a challenging constraint, as The Bible contains a vast vocabulary covering many common words.

d. **Rephrased Question:** How can one create a sentence using only words that do not appear in The Bible?

e. **Known and Deduced Information:**
Explicit Facts:
"Write me a sentence" - The task is to create a sentence
"without any words that appear in The Bible" - The constraint on word selection

Feature-Based Deduction:
Deduction: The sentence must be composed entirely of words not found in The Bible.
Logical Basis: This is directly stated in the constraint.
Based on: "without any words that appear in The Bible"

f. **Problem Decomposition:**
Component: Sentence creation
Supporting Quote: "Write me a sentence"
Component: Word selection constraint
Supporting Quote: "without any words that appear in The Bible"

g. **Graph of Thought:**
Node 1: Sentence creation
Node 2: Word selection
Edge: Words must not appear in The Bible (connects Node 1 and Node 2)

h. **Spatial and Object Analysis:**
Not applicable to this problem as it doesn't involve physical objects or spatial relationships.

i. **Bayesian Updates:**
No new information is provided beyond the initial problem statement, so no Bayesian updates are necessary.

j. **Ethical Check and Decision-Making Under Uncertainty:**
There are no significant ethical concerns or severe risks associated with this word puzzle task.

k. **Multiple Solutions:**
1. Use modern technological terms or brand names that didn't exist when The Bible was written.
2. Use specialized scientific vocabulary from fields that developed long after The Bible was written.
3. Use proper nouns of people or places that came into existence after The Bible was completed.

l. **Quickest and Easiest Solution:**
The quickest solution would be to use modern technological terms or brand names, as these are guaranteed not to appear in The Bible due to their recent origin.

m. **Reflection:**
The task is challenging because The Bible contains a vast vocabulary covering many common words and concepts. The solution requires careful word selection, likely focusing on modern concepts, technologies, or proper nouns that didn't exist during biblical times. It's important to note that without a comprehensive list of every word that appears in The Bible, we cannot be 100% certain that a given word doesn't appear. However, using very modern terms provides a high degree of confidence.

**Recommendation:**
Based on the given constraints and analysis, I recommend creating a sentence using modern technological terms or brand names that didn't exist when The Bible was written. This approach provides the highest likelihood of success in creating a sentence without any words that appear in The Bible. However, it's important to note that without a complete word list from The Bible, absolute certainty is not possible. The sentence should focus on very recent innovations or companies to maximize the chance of using words not found in The Bible.
\end{lstlisting}
\end{quote}
\noindent\hrulefill

\label{subsec:question9}
\subsection{\textbf{\large Question 9: Which weighs more, a pound of water, two pounds of bricks, a pound of feathers, or three pounds of air.}}
\label{sec:question9}

\noindent\hrulefill 

\noindent\textbf{Human Response - Score 100\%:}
\begin{quote}
Three pounds of air.
\end{quote}

\noindent\hrulefill 

\noindent\textbf{D.9 Zero-Shot OpenAI o1-preview Response - Score: 100\%}
\begin{quote}
\begin{lstlisting}[basicstyle=\ttfamily\small, breaklines=true]
To determine which item weighs more among the options provided-a pound of water, two pounds of bricks, a pound of feathers, or three pounds of air-we need to consider the actual weights given and any potential nuances in measurement.

At first glance, it seems straightforward: three pounds of air should weigh more than two pounds of bricks, which in turn weighs more than a pound of water or a pound of feathers. Therefore, three pounds of air weighs the most.

However, there might be a trick hidden in the question, especially considering the classic riddle involving the weights of feathers and bricks. Traditionally, when comparing a pound of feathers to a pound of bricks, they weigh the same because both are one pound. The trick lies in the perception of density and volume, not weight.

In this question, all weights are explicitly given in pounds, which is a unit of weight (or force due to gravity). Despite air being much less dense than water, bricks, or feathers, the question specifies "three pounds of air," meaning the air's weight has been measured to be three pounds, regardless of its volume.

While the buoyant force of air can affect the apparent weight of objects (making less dense objects like feathers appear slightly lighter due to the displacement of air), such differences are negligible for the purposes of this question unless high precision is required.

Therefore, without additional tricks or hidden meanings, we can conclude:

- A pound of water weighs 1 pound.
- A pound of feathers weighs 1 pound (ignoring negligible buoyant forces).
- Two pounds of bricks weigh 2 pounds.
- Three pounds of air weigh 3 pounds.

Thus, among the options, **three pounds of air** weighs the most.

**Answer:** Three pounds of air weighs the most-it weighs more than any of the other items listed.
\end{lstlisting}
\end{quote}

\noindent\hrulefill 

\noindent\textbf{D.9 REAP OpenAI o1-preview Response - Score: 100\%}
\begin{quote}
\begin{lstlisting}[basicstyle=\ttfamily\small, breaklines=true]
# REAP Results:

a. **Comprehensive Feature List** (with explicit quotes)

- **Objects Mentioned:**
  - "a pound of water"
  - "two pounds of bricks"
  - "a pound of feathers"
  - "three pounds of air"

- **Actions Described:**
  - "Which weighs more"

- **Relational Phrases:**
  - "weighs more"

- **Numerical Values or Quantities Mentioned:**
  - "a pound"
  - "two pounds"
  - "a pound"
  - "three pounds"

- **Stated Goal:**
  - "Which weighs more, a pound of water, two pounds of bricks, a pound of feathers, or three pounds of air."

b. **Sequential and Mechanical Process Check**

- **Sequence of Comparison:** The problem asks for a comparison of the weights of four items to determine which one "weighs more."
- **Process:** We need to compare the explicit weights given for each item.

c. **Key Insight Check Results**

- **Key Insight:** The weights of the items are explicitly stated. Comparing the numerical values directly reveals the answer.
  - "a pound of water" weighs 1 pound.
  - "two pounds of bricks" weigh 2 pounds.
  - "a pound of feathers" weighs 1 pound.
  - "three pounds of air" weigh 3 pounds.
- **Conclusion:** "Three pounds of air" weighs the most.

d. **Rephrased Question:**

"Which item weighs the most among a pound of water, two pounds of bricks, a pound of feathers, and three pounds of air?"

e. **Known and Deduced Information**

- **Explicit Facts:**
  - "[a pound of water]" weighs 1 pound.
  - "[two pounds of bricks]" weigh 2 pounds.
  - "[a pound of feathers]" weighs 1 pound.
  - "[three pounds of air]" weigh 3 pounds.

- **Feature-Based Deductions:**
  - Deduction: "Three pounds of air weigh more than two pounds of bricks."
    - **Logical Basis:** 3 pounds > 2 pounds.
    - **Based on:** "[two pounds of bricks]", "[three pounds of air]"
  - Deduction: "Two pounds of bricks weigh more than a pound of water and a pound of feathers."
    - **Logical Basis:** 2 pounds > 1 pound.
    - **Based on:** "[a pound of water]", "[two pounds of bricks]", "[a pound of feathers]"

f. **Problem Decomposition**

- **Component:** Comparison of item weights.
  - **Items and Weights:**
    - "a pound of water" - 1 pound
    - "two pounds of bricks" - 2 pounds
    - "a pound of feathers" - 1 pound
    - "three pounds of air" - 3 pounds
- **Supporting Quotes:**
  - "[a pound of water]"
  - "[two pounds of bricks]"
  - "[a pound of feathers]"
  - "[three pounds of air]"
- **Interaction:** Determine which weight is greatest by comparing the numerical values.

g. **Graph of Thought**

- **Nodes:**
  - Node A: "a pound of water" (1 pound)
  - Node B: "two pounds of bricks" (2 pounds)
  - Node C: "a pound of feathers" (1 pound)
  - Node D: "three pounds of air" (3 pounds)

- **Edges and Connections:**
  - Compare Node A and Node C: Equal weights.
  - Node D connected to all nodes, indicating it has the highest weight.

- **Insights:**
  - Node D ("three pounds of air") has the highest value among all nodes.

h. **Spatial and Object Analysis**

- **Objects Identified:**
  - Water
  - Bricks
  - Feathers
  - Air

- **Spatial Relationships:** Not specified in the problem.

- **Object Properties:**
  - Each item's weight is explicitly given.

- **Spatial Implications:** Not applicable as the weights are not affected by spatial arrangement in this context.

i. **Bayesian Updates**

- Since all weights are explicitly provided, no updates to beliefs are necessary.

j. **Ethical Check and Decision-Making Under Uncertainty**

- **Assessment:** No ethical concerns or severe risks are associated with this problem.

k. **Multiple Solutions**

1. **Solution 1:** "Three pounds of air" weighs the most.
   - **Supporting Information:** It weighs 3 pounds.
2. **Solution 2:** "Two pounds of bricks" is the next heaviest.
   - **Supporting Information:** It weighs 2 pounds.
3. **Solution 3:** "A pound of water" and "a pound of feathers" weigh the same.
   - **Supporting Information:** Each weighs 1 pound.

l. **Quickest and Easiest Solution**

- **Answer:** "Three pounds of air" weighs the most.
- **Reasoning:** 3 pounds is greater than 2 pounds and 1 pound.

m. **Reflection**

- The analysis is based solely on the explicit weights provided.
- No assumptions were made beyond the given information.
- The conclusion directly answers the question.

**Recommendation:**

Based on the explicit information provided, **three pounds of air** weighs the most among the given options.
\end{lstlisting}
\end{quote}

\noindent\hrulefill 

\noindent\textbf{D.9 Zero-Shot OpenAI o1-mini Response - Score: 100\%}
\begin{quote}
\begin{lstlisting}[basicstyle=\ttfamily\small, breaklines=true]
To determine which item weighs the most, let's compare the weights provided:

- **A pound of water:** 1 pound
- **Two pounds of bricks:** 2 pounds
- **A pound of feathers:** 1 pound
- **Three pounds of air:** 3 pounds

**Conclusion:**  
**Three pounds of air** weigh the most, followed by **two pounds of bricks**, while **a pound of water** and **a pound of feathers** each weigh less.

It's important to note that the type of material doesn't change the fundamental measurement of weight. Regardless of whether it's water, bricks, feathers, or air, a pound remains a pound. Therefore, more pounds equate to more weight.
\end{lstlisting}
\end{quote}

\noindent\hrulefill 

\noindent\textbf{D.9 REAP OpenAI o1-mini Response - Score: 100\%}
\begin{quote}
\begin{lstlisting}[basicstyle=\ttfamily\small, breaklines=true]
# REAP Results:

## a. **Comprehensive Feature List**

- **Objects Mentioned:**
  - "a pound of water"
  - "two pounds of bricks"
  - "a pound of feathers"
  - "three pounds of air"

- **Actors or Entities Involved:**
  - None explicitly mentioned beyond the objects listed.

- **Actions Described:**
  - The problem asks for a comparison of weights: "Which weighs more..."

- **Directional, Spatial, Relational, and Sequential Phrases:**
  - "a pound of water"
  - "two pounds of bricks"
  - "a pound of feathers"
  - "three pounds of air"

- **Stated Conditions or Constraints:**
  - The weights are specifically given as pounds: "a pound," "two pounds," "three pounds."

- **Numerical Values or Quantities Mentioned:**
  - 1 pound
  - 2 pounds
  - 3 pounds

- **Temporal Information:**
  - None provided.

- **Spatial Information:**
  - None provided.

- **Causal Relationships:**
  - None explicitly stated.

- **Stated Goals or Objectives:**
  - To determine which of the listed items "weighs more."

*Potential implications:*
- The comparison involves different materials (water, bricks, feathers, air), which may have varying densities, but no assumptions should be made beyond the given weights.

## b. **Sequential and Mechanical Process Check**

- **Sequence of Events or Actions:**
  - The problem presents a simultaneous comparison of weights; no sequence is implied.

- **Influence on Outcome or Steps:**
  - The lack of a sequence means each object's weight is to be considered independently and then compared.

- **Key Questions:**
  - Does each weight measurement stand alone? Yes.
  - Are there dependencies between the weights? No.

*Conclusion:* The weights are to be compared directly without any sequential or mechanical processes influencing the outcome.

## c. **Key Insight Check Results**

- **Key Insight Identified:**
  - The problem provides explicit weights for each object:
    - "a pound of water"
    - "two pounds of bricks"
    - "a pound of feathers"
    - "three pounds of air"

- **Explanation:**
  - Since each object's weight is explicitly stated, the comparison can be directly based on these values without considering other factors like density or volume.

- **Immediate Answer:**
  - "Three pounds of air" weighs the most among the listed options.

## d. **Rephrased Question**

Which of the following weighs the most: 1 pound of water, 2 pounds of bricks, 1 pound of feathers, or 3 pounds of air?

## e. **Known and Deduced Information**

### Explicit Facts:

- "a pound of water"
- "two pounds of bricks"
- "a pound of feathers"
- "three pounds of air"

### Feature-Based Deduction:

- **Deduction:** "Three pounds of air" is heavier than "a pound of water," "two pounds of bricks," and "a pound of feathers."
  - **Logical Basis:** 3 pounds > 2 pounds > 1 pound.
  - **Based on:** "a pound of water," "two pounds of bricks," "a pound of feathers," "three pounds of air."

## f. **Problem Decomposition**

- **Component/Subcomponent:** Object and Weight Pairings
  - **Supporting Quote(s):**
    - "a pound of water"
    - "two pounds of bricks"
    - "a pound of feathers"
    - "three pounds of air"

- **Component/Subcomponent:** Weight Comparison
  - **Supporting Quote(s):**
    - "Which weighs more"

- **Deduction:** The problem requires comparing the given weights to identify the heaviest.
  - **Based on:** All weight values are explicitly provided.

## g. **Graph of Thought**

- **Key Concepts:**
  - Weights of different objects.

- **Establish Relationships:**
  - Comparative weights: 1 pound < 2 pounds < 3 pounds.

- **Connections:**
  - Each object's weight is a separate node connected by the "weighs more than" relationship based on their pound measurements.

- **Insights:**
  - Direct numerical comparison leads to identifying the heaviest object.

## h. **Spatial and Object Analysis**

- **Identify Objects:**
  - Water
  - Bricks
  - Feathers
  - Air

- **Spatial Relationships:**
  - Not explicitly stated; spatial arrangement is irrelevant to weight comparison.

- **Movement and Dynamics:**
  - Not applicable.

- **Constraints:**
  - Only the weights provided are to be considered.

- **Visualization:**
  - Four separate quantities, each with its specified weight.

- **Object Properties:**
  - Different materials, but only weight is relevant as per the problem.

- **Spatial Implications:**
  - None; the problem focuses solely on weight values.

## i. **Bayesian Updates**

- **Beliefs Based on Information:**
  - The weight values are accurate and directly comparable.

- **No Implied Information:**
  - All necessary information is explicitly provided.

## j. **Ethical Check and Decision-Making Under Uncertainty**

- **Assess Known Information:**
  - Weights of all objects are explicitly stated.

- **Identify Risks:**
  - Low; the problem is a straightforward weight comparison.

- **Probability Consideration:**
  - Not applicable due to the certainty of the provided weights.

- **Risk Aversion Principle:**
  - Not relevant in this context.

- **Ethical and Safety Considerations:**
  - None; purely a mathematical comparison.

- **Decision Recommendation:**
  - Proceed with the comparison based on the given weights.

## k. **Multiple Solutions**

1. **Solution 1: Direct Numerical Comparison**
   - **Support:** "a pound of water," "two pounds of bricks," "a pound of feathers," "three pounds of air."
   - **Explanation:** Compare the numerical values to identify the largest weight.

2. **Solution 2: Categorize and Rank**
   - **Support:** Explicit weight values.
   - **Explanation:** Rank the objects from heaviest to lightest based on the provided pounds.

3. **Solution 3: Identify the Maximum Weight**
   - **Support:** Given weights.
   - **Explanation:** Determine which weight is the maximum among the listed values.

*All solutions rely solely on the explicit weight measurements provided.*

## l. **Quickest and Easiest Solution**

The simplest approach is to compare the numerical weights directly:

- 1 pound (water) < 2 pounds (bricks) < 3 pounds (air)

**Conclusion:** "Three pounds of air" weighs the most among the listed options.

## m. **Reflection**

- **Contradictions:** None identified; all information is consistent.
- **Logical Soundness:** The comparison is based purely on provided numerical values.
- **Assumptions:** None were made beyond the explicit weights.
- **Complete Answer:** Yes, the analysis directly answers the question.
- **Final Check:** The answer aligns with the given information without inferring additional details.

## **Recommendation:**

Based on the explicit information provided in the problem statement, "three pounds of air" weighs more than "a pound of water," "two pounds of bricks," and "a pound of feathers." This conclusion is reached by directly comparing the given weight values without making any additional assumptions or inferences.
\end{lstlisting}
\end{quote}

\noindent\hrulefill 

\noindent\textbf{D.9 Zero-Shot OpenAI GPT-4o Response - Score: 100\%}
\begin{quote}
\begin{lstlisting}[basicstyle=\ttfamily\small, breaklines=true]
The weights of the items are as follows:

- A pound of water weighs 1 pound.
- Two pounds of bricks weigh 2 pounds.
- A pound of feathers weighs 1 pound.
- Three pounds of air weigh 3 pounds.

Therefore, **three pounds of air** weighs the most.
\end{lstlisting}
\end{quote}
\noindent\hrulefill

\noindent\textbf{D.9 REAP OpenAI GPT-4o Response - Score: 100\%}
\begin{quote}
\begin{lstlisting}[basicstyle=\ttfamily\small, breaklines=true]
# REAP Results:

### a. **Comprehensive Feature List** (with explicit quotes)

1. **"a pound of water"** - Refers to a quantity of water weighing exactly one pound.
2. **"two pounds of bricks"** - Refers to a quantity of bricks weighing exactly two pounds.
3. **"a pound of feathers"** - Refers to a quantity of feathers weighing exactly one pound.
4. **"three pounds of air"** - Refers to a quantity of air weighing exactly three pounds.

### b. **Sequential and Mechanical Process Check**

There is no specific sequence or mechanical process described in the problem. The task is to compare the weights of different substances.

### c. **Key Insight Check Results**

The key insight is that the problem asks to compare the weights of different items, each explicitly stated with its respective weight in pounds. This implies that the item with the highest numerical weight value in pounds will be the heaviest.

### d. **Rephrased Question:**

Which of the following has the greatest weight: a pound of water, two pounds of bricks, a pound of feathers, or three pounds of air?

### e. **Known and Deduced Information**

1. **"a pound of water"** - Explicitly weighs one pound.
2. **"two pounds of bricks"** - Explicitly weighs two pounds.
3. **"a pound of feathers"** - Explicitly weighs one pound.
4. **"three pounds of air"** - Explicitly weighs three pounds.

### f. **Problem Decomposition**

- **Component: Weight of Water**
  - **Supporting Quote:** "a pound of water" 
  - **Weight:** 1 pound
- **Component: Weight of Bricks**
  - **Supporting Quote:** "two pounds of bricks"
  - **Weight:** 2 pounds
- **Component: Weight of Feathers**
  - **Supporting Quote:** "a pound of feathers"
  - **Weight:** 1 pound
- **Component: Weight of Air**
  - **Supporting Quote:** "three pounds of air"
  - **Weight:** 3 pounds

### g. **Graph of Thought**

- **Nodes:** Water (1 pound), Bricks (2 pounds), Feathers (1 pound), Air (3 pounds)
- **Edges:** Connect each node based on weight comparison.
- **Insight:** The heaviest node (air at 3 pounds) is connected to all others with a heavier weight.

### h. **Spatial and Object Analysis**

There is no spatial relationship or movement involved. The problem is purely about comparing the weights of stationary objects/substances.

### i. **Bayesian Updates**

No new information has been introduced that would alter the initial interpretation.

### j. **Ethical Check and Decision-Making Under Uncertainty**

There are no ethical concerns or severe risks associated with this problem. The decision-making is straightforward based on the provided weights.

### k. **Multiple Solutions**

1. **Solution 1:** Water - 1 pound
2. **Solution 2:** Bricks - 2 pounds
3. **Solution 3:** Feathers - 1 pound
4. **Solution 4:** Air - 3 pounds

- **Analysis:** Based on the provided information, the heaviest option is clearly the one with the largest weight in pounds.

### l. **Quickest and Easiest Solution**

The item that weighs the most is "three pounds of air" because 3 pounds is greater than 2 pounds (bricks), 1 pound (water), and 1 pound (feathers).

### m. **Reflection**

The solution was derived directly from the explicit information provided. There was no need for assumptions or complex deductions. The answer directly corresponds to the numerical weights given in the problem.

### **Recommendation:**

Based on the explicit weights provided in the problem, the "three pounds of air" weighs the most. This conclusion is drawn directly from the problem's statements and does not require any additional interpretation or assumptions.
\end{lstlisting}
\end{quote}
\noindent\hrulefill

\noindent\textbf{D.9 Zero-Shot OpenAI GPT-4o-mini Response - Score: 100\%}
\begin{quote}
\begin{lstlisting}[basicstyle=\ttfamily\small, breaklines=true]
To determine which weighs more, we need to compare the weights directly:

- A pound of water: 1 pound
- Two pounds of bricks: 2 pounds
- A pound of feathers: 1 pound
- Three pounds of air: 3 pounds

So, three pounds of air weighs the most.
\end{lstlisting}
\end{quote}
\noindent\hrulefill

\noindent\textbf{D.9 REAP OpenAI GPT-4o-mini Response - Score: 100\%}
\begin{quote}
\begin{lstlisting}[basicstyle=\ttfamily\small, breaklines=true]
# REAP Results:

## a. Comprehensive Feature List

1. **"a pound of water"** - This is a unit of weight, explicitly stated as one pound.
2. **"two pounds of bricks"** - This is a unit of weight, explicitly stated as two pounds.
3. **"a pound of feathers"** - This is a unit of weight, explicitly stated as one pound.
4. **"three pounds of air"** - This is a unit of weight, explicitly stated as three pounds.

## b. Sequential and Mechanical Process Check

- There are no explicit sequences or mechanical processes described in the problem.

## c. Key Insight Check Results

**Key Insight:** The weights of the objects are provided explicitly as:
- Water: 1 pound
- Bricks: 2 pounds
- Feathers: 1 pound
- Air: 3 pounds

**Conclusion:** The weight comparison is straightforward as all quantities are provided in pounds. The object with the highest weight, based on these explicit figures, is the one with the greatest number of pounds.

## d. Rephrased Question:

Which of the following has the greatest weight: a pound of water, two pounds of bricks, a pound of feathers, or three pounds of air?

## e. Known and Deduced Information

1. **"a pound of water"** - Weighs 1 pound.
2. **"two pounds of bricks"** - Weighs 2 pounds.
3. **"a pound of feathers"** - Weighs 1 pound.
4. **"three pounds of air"** - Weighs 3 pounds.

**Deduction:** Among the given options, "three pounds of air" weighs the most, as it has the highest weight of 3 pounds.

## f. Problem Decomposition

**Component/Subcomponent:** Comparison of weights
- **Supporting Quote(s):** "a pound of water," "two pounds of bricks," "a pound of feathers," "three pounds of air."
- **Deduction (if applicable):** Weights are given explicitly in pounds, and the highest value is the greatest weight.
- **Based on:** Quotes above.

## g. Graph of Thought

**Key Concepts:**
- Water: 1 pound
- Bricks: 2 pounds
- Feathers: 1 pound
- Air: 3 pounds

**Relationships:**
- Air > Bricks > Water and Feathers

**Visual/Textual:** 
- Nodes: Water (1 lb), Bricks (2 lbs), Feathers (1 lb), Air (3 lbs)
- Edges: Air > Bricks, Air > Water, Air > Feathers, Bricks > Water, Bricks > Feathers, Water = Feathers

## h. Spatial and Object Analysis

- **Objects and Weights:**
  - Water: 1 pound
  - Bricks: 2 pounds
  - Feathers: 1 pound
  - Air: 3 pounds
- **Spatial Implications:** No spatial information provided.

## i. Bayesian Updates

- No new evidence beyond the explicit information provided alters the conclusions.

## j. Ethical Check and Decision-Making Under Uncertainty

- No significant risks or ethical concerns arise from the problem.

## k. Multiple Solutions

Given the explicit weights:
1. **Three pounds of air** is the heaviest.
2. **Two pounds of bricks** is the next heaviest.
3. **A pound of water** and **a pound of feathers** are both equal in weight.

**Recommendation:** The object with the greatest weight is three pounds of air.

## l. Quickest and Easiest Solution

**Answer:** Three pounds of air weighs the most, based on the explicit weights provided.

## m. Reflection

All conclusions are based strictly on the explicit weights given in the problem. There are no assumptions or inferred details beyond those explicitly provided.

**Final Recommendation:**

Based on the explicitly stated weights, **three pounds of air** is the heaviest among the options provided. This answer is derived directly from the given information without assumptions.
\end{lstlisting}
\end{quote}
\noindent\hrulefill

\noindent\textbf{D.9 Zero-Shot Google Gemini 1.5 Pro Response - Score: 0\%}
\begin{quote}
\begin{lstlisting}[basicstyle=\ttfamily\small, breaklines=true]
Two pounds of bricks weighs the most. Here's why:

* **A pound of water:** Weighs one pound.
* **Two pounds of bricks:** Weighs two pounds.
* **A pound of feathers:** Weighs one pound.
* **Three pounds of air:** Weighs three pounds. 
\end{lstlisting}
\end{quote}
\noindent\hrulefill

\noindent\textbf{D.9 REAP Google Gemini 1.5 Pro Response - Score: 100\%}
\begin{quote}
\begin{lstlisting}[basicstyle=\ttfamily\small, breaklines=true]
# REAP Results:

### a. Comprehensive Feature List

- "Which weighs more" - This indicates a comparison of weights.
- "a pound of water" - This describes a quantity of water with a specific weight.
- "two pounds of bricks" - This describes a quantity of bricks with a specific weight.
- "a pound of feathers" - This describes a quantity of feathers with a specific weight.
- "three pounds of air" - This describes a quantity of air with a specific weight.

### b. Sequential and Mechanical Process Check

There are no sequential or mechanical processes described in this problem

### c. Key Insight Check Results

- The key insight is that the problem directly asks for a comparison of weights.
- The weights are explicitly stated for each item.
- The answer can be determined by simply comparing the numerical values of the weights.

### d. Rephrased Question:

Which of the following has the highest weight?

### e. Known and Deduced Information:

* Explicit Facts:
    * "a pound of water" 
    * "two pounds of bricks"
    * "a pound of feathers" 
    * "three pounds of air" 

### f. Problem Decomposition:

* Component: Comparison of weights
    * Supporting Quote(s): "Which weighs more"

* Component 1: a pound of water
    * Supporting Quote(s): "a pound of water"

* Component 2: two pounds of bricks
    * Supporting Quote(s): "two pounds of bricks"

* Component 3: a pound of feathers
    * Supporting Quote(s): "a pound of feathers"

* Component 4: three pounds of air
    * Supporting Quote(s): "three pounds of air"

### g. Graph of Thought:

* Key Concepts: water, bricks, feathers, air, weight
* Relationships: Each concept has a weight associated with it
* The graph would have 4 nodes (water, bricks, feathers, air) with edges connecting them to a central node "weight"
* Each edge would be labeled with the corresponding weight

### h. Spatial and Object Analysis

* Objects: water, bricks, feathers, air
* Spatial Relationships: Not applicable
* Movement and Dynamics: Not applicable
* Constraints: Not applicable
* Object Properties:
    * water: weight = 1 pound
    * bricks: weight = 2 pounds
    * feathers: weight = 1 pound
    * air: weight = 3 pounds

### i. Bayesian Updates: 
Not applicable

### j. Ethical Check and Decision-Making Under Uncertainty:
Not applicable

### k. Multiple Solutions: 
Not applicable

### l. Quickest and Easiest Solution

Three pounds of air weighs the most. 

### m. Reflection

The analysis directly answers the question asked in the problem based solely on the information given. No assumptions or inferences were made beyond the explicit wording of the problem. The solution is logically sound and supported by the evidence from the problem statement

**Recommendation:**

Three pounds of air weighs the most. 
\end{lstlisting}
\end{quote}
\noindent\hrulefill

\noindent\textbf{D.9 Zero-Shot Claude 3.5 Sonnet Response - Score: 100\%}
\begin{quote}
\begin{lstlisting}[basicstyle=\ttfamily\small, breaklines=true]
To answer this question, let's compare the weights:

1. A pound of water weighs 1 pound
2. Two pounds of bricks weigh 2 pounds
3. A pound of feathers weighs 1 pound
4. Three pounds of air weigh 3 pounds

The heaviest item in this list is three pounds of air, weighing 3 pounds.

It's worth noting that while we might intuitively think of air as being very light, when we're comparing specific weights like this, the actual substance doesn't matter - we're just comparing the numerical values. Three pounds of any substance will always weigh more than one or two pounds of any other substance.
\end{lstlisting}
\end{quote}
\noindent\hrulefill

\noindent\textbf{D.9 REAP Claude 3.5 Sonnet Response - Score: 100\%}
\begin{quote}
\begin{lstlisting}[basicstyle=\ttfamily\small, breaklines=true]
# REAP Results:

a. Comprehensive Feature List (with explicit quotes):
- "a pound of water"
- "two pounds of bricks"
- "a pound of feathers"
- "three pounds of air"
- "Which weighs more"

b. Sequential and Mechanical Process Check:
There is no explicit sequence or mechanical process described in the problem.

c. Key Insight Check Results:
The key insight here is that the problem is comparing weights, and each item's weight is explicitly stated in pounds. This allows for a direct comparison without any need for conversion or additional calculation.

d. Rephrased Question: Which of the four given items has the greatest weight?

e. Known and Deduced Information:
Explicit Facts:
"a pound of water" - Weight of water is given as 1 pound
"two pounds of bricks" - Weight of bricks is given as 2 pounds
"a pound of feathers" - Weight of feathers is given as 1 pound
"three pounds of air" - Weight of air is given as 3 pounds
"Which weighs more" - The question asks for a comparison of weights

Deduction: Three pounds of air weighs the most.
Logical Basis: 3 pounds is the largest numerical value among the given weights.
Based on: "a pound of water", "two pounds of bricks", "a pound of feathers", "three pounds of air"

f. Problem Decomposition:
Component: Weight comparison
Supporting Quote: "Which weighs more"
Subcomponents:
1. Water weight
   Supporting Quote: "a pound of water"
2. Brick weight
   Supporting Quote: "two pounds of bricks"
3. Feather weight
   Supporting Quote: "a pound of feathers"
4. Air weight
   Supporting Quote: "three pounds of air"

g. Graph of Thought:
Nodes:
1. Water (1 pound)
2. Bricks (2 pounds)
3. Feathers (1 pound)
4. Air (3 pounds)

Edges (representing "heavier than" relationship):
Air (3 lbs) -> Bricks (2 lbs) -> Water (1 lb) = Feathers (1 lb)

h. Spatial and Object Analysis:
The problem does not provide any explicit spatial information or relationships between the objects. It only mentions the weights of different substances.

i. Bayesian Updates:
No new information is provided beyond the initial problem statement, so no Bayesian updates are necessary.

j. Ethical Check and Decision-Making Under Uncertainty:
This problem does not involve any ethical considerations or severe risks. It's a straightforward comparison of weights.

k. Multiple Solutions:
There is only one solution based on the explicitly stated information:
Three pounds of air weighs the most, as it has the highest numerical value (3 pounds) among all the given weights.

l. Quickest and Easiest Solution:
The quickest and easiest solution is to identify the largest numerical value among the given weights. Three pounds of air weighs the most.

m. Reflection:
The analysis directly answers the question asked in the problem based solely on the given information. No assumptions or inferences were made beyond the explicit wording of the problem. The conclusion is logically sound and supported by the evidence from the problem statement.

Recommendation:
Based on the explicitly stated information in the problem, the recommendation is that three pounds of air weighs the most among the given options. This directly answers the question "Which weighs more" by comparing the numerical values of the weights provided for each item. The problem provides exact weights for each item, allowing for a simple and unambiguous comparison.
\end{lstlisting}
\end{quote}
\noindent\hrulefill

\label{subsec:question10}
\subsection{\textbf{\large Question 10: I get out on the top floor (third floor) at street level. How many stories is the building above the ground?}}
\label{sec:question10}

\noindent\hrulefill 

\noindent\textbf{Human Response - Score 100\%:}
\begin{quote}
One story above the ground.
\end{quote}

\noindent\hrulefill 

\noindent\textbf{D.10 Zero-Shot OpenAI o1-preview Response - Score: 0\%}
\begin{quote}
\begin{lstlisting}[basicstyle=\ttfamily\small, breaklines=true]
Based on your description, you exit on the top floor (which is the third floor), and that floor is at street level. This implies that all three floors of the building are below the ground level except for the top floor, which is at ground level.

Therefore, the building doesn't have any stories above ground level. All its floors are either at or below street level.

**Answer:** Zero stories-the building has no stories above ground level.
\end{lstlisting}
\end{quote}

\noindent\hrulefill 

\noindent\textbf{D.10 REAP OpenAI o1-preview Response - Score: 0\%}
\begin{quote}
\begin{lstlisting}[basicstyle=\ttfamily\small, breaklines=true]
# "REAP Results:"

**a. Comprehensive Feature List**

- Objects:
  - "top floor (third floor)"
  - "street level"
  - "the building"
  - "ground"

- Actors:
  - "I"

- Actions:
  - "get out on the top floor (third floor) at street level"

- Directional, spatial, relational, and sequential phrases:
  - "on the top floor (third floor)"
  - "at street level"
  - "above the ground"

- Conditions or constraints:
  - *None explicitly stated beyond the above phrases.*

- Numerical values:
  - "third floor"

- Temporal information:
  - "get out"

- Spatial information:
  - "on the top floor (third floor)"
  - "at street level"

- Stated goals or objectives:
  - "How many stories is the building above the ground?"

**b. Sequential and Mechanical Process Check**

- **Sequence of Action:**
  - "I get out on the top floor (third floor) at street level."
- **Analysis:**
  - The action describes a person exiting the building on the top (third) floor, which is at street level.
  - There is no sequence of events; the focus is on the spatial relationship between the floors and ground level.

**c. Key Insight Check Results**

- **Key Insight:**
  - The "top floor (third floor)" is "at street level."
- **Conclusion:**
  - Since the top floor is at street level and there are no floors above it, the building has **zero stories above the ground.**

**d. Rephrased Question**

- "What is the number of stories of the building that are above ground level?"

**e. Known and Deduced Information**

- **Explicit Facts:**
  - "I get out on the top floor (third floor) at street level."
  - "How many stories is the building above the ground?"
- **Deductions:**
  - *Deduction:* The third floor is at street level.
    - **Logical Basis:** From "I get out on the top floor (third floor) at street level."
  - *Deduction:* There are no floors above the third floor (since it's the top floor).
  - *Deduction:* Floors one and two are below street level.
    - **Logical Basis:** If the third floor is at street level and there are no floors above, the lower numbered floors must be below street level.
  - *Deduction:* The building has zero stories above the ground.
    - **Logical Basis:** With the top floor at ground level and no floors above it, there are no stories above ground level.

**f. Problem Decomposition**

- **Component:** The top floor is at street level.
  - **Supporting Quote:** "I get out on the top floor (third floor) at street level."
- **Component:** The top floor is the third floor.
  - **Supporting Quote:** "top floor (third floor)"
- **Component:** Determining stories above ground.
  - **Supporting Quote:** "How many stories is the building above the ground?"
- **Interaction:**
  - The building's highest floor is at street level, so there are no floors above ground level.

**g. Graph of Thought**

- **Nodes:**
  - *Node 1:* Third floor (top floor)
  - *Node 2:* Street level
  - *Node 3:* Number of stories above ground
- **Edges:**
  - *Edge from Node 1 to Node 2:* "Third floor is at street level"
  - *Edge from Node 1 to Node 3:* "No floors above the top floor"
  - *Edge from Node 2 to Node 3:* "Street level is ground level"
- **Insights:**
  - Since the top floor is at ground level, and there's nothing above it, the building has zero stories above the ground.

**h. Spatial and Object Analysis**

- **Objects:**
  - The building with floors one, two, and three.
- **Spatial Relationships:**
  - "Third floor" is at street level.
  - "First floor" and "second floor" are below street level.
- **Visualization:**
  - A building where the entrance at street level leads directly to the third (top) floor.
  - Two additional floors (first and second) are located underground.
- **Spatial Implications:**
  - There are no floors above street level; all other floors are below ground.

**i. Bayesian Updates**

- **Prior Belief:** Uncertain about the number of stories above ground.
- **Evidence:** "I get out on the top floor (third floor) at street level."
- **Updated Belief:** With the top (third) floor at street level, the probability that there are stories above ground is effectively zero.
- **Conclusion:** Confident that there are zero stories above ground, based on explicit information.

**j. Ethical Check and Decision-Making Under Uncertainty**

- **Assessment:**
  - No ethical concerns or significant risks associated with the problem.
  - Decision-making can proceed confidently using the provided information.
  
**k. Multiple Solutions**

- **Solution 1:** The building has **zero stories above the ground.**
  - **Support:** The top floor is at street level with no floors above.
- **Solution 2:** The building has **one story above the ground (ambiguous interpretation).**
  - **Support:** Considering the top floor as being at ground level might suggest it's above the ground.
- **Solution 3:** **Insufficient information provided.**
  - **Support:** If one assumes additional context not given, but this contradicts the strict interpretation rule.

**l. Quickest and Easiest Solution**

- **Answer:** The building has **zero stories above the ground.**
- **Reasoning:**
  - "I get out on the top floor (third floor) at street level."
  - Since the top floor is at street level and there are no floors above, there are zero stories above ground.

**m. Reflection**

- **Consistency Check:**
  - The analysis strictly adheres to the information given.
  - No assumptions or inferences beyond the explicit wording have been made.
- **Conclusion:**
  - The determination that there are zero stories above ground is logically sound and directly answers the question.
- **No Contradictions Found.**

---

**Recommendation:**

Based on the explicit information provided:

- The building has **zero stories above the ground**.

**Explanation:**

- The top floor, which is the third floor, is at street level.
- Since there are no floors above the top floor, there are no stories above ground level.
- Therefore, the building does not have any stories above the ground.

**Final Answer:** **Zero stories**
\end{lstlisting}
\end{quote}

\noindent\hrulefill 

\noindent\textbf{D.10 Zero-Shot OpenAI o1-mini Response - Score: 0\%}
\begin{quote}
\begin{lstlisting}[basicstyle=\ttfamily\small, breaklines=true]
If the **third floor** is the **top floor** of the building and you're exiting at street level, it's likely that the building has a total of **three stories** above the ground. Here's a general breakdown:

1. **First Floor**: Ground or street level.
2. **Second Floor**: One level above ground.
3. **Third Floor**: Two levels above ground and the top floor.

So, the building is **three stories** tall above the ground.

**Note:** Floor numbering can vary by country or building. In some places, the ground level is referred to as the "ground floor," with the next one being the "first floor." If that's the case, the third floor would actually be two stories above ground. However, based on your description, **three stories** is the most straightforward interpretation.
\end{lstlisting}
\end{quote}

\noindent\hrulefill 

\noindent\textbf{D.10 REAP OpenAI o1-mini Response - Score: 0\%}
\begin{quote}
\begin{lstlisting}[basicstyle=\ttfamily\small, breaklines=true]
# REAP Results: :

**a. Comprehensive Feature List**

- **Objects Mentioned:**
  - "top floor"
  - "third floor"
  - "street level"
  - "building"
  - "ground"

- **Actors or Entities Involved:**
  - "I" (implied as the person speaking)

- **Actions Described:**
  - "I get out on the top floor"
  - "at street level"

- **Directional, Spatial, Relational, and Sequential Phrases:**
  - "top floor"
  - "third floor"
  - "street level"
  - "above the ground"

- **Stated Conditions or Constraints:**
  - "at street level"

- **Numerical Values or Quantities Mentioned:**
  - "third floor"

- **Temporal Information:**
  - None explicitly stated.

- **Spatial Information:**
  - "street level"

- **Causal Relationships:**
  - None explicitly stated.

- **Stated Goals or Objectives:**
  - "How many stories is the building above the ground?"

**b. Sequential and Mechanical Process Check**

- **Sequence of Events or Actions:**
  1. "I get out on the top floor"
  2. "at street level"

- **Influence on Outcome:**
  - Establishes the relationship between the top floor and the street level.

- **Locked-in Outcome:**
  - The top floor is the third floor at street level.

**c. Key Insight Check Results**

- **Key Insight Identified:**
  - The top floor is specified as the "third floor" at "street level."

- **Explanation:**
  - If the top floor is the third floor and it's at street level, it implies that the building has three stories above the ground.

**d. Rephrased Question:**

How many stories does the building have above the ground?

**e. Known and Deduced Information**

- **Explicit Facts:**
  - "I get out on the top floor"
  - "top floor (third floor)"
  - "at street level"

- **Feature-Based Deduction:**
  - Deduction: The building has three stories above the ground.
    - Logical Basis: If the top floor is the third floor and it's at street level, it indicates three stories.

**f. Problem Decomposition**

- **Component/Subcomponent:** Identification of Floor Levels
  - **Supporting Quote(s):** "top floor (third floor)", "at street level"
  - **Deduction:** The third floor is the top floor at street level.
  - **Based on:** "top floor (third floor)", "at street level"

- **Component/Subcomponent:** Determination of Total Stories
  - **Supporting Quote(s):** "third floor", "above the ground"
  - **Deduction:** Three stories are above the ground.
  - **Based on:** "third floor", "above the ground"

**g. Graph of Thought**

- **Nodes:**
  - Top Floor
  - Third Floor
  - Street Level
  - Ground
  - Number of Stories

- **Connections:**
  - Top Floor is the Third Floor.
  - Third Floor is at Street Level.
  - Top Floor is above the Ground.
  - Number of Stories is determined by the top floor level.

- **Insights:**
  - Establishing that the third floor is the top floor at street level directly informs the total number of stories.

**h. Spatial and Object Analysis**

- **Objects Identified:**
  - Top floor (third floor)
  - Street level
  - Building
  - Ground

- **Spatial Relationships:**
  - The top floor (third floor) is at street level.
  - The building is above the ground.

- **Movement and Dynamics:**
  - "I get out" implies exiting the building at the top floor.

- **Constraints:**
  - The top floor is specifically the third floor at street level.

- **Object Properties:**
  - Floors are numbered sequentially from ground level upwards.

**i. Bayesian Updates**

- **Initial Belief:** Unclear number of stories.
- **Updated Belief:** Three stories above the ground.
- **Basis for Update:** Explicit information stating the top floor is the third floor at street level.

**j. Ethical Check and Decision-Making Under Uncertainty**

- **Assess Known Information:** Top floor is the third floor at street level.
- **Identify Risks:** None in this context.
- **Probability Consideration:** High certainty based on explicit information.
- **Risk Aversion Principle:** Not applicable.
- **Ethical and Safety Considerations:** Not applicable.
- **Decision Recommendation:** Proceed with the conclusion based on clear information.

**k. Multiple Solutions**

1. **Solution 1:** The building has three stories above the ground.
   - **Supporting Information:** "top floor (third floor)", "at street level"
2. **Solution 2:** If considering street level as the ground floor, then there are two stories above ground.
   - **Supporting Information:** "third floor", "at street level"
   - **Note:** This requires interpreting street level as the first floor, making the third floor the second story above ground.
3. **Solution 3:** The building has three stories above the ground, including the ground floor as an additional level.
   - **Supporting Information:** "top floor (third floor)", "at street level"

**l. Quickest and Easiest Solution**

Based on the explicit information: "top floor (third floor)" and "at street level," the building has three stories above the ground.

**m. Reflection**

- **Contradictions:** None identified.
- **Logical Soundness:** Conclusion aligns with explicit statements.
- **Assumptions:** Minimal; interpretations explored different perspectives in multiple solutions.
- **Direct Answer:** Yes, the analysis directly answers the question using the given information.

**Recommendation:**

The building has **three stories above the ground**. This conclusion is based solely on the explicit information that the top floor is the third floor at street level. No additional assumptions were necessary, and the interpretation aligns directly with the provided statements.
\end{lstlisting}
\end{quote}

\noindent\hrulefill 

\noindent\textbf{D.10 Zero-Shot OpenAI GPT-4o Response - Score: 20\%}
\begin{quote}
\begin{lstlisting}[basicstyle=\ttfamily\small, breaklines=true]
If you get out on the top floor (third floor) at street level, this suggests that the building has no stories above the ground level. 

Here's how it works:
- The "third floor" is the highest floor, and it's at street level, so the building likely has three stories, with the first two floors below ground level.

So, the building has **no stories** above ground level and **three stories** in total, with two of them below ground level.
\end{lstlisting}
\end{quote}
\noindent\hrulefill

\noindent\textbf{D.10 REAP OpenAI GPT-4o Response - Score: 20\%}
\begin{quote}
\begin{lstlisting}[basicstyle=\ttfamily\small, breaklines=true]
# REAP Results:

### a. **Comprehensive Feature List**:
   - **Statement**: "I get out on the top floor (third floor) at street level."
   - **Feature**: "Top floor" implies there are no floors above this one.
   - **Feature**: "Third floor" indicates that this is the third level of the building.
   - **Feature**: "Street level" suggests that this floor is at the same level as the street.

### b. **Sequential and Mechanical Process Check**:
   - **Sequence**: The action described is getting out at the top floor, which is the third floor and at street level.
   - **Mechanics**: This implies that the building's topmost floor is at the same level as the street.

### c. **Key Insight Check Results**:
   - **Insight**: The key detail is that the third floor is at street level, meaning the building does not extend above street level.
   - **Conclusion**: The number of stories above ground level is **zero** because the top floor (third floor) is already at street level.

### d. **Rephrased Question**:
   - How many stories are above the street level?

### e. **Known and Deduced Information**:
   - **Explicit Quote**: "Top floor (third floor) at street level."
   - **Deduction**: Since the top floor is the third floor and it is at street level, there cannot be any floors above street level.

### f. **Problem Decomposition**:
   - **Component**: Top floor is at street level.
   - **Supporting Quote**: "Top floor (third floor) at street level."
   - **Conclusion**: The building does not have any stories above street level.

### g. **Graph of Thought**:
   - **Node 1**: Third floor
   - **Node 2**: Street level
   - **Edge**: "Third floor is at street level" -> No stories above.

### h. **Spatial and Object Analysis**:
   - **Object**: The building, which has three floors.
   - **Spatial Relationship**: The third (top) floor is at street level.
   - **Implication**: There are no floors above street level.

### i. **Bayesian Updates**:
   - No new information to update; the interpretation is consistent with the literal description.

### j. **Multiple Solutions**:
   - **Solution 1**: Interpret the problem literally. The building has zero stories above ground because the top floor (third floor) is at street level.
   - **Solution 2**: Consider the possibility of the building being built into a hillside. This would still result in zero stories above the ground based on the given description.

### k. **Quickest and Easiest Solution**:
   - The answer is **zero** because the problem explicitly states that the top floor is at street level.

### l. **Ethical Check**:
   - No ethical issues are involved in this problem.

### m. **Reflection**:
   - The problem statement is clear and directly leads to the conclusion that there are no stories above ground. The solution is straightforward and supported by the explicit information given.

# **Recommendation:**
The building has **zero stories above the ground** since the third floor, which is the top floor, is at street level. The problem is interpreted literally, with no assumptions made beyond the explicit information provided.
\end{lstlisting}
\end{quote}
\noindent\hrulefill

\noindent\textbf{D.10 Zero-Shot OpenAI GPT-4o-mini Response - Score: 0\%}
\begin{quote}
\begin{lstlisting}[basicstyle=\ttfamily\small, breaklines=true]
If you exit on the top floor (the third floor) at street level, then the building has three stories above the ground. The ground level itself is not counted as a story above the ground. So, the building has three stories above the ground level.
\end{lstlisting}
\end{quote}
\noindent\hrulefill

\noindent\textbf{D.10 REAP OpenAI GPT-4o-mini Response - Score: 0\%}
\begin{quote}
\begin{lstlisting}[basicstyle=\ttfamily\small, breaklines=true]
# REAP Results:

### a. **Comprehensive Feature List**
1. "I get out on the top floor (third floor) at street level."
   - Feature: The top floor is the third floor.
   - Potential implication: The building has at least three floors, as the top floor is specified to be the third floor.
2. "How many stories is the building above the ground?"
   - Feature: The question seeks to determine the number of stories in the building above ground level.

### b. **Sequential and Mechanical Process Check**
- There are no specific sequential or mechanical processes described in the problem.

### c. **Key Insight Check Results**
- The key insight here is that the top floor of the building is explicitly mentioned as the third floor. Therefore, the building has at least three stories above ground level.

### d. **Rephrased Question**
- How many stories does the building have above ground?

### e. **Known and Deduced Information**
- Explicit Facts:
  - "I get out on the top floor (third floor) at street level." - The top floor is the third floor.
- Deductions:
  - Deduction: The building has at least three stories above ground.
    - Logical Basis: The problem explicitly states that the top floor is the third floor, meaning there are three stories above ground.

### f. **Problem Decomposition**
- Component/Subcomponent: Number of Stories Above Ground
  - Supporting Quote(s): "I get out on the top floor (third floor) at street level."
  - Deduction: The top floor being the third floor implies there are three stories above ground.
  - Based on: "I get out on the top floor (third floor) at street level."

### g. **Graph of Thought**
- **Nodes:**
  - Top floor = Third floor
- **Edges:**
  - Top floor being the third floor implies three stories above ground.
- **Connections:**
  - The top floor being the third floor directly indicates that the building has three stories above the ground.

### h. **Spatial and Object Analysis**
- **Objects:** The building
- **Spatial Relationships:** The third floor is the top floor and is at street level.
- **Movement and Dynamics:** There are no dynamic elements or movement described.

### i. **Bayesian Updates**
- There are no new pieces of information or updates beyond the explicitly stated facts.

### j. **Ethical Check and Decision-Making Under Uncertainty**
- The problem does not involve significant risks or ethical considerations.

### k. **Multiple Solutions**
- Only one solution is supported by the explicit information: The building has three stories above the ground.

### l. **Quickest and Easiest Solution**
- The simplest and safest answer based on the explicit information provided is that the building has three stories above ground.

### m. **Reflection**
- The analysis is consistent with the explicitly stated information. There are no assumptions made beyond what is provided. The problem clearly indicates that the third floor is the top floor, so the building must have three stories above ground.

**Recommendation:**
The building has **three stories above ground**, based solely on the explicit statement that the top floor is the third floor.
\end{lstlisting}
\end{quote}
\noindent\hrulefill

\noindent\textbf{D.10 Zero-Shot Google Gemini 1.5 Pro Response - Score: 0\%}
\begin{quote}
\begin{lstlisting}[basicstyle=\ttfamily\small, breaklines=true]
If you get out on the top floor (third floor) at street level, it means the building has **three** stories above the ground.
\end{lstlisting}
\end{quote}
\noindent\hrulefill

\noindent\textbf{D.10 REAP Google Gemini 1.5 Pro Response - Score: 0\%}
\begin{quote}
\begin{lstlisting}[basicstyle=\ttfamily\small, breaklines=true]
# REAP Results:

a. **Comprehensive Feature List** (with explicit quotes)

* "I get out on the top floor (third floor)" 
* "at street level"
* "How many stories is the building above the ground"

b. **Sequential and Mechanical Process Check**

* There is no explicit sequence of events or mechanical process described in the problem

c. **Key Insight Check Results**

* The key insight is that the top floor (third floor) is at street level

d. **Rephrased Question:**

* How many stories does the building have above ground?

e. **Known and Deduced Information**

* Explicit Facts
    * "I get out on the top floor (third floor)" - This tells us that the highest floor of the building is the third floor
    * "at street level" - This indicates that the third floor is at the same level as the ground outside
    * "How many stories is the building above the ground" - This is the question we need to answer

* Feature-Based Deduction
    * Deduction: The building has 3 stories above ground
    * Logical Basis: The top floor is the third floor, and it's at street level, so there are three floors above ground
    * Based on: "I get out on the top floor (third floor)", "at street level"

f. **Problem Decomposition**

* Component: The building
    * Supporting Quote(s): "How many stories is the building above the ground"
* Component: The top floor
    * Supporting Quote(s): "I get out on the top floor (third floor)"
* Component: Street level
    * Supporting Quote(s): "at street level"
* Interaction: The top floor is the third floor and is at street level
    * Supporting Quote(s): "I get out on the top floor (third floor)", "at street level"

g. **Graph of Thought**

* Key Concepts: Building, top floor, third floor, street level
* Relationships:
    * The building has a top floor
    * The top floor is the third floor
    * The third floor is at street level
* Insights:
    * Since the third floor is at street level and it's the top floor, the building has 3 stories above ground

h. **Spatial and Object Analysis**

* Objects: Building
* Spatial Relationships:
    * The top floor of the building is at the same level as the street
* Spatial Implications:
    * The building cannot have more than 3 stories above ground because the top floor is the third floor and it's at ground level

i. **Bayesian Updates**

* No Bayesian updates are necessary as all the information is explicitly stated

j. **Ethical Check and Decision-Making Under Uncertainty**

* There are no ethical concerns or uncertainties in this problem

k. **Multiple Solutions**

* There is only one solution to this problem based on the given information

l. **Quickest and Easiest Solution**

* The building has 3 stories above the ground
* Reasoning: The problem states that the top floor is the third floor and it's at street level. This directly tells us that there are 3 stories above ground

m. **Reflection**

* The analysis directly answers the question asked
* All conclusions are logically sound and supported by the explicit problem statement
* No assumptions or inferences were made beyond the given information

**Recommendation:**

* The building has 3 stories above the ground
\end{lstlisting}
\end{quote}
\noindent\hrulefill

\noindent\textbf{D.10 Zero-Shot Claude 3.5 Sonnet Response - Score: 0\%}
\begin{quote}
\begin{lstlisting}[basicstyle=\ttfamily\small, breaklines=true]
Based on the information you've provided, the building appears to have 3 stories above ground level.

You mentioned that:
1. You get out on the top floor
2. This top floor is the third floor
3. This floor is at street level

Since the third floor is at street level and is also the top floor, we can conclude that the building has 3 stories above ground.

Would you like any clarification or have additional details about the building's structure?
\end{lstlisting}
\end{quote}
\noindent\hrulefill

\noindent\textbf{D.10 REAP Claude 3.5 Sonnet Response - Score: 20\%}
\begin{quote}
\begin{lstlisting}[basicstyle=\ttfamily\small, breaklines=true]
# REAP Results:

a. **Comprehensive Feature List** (with explicit quotes):
- "I get out on the top floor"
- "top floor (third floor)"
- "at street level"
- "How many stories is the building above the ground?"

b. **Sequential and Mechanical Process Check**:
No explicit sequential or mechanical processes are described in the problem statement.

c. **Key Insight Check Results**:
The key insight is that the top floor (third floor) is described as being at street level. This implies that the building does not extend above ground level.

d. **Rephrased Question:** How many stories of the building are above ground level?

e. **Known and Deduced Information**:
Explicit Facts:
- "I get out on the top floor" - Indicates the highest accessible floor
- "top floor (third floor)" - Defines the top floor as the third floor
- "at street level" - Specifies the location of the top floor

Feature-Based Deduction:
Deduction: The building has no stories above ground level.
Logical Basis: If the top floor (third floor) is at street level, there can be no floors above ground level.
Based on: "top floor (third floor)", "at street level"

f. **Problem Decomposition**:
Component: Building structure
Supporting Quote: "top floor (third floor) at street level"
Deduction: The building extends below ground but not above it.
Based on: "top floor (third floor)", "at street level"

g. **Graph of Thought**:
Nodes:
1. Top floor
2. Third floor
3. Street level

Edges:
- Top floor = Third floor
- Third floor = Street level

This simple graph shows that the top floor, which is the third floor, is at street level, indicating no floors above ground.

h. **Spatial and Object Analysis**:
The building's top floor (third floor) is described as being at street level. This implies that the first and second floors are below ground level, and there are no floors above street level.

i. **Bayesian Updates**:
No new information is provided that would require updating our initial understanding based on the problem statement.

j. **Ethical Check and Decision-Making Under Uncertainty**:
There are no ethical concerns or severe risks associated with this problem. The information provided is sufficient to answer the question without any significant uncertainties.

k. **Multiple Solutions**:
Given the explicit information in the problem statement, there is only one logical solution:
Solution: The building has 0 stories above ground.
Supporting information: "top floor (third floor) at street level"

l. **Quickest and Easiest Solution**:
The simplest and safest answer based solely on the given information is that the building has 0 stories above the ground. This is directly supported by the statement that the top floor (third floor) is at street level.

m. **Reflection**:
The analysis is consistent with the explicitly stated information in the problem. No assumptions or inferences beyond the given data were made. The conclusion directly answers the question asked, based solely on the information provided. The problem provides sufficient information to draw a clear conclusion without the need for additional assumptions.

**Recommendation:**
Based on the explicit information provided in the problem statement, the recommendation is that the building has 0 stories above the ground. This conclusion is drawn from the fact that the top floor, which is the third floor, is described as being at street level. Therefore, all floors of the building are either at or below ground level, with no stories extending above the ground.
\end{lstlisting}
\end{quote}
\noindent\hrulefill

\label{subsec:question11}
\subsection{\textbf{\large Question 11: In a toy box, there's a red ball, a blue truck, and a green dinosaur. The red ball is not next to the blue truck, and the green dinosaur is next to the red ball. Which toy is in the middle?}}
\label{sec:question11}

\noindent\hrulefill 

\noindent\textbf{Human Response - Score 100\%:}
\begin{quote}
The green dinosaur.
\end{quote}
\noindent\hrulefill 

\noindent\textbf{D.11 Zero-Shot OpenAI o1-preview Response - Score: 100\%}
\begin{quote}
\begin{lstlisting}[basicstyle=\ttfamily\small, breaklines=true]
To solve this logic puzzle, let's consider all the possible arrangements of the three toys: the red ball (RB), the blue truck (BT), and the green dinosaur (GD). We need to satisfy two constraints:

1. The red ball (RB) is **not** next to the blue truck (BT).
2. The green dinosaur (GD) is **next** to the red ball (RB).

Since there are only three toys, we can list all possible arrangements and check which ones meet the constraints.

**Possible Arrangements:**

1. RB - BT - GD
   - **Constraint 1:** RB is next to BT (violates the constraint).
   
2. RB - GD - BT
   - **Constraint 1:** RB is **not** next to BT (satisfies the constraint).
   - **Constraint 2:** GD is next to RB (satisfies the constraint).

3. GD - RB - BT
   - **Constraint 1:** RB is next to BT (violates the constraint).

4. GD - BT - RB
   - **Constraint 1:** RB is next to BT (violates the constraint).

5. BT - RB - GD
   - **Constraint 1:** RB is next to BT (violates the constraint).

6. BT - GD - RB
   - **Constraint 1:** RB is **not** next to BT (satisfies the constraint).
   - **Constraint 2:** GD is next to RB (satisfies the constraint).

Only arrangements **2** and **6** satisfy both constraints. In both cases, the green dinosaur (GD) is the toy in the middle.

**Conclusion:**

**The toy in the middle is the green dinosaur.**

**Answer:** The green dinosaur is in the middle.
\end{lstlisting}
\end{quote}

\noindent\hrulefill 

\noindent\textbf{D.11 REAP OpenAI o1-preview Response - Score: 100\%}
\begin{quote}
\begin{lstlisting}[basicstyle=\ttfamily\small, breaklines=true]
# REAP Results:

a. **Comprehensive Feature List**

- Objects mentioned:
  - "a red ball"
  - "a blue truck"
  - "a green dinosaur"

- Directional, spatial, relational phrases:
  - "The red ball is not next to the blue truck"
  - "The green dinosaur is next to the red ball"

- Stated conditions or constraints:
  - "The red ball is not next to the blue truck"
  - "The green dinosaur is next to the red ball"

- Stated goals or objectives:
  - "Which toy is in the middle?"

b. **Sequential and Mechanical Process Check**

- No sequential, cyclical, or mechanical processes are described in the problem statement.

c. **Key Insight Check Results**

- The key insight is that the green dinosaur is next to the red ball, and the red ball is not next to the blue truck. With only three toys, this helps determine their positions.

d. **Rephrased Question**

- "Identify which toy is positioned between the other two."

e. **Known and Deduced Information**

**Explicit Facts:**

- "[In a toy box, there's a red ball, a blue truck, and a green dinosaur.]"
- "[The red ball is not next to the blue truck]"
- "[The green dinosaur is next to the red ball]"

**Feature-Based Deductions:**

- **Deduction 1:**
  - Deduction: "The red ball is next to the green dinosaur"
  - Logical Basis: Explicitly stated in "[The green dinosaur is next to the red ball]"

- **Deduction 2:**
  - Deduction: "The blue truck is next to the green dinosaur"
  - Logical Basis: There are only three toys; since the red ball is not next to the blue truck and the green dinosaur is next to the red ball, the blue truck must be next to the green dinosaur.
  - Based on: "[In a toy box, there's a red ball, a blue truck, and a green dinosaur.]", "[The red ball is not next to the blue truck]", "[The green dinosaur is next to the red ball]"

f. **Problem Decomposition**

- **Component 1: Toys in the toy box**
  - Description: The three toys present.
  - Supporting Quote(s): "[In a toy box, there's a red ball, a blue truck, and a green dinosaur.]"

- **Component 2: Red ball is not next to blue truck**
  - Description: Red ball and blue truck are not adjacent.
  - Supporting Quote(s): "[The red ball is not next to the blue truck]"

- **Component 3: Green dinosaur is next to red ball**
  - Description: Green dinosaur is adjacent to red ball.
  - Supporting Quote(s): "[The green dinosaur is next to the red ball]"

- **Component 4: Blue truck is next to green dinosaur**
  - Deduction: Blue truck must be adjacent to green dinosaur.
  - Based on: The only remaining possibility given the other constraints.
  - Based on: "[The red ball is not next to the blue truck]", "[The green dinosaur is next to the red ball]", "[In a toy box, there's a red ball, a blue truck, and a green dinosaur.]"

g. **Graph of Thought**

- **Nodes:**
  - Red Ball
  - Green Dinosaur
  - Blue Truck

- **Edges:**
  - Red Ball - *next to* - Green Dinosaur
  - Green Dinosaur - *next to* - Blue Truck
  - Red Ball - *not next to* - Blue Truck

- **Insights:**
  - The green dinosaur connects the red ball and the blue truck.
  - The green dinosaur is between the red ball and the blue truck.

h. **Spatial and Object Analysis**

- **Objects:**
  - "a red ball"
  - "a blue truck"
  - "a green dinosaur"

- **Spatial Relationships:**
  - The green dinosaur is next to the red ball: "[The green dinosaur is next to the red ball]"
  - The red ball is not next to the blue truck: "[The red ball is not next to the blue truck]"
  - With only three toys, the blue truck must be next to the green dinosaur.

- **Visualization:**
  - Possible arrangements:
    1. [Red Ball] - [Green Dinosaur] - [Blue Truck]
    2. [Blue Truck] - [Green Dinosaur] - [Red Ball]
  - In both arrangements, the green dinosaur is in the middle.

i. **Bayesian Updates**

- The certainty that the green dinosaur is in the middle is 100% based on explicit information and logical deductions.

j. **Ethical Check and Decision-Making Under Uncertainty**

- Not applicable; there are no ethical concerns or severe risks in this problem.

k. **Multiple Solutions**

- **Solution 1:**
  - Arrangement: [Red Ball] - [Green Dinosaur] - [Blue Truck]
  - Supports:
    - "[The green dinosaur is next to the red ball]"
    - "[The red ball is not next to the blue truck]"

- **Solution 2:**
  - Arrangement: [Blue Truck] - [Green Dinosaur] - [Red Ball]
  - Supports:
    - "[The green dinosaur is next to the red ball]"
    - "[The red ball is not next to the blue truck]"
- **Conclusion:**
  - In both solutions, the green dinosaur is the toy in the middle.

l. **Quickest and Easiest Solution**

- The green dinosaur is in the middle.
- **Reasoning:**
  - The green dinosaur is next to the red ball.
  - The red ball is not next to the blue truck.
  - Therefore, the green dinosaur must be between the red ball and the blue truck.

m. **Reflection**

- All deductions are directly supported by the explicit statements.
- Both possible arrangements lead to the same answer.
- No assumptions beyond the given information were made.
- The analysis directly answers the question: "Which toy is in the middle?"

**Recommendation:**

Based on the explicit information provided, the toy in the middle is **the green dinosaur**. This conclusion is logically derived from the given conditions without making any unwarranted assumptions.
\end{lstlisting}
\end{quote}

\noindent\hrulefill 

\noindent\textbf{D.11 Zero-Shot OpenAI o1-mini Response - Score: 100\%}
\begin{quote}
\begin{lstlisting}[basicstyle=\ttfamily\small, breaklines=true]
Let's analyze the possible arrangements of the three toys: red ball (RB), blue truck (BT), and green dinosaur (GD).

**Given Constraints:**
1. **RB is not next to BT.**
2. **GD is next to RB.**

**Possible Arrangements:**

1. **RB, BT, GD**  
   - **Issue:** RB is next to BT. **(Violates Constraint 1)**

2. **RB, GD, BT**  
   - **RB is not next to BT.**  
   - **GD is next to RB.**  
   - **Valid Arrangement**

3. **BT, RB, GD**  
   - **Issue:** RB is next to BT. **(Violates Constraint 1)**

4. **BT, GD, RB**  
   - **RB is not next to BT.**  
   - **GD is next to RB.**  
   - **Valid Arrangement**

5. **GD, RB, BT**  
   - **Issue:** RB is next to BT. **(Violates Constraint 1)**

6. **GD, BT, RB**  
   - **Issue:** RB is next to BT. **(Violates Constraint 1)**

**Valid Arrangements:**
- **RB, GD, BT**
- **BT, GD, RB**

In both valid arrangements, the **green dinosaur (GD)** is positioned in the middle.

**Answer:** The green dinosaur is the middle toy.
\end{lstlisting}
\end{quote}

\noindent\hrulefill 

\noindent\textbf{D.11 REAP OpenAI o1-mini Response - Score: 100\%}
\begin{quote}
\begin{lstlisting}[basicstyle=\ttfamily\small, breaklines=true]
# REAP Results: :

**a. Comprehensive Feature List**

- **Objects Mentioned:**
  - "a red ball"
  - "a blue truck"
  - "a green dinosaur"

- **Location:**
  - "In a toy box"

- **Relationships and Conditions:**
  - "The red ball is not next to the blue truck"
  - "the green dinosaur is next to the red ball"

- **Question:**
  - "Which toy is in the middle?"

**b. Sequential and Mechanical Process Check**

- **Sequence of Relationships:**
  - The red ball is adjacent to the green dinosaur.
  - The red ball is not adjacent to the blue truck.

- **Influence on Outcome:**
  - These relationships determine the positional arrangement of the toys within the toy box.

**c. Key Insight Check Results**

- **Key Insight Identified:**
  - The red ball must be positioned between the green dinosaur and the blue truck to satisfy both conditions:
    - "the green dinosaur is next to the red ball"
    - "The red ball is not next to the blue truck"

**d. Rephrased Question**

Which one of the three toys-the red ball, blue truck, or green dinosaur-is positioned in the center of the toy box?

**e. Known and Deduced Information**

- **Explicit Facts:**
  - "a red ball"
  - "a blue truck"
  - "a green dinosaur"
  - "In a toy box"
  - "The red ball is not next to the blue truck"
  - "the green dinosaur is next to the red ball"

- **Feature-Based Deduction:**
  - **Deduction:** The red ball must be in the middle position.
    - **Logical Basis:** Since the red ball is next to the green dinosaur and not next to the blue truck, it must be between them.
    - **Based on:** "The red ball is not next to the blue truck", "the green dinosaur is next to the red ball"

**f. Problem Decomposition**

- **Components:**
  - **Toy 1:** "a red ball"
  - **Toy 2:** "a blue truck"
  - **Toy 3:** "a green dinosaur"

- **Relationships:**
  - **Red Ball and Blue Truck:**
    - **Supporting Quote:** "The red ball is not next to the blue truck"
  - **Red Ball and Green Dinosaur:**
    - **Supporting Quote:** "the green dinosaur is next to the red ball"

- **Positional Arrangement:**
  - **Red Ball Positioning:**
    - Must be adjacent to the green dinosaur and not adjacent to the blue truck.

**g. Graph of Thought**

- **Nodes:**
  - Red Ball
  - Blue Truck
  - Green Dinosaur

- **Edges:**
  - Green Dinosaur \\(\leftrightarrow\\) Red Ball (adjacent)
  - Red Ball \\(\not\\subset\\) Blue Truck (not adjacent)


- **Connections & Insights:**
  - The adjacency between the red ball and green dinosaur implies they must be next to each other.
  - The non-adjacency between the red ball and blue truck implies the blue truck must be positioned away from the red ball, likely at an end.

**h. Spatial and Object Analysis**

- **Objects Identified:**
  - Red Ball
  - Blue Truck
  - Green Dinosaur

- **Spatial Relationships:**
  - "the green dinosaur is next to the red ball" - indicating proximity.
  - "The red ball is not next to the blue truck" - indicating separation.

- **Positioning:**
  - To satisfy both relationships, the red ball must be between the green dinosaur and the blue truck.

- **Visualization:**
  - Arrangement: Green Dinosaur | Red Ball | Blue Truck

**i. Bayesian Updates**

- **No new information** is introduced that alters the initial relationships or positional deductions.

**j. Ethical Check and Decision-Making Under Uncertainty**

- **Not Applicable:** The problem pertains to spatial arrangement of toys with no ethical or safety implications.

**k. Multiple Solutions**

- **Potential Arrangements:**
  1. Green Dinosaur | Red Ball | Blue Truck
  2. Blue Truck | Green Dinosaur | Red Ball

- **Evaluation:**
  - Arrangement 2 violates the condition "the red ball is not next to the blue truck".
  - **Valid Solution:** Green Dinosaur | Red Ball | Blue Truck

**l. Quickest and Easiest Solution**

Based on the explicit information:
- Position the red ball in the middle.
- Place the green dinosaur next to the red ball.
- Position the blue truck at the end, ensuring it is not next to the red ball.

**m. Reflection**

- The solution adheres strictly to the provided relationships without making any unwarranted assumptions.
- All conclusions directly stem from the explicit statements in the problem.

**Recommendation:**

The green dinosaur is in the middle of the toy box.
\end{lstlisting}
\end{quote}

\noindent\hrulefill 

\noindent\textbf{D.11 Zero-Shot OpenAI GPT-4o Response - Score: 100\%}
\begin{quote}
\begin{lstlisting}[basicstyle=\ttfamily\small, breaklines=true]
Let's break down the information:

1. **Red Ball**: It's not next to the blue truck.
2. **Green Dinosaur**: It's next to the red ball.

Since the green dinosaur must be next to the red ball, and the red ball cannot be next to the blue truck, the only possible arrangement that satisfies both conditions is:

- The **Green Dinosaur** is next to the **Red Ball**.
- The **Blue Truck** is not next to the **Red Ball**.

This means the order must be: **Blue Truck** - **Green Dinosaur** - **Red Ball**.

In this arrangement, the **Green Dinosaur** is in the middle.
\end{lstlisting}
\end{quote}
\noindent\hrulefill

\noindent\textbf{D.11 REAP OpenAI GPT-4o Response - Score: 100\%}
\begin{quote}
\begin{lstlisting}[basicstyle=\ttfamily\small, breaklines=true]
# REAP Results:

### a. **Comprehensive Feature List (with explicit quotes)**

1. **Objects:**
   - "red ball"
   - "blue truck"
   - "green dinosaur"

2. **Actors/Entities:**
   - "red ball"
   - "blue truck"
   - "green dinosaur"

3. **Actions/Conditions:**
   - "The red ball is not next to the blue truck."
   - "The green dinosaur is next to the red ball."

4. **Directional/Relational/Sequential Phrases:**
   - "The red ball is not next to the blue truck."
   - "The green dinosaur is next to the red ball."

### b. **Sequential and Mechanical Process Check**

1. The sequence of information provided:
   - The red ball is **not** next to the blue truck.
   - The green dinosaur **is** next to the red ball.

2. **Key Questions:**
   - Does the position of the red ball being "not next to" the blue truck force a particular arrangement?
   - Does the fact that the green dinosaur is "next to" the red ball lock in any specific arrangement?

### c. **Key Insight Check Results**

- **Key Insight:** The relationship "The green dinosaur is next to the red ball" and "The red ball is not next to the blue truck" suggests that the green dinosaur must be on one side of the red ball, with the blue truck on the opposite side.

### d. **Rephrased Question:**

Which toy is positioned in the middle of the three?

### e. **Known and Deduced Information**

1. **Explicit Facts:**
   - "[The red ball is not next to the blue truck.]" - This means the red ball and blue truck must have at least one toy between them.
   - "[The green dinosaur is next to the red ball.]" - This means the green dinosaur is either to the left or right of the red ball.

2. **Feature-Based Deduction:**
   - Deduction: The blue truck must be on the opposite side of the green dinosaur with respect to the red ball. 
   - Logical Basis: Since the red ball cannot be next to the blue truck, and the green dinosaur is next to the red ball, the green dinosaur must be between the red ball and the blue truck.
   - Based on: "[The red ball is not next to the blue truck.]" and "[The green dinosaur is next to the red ball.]"

### f. **Problem Decomposition**

- **Component 1:** Red ball
  - **Supporting Quote:** "The red ball is not next to the blue truck."
- **Component 2:** Blue truck
  - **Supporting Quote:** "The red ball is not next to the blue truck."
- **Component 3:** Green dinosaur
  - **Supporting Quote:** "The green dinosaur is next to the red ball."
- **Interaction:** The green dinosaur is next to the red ball, meaning the blue truck is on the opposite side.

### g. **Graph of Thought**

1. **Nodes:** 
   - Red ball, Blue truck, Green dinosaur
2. **Edges:** 
   - "The green dinosaur is next to the red ball."
   - "The red ball is not next to the blue truck."
3. **Connections:** 
   - Green dinosaur -> Red ball -> (not) Blue truck
   - Therefore, the green dinosaur is between the red ball and the blue truck.

### h. **Spatial and Object Analysis**

- **Objects Identified:**
   - Red ball
   - Blue truck
   - Green dinosaur

- **Spatial Relationships:**
   - The green dinosaur is directly adjacent to the red ball.
   - The blue truck is not adjacent to the red ball, implying it is adjacent to the green dinosaur.

- **Visualization:**
   - The red ball is on one end, the green dinosaur in the middle, and the blue truck on the other end.

### i. **Bayesian Updates**

No additional information was provided, so no Bayesian updates are needed.

### j. **Ethical Check and Decision-Making Under Uncertainty**

- There are no significant ethical or safety concerns in this problem. The focus is purely on logical placement.

### k. **Multiple Solutions**

- **Solution 1:** The green dinosaur is in the middle because it is explicitly stated to be next to the red ball, and the red ball cannot be next to the blue truck.
- **Solution 2:** If there were any ambiguity, we could consider rotating the positions, but the problem's constraints only logically allow the green dinosaur to be in the middle.

### l. **Quickest and Easiest Solution**

- The green dinosaur is in the middle. This solution is directly supported by the explicit information that the green dinosaur is next to the red ball, and the red ball is not next to the blue truck.

### m. **Reflection**

- This analysis strictly adhered to the problem's explicit information, ensuring no assumptions or inferences were made beyond what was given. The solution directly answers the question based on the literal interpretation of the provided details.

# **Recommendation:**
The toy in the middle is the green dinosaur. This conclusion is reached by understanding the stated relationships between the toys and ensuring no assumptions are made beyond the given information.
\end{lstlisting}
\end{quote}
\noindent\hrulefill

\noindent\textbf{D.11 Zero-Shot OpenAI GPT-4o-mini Response - Score: 0\%}
\begin{quote}
\begin{lstlisting}[basicstyle=\ttfamily\small, breaklines=true]
To determine which toy is in the middle, let's analyze the clues given:

1. The red ball is not next to the blue truck.
2. The green dinosaur is next to the red ball.

Let's start by considering the possible arrangements:

1. If the green dinosaur is next to the red ball, the possible arrangements of these two toys are:
   - Green dinosaur - Red ball
   - Red ball - Green dinosaur

2. The third toy, the blue truck, must be placed in such a way that it is not next to the red ball.

Let's test each possible arrangement:

**Arrangement 1:** Green dinosaur - Red ball - Blue truck
   - Here, the green dinosaur is next to the red ball.
   - The red ball is not next to the blue truck, so this arrangement satisfies both conditions.

**Arrangement 2:** Blue truck - Green dinosaur - Red ball
   - Here, the green dinosaur is next to the red ball.
   - The red ball is not next to the blue truck, so this arrangement also satisfies both conditions.

**Arrangement 3:** Green dinosaur - Blue truck - Red ball
   - Here, the green dinosaur is next to the red ball.
   - The red ball is next to the blue truck, which violates the first condition.

**Arrangement 4:** Red ball - Green dinosaur - Blue truck
   - Here, the green dinosaur is next to the red ball.
   - The red ball is next to the blue truck, which violates the first condition.

From the valid arrangements:

- In Arrangement 1 (Green dinosaur - Red ball - Blue truck), the red ball is in the middle.
- In Arrangement 2 (Blue truck - Green dinosaur - Red ball), the green dinosaur is in the middle.

Therefore, the toy in the middle can be either the red ball or the green dinosaur, depending on the specific arrangement. However, if only one arrangement is required, the more definitive answer based on both conditions and typical placement would be:

**The red ball is in the middle in the arrangement where the green dinosaur is on one side and the blue truck is on the other.**
\end{lstlisting}
\end{quote}
\noindent\hrulefill

\noindent\textbf{D.11 REAP OpenAI GPT-4o-mini Response - Score: 100\%}
\begin{quote}
\begin{lstlisting}[basicstyle=\ttfamily\small, breaklines=true]
# REAP Results:

## a. **Comprehensive Feature List**
1. **Objects Mentioned:**
   - "red ball"
   - "blue truck"
   - "green dinosaur"

2. **Actions and Relationships:**
   - "The red ball is not next to the blue truck."
   - "The green dinosaur is next to the red ball."

3. **Positional Details:**
   - "The red ball is not next to the blue truck." (implies a separation between the red ball and the blue truck)
   - "The green dinosaur is next to the red ball." (implies adjacency between the green dinosaur and the red ball)

## b. **Sequential and Mechanical Process Check**
1. **Sequence and Actions:**
   - The red ball must be placed in a position where it is not adjacent to the blue truck.
   - The green dinosaur must be adjacent to the red ball.

2. **Implications:**
   - Since the red ball and the green dinosaur are next to each other, the green dinosaur cannot be next to the blue truck if the red ball is not next to the blue truck.

## c. **Key Insight Check Results**
- The critical insight is that the green dinosaur is next to the red ball, and the red ball is not next to the blue truck. Therefore, the green dinosaur must be in the middle because it is the only way to satisfy the adjacency condition for the red ball and the separation condition for the blue truck.

## d. **Rephrased Question**
- Which toy is in the middle?

## e. **Known and Deduced Information**
1. **Explicit Facts:**
   - "The red ball is not next to the blue truck."
   - "The green dinosaur is next to the red ball."

2. **Feature-Based Deduction:**
   - Deduction: The green dinosaur is in the middle.
     - Logical Basis: To satisfy the condition that the green dinosaur is next to the red ball and the red ball is not next to the blue truck, the green dinosaur must be between the red ball and the blue truck.

## f. **Problem Decomposition**
1. **Component/Subcomponent:**
   - **Green Dinosaur's Position:**
     - Supporting Quote: "The green dinosaur is next to the red ball."
     - Deduction: Since the red ball must be next to the green dinosaur and the red ball cannot be next to the blue truck, the green dinosaur must be in the middle.

2. **Component/Subcomponent:**
   - **Red Ball and Blue Truck Position:**
     - Supporting Quote: "The red ball is not next to the blue truck."
     - Deduction: The red ball must be on one side of the green dinosaur and the blue truck must be on the other side.

## g. **Graph of Thought**
1. **Nodes and Relationships:**
   - **Green Dinosaur** - **Next to** - **Red Ball**
   - **Red Ball** - **Not Next to** - **Blue Truck**
   - **Green Dinosaur** - **Must be Middle** - Based on adjacency constraints and separation requirements.

2. **Exploration:**
   - The green dinosaur must be in the middle to satisfy both conditions: being next to the red ball and ensuring the red ball is not next to the blue truck.

## h. **Spatial and Object Analysis**
1. **Objects and Positions:**
   - **Green Dinosaur**: Must be between the red ball and the blue truck.
   - **Red Ball**: Must be next to the green dinosaur.
   - **Blue Truck**: Must be on the other side of the green dinosaur from the red ball.

2. **Constraints:**
   - The green dinosaur is effectively positioned in the middle to satisfy all conditions given.

## i. **Bayesian Updates**
- Based on the explicit information provided, no additional updates are necessary as the problem's requirements are clear.

## j. **Ethical Check and Decision-Making Under Uncertainty**
- There are no significant risks or ethical concerns in this problem as it involves arranging toys based on provided constraints.

## k. **Multiple Solutions**
- There is only one solution that satisfies all conditions: the green dinosaur must be in the middle.

## l. **Quickest and Easiest Solution**
- The simplest solution based on the given information is that the green dinosaur is in the middle.

## m. **Reflection**
- The solution aligns with all the explicit conditions provided in the problem statement. No assumptions or extrapolations beyond the stated information have been made.

**Recommendation:**

The toy that is in the middle is the **green dinosaur**. This conclusion is based on the provided constraints that the green dinosaur must be next to the red ball, and the red ball cannot be next to the blue truck.
\end{lstlisting}
\end{quote}
\noindent\hrulefill

\noindent\textbf{D.11 Zero-Shot Google Gemini 1.5 Pro Response - Score: 0\%}
\begin{quote}
\begin{lstlisting}[basicstyle=\ttfamily\small, breaklines=true]
The red ball is in the middle. Here's why:

* **The red ball is not next to the blue truck:** This means they can't be touching.
* **The green dinosaur is next to the red ball:** This means they must be touching.

To satisfy both conditions, the only possible arrangement is:

* Blue Truck - Red Ball - Green Dinosaur
\end{lstlisting}
\end{quote}
\noindent\hrulefill

\noindent\textbf{D.11 REAP Google Gemini 1.5 Pro Response - Score: 20\%}
\begin{quote}
\begin{lstlisting}[basicstyle=\ttfamily\small, breaklines=true]
# REAP Results:

a. **Comprehensive Feature List** (with explicit quotes)

- "In a toy box" 
- "there's a red ball" 
- "a blue truck"
- "and a green dinosaur" 
- "The red ball is not next to the blue truck"
- "and the green dinosaur is next to the red ball" 
- "Which toy is in the middle?"

b. **Sequential and Mechanical Process Check**

- No explicit sequential or mechanical processes are described.

c. **Key Insight Check Results**

- No immediate key insight that leads to a direct answer.

d. **Rephrased Question:**

- Which toy is positioned between the other two toys?

e. **Known and Deduced Information**

* Explicit Facts:

- "In a toy box" - The toys are located within a container.
- "there's a red ball" - A red ball exists in the toy box
- "a blue truck" - A blue truck exists in the toy box.
- "and a green dinosaur" - A green dinosaur exists in the toy box
- "The red ball is not next to the blue truck"
- "and the green dinosaur is next to the red ball" 

* Feature-Based Deduction:

- Deduction: The red ball must be in the middle.
- Logical Basis: The green dinosaur is next to the red ball, and the red ball is not next to the blue truck. This arrangement forces the red ball to be between the other two toys.
- Based on: "The red ball is not next to the blue truck", "and the green dinosaur is next to the red ball"

f. **Problem Decomposition**

* Component/Subcomponent: Three toys in a linear arrangement within the toy box
* Supporting Quote(s): "In a toy box", "there's a red ball", "a blue truck", "and a green dinosaur"
* Deduction: The toys must be in a linear arrangement because the problem asks which one is "in the middle," implying a sequence.

* Component/Subcomponent: Relative positions of the toys
* Supporting Quote(s): "The red ball is not next to the blue truck", "and the green dinosaur is next to the red ball"

* Component/Subcomponent: Identifying the middle toy
* Supporting Quote(s): "Which toy is in the middle?"

g. **Graph of Thought**

* Key Concepts: Red Ball, Blue Truck, Green Dinosaur, Middle Position
* Relationships:
  - Green Dinosaur is next to Red Ball
  - Red Ball is not next to Blue Truck
* Graph (Textual):
  - Nodes: Red Ball, Blue Truck, Green Dinosaur
  - Edges:
    - Green Dinosaur -- Next to --> Red Ball
    - Red Ball -- Not next to --> Blue Truck
* Insights:
  - The only arrangement that satisfies both relationships is: Blue Truck -- Green Dinosaur -- Red Ball OR Red Ball -- Green Dinosaur -- Blue Truck. In either case, the Red Ball is in the middle

h. **Spatial and Object Analysis**

* Objects: Red ball, Blue truck, Green dinosaur, Toy box
* Spatial Relationships:
  - Green dinosaur is next to the red ball
  - Red ball is not next to the blue truck
* Constraints: The toys are confined within the toy box
* Object Properties: Not explicitly mentioned
* Spatial Implications: The spatial arrangement directly answers the question of which toy is in the middle

i. **Bayesian Updates** 
* Not applicable to this problem

j. **Ethical Check and Decision-Making Under Uncertainty**
* Not applicable to this problem

k. **Multiple Solutions**

* There are two possible solutions that satisfy all the given conditions:
  1. Blue Truck - Green Dinosaur - Red Ball
  2. Red Ball - Green Dinosaur - Blue Truck

l. **Quickest and Easiest Solution**

The red ball is in the middle.

* Reasoning:
  - The green dinosaur is next to the red ball ("and the green dinosaur is next to the red ball").
  - The red ball is not next to the blue truck ("The red ball is not next to the blue truck").
  - The only way to satisfy both conditions is if the red ball is positioned between the green dinosaur and the blue truck.

m. **Reflection**

* The analysis directly answers the question asked in the problem.
* All conclusions are logically sound and supported by the explicit information given in the problem statement
* No assumptions or inferences were made beyond the given information
* The solution is straightforward and doesn't require any complex reasoning or additional information

**Recommendation:**

The red ball is in the middle. 
\end{lstlisting}
\end{quote}
\noindent\hrulefill

\noindent\textbf{D.11 Zero-Shot Claude 3.5 Sonnet Response - Score: 100\%}
\begin{quote}
\begin{lstlisting}[basicstyle=\ttfamily\small, breaklines=true]
Let's think through this step-by-step:

1. We have three toys: a red ball, a blue truck, and a green dinosaur.

2. We're given two pieces of information:
   - The red ball is not next to the blue truck.
   - The green dinosaur is next to the red ball.

3. Given these conditions, there's only one possible arrangement:
   Blue truck - Green dinosaur - Red ball

4. The green dinosaur must be in the middle to satisfy both conditions:
   - It allows the red ball to be not next to the blue truck.
   - It allows the green dinosaur to be next to the red ball.

Therefore, the green dinosaur is in the middle of the toy box.
\end{lstlisting}
\end{quote}
\noindent\hrulefill

\noindent\textbf{D.11 REAP Claude 3.5 Sonnet Response - Score: 100\%}
\begin{quote}
\begin{lstlisting}[basicstyle=\ttfamily\small, breaklines=true]
# REAP Results:

a. **Comprehensive Feature List** (with explicit quotes):
1. "In a toy box, there's a red ball, a blue truck, and a green dinosaur."
   - Three toys are mentioned in a toy box.
2. "The red ball is not next to the blue truck"
   - Specifies a spatial relationship between two toys.
3. "the green dinosaur is next to the red ball"
   - Specifies another spatial relationship between two toys.
4. "Which toy is in the middle?"
   - The question being asked about the arrangement of toys.

b. **Sequential and Mechanical Process Check**:
There is no explicit sequential or mechanical process described in this problem. The problem presents a static arrangement of toys.

c. **Key Insight Check Results**:
The key insight is that with three objects and the given relationships, there is only one possible arrangement that satisfies all conditions:
blue truck - green dinosaur - red ball
This arrangement is the only one where the red ball is not next to the blue truck, and the green dinosaur is next to the red ball.

d. **Rephrased Question:** Which toy is in the middle of the arrangement in the toy box?

e. **Known and Deduced Information**:
Explicit Facts:
- "In a toy box, there's a red ball, a blue truck, and a green dinosaur." - Lists the toys present.
- "The red ball is not next to the blue truck" - Defines a spatial relationship.
- "the green dinosaur is next to the red ball" - Defines another spatial relationship.

Feature-Based Deduction:
Deduction: The green dinosaur must be in the middle.
Logical Basis: Given that there are three toys and two stated relationships, the only arrangement that satisfies both conditions is blue truck - green dinosaur - red ball.
Based on: "The red ball is not next to the blue truck", "the green dinosaur is next to the red ball"

f. **Problem Decomposition**:
Component 1: Toys in the box
Supporting Quote: "In a toy box, there's a red ball, a blue truck, and a green dinosaur."

Component 2: Spatial relationship between red ball and blue truck
Supporting Quote: "The red ball is not next to the blue truck"

Component 3: Spatial relationship between green dinosaur and red ball
Supporting Quote: "the green dinosaur is next to the red ball"

g. **Graph of Thought**:
Nodes: Red Ball, Blue Truck, Green Dinosaur
Edges: 
- Red Ball -- (not next to) --> Blue Truck
- Green Dinosaur -- (next to) --> Red Ball

This graph shows that the only possible arrangement is Blue Truck - Green Dinosaur - Red Ball.

h. **Spatial and Object Analysis**:
Objects: Red ball, Blue truck, Green dinosaur
Spatial Relationships:
- The red ball is not adjacent to the blue truck
- The green dinosaur is adjacent to the red ball
Visualization: In a linear arrangement, the only possible order is Blue truck - Green dinosaur - Red ball

i. **Bayesian Updates**:
No new information is provided beyond the initial problem statement, so no Bayesian updates are necessary.

j. **Ethical Check and Decision-Making Under Uncertainty**:
This problem does not involve any ethical considerations or severe risks. It's a straightforward logical puzzle about the arrangement of toys.

k. **Multiple Solutions**:
Given the constraints of the problem, there is only one possible solution:
Solution 1: The green dinosaur is in the middle.
Supporting information: 
- "The red ball is not next to the blue truck"
- "the green dinosaur is next to the red ball"
These two statements, combined with the fact that there are only three toys, logically lead to a single possible arrangement.

l. **Quickest and Easiest Solution**:
The green dinosaur is in the middle.
Reasoning: This is the only arrangement that satisfies all the given conditions. The red ball can't be in the middle because it would be next to the blue truck. The blue truck can't be in the middle because it would be next to the red ball. Therefore, the green dinosaur must be in the middle to be next to the red ball while keeping the red ball not next to the blue truck.

m. **Reflection**:
The analysis is logically sound and based entirely on the given information. No assumptions were made beyond what was explicitly stated in the problem. The solution directly answers the question asked about which toy is in the middle. The problem provides sufficient information to reach a definitive conclusion without any ambiguity.

**Recommendation**:
Based on the given information and logical deduction, the green dinosaur is in the middle of the arrangement in the toy box. This conclusion is reached by considering the spatial relationships explicitly stated in the problem and is the only arrangement that satisfies all given conditions.
\end{lstlisting}
\end{quote}
\noindent\hrulefill

\label{subsec:question12}
\subsection{\textbf{\large Question 12: Four children - Alex, Bella, Charlie, and Dana - are sitting around a picnic table. Alex is facing Bella. Charlie is sitting to the right of Bella. Who is sitting to the left of Alex?}}
\label{sec:question12}

\noindent\hrulefill 

\noindent\textbf{Human Response - Score 100\%:}
\begin{quote}
Dana
\end{quote}
\noindent\hrulefill 

\noindent\textbf{D.12 Zero-Shot OpenAI o1-preview Response - Score: 0\%}
\begin{quote}
\begin{lstlisting}[basicstyle=\ttfamily\small, breaklines=true]
To solve this problem, let's visualize the seating arrangement step by step using the given clues.

**1. Determine Alex and Bella's positions:**
- **Alex is facing Bella**, which means they are sitting opposite each other.
- Let's place **Alex** at the **South** position of the table and **Bella** at the **North** position.

**2. Determine Charlie's position relative to Bella:**
- **Charlie is sitting to the right of Bella.**
- Since Bella is facing South (toward Alex), her right would be to the **East**.
- So, **Charlie** is sitting at the **East** position.

**3. Determine Dana's position:**
- The only remaining position is the **West**, so **Dana** must be sitting there.

**Seating Summary:**
- **North**: Bella
- **South**: Alex
- **East**: Charlie
- **West**: Dana

**4. Find out who is to the left of Alex:**
- Alex is facing North, so his left is towards the **East**.
- The person sitting at the East is **Charlie**.

**Answer: Charlie is sitting to the left of Alex.**

**Answer:** Charlie
\end{lstlisting}
\end{quote}
\noindent\hrulefill 

\noindent\textbf{D.12 REAP OpenAI o1-preview Response - Score: 0\%}
\begin{quote}
\begin{lstlisting}[basicstyle=\ttfamily\small, breaklines=true]
**REAP Results:**

a. **Comprehensive Feature List** (with explicit quotes)

1. **Entities:**
   - "Four children - Alex, Bella, Charlie, and Dana"

2. **Actions/Positions:**
   - "are sitting around a picnic table."
   - "Alex is facing Bella."
   - "Charlie is sitting to the right of Bella."

3. **Question:**
   - "Who is sitting to the left of Alex?"

b. **Sequential and Mechanical Process Check**

- No sequential or mechanical processes are described.
- Spatial relationships are provided:
  - "Alex is facing Bella."
  - "Charlie is sitting to the right of Bella."

c. **Key Insight Check Results**

- Key details:
  - Alex and Bella are facing each other.
  - Charlie is to the right of Bella.
- Without assumptions about the table shape or orientation, we cannot determine who is to the left of Alex.

d. **Rephrased Question:**

"Determine who is sitting to the left of Alex based on the given information."

e. **Known and Deduced Information**

**Explicit Facts:**

1. "Four children - Alex, Bella, Charlie, and Dana - are sitting around a picnic table."
2. "Alex is facing Bella."
3. "Charlie is sitting to the right of Bella."
4. "Who is sitting to the left of Alex?"

**Deductions:**

- No certain deductions can be made beyond the explicit facts.

f. **Problem Decomposition**

1. **Participants:**
   - Description: Four children are involved.
   - Supporting Quote: "Four children - Alex, Bella, Charlie, and Dana - are sitting around a picnic table."

2. **Alex and Bella's Relationship:**
   - Description: Alex is directly facing Bella.
   - Supporting Quote: "Alex is facing Bella."

3. **Charlie's Position Relative to Bella:**
   - Description: Charlie is to the right of Bella.
   - Supporting Quote: "Charlie is sitting to the right of Bella."

4. **Unknown Position of Dana:**
   - Description: No information about Dana's position.
   - Supporting Quote: Not provided.

g. **Graph of Thought**

- **Nodes (Children):** Alex, Bella, Charlie, Dana
- **Edges (Relationships):**
  - Alex <-> Bella (facing each other)
  - Bella -> Charlie (Charlie is to the right of Bella)
- **Insights:**
  - The positions of Alex, Bella, and Charlie are partially related.
  - Dana's position is unspecified.

h. **Spatial and Object Analysis**

- **Objects:**
  - "Four children - Alex, Bella, Charlie, and Dana"
  - "a picnic table"

- **Spatial Relationships:**
  - Alex is opposite Bella.
  - Charlie is to the right of Bella.

- **Visualization Limitations:**
  - The shape and orientation of the table are unspecified.
  - Insufficient data to create an accurate seating diagram.

i. **Bayesian Updates**

- No probability updates can be made due to lack of explicit information.

j. **Ethical Check and Decision-Making Under Uncertainty**

- No ethical concerns or severe risks are present.
- Decision: Acknowledge the lack of sufficient information.

k. **Multiple Solutions**

- **Solution 1:** Dana is sitting to the left of Alex.
  - Based on possibility, but not explicitly supported.

- **Solution 2:** Charlie is sitting to the left of Alex.
  - Based on possibility, but not explicitly supported.

- **Solution 3:** Bella is sitting to the left of Alex.
  - Contradicts "Alex is facing Bella," so not possible.

- **Note:** All solutions require assumptions beyond explicit information.

l. **Quickest and Easiest Solution**

- Answer: "Cannot determine who is sitting to the left of Alex based on the given information."

m. **Reflection**

- All steps adhere strictly to the explicit problem statements.
- No assumptions were made beyond the given information.
- The problem lacks enough data to provide a definitive answer.
- Attempting to assign positions would violate the rules against assumptions.

**Recommendation:**

Based on the explicit information provided, it is not possible to determine who is sitting to the left of Alex. The problem lacks sufficient information to answer the question definitively without making assumptions beyond what is explicitly stated. Therefore, we cannot determine who is sitting to the left of Alex.
\end{lstlisting}
\end{quote}
\noindent\hrulefill 

\noindent\textbf{D.12 Zero-Shot OpenAI o1-mini Response - Score: 0\%}
\begin{quote}
\begin{lstlisting}[basicstyle=\ttfamily\small, breaklines=true]
Let's break down the seating arrangement step by step:

1. **Positioning Alex and Bella:**
   - **Alex** is facing **Bella**, which means they are sitting directly opposite each other around the picnic table.

2. **Identifying Charlie's Position:**
   - **Charlie** is sitting to the **right** of **Bella**.
   - From Bella's perspective, facing Alex, Charlie would be on Bella's **East** side.

3. **Determining Dana's Position:**
   - With Alex, Bella, and Charlie positioned, the only remaining spot is the **West** side, where **Dana** sits.

4. **Finding Who is to the Left of Alex:**
   - **Alex** is facing **Bella** (South).
   - To Alex's **left** (from his perspective) is the **East** side of the table.
   - **Charlie** is seated on the East side.

**Conclusion:**  
**Charlie** is sitting to the left of Alex.
\end{lstlisting}
\end{quote}
\noindent\hrulefill 

\noindent\textbf{D.12 REAP OpenAI o1-mini Response - Score: 100\%}
\begin{quote}
\begin{lstlisting}[basicstyle=\ttfamily\small, breaklines=true]
# REAP Results:

**a. Comprehensive Feature List**
- "Four children - Alex, Bella, Charlie, and Dana - are sitting around a picnic table."
- "Alex is facing Bella."
- "Charlie is sitting to the right of Bella."
- "Who is sitting to the left of Alex?"

**b. Sequential and Mechanical Process Check**
- **Sequence of Seating Positions:**
  1. Four children are seated around a picnic table.
  2. Alex faces Bella.
  3. Charlie sits to the right of Bella.
- **Influence on Outcome:**
  - The seating arrangement is determined by the positions relative to each other.
  - Understanding who is to the right and facing whom helps establish the circular seating order.

**c. Key Insight Check Results**
- **Key Insight:** Since Alex is facing Bella and Charlie is to Bella's right, we can determine the positions of all children around the table to identify who is to the left of Alex.

**d. Rephrased Question:**
- "Given the seating arrangement around the picnic table, who is positioned to the left of Alex?"

**e. Known and Deduced Information**
- **Explicit Facts:**
  - "Four children - Alex, Bella, Charlie, and Dana - are sitting around a picnic table."
  - "Alex is facing Bella."
  - "Charlie is sitting to the right of Bella."
  - "Who is sitting to the left of Alex?"
- **Feature-Based Deduction:**
  - Deduction: Dana is sitting to the left of Alex.
    - Logical Basis: If Alex faces Bella and Charlie is to Bella's right, the only remaining position to Alex's left is occupied by Dana.
    - Based on: "Alex is facing Bella," "Charlie is sitting to the right of Bella."

**f. Problem Decomposition**
- **Component/Subcomponent: Seating Arrangement**
  - **Supporting Quote(s):** "Four children - Alex, Bella, Charlie, and Dana - are sitting around a picnic table."
- **Component/Subcomponent: Facing Direction**
  - **Supporting Quote(s):** "Alex is facing Bella."
- **Component/Subcomponent: Relative Position**
  - **Supporting Quote(s):** "Charlie is sitting to the right of Bella."
- **Component/Subcomponent: Query on Position**
  - **Supporting Quote(s):** "Who is sitting to the left of Alex?"

**g. Graph of Thought**
- **Nodes:**
  - Children: Alex, Bella, Charlie, Dana
  - Positions: Facing direction, right, left
- **Connections:**
  - Alex <-> Bella (facing each other)
  - Bella -> Charlie (Charlie is to Bella's right)
  - Deduce Dana's position based on the above connections

**h. Spatial and Object Analysis**
- **Objects:** Picnic table, four children
- **Spatial Relationships:**
  - Alex is directly opposite Bella.
  - Charlie is immediately to Bella's right from her seated position.
- **Movement and Dynamics:** Static seating arrangement with defined facing and relative positions.
- **Constraints:** Fixed positions based on facing and right-hand side relationships.
- **Visualization:** Circular table with Alex and Bella opposite each other; Charlie to Bella's right implies Dana occupies the remaining seat opposite Charlie, which is to Alex's left.

**i. Bayesian Updates**
- **Initial Belief:** Dana's position is uncertain.
- **Updated Belief:** Based on explicit seating relationships, Dana must be to the left of Alex.
  
**j. Ethical Check and Decision-Making Under Uncertainty**
- Not applicable in this context as the problem involves a logical seating arrangement without ethical implications.

**k. Multiple Solutions**
1. **Solution 1:** Dana is sitting to the left of Alex.
   - **Supporting Information:** "Alex is facing Bella," "Charlie is sitting to the right of Bella."
2. **Solution 2:** Insufficient information to consider alternative positions.
   - **Supporting Information:** All other positions except Dana's would contradict the given relationships.
3. **Solution 3:** N/A
   - **Supporting Information:** Only one consistent seating arrangement fits all given facts.

**l. Quickest and Easiest Solution**
- **Answer:** Dana is sitting to the left of Alex.
- **Reasoning:** Given that Alex faces Bella and Charlie is to Bella's right, the only remaining position is to Alex's left, occupied by Dana.

**m. Reflection**
- The analysis strictly adheres to the explicitly stated information without making unwarranted assumptions.
- The seating arrangement logically leads to Dana being to the left of Alex.
- No contradictions are found between the deductions and the provided facts.

**Recommendation:**
Based on the explicit information provided, Dana is sitting to the left of Alex. This conclusion directly answers the question using only the given relationships and positions, ensuring accuracy without introducing assumptions.
\end{lstlisting}
\end{quote}

\noindent\hrulefill 

\noindent\textbf{D.12 Zero-Shot OpenAI GPT-4o Response - Score: 0\%}
\begin{quote}
\begin{lstlisting}[basicstyle=\ttfamily\small, breaklines=true]
To solve this, let's visualize the arrangement of the four children around the picnic table based on the given information:

1. **Alex is facing Bella.**
   - This means they are sitting opposite each other.

2. **Charlie is sitting to the right of Bella.**
   - Since Charlie is to Bella's right, and Bella is facing Alex, Charlie must be sitting to Bella's right, which means to the left of Alex.

Thus, the seating arrangement is:

- Alex is facing Bella.
- Charlie is to the right of Bella (which places Charlie to the left of Alex).
- This leaves Dana as the person sitting opposite Charlie and to the left of Bella.

So, **Charlie** is sitting to the left of Alex.
\end{lstlisting}
\end{quote}
\noindent\hrulefill

\noindent\textbf{D.12 REAP OpenAI GPT-4o Response - Score: 100\%}
\begin{quote}
\begin{lstlisting}[basicstyle=\ttfamily\small, breaklines=true]
# REAP Results:

### a. **Comprehensive Feature List** (with explicit quotes)

- **"Four children - Alex, Bella, Charlie, and Dana - are sitting around a picnic table."**  
  - This indicates that there are four distinct children, each occupying a seat around a table.
- **"Alex is facing Bella."**  
  - Alex and Bella are sitting opposite each other.
- **"Charlie is sitting to the right of Bella."**  
  - Charlie's position is relative to Bella's.
- **"Who is sitting to the left of Alex?"**  
  - This is the question we need to answer based on the provided information.

### b. **Sequential and Mechanical Process Check**

- The seating arrangement involves a circular setup around a picnic table, meaning the sequence of positions is interconnected.
- The relative positions of each child influence the positions of the others.

### c. **Key Insight Check Results**

- The key insight is the circular nature of the seating arrangement. Since Alex is facing Bella, Bella is opposite Alex, and Charlie is to the right of Bella, this constrains the possible positions of all children.

### d. **Rephrased Question:** 

- Who is sitting in the position directly to the left of Alex in this circular arrangement?

### e. **Known and Deduced Information**

- **"Alex is facing Bella."**  
  - Alex and Bella are sitting opposite each other.  
  - Logical Reverse: Bella is also facing Alex.
- **"Charlie is sitting to the right of Bella."**  
  - Deduction: Since Bella and Alex are opposite, Charlie is either to the left or right of Alex depending on the direction we are considering.
- **"Who is sitting to the left of Alex?"**  
  - The question requires determining the position directly to the left of Alex in this setup.

### f. **Problem Decomposition**

- **Component:** Alex and Bella's Positions  
  - **Supporting Quote:** "Alex is facing Bella."
  - **Deduction:** Alex and Bella are opposite each other.
- **Component:** Charlie's Position  
  - **Supporting Quote:** "Charlie is sitting to the right of Bella."
  - **Deduction:** If Bella is facing Alex, Charlie must be either to Alex's left or right, depending on the circular arrangement.
- **Component:** Left of Alex  
  - **Supporting Quote:** "Who is sitting to the left of Alex?"
  - **Deduction:** Based on the positioning, we need to determine which of the remaining children (Dana or Charlie) fits the description.

### g. **Graph of Thought** 

- **Nodes:** Alex, Bella, Charlie, Dana
- **Connections:** 
  - Alex is opposite Bella.
  - Charlie is to the right of Bella.
  - Dana is the remaining child who must fill the remaining seat.
- **Pathways:** 
  - If Charlie is to the right of Bella, then Dana must be the one to the left of Alex (as there are only four positions).

### h. **Spatial and Object Analysis**

- **Objects Identified:** Alex, Bella, Charlie, Dana
- **Spatial Relationships:**
  - Alex and Bella are opposite each other.
  - Charlie is to the right of Bella, indicating a specific direction relative to Bella's facing direction.
  - Since the seating is circular, Dana must occupy the remaining position.
- **Spatial Implications:** The circular arrangement means that the left of Alex must be filled by the only remaining child after considering the positions of Bella and Charlie.

### i. **Bayesian Updates**

- No new explicit information was introduced to change the understanding of the seating positions. The positions were determined based on the problem's explicit information.

### j. **Ethical Check and Decision-Making Under Uncertainty**

- There are no ethical implications or severe risks involved in this problem. The focus is on correctly interpreting spatial relationships.

### k. **Multiple Solutions**

1. **Solution 1:**  
   - **Based on the analysis, Dana must be sitting to the left of Alex.**
   - **Support:** Since Charlie is to the right of Bella and Bella is opposite Alex, the only remaining position for Dana is to Alex's left.
   
2. **Solution 2:**  
   - **If we consider an alternative directional interpretation (i.e., clockwise vs. counterclockwise for "right"), the answer could differ, but this isn't explicitly supported by the problem statement.**
   
3. **Solution 3:**  
   - **Reinterpretation:** If the problem meant something else by "right," we might get a different solution, but we stick with the literal interpretation that aligns with the common understanding of circular seating.

### l. **Quickest and Easiest Solution**

- **Dana is sitting to the left of Alex** is the simplest and safest answer based on the given information. The problem directly leads to this conclusion without needing to consider alternative or complex interpretations.

### m. **Reflection**

- The solution relies heavily on the assumption of standard circular seating and the relative positions as described.
- The reasoning process adhered strictly to the problem's explicit wording, ensuring that no assumptions beyond the given information were made.
- There is no indication that alternative interpretations should be considered beyond what is stated.

### **Recommendation:**

Dana is sitting to the left of Alex. This conclusion is reached through a clear and direct interpretation of the problem statement, ensuring the solution aligns with the provided information while avoiding unnecessary assumptions.
\end{lstlisting}
\end{quote}
\noindent\hrulefill

\noindent\textbf{D.12 Zero-Shot OpenAI GPT-4o-mini Response - Score: 100\%}
\begin{quote}
\begin{lstlisting}[basicstyle=\ttfamily\small, breaklines=true]
To determine who is sitting to the left of Alex, we need to establish the seating arrangement of all four children around the picnic table based on the given information:

1. **Alex is facing Bella.** This means Alex and Bella are directly opposite each other.

2. **Charlie is sitting to the right of Bella.** Since Charlie is to the right of Bella, in a clockwise direction from Bella's perspective, Charlie would be seated next to Bella.

The seating arrangement is:

Dana - Alex - Charlie - Bella

Thus, Dana is sitting to the left of Alex.
\end{lstlisting}
\end{quote}
\noindent\hrulefill

\noindent\textbf{D.12 REAP OpenAI GPT-4o-mini Response - Score: 100\%}
\begin{quote}
\begin{lstlisting}[basicstyle=\ttfamily\small, breaklines=true]
# REAP Results:

**a. Comprehensive Feature List**

1. "Four children - Alex, Bella, Charlie, and Dana - are sitting around a picnic table."
   - Objects: Picnic table.
   - Actors/Entities: Alex, Bella, Charlie, Dana.
   - This establishes the setup but does not specify exact seating positions.

2. "Alex is facing Bella."
   - Action: Alex is facing Bella.
   - Relational/Directional: Alex and Bella are opposite each other.

3. "Charlie is sitting to the right of Bella."
   - Action: Charlie is to the right of Bella.
   - Relational/Directional: Charlie is on Bella's right side.

**b. Sequential and Mechanical Process Check**

- No specific sequential or mechanical processes are described. The problem provides static positional information about the seating arrangement.

**c. Key Insight Check Results**

- The key insight is understanding the seating arrangement based on the given directions:
  - Alex is facing Bella.
  - Charlie is to the right of Bella.

**d. Rephrased Question**

- Who is sitting to the left of Alex?

**e. Known and Deduced Information**

1. "Alex is facing Bella."
   - Deduction: Bella is directly opposite Alex.

2. "Charlie is sitting to the right of Bella."
   - Deduction: Charlie's position is relative to Bella's right.

**f. Problem Decomposition**

1. Seating arrangement needs to be determined.
   - Components: Alex, Bella, Charlie, Dana.
   - Spatial relationships to be established based on given information:
     - Alex faces Bella (opposite positions).
     - Charlie is to Bella's right.
   - This implies:
     - Bella is to Alex's left.
     - Charlie is to Bella's right, which is to Alex's front-right.

2. Finding the position of Dana is necessary to complete the seating arrangement.

**g. Graph of Thought**

- **Nodes**: Alex, Bella, Charlie, Dana.
- **Edges**:
  - Alex <-> Bella (opposite).
  - Bella -> Charlie (right side).
  - From this, Dana must be on the remaining seat.

- **Connections**:
  - Since Alex faces Bella, and Charlie is to Bella's right, Dana must be positioned directly to the left of Alex (opposite Charlie's position).

**h. Spatial and Object Analysis**

- **Objects**: Picnic table.
- **Spatial Relationships**:
  - Alex is opposite Bella.
  - Charlie is to the right of Bella.
  - Dana occupies the remaining position.
- **Movement and Dynamics**: No movement or dynamics specified; only static positions.

**i. Bayesian Updates**

- No new information updates the belief about the positions beyond what is explicitly stated.

**j. Ethical Check and Decision-Making Under Uncertainty**

- The problem does not involve any risk or ethical concerns.

**k. Multiple Solutions**

1. **Solution 1**:
   - Alex faces Bella.
   - Charlie is to Bella's right.
   - Therefore, Dana must be to the left of Alex.

2. **Solution 2**:
   - Alex faces Bella.
   - If Dana were to be placed anywhere other than to the left of Alex, it would contradict the given directions.

**l. Quickest and Easiest Solution**

- The simplest solution based on explicit information is:
  - Dana is sitting to the left of Alex.

**m. Reflection**

- All information used is explicitly stated or directly deduced from explicit quotes.
- The positioning of Dana is logically consistent with the given spatial relationships and directly answers the question.

**Recommendation:**

The child sitting to the left of Alex is Dana, as deduced from the given spatial relationships and seating arrangement.
\end{lstlisting}
\end{quote}
\noindent\hrulefill

\noindent\textbf{D.12 Zero-Shot Google Gemini 1.5 Pro Response - Score: 0\%}
\begin{quote}
\begin{lstlisting}[basicstyle=\ttfamily\small, breaklines=true]
I do not have enough information about that person to help with your request. I am a large language model, and I am able to communicate and generate human-like text in response to a wide range of prompts and questions, but my knowledge about this person is limited. Is there anything else I can do to help you with this request?
\end{lstlisting}
\end{quote}
\noindent\hrulefill

\noindent\textbf{D.12 REAP Google Gemini 1.5 Pro Response - Score: 100\%}
\begin{quote}
\begin{lstlisting}[basicstyle=\ttfamily\small, breaklines=true]
# REAP Results:

a. **Comprehensive Feature List** (with explicit quotes)

* "Four children - Alex, Bella, Charlie, and Dana - are sitting around a picnic table." 
* "Alex is facing Bella." 
* "Charlie is sitting to the right of Bella." 
* "Who is sitting to the left of Alex?" 

b. **Sequential and Mechanical Process Check**

* There is no explicit sequence of events or mechanical processes described in the problem

c. **Key Insight Check Results**

* No immediate key insight that leads to a direct answer or significant simplification

d. **Rephrased Question:**

* Who is to the left of Alex?

e. **Known and Deduced Information**

* Explicit Facts:
    * "Four children - Alex, Bella, Charlie, and Dana - are sitting around a picnic table." - Introduces the individuals and the setting
    * "Alex is facing Bella." - Establishes a directional relationship between Alex and Bella
    * "Charlie is sitting to the right of Bella." - Establishes a directional relationship between Charlie and Bella
    * "Who is sitting to the left of Alex?" - The question to be answered

* Feature-Based Deduction
    * Deduction: Bella is facing Alex - Logical Basis: If Alex is facing Bella, then Bella must also be facing Alex - Based on: "Alex is facing Bella"
    * Deduction: Dana is sitting to the left of Bella and to the right of Alex - Logical Basis: There are four people around the table. We know the relative positions of three. The fourth must occupy the remaining positions. - Based on: "Four children - Alex, Bella, Charlie, and Dana - are sitting around a picnic table.", "Alex is facing Bella.", "Charlie is sitting to the right of Bella."

f. **Problem Decomposition**

* Component/Subcomponent: Circular arrangement of four people
    * Supporting Quote(s): "Four children - Alex, Bella, Charlie, and Dana - are sitting around a picnic table."

* Component/Subcomponent: Facing relationship between Alex and Bella
    * Supporting Quote(s): "Alex is facing Bella."

* Component/Subcomponent: Relative position of Charlie to Bella
    * Supporting Quote(s): "Charlie is sitting to the right of Bella."

* Component/Subcomponent: The question of who is to the left of Alex
    * Supporting Quote(s): "Who is sitting to the left of Alex?"

g. **Graph of Thought**

* Key Concepts: Alex, Bella, Charlie, Dana, picnic table, facing, right, left
* Relationships:
    * Alex is facing Bella
    * Charlie is to the right of Bella
    * (Implied) The children are in a circular arrangement around the table
* Graph (Textual):
    * Alex --faces--> Bella
    * Bella --right--> Charlie
    * (Implied) Circular arrangement: Alex - Bella - Charlie - Dana

h. **Spatial and Object Analysis**

* Objects: Alex, Bella, Charlie, Dana, picnic table
* Spatial Relationships:
    * Alex and Bella are facing each other across the table
    * Charlie is to Bella's right, implying they are sitting next to each other
    * (Implied) The four children are arranged in a circle around the picnic table
* Constraints: The circular arrangement of the table and the fixed positions of the children once seated

i. **Bayesian Updates**

* Initial Belief: Any of the three other children (Bella, Charlie, or Dana) could potentially be to the left of Alex
* Update 1: Alex is facing Bella, so Bella cannot be to Alex's left
* Update 2: Charlie is to Bella's right, fixing his position relative to Bella and indirectly to Alex

j. **Ethical Check and Decision-Making Under Uncertainty**

* No ethical concerns in this problem

k. **Multiple Solutions**

* Based on the explicit information and deductions, there is only one possible solution

l. **Quickest and Easiest Solution**

* Dana is sitting to the left of Alex

m. **Reflection**

The analysis directly answers the question based solely on the given information. No assumptions or inferences beyond the explicit wording of the problem were made. The conclusion is logically sound and supported by the evidence from the problem statement

**Recommendation:**

The person sitting to the left of Alex is Dana. 
\end{lstlisting}
\end{quote}
\noindent\hrulefill

\noindent\textbf{D.12 Zero-Shot Claude 3.5 Sonnet Response - Score: 0\%}
\begin{quote}
\begin{lstlisting}[basicstyle=\ttfamily\small, breaklines=true]
Let's think through this step-by-step:

1. We have four children sitting around a picnic table: Alex, Bella, Charlie, and Dana.

2. We know that Alex is facing Bella. This means they are sitting opposite each other.

3. Charlie is sitting to the right of Bella.

4. We need to determine who is sitting to the left of Alex.

5. Let's visualize the seating arrangement:
   - Alex is facing Bella
   - Charlie is to Bella's right
   - This leaves only one spot for Dana, which is to Bella's left

6. Since Alex is facing Bella, Dana (who is to Bella's left) would be to Alex's right.

7. Therefore, the person sitting to Alex's left must be Charlie.

So, Charlie is sitting to the left of Alex.
\end{lstlisting}
\end{quote}
\noindent\hrulefill

\noindent\textbf{D.12 REAP Claude 3.5 Sonnet Response - Score: 100\%}
\begin{quote}
\begin{lstlisting}[basicstyle=\ttfamily\small, breaklines=true]
# REAP Results:

a. Comprehensive Feature List (with explicit quotes):
1. "Four children - Alex, Bella, Charlie, and Dana"
2. "sitting around a picnic table"
3. "Alex is facing Bella"
4. "Charlie is sitting to the right of Bella"
5. "Who is sitting to the left of Alex?"

b. Sequential and Mechanical Process Check:
There is no explicit sequential or mechanical process described in this problem.

c. Key Insight Check Results:
The key insight is that the children are sitting around a picnic table, which implies a circular seating arrangement. This, combined with the given positions of Alex, Bella, and Charlie, allows us to deduce the position of Dana.

d. Rephrased Question: Who is sitting to the left of Alex?

e. Known and Deduced Information:

Explicit Facts:
- "Four children - Alex, Bella, Charlie, and Dana" - This establishes the participants.
- "sitting around a picnic table" - This describes their arrangement.
- "Alex is facing Bella" - This establishes a direct relationship between Alex and Bella's positions.
- "Charlie is sitting to the right of Bella" - This establishes Charlie's position relative to Bella.

Feature-Based Deduction:
Deduction: Dana must be sitting to the left of Alex.
Logical Basis: In a circular arrangement with four people, if Alex is facing Bella and Charlie is to Bella's right, the only remaining position (to Alex's left) must be occupied by Dana.
Based on: "Four children - Alex, Bella, Charlie, and Dana", "sitting around a picnic table", "Alex is facing Bella", "Charlie is sitting to the right of Bella"

f. Problem Decomposition:

Component 1: Seating Arrangement
Supporting Quote: "sitting around a picnic table"
Deduction: The seating arrangement is circular.
Based on: "sitting around a picnic table"

Component 2: Alex's Position
Supporting Quote: "Alex is facing Bella"

Component 3: Charlie's Position
Supporting Quote: "Charlie is sitting to the right of Bella"

Component 4: Dana's Position
Deduction: Dana must be in the remaining position, to the left of Alex.
Based on: All supporting quotes and deductions above.

g. Graph of Thought:
Nodes: Alex, Bella, Charlie, Dana
Edges: 
- Alex faces Bella
- Charlie is to the right of Bella
- Dana is to the left of Alex (deduced)

h. Spatial and Object Analysis:
Objects: Picnic table, four children (Alex, Bella, Charlie, Dana)
Spatial Relationships:
- The children are arranged in a circle around the picnic table.
- Alex is directly across from Bella.
- Charlie is to the right of Bella.
- By deduction, Dana must be to the left of Alex.

i. Bayesian Updates:
Not applicable in this problem as all information is given upfront and there are no updates to prior beliefs.

j. Ethical Check and Decision-Making Under Uncertainty:
There are no ethical concerns or severe risks associated with this problem. The solution can be deduced from the given information without any safety considerations.

k. Multiple Solutions:
There is only one solution that satisfies all the given conditions.

l. Quickest and Easiest Solution:
Dana is sitting to the left of Alex.
Reasoning: Given the circular arrangement and the positions of Alex, Bella, and Charlie, Dana must occupy the remaining position to the left of Alex.

m. Reflection:
The analysis adheres strictly to the given information without making unwarranted assumptions. The solution directly answers the question asked and is logically sound based on the explicit problem statement. No contradictions or mistakes were identified in the reasoning process.

Recommendation:
Based on the given information and deductions, the answer to the question "Who is sitting to the left of Alex?" is Dana. This recommendation is derived solely from the explicitly stated information about the seating arrangement and the relative positions of Alex, Bella, and Charlie. The circular nature of the seating arrangement around a picnic table allows for a definitive conclusion without any need for additional assumptions or information.
\end{lstlisting}
\end{quote}
\noindent\hrulefill

\label{subsec:question13}
\subsection{\textbf{\large Question 13: A man leaves home at 0m elevation, makes a left turn and walks straight for a km and reaches 300m elevation, makes another left turn and walks straight for a km and reaches 500m elevation, makes another left turn and walks straight for a km and reaches 900m elevation, and turns left again and walks straight for a km. How far away is he from his starting point and what is his final elevation?}}
\label{sec:question13}

\noindent\hrulefill 

\noindent\textbf{Human Response - Score 100\%:}
\begin{quote}
He is back at his starting point and at 0m elevation.
\end{quote}

\noindent\hrulefill 

\noindent\textbf{D.13 Zero-Shot OpenAI o1-preview Response - Score: 100\%}
\begin{quote}
\begin{lstlisting}[basicstyle=\ttfamily\small, breaklines=true]
To solve this problem, let's interpret the man's movements carefully by mapping his path in a three-dimensional coordinate system where the \( x \)-axis points east, the \( y \)-axis points north, and the \( z \)-axis represents elevation.

**Starting Point**:
- Coordinates: \( (0, 0, 0\,\text{m}) \)

**First Leg**:
- Direction: He makes a left turn from his initial facing direction (assuming north), so he moves west.
- Movement: Walks \( 1\,\text{km} \) west.
- Elevation Change: Gains \( 300\,\text{m} \).
- Coordinates after first leg: \( (-1\,\text{km}, 0, 300\,\text{m}) \).

**Second Leg**:
- Direction: Another left turn, now facing south.
- Movement: Walks \( 1\,\text{km} \) south.
- Elevation Change: Gains \( 200\,\text{m} \).
- Coordinates after second leg: \( (-1\,\text{km}, -1\,\text{km}, 500\,\text{m}) \).

**Third Leg**:
- Direction: Left turn again, now facing east.
- Movement: Walks \( 1\,\text{km} \) east.
- Elevation Change: Gains \( 400\,\text{m} \).
- Coordinates after third leg: \( (0, -1\,\text{km}, 900\,\text{m}) \).

**Fourth Leg**:
- Direction: Left turn, now facing north.
- Movement: Walks \( 1\,\text{km} \) north.
- Elevation Change: Unknown; this is what we need to find.
- Coordinates after fourth leg: \( (0, 0, Z) \), where \( Z \) is the final elevation.

Since horizontally he returns to his starting point (\( x = 0, y = 0 \)), the horizontal distance from his starting point is **\( 0\,\text{km} \)**.

Now, to find the final elevation (\( Z \)), consider the total elevation change over the closed loop. Since he ends at his starting horizontal position, for the loop to be "closed" in three dimensions, the total elevation change should return to the starting elevation. Therefore, the sum of the elevation gains and losses over the four legs should be zero.

Total elevation gain:
- First Leg: \( +300\,\text{m} \)
- Second Leg: \( +200\,\text{m} \)
- Third Leg: \( +400\,\text{m} \)
- Fourth Leg: \( \Delta Z = Z - 900\,\text{m} \) (elevation change from the third leg's end to the final elevation \( Z \)).

Setting the sum of elevation changes around the loop to zero:
\[
+300\,\text{m} + 200\,\text{m} + 400\,\text{m} + (Z - 900\,\text{m}) = 0.
\]
Simplify:
\[
900\,\text{m} + Z - 900\,\text{m} = 0
\]
\[
Z = 0\,\text{m}.
\]
So, his final elevation is **\( 0\,\text{m} \)**.

**Answer:**
He's back where he started - 0 km away at an elevation of 0 meters (the same as his starting point).
\end{lstlisting}
\end{quote}

\noindent\hrulefill 

\noindent\textbf{D.13 REAP OpenAI o1-preview Response - Score: 0\%}
\begin{quote}
\begin{lstlisting}[basicstyle=\ttfamily\small, breaklines=true]
# REAP Results:

a. **Comprehensive Feature List** (with explicit quotes)

- "A man"
- "leaves home at 0m elevation"
- "makes a left turn"
- "walks straight for a km"
- "reaches 300m elevation"
- "makes another left turn"
- "walks straight for a km"
- "reaches 500m elevation"
- "makes another left turn"
- "walks straight for a km"
- "reaches 900m elevation"
- "turns left again"
- "walks straight for a km"
- Question: "How far away is he from his starting point and what is his final elevation?"

b. **Sequential and Mechanical Process Check**

- **Sequence of Events:**
  1. "A man leaves home at 0m elevation"
  2. He "makes a left turn" and "walks straight for a km" and "reaches 300m elevation"
  3. He "makes another left turn" and "walks straight for a km" and "reaches 500m elevation"
  4. He "makes another left turn" and "walks straight for a km" and "reaches 900m elevation"
  5. He "turns left again" and "walks straight for a km"
- **Analysis of Sequence:**
  - The man makes a total of four left turns.
  - He walks "straight for a km" after each turn.
  - Elevations at the end of each km are 300m, 500m, and 900m.
  - The final elevation after the last km is not provided.

c. **Key Insight Check Results**

- No immediate key insight found that directly answers the question based solely on the explicit information.
- The sequence suggests possible patterns, but conclusions cannot be drawn without making assumptions.

d. **Rephrased Question**

- "What is the man's distance from his starting point, and what is his final elevation?"

e. **Known and Deduced Information**

- **Explicit Facts:**
  - "A man leaves home at 0m elevation"
  - "makes a left turn" and "walks straight for a km" and "reaches 300m elevation"
  - "makes another left turn" and "walks straight for a km" and "reaches 500m elevation"
  - "makes another left turn" and "walks straight for a km" and "reaches 900m elevation"
  - "turns left again" and "walks straight for a km"
- **Deductions:**
  - No deductions can be made that are 100% certain based on the explicit wording.

f. **Problem Decomposition**

- **Component 1: Departure**
  - Description: The man starts at "home at 0m elevation".
  - Supporting Quote: "A man leaves home at 0m elevation"
- **Component 2: First Leg**
  - Description: He "makes a left turn", "walks straight for a km", and "reaches 300m elevation".
  - Supporting Quotes: "makes a left turn", "walks straight for a km", "reaches 300m elevation"
- **Component 3: Second Leg**
  - Description: He "makes another left turn", "walks straight for a km", and "reaches 500m elevation".
  - Supporting Quotes: "makes another left turn", "walks straight for a km", "reaches 500m elevation"
- **Component 4: Third Leg**
  - Description: He "makes another left turn", "walks straight for a km", and "reaches 900m elevation".
  - Supporting Quotes: "makes another left turn", "walks straight for a km", "reaches 900m elevation"
- **Component 5: Fourth Leg**
  - Description: He "turns left again" and "walks straight for a km".
  - Supporting Quotes: "turns left again", "walks straight for a km"
- **Interaction:**
  - Each leg involves a left turn and a 1 km walk.
  - Elevation changes occur after each leg except the last.

g. **Graph of Thought**

- **Nodes (Points in Sequence):**
  1. Starting point at "0m elevation"
  2. After first km at "300m elevation"
  3. After second km at "500m elevation"
  4. After third km at "900m elevation"
  5. After fourth km (final point), elevation unknown
- **Edges (Paths Traveled):**
  - Each edge represents "walks straight for a km" after a left turn.
- **Connections & Insights:**
  - The path consists of four sequential legs with left turns.
  - Elevation increases are given for the first three legs.
  - Without explicit spatial coordinates, the relationship between starting and ending points is undetermined.

h. **Spatial and Object Analysis**

- **Objects Identified:**
  - "A man"
  - "home"
- **Spatial Relationships:**
  - The man begins at "home at 0m elevation".
  - He makes sequential left turns and walks "straight for a km" each time.
- **Movement and Dynamics:**
  - Elevation increases to 300m, then 500m, then 900m after the first three legs.
  - Final elevation after the fourth leg is not provided.
- **Constraints:**
  - Distance of each leg is "a km".
  - Elevation change details are incomplete for the final leg.

i. **Bayesian Updates**

- **Assessment:**
  - Based solely on explicit statements, probabilities cannot be updated regarding the man's final position or elevation.
  - There is uncertainty due to lack of explicit information.

j. **Ethical Check and Decision-Making Under Uncertainty**

- No ethical considerations are applicable to this problem.

k. **Multiple Solutions**

1. **Solution 1: Insufficient Information**
   - **Explanation:** There is not enough explicit information to determine the man's final distance from the starting point or his final elevation.
   - **Supporting Information:** The problem does not provide the man's final elevation or confirm his return to the starting point.

2. **Solution 2: Assumed Return to Starting Point**
   - **Explanation:** Assuming that four left turns and equal distances bring him back to the starting point at "0m elevation".
   - **Supporting Information:** This is not explicitly stated; therefore, we cannot rely on this assumption.

3. **Solution 3: Distance of 1 km from Starting Point**
   - **Explanation:** The man is 1 km away from his starting point after the last leg, but his final elevation is unknown.
   - **Supporting Information:** Without explicit directions, this is speculative.

l. **Quickest and Easiest Solution**

- **Answer:** Based on the explicit information given, we cannot determine how far the man is from his starting point or his final elevation.

m. **Reflection**

- **Consistency Check:**
  - All information used is directly from the problem statement.
- **Analysis:**
  - There is no explicit statement indicating the man's final elevation.
  - The problem does not confirm whether the man returns to his starting point.
  - Making assumptions about his final position or elevation would violate the Strict Interpretation Rule.
- **Conclusion:**
  - The problem lacks sufficient explicit information to answer the question definitively.

**Recommendation:**

Based solely on the explicit information provided, we cannot determine how far the man is from his starting point or what his final elevation is. The problem does not supply enough data to answer the question without making assumptions, which is not permitted under the Strict Interpretation Rule. Therefore, the answer is:

There is insufficient information to determine the man's distance from his starting point and his final elevation based solely on the given details.
\end{lstlisting}
\end{quote}

\noindent\hrulefill 

\noindent\textbf{D.13 Zero-Shot OpenAI o1-mini Response - Score: 0\%}
\begin{quote}
\begin{lstlisting}[basicstyle=\ttfamily\small, breaklines=true]
Let's break down the man's journey step by step to determine his final position and elevation.

### Step-by-Step Analysis:

1. **Starting Point:**
   - **Position:** Home at \((0, 0)\) in the horizontal plane.
   - **Elevation:** \(0 \, \text{m}\).

2. **First Move:**
   - **Action:** Makes a left turn and walks \(1 \, \text{km}\).
   - **Elevation Gain:** Reaches \(300 \, \text{m}\).
   - **Horizontal Displacement:** Since he's walking uphill, the actual horizontal distance covered is slightly less than \(1 \, \text{km}\). However, for simplicity, we'll focus on elevation changes.
   - **Position:** Now at a new horizontal location (let's assume due north for consistency).

3. **Second Move:**
   - **Action:** Makes another left turn and walks \(1 \, \text{km}\).
   - **Elevation Gain:** Reaches \(500 \, \text{m}\).
   - **Position:** After two left turns (assuming 90-degree turns), he's effectively moving west.

4. **Third Move:**
   - **Action:** Makes a third left turn and walks \(1 \, \text{km}\).
   - **Elevation Gain:** Reaches \(900 \, \text{m}\).
   - **Position:** After three left turns, he's moving south.

5. **Fourth Move:**
   - **Action:** Makes a final left turn and walks \(1 \, \text{km}\).
   - **Elevation Gain:** Based on the pattern of elevation changes (\(+300 \, \text{m}\), \(+200 \, \text{m}\), \(+400 \, \text{m}\)), it's reasonable to infer the next gain is \(+200 \, \text{m}\), reaching a final elevation of \(1100 \, \text{m}\).
   - **Position:** After four left turns, he'll be moving east, effectively returning to the original horizontal starting point.

### Final Calculations:

- **Horizontal Displacement:** After four left turns (each being 90 degrees), he returns to his original horizontal position. Therefore, his horizontal displacement is effectively \(0 \, \text{km}\).
  
- **Elevation:**
  - **Final Elevation:** \(1100 \, \text{m}\).

- **Distance from Starting Point:**
  - Since there's no horizontal displacement, the distance is purely vertical.
  - **Distance:** \(1100 \, \text{m}\) away from the starting point.

### Conclusion:

**He is 1.5 kilometres from home and stands at an elevation of 1,100 metres.**
\end{lstlisting}
\end{quote}

\noindent\hrulefill 

\noindent\textbf{D.13 REAP OpenAI o1-mini Response - Score: 40\%}
\begin{quote}
\begin{lstlisting}[basicstyle=\ttfamily\small, breaklines=true]
# REAP Results:

### a. **Comprehensive Feature List**
- **Objects Mentioned:**
  - "home"
  - "left turn"
  - "walks straight"
  - "km" (kilometer)
  - "elevation"
- **Actors or Entities Involved:**
  - "A man"
- **Actions Described:**
  - "leaves home"
  - "makes a left turn"
  - "walks straight for a km"
  - "reaches [elevation]m"
  - "turns left again"
- **Directional, Spatial, Relational, and Sequential Phrases:**
  - "left turn"
  - "walks straight"
  - "reaches [elevation]m"
  - "turns left again"
  - "final elevation"
  - "starting point"
- **Stated Conditions or Constraints:**
  - "leaves home at 0m elevation"
  - Each walk is "for a km"
- **Numerical Values or Quantities Mentioned:**
  - "0m elevation"
  - "300m elevation"
  - "500m elevation"
  - "900m elevation"
  - "1 km" (repeated for each walk)
- **Temporal Information:**
  - Sequence of actions: leaves home -> first left turn and walk -> second left turn and walk -> third left turn and walk -> final left turn and walk
- **Spatial Information:**
  - "0m elevation" (starting point)
  - Each movement involves a "left turn" and walking "straight for a km"
- **Causal Relationships:**
  - Each "left turn" followed by walking straight for a km results in reaching a higher elevation
- **Stated Goals or Objectives:**
  - Determine how far away the man is from his starting point
  - Determine his final elevation

### b. **Sequential and Mechanical Process Check**
- **Sequence of Events:**
  1. Leaves home at 0m elevation
  2. Makes a left turn and walks straight for a km -> reaches 300m elevation
  3. Makes another left turn and walks straight for a km -> reaches 500m elevation
  4. Makes another left turn and walks straight for a km -> reaches 900m elevation
  5. Turns left again and walks straight for a km
- **Influence of Sequence:**
  - Each left turn changes the man's direction, potentially forming a geometric path (e.g., square or rectangular)
  - Elevation increases with each segment walked
- **Key Questions Addressed:**
  - Does the sequence lock in a particular outcome? Potentially, yes, the path may form a closed shape
  - If the problem involves a mechanical process, how does each step impact the overall system? Each step involves a turn and elevation gain
  - Are there any steps or sequences that must logically follow one another? Yes, each left turn and subsequent walk follows the previous movement

### c. **Key Insight Check Results**
- **Key Insight Identified:**
  - The man makes four consecutive left turns while walking straight for a kilometer each time. This suggests he may be walking in a square or a similar closed geometric path.
- **Implication of Key Insight:**
  - If the man completes a square path, his final position should be adjacent to his starting point, potentially at a different elevation.
- **Confidence in Insight:**
  - This is based solely on the explicit information of making consecutive left turns and walking equal distances.

### d. **Rephrased Question:**
How far is the man from his starting point after walking in four straight segments of 1 km each, making consecutive left turns, and what is his final elevation?

### e. **Known and Deduced Information**

**Explicit Facts:**
- "A man leaves home at 0m elevation."
- "makes a left turn and walks straight for a km and reaches 300m elevation."
- "makes another left turn and walks straight for a km and reaches 500m elevation."
- "makes another left turn and walks straight for a km and reaches 900m elevation."
- "turns left again and walks straight for a km."
- "starting point"
- "final elevation"

**Feature-Based Deduction:**
- **Deduction:** The man is likely walking in a square path.
  - **Logical Basis:** Four consecutive left turns with equal walking distances typically result in a square or rectangular path.
  - **Based on:** "makes a left turn and walks straight for a km," repeated four times.

### f. **Problem Decomposition**

**Component/Subcomponent:** Starting Point
- **Supporting Quote(s):** "A man leaves home at 0m elevation."
- **Deduction:** The man's initial position is at home with an elevation of 0m.

**Component/Subcomponent:** First Segment
- **Supporting Quote(s):** "makes a left turn and walks straight for a km and reaches 300m elevation."
- **Deduction:** After the first 1 km walk and a left turn, his elevation increases to 300m.

**Component/Subcomponent:** Second Segment
- **Supporting Quote(s):** "makes another left turn and walks straight for a km and reaches 500m elevation."
- **Deduction:** After the second 1 km walk and a left turn, his elevation increases to 500m.

**Component/Subcomponent:** Third Segment
- **Supporting Quote(s):** "makes another left turn and walks straight for a km and reaches 900m elevation."
- **Deduction:** After the third 1 km walk and a left turn, his elevation increases to 900m.

**Component/Subcomponent:** Fourth Segment
- **Supporting Quote(s):** "turns left again and walks straight for a km."
- **Deduction:** After the fourth left turn and 1 km walk, his path likely returns close to the starting direction.

### g. **Graph of Thought**
- **Nodes (Key Concepts):**
  - Starting Point (Home at 0m)
  - First Walk (1 km, +300m elevation)
  - Second Walk (1 km, +500m elevation)
  - Third Walk (1 km, +900m elevation)
  - Fourth Walk (1 km, unknown elevation)
- **Edges (Relationships):**
  - Sequence: Starting Point -> First Walk -> Second Walk -> Third Walk -> Fourth Walk
  - Each walk is connected via a left turn
- **Potential Pathways:**
  - If forming a square, the man should end near the starting point but at a higher elevation
- **Solutions & Obstacles:**
  - Calculating the final horizontal displacement based on the square path
  - Determining the cumulative elevation change

### h. **Spatial and Object Analysis**
- **Identify Objects:**
  - Man, Path segments (each 1 km)
- **Spatial Relationships:**
  - Each path segment is connected via a left turn, suggesting a square or rectangular loop
- **Movement and Dynamics:**
  - Forward movement each time with a change in elevation
- **Constraints:**
  - Each walk is exactly 1 km
  - Left turns are consistent
- **Visualization:**
  - A square path with each side being 1 km, starting and ending at positions relative to the starting point
- **Object Properties:**
  - Path segments have equal length (1 km)
  - Elevation increases with each segment
- **Spatial Implications:**
  - The man moves in a geometric pattern, affecting his final horizontal distance from the start

### i. **Bayesian Updates**
- **Initial Belief:** The man is moving in a straight line with consistent elevation gains.
- **Updated Belief:** Given the consecutive left turns and equal distances, the path is likely a square or loop.
- **Reliance on Explicit Information:** Only using the stated turns and distances to update the understanding of the path geometry.

### j. **Ethical Check and Decision-Making Under Uncertainty**
- **Assessment:** The problem does not involve ethical considerations or risks.
- **Decision-Making:** Proceed with mathematical analysis based on explicit information.

### k. **Multiple Solutions**
1. **Solution 1: Square Path Assumption**
   - **Supported Information:** Four consecutive left turns with equal distances suggest a square.
   - **Method:** Calculate the return distance using square geometry; cumulative elevation changes.
   
2. **Solution 2: Spiral Path Interpretation**
   - **Supported Information:** Elevation increases with each segment.
   - **Method:** Consider a helical or spiral path, affecting both horizontal and vertical displacement.
   
3. **Solution 3: Linear Path with Turns Upward**
   - **Supported Information:** Each left turn involves moving upward in elevation.
   - **Method:** Calculate net elevation gain and horizontal displacement separately.

**Note:** Only Solution 1 aligns closely with explicit information without additional assumptions.

### l. **Quickest and Easiest Solution**
- **Solution Selected:** Square Path Assumption
- **Reasoning:** The explicit information of four left turns with equal walking distances strongly suggests a square path, allowing for straightforward calculation of final horizontal distance and elevation.
- **Cited Information:** "makes a left turn and walks straight for a km," repeated four times with specified elevations.

### m. **Reflection**
- **Contradictions:** None identified; the square path assumption aligns with the given information.
- **Logical Soundness:** The analysis follows the explicit sequence of actions without making unsupported assumptions.
- **Information Sufficiency:** The problem provides enough information to model the path geometrically and calculate the required distances and elevations.
- **Direct Answer:** Yes, the analysis directly answers the question based on the given information.

### **Recommendation:**
Based on the explicit information provided, the man walks in a square path, making four consecutive left turns, each for 1 km. His final position brings him adjacent to his starting point horizontally, but at a higher elevation due to the cumulative elevation gains.

- **Final Distance from Starting Point:** Since he completes a square, his final horizontal distance from the starting point is 0 km (he returns to the same horizontal location).
  
- **Final Elevation:** Summing the elevation gains:
  - After first walk: 300m
  - After second walk: 500m
  - After third walk: 900m
  - After fourth walk: Assuming continued elevation gain, pattern suggests the next elevation could be 1800m (900m + 900m). However, the problem does not specify the final elevation after the fourth walk.

**Uncertainties:**
- The final elevation after the fourth walk is not explicitly stated. If following the pattern, it might increase further, but the problem does not provide this information.

**Final Answer:**
The man is 0 kilometers away from his starting point horizontally, and his final elevation is not explicitly stated but follows an increasing pattern reaching at least 900 meters.
\end{lstisting}
\end{quote}

\noindent\hrulefill % This adds a horizontal line between the responses


\noindent\textbf{D.13 Zero-Shot OpenAI GPT-4o Response - Score: 40\%}
\begin{quote}
\begin{lstlisting}[basicstyle=\ttfamily\small, breaklines=true]
To solve this problem, let's break it down step by step:

### Steps:

1. **First leg:**
   - The man starts at 0 meters elevation and walks 1 km, ending at 300 meters elevation.

2. **Second leg:**
   - He makes a left turn and walks another 1 km, ending at 500 meters elevation.

3. **Third leg:**
   - He makes another left turn and walks 1 km, ending at 900 meters elevation.

4. **Fourth leg:**
   - He makes a final left turn and walks 1 km.

### Final Elevation:

The final elevation after the fourth leg is not provided in the problem, but we can determine that there is a consistent pattern of elevation gain: 
- From 0 to 300 meters in the first leg (+300m)
- From 300 to 500 meters in the second leg (+200m)
- From 500 to 900 meters in the third leg (+400m)

If the pattern continues, the fourth leg would have an elevation increase that follows the pattern. Notice that the elevation increases by 100 meters more than the previous increase each time:
- 300m increase in the first leg
- 200m increase in the second leg
- 400m increase in the third leg

Thus, if the pattern holds, the elevation increase in the fourth leg should be \( 400m + 100m = 500m \).

So, the final elevation after the fourth leg is \( 900m + 500m = 1400m \).

### Distance from Starting Point:

Since the man has made three left turns, he is essentially completing a square path. If the starting point is at the origin \((0, 0)\), the path is as follows:

- First leg: He moves in one direction (say positive x) to \((1, 300)\).
- Second leg: He turns left and moves 1 km in another direction (say positive y) to \((1, 500)\).
- Third leg: He turns left and moves 1 km in the opposite direction (say negative x) to \((0, 900)\).
- Fourth leg: He turns left and moves back towards the original direction (negative y).

In a square path with each side being 1 km, the man will return to a point along the starting y-axis but higher up in elevation.

After walking the fourth leg of 1 km, he will be at the point \((0, 1400)\), which is directly above the starting point in the vertical direction.

Therefore:

- **Distance from starting point**: The horizontal distance is 0 km because he is back at the same x and y coordinates where he started.
- **Final elevation**: 1400 meters.

So, the man is **0 km** away from his starting point, and his final elevation is **1400 meters**.
\end{lstlisting}
\end{quote}
\noindent\hrulefill

\noindent\textbf{D.13 REAP OpenAI GPT-4o Response - Score: 40\%}
\begin{quote}
\begin{lstlisting}[basicstyle=\ttfamily\small, breaklines=true]
# REAP Results:

## a. **Comprehensive Feature List** (with explicit quotes)

1. **Object/Entity**: "A man"
2. **Action**: "leaves home at 0m elevation"
3. **Action**: "makes a left turn and walks straight for a km"
4. **State/Condition**: "reaches 300m elevation"
5. **Action**: "makes another left turn and walks straight for a km"
6. **State/Condition**: "reaches 500m elevation"
7. **Action**: "makes another left turn and walks straight for a km"
8. **State/Condition**: "reaches 900m elevation"
9. **Action**: "turns left again and walks straight for a km"
10. **Final Query**: "How far away is he from his starting point and what is his final elevation?"

## b. **Sequential and Mechanical Process Check**

- **Sequence of Movements**:
  1. The man starts at home (0m elevation).
  2. He makes a left turn and walks 1 km, reaching 300m elevation.
  3. He makes another left turn and walks 1 km, reaching 500m elevation.
  4. He makes another left turn and walks 1 km, reaching 900m elevation.
  5. He makes a final left turn and walks 1 km.

- **Mechanical Process**:
  - Each segment involves a left turn and a walk of exactly 1 km.
  - Elevation increases with each segment, starting at 0m and reaching 300m, 500m, and finally 900m.

## c. **Key Insight Check Results**

- **Key Insight**: The sequence of left turns suggests a square or rectangular path, but the elevation changes drastically in each leg. The final distance from the starting point needs to account for both horizontal and vertical components.

## d. **Rephrased Question**:

- "What is the distance from the man's starting point after completing the square-like path, considering both horizontal and vertical components, and what is his final elevation?"

## e. **Known and Deduced Information**

1. **Explicit Facts**:
   - "[A man] leaves home at 0m elevation."
   - "Makes a left turn and walks straight for a km and reaches 300m elevation."
   - "Makes another left turn and walks straight for a km and reaches 500m elevation."
   - "Makes another left turn and walks straight for a km and reaches 900m elevation."
   - "Turns left again and walks straight for a km."
   
2. **Deductions**:
   - Deduction: The path likely forms a square, with each side being 1 km long.
   - Deduction: The final elevation after the last leg isn't explicitly stated, but the pattern suggests the man might reach an elevation higher than 900m.

## f. **Problem Decomposition**

1. **Component**: **Man's path**  
   - **Supporting Quote**: "makes a left turn and walks straight for a km."
   - **Deduction**: Forms a square path.

2. **Component**: **Elevation Changes**  
   - **Supporting Quote**: "reaches 300m, 500m, 900m elevation."
   - **Deduction**: Elevation increases with each segment.

3. **Component**: **Final Position**  
   - **Supporting Quote**: "walks straight for a km." 
   - **Deduction**: The final position might bring him back close to the starting point horizontally.

## g. **Graph of Thought** (describe connections & insights)

- **Nodes**: 
  - Start at 0m elevation.
  - Point after 1st segment: 1 km from start, 300m elevation.
  - Point after 2nd segment: 2 km from start, 500m elevation.
  - Point after 3rd segment: 3 km from start, 900m elevation.
  - Final segment (Unknown elevation): 4 km total path traveled.

- **Edges**: Each edge represents a 1 km segment with a left turn and elevation change.

- **Insight**: The sequence of elevations and turns suggest a continuous increase in altitude, potentially forming a spiral pattern or a mountain path.

## h. **Spatial and Object Analysis**

- **Objects**:
  - The man
  - The path
  - Elevation markers (0m, 300m, 500m, 900m)

- **Spatial Relationships**:
  - The path suggests a square with sides of 1 km.
  - Elevation changes significantly between segments.

- **Movement and Dynamics**:
  - The man consistently moves left after each segment.
  - Elevation increases as he moves forward.

## i. **Bayesian Updates** 

- **Initial Belief**: The man follows a square path, returning to his starting point.
- **Update**: The elevation changes complicate this assumption; the man might end at a higher elevation, indicating a non-flat terrain.

## j. **Ethical Check and Decision-Making Under Uncertainty**

- **Risk Assessment**: There is no direct risk or ethical concern in solving this problem.
- **Decision**: Continue with the logical analysis based on explicit information.

## k. **Multiple Solutions**

1. **Solution 1**: 
   - The man returns to his starting point horizontally but is at a higher elevation.
   - **Support**: The path forms a square, so he returns to the horizontal starting point.
   
2. **Solution 2**:
   - The man ends at a location slightly offset from the starting point due to elevation changes.
   - **Support**: The elevation changes might alter the direct horizontal return.

## l. **Quickest and Easiest Solution**

- The man ends at a distance of 0 km from his starting point horizontally, but at an elevation of 900m or higher.

## m. **Reflection**

- The analysis directly answers the problem's query using the explicitly stated information.
- The only assumption made is the consistent elevation increase, which aligns with the problem's pattern.

---

**Recommendation**:

- The man is likely at his starting point horizontally (0 km away) but at an elevated position. Based on the pattern, his final elevation could be 1200m if the pattern continues, though this isn't explicitly stated. The most certain answer is that he's at 0 km horizontally from the start and at an elevation of at least 900m.
\end{lstlisting}
\end{quote}
\noindent\hrulefill

\noindent\textbf{D.13 Zero-Shot OpenAI GPT-4o-mini Response - Score: 0\%}
\begin{quote}
\begin{lstlisting}[basicstyle=\ttfamily\small, breaklines=true]
Let's break down the problem step by step to determine the distance from the starting point and the final elevation.

1. **Initial Position and Path:**
   - The man starts at the origin, \((0, 0, 0)\), where the coordinates are in the format \((x, y, z)\), with \(z\) representing the elevation.

2. **First Move:**
   - He walks 1 km to the left. Let's assume he turns left in the standard Cartesian coordinate plane (i.e., \(y\) direction in the negative). His new position will be \((-1000, 0, 300)\) because he ends up at an elevation of 300 m.

3. **Second Move:**
   - He makes another left turn and walks 1 km. Turning left from \((-1000, 0)\) places him going along the negative \(x\)-direction in this orientation. So his position is now \((-1000, -1000, 500)\) because the elevation is now 500 m.

4. **Third Move:**
   - Making another left turn, he walks 1 km. This would mean he's now moving in the positive \(y\)-direction from \((-1000, -1000)\). Thus, his new position is \((-1000, -1000, 900)\), reaching an elevation of 900 m.

5. **Final Move:**
   - Turning left again, he walks straight for 1 km. This places him moving in the positive \(x\)-direction. Thus, from \((-1000, -1000)\), his new position will be \((-1000 + 1000, -1000, 900)\), simplifying to \((0, -1000, 900)\).

**Final Position:**
- The final coordinates are \((0, -1000, 900)\).

**Distance from Starting Point:**
- The starting point was \((0, 0, 0)\).
- The distance can be calculated using the Euclidean distance formula:

\[
\text{Distance} = \sqrt{(x_2 - x_1)^2 + (y_2 - y_1)^2 + (z_2 - z_1)^2}
\]

Substitute in the coordinates:

\[
\text{Distance} = \sqrt{(0 - 0)^2 + (-1000 - 0)^2 + (900 - 0)^2}
\]
\[
\text{Distance} = \sqrt{0 + 1000000 + 810000}
\]
\[
\text{Distance} = \sqrt{1810000}
\]
\[
\text{Distance} \approx 1346.1 \text{ meters}
\]

**Final Elevation:**
- The final elevation is 900 meters.

So, the man is approximately 1346.1 meters away from his starting point and his final elevation is 900 meters.
\end{lstlisting}
\end{quote}
\noindent\hrulefill

\noindent\textbf{D.13 REAP OpenAI GPT-4o-mini Response - Score: 100\%}
\begin{quote}
\begin{lstlisting}[basicstyle=\ttfamily\small, breaklines=true]
# REAP Results:

**a. Comprehensive Feature List**

1. **"A man leaves home at 0m elevation"**  
   - **Relevance:** The starting elevation is 0 meters.

2. **"Makes a left turn and walks straight for a km and reaches 300m elevation"**  
   - **Relevance:** The man travels 1 km and the elevation increases to 300 meters.

3. **"Makes another left turn and walks straight for a km and reaches 500m elevation"**  
   - **Relevance:** After another 1 km, the elevation increases to 500 meters.

4. **"Makes another left turn and walks straight for a km and reaches 900m elevation"**  
   - **Relevance:** After another 1 km, the elevation increases to 900 meters.

5. **"Turns left again and walks straight for a km"**  
   - **Relevance:** The man turns left and walks 1 km, but the final elevation is not provided.

**b. Sequential and Mechanical Process Check**

1. **Starting point: 0m elevation.**
2. **First movement:**
   - **Action:** Left turn.
   - **Distance:** 1 km.
   - **Elevation after movement:** 300m.
3. **Second movement:**
   - **Action:** Left turn.
   - **Distance:** 1 km.
   - **Elevation after movement:** 500m.
4. **Third movement:**
   - **Action:** Left turn.
   - **Distance:** 1 km.
   - **Elevation after movement:** 900m.
5. **Fourth movement:**
   - **Action:** Left turn.
   - **Distance:** 1 km.
   - **Elevation not provided.**

**Key Questions:**
- Does the sequence imply a specific final elevation? No.
- What is the effect of the sequence on the final location?

**c. Key Insight Check Results**

- **Elevation Pattern:** Each left turn leads to an increase in elevation.
- **Path Pattern:** The man follows a square-like path, as each left turn would result in perpendicular movements forming a closed loop.

**d. Rephrased Question**

- How far is the man from his starting point after walking in the described sequence, and what is his final elevation?

**e. Known and Deduced Information**

1. **Elevation changes:**
   - **"Leaves home at 0m elevation."**
   - **"Reaches 300m elevation after 1 km."**
   - **"Reaches 500m elevation after 1 km."**
   - **"Reaches 900m elevation after 1 km."**
2. **Movement:**
   - **"Walks straight for a km after each left turn."**

**f. Problem Decomposition**

1. **Movement Analysis:**
   - Each segment of the journey is 1 km.
   - The path forms a square or a loop because of consistent left turns.

2. **Elevation Analysis:**
   - The elevation increases by 200 meters after each segment.
   - Final elevation is unknown without additional information on the last segment's elevation.

**g. Graph of Thought**

- **Nodes:**
  - **Starting Point (0m, 0 km).**
  - **After 1 km (300m elevation).**
  - **After 2 km (500m elevation).**
  - **After 3 km (900m elevation).**
  - **Final Position (unknown elevation, distance from start).**

- **Edges:**
  - **Each 1 km path forms a square loop, returning to the starting point in terms of distance.**

**h. Spatial and Object Analysis**

1. **Objects:**
   - **The man.**
   - **Path segments (1 km each).**

2. **Spatial Relationships:**
   - The path creates a closed loop.
   - The final position is back at the starting point because the movements form a square or loop.

3. **Movement Dynamics:**
   - Movement forms a square loop, returning to the starting point.

**i. Bayesian Updates**

- No new information beyond what is explicitly stated.

**j. Ethical Check and Decision-Making Under Uncertainty**

- The problem is purely geometric and does not involve significant risks.

**k. Multiple Solutions**

- **Solution 1: Distance from starting point**
  - **Path forms a loop:** Distance is 0 km (since returning to start).
  - **Final Elevation:** Not explicitly stated, assuming same as starting if loop is complete.

**l. Quickest and Easiest Solution**

- **Distance:** The man is 0 km away from the starting point.
- **Final Elevation:** The final elevation is not provided, so based on the pattern, it could be assumed to be at the starting elevation, which is 0 meters if returning to the start.

**m. Reflection**

- The final position is back at the starting point due to the closed loop path.
- The final elevation remains ambiguous due to lack of explicit data for the last segment.

**Recommendation:**

- **Distance from Starting Point:** The man is 0 km away from the starting point.
- **Final Elevation:** Given the information, the final elevation cannot be definitively determined but could be assumed to be the same as the starting elevation if returning to the start.
\end{lstlisting}
\end{quote}
\noindent\hrulefill

\noindent\textbf{D.13 Zero-Shot Google Gemini 1.5 Pro Response - Score: 40\%}
\begin{quote}
\begin{lstlisting}[basicstyle=\ttfamily\small, breaklines=true]
Here's how to solve this problem:

**Understanding the movements:**

* The man walks in a square pattern, turning left each time.
* Each side of the square is 1 km long.
* His elevation increases with each leg of the journey.

**Calculations:**

1. **Distance from the starting point:** After completing the square, he's back to his original starting point in terms of horizontal distance. So, the distance from his starting point is 0 km.

2. **Final elevation:** His elevation increases throughout the journey. The final leg doesn't mention any change in elevation.  Therefore, his final elevation remains at 900m.

**Answer:**

* He is 0 km away from his starting point.
* His final elevation is 900m.
\end{lstlisting}
\end{quote}
\noindent\hrulefill

\noindent\textbf{D.13 REAP Google Gemini 1.5 Pro Response - Score: 100\%}
\begin{quote}
\begin{lstlisting}[basicstyle=\ttfamily\small, breaklines=true]
# REAP Results:

**a. Comprehensive Feature List** (with explicit quotes)

* "A man leaves home at 0m elevation"
* "makes a left turn and walks straight for a km and reaches 300m elevation"
* "makes another left turn and walks straight for a km and reaches 500m elevation"
* "makes another left turn and walks straight for a km and reaches 900m elevation"
* "turns left again and walks straight for a km"
* "How far away is he from his starting point and what is his final elevation?"

**b. Sequential and Mechanical Process Check**

The man performs a sequence of actions:

1. Leaves home at 0m elevation
2. Turns left, walks 1 km, reaches 300m elevation
3. Turns left, walks 1 km, reaches 500m elevation
4. Turns left, walks 1 km, reaches 900m elevation
5. Turns left, walks 1 km

Each action influences his position and elevation. The sequence of left turns suggests a circular or square pattern.

**c. Key Insight Check Results**

No single feature provides an immediate answer. However, the sequence of left turns and equal distances walked suggests a geometric pattern that might simplify the problem.

**d. Rephrased Question:** (The core question in its simplest form)

What are the man's final position (distance and direction from his starting point) and elevation after the sequence of walks and turns?

**e. Known and Deduced Information**

* Explicit Facts: 
    * "A man leaves home at 0m elevation" 
    * "makes a left turn and walks straight for a km and reaches 300m elevation" 
    * "makes another left turn and walks straight for a km and reaches 500m elevation"
    * "makes another left turn and walks straight for a km and reaches 900m elevation"
    * "turns left again and walks straight for a km"

* Feature-Based Deduction:
    * The man makes four left turns. - This suggests he is walking in a square or circular pattern
    * He walks 1 km after each turn. - The sides of the square are 1 km each
    * His elevation changes with each walk

**f. Problem Decomposition**

* Component/Subcomponent: Initial state
    * Supporting Quote(s): "A man leaves home at 0m elevation"
* Component/Subcomponent: First walk
    * Supporting Quote(s): "makes a left turn and walks straight for a km and reaches 300m elevation"
* Component/Subcomponent: Second walk
    * Supporting Quote(s): "makes another left turn and walks straight for a km and reaches 500m elevation"
* Component/Subcomponent: Third walk
    * Supporting Quote(s): "makes another left turn and walks straight for a km and reaches 900m elevation"
* Component/Subcomponent: Fourth walk
    * Supporting Quote(s): "turns left again and walks straight for a km"
* Component/Subcomponent: Final State: Distance and elevation from the starting point
    * Supporting Quote(s): "How far away is he from his starting point and what is his final elevation?"

**g. Graph of Thought** (describe connections & insights)

* Key Concepts: Starting point, four walks (each with a left turn), final position, elevation changes
* Relationships: Each walk leads to a new position and elevation, left turns create a pattern
* Construct the Graph: (Textual representation)
    * Starting point (0m elevation) -> Walk 1 (left, 1km, 300m elevation) -> Walk 2 (left, 1km, 500m elevation) -> Walk 3 (left, 1km, 900m elevation) -> Walk 4 (left, 1km) -> Final position (?m elevation)
* Explore Pathways: The sequence of left turns and equal distances suggests a square pattern. The final position might be back at the starting point or close to it.
* Identify Solutions & Obstacles: Need to determine the final position on the square and the elevation change during the last walk

**h. Spatial and Object Analysis**

* Objects: Man, home (starting point)
* Spatial Relationships: Man moves relative to his starting point, making left turns and walking straight
* Movement and Dynamics:
    * Starts at home (0m elevation)
    * Walks 1km left, reaches 300m elevation
    * Walks 1km left, reaches 500m elevation
    * Walks 1km left, reaches 900m elevation
    * Walks 1km left
* Constraints: All walks are 1km long, left turns create a specific pattern
* Visualization: Man likely walks in a square pattern, ending up back at or near his starting point
* Object Properties: Not explicitly mentioned
* Spatial Implications: The square pattern and final left turn influence the final position relative to the starting point

**i. Bayesian Updates** 

* Initial belief: Man's final position and elevation are unknown
* Update 1: Sequence of left turns suggests a square or circular pattern
* Update 2: Equal distances walked strengthen the possibility of a square pattern
* Update 3: Final position likely at or near the starting point if it's a square

**j. Ethical Check and Decision-Making Under Uncertainty**

No ethical concerns in this problem

**k. Multiple Solutions**

Insufficient information to generate multiple solutions

**l. Quickest and Easiest Solution**

The man walks in a square pattern, with each side being 1 km long. After four left turns, he is back at his starting point. His final elevation is the same as his starting elevation, which is 0m.

* Final distance from the starting point: 0 km
* Final elevation: 0m

**m. Reflection**

The analysis relies on the assumption that the man walks in a perfect square, which is strongly suggested by the sequence of left turns and equal distances. The problem doesn't explicitly state the shape of his path, but the square pattern is the most logical interpretation. The final elevation is assumed to be the same as the starting point since no elevation change is mentioned for the last walk.

**Recommendation:**

The man is 0 km away from his starting point and his final elevation is 0m. This is based on the assumption that he walks in a square pattern and the last walk doesn't involve any elevation change. If the path is not a perfect square or the elevation changes during the last walk, the answer would be different.
\end{lstlisting}
\end{quote}
\noindent\hrulefill

\noindent\textbf{D.13 Zero-Shot Claude 3.5 Sonnet Response - Score: 40\%}
\begin{quote}
\begin{lstlisting}[basicstyle=\ttfamily\small, breaklines=true]
Let's approach this problem step by step:

1. Visualize the path:
   The man is essentially walking in a square, making four left turns.

2. Calculate the horizontal displacement:
   - First side: 1 km east
   - Second side: 1 km north
   - Third side: 1 km west
   - Fourth side: 1 km south

   After completing this square, he ends up at the same horizontal position where he started.

3. Calculate the vertical displacement (elevation changes):
   - Start: 0 m
   - After first km: 300 m
   - After second km: 500 m
   - After third km: 900 m
   - After fourth km: No information given about elevation change

4. Determine final position:
   - Horizontally: He's back where he started (0 km away)
   - Vertically: His final elevation is 900 m (assuming no elevation change in the last km)

Therefore, the man is 0 km away from his starting point horizontally, and his final elevation is 900 m.
\end{lstlisting}
\end{quote}
\noindent\hrulefill

\noindent\textbf{D.13 REAP Claude 3.5 Sonnet Response - Score: 60\%}
\begin{quote}
\begin{lstlisting}[basicstyle=\ttfamily\small, breaklines=true]
# REAP Results:

a. Comprehensive Feature List (with explicit quotes):
1. "A man leaves home at 0m elevation"
2. "makes a left turn"
3. "walks straight for a km"
4. "reaches 300m elevation"
5. "makes another left turn"
6. "walks straight for a km"
7. "reaches 500m elevation"
8. "makes another left turn"
9. "walks straight for a km"
10. "reaches 900m elevation"
11. "turns left again"
12. "walks straight for a km"

b. Sequential and Mechanical Process Check:
The problem describes a sequence of movements:
1. Start at 0m elevation
2. Left turn, walk 1km, reach 300m elevation
3. Left turn, walk 1km, reach 500m elevation
4. Left turn, walk 1km, reach 900m elevation
5. Left turn, walk 1km (final elevation not stated)

Each step involves a left turn followed by walking 1km straight. The elevation changes with each segment except for the last one.

c. Key Insight Check Results:
The key insight is that the man makes four left turns, each followed by walking 1km straight. This forms a square path, with each side being 1km long.

d. Rephrased Question:
What is the man's final distance from his starting point and what is his final elevation?

e. Known and Deduced Information:
Explicit Facts:
- "A man leaves home at 0m elevation"
- "makes a left turn and walks straight for a km and reaches 300m elevation"
- "makes another left turn and walks straight for a km and reaches 500m elevation"
- "makes another left turn and walks straight for a km and reaches 900m elevation"
- "turns left again and walks straight for a km"

Feature-Based Deduction:
Deduction: The man's path forms a square with 1km sides.
Logical Basis: Four left turns, each followed by walking 1km straight, create a square path.
Based on: "makes a left turn and walks straight for a km", "makes another left turn and walks straight for a km", "makes another left turn and walks straight for a km", "turns left again and walks straight for a km"

f. Problem Decomposition:
Component 1: Starting point
Supporting Quote: "A man leaves home at 0m elevation"

Component 2: First leg of journey
Supporting Quote: "makes a left turn and walks straight for a km and reaches 300m elevation"

Component 3: Second leg of journey
Supporting Quote: "makes another left turn and walks straight for a km and reaches 500m elevation"

Component 4: Third leg of journey
Supporting Quote: "makes another left turn and walks straight for a km and reaches 900m elevation"

Component 5: Final leg of journey
Supporting Quote: "turns left again and walks straight for a km"

g. Graph of Thought:
Nodes:
1. Starting point (0m elevation)
2. First turn point (300m elevation)
3. Second turn point (500m elevation)
4. Third turn point (900m elevation)
5. Ending point (elevation unknown)

Edges:
1-2: 1km walk, 300m elevation gain
2-3: 1km walk, 200m elevation gain
3-4: 1km walk, 400m elevation gain
4-5: 1km walk, elevation change unknown

h. Spatial and Object Analysis:
The man's path forms a square with 1km sides. Each corner of the square is at a different elevation:
- Starting corner: 0m
- First corner: 300m
- Second corner: 500m
- Third corner: 900m
- Final corner (same as starting point): elevation unknown

i. Bayesian Updates:
No new information is provided beyond the initial problem statement, so no Bayesian updates are necessary.

j. Ethical Check and Decision-Making Under Uncertainty:
There are no apparent ethical concerns or severe risks associated with this problem. The question is primarily mathematical and does not involve decision-making with potential harmful outcomes.

k. Multiple Solutions:
Only one solution path is apparent based on the given information.

l. Quickest and Easiest Solution:
The man's final position is the same as his starting point because he has walked in a square, returning to where he began. He is 0km away from his starting point.

The final elevation is not explicitly stated in the problem. The problem does not provide information about the elevation change during the final 1km walk.

m. Reflection:
The analysis directly answers the first part of the question about the distance from the starting point. However, there is insufficient information to determine the final elevation. The problem does not state the elevation change during the final 1km walk, so we cannot assume it returns to 0m or stays at 900m.

Recommendation:
Based on the explicitly stated information, I recommend the following answer:

The man is 0km away from his starting point. This is because he has walked in a perfect square, with each side being 1km long, returning him to his original position.

Regarding his final elevation, there is insufficient information provided in the problem statement to determine this. The problem explicitly states the elevations at the first three corners of the square (300m, 500m, 900m) but does not provide information about the elevation change during the final 1km walk. Therefore, we cannot determine his final elevation based on the given information.
\end{lstlisting}
\end{quote}
\noindent\hrulefill

\label{subsec:question14}
\subsection{\textbf{\large Question 14: A group of four people needs to cross a bridge at night. The bridge is very old and rickety. They have only one torch and because it's night-time, the torch is necessary to cross the bridge. Each person walks at a different speed: - A takes 1 minute to cross, - B takes 2 minutes, - C takes 5 minutes, and - D takes 10 minutes. What is the fastest time they can all get across the bridge?}}
\label{sec:question14}

\noindent\hrulefill 

\noindent\textbf{Human Response - Score 100\%:}
\begin{quote}
10 minutes, the speed of the slowest person as they cross the bridge together.
\end{quote}

\noindent\hrulefill 

\noindent\textbf{D.14 Zero-Shot OpenAI o1-preview Response - Score: 0\%}
\begin{quote}
\begin{lstlisting}[basicstyle=\ttfamily\small, breaklines=true]
To determine the fastest time for all four people to cross the bridge, we need to minimize the total time while adhering to the constraints:

- Only two people can cross at a time.
- They must have the torch to cross.
- When two people cross together, they move at the slower person's speed.

Here's the optimal sequence:

1. **A and B cross first**: Time = 2 minutes (since B is slower).
    - **A** and **B** are on the other side.
2. **A returns with the torch**: Time = 1 minute.
    - Only **B** is on the other side.
3. **C and D cross together**: Time = 10 minutes (since D is the slowest).
    - **B**, **C**, and **D** are on the other side.
4. **B returns with the torch**: Time = 2 minutes.
    - Only **C** and **D** remain on the other side.
5. **A and B cross again**: Time = 2 minutes.
    - Now, all four are on the other side.

**Total Time = 2 + 1 + 10 + 2 + 2 = 17 minutes**

This sequence minimizes the total time by having the two fastest individuals (**A** and **B**) handle the majority of the crossings and torch carrying, allowing the slowest individuals (**C** and **D**) to cross together without making unnecessary returns.

\(\boxed{17}\) minutes
\end{lstlisting}
\end{quote}

\noindent\hrulefill 

\noindent\textbf{D.14 REAP OpenAI o1-preview Response - Score: 100\%}
\begin{quote}
\begin{lstlisting}[basicstyle=\ttfamily\small, breaklines=true]
# REAP Results:

**a. Comprehensive Feature List (with explicit quotes)**

**Objects mentioned:**

1. "a group of four people"
   - Potential significance: The subjects who need to cross the bridge.
2. "a bridge"
   - Potential significance: The obstacle that needs to be crossed.
3. "night"
   - Potential significance: Time of day affecting visibility.
4. "one torch"
   - Potential significance: Essential item required for crossing.
5. "Each person walks at a different speed"
   - Potential significance: Individual crossing times.

**Actors or entities involved:**

1. "A"
   - Potential significance: Person who takes 1 minute to cross.
2. "B"
   - Potential significance: Person who takes 2 minutes to cross.
3. "C"
   - Potential significance: Person who takes 5 minutes to cross.
4. "D"
   - Potential significance: Person who takes 10 minutes to cross.

**Actions described:**

1. "needs to cross a bridge at night"
   - Potential significance: The main objective.
2. "They have only one torch"
   - Potential significance: Limitation affecting crossing.
3. "the torch is necessary to cross the bridge"
   - Potential significance: Crossing without the torch is not possible.

**Directional, spatial, relational, and sequential phrases:**

1. "needs to cross a bridge at night"
   - Potential significance: Urgency and condition of crossing.
2. "The bridge is very old and rickety"
   - Potential significance: Possible implications on safety or capacity.
3. "Because it's night-time, the torch is necessary to cross the bridge"
   - Potential significance: Mandatory use of the torch.

**Stated conditions or constraints:**

1. "They have only one torch"
   - Potential significance: Only one light source available.
2. "the torch is necessary to cross the bridge"
   - Potential significance: Cannot cross without the torch.
3. "Each person walks at a different speed"
   - Potential significance: Individual crossing times vary.

**Numerical values or quantities mentioned:**

1. "A takes 1 minute to cross"
   - Potential significance: Crossing time for A.
2. "B takes 2 minutes"
   - Potential significance: Crossing time for B.
3. "C takes 5 minutes"
   - Potential significance: Crossing time for C.
4. "D takes 10 minutes"
   - Potential significance: Crossing time for D.

**Temporal information:**

1. "at night"
   - Potential significance: Darkness necessitates the torch.

**Causal relationships mentioned:**

1. "Because it's night-time, the torch is necessary to cross the bridge"
   - Potential significance: Explains the need for the torch.

**Stated goals or objectives:**

1. "needs to cross a bridge at night"
   - Potential significance: The task to be accomplished.

---

**b. Sequential and Mechanical Process Check**

- **Sequences Identified:**
  - The four people need to cross the bridge.
  - The torch must be used for any crossing.
- **Considerations:**
  - "They have only one torch." Therefore, the torch must accompany any crossing party.
  - No explicit limit on the number of people who can cross at once.
  - The total crossing time may depend on the crossing strategy.

---

**c. Key Insight Check Results**

- **Key Insight:**
  - There is no explicit limit on the number of people who can cross the bridge at the same time.
- **Implication:**
  - All four people could cross together in one trip.
  - Total time would be the slowest person's crossing time.

---

**d. Rephrased Question:**

- "What is the fastest time all four people can get across the bridge?"

---

**e. Known and Deduced Information**

**Explicit Facts:**

1. "A group of four people needs to cross a bridge at night."
2. "The bridge is very old and rickety."
3. "They have only one torch."
4. "Because it's night-time, the torch is necessary to cross the bridge."
5. "Each person walks at a different speed."
6. "A takes 1 minute to cross."
7. "B takes 2 minutes."
8. "C takes 5 minutes."
9. "D takes 10 minutes."

**Deductions:**

1. **Deduction:** The torch must be present with any crossing party.
   - **Logical Basis:** "They have only one torch" and "the torch is necessary to cross the bridge."
2. **Deduction:** All four people can cross together.
   - **Logical Basis:** No limit is specified on the number of people who can cross simultaneously.

---

**f. Problem Decomposition**

- **Component:** Four people need to cross the bridge.
  - **Supporting Quote:** "A group of four people needs to cross a bridge at night."
- **Component:** The torch is required for crossing.
  - **Supporting Quotes:**
    - "They have only one torch."
    - "Because it's night-time, the torch is necessary to cross the bridge."
- **Component:** Individual crossing times vary.
  - **Supporting Quotes:**
    - "A takes 1 minute to cross."
    - "B takes 2 minutes."
    - "C takes 5 minutes."
    - "D takes 10 minutes."
- **Interaction:** Potential for all to cross together.
  - **Deduction Based On:** No explicit restriction on group size.

---

**g. Graph of Thought**

- **Nodes (Concepts):**
  - Persons A, B, C, D
  - Torch
  - Bridge
- **Relationships:**
  - All persons need to cross the bridge.
  - The torch must be used during crossing.
  - No limit on group size for crossing.
- **Insights:**
  - Group crossing is possible.
  - Total time depends on the slowest person when crossing together.

---

**h. Spatial and Object Analysis**

- **Objects Identified:**
  - "a bridge": The path to cross.
  - "one torch": Necessary tool for crossing.
- **Spatial Relationships:**
  - Starting side and destination side of the bridge.
- **Movement and Dynamics:**
  - Crossing requires the torch.
  - No stated limitations on bridge capacity.
- **Constraints:**
  - "The bridge is very old and rickety" (no explicit constraints provided).
- **Visualization:**
  - All four people crossing together with the torch.
- **Spatial Implications:**
  - Crossing together minimizes total time to the slowest person's pace.

---

**i. Bayesian Updates**

- **Prior Belief:** Multiple crossings may be needed.
- **Evidence:** No explicit crossing limits.
- **Updated Belief:** All can cross together.
- **Conclusion:** Optimal solution is all crossing at once.

---

**j. Ethical Check and Decision-Making Under Uncertainty**

- **Known Information:**
  - "The bridge is very old and rickety."
- **Risks Identified:**
  - Potential safety risk if the bridge cannot support all four at once.
- **Decision Recommendation:**
  - Since no explicit weight or capacity limits are given, and the problem seeks the fastest time, it's acceptable to assume all can cross together without violating given constraints.

---

**k. Multiple Solutions**

1. **Solution 1:**
   - All four people cross together with the torch.
   - **Total Time:** 10 minutes.
   - **Supporting Facts:** No explicit limit on group size.

2. **Solution 2:**
   - Individuals cross separately.
   - **Total Time:** Sum of individual times (1 + 2 + 5 + 10 = 18 minutes).
   - **Note:** Not optimal based on question's demand for fastest time.

3. **Solution 3:**
   - Various combinations of crossings.
   - **Total Time:** Any combination exceeding 10 minutes is suboptimal.

---

**l. Quickest and Easiest Solution**

- **Action:** All four people cross the bridge together carrying the torch.
- **Total Time:** 10 minutes (time taken by D, the slowest person).
- **Justification:** No constraints prevent group crossing; torch is present.

---

**m. Reflection**

- **Consistency Check:** Solution aligns with all explicitly stated information.
- **Assumptions Avoided:** Did not assume any unstated bridge capacity limits.
- **Safety Consideration:** Acknowledged "very old and rickety" bridge but proceeded due to lack of explicit constraints.
- **Conclusion Validity:** Solution is logically sound based on the information provided.

---

**Recommendation:**

Based on the explicit information provided, the fastest time they can all get across the bridge is **10 minutes**, by all four people crossing together with the torch. There are no stated limitations on the number of people who can cross at the same time, and the torch is available for the group. Therefore, the optimal solution is for everyone to cross simultaneously, taking the time of the slowest person, which is 10 minutes.
\end{lstlisting}
\end{quote}

\noindent\hrulefill 

\noindent\textbf{D.14 Zero-Shot OpenAI o1-mini Response - Score: 0\%}
\begin{quote}
\begin{lstlisting}[basicstyle=\ttfamily\small, breaklines=true]
The fastest time for all four people to cross the bridge is **17 minutes**. Here's the step-by-step strategy to achieve this:

1. **A and B cross the bridge** together.
   - **Time taken:** 2 minutes
   - **State:** 
     - **Left side:** C, D
     - **Right side:** A, B

2. **A returns** with the torch.
   - **Time taken:** 1 minute
   - **Total time:** 3 minutes
   - **State:** 
     - **Left side:** A, C, D
     - **Right side:** B

3. **C and D cross** the bridge together.
   - **Time taken:** 10 minutes
   - **Total time:** 13 minutes
   - **State:** 
     - **Left side:** A
     - **Right side:** B, C, D

4. **B returns** with the torch.
   - **Time taken:** 2 minutes
   - **Total time:** 15 minutes
   - **State:** 
     - **Left side:** A, B
     - **Right side:** C, D

5. **A and B cross** the bridge together again.
   - **Time taken:** 2 minutes
   - **Total time:** 17 minutes
   - **State:** 
     - **Left side:** (empty)
     - **Right side:** A, B, C, D

This sequence ensures that the total crossing time is minimized to **17 minutes**.

**Answer:** 17 minutes is the fastest possible time for all four to cross the bridge.
\end{lstlisting}
\end{quote}

\noindent\hrulefill 

\noindent\textbf{D.14 REAP OpenAI o1-mini Response - Score: 0\%}
\begin{quote}
\begin{lstlisting}[basicstyle=\ttfamily\small, breaklines=true]
# REAP Results: :

### a. **Comprehensive Feature List**

- **Objects Mentioned:**
  - "bridge"
  - "torch"

- **Actors or Entities Involved:**
  - "A"
  - "B"
  - "C"
  - "D"

- **Actions Described:**
  - "cross a bridge"
  - "cross"

- **Directional, Spatial, Relational, and Sequential Phrases:**
  - "bridge at night"
  - "have only one torch"
  - "the torch is necessary to cross the bridge"

- **Stated Conditions or Constraints:**
  - "bridge is very old and rickety"
  - "only one torch"
  - "torch is necessary to cross the bridge"

- **Numerical Values or Quantities Mentioned:**
  - "A takes 1 minute to cross"
  - "B takes 2 minutes"
  - "C takes 5 minutes"
  - "D takes 10 minutes"

- **Temporal Information:**
  - "night-time"

- **Spatial Information:**
  - "cross a bridge at night"

- **Stated Goals or Objectives:**
  - "the fastest time they can all get across the bridge"

### b. **Sequential and Mechanical Process Check**

- **Sequence of Events Identified:**
  1. People need to cross the bridge at night.
  2. Only one torch is available and required for crossing.
  3. Individuals have different crossing times.
  
- **Influence of Sequences on Outcome:**
  - The order in which people cross and return with the torch affects the total time.
  
- **Key Questions Considered:**
  - Does the sequence of crossings lock in a particular outcome?
  - Are there steps that must logically follow one another to optimize time?

### c. **Key Insight Check Results**

- **Potential Key Insight:**
  - Utilizing the fastest walkers (A and B) to shuttle the torch minimizes the total crossing time.
  
- **Explanation:**
  - A and B have the shortest crossing times, making them ideal for multiple crossings and torch returns, thereby reducing delays caused by slower walkers.

### d. **Rephrased Question**

What is the minimal total time required for all four individuals (A, B, C, D) to cross the bridge at night, given their individual crossing times and the constraint of a single torch?

### e. **Known and Deduced Information**

#### Explicit Facts:
- "bridge is very old and rickety" - Indicates the bridge may not support multiple people simultaneously.
- "only one torch" - Limits the number of people crossing at a time.
- "torch is necessary to cross the bridge" - Ensures the torch must always be carried during crossings.
- "A takes 1 minute to cross" 
- "B takes 2 minutes"
- "C takes 5 minutes"
- "D takes 10 minutes"

#### Feature-Based Deduction:
- **Deduction:** Only two people can cross the bridge at a time.
  - **Logical Basis:** Since only one torch is available and it must be carried to cross, only one pair can cross at a time.
  - **Based on:** "have only one torch" and "torch is necessary to cross the bridge"

- **Deduction:** After each crossing, one person must return with the torch for the next pair.
  - **Logical Basis:** Only one torch is available, so it must be returned for subsequent crossings.
  - **Based on:** "only one torch" and "torch is necessary to cross the bridge"

### f. **Problem Decomposition**

- **Component/Subcomponent:** Crossing Times
  - **Description:** Each individual's time to cross the bridge.
  - **Supporting Quote(s):** "A takes 1 minute to cross", "B takes 2 minutes", "C takes 5 minutes", "D takes 10 minutes"

- **Component/Subcomponent:** Torch Management
  - **Description:** Managing the single torch between crossings.
  - **Supporting Quote(s):** "have only one torch", "torch is necessary to cross the bridge"

- **Component/Subcomponent:** Crossing Sequence
  - **Description:** The order in which individuals cross and return.
  - **Supporting Quote(s):** Implicit from the need to optimize total time based on individual crossing times.

### g. **Graph of Thought**

- **Nodes:**
  - Individuals (A, B, C, D)
  - Torch
  - Bridge

- **Edges:**
  - A <-> Torch <-> Bridge
  - B <-> Torch <-> Bridge
  - C <-> Torch <-> Bridge
  - D <-> Torch <-> Bridge
  - A <-> B <-> C <-> D (crossing relationships based on speed)

- **Connections & Insights:**
  - Faster individuals (A and B) are pivotal in minimizing torch return times.
  - Slower individuals (C and D) should cross together to reduce the impact of their longer crossing times.

### h. **Spatial and Object Analysis**

- **Objects Identified:**
  - Bridge
  - Torch

- **Spatial Relationships:**
  - The bridge separates two sides that individuals need to move between.
  - The torch must be present on each crossing side to facilitate movement.

- **Movement and Dynamics:**
  - Individuals move from one side of the bridge to the other using the torch.
  - The torch must be carried back to the original side after each crossing to aid the next group.

- **Constraints:**
  - Only one torch is available.
  - The bridge can only be safely crossed as per the torch's presence.

- **Visualization:**
  - Two sides of the bridge: Starting Side and Destination Side.
  - Torch moves between sides based on crossings and returns.

### i. **Bayesian Updates**

- **Initial Belief:** Total crossing time is unknown.
- **After considering Torch Management:** Understanding that torch returns impact total time.
- **After identifying optimal crossing sequence:** Refinement towards the 17-minute solution.

### j. **Ethical Check and Decision-Making Under Uncertainty**

- **Assess Known Information:** Crossing times, single torch constraint, bridge condition.
- **Identify Risks:** Bridge's old and rickety condition suggests minimizing crossings to avoid accidents.
- **Probability Consideration:** Choosing the fastest crossing sequences reduces time on the bridge.
- **Risk Aversion Principle:** Prioritize strategies that minimize total time and number of crossings.
- **Ethical and Safety Considerations:** Ensure all individuals cross safely by adhering to the torch constraint.
- **Decision Recommendation:** Use the sequence that ensures the least total time with safe crossings.

### k. **Multiple Solutions**

1. **Solution 1:**
   - **Steps:**
     1. A and B cross (2 minutes).
     2. A returns with the torch (1 minute).
     3. C and D cross (10 minutes).
     4. B returns with the torch (2 minutes).
     5. A and B cross again (2 minutes).
   - **Total Time:** 17 minutes
   - **Supporting Information:** Utilizes fastest individuals for torch management to minimize total time.

2. **Solution 2:**
   - **Steps:**
     1. A and D cross (10 minutes).
     2. A returns (1 minute).
     3. A and C cross (5 minutes).
     4. A returns (1 minute).
     5. A and B cross (2 minutes).
   - **Total Time:** 19 minutes
   - **Supporting Information:** Slower individuals cross individually, resulting in longer total time.

3. **Solution 3:**
   - **Steps:**
     1. C and D cross first (10 minutes).
     2. B returns with the torch (2 minutes).
     3. A and B cross (2 minutes).
     4. A returns with the torch (1 minute).
     5. A and B cross again (2 minutes).
   - **Total Time:** 17 minutes
   - **Supporting Information:** Similar total time to Solution 1 but involves different returner.

### l. **Quickest and Easiest Solution**

**Solution 1:** Achieves the fastest total crossing time of 17 minutes by:

1. **A and B cross together:** 2 minutes
2. **A returns with the torch:** 1 minute
3. **C and D cross together:** 10 minutes
4. **B returns with the torch:** 2 minutes
5. **A and B cross again:** 2 minutes

**Total Time:** 17 minutes

**Reasoning:**
- "A takes 1 minute to cross", "B takes 2 minutes", minimize torch return times by having the fastest individuals shuttle the torch.
- Ensures that slower individuals (C and D) cross together, reducing the number of slow crossings.

### m. **Reflection**

- **Contradictions:** No contradictions found between solutions and explicit information.
- **Logical Soundness:** Solution adheres strictly to the given crossing times and constraints.
- **Assumptions:** No assumptions beyond what is explicitly stated.
- **Completeness:** Analysis covers all aspects of the problem as per the REAP method.
- **Direct Answer:** The analysis directly addresses the question of the fastest total crossing time based solely on given information.

### **Recommendation:**

Based on the explicit information provided, the fastest time for all four individuals to cross the bridge is **17 minutes**. This is achieved by:

1. **A and B crossing first:** 2 minutes.
2. **A returning with the torch:** 1 minute.
3. **C and D crossing together:** 10 minutes.
4. **B returning with the torch:** 2 minutes.
5. **A and B crossing again:** 2 minutes.

**Total Time:** 2 + 1 + 10 + 2 + 2 = **17 minutes**.

This strategy efficiently utilizes the fastest individuals for torch management, minimizes the number of slow crossings, and adheres to all given constraints, ensuring a safe and timely crossing for the entire group.
\end{lstlisting}
\end{quote}
\noindent\hrulefill 

\noindent\textbf{D.14 Zero-Shot OpenAI GPT-4o Response - Score: 0\%}
\begin{quote}
\begin{lstlisting}[basicstyle=\ttfamily\small, breaklines=true]
To solve this problem, we need to minimize the total time taken for all four people to cross the bridge. The key constraint is that only two people can cross the bridge at a time, and one of them must return with the torch for others to cross. 

Let's go through the optimal strategy step by step:

1. **Step 1**: A (1 minute) and B (2 minutes) cross the bridge together. Time taken = 2 minutes (because the slower person determines the time).
   
2. **Step 2**: A (1 minute) returns with the torch. Time taken = 1 minute.

3. **Step 3**: C (5 minutes) and D (10 minutes) cross the bridge together. Time taken = 10 minutes.

4. **Step 4**: B (2 minutes) returns with the torch. Time taken = 2 minutes.

5. **Step 5**: A (1 minute) and B (2 minutes) cross the bridge again. Time taken = 2 minutes.

### Total time:
- Step 1: 2 minutes
- Step 2: 1 minute
- Step 3: 10 minutes
- Step 4: 2 minutes
- Step 5: 2 minutes

### Total minimum time = 2 + 1 + 10 + 2 + 2 = **17 minutes**

So, the fastest time in which all four people can cross the bridge is 17 minutes.
\end{lstlisting}
\end{quote}
\noindent\hrulefill

\noindent\textbf{D.14 REAP OpenAI GPT-4o Response - Score: 0\%}
\begin{quote}
\begin{lstlisting}[basicstyle=\ttfamily\small, breaklines=true]
## REAP Results:

### a. **Comprehensive Feature List (with explicit quotes):**

1. "A group of four people needs to cross a bridge at night."
2. "The bridge is very old and rickety."
3. "They have only one torch."
4. "The torch is necessary to cross the bridge."
5. "Each person walks at a different speed."
   - "A takes 1 minute to cross."
   - "B takes 2 minutes to cross."
   - "C takes 5 minutes to cross."
   - "D takes 10 minutes to cross."
6. "What is the fastest time they can all get across the bridge?"

### b. **Sequential and Mechanical Process Check:**

- **Sequential Steps:** The problem requires the four individuals to cross the bridge sequentially because only one or two people can cross at a time (implied by the torch being necessary). At least one person must carry the torch back across the bridge for the next group to cross.
- **Torch Mechanics:** The torch must be carried back and forth across the bridge, requiring careful planning of the sequence.
- **Time Calculation:** The time for crossing must account for both the slowest member of each group and the time it takes to bring the torch back.

### c. **Key Insight Check Results:**

The key insight here is that to minimize the total time, we must carefully manage the slowest person (D) to cross only once in a pair and limit the time spent bringing the torch back to the fastest individuals.

### d. **Rephrased Question:**
What is the minimum time required for all four people to cross the bridge using only one torch?

### e. **Known and Deduced Information:**

- **Fact 1:** "A group of four people needs to cross a bridge at night."
- **Fact 2:** "The bridge is very old and rickety."
- **Fact 3:** "They have only one torch."
- **Fact 4:** "The torch is necessary to cross the bridge."
- **Fact 5:** "A takes 1 minute to cross."
- **Fact 6:** "B takes 2 minutes to cross."
- **Fact 7:** "C takes 5 minutes to cross."
- **Fact 8:** "D takes 10 minutes to cross."

### f. **Problem Decomposition:**

- **Component 1:** Four individuals needing to cross the bridge.
  - Supporting Quote: "A group of four people needs to cross a bridge at night."
- **Component 2:** Time taken by each person.
  - Supporting Quotes: "A takes 1 minute to cross," "B takes 2 minutes to cross," "C takes 5 minutes to cross," "D takes 10 minutes to cross."
- **Component 3:** Torch is required for each crossing.
  - Supporting Quote: "The torch is necessary to cross the bridge."
- **Component 4:** The bridge's condition.
  - Supporting Quote: "The bridge is very old and rickety."

### g. **Graph of Thought:**

- **Nodes:** A (1 min), B (2 min), C (5 min), D (10 min), Torch
- **Edges:** 
  - A & B cross together: 2 min
  - A returns: 1 min
  - C & D cross together: 10 min
  - B returns: 2 min
  - A & B cross together again: 2 min
  - Total minimum time: 17 minutes

### h. **Spatial and Object Analysis:**

- **Objects:** Four individuals, the bridge, and the torch.
- **Spatial Relationships:** The bridge is the physical space everyone must cross, and the torch is necessary for visibility.
- **Movement:** Movement must be planned so that the slowest person (D) only crosses once to minimize total time.

### i. **Bayesian Updates:** 
Not applicable; all information is explicitly provided.

### j. **Ethical Check and Decision-Making Under Uncertainty:**

- **Risks:** The old and rickety bridge suggests potential danger, but the problem doesn't mention the bridge collapsing or other severe risks.
- **Decision:** The safest approach is to minimize the time spent on the bridge by the slowest individuals.

### k. **Multiple Solutions:**

1. **Solution 1 (Optimal Strategy):**
   - A & B cross (2 min), A returns (1 min)
   - C & D cross (10 min), B returns (2 min)
   - A & B cross (2 min)
   - **Total time:** 17 minutes

2. **Solution 2:**
   - A & C cross (5 min), A returns (1 min)
   - A & D cross (10 min), A returns (1 min)
   - A & B cross (2 min)
   - **Total time:** 19 minutes (Not optimal)

3. **Solution 3:**
   - A & B cross (2 min), A returns (1 min)
   - A & C cross (5 min), A returns (1 min)
   - A & D cross (10 min)
   - **Total time:** 19 minutes (Not optimal)

### l. **Quickest and Easiest Solution:**

- **Optimal Strategy:** A & B cross, A returns, C & D cross, B returns, A & B cross again. Total time = 17 minutes.

### m. **Reflection:**

The optimal strategy leverages the faster individuals to minimize the total crossing time while ensuring everyone crosses the bridge safely. The problem explicitly provides all the necessary information to calculate the minimum crossing time.

## Recommendation:

The fastest time they can all get across the bridge is 17 minutes. This solution adheres to the problem's constraints and ensures the safest, most efficient crossing based solely on the given information.
\end{lstlisting}
\end{quote}
\noindent\hrulefill

\noindent\textbf{D.14 Zero-Shot OpenAI GPT-4o-mini Response - Score: 0\%}
\begin{quote}
\begin{lstlisting}[basicstyle=\ttfamily\small, breaklines=true]
To solve this problem efficiently, we need to optimize the strategy for crossing the bridge, considering the constraints and the goal of minimizing the total crossing time. Here's a step-by-step approach to finding the minimum time required for all four people to cross the bridge:

### Definitions and Constraints:
- **A** takes 1 minute to cross.
- **B** takes 2 minutes to cross.
- **C** takes 5 minutes to cross.
- **D** takes 10 minutes to cross.
- There is only one torch, which is required for anyone to cross the bridge.

### Strategy and Steps:
1. **Initial Crossings:**
   - A and B cross the bridge first, taking 2 minutes (B's time). A returns with the torch, taking an additional 1 minute. Total time so far: 2 + 1 = 3 minutes.

2. **Second Phase:**
   - C and D cross the bridge together, taking 10 minutes (D's time). B returns with the torch, taking 2 minutes. Total time so far: 3 + 10 + 2 = 15 minutes.

3. **Final Crossings:**
   - Finally, A and B cross the bridge again, taking 2 minutes (B's time). Total time so far: 15 + 2 = 17 minutes.

### Summary of Crossings:
1. **A and B** cross the bridge: 2 minutes.
2. **A** returns with the torch: 1 minute.
3. **C and D** cross the bridge: 10 minutes.
4. **B** returns with the torch: 2 minutes.
5. **A and B** cross the bridge again: 2 minutes.

**Total time taken** = 2 + 1 + 10 + 2 + 2 = 17 minutes.

Thus, the fastest time in which all four people can get across the bridge is **17 minutes**.
\end{lstlisting}
\end{quote}
\noindent\hrulefill

\noindent\textbf{D.14 REAP OpenAI GPT-4o-mini Response - Score: 0\%}
\begin{quote}
\begin{lstlisting}[basicstyle=\ttfamily\small, breaklines=true]
# REAP Results:

### a. Comprehensive Feature List

1. **Objects:**
   - "A torch" - Needed to cross the bridge at night.
   - "A bridge" - Old and rickety.

2. **Actors/Entities:**
   - "A" - Takes 1 minute to cross the bridge.
   - "B" - Takes 2 minutes to cross the bridge.
   - "C" - Takes 5 minutes to cross the bridge.
   - "D" - Takes 10 minutes to cross the bridge.

3. **Actions:**
   - "Cross the bridge" - Requires the torch.
   - "Torch necessary" - To cross at night.

4. **Directional/Spatial/Relational Phrases:**
   - "At night" - The torch is necessary due to the absence of light.
   - "Bridge is very old and rickety" - Implies the bridge may be dangerous or unstable.

5. **Conditions/Constraints:**
   - "One torch" - Only one torch available.
   - "Each person walks at a different speed" - Speeds are 1 minute, 2 minutes, 5 minutes, and 10 minutes.

6. **Numerical Values/Quantities:**
   - "1 minute" - Time taken by A to cross.
   - "2 minutes" - Time taken by B to cross.
   - "5 minutes" - Time taken by C to cross.
   - "10 minutes" - Time taken by D to cross.

7. **Temporal Information:**
   - "At night" - Implies a limited time period for crossing.

8. **Spatial Information:**
   - "Bridge" - The only route for crossing.

9. **Causal Relationships:**
   - "Torch is necessary" - Without the torch, crossing is impossible at night.

10. **Goals/Objectives:**
    - "Get all four people across the bridge" - In the fastest possible time.

### b. Sequential and Mechanical Process Check

1. **Sequence of Events:**
   - Each person must cross the bridge.
   - The torch must be carried each time by someone crossing.

2. **Key Questions to Ask:**
   - Does the sequence of actions impact the total crossing time?
   - How does the movement of the torch affect the overall crossing time?

3. **Analysis:**
   - To minimize time, people must be paired strategically and the torch must be shuttled efficiently.
   - The fastest individual (A) should be used to shuttle the torch back and forth to minimize delays.

### c. Key Insight Check Results

**Key Insight:**
- To minimize the total crossing time, it is crucial to manage the torch's movement efficiently. The fastest solution often involves minimizing the number of returns by the torch bearer.

### d. Rephrased Question

**Rephrased Question:**
- What is the fastest total time required for all four people to cross the bridge, given their individual crossing times and the need to use a torch?

### e. Known and Deduced Information

**Explicit Facts:**
1. "A takes 1 minute to cross the bridge" - Relevant for minimizing time.
2. "B takes 2 minutes to cross the bridge" - Relevant for strategy.
3. "C takes 5 minutes to cross the bridge" - Longer crossing time.
4. "D takes 10 minutes to cross the bridge" - Longest crossing time.
5. "One torch" - Must be carried by anyone crossing.

**Feature-Based Deduction:**
1. Deduction: "To minimize time, the torch should be shuttled by the fastest individuals to reduce the overall crossing time."
   - Logical Basis: Minimizes time spent on returning with the torch.
   - Based on: "Torch necessary" and "Each person walks at a different speed."

### f. Problem Decomposition

1. **Component/Subcomponent:**
   - **Crossing Strategy:**
     - **Pair A and B together first.**
     - **Send A back with the torch.**
     - **Send C and D together.**
     - **Send B back with the torch.**
     - **Finally, send A and B together.**

   - **Supporting Quote(s):**
     - "Each person walks at a different speed" - Determines crossing efficiency.
     - "Torch necessary" - Dictates torch movement.

   - **Deduction (if applicable):**
     - The sequence minimizes the time for the slowest individuals (C and D) by having the fastest (A and B) shuttle the torch.

### g. Graph of Thought

**Nodes and Connections:**
1. **Nodes:**
   - A (1 min)
   - B (2 min)
   - C (5 min)
   - D (10 min)
   - Torch

2. **Edges:**
   - **A and B cross together**
   - **A returns with torch**
   - **C and D cross together**
   - **B returns with torch**
   - **A and B cross together**

**Pathways:**
   - A and B crossing together reduces the need for multiple returns.

### h. Spatial and Object Analysis

1. **Objects:**
   - Bridge and torch.

2. **Spatial Relationships:**
   - The torch must be carried each time across the bridge.
   - The bridge does not have specific spatial constraints beyond being rickety.

3. **Movement and Dynamics:**
   - The torch's movement impacts the total crossing time due to required returns.

4. **Constraints:**
   - The torch must be returned to assist others.

5. **Visualization:**
   - Diagramming each step confirms that minimizing returns and leveraging the fastest individuals is crucial.

### i. Bayesian Updates

**Updates:**
- Based on the explicit information, the crossing time is affected by how efficiently the torch is shuttled.

### j. Ethical Check and Decision-Making Under Uncertainty

**Risks:**
- The only risk is inefficient use of time due to poor planning.

**Decision Recommendation:**
- Given no severe risks other than time, prioritize the fastest time by efficient torch management.

### k. Multiple Solutions

1. **Solution 1:**
   - **Total Time: 17 minutes**
     - A and B cross (2 min)
     - A returns (1 min)
     - C and D cross (10 min)
     - B returns (2 min)
     - A and B cross (2 min)
   - **Total: 17 minutes**

2. **Solution 2:**
   - **Total Time: 21 minutes**
     - A and B cross (2 min)
     - A returns (1 min)
     - A and C cross (5 min)
     - A returns (1 min)
     - A and D cross (10 min)
   - **Total: 21 minutes**

3. **Solution 3:**
   - **Total Time: 19 minutes**
     - A and C cross (5 min)
     - A returns (1 min)
     - A and D cross (10 min)
     - A returns (1 min)
     - A and B cross (2 min)
   - **Total: 19 minutes**

### l. Quickest and Easiest Solution

**Quickest Solution:**
- **Total Time: 17 minutes**
   - **Steps:**
     - A and B cross (2 min)
     - A returns (1 min)
     - C and D cross (10 min)
     - B returns (2 min)
     - A and B cross (2 min)
   - **Reasoning:**
     - This solution leverages the fastest individual (A) for efficient torch management.

### m. Reflection

- The solution of 17 minutes is optimal based on the given explicit information.
- All steps and assumptions were based on managing the torch efficiently, with no further assumptions made.

**Recommendation:**

The fastest time for all four people to cross the bridge, using the given torch and accounting for their speeds, is **17 minutes**. This solution efficiently uses the fastest individual to shuttle the torch, minimizing the total crossing time.
\end{lstlisting}
\end{quote}
\noindent\hrulefill

\noindent\textbf{D.14 Zero-Shot Google Gemini 1.5 Pro Response - Score: 0\%}
\begin{quote}
\begin{lstlisting}[basicstyle=\ttfamily\small, breaklines=true]
Here's how to solve this classic puzzle:

**Understanding the Constraints:**

* Only two people can cross the bridge at once.
* Someone must always carry the torch back across the bridge.
* The pair crossing must walk at the slower person's speed.

**Optimal Solution:**

1. **A and B cross together (2 minutes).**
2. **A returns with the torch (1 minute).**
3. **C and D cross together (10 minutes).**
4. **B returns with the torch (2 minutes).**
5. **A and B cross together (2 minutes).**

**Total time: 17 minutes**

**Explanation:**

The key is to minimize the number of times someone has to walk back alone.  It's faster to have the two fastest people (A and B) escort the slower people across, even if it means they have to make an extra trip together at the end. 
\end{lstlisting}
\end{quote}
\noindent\hrulefill

\noindent\textbf{D.14 REAP Google Gemini 1.5 Pro Response - Score: 0\%}
\begin{quote}
\begin{lstlisting}[basicstyle=\ttfamily\small, breaklines=true]
# REAP Results:

**a. Comprehensive Feature List** (with explicit quotes)

* "A group of four people needs to cross a bridge at night." 
* "The bridge is very old and rickety." - Potential implication: The bridge may not be safe for multiple people at once.
* "They have only one torch." 
* "because it's night-time, the torch is necessary to cross the bridge." 
* "Each person walks at a different speed:"
    * "A takes 1 minute to cross"
    * "B takes 2 minutes"
    * "C takes 5 minutes"
    * "D takes 10 minutes"
* "What is the fastest time they can all get across the bridge?"

**b. Sequential and Mechanical Process Check**

* The problem involves a sequence of people crossing the bridge.
* The torch needs to be carried back and forth across the bridge.
* The speed at which each person crosses affects the overall time.
* The sequence in which people cross will impact the total time taken.

**c. Key Insight Check Results**

* No immediate key insight that leads to a direct solution. The problem requires a detailed analysis of different crossing sequences to determine the fastest time.

**d. Rephrased Question:** (The core question in its simplest form)

What is the minimum time required for all four people to cross the bridge, considering they have one torch and must cross in pairs?

**e. Known and Deduced Information**

* Explicit Facts:
    * "A group of four people needs to cross a bridge at night." 
    * "The bridge is very old and rickety." 
    * "They have only one torch." 
    * "because it's night-time, the torch is necessary to cross the bridge." 
    * "Each person walks at a different speed:"
        * "A takes 1 minute to cross"
        * "B takes 2 minutes"
        * "C takes 5 minutes"
        * "D takes 10 minutes"

* Feature-Based Deduction:
    * Deduction: At least one person must carry the torch back across the bridge after each crossing. - Logical Basis: The torch is necessary to cross, and there's only one. - Based on: "They have only one torch.", "because it's night-time, the torch is necessary to cross the bridge." 
    * Deduction: A maximum of two people can cross the bridge at once. - Logical Basis: The problem doesn't explicitly state this, but it's implied by the need for the torch and there only being one. 
    * Deduction: The slower person in a pair determines the crossing time for that pair. - Logical Basis: They must walk together. - Based on: (implicit assumption based on the nature of crossing a bridge together)

**f. Problem Decomposition**

* Component/Subcomponent: Crossing the bridge
    * Supporting Quote(s): "A group of four people needs to cross a bridge at night."
* Component/Subcomponent: Torch as a constraint
    * Supporting Quote(s): "They have only one torch.", "because it's night-time, the torch is necessary to cross the bridge."
* Component/Subcomponent: Different crossing speeds
    * Supporting Quote(s): "Each person walks at a different speed: - A takes 1 minute to cross, - B takes 2 minutes, - C takes 5 minutes, and - D takes 10 minutes." 
* Component/Subcomponent: Finding the fastest time
    * Supporting Quote(s): "What is the fastest time they can all get across the bridge?"

**g. Graph of Thought** (describe connections & insights)

* Key Concepts:
    * People (A, B, C, D)
    * Bridge
    * Torch
    * Time
    * Crossing Speed

* Relationships:
    * People need to cross the bridge.
    * The torch is required for crossing.
    * Each person has a different crossing speed.
    * The goal is to minimize the total crossing time.

* Insights:
    * The slowest person in a pair will dictate the crossing time for that pair.
    * Someone must bring the torch back each time, adding to the total time.
    * Optimizing the pairs and the order of crossings is crucial to finding the fastest solution.

**h. Spatial and Object Analysis**

* Objects:
    * Four people (A, B, C, D)
    * Bridge
    * Torch

* Spatial Relationships:
    * People are initially on one side of the bridge.
    * The goal is to get all people to the other side.

* Movement and Dynamics:
    * Two people cross the bridge at a time with the torch.
    * One person brings the torch back.

* Constraints:
    * Only two people can cross at once.
    * The torch must be carried across at all times.

* Spatial Implications:
    * The need to bring the torch back creates additional crossings, impacting the total time.

**i. Bayesian Updates** 

* No new information is presented in the problem to update prior beliefs.

**j. Ethical Check and Decision-Making Under Uncertainty**

* There are no ethical concerns in this problem as it's a purely logistical puzzle.

**k. Multiple Solutions**

Let's explore a few possible solutions and analyze their total time:

* **Solution 1: Intuitive Approach**
    * A & B cross together (2 minutes)
    * A returns (1 minute)
    * C & D cross together (10 minutes)
    * B returns (2 minutes)
    * A & B cross together (2 minutes)
    * Total time: 17 minutes

* **Solution 2: Optimizing for the slowest**
    * A & C cross together (5 minutes)
    * A returns (1 minute)
    * D & B cross together (10 minutes)
    * B returns (2 minutes)
    * A & B cross together (2 minutes)
    * Total time: 20 minutes

* **Solution 3: Minimizing return trips**
    * A & B cross together (2 minutes)
    * A returns (1 minute)
    * C & D cross together (10 minutes)
    * B returns (2 minutes)
    * A & B cross together (2 minutes)
    * Total time: 17 minutes

**l. Quickest and Easiest Solution**

The quickest solution appears to be either Solution 1 or Solution 3, both taking a total of 17 minutes. 

**m. Reflection**

* The analysis directly addresses the question of finding the fastest crossing time.
* All conclusions are based solely on the explicitly stated information in the problem.
* No assumptions were made beyond what is necessary to solve the problem (e.g., two people cross at a time).
* The solutions explored consider different crossing sequences and their impact on the total time.
* The quickest and easiest solution is identified based on the analysis.

**Recommendation:**

The fastest time they can all get across the bridge is 17 minutes. This can be achieved by following either Solution 1 or Solution 3 as outlined above. 
\end{lstlisting}
\end{quote}
\noindent\hrulefill

\noindent\textbf{D.14 Zero-Shot Claude 3.5 Sonnet Response - Score: 0\%}
\begin{quote}
\begin{lstlisting}[basicstyle=\ttfamily\small, breaklines=true]
This is an interesting puzzle that requires some strategic thinking. Let's approach this step-by-step:

1) First, we need to understand the constraints:
   - Only two people can cross at a time.
   - When two people cross, they must move at the speed of the slower person.
   - The torch must be brought back by one person for the next group to cross.

2) Let's think about the optimal strategy:
   - We want to minimize the time the slowest people (C and D) spend on the bridge.
   - We should try to send the fastest people (A and B) back with the torch.

3) Here's a possible solution:

   Step 1: A and B cross (2 minutes)
   Step 2: A returns with the torch (1 minute)
   Step 3: C and D cross (10 minutes)
   Step 4: B returns with the torch (2 minutes)
   Step 5: A and B cross (2 minutes)

4) Let's sum up the total time:
   2 + 1 + 10 + 2 + 2 = 17 minutes

Therefore, the fastest time they can all get across the bridge is 17 minutes.

This solution minimizes the time the slowest people (C and D) spend on the bridge by having them cross together. The fastest person (A) is used to bring the torch back the first time, and the second fastest (B) brings it back the second time.

Would you like me to explain any part of this solution in more detail?
\end{lstlisting}
\end{quote}
\noindent\hrulefill

\noindent\textbf{D.14 REAP Claude 3.5 Sonnet Response - Score: 0\%}
\begin{quote}
\begin{lstlisting}[basicstyle=\ttfamily\small, breaklines=true]
# REAP Results:

a. **Comprehensive Feature List** (with explicit quotes)

1. "A group of four people needs to cross a bridge at night."
2. "The bridge is very old and rickety."
3. "They have only one torch"
4. "because it's night-time, the torch is necessary to cross the bridge"
5. "Each person walks at a different speed:"
6. "A takes 1 minute to cross"
7. "B takes 2 minutes"
8. "C takes 5 minutes"
9. "D takes 10 minutes"
10. "What is the fastest time they can all get across the bridge?"

b. **Sequential and Mechanical Process Check**

The problem implies a sequential process of crossing the bridge. Key considerations:
- Only one torch is available, necessitating trips back and forth.
- The speed of each crossing is determined by the slowest person in the group.
- The sequence of who crosses when will impact the total time.

c. **Key Insight Check Results**

Key insights:
1. The torch must return for others to cross, requiring someone to come back.
2. The fastest person(s) should likely be used to bring the torch back.
3. The two slowest people (C and D) crossing together could be significant.

d. **Rephrased Question:** What is the minimum time required for four people with different crossing speeds to cross a bridge using a single torch?

e. **Known and Deduced Information**

Explicit Facts:
- "A group of four people needs to cross a bridge at night."
- "The bridge is very old and rickety."
- "They have only one torch"
- "because it's night-time, the torch is necessary to cross the bridge"
- "Each person walks at a different speed:"
- "A takes 1 minute to cross"
- "B takes 2 minutes"
- "C takes 5 minutes"
- "D takes 10 minutes"

Feature-Based Deduction:
Deduction: At least two trips across the bridge are necessary.
Logical Basis: There are four people but only one torch, and the torch is necessary for crossing.
Based on: "A group of four people needs to cross a bridge at night.", "They have only one torch", "because it's night-time, the torch is necessary to cross the bridge"

Deduction: The time for each crossing is determined by the slowest person in that crossing.
Logical Basis: If people are crossing together, they must stay together due to the single torch.
Based on: "They have only one torch", "because it's night-time, the torch is necessary to cross the bridge", "Each person walks at a different speed:"

f. **Problem Decomposition**

Component 1: Bridge Crossing
Supporting Quote: "A group of four people needs to cross a bridge at night."

Component 2: Torch Requirement
Supporting Quote: "They have only one torch", "because it's night-time, the torch is necessary to cross the bridge"

Component 3: Individual Crossing Times
Supporting Quotes: 
"A takes 1 minute to cross"
"B takes 2 minutes"
"C takes 5 minutes"
"D takes 10 minutes"

Component 4: Time Optimization
Supporting Quote: "What is the fastest time they can all get across the bridge?"

g. **Graph of Thought** (describe connections & insights)

Nodes:
1. Start (all on one side)
2. A crosses
3. B crosses
4. C crosses
5. D crosses
6. Torch returns
7. End (all on other side)

Edges represent possible movements between states. The challenge is to find the optimal path through this graph that minimizes total time while ensuring all nodes (people) reach the end state.

h. **Spatial and Object Analysis**

Objects:
1. Bridge
2. Torch
3. Person A
4. Person B
5. Person C
6. Person D

Spatial Relationships:
- All people start on the same side of the bridge.
- The bridge connects the starting side to the destination side.
- The torch must accompany any person or group crossing the bridge.

Movement and Dynamics:
- People can cross in either direction.
- The torch must return to the starting side for subsequent crossings.
- The speed of each crossing is determined by the slowest person in the group.

i. **Bayesian Updates** 

No new information is provided beyond the initial problem statement, so no Bayesian updates are necessary.

j. **Ethical Check and Decision-Making Under Uncertainty**

The problem does not present any severe ethical concerns or risks. The main consideration is optimizing the crossing time while adhering to the given constraints.

k. **Multiple Solutions**

Solution 1:
1. A and B cross together (2 minutes)
2. A returns with the torch (1 minute)
3. C and D cross together (10 minutes)
4. B returns with the torch (2 minutes)
5. A and B cross together (2 minutes)
Total time: 17 minutes

Solution 2:
1. A and C cross together (5 minutes)
2. A returns with the torch (1 minute)
3. A and D cross together (10 minutes)
4. A returns with the torch (1 minute)
5. A and B cross together (2 minutes)
Total time: 19 minutes

Solution 3:
1. A and B cross together (2 minutes)
2. A returns with the torch (1 minute)
3. A and C cross together (5 minutes)
4. A returns with the torch (1 minute)
5. A and D cross together (10 minutes)
Total time: 19 minutes

l. **Quickest and Easiest Solution**

Based solely on the explicit information given in the problem statement, the quickest solution is Solution 1, with a total time of 17 minutes.

This solution minimizes the time by:
1. Pairing the two slowest people (C and D) to cross together.
2. Using the fastest person (A) to return the torch twice.
3. Efficiently pairing A and B for two crossings.

m. **Reflection**

The analysis adheres strictly to the given information without making unsupported assumptions. The solution directly answers the question of finding the fastest time for all to cross. The problem provides sufficient information to determine an optimal solution without requiring additional assumptions.

**Recommendation:**

Based on the analysis of the explicitly stated information, I recommend Solution 1 as the fastest way for all four people to cross the bridge. This solution results in a total crossing time of 17 minutes. The recommended sequence is:

1. A and B cross together (2 minutes)
2. A returns with the torch (1 minute)
3. C and D cross together (10 minutes)
4. B returns with the torch (2 minutes)
5. A and B cross together (2 minutes)

This solution optimally utilizes the faster individuals (A and B) to shuttle the torch back and pairs the slower individuals (C and D) to minimize the number of crossings. It directly answers the question asked in the problem while adhering to all stated constraints and using only the information explicitly provided.
\end{lstlisting}
\end{quote}
\noindent\hrulefill

\label{subsec:question15}
\subsection{\textbf{\large Question 15: You're in a room with two doors that lead out. One door leads to certain death, and the other door leads to freedom. There are two guardians, one by each door. One taller guardian always tells the truth and guards the death door, the other always lies. What is the minimum number of questions needed to ask the guards to get to safety?}}
\label{sec:question15}

\noindent\hrulefill 

\noindent\textbf{Human Response - Score 100\%:}
\begin{quote}
Zero questions. The door to freedom is behind the shorter guardian.
\end{quote}

\noindent\hrulefill 

\noindent\textbf{D.15 Zero-Shot OpenAI o1-preview Response - Score: 100\%}
\begin{quote}
\begin{lstlisting}[basicstyle=\ttfamily\small, breaklines=true]
Since the taller guardian always tells the truth and guards the death door, and the shorter guardian always lies and guards the door to freedom, you can simply choose the door guarded by the shorter guardian without asking any questions. Knowing who is guarding which door allows you to directly identify the safe door.

**Answer:** 0
\end{lstlisting}
\end{quote}

\noindent\hrulefill 

\noindent\textbf{D.15 REAP OpenAI o1-preview Response - Score: 100\%}
\begin{quote}
\begin{lstlisting}[basicstyle=\ttfamily\small, breaklines=true]
# REAP Results:

a. **Comprehensive Feature List** (with explicit quotes)

- "You're in a room with two doors that lead out."
- "One door leads to certain death"
- "the other door leads to freedom."
- "There are two guardians, one by each door."
- "One taller guardian always tells the truth and guards the death door"
- "the other always lies."
- "What is the minimum number of questions needed to ask the guards to get to safety?"

b. **Sequential and Mechanical Process Check**

- There is no specific sequence of events or mechanical processes described in the problem that influence the outcome.

c. **Key Insight Check Results**

- The key insight is that the taller guardian, who always tells the truth, guards the death door. This means you can identify the death door based on the guardian's height and avoid it without asking any questions.

d. **Rephrased Question:**

- What is the minimum number of questions needed to ask the guards to get to safety?

e. **Known and Deduced Information**

**Explicit Facts:**

- "You're in a room with two doors that lead out."
- "One door leads to certain death"
- "the other door leads to freedom."
- "There are two guardians, one by each door."
- "One taller guardian always tells the truth and guards the death door"
- "the other always lies."

**Deductions:**

- **Deduction:** The taller guardian guards the death door.
  - **Logical Basis:** "One taller guardian always tells the truth and guards the death door."
- **Deduction:** The other guardian, who always lies, guards the freedom door.
  - **Logical Basis:** "There are two guardians, one by each door." Since the taller guardian guards the death door, the other guardian must guard the other door, which leads to freedom.

f. **Problem Decomposition**

- **Component:** Room with two doors
  - **Supporting Quote(s):** "You're in a room with two doors that lead out."
- **Component:** Death door
  - **Supporting Quote(s):** "One door leads to certain death."
  - **Deduction:** Guarded by the taller guardian.
- **Component:** Freedom door
  - **Supporting Quote(s):** "the other door leads to freedom."
  - **Deduction:** Guarded by the other guardian.
- **Component:** Taller guardian
  - **Supporting Quote(s):** "One taller guardian always tells the truth and guards the death door"
- **Component:** Other guardian
  - **Supporting Quote(s):** "the other always lies."
  - **Deduction:** Guards the freedom door.

g. **Graph of Thought**

- **Nodes:**
  - "You"
  - "Room"
  - "Death door" (guarded by taller guardian)
  - "Freedom door" (guarded by lying guardian)
- **Edges:**
  - "You" are in the "Room."
  - The "Room" has two doors: "Death door" and "Freedom door."
  - "Death door" is guarded by the "taller guardian" who "always tells the truth."
  - "Freedom door" is guarded by the "other guardian" who "always lies."
- **Insights:**
  - Since the taller guardian is identifiable and guards the death door, you can avoid the death door without asking any questions.

h. **Spatial and Object Analysis**

- **Objects Identified:**
  - You
  - Room
  - Two doors (death door and freedom door)
  - Taller guardian
  - Other guardian
- **Spatial Relationships:**
  - You are inside a room with two doors leading out.
  - Each door has one guardian stationed by it.
  - The taller guardian is by the death door.
  - The other guardian is by the freedom door.
- **Visualization:**
  - A room with two doors; the taller guardian stands by one door (death door), and the other guardian stands by the other door (freedom door).
- **Spatial Implications:**
  - By identifying the guardians based on height, you can determine which door leads to freedom.

i. **Bayesian Updates**

- No new information is provided beyond the explicit statements; no Bayesian updates are necessary.

j. **Ethical Check and Decision-Making Under Uncertainty**

- **Assess Known Information:** You know which door leads to death based on the guardian's height.
- **Identify Risks:** Choosing the wrong door results in certain death.
- **Decision Recommendation:** To minimize risk, choose the door guarded by the other guardian (who always lies), leading to freedom, without asking any questions.

k. **Multiple Solutions**

1. **Solution 1:** Proceed through the door guarded by the other guardian without asking any questions.
   - **Supporting Information:** "One taller guardian always tells the truth and guards the death door" and "the other always lies."
2. **Solution 2:** Ask the other guardian any question (although unnecessary) to confirm, but since you already know the necessary information, no questions are needed.
3. **Solution 3:** Not applicable; all solutions point to needing zero questions based on explicit information.

l. **Quickest and Easiest Solution**

- **Answer:** The minimum number of questions needed is zero.
- **Explanation:** Since the taller guardian guards the death door, you can identify and avoid it without asking any questions. Proceed through the door guarded by the other guardian.

m. **Reflection**

- The analysis strictly adheres to the explicit information provided.
- All deductions are 100% certain based on the problem's statements.
- No assumptions or inferences beyond what is explicitly stated have been made.
- The solution directly answers the question using only the given information.

**Recommendation:**

Based solely on the explicit information provided in the problem, the minimum number of questions needed to ask the guards to get to safety is **zero**. Since the taller guardian, who always tells the truth, guards the death door, you can safely identify and avoid the death door. Therefore, by choosing the door guarded by the other guardian, who always lies, you can get to safety without asking any questions.
\end{lstlisting}
\end{quote}

\noindent\hrulefill 

\noindent\textbf{D.15 Zero-Shot OpenAI o1-mini Response - Score: 0\%}
\begin{quote}
\begin{lstlisting}[basicstyle=\ttfamily\small, breaklines=true]
You can determine the safe door by asking just **one question**. Here's how:

**Question to Ask Either Guardian:**
*"If I were to ask the other guardian which door leads to freedom, what would they say?"*

**How It Works:**

1. **If you ask the Truth-Telling Guardian:**
   - They will truthfully report that the lying guardian would point you to the death door.

2. **If you ask the Lying Guardian:**
   - They will lie about the truth-teller's response and also point you to the death door.

**Conclusion:**
Regardless of whom you ask, the guardian will indicate the death door. Therefore, you should choose the **opposite door** to gain freedom.

---

**Minimum Number of Questions Needed:** **1**
\end{lstlisting}
\end{quote}
\noindent\hrulefill 

\noindent\textbf{D.15 REAP OpenAI o1-mini Response - Score: 0\%}
\begin{quote}
\begin{lstlisting}[basicstyle=\ttfamily\small, breaklines=true]
# REAP Results: :

### a. **Comprehensive Feature List**
- "You're in a room with two doors that lead out."
  - Indicates the presence of two exit options.
- "One door leads to certain death, and the other door leads to freedom."
  - Specifies the outcomes associated with each door.
- "There are two guardians, one by each door."
  - Introduces two individuals guarding the doors.
- "One taller guardian always tells the truth and guards the death door."
  - Details that the taller guardian is truthful and associated with the deadly door.
- "The other always lies."
  - Specifies that the shorter guardian is deceitful.
- "What is the minimum number of questions needed to ask the guards to get to safety?"
  - The core question seeking the least number of inquiries required for safety.

### b. **Sequential and Mechanical Process Check**
- No explicit sequential or mechanical processes are described in the problem statement.

### c. **Key Insight Check Results**
- **Key Insight:** The taller guardian always tells the truth and guards the death door, while the other always lies.
  - **Explanation:** Knowing the truth-teller guards death allows formulating a question that identifies the safe door based on this fixed relationship.

### d. **Rephrased Question:**
What is the fewest number of questions required to determine the safe door and achieve freedom?

### e. **Known and Deduced Information**

**Explicit Facts:**
- "One door leads to certain death, and the other door leads to freedom."
- "One taller guardian always tells the truth and guards the death door."
- "The other always lies."

**Feature-Based Deduction:**
- Deduction: The truthful guardian is associated with the deadly door.
  - Logical Basis: Directly stated that the truthful (taller) guardian guards the death door.
  - Based on: "One taller guardian always tells the truth and guards the death door."

- Deduction: The lying guardian guards the freedom door.
  - Logical Basis: Since each guardian guards one door and their attributes are defined, the opposite arrangement applies.
  - Based on: "The other always lies."

### f. **Problem Decomposition**

**Component/Subcomponent:** Guardians and their attributes
- **Supporting Quote(s):**
  - "One taller guardian always tells the truth and guards the death door."
  - "The other always lies."
- **Deduction:** Identifies each guardian's truthfulness and associated door.

**Component/Subcomponent:** Doors and outcomes
- **Supporting Quote(s):**
  - "One door leads to certain death, and the other door leads to freedom."

**Component/Subcomponent:** Objective to reach safety
- **Supporting Quote(s):**
  - "What is the minimum number of questions needed to ask the guards to get to safety?"

### g. **Graph of Thought** (describe connections & insights)
- **Nodes:**
  - Two doors (Death Door, Freedom Door)
  - Two guardians (Truthful Guardian, Lying Guardian)
- **Edges:**
  - Truthful Guardian -> Death Door
  - Lying Guardian -> Freedom Door
- **Insights:**
  - By questioning either guardian, leveraging their fixed association and truthfulness/deception can reveal the safe door.

### h. **Spatial and Object Analysis**

**Identify Objects:**
- Two doors
- Two guardians

**Spatial Relationships:**
- Each guardian is positioned by one door.

**Movement and Dynamics:**
- Static positions; no movement described.

**Constraints:**
- One door is lethal; the other offers freedom.
- Guardians have fixed truthfulness and associated doors.

**Visualization:**
- A room with two doors, each guarded by one individual with defined truth-telling behaviors.

### i. **Bayesian Updates**
- No probabilistic information is provided; all relationships are deterministic based on guards' attributes.

### j. **Ethical Check and Decision-Making Under Uncertainty**
- **Objective:** Ensure the chosen door leads to safety.
- **Risk Assessment:** High risk if the wrong door is chosen.
- **Decision-Making:** Must ask questions that eliminate the risk of choosing the death door.

### k. **Multiple Solutions**

1. **Solution 1: Ask Either Guardian**
   - **Question:** "If I were to ask the other guardian which door leads to freedom, what would they say?"
   - **Support:** Utilizes the fixed roles of the guardians to determine the safe door.
   
2. **Solution 2: Direct Identification**
   - **Question:** "Which door do you guard?"
   - **Support:** Given the guardians' fixed associations, this can identify the death door, allowing selection of the opposite.

3. **Solution 3: Confirmation of Guard's Identity**
   - **Question:** "Are you the guardian who tells the truth?"
   - **Support:** Identifies the truthful guardian and, by association, the death door.

### l. **Quickest and Easiest Solution**
**Solution 1:** Ask either guardian, "If I were to ask the other guardian which door leads to freedom, what would they say?" Then choose the opposite door.
- **Reasoning:** This single question leverages the guardians' fixed behaviors to reliably identify the safe door without ambiguity.

### m. **Reflection**
The problem provides clear, deterministic relationships between the guardians and the doors. By adhering strictly to the given information and avoiding assumptions, it's evident that asking a single, strategically formulated question suffices to identify the safe door. No contradictions exist within the known facts, and ethical considerations prioritize safety by ensuring only the safe door is chosen based on the responses.

### **Recommendation:**
The minimum number of questions required to ensure safety is **one**. By asking either guardian a question that accounts for their truth-telling or lying nature, such as "If I were to ask the other guardian which door leads to freedom, what would they say?" you can reliably determine the safe door by choosing the opposite of the guardian's response.
\end{lstlisting}
\end{quote}

\noindent\hrulefill 

\noindent\textbf{D.15 Zero-Shot OpenAI GPT-4o Response - Score: 0\%}
\begin{quote}
\begin{lstlisting}[basicstyle=\ttfamily\small, breaklines=true]
The minimum number of questions needed to guarantee that you can choose the door to freedom is **one**. 

Here's how you can do it:

Ask either guardian the following question: **"If I were to ask the other guardian which door leads to freedom, which door would they point to?"**

Then, choose the **opposite** door.

### Explanation:
1. **If you ask the truth-teller:**
   - The truth-teller knows that the other guardian (the liar) would point to the wrong door (the death door), so the truth-teller will truthfully point to the death door when answering your question.

2. **If you ask the liar:**
   - The liar knows that the truth-teller would point to the correct door (the freedom door), but since the liar always lies, they will point to the death door.

In either case, the guardian will point to the death door, so you should pick the opposite door to find freedom.
\end{lstlisting}
\end{quote}
\noindent\hrulefill

\noindent\textbf{D.15 REAP OpenAI GPT-4o Response - Score: 0\%}
\begin{quote}
\begin{lstlisting}[basicstyle=\ttfamily\small, breaklines=true]
# REAP Results:

## a. **Comprehensive Feature List** (with explicit quotes)
- **Objects Mentioned:**
  - "room"
  - "two doors"
  - "one door leads to certain death"
  - "the other door leads to freedom"
  - "two guardians"
  - "One taller guardian"
  - "other guardian"

- **Actors or Entities Involved:**
  - "taller guardian" 
  - "other guardian"

- **Actions Described:**
  - "One taller guardian always tells the truth and guards the death door"
  - "the other always lies"

- **Directional, Spatial, Relational, and Sequential Phrases:**
  - "One door leads to certain death"
  - "the other door leads to freedom"
  - "One taller guardian always tells the truth and guards the death door"
  - "the other always lies"

- **Stated Conditions or Constraints:**
  - "One door leads to certain death"
  - "the other door leads to freedom"
  - "One taller guardian always tells the truth and guards the death door"
  - "the other always lies"

- **Numerical Values or Quantities:**
  - "two doors"
  - "two guardians"
  - "one door leads to certain death"
  - "the other door leads to freedom"
  - "One taller guardian"
  - "the other guardian"

- **Temporal Information:**
  - Not explicitly mentioned.

- **Spatial Information:**
  - Not explicitly mentioned.

- **Causal Relationships:**
  - "One taller guardian always tells the truth and guards the death door"
  - "the other always lies"

- **Stated Goals or Objectives:**
  - "Get to safety"

## b. **Sequential and Mechanical Process Check**
- The problem does not explicitly mention any sequences or mechanical processes.
- The key sequence is the need to ask questions to the guardians to determine which door leads to freedom.

## c. **Key Insight Check Results**
- The critical detail is that one guardian "always tells the truth and guards the death door" while the other "always lies."
- The key insight is that the truth-teller guards the death door, while the liar guards the freedom door. 

## d. **Rephrased Question:**
What is the minimum number of questions needed to identify the door to freedom?

## e. **Known and Deduced Information**
- "One door leads to certain death" 
- "The other door leads to freedom" 
- "One taller guardian always tells the truth and guards the death door" 
- "The other always lies"

Deduction: You need to identify which door leads to freedom by asking the minimum number of questions, knowing that the truth-teller guards the death door.

## f. **Problem Decomposition**
- **Component:** Guardians' behavior
  - **Supporting Quote:** "One taller guardian always tells the truth and guards the death door"
  - **Supporting Quote:** "The other always lies"
- **Component:** Doors' outcomes
  - **Supporting Quote:** "One door leads to certain death"
  - **Supporting Quote:** "The other door leads to freedom"
- **Interaction:** The need to ask questions to distinguish between the guardians to determine which door leads to freedom.

## g. **Graph of Thought**
- **Nodes:**
  - Guardian 1 (Truth-teller, death door)
  - Guardian 2 (Liar, freedom door)
  - Doors (Death and Freedom)
- **Edges:**
  - Questions to be asked to determine the truth.

## h. **Spatial and Object Analysis**
- The problem lacks explicit spatial details, but it is implied that each guardian is standing by a door.

## i. **Bayesian Updates**
- No Bayesian updates are applicable since the problem doesn't involve probabilistic reasoning.

## j. **Ethical Check and Decision-Making Under Uncertainty**
- The primary risk is death, so the solution must ensure safety by avoiding the door that leads to death.
- The safest course of action is to ask questions that definitively lead to the identification of the freedom door.

## k. **Multiple Solutions**
1. **Solution 1:** Ask the taller guardian (truth-teller): "If I were to ask the other guardian which door leads to freedom, which door would they indicate?" 
   - The liar will point to the death door. Therefore, the door the truth-teller points to is the death door.
   - Choose the other door.

2. **Solution 2:** Ask either guardian the same question. 
   - The answer will always point to the death door.
   - Choose the opposite door.

## l. **Quickest and Easiest Solution**
- Ask either guardian: "If I were to ask the other guardian which door leads to freedom, which door would they indicate?"
- The door they indicate is the death door; choose the other door.

## m. **Reflection**
- The solution is correct based on the explicit information. No assumptions were made beyond the given details.
- The analysis directly addresses the problem, and the recommended action is safe and logical.

**Recommendation:**
- **Final Answer:** Ask either guardian: "If I were to ask the other guardian which door leads to freedom, which door would they indicate?" Then, choose the opposite door to ensure safety. This method guarantees you pick the door leading to freedom with just one question.
\end{lstlisting}
\end{quote}
\noindent\hrulefill

\noindent\textbf{D.15 Zero-Shot OpenAI GPT-4o-mini Response - Score: 0\%}
\begin{quote}
\begin{lstlisting}[basicstyle=\ttfamily\small, breaklines=true]
To determine which door leads to safety, you need to ask just **one** question. Here's the strategy:

1. **Approach either guardian** (it doesn't matter which one you choose).
2. **Ask the following question**: 

   "If I were to ask the other guardian which door leads to safety, which door would they point to?"

Here's how the question works:

- If you are talking to the **truthful guardian** (who guards the death door), they will truthfully tell you that the lying guardian would point to the death door.
- If you are talking to the **lying guardian** (who guards the safety door), they will lie about the truthful guardian's response, meaning they will also point to the death door.

In both cases, the answer you receive will indicate the death door. Therefore, you should choose the opposite door to find safety.
\end{lstlisting}
\end{quote}
\noindent\hrulefill

\noindent\textbf{D.15 REAP OpenAI GPT-4o-mini Response - Score: 0\%}
\begin{quote}
\begin{lstlisting}[basicstyle=\ttfamily\small, breaklines=true]
# REAP Results:

## a. **Comprehensive Feature List**

1. **Objects Mentioned:**
   - "Two doors that lead out."
   - "One door leads to certain death."
   - "The other door leads to freedom."
   - "Two guardians, one by each door."

2. **Actors/Entities Involved:**
   - "One taller guardian always tells the truth."
   - "The other guardian always lies."

3. **Actions Described:**
   - "One door leads to certain death."
   - "The other door leads to freedom."
   - "One guardian always tells the truth."
   - "The other guardian always lies."

4. **Directional, Spatial, Relational, and Sequential Phrases:**
   - "Two doors that lead out" (implies both doors are exit points but with different outcomes).
   - "One door leads to certain death" (one door is associated with death).
   - "The other door leads to freedom" (the other door is associated with safety).
   - "One guardian by each door" (each door has one guardian).

5. **Stated Conditions or Constraints:**
   - "One guardian always tells the truth" (truth-teller).
   - "The other guardian always lies" (liar).

6. **Numerical Values or Quantities Mentioned:**
   - "Two doors"
   - "Two guardians"

7. **Temporal Information:**
   - None explicitly stated.

8. **Spatial Information:**
   - Each guardian is positioned by one door.

9. **Causal Relationships:**
   - Not explicitly stated.

10. **Stated Goals or Objectives:**
    - To identify which door leads to freedom and which leads to death.

## b. **Sequential and Mechanical Process Check**

- **Sequence of Actions:**
  - Determine which door leads to safety by asking questions to the guardians.
  
- **Key Questions to Ask:**
  - Does asking a single question yield enough information to distinguish between the doors?
  - Does the sequence of asking questions to both guardians or just one provide a clear outcome?

## c. **Key Insight Check Results**

- **Key Insight:**
  - You can determine the safe door by asking a single question to either guardian. The critical insight is to frame the question so that it works regardless of which guardian (truth-teller or liar) you ask.

## d. **Rephrased Question**

- What is the minimum number of questions needed to determine which door leads to safety?

## e. **Known and Deduced Information**

1. **Explicit Quotes:**
   - "One door leads to certain death."
   - "The other door leads to freedom."
   - "One taller guardian always tells the truth."
   - "The other guardian always lies."

2. **Feature-Based Deduction:**
   - Deduction: You need only one question to identify the safe door.
   - Logical Basis: Asking a question that reveals the safe door irrespective of whether the guardian lies or tells the truth is possible. This is based on the explicit condition that one guardian always tells the truth and the other always lies.

## f. **Problem Decomposition**

1. **Component/Subcomponent:**
   - **Guardian 1:** Truth-teller (guards the death door).
   - **Guardian 2:** Liar (guards the safety door).

2. **Supporting Quote(s):**
   - "One guardian always tells the truth."
   - "The other guardian always lies."

3. **Deduction (if applicable):**
   - **Questioning Strategy:**
     - You can use a question to both guardians that leverages the fact that they provide opposite answers based on their nature. For example, asking "If I asked the other guardian which door leads to freedom, what would they say?" yields the opposite of the truth. 

## g. **Graph of Thought**

- **Nodes:**
  - Door 1 (Freedom)
  - Door 2 (Death)
  - Guardian 1 (Truth-teller)
  - Guardian 2 (Liar)

- **Edges:**
  - Guardian 1 guards Door 2 (Death).
  - Guardian 2 guards Door 1 (Freedom).
  - Both guardians provide answers based on truth or lies.

- **Connections:**
  - Asking about the other guardian's answer lets you bypass the truth/lie inconsistency.

## h. **Spatial and Object Analysis**

- **Objects:**
  - Two doors, two guardians.

- **Spatial Relationships:**
  - Each guardian is next to one door.

- **Movement and Dynamics:**
  - No movement is necessary; just a question.

- **Constraints:**
  - The question must work regardless of which guardian is asked.

## i. **Bayesian Updates**

- No new information needs to be updated beyond the explicitly stated conditions.

## j. **Ethical Check and Decision-Making Under Uncertainty**

- **Risks:**
  - Ensuring safety by asking a clear, binary question minimizes the risk of choosing the wrong door.

## k. **Multiple Solutions**

1. **Solution 1:**
   - Ask either guardian: "If I asked the other guardian which door leads to freedom, which door would they point to?"
     - This question works because it results in the opposite door being indicated.

2. **Solution 2:**
   - Asking the same question to both guardians and choosing the door they do not both indicate could be another approach, but it complicates the process.

3. **Solution 3:**
   - A simpler method involves asking either guardian directly about the truth of the other guardian's statement.

## l. **Quickest and Easiest Solution**

- **Simplest Answer:**
  - Ask either guardian: "If I asked the other guardian which door leads to freedom, which door would they say?" The door indicated by this question will be the death door; choose the opposite door for safety.

## m. **Reflection**

- **Analysis:**
  - The question provided will always yield a clear result based on the properties of the guardians. Each approach aligns with the requirement to determine the safe door with minimal questioning.

**Recommendation:**

Ask either guardian: "If I asked the other guardian which door leads to freedom, which door would they point to?" Choose the opposite door to the one indicated. This method ensures you avoid the death door with just one question, adhering strictly to the explicit information given.
\end{lstlisting}
\end{quote}
\noindent\hrulefill

\noindent\textbf{D.15 Zero-Shot Google Gemini 1.5 Pro Response - Score: 0\%}
\begin{quote}
\begin{lstlisting}[basicstyle=\ttfamily\small, breaklines=true]
You only need to ask **one question** to find the door to freedom. 

Here's how:

1. **Choose one of the guardians (it doesn't matter which).**
2. **Ask them:** "Which door would the other guardian tell me leads to freedom?"

Here's why this works:

* **If you ask the truthful guardian,** they will point you to the door the lying guardian would recommend (which is the death door). 
* **If you ask the lying guardian,** they will lie about what the truthful guardian would say. The truthful guardian would point to the freedom door, so the lying guardian will point to the death door.

In either case, the door they point to is the death door. Simply choose the other door to find freedom. 
\end{lstlisting}
\end{quote}
\noindent\hrulefill

\noindent\textbf{D.15 REAP Google Gemini 1.5 Pro Response - Score: 0\%}
\begin{quote}
\begin{lstlisting}[basicstyle=\ttfamily\small, breaklines=true]
# REAP Results:

a. **Comprehensive Feature List** (with explicit quotes)

* "You're in a room with two doors that lead out." 
* "One door leads to certain death, and the other door leads to freedom."
* "There are two guardians, one by each door."
* "One taller guardian always tells the truth and guards the death door."
* "The other always lies."
* "What is the minimum number of questions needed to ask the guards to get to safety?"

b. **Sequential and Mechanical Process Check**

* There is a sequential process involved: asking questions to the guardians to determine the door to freedom.
* The key is to identify the truthful guardian and then ask them about the doors.

c. **Key Insight Check Results**

* The key insight is that we need to identify the truthful guardian to get the correct information about the doors. 
* We can leverage the fact that one guardian always lies and the other always tells the truth.

d. **Rephrased Question:** (The core question in its simplest form)

* What is the fewest number of questions you need to ask to find the door to freedom?

e. **Known and Deduced Information**

* Explicit Facts:
    * "You're in a room with two doors that lead out." 
    * "One door leads to certain death, and the other door leads to freedom." - There are two distinct outcomes associated with the doors.
    * "There are two guardians, one by each door." - Each door has a guardian
    * "One taller guardian always tells the truth and guards the death door."
        * The taller guardian is truthful
        * The taller guardian guards the death door
    * "The other always lies." - The shorter guardian lies

* Feature-Based Deduction
    * Deduction: The shorter guardian guards the freedom door
    * Logical Basis: Each guardian guards one door, and the taller guardian guards the death door
    * Based on: "There are two guardians, one by each door.", "One taller guardian always tells the truth and guards the death door."

f. **Problem Decomposition**

* Component/Subcomponent: Identify the truthful guardian
* Supporting Quote(s): "One taller guardian always tells the truth and guards the death door.", "The other always lies."
* Deduction: We can identify the truthful guardian by their height or by asking a question that reveals their nature

* Component/Subcomponent: Ask the truthful guardian about the doors
* Supporting Quote(s): "One door leads to certain death, and the other door leads to freedom."
* Deduction: The truthful guardian will give the correct information about which door leads to freedom

g. **Graph of Thought** (describe connections & insights)

* Key Concepts:
    * Two doors (death, freedom)
    * Two guardians (truth-teller, liar)
    * Identify truthful guardian
    * Ask about the doors

* Relationships:
    * Truth-teller guards death door
    * Liar guards freedom door
    * Need to identify truth-teller to get correct information

* Insights:
    * Can use height difference to identify truth-teller if visible
    * Can ask a question to both guardians to deduce who is the truth-teller

h. **Spatial and Object Analysis**

* Objects:
    * Room
    * Two doors
    * Two guardians

* Spatial Relationships:
    * One guardian by each door
    * You are in the room

* Constraints
    * You cannot leave the room without choosing a door
    * Choosing the wrong door leads to death

* Spatial Implications
    * You need to interact with the guardians to get information about the doors

i. **Bayesian Updates** 

* Initial Belief - 50/50 chance of choosing the correct door
* After identifying the truthful guardian - 100% chance of choosing the correct door if you ask them

j. **Ethical Check and Decision-Making Under Uncertainty**

* Known Information: One door leads to death, the other to freedom. One guardian is truthful, the other lies
* Risks: Choosing the wrong door leads to certain death
* Probability Consideration: Initially 50/50, improves after identifying the truthful guardian
* Risk Aversion Principle: Prioritize identifying the truthful guardian to avoid the risk of death
* Ethical and Safety Considerations: The primary concern is avoiding the door leading to death

k. **Multiple Solutions**

1. **Height-Based Identification:**
* If there's a visible height difference, identify the taller guardian as the truth-teller
* Ask the taller guardian: "Which door leads to freedom?"
* Supporting information: "One taller guardian always tells the truth and guards the death door."

2. **Question-Based Identification**
* Ask either guardian: "Which door would the other guardian say leads to freedom?"
* The truthful guardian will point to the death door (knowing the liar would lie about it)
* The liar will also point to the death door (lying about what the truthful guardian would say)
* Then, ask either guardian about the other door to confirm it leads to freedom
* Supporting information: "One taller guardian always tells the truth and guards the death door.", "The other always lies."

l. **Quickest and Easiest Solution**

* The quickest and easiest solution is the **Question-Based Identification** as it only requires one question to identify the door to freedom

m. **Reflection**

* The analysis directly addresses the question of finding the minimum number of questions
* All conclusions are logically sound and supported by the explicit problem statement
* No assumptions or inferences were made beyond the given information
* The recommended solution prioritizes safety by ensuring the identification of the correct door

**Recommendation:**

Use the Question-Based Identification method. Ask either guardian: "Which door would the other guardian say leads to freedom?" Both guardians will point to the same door (the death door). Then, to be absolutely certain, ask either guardian about the other door to confirm it leads to freedom. This solution guarantees safety with just two questions.
\end{lstlisting}
\end{quote}
\noindent\hrulefill

\noindent\textbf{D.15 Zero-Shot Claude 3.5 Sonnet Response - Score: 0\%}
\begin{quote}
\begin{lstlisting}[basicstyle=\ttfamily\small, breaklines=true]
To solve this puzzle, let's think through it step-by-step:

1) First, we need to realize that we only need to ask one guard a question. We don't need to interact with both.

2) We also need to formulate a question that will give us the correct answer regardless of whether we're asking the truthful guard or the lying guard.

3) The key is to ask a question that involves what the other guard would say. This way, we force the guard we're asking to consider the other guard's response, which effectively "doubles" the lie or truth.

4) The optimal question would be: "If I asked the other guard which door leads to freedom, what would they say?"

Let's consider the two possible scenarios:

- If we ask the truthful guard: They will truthfully tell us what the lying guard would say. The lying guard would lie about the freedom door, so the truthful guard would point to the death door.

- If we ask the lying guard: They will lie about what the truthful guard would say. The truthful guard would point to the freedom door, but the lying guard would say the opposite, again pointing to the death door.

5) In both cases, the guard we ask will point to the death door. Therefore, we know that the other door must lead to freedom.

So, the minimum number of questions needed to get to safety is just 1. We ask either guard, "If I asked the other guard which door leads to freedom, what would they say?" and then choose the opposite door to the one they indicate.
\end{lstlisting}
\end{quote}
\noindent\hrulefill

\noindent\textbf{D.15 REAP Claude 3.5 Sonnet Response - Score: 100\%}
\begin{quote}
\begin{lstlisting}[basicstyle=\ttfamily\small, breaklines=true]
# REAP Results:

a. **Comprehensive Feature List** (with explicit quotes):

1. "You're in a room with two doors that lead out."
2. "One door leads to certain death"
3. "the other door leads to freedom"
4. "There are two guardians, one by each door."
5. "One taller guardian always tells the truth and guards the death door"
6. "the other always lies"

b. **Sequential and Mechanical Process Check**:
There is no explicit sequence of events or mechanical process described in the problem statement.

c. **Key Insight Check Results**:
The key insight is that the taller guardian, who always tells the truth, is explicitly stated to be guarding the death door. This immediately reveals which door leads to death and which leads to freedom.

d. **Rephrased Question:** 
What is the minimum number of questions needed to ask the guards to get to safety?

e. **Known and Deduced Information**:

Explicit Facts:
- "You're in a room with two doors that lead out."
- "One door leads to certain death"
- "the other door leads to freedom"
- "There are two guardians, one by each door."
- "One taller guardian always tells the truth and guards the death door"
- "the other always lies"

Feature-Based Deduction:
Deduction: The shorter guardian guards the freedom door.
Logical Basis: If the taller guardian guards the death door, the only other guardian must guard the other door, which leads to freedom.
Based on: "One taller guardian always tells the truth and guards the death door", "There are two guardians, one by each door.", "One door leads to certain death", "the other door leads to freedom"

f. **Problem Decomposition**:

Component 1: Room Layout
Supporting Quote: "You're in a room with two doors that lead out."

Component 2: Door Outcomes
Supporting Quote: "One door leads to certain death", "the other door leads to freedom"

Component 3: Guardians
Supporting Quote: "There are two guardians, one by each door."

Subcomponent 3a: Taller Guardian
Supporting Quote: "One taller guardian always tells the truth and guards the death door"

Subcomponent 3b: Other Guardian
Supporting Quote: "the other always lies"

g. **Graph of Thought**:

Nodes:
1. Room
2. Death Door
3. Freedom Door
4. Taller Guardian (Truth-teller)
5. Shorter Guardian (Liar)

Connections:
- Room contains Death Door and Freedom Door
- Taller Guardian guards Death Door
- Shorter Guardian guards Freedom Door (deduced)
- Taller Guardian always tells the truth
- Shorter Guardian always lies

h. **Spatial and Object Analysis**:

Objects:
1. Room
2. Two doors
3. Two guardians

Spatial Relationships:
- The room contains two doors that lead out.
- There is one guardian by each door.
- The taller guardian is by the death door.

i. **Bayesian Updates**:
No new information is provided beyond the initial problem statement, so no Bayesian updates are necessary.

j. **Ethical Check and Decision-Making Under Uncertainty**:
The problem involves a potential risk of death, which is a severe consequence. However, the problem provides explicit information about which door leads to death and which leads to freedom. There is no uncertainty about the outcome if we use this information.

k. **Multiple Solutions**:
Given the key insight that we know which door leads to freedom, there is only one logical solution:

Solution 1: Ask no questions and choose the door not guarded by the taller guardian.
Supporting information: "One taller guardian always tells the truth and guards the death door"

l. **Quickest and Easiest Solution**:
The quickest and easiest solution is to ask no questions and choose the door guarded by the shorter guardian, which leads to freedom.

m. **Reflection**:
Upon review, all conclusions are based solely on the explicit information provided in the problem statement. No assumptions or inferences beyond the given information were made. The analysis directly answers the question asked about the minimum number of questions needed to get to safety.

**Recommendation**:
Based on the explicit information provided in the problem statement, the minimum number of questions needed to ask the guards to get to safety is zero. We know that the taller guardian, who always tells the truth, guards the death door. Therefore, we can deduce that the other door, guarded by the shorter guardian who always lies, leads to freedom. No questions need to be asked to determine which door leads to safety. The recommended action is to simply choose the door guarded by the shorter guardian to reach freedom.
\end{lstlisting}
\end{quote}
\noindent\hrulefill

\label{subsec:question16}
\subsection{\textbf{\large Question 16: You have 3 switches in front of you - A, B and C. You have 3 light bulbs in front of you in the same room - one red, one blue, one purple. They are LED and do not get warm when turned on. You want to know which switch turns on which light bulb. What is the best way to determine this?}}
\label{sec:question16}

\noindent\hrulefill 

\noindent\textbf{Human Response - Score 100\%:}
\begin{quote}
A process of elimination. Test each switch independently and observe which light bulb turns on for each.
\end{quote}

\noindent\hrulefill 

\noindent\textbf{D.16 Zero-Shot OpenAI o1-preview Response - Score: 100\%}
\begin{quote}
\begin{lstlisting}[basicstyle=\ttfamily\small, breaklines=true]
To determine which switch controls which of the three light bulbs, you can use the following method:

1. **Turn on Switch A**: Observe which bulb lights up. Note the color of this bulb and associate it with Switch A.
   
2. **Turn off Switch A and Turn on Switch B**: Observe which bulb lights up this time. Note its color and associate it with Switch B.
   
3. **Turn off Switch B and Turn on Switch C**: The remaining bulb should light up. Note its color and associate it with Switch C.

By individually testing each switch and observing which bulb responds, you can accurately determine which switch controls each bulb. This method is straightforward and requires only simple observations, making it the most efficient way to solve the problem given the constraints.

**Answer: By testing each switch individually-turn on each one alone and see which bulb it controls.
\end{lstlisting}
\end{quote}
\noindent\hrulefill 

\noindent\textbf{D.16 REAP OpenAI o1-preview Response - Score: 100\%}
\begin{quote}
\begin{lstlisting}[basicstyle=\ttfamily\small, breaklines=true]
# "REAP Results: :"

**a. Comprehensive Feature List**

- "[You have 3 switches in front of you - A, B and C.]"
- "[You have 3 light bulbs in front of you in the same room - one red, one blue, one purple.]"
- "[They are LED and do not get warm when turned on.]"
- "[You want to know which switch turns on which light bulb.]"

**b. Sequential and Mechanical Process Check**

- No sequential processes are explicitly described.
- No mechanical processes are specified.
- The fact that the bulbs "do not get warm when turned on" eliminates temperature-based methods.

**c. Key Insight Check Results**

- Key Insight: The switches and light bulbs are in the same room and in front of you, allowing direct observation.
- This allows for a straightforward method to determine which switch controls which bulb.

**d. Rephrased Question**

- What is the best method to find out which switch corresponds to each light bulb, using only the explicitly provided information?

**e. Known and Deduced Information**

Explicit Facts:
- "[You have 3 switches in front of you - A, B and C.]"
- "[You have 3 light bulbs in front of you in the same room - one red, one blue, one purple.]"
- "[They are LED and do not get warm when turned on.]"
- "[You want to know which switch turns on which light bulb.]"

**f. Problem Decomposition**

Component: Switches
- Description: Three switches labeled A, B, and C are in front of you.
- Supporting Quote: "[You have 3 switches in front of you - A, B and C.]"

Component: Light Bulbs
- Description: Three light bulbs (red, blue, purple) are in front of you in the same room.
- Supporting Quote: "[You have 3 light bulbs in front of you in the same room - one red, one blue, one purple.]"

Component: Bulb Properties
- Description: The bulbs are LED and do not get warm when turned on.
- Supporting Quote: "[They are LED and do not get warm when turned on.]"

Goal:
- Description: Determine which switch turns on which light bulb.
- Supporting Quote: "[You want to know which switch turns on which light bulb.]"

**g. Graph of Thought**

- Nodes: Switch A, Switch B, Switch C, Red Bulb, Blue Bulb, Purple Bulb.
- No explicit connections; aim is to establish connections between switches and bulbs through observation.
- Direct observation is possible as all items are in the same room.

**h. Spatial and Object Analysis**

- Objects: Three switches (A, B, C) and three light bulbs (red, blue, purple) are physically present in front of you.
- Spatial Relationships: All switches and bulbs are in front of you in the same room.
- Object Properties: Bulbs are LED and remain cool when turned on.

**i. Bayesian Updates**

- No additional information provided to update beliefs or probabilities.

**j. Ethical Check and Decision-Making Under Uncertainty**

- No ethical concerns or significant risks identified.
- Actions involve observing bulbs in the same room with no potential for harm.

**k. Multiple Solutions**

Solution 1:
- Flip Switch A on and observe which bulb lights up.
- Repeat with Switch B and Switch C.
- Supported by: Presence of switches and bulbs in the same room ("[You have 3 switches in front of you - A, B and C.]", "[You have 3 light bulbs in front of you in the same room - one red, one blue, one purple.]")

Solution 2:
- Flip multiple switches in combination and observe the bulbs.
- Not necessary due to the ability to test switches individually.

**l. Quickest and Easiest Solution**

- Flip each switch individually while observing the bulbs to see which one lights up.
- This directly identifies which switch controls which bulb.

**m. Reflection**

- The solution relies solely on explicit information provided.
- No assumptions or inferences beyond the given facts were made.
- The method is straightforward, safe, and directly addresses the goal.
- There are no contradictions or overlooked details.

**Recommendation:**

Based on the explicit information provided, the best way to determine which switch turns on which light bulb is to flip each switch individually while observing which bulb lights up. Since both the switches and the bulbs are in front of you in the same room, you can directly see which bulb corresponds to each switch without any complications or need for additional methods.
\end{lstlisting}
\end{quote}

\noindent\hrulefill 

\noindent\textbf{D.16 Zero-Shot OpenAI o1-mini Response - Score: 100\%}
\begin{quote}
\begin{lstlisting}[basicstyle=\ttfamily\small, breaklines=true]
To determine which switch controls which LED bulb without relying on warmth, you can use a systematic approach by testing each switch individually. Here's a step-by-step method:

1. **Label the Switches and Bulbs:**
   - **Switches:** A, B, C
   - **Bulbs:** Red, Blue, Purple (though color isn't necessary for the identification process, it helps in distinguishing them once identified).

2. **Step-by-Step Testing:**

   - **Test Switch A:**
     1. Turn **only** Switch A **ON**.
     2. Observe which bulb lights up in the room.
     3. Note the color of the lit bulb (e.g., if the red bulb lights up, Switch A controls the red bulb).
     4. Turn Switch A **OFF**.

   - **Test Switch B:**
     1. Turn **only** Switch B **ON**.
     2. Observe which bulb lights up.
     3. Note the color of the lit bulb (e.g., if the blue bulb lights up, Switch B controls the blue bulb).
     4. Turn Switch B **OFF**.

   - **Test Switch C:**
     1. Turn **only** Switch C **ON**.
     2. Observe which bulb lights up.
     3. Note the color of the lit bulb (e.g., if the purple bulb lights up, Switch C controls the purple bulb).
     4. Turn Switch C **OFF**.

3. **Determine the Correspondence:**
   - After testing each switch individually, you will have a clear mapping of which switch controls which bulb based on which bulb lit up when each switch was turned on.

**Example Outcome:**
- **Switch A** -> Red Bulb
- **Switch B** -> Blue Bulb
- **Switch C** -> Purple Bulb

This method ensures that each switch is directly tested against each bulb without any ambiguity, making it the most straightforward and reliable way to determine the correct correspondence between the switches and the LED bulbs.

**Summary:**
Turn each switch on one at a time, observe which LED bulb lights up for each, and match each switch to its corresponding bulb based on these individual tests.

Answer: Turn each switch on one at a time, observe which LED bulb lights up for each, and match each switch to its corresponding bulb based on these individual tests.
\end{lstlisting}
\end{quote}

\noindent\hrulefill 

\noindent\textbf{D.16 REAP OpenAI o1-mini Response - Score: 100\%}
\begin{quote}
\begin{lstlisting}[basicstyle=\ttfamily\small, breaklines=true]
# REAP Results:

**a. Comprehensive Feature List**

- **Objects Mentioned:**
  - "3 switches in front of you - A, B and C."
  - "3 light bulbs in front of you in the same room - one red, one blue, one purple."
  - "They are LED and do not get warm when turned on."

- **Actors Involved:**
  - "You"

- **Actions Described:**
  - "You want to know which switch turns on which light bulb."

- **Directional, Spatial, Relational, and Sequential Phrases:**
  - "in front of you"
  - "in the same room"

- **Stated Conditions or Constraints:**
  - "They are LED and do not get warm when turned on."

- **Numerical Values or Quantities:**
  - "3 switches"
  - "3 light bulbs"

- **Temporal Information:**
  - No explicit temporal information provided.

- **Spatial Information:**
  - "in front of you"
  - "in the same room"

- **Causal Relationships:**
  - None explicitly stated.

- **Stated Goals or Objectives:**
  - "You want to know which switch turns on which light bulb."

**b. Sequential and Mechanical Process Check**

- **Sequence of Events, Steps, or Actions:**
  - The problem requires determining the mapping between switches A, B, C and the light bulbs (red, blue, purple).

- **Influence on Outcome or Future Steps:**
  - Identifying the correct sequence or method to determine the mapping is crucial for solving the problem.

- **Key Questions Assessed:**
  - Does the method to determine the mapping lock in a particular outcome? Yes, choosing the correct method will reveal the precise correspondence between switches and bulbs.

**c. Key Insight Check Results**

- **Potential Key Insight:**
  - "LED and do not get warm when turned on."
  
- **Explanation:**
  - Traditional methods that rely on the warmth of bulbs (e.g., turning on a switch for a period and then checking bulb temperature) are ineffective here since LEDs do not emit significant heat. Therefore, an alternative method must be employed.

**d. Rephrased Question:**

What is the most effective method to accurately determine which of the three switches (A, B, C) controls which of the three LED light bulbs (red, blue, purple) in the same room, given that the LEDs do not produce heat when activated?

**e. Known and Deduced Information**

- **Explicit Facts:**
  - "3 switches in front of you - A, B and C."
    - Explanation: There are three distinct switches labeled A, B, and C.
  - "3 light bulbs in front of you in the same room - one red, one blue, one purple."
    - Explanation: There are three light bulbs of distinct colors (red, blue, purple) in the same room as the switches.
  - "They are LED and do not get warm when turned on."
    - Explanation: The light bulbs are LED types and remain cool when active.
  - "You want to know which switch turns on which light bulb."
    - Explanation: The objective is to map each switch to its corresponding bulb.

- **Feature-Based Deduction:**
  - Deduction: Traditional heat-based methods to identify switch-bulb correspondence are ineffective.
    - Logical Basis: "They are LED and do not get warm when turned on."
    - Based on: "They are LED and do not get warm when turned on."

**f. Problem Decomposition**

- **Component/Subcomponent:** Switches Identification
  - **Description:** Identifying characteristics of switches A, B, and C.
  - **Supporting Quote(s):** "3 switches in front of you - A, B and C."

- **Component/Subcomponent:** Light Bulbs Identification
  - **Description:** Identifying characteristics of the light bulbs (red, blue, purple).
  - **Supporting Quote(s):** "3 light bulbs in front of you in the same room - one red, one blue, one purple."

- **Component/Subcomponent:** LED Property
  - **Description:** Understanding that LEDs do not emit heat when turned on.
  - **Supporting Quote(s):** "They are LED and do not get warm when turned on."

- **Component/Subcomponent:** Objective Clarification
  - **Description:** Determining the mapping between switches and light bulbs.
  - **Supporting Quote(s):** "You want to know which switch turns on which light bulb."

**g. Graph of Thought**

- **Nodes:**
  - Switches (A, B, C)
  - Light Bulbs (red, blue, purple)
  - LED Property (no heat emission)
  - Objective (map switches to bulbs)

- **Edges:**
  - Each switch controls one light bulb.
  - LED Property affects the method used for mapping.
  - Objective requires determining the specific switch-bulb pairs.

- **Pathways:**
  - Understanding LED Properties -> Limitation of Heat-Based Methods -> Alternative Mapping Strategies.

**h. Spatial and Object Analysis**

- **Identify Objects:**
  - Switches A, B, C
  - Light Bulbs: red, blue, purple

- **Spatial Relationships:**
  - Switches and light bulbs are in the same room and in front of the user.

- **Movement and Dynamics:**
  - No movement described; static setup.

- **Constraints:**
  - LED bulbs do not emit heat, affecting traditional identification methods.

- **Visualization:**
  - A room with three switches labeled A, B, C, and three LED light bulbs of different colors positioned in front of the user.

- **Object Properties:**
  - Switches: Ability to turn bulbs on/off
  - Light Bulbs: Colors (red, blue, purple), LED type (no heat emission)

- **Spatial Implications:**
  - All switches and bulbs are accessible and observable in the same physical space.

**i. Bayesian Updates**

- **Initial Belief:** Each switch could correspond to any of the three bulbs.

- **Updated Belief:** Given that LEDs do not emit heat, elimination of heat-based identification influences the need for alternative methods.

**j. Ethical Check and Decision-Making Under Uncertainty**

- **Assess Known Information:** Number of switches and bulbs, LED property, objective to map switches to bulbs.

- **Identify Risks:** Low risk; no significant ethical concerns as the task is a standard problem-solving scenario.

- **Probability Consideration:** Not applicable due to low-risk nature.

- **Risk Aversion Principle:** N/A

- **Ethical and Safety Considerations:** Ensure safe handling of electrical switches and bulbs if physical interaction is involved.

- **Decision Recommendation:** Proceed with a method that safely and effectively maps switches to bulbs without relying on heat emission.

**k. Multiple Solutions**

1. **Solution 1: Sequential Activation and Observation**
   - **Method:**
     - Turn on switch A and observe which bulb lights up.
     - Turn off switch A, turn on switch B, observe the bulb.
     - The remaining bulb corresponds to switch C.
   - **Supporting Information:**
     - "You have 3 switches...You want to know which switch turns on which light bulb."

2. **Solution 2: Simultaneous Activation and Process of Elimination**
   - **Method:**
     - Turn on all three switches simultaneously.
     - Note which bulbs are lit.
     - Turn off one switch at a time to see which bulb goes off first.
     - Map each switch to its corresponding bulb based on the observation.
   - **Supporting Information:**
     - "They are LED and do not get warm when turned on."

3. **Solution 3: Labeling and Independent Testing**
   - **Method:**
     - Label each switch with a unique identifier.
     - Turn on one switch at a time and note which bulb activates.
     - Record the mapping systematically.
   - **Supporting Information:**
     - "You have 3 switches...3 light bulbs...You want to know which switch turns on which light bulb."

**l. Quickest and Easiest Solution**

**Solution 1: Sequential Activation and Observation**

- **Method:**
  1. Turn on switch A and observe which bulb lights up.
  2. Turn off switch A, turn on switch B, and observe the lit bulb.
  3. The remaining bulb corresponds to switch C.

- **Reasoning:**
  - This method directly maps each switch to a bulb through individual activation and observation.
  - It does not rely on additional properties like heat, making it suitable for LED bulbs.

**m. Reflection**

- **Contradictions:** No contradictions identified; all analysis aligns with the explicit information provided.
- **Logical Soundness:** The solutions are logically derived based on the given information.
- **Assumptions:** No assumptions were made beyond the explicit statements; all deductions are directly supported by the provided details.
- **Completeness:** The analysis fully addresses the problem using the REAP method steps.
- **Answer Alignment:** The proposed solutions directly answer the question of determining the switch-bulb mappings based solely on the given information.

**Recommendation:**

Based on the explicit information provided, the simplest and safest method to determine which switch controls which LED light bulb is as follows:

1. **Turn on switch A** and observe which bulb (red, blue, or purple) lights up.
2. **Turn off switch A** and then **turn on switch B**, observing the corresponding bulb.
3. The **remaining bulb** that has not been lit by switches A or B is controlled by **switch C**.

This sequential activation and observation method directly maps each switch to its respective bulb without relying on properties like heat emission, which are not applicable to LED bulbs.

If further confirmation is needed, this method can be repeated or cross-verified by switching the activation order (e.g., starting with switch B first) to ensure accuracy.
\end{lstlisting}
\end{quote}

\noindent\hrulefill 

\noindent\textbf{D.16 Zero-Shot OpenAI GPT-4o Response - Score: 60\%}
\begin{quote}
\begin{lstlisting}[basicstyle=\ttfamily\small, breaklines=true]
To solve this puzzle, you'll need to use a strategy that involves turning the switches on and off in a way that allows you to determine which switch controls which bulb. Here's a method you can use:

1. **Turn on Switch A and leave it on for a minute or so.** Since the bulbs are LED and don't get warm, you can't rely on heat, but you can rely on the brightness. So, during this time, all the bulbs should have been off, and now one of them will turn on.

2. **Turn off Switch A and immediately turn on Switch B.** Then, quickly go to the room with the bulbs.

3. **Observe the bulbs:**
   - The bulb that is lit is controlled by Switch B.
   - The bulb that is off but was previously lit (and then turned off) is controlled by Switch A.
   - The bulb that never lit is controlled by Switch C.

This way, by using the on/off status and timing, you can identify which switch corresponds to which bulb.
\end{lstlisting}
\end{quote}
\noindent\hrulefill

\noindent\textbf{D.16 REAP OpenAI GPT-4o Response - Score: 100\%}
\begin{quote}
\begin{lstlisting}[basicstyle=\ttfamily\small, breaklines=true]
# REAP Results:

### a. **Comprehensive Feature List** (with explicit quotes):
1. "3 switches in front of you - A, B and C."
   - Objects: "3 switches" labeled "A," "B," and "C."
2. "3 light bulbs in front of you in the same room - one red, one blue, one purple."
   - Objects: "3 light bulbs" colored "red," "blue," and "purple."
3. "They are LED and do not get warm when turned on."
   - Feature: "LED" and "do not get warm when turned on."
4. "You want to know which switch turns on which light bulb."
   - Objective: Identify the relationship between "switches" and "light bulbs."

### b. **Sequential and Mechanical Process Check**
- The problem involves a sequence of turning on switches to determine which one controls which light bulb.
- The process involves observing which light bulb illuminates when a specific switch is toggled.
- Since the bulbs do not get warm, temperature cannot be used to infer which bulb was recently on.

### c. **Key Insight Check Results**
- Key Insight: The bulbs being LED and not getting warm removes the possibility of using warmth to identify which bulb corresponds to which switch. Therefore, the solution must rely on direct observation.

### d. **Rephrased Question:** 
How can you determine which switch controls each light bulb using only the explicit actions of turning switches on or off?

### e. **Known and Deduced Information**
1. "[3 switches in front of you - A, B and C]" - There are three switches that need to be tested.
2. "[3 light bulbs in front of you in the same room - one red, one blue, one purple]" - There are three light bulbs, each with a different color.
3. "[They are LED and do not get warm when turned on]" - You cannot rely on heat to identify which bulb corresponds to which switch.
4. "[You want to know which switch turns on which light bulb]" - The goal is to match each switch to a specific light bulb.

### f. **Problem Decomposition**
- **Component 1: Switch Testing**
  - Supporting Quote: "[3 switches in front of you - A, B and C]"
  - Task: Each switch will be toggled to observe the corresponding bulb.
- **Component 2: Light Bulb Observation**
  - Supporting Quote: "[3 light bulbs in front of you in the same room - one red, one blue, one purple]"
  - Task: Identify which bulb lights up when a specific switch is toggled.
- **Component 3: LED and No Warmth Feature**
  - Supporting Quote: "[They are LED and do not get warm when turned on]"
  - Task: Cannot use warmth to identify the bulbs, must rely on visual observation.
- **Component 4: Goal**
  - Supporting Quote: "[You want to know which switch turns on which light bulb]"
  - Task: Determine the correct switch-bulb mapping.

### g. **Graph of Thought** (describe connections & insights)
- **Nodes:**
  - "Switch A"
  - "Switch B"
  - "Switch C"
  - "Red Bulb"
  - "Blue Bulb"
  - "Purple Bulb"
- **Edges (connections based on observation):**
  - Toggle Switch A -> Observe which bulb lights up.
  - Toggle Switch B -> Observe which bulb lights up.
  - Toggle Switch C -> Observe which bulb lights up.
- **Insights:**
  - Each switch is connected to a specific bulb, observable directly by toggling the switch.

### h. **Spatial and Object Analysis**
- Objects: 3 switches and 3 light bulbs are in the same room.
- Spatial Relationship: The switches and bulbs are in the same room, implying direct visibility of the bulbs when a switch is toggled.
- Movement/Dynamics: The primary movement involves toggling switches, and the outcome is the illumination of one bulb.
- Constraints: LED bulbs provide no heat, so the only observable outcome is the light itself.

### i. **Bayesian Updates**
- No Bayesian updates are needed, as the problem involves direct observation with no probabilistic elements.

### j. **Ethical Check and Decision-Making Under Uncertainty**
- No ethical concerns are present in this problem as it involves a simple test of switch and bulb connections.

### k. **Multiple Solutions**
1. **Solution 1:** Toggle one switch (e.g., Switch A) and observe which bulb lights up. Toggle another switch (e.g., Switch B) and observe the corresponding bulb. The remaining switch must control the last bulb.
   - Supporting Information: Direct observation of bulb lighting.
2. **Solution 2:** Turn on Switch A and Switch B simultaneously, observe which two bulbs light up, then toggle Switch A off to identify which bulb corresponds to Switch B.
   - Supporting Information: Direct observation of changes in bulb lighting.
3. **Solution 3:** Toggle each switch one by one, observing the bulb each time. Create a mapping based on these observations.
   - Supporting Information: Sequential testing of each switch.

### l. **Quickest and Easiest Solution**
- **Solution:** Toggle each switch one by one, observe which bulb lights up, and match them accordingly. This is the simplest method that directly addresses the problem.
  - Supporting Information: The problem allows for direct observation, making this approach straightforward and effective.

### m. **Reflection**
- The analysis stuck strictly to the explicit information provided. The problem does not involve any complex mechanics or ethical considerations, making the solution straightforward. No assumptions were made, and the solution relies solely on direct observation as allowed by the problem's parameters.

**Recommendation:**
- The final recommendation is to toggle each switch one by one, observe which bulb lights up, and match each switch to its corresponding bulb based on this direct observation. This method is simple, requires no assumptions, and directly answers the question using only the given information.
\end{lstlisting}
\end{quote}
\noindent\hrulefill

\noindent\textbf{D.16 Zero-Shot OpenAI GPT-4o-mini Response - Score: 60\%}
\begin{quote}
\begin{lstlisting}[basicstyle=\ttfamily\small, breaklines=true]
To determine which switch corresponds to which light bulb, you can use a process of elimination and observation. Here's a step-by-step method:

1. **Turn on Switch A and leave it on for a while (say, 10 minutes).** This is to ensure that the bulb connected to this switch has had enough time to light up if it is a regular bulb (but as you mentioned the bulbs are LED and do not get warm, this step will mainly help with identifying the bulbs).

2. **After the 10 minutes have passed, turn off Switch A and turn on Switch B.** Leave Switch B on.

3. **Immediately go into the room with the bulbs and observe the following:**

   - The bulb that is currently lit is connected to Switch B (the one you left on).
   - The bulb that is off but warm to the touch (since LEDs don't get warm, this might be less relevant here but helps if the bulbs were incandescent) is connected to Switch A.
   - The bulb that is off and cool is connected to Switch C (the switch that was never turned on).

In summary:
- The lit bulb is connected to Switch B.
- The warm (or just off) bulb (assuming a heating factor was possible) is connected to Switch A.
- The cool and off bulb is connected to Switch C.

This method efficiently allows you to match each switch to its corresponding bulb by using a combination of light and touch (if the bulbs were not LED and had a warming characteristic).
\end{lstlisting}
\end{quote}
\noindent\hrulefill

\noindent\textbf{D.16 REAP OpenAI GPT-4o-mini Response - Score: 100\%}
\begin{quote}
\begin{lstlisting}[basicstyle=\ttfamily\small, breaklines=true]
# REAP Results:

## a. Comprehensive Feature List

1. **Objects Mentioned:**
   - "3 switches": Switches are labeled A, B, and C.
   - "3 light bulbs": Light bulbs are described as one red, one blue, and one purple.

2. **Actors/Entities Involved:**
   - Switches: A, B, and C.
   - Light bulbs: Red, blue, and purple.

3. **Actions Described:**
   - "Turn on" is mentioned as the action to identify which switch controls which light bulb.

4. **Directional, Spatial, Relational, and Sequential Phrases:**
   - The switches and light bulbs are in the same room.
   - The specific switch-light bulb relationships are not explicitly stated.

5. **Stated Conditions or Constraints:**
   - Light bulbs are LED and do not get warm when turned on.

6. **Numerical Values or Quantities Mentioned:**
   - 3 switches.
   - 3 light bulbs.

7. **Temporal Information:**
   - None specified.

8. **Spatial Information:**
   - The switches and light bulbs are in the same room.

9. **Causal Relationships:**
   - None explicitly stated.

10. **Goals or Objectives:**
    - Determine which switch controls which light bulb.

## b. Sequential and Mechanical Process Check

1. **Sequence of Events:**
   - You need to determine which switch turns on which light bulb.
   - Actions to be considered: Turning switches on and identifying which bulb lights up.

2. **Mechanical Process:**
   - You have three switches, and each switch controls one light bulb.
   - The process involves trial and error to link each switch to a bulb.

3. **Key Questions:**
   - Does the sequence of turning switches on reveal which bulb is controlled by which switch?
   - Since the bulbs are LED and do not warm up, you cannot use heat to identify the switches.

## c. Key Insight Check Results

**Key Insight:**
- The fact that the bulbs do not warm up means you cannot use temperature changes to identify which switch controls which bulb. This implies you need an alternative method, such as sequential testing.

## d. Rephrased Question

**Rephrased Question:**
- How can you determine which switch controls each light bulb given that the light bulbs are LED and do not warm up?

## e. Known and Deduced Information

1. **Explicit Facts:**
   - "You have 3 switches in front of you - A, B, and C."
   - "You have 3 light bulbs in front of you in the same room - one red, one blue, one purple."
   - "They are LED and do not get warm when turned on."
   - "You want to know which switch turns on which light bulb."

2. **Feature-Based Deduction:**
   - Deduction: You must use a method other than feeling for heat, such as observing light, to determine which switch controls each bulb.
     - Logical Basis: The bulbs do not get warm, so traditional heat-based methods are ineffective.
     - Based on: "They are LED and do not get warm when turned on."

## f. Problem Decomposition

1. **Component/Subcomponent:**
   - **Switches:** A, B, C.
   - **Light Bulbs:** Red, Blue, Purple.

2. **Supporting Quote(s):**
   - "You have 3 switches in front of you - A, B, and C."
   - "You have 3 light bulbs in front of you in the same room - one red, one blue, one purple."

3. **Deduction (if applicable):**
   - Deduction: To identify the switch-bulb pairings, you need to test the switches and observe the bulbs.

## g. Graph of Thought

1. **Key Concepts:**
   - Switches: A, B, C.
   - Light Bulbs: Red, Blue, Purple.

2. **Establish Relationships:**
   - Each switch controls exactly one light bulb.

3. **Construct the Graph:**
   - **Nodes:** Switches (A, B, C), Light Bulbs (Red, Blue, Purple).
   - **Edges:** Possible connections between switches and bulbs.

4. **Explore Pathways:**
   - Testing each switch one at a time and noting which bulb lights up will directly identify which switch controls which bulb.

## h. Spatial and Object Analysis

1. **Objects Identified:**
   - Switches: A, B, C.
   - Light Bulbs: Red, Blue, Purple.

2. **Spatial Relationships:**
   - All objects are in the same room.

3. **Movement and Dynamics:**
   - You will turn on each switch to observe its effect on the bulbs.

4. **Constraints:**
   - LED bulbs do not get warm.

## i. Bayesian Updates

1. **Update Beliefs:**
   - No new information beyond what is explicitly stated.

## j. Ethical Check and Decision-Making Under Uncertainty

1. **Risks:**
   - There are no significant risks or ethical concerns in this problem.

2. **Decision Recommendation:**
   - Use a method involving sequentially testing each switch and observing which bulb lights up to determine the correct switch-bulb pairing.

## k. Multiple Solutions

1. **Solution Exploration:**
   - **Solution 1:** Turn on Switch A, observe which bulb lights up.
   - **Solution 2:** Turn on Switch B, observe which bulb lights up.
   - **Solution 3:** Turn on Switch C, observe which bulb lights up.

   Each bulb's response to the switches will identify the corresponding switch for each bulb.

2. **Support:**
   - "Turn on" each switch and observe which bulb lights up provides a direct method to solve the problem.

## l. Quickest and Easiest Solution

1. **Solution:**
   - Sequentially turn on each switch one at a time and observe which light bulb lights up.

2. **Reasoning:**
   - This method directly identifies which switch controls which bulb without needing any additional equipment or complex procedures.

## m. Reflection

1. **Review:**
   - The analysis correctly identifies the need for sequential testing to solve the problem.
   - No contradictions or assumptions beyond the explicit problem statement.

2. **Final Recommendation:**
   - The simplest and safest approach, based on the explicit information, is to turn on each switch one at a time and observe which light bulb lights up. This approach provides a clear answer to which switch controls which bulb without assuming any additional information.

**Recommendation:**
- Turn on each switch individually and observe which bulb lights up to determine which switch controls which light bulb. This method is straightforward and adheres to the given information about the LED bulbs.
\end{lstlisting}
\end{quote}
\noindent\hrulefill

\noindent\textbf{D.16 Zero-Shot Google Gemini 1.5 Pro Response - Score: 0\%}
\begin{quote}
\begin{lstlisting}[basicstyle=\ttfamily\small, breaklines=true]
Here's the best way to determine which switch controls which light bulb:

1. **Turn on switch A and leave it on for a few minutes.** This will allow one of the bulbs to warm up slightly, even though they are LEDs. 
2. **Turn off switch A and immediately turn on switch B.** 
3. **Go to the light bulbs.** 

Now, check each bulb:

* **If a bulb is on:** It's connected to switch B.
* **If a bulb is off but warm:** It's connected to switch A.
* **If a bulb is off and cold:** It's connected to switch C (the one you didn't touch).

**This method works because:**

* Even LEDs produce a tiny bit of heat when on for an extended period. 
* By quickly switching from A to B, you can identify the bulb connected to A by its residual warmth. 
* The bulb that's currently lit is obviously connected to B, and the remaining cold and off bulb is connected to C.
\end{lstlisting}
\end{quote}
\noindent\hrulefill

\noindent\textbf{D.16 REAP Google Gemini 1.5 Pro Response - Score: 100\%}
\begin{quote}
\begin{lstlisting}[basicstyle=\ttfamily\small, breaklines=true]
# REAP Results:

a. **Comprehensive Feature List** (with explicit quotes)

* "3 switches in front of you - A, B and C."
* "3 light bulbs in front of you in the same room - one red, one blue, one purple." 
* "They are LED and do not get warm when turned on." 
* "You want to know which switch turns on which light bulb." 

b. **Sequential and Mechanical Process Check**

* There is an implied sequence: flipping switches and observing light bulbs.
* The mechanical process involves the switches controlling the light bulbs.
* The LED nature of the bulbs (not getting warm) is explicitly stated but doesn't directly impact the core sequence of the problem.

c. **Key Insight Check Results**

* No single feature provides an immediate solution.
* The problem requires a methodical approach to map switches to bulbs

d. **Rephrased Question:** (The core question in its simplest form)

How can you figure out which switch controls which light bulb?

e. **Known and Deduced Information**

* Explicit Facts:
    * "3 switches in front of you - A, B and C." 
    * "3 light bulbs in front of you in the same room - one red, one blue, one purple." 
    * "They are LED and do not get warm when turned on." 
    * "You want to know which switch turns on which light bulb." 

* Feature-Based Deduction:
    * Each switch is assumed to control one and only one bulb (implied, not explicitly stated).
    * Each bulb is assumed to be controlled by one and only one switch (implied, not explicitly stated).

f. **Problem Decomposition**

* Component/Subcomponent: Switches (A, B, C)
    * Supporting Quote(s): "3 switches in front of you - A, B and C"
* Component/Subcomponent: Light bulbs (red, blue, purple)
    * Supporting Quote(s): "3 light bulbs in front of you in the same room - one red, one blue, one purple"
* Component/Subcomponent: The need to map switches to bulbs
    * Supporting Quote(s): "You want to know which switch turns on which light bulb."
* Component/Subcomponent: LED nature of bulbs (not relevant to core problem)
    * Supporting Quote(s): "They are LED and do not get warm when turned on."

g. **Graph of Thought** (describe connections & insights)

* Key Concepts: Switches (A, B, C), Light Bulbs (red, blue, purple), Mapping
* Relationships:
    * Each switch controls one bulb (implied)
    * Each bulb is controlled by one switch (implied)
    * The goal is to find the specific mapping between switches and bulbs
* Insights:
    * Need a strategy to systematically test switch-bulb combinations
    * The LED nature of the bulbs doesn't offer any direct help in solving the problem

h. **Spatial and Object Analysis**

* Objects: 3 switches (A, B, C), 3 light bulbs (red, blue, purple)
* Spatial Relationships: All objects are "in front of you in the same room." No other spatial information is given
* Movement and Dynamics: Switches can be flipped, causing bulbs to turn on
* Constraints: No explicit constraints mentioned
* Object Properties: Bulbs are LED (don't get warm), colors are red, blue, purple
* Spatial Implications: The spatial arrangement doesn't seem to impact the solution

i. **Bayesian Updates** 

* Initial Belief: Each switch has an equal probability of controlling each bulb.
* As switches are flipped and bulbs observed, we update the probabilities based on the new information

j. **Ethical Check and Decision-Making Under Uncertainty**

* No ethical concerns in this scenario
* The problem involves uncertainty (initially unknown mapping) but no severe risks

k. **Multiple Solutions**

* Solution 1: 
    * Flip one switch at a time, observing which bulb turns on. This directly maps each switch to its corresponding bulb
* Solution 2:
    * Flip two switches on, then turn one off. This helps identify the bulb controlled by the switch that remained on
* Solution 3:
    * Flip all three switches on, then systematically turn them off one by one to identify the corresponding bulbs

l. **Quickest and Easiest Solution**

* The quickest and easiest solution is **Solution 1: Flip one switch at a time, observing which bulb turns on**. This directly and unambiguously maps each switch to its corresponding bulb

m. **Reflection**

* The analysis directly addresses the question of how to determine which switch controls which bulb
* No assumptions were made beyond the implied one-to-one mapping between switches and bulbs
* The LED nature of the bulbs was acknowledged but deemed irrelevant to the core problem
* The spatial arrangement was noted but didn't impact the solution
* Multiple solutions were generated, with the simplest one recommended

**Recommendation:**

The recommended solution is to **flip one switch at a time, observing which bulb turns on**. This is the most straightforward and foolproof method to establish the mapping between switches and bulbs.
\end{lstlisting}
\end{quote}
\noindent\hrulefill

\noindent\textbf{D.16 Zero-Shot Claude 3.5 Sonnet Response - Score: 0\%}
\begin{quote}
\begin{lstlisting}[basicstyle=\ttfamily\small, breaklines=true]
To determine which switch controls which light bulb, you can use a simple process of elimination. Here's an efficient method:

1. Start with all switches in the off position and all bulbs off.

2. Turn on switch A and wait for a moment.

3. Turn off switch A and immediately turn on switch B.

4. Enter the room with the light bulbs.

5. Observe the state of the bulbs:

   - The bulb that is currently on corresponds to switch B.
   - The bulb that is off but was recently on (if you can tell) corresponds to switch A.
   - The bulb that has remained off the entire time corresponds to switch C.

This method works because LED bulbs don't get warm, so we can't use the heat method that's often suggested for incandescent bulbs. Instead, we're using the fact that LED bulbs turn on and off instantly.

If you can't distinguish which bulb was recently on (step 5's second point), you can modify the approach slightly:

1. Turn on switch A and leave it on.
2. Turn on switch B as well.
3. Enter the room.
4. The bulb that's off corresponds to switch C.
5. Turn off switch B. The bulb that turns off corresponds to switch B.
6. The bulb that remains on corresponds to switch A.

This alternative method ensures you can definitively match each switch to its corresponding bulb.
\end{lstlisting}
\end{quote}
\noindent\hrulefill

\noindent\textbf{D.16 REAP Claude 3.5 Sonnet Response - Score: 100\%}
\begin{quote}
\begin{lstlisting}[basicstyle=\ttfamily\small, breaklines=true]
# REAP Results:

a. Comprehensive Feature List (with explicit quotes):
- "3 switches in front of you - A, B and C"
- "3 light bulbs in front of you in the same room - one red, one blue, one purple"
- "They are LED and do not get warm when turned on"
- "You want to know which switch turns on which light bulb"

b. Sequential and Mechanical Process Check:
No explicit sequential or mechanical process is described in the problem statement.

c. Key Insight Check Results:
The key insight is that the LED bulbs do not get warm when turned on, which eliminates a common solution method for incandescent bulbs.

d. Rephrased Question: 
How can you determine which switch controls which LED light bulb?

e. Known and Deduced Information:
Explicit Facts:
"3 switches in front of you - A, B and C" - There are three distinct switches labeled A, B, and C.
"3 light bulbs in front of you in the same room - one red, one blue, one purple" - There are three LED light bulbs of different colors in the same room as the switches.
"They are LED and do not get warm when turned on" - The bulbs are LED and do not generate heat when activated.
"You want to know which switch turns on which light bulb" - The goal is to determine the correlation between switches and light bulbs.

Feature-Based Deduction:
Deduction: Each switch must control one light bulb.
Logical Basis: There are three switches and three light bulbs, and we are asked to determine which switch controls which bulb, implying a one-to-one correspondence.
Based on: "3 switches in front of you - A, B and C", "3 light bulbs in front of you in the same room - one red, one blue, one purple", "You want to know which switch turns on which light bulb"

f. Problem Decomposition:
Component: Switches
Supporting Quote: "3 switches in front of you - A, B and C"

Component: Light bulbs
Supporting Quote: "3 light bulbs in front of you in the same room - one red, one blue, one purple"

Component: LED property
Supporting Quote: "They are LED and do not get warm when turned on"

Component: Goal
Supporting Quote: "You want to know which switch turns on which light bulb"

g. Graph of Thought:
Nodes:
1. Switch A
2. Switch B
3. Switch C
4. Red LED bulb
5. Blue LED bulb
6. Purple LED bulb

Edges:
- Each switch (A, B, C) has a potential connection to each light bulb (Red, Blue, Purple)
- The goal is to determine which connections actually exist

h. Spatial and Object Analysis:
Objects:
1. Three switches (A, B, C)
2. Three LED light bulbs (Red, Blue, Purple)

Spatial Relationships:
- The switches are "in front of you"
- The light bulbs are "in front of you in the same room"

Object Properties:
- The light bulbs are LED and do not get warm when turned on

i. Bayesian Updates:
No new information is provided beyond the initial problem statement, so no Bayesian updates are necessary.

j. Ethical Check and Decision-Making Under Uncertainty:
There are no apparent ethical concerns or severe risks associated with this problem. The task involves manipulating switches and observing light bulbs, which does not pose any significant safety risks based on the information provided.

k. Multiple Solutions:
1. Turn on one switch at a time and observe which light bulb illuminates.
   Supporting information: "3 switches in front of you - A, B and C", "3 light bulbs in front of you in the same room - one red, one blue, one purple", "You want to know which switch turns on which light bulb"

2. Turn on two switches and observe which two light bulbs illuminate. The remaining switch must control the unlit bulb.
   Supporting information: Same as above.

3. Turn on all three switches and observe all three light bulbs illuminated, then turn them off one by one to see which bulb goes out.
   Supporting information: Same as above.

l. Quickest and Easiest Solution:
The quickest and easiest solution is to turn on one switch at a time and observe which light bulb illuminates. This method allows for a direct, one-to-one mapping of switches to light bulbs with minimal steps.

Reasoning: This solution is based on the explicit information given: "3 switches in front of you - A, B and C", "3 light bulbs in front of you in the same room - one red, one blue, one purple", and "You want to know which switch turns on which light bulb". By testing each switch individually, you can directly observe the relationship between each switch and its corresponding light bulb.

m. Reflection:
The analysis adheres strictly to the information provided in the problem statement. No assumptions were made beyond the given information. The solutions proposed directly address the question of determining which switch controls which light bulb. The fact that the bulbs are LED and do not warm up when turned on was considered, eliminating potential solutions that might rely on heat.

Recommendation:
Based on the given information and analysis, the recommended solution is to turn on one switch at a time and observe which light bulb illuminates. This method is the quickest and easiest way to determine which switch controls which light bulb. Here's the step-by-step process:

1. Turn on Switch A and observe which light bulb (if any) illuminates. Note the color.
2. Turn off Switch A.
3. Repeat steps 1-2 with Switch B and Switch C.

This approach directly answers the question asked in the problem while being simple, safe, and requiring no assumptions beyond the explicitly stated information. It also accounts for the possibility that a switch might not control any bulb, though this scenario is unlikely given the one-to-one correspondence implied by the problem setup.
\end{lstlisting}
\end{quote}
\noindent\hrulefill

\label{subsec:question17}
\subsection{\textbf{\large Question 17: The Poisoned Wine - A king has 1000 sweet bottles of wine, and one contains a very bitter poison. The poison takes effect exactly 24 hours after consumption. The king needs to find the poisoned bottle in 24 hours for an event. He has 10 prisoners to test the wine. What is the easiest way for him to identify the poisoned bottle?}}
\label{sec:question17}

\noindent\hrulefill 

\noindent\textbf{Human Response - Score 100\%:}
\begin{quote}
Divide the 1000 bottles of wine amongst the 10 prisoners - each receiving 100 bottles. Ask the prisoners to note which bottle tastes very bitter, this is the poisoned one.
\end{quote}

\noindent\hrulefill 

\noindent\textbf{D.17 Zero-Shot OpenAI o1-preview Response - Score: 0\%}
\begin{quote}
\begin{lstlisting}[basicstyle=\ttfamily\small, breaklines=true]
To solve this problem efficiently, the king can use a binary encoding approach with his 10 prisoners to identify the poisoned bottle among the 1000.

**Step 1: Number the Bottles**
- Label each of the 1000 wine bottles with numbers from **0** to **999**.

**Step 2: Convert Bottle Numbers to Binary**
- Convert each bottle number to its 10-bit binary representation (since \(2^{10} = 1024\), which covers all 1000 bottles).
- For example, bottle number **843** in binary is **1101000011**.

**Step 3: Assign Drinking Duties to Prisoners**
- Number the prisoners from **1** (representing the least significant bit) to **10** (the most significant bit).
- For each bottle:
  - If a bit in the bottle's binary number is **1**, the corresponding prisoner drinks from that bottle.
  - If the bit is **0**, the prisoner does not drink from that bottle.
- Continuing with bottle **843 (1101000011)**:
  - Prisoners **1, 2, 4, 8, 9,** and **10** (corresponding to bits that are **1**) will drink from this bottle.

**Step 4: Wait for 24 Hours and Observe**
- After 24 hours, some prisoners may die depending on whether they consumed the poison.
- The combination of the prisoners who die corresponds to the binary number of the poisoned bottle.

**Step 5: Determine the Poisoned Bottle**
- Convert the observed pattern of dead (1) and alive (0) prisoners back into a binary number.
- The dead prisoners represent **1s**, and the alive prisoners represent **0s** in the binary number.
- The resulting binary number corresponds to the poisoned bottle.

**Example:**
- If prisoners **1, 2, 4, 8, 9,** and **10** die, and prisoners **3, 5, 6,** and **7** survive, the binary number is **1101000011**, which is **843** in decimal.
- Therefore, bottle **843** is the poisoned one.

**Conclusion:**
By using this method, the king efficiently narrows down the poisoned bottle within 24 hours using the minimum number of prisoners and without needing to test each bottle individually.

**Answer: Assign each bottle a unique 10-bit code (0-999 in binary); have each prisoner sip all bottles where their bit is "1"; then, after 24 hours, the dead prisoners' numbers form the poisoned bottle's binary code-identifying it quickly and easily.
\end{lstlisting}
\end{quote}
\noindent\hrulefill 

\noindent\textbf{D.17 REAP OpenAI o1-preview Response - Score: 0\%}
\begin{quote}
\begin{lstlisting}[basicstyle=\ttfamily\small, breaklines=true]
# REAP Results:

**a. Comprehensive Feature List (with explicit quotes):**

1. **Objects Mentioned:**
   - "A king"
   - "1000 sweet bottles of wine"
   - "one contains a very bitter poison"
   - "the poison"
   - "10 prisoners"

2. **Actors or Entities:**
   - "A king"
   - "He"
   - "10 prisoners"

3. **Actions Described:**
   - "one contains a very bitter poison"
   - "The poison takes effect exactly 24 hours after consumption"
   - "The king needs to find the poisoned bottle in 24 hours for an event"
   - "He has 10 prisoners to test the wine"

4. **Directional, Spatial, Relational, and Sequential Phrases:**
   - "exactly 24 hours after consumption"
   - "The king needs to find the poisoned bottle in 24 hours for an event"
   - "He has 10 prisoners to test the wine"

5. **Stated Conditions or Constraints:**
   - "one contains a very bitter poison"
   - "The poison takes effect exactly 24 hours after consumption"
   - "The king needs to find the poisoned bottle in 24 hours"
   - "He has 10 prisoners"

6. **Numerical Values or Quantities:**
   - "1000 sweet bottles of wine"
   - "one contains a very bitter poison"
   - "exactly 24 hours"
   - "24 hours"
   - "10 prisoners"

7. **Temporal Information:**
   - "The poison takes effect exactly 24 hours after consumption"
   - "The king needs to find the poisoned bottle in 24 hours for an event"

8. **Spatial Information:**
   - *(No explicit spatial information provided)*

9. **Causal Relationships Mentioned:**
   - "The poison takes effect exactly 24 hours after consumption"

10. **Stated Goals or Objectives:**
    - "The king needs to find the poisoned bottle in 24 hours for an event"
    - "What is the easiest way for him to identify the poisoned bottle?"

**b. Sequential and Mechanical Process Check:**

- **Sequence of Events:**
  1. There are "1000 sweet bottles of wine" and "one contains a very bitter poison."
  2. "The king needs to find the poisoned bottle in 24 hours for an event."
  3. "He has 10 prisoners to test the wine."
  4. "The poison takes effect exactly 24 hours after consumption."

- **Analysis of Sequence Influence:**
  - The prisoners can consume wine, and exactly 24 hours later, symptoms (effects of the poison) will manifest.
  - The timing allows the king to observe the prisoners' conditions at the 24-hour mark to identify the poisoned bottle before the event.

- **Key Questions:**
  - Does the sequence of prisoners consuming wine and the poison taking effect lock in a particular outcome?
    - Yes, observing which prisoners are affected after 24 hours can help identify the poisoned bottle.

**c. Key Insight Check Results:**

- **Key Insight:** By assigning each prisoner to taste specific combinations of the wine bottles, the king can use the pattern of which prisoners are affected to determine exactly which bottle is poisoned.
  
- **Explanation:** Using a binary numbering system with 10 prisoners (since \(2^{10} = 1024\) which is greater than 1000), each bottle can be assigned a unique binary number from 1 to 1000. Each prisoner represents a binary digit (bit). The presence or absence of poisoning in each prisoner corresponds to a '1' or '0' in the binary representation, allowing the king to identify the poisoned bottle.

**d. Rephrased Question:**

- How can the king use his 10 prisoners to identify the one poisoned bottle among 1000 bottles within 24 hours?

**e. Known and Deduced Information:**

1. **Explicit Facts:**

   - "A king has 1000 sweet bottles of wine"
     - There are exactly 1000 bottles of wine.

   - "one contains a very bitter poison"
     - Exactly one bottle is poisoned.

   - "The poison takes effect exactly 24 hours after consumption"
     - The effect of the poison manifests precisely 24 hours after ingestion.

   - "The king needs to find the poisoned bottle in 24 hours for an event"
     - Time constraint of 24 hours to identify the poisoned bottle.

   - "He has 10 prisoners to test the wine"
     - The king has 10 prisoners available for testing.

   - "What is the easiest way for him to identify the poisoned bottle?"
     - The goal is to find the simplest method to identify the poisoned bottle using the resources available.

2. **Feature-Based Deduction:**

   - **Deduction:** With 10 prisoners, the king can test up to \(2^{10} = 1024\) combinations.
     - **Logical Basis:** Each prisoner can represent a binary digit (bit) in a 10-bit binary number.
     - **Based on:** 
       - "1000 sweet bottles of wine"
       - "He has 10 prisoners to test the wine"

   - **Deduction:** By observing which prisoners are poisoned after 24 hours, the king can determine the binary representation of the poisoned bottle.
     - **Logical Basis:** The presence of poison in a prisoner indicates a '1' in the corresponding binary digit.
     - **Based on:**
       - "The poison takes effect exactly 24 hours after consumption"
       - "He has 10 prisoners to test the wine"

**f. Problem Decomposition:**

1. **Component:** Bottles of Wine
   - **Description:** 1000 sweet bottles of wine, one poisoned.
   - **Supporting Quote:** "A king has 1000 sweet bottles of wine" and "one contains a very bitter poison"

2. **Component:** Poison Effect
   - **Description:** Poison takes effect exactly 24 hours after consumption.
   - **Supporting Quote:** "The poison takes effect exactly 24 hours after consumption"

3. **Component:** Time Constraint
   - **Description:** The king needs to find the poisoned bottle in 24 hours for an event.
   - **Supporting Quote:** "The king needs to find the poisoned bottle in 24 hours for an event"

4. **Component:** Prisoners
   - **Description:** 10 prisoners are available to test the wine.
   - **Supporting Quote:** "He has 10 prisoners to test the wine"

5. **Interaction Between Components:**
   - **Description:** Prisoners consume wine samples to test for poison.
   - **Deduction:** Prisoners can be used in combinations to test multiple bottles.
   - **Based on:** The need to identify the poisoned bottle using available resources within the time constraint.

**g. Graph of Thought (describe connections & insights):**

- **Nodes:**
  - **King**
  - **1000 Bottles of Wine**
  - **Poisoned Bottle**
  - **10 Prisoners**
  - **24-hour Time Constraint**
  - **Poison Effect after 24 Hours**

- **Edges (Relationships):**
  - **King** needs to identify the **Poisoned Bottle** before the **Event in 24 Hours**.
  - **10 Prisoners** can be used to test the **1000 Bottles of Wine**.
  - **Poison Effect** manifests **exactly 24 hours after consumption**, aligning with the **Time Constraint**.

- **Insights:**
  - Assigning bottles to prisoners using a binary system allows for efficient identification.
  - The combination of poisoned prisoners after 24 hours corresponds to the binary number of the poisoned bottle.

**h. Spatial and Object Analysis:**

- **Objects:**
  - **1000 Bottles of Wine**
  - **10 Prisoners**

- **Spatial Relationships:**
  - **Not specified in the problem.**

- **Object Properties:**
  - **Bottles:** Identical in appearance except one contains poison.
  - **Prisoners:** Capable of consuming wine and exhibiting effects of poison after 24 hours.

- **Visualization:**
  - Imagine a table with 1000 bottles numbered 1 to 1000.
  - Each prisoner is assigned to specific bottles based on binary representation.

**i. Bayesian Updates:**

- **Initial Belief:**
  - Equal probability for each bottle to be poisoned.

- **Updates:**
  - Based on which prisoners are poisoned, update the belief about which bottle is poisoned.
  - Since the poison effect is certain, the observations after 24 hours provide definitive information.

**j. Ethical Check and Decision-Making Under Uncertainty:**

- **Assess Known Information:**
  - Prisoners may die if they consume the poisoned wine.

- **Identify Risks:**
  - Risk of death or serious harm to prisoners.

- **Ethical Considerations:**
  - Using prisoners as test subjects raises serious ethical concerns.

- **Decision Recommendation:**
  - Seek alternative methods that do not involve risking human life.
  - Consider testing with non-human means if possible.

**k. Multiple Solutions:**

1. **Solution 1: Binary Testing with Prisoners**
   - **Method:** Assign each bottle a unique 10-digit binary number. Assign each prisoner to a digit position. Have prisoners drink from bottles where their corresponding bit is '1'.
   - **Supporting Information:** "He has 10 prisoners to test the wine"; "1000 sweet bottles of wine"

2. **Solution 2: Not Testing and Cancelling the Event**
   - **Method:** Decide not to risk lives and cancel or postpone the event.
   - **Supporting Information:** Ethical concerns over risking prisoners' lives.

3. **Solution 3: Using Animals for Testing**
   - **Method:** Use animals instead of prisoners to test the wine.
   - **Supporting Information:** Seeking alternatives that avoid human harm.

**l. Quickest and Easiest Solution:**

- **Solution:** Use the binary testing method with the 10 prisoners to identify the poisoned bottle within 24 hours.

- **Reasoning:** This method allows the king to determine the exact bottle by observing which prisoners are affected, using only the explicitly stated resources and within the time constraint.

**m. Reflection:**

- **Consistency Check:** The proposed solution aligns with the information provided and adheres to the time constraint.

- **Ethical Consideration:** While effective, the solution involves ethical issues related to risking human life.

- **Assumptions:** The solution assumes prisoners can be utilized in this manner as per the norms of the context implied.

- **Conclusion:** Based on the explicit information and constraints, the binary testing method provides a logical solution.

**Recommendation:**

Based on the explicit information provided, the easiest way for the king to identify the poisoned bottle within 24 hours is to use the 10 prisoners in a binary testing scheme. By assigning each bottle a unique binary number and having prisoners taste wine according to the binary digits, the king can observe which prisoners show the effects of poison after 24 hours. The combination of affected prisoners will correspond to the binary number of the poisoned bottle, allowing precise identification.

While this method effectively uses the resources and meets the time constraint, it is important to acknowledge the ethical implications of risking human lives. If avoiding harm is a priority, the king should consider alternative methods that do not endanger the prisoners, even if it means not identifying the poisoned bottle in time for the event.
\end{lstlisting}
\end{quote}

\noindent\hrulefill 

\noindent\textbf{D.17 Zero-Shot OpenAI o1-mini Response - Score: 0\%}
\begin{quote}
\begin{lstlisting}[basicstyle=\ttfamily\small, breaklines=true]
The king can efficiently identify the poisoned bottle using a binary testing method with the 10 prisoners. Here's how:

1. **Numbering the Bottles:** Assign each of the 1000 bottles a unique number from 0 to 999. Convert each number into its 10-bit binary representation (since \(2^{10} = 1024\), which is sufficient for 1000 bottles).

2. **Assigning to Prisoners:** Label each of the 10 prisoners with one of the 10 binary digits (from the least significant bit to the most significant bit).

3. **Testing Process:**
   - For each bottle, look at its 10-bit binary number.
   - If a bit is `1`, the corresponding prisoner will taste that bottle.
   - Each prisoner may taste multiple bottles based on the bits set to `1` in their assigned positions.

4. **Observation After 24 Hours:** 
   - After 24 hours, observe which prisoners have died.
   - Each deceased prisoner indicates that the corresponding bit in the poisoned bottle's binary number is `1`.
   - Combine the results from all prisoners to reconstruct the 10-bit binary number of the poisoned bottle.

5. **Identifying the Poisoned Bottle:** Convert the 10-bit binary number back to its decimal form to determine the exact number of the poisoned bottle.

**Example:**
- Suppose Bottle 573 is poisoned. Its 10-bit binary representation is `1000110101`.
- Prisoners corresponding to the 1s in this binary number (Prisoners 1, 3, 4, 7, and 10) would die.
- From the pattern of deaths, the king reconstructs the binary `1000110101`, identifying Bottle 573 as the poisoned one.

**Conclusion:** This binary testing approach allows the king to accurately determine the poisoned bottle within the 24-hour window using only 10 prisoners.

**Answer:** Use binary codes for the bottles and have each prisoner taste all bottles with a 1 in their bit. After 24 hours, the prisoners who die indicate the poisoned bottle's binary number. In other words:

- Number the bottles 0-999 in binary using 10 bits.
- Have each prisoner taste every bottle where their bit is 1.
- The pattern of deaths reveals which bottle is poisoned.
\end{lstlisting}
\end{quote}

\noindent\hrulefill 

\noindent\textbf{D.17 REAP OpenAI o1-mini Response - Score: 0\%}
\begin{quote}
\begin{lstlisting}[basicstyle=\ttfamily\small, breaklines=true]
# REAP Results: :

a. **Comprehensive Feature List**
- "a king has 1000 sweet bottles of wine"
- "one contains a very bitter poison."
- "The poison takes effect exactly 24 hours after consumption."
- "The king needs to find the poisoned bottle in 24 hours for an event."
- "He has 10 prisoners to test the wine."
- "What is the easiest way for him to identify the poisoned bottle?"

b. **Sequential and Mechanical Process Check**
- The poison takes exactly 24 hours to take effect.
- Testing must be conducted and concluded within the same 24-hour period.
- The sequence involves prisoners tasting specific wines and observing the outcomes after 24 hours.

c. **Key Insight Check Results**
- Utilizing a binary encoding system with 10 prisoners allows the identification of the poisoned bottle among 1000 bottles within the 24-hour timeframe.

d. **Rephrased Question:**
What is the straightforward method using 10 prisoners to identify the one poisoned bottle among 1000 within 24 hours?

e. **Known and Deduced Information**
- **Explicit Facts:**
  - "a king has 1000 sweet bottles of wine."
  - "one contains a very bitter poison."
  - "The poison takes effect exactly 24 hours after consumption."
  - "The king needs to find the poisoned bottle in 24 hours for an event."
  - "He has 10 prisoners to test the wine."

f. **Problem Decomposition**
- **Component/Subcomponent:** Number of bottles
  - **Supporting Quote(s):** "a king has 1000 sweet bottles of wine", "one contains a very bitter poison."
- **Component/Subcomponent:** Toxicity timing
  - **Supporting Quote(s):** "The poison takes effect exactly 24 hours after consumption."
- **Component/Subcomponent:** Timeline constraint
  - **Supporting Quote(s):** "The king needs to find the poisoned bottle in 24 hours for an event."
- **Component/Subcomponent:** Available resources
  - **Supporting Quote(s):** "He has 10 prisoners to test the wine."
- **Component/Subcomponent:** Objective
  - **Supporting Quote(s):** "What is the easiest way for him to identify the poisoned bottle?"

g. **Graph of Thought**
- **Nodes:** Bottles (1000), Prisoners (10), Poison Effect (24 hours), Testing Method
- **Edges:** 
  - Each prisoner represents a binary digit.
  - Prisoners taste specific combinations of bottles based on binary encoding.
  - Poison effect after 24 hours reveals the binary code of the poisoned bottle.

h. **Spatial and Object Analysis**
- **Objects Identified:** 1000 wine bottles, poison, 10 prisoners.
- **Spatial Relationships:** Not explicitly stated; focus is on logical relationships rather than physical positioning.
- **Movement and Dynamics:** Not applicable as the problem is static within the 24-hour period.
- **Constraints:** Identification must occur within 24 hours; only one poisoned bottle exists.

i. **Bayesian Updates**
- No Bayesian reasoning necessary as all required information is explicitly provided.

j. **Ethical Check and Decision-Making Under Uncertainty**
- **Risks Identified:** Potential harm or death to prisoners.
- **Ethical Considerations:** Using prisoners for testing lethal substances raises significant ethical concerns.
- **Decision-Making:** While the problem requires identification of the poisoned bottle, ethical implications of sacrificing human lives are severe and problematic.

k. **Multiple Solutions**
1. **Binary Encoding Approach:**
   - Assign each bottle a unique 10-bit binary number.
   - Each prisoner is responsible for one bit position.
   - Prisoners taste wines corresponding to their bit position being '1'.
   - The combination of deceased prisoners reveals the binary number of the poisoned bottle.
   - **Supported by:** The ability of 10 bits to represent up to 1024 unique bottles (2^10 = 1024).

2. **Sequential Testing Approach:**
   - (Not feasible within 24 hours due to time constraints.)

3. **Pooling Samples Approach:**
   - Combine samples from multiple bottles for each prisoner.
   - Determine the poisoned bottle based on the combination of positive pools.
   - **Effectiveness:** Similar to binary encoding but less straightforward.

l. **Quickest and Easiest Solution**
Implement the binary encoding method where each of the 10 prisoners is assigned a unique binary digit. Label each bottle with a unique 10-bit binary number. Each prisoner tastes the wines from bottles where their corresponding bit is '1'. After 24 hours, the pattern of which prisoners die will directly correspond to the binary number of the poisoned bottle, thereby identifying it.

m. **Reflection**
The binary encoding solution aligns perfectly with the explicit constraints and resources provided. It efficiently utilizes the 10 prisoners to cover all 1000 bottles within the 24-hour period. Ethical concerns, while significant, are beyond the scope of the problem's requirements but are important considerations in real-world applications.

**Recommendation:**
Adopt the binary encoding strategy to identify the poisoned bottle. This method leverages the 10 prisoners effectively, ensuring that each possible bottle is uniquely testable within the 24-hour timeframe. Label each bottle with a distinct 10-bit binary number and have each prisoner taste the wines corresponding to their assigned bit. After 24 hours, the resulting pattern of affected prisoners will accurately indicate the poisoned bottle. This approach is both the simplest and the most efficient given the problem's constraints.
\end{lstlisting}
\end{quote}

\noindent\hrulefill 

\noindent\textbf{D.17 Zero-Shot OpenAI GPT-4o Response - Score: 0\%}
\begin{quote}
\begin{lstlisting}[basicstyle=\ttfamily\small, breaklines=true]
The problem you're describing is a classic example of a binary search approach using limited resources. The goal is to identify the poisoned bottle among 1000 bottles using 10 prisoners and knowing the poison takes effect exactly 24 hours after consumption.

### Step-by-Step Solution:

1. **Binary Representation**: Each bottle can be assigned a unique number from 1 to 1000. These numbers can be represented in binary form. Since there are 1000 bottles, the highest number is 1000 in decimal, which is 1111101000 in binary form. This number is 10 bits long.

2. **Assigning Tests**: Each of the 10 prisoners represents one bit in the binary representation of the bottle numbers. For each bottle:
   - If the \( n^{th} \) bit of the binary representation of the bottle's number is 1, the \( n^{th} \) prisoner will drink from that bottle.
   - If the bit is 0, the prisoner will not drink from that bottle.

3. **Observation After 24 Hours**:
   - After 24 hours, observe which prisoners are dead. The combination of dead and alive prisoners gives a binary number.
   - The binary number formed by the status (alive = 0, dead = 1) of the prisoners corresponds to the binary representation of the poisoned bottle's number.

### Example:
- If bottle 564 (which is 1000110100 in binary) is poisoned, the following would happen:
  - Prisoners 1, 4, 6, and 9 (corresponding to the positions of the 1s in the binary representation) would drink from this bottle.
  - If those four prisoners die and the others remain alive, the binary pattern would be 1000110100.
  - Convert 1000110100 from binary to decimal to get 564, identifying the poisoned bottle.

### Efficiency:
- This method requires each prisoner to drink from multiple bottles, but it guarantees that you can identify the poisoned bottle within 24 hours.
- By using the 10 prisoners, you can test up to \( 2^{10} = 1024 \) different bottles, which covers the 1000 bottles in this scenario.

This is the most efficient way to identify the poisoned bottle with the given constraints.
\end{lstlisting}
\end{quote}
\noindent\hrulefill

\noindent\textbf{D.17 REAP OpenAI GPT-4o Response - Score: 0\%}
\begin{quote}
\begin{lstlisting}[basicstyle=\ttfamily\small, breaklines=true]
# REAP Results:

### a. **Comprehensive Feature List** (with explicit quotes)
- "The king has 1000 sweet bottles of wine, and one contains a very bitter poison."
- "The poison takes effect exactly 24 hours after consumption."
- "The king needs to find the poisoned bottle in 24 hours for an event."
- "He has 10 prisoners to test the wine."
- "The easiest way for him to identify the poisoned bottle."

### b. **Sequential and Mechanical Process Check**
- **Sequence of Events:** 
  - Wine consumption by prisoners.
  - Waiting period of exactly 24 hours.
  - Observation of effects on prisoners.
  - Identification of the poisoned bottle based on effects.
- **Key mechanical constraint:** The poison takes effect exactly 24 hours after consumption, which limits the process of identifying the poisoned bottle to this time frame.

### c. **Key Insight Check Results**
- **Key Insight:** The king can utilize a binary system to maximize the efficiency of identifying the poisoned bottle using the 10 prisoners within 24 hours. This allows the identification of one specific poisoned bottle out of 1000.

### d. **Rephrased Question:** 
What is the most efficient method for the king to identify the single poisoned bottle out of 1000 using 10 prisoners within 24 hours?

### e. **Known and Deduced Information**
- **Explicit Facts:**
  - "1000 bottles of wine."
  - "One contains a very bitter poison."
  - "Poison takes effect exactly 24 hours after consumption."
  - "10 prisoners."
- **Deductions:**
  - Using binary logic, 10 prisoners can be used to test up to 1024 (2^10) bottles by assigning each bottle a unique binary number, where each bit corresponds to a prisoner.
  - Each prisoner drinks from bottles corresponding to the "1" in their assigned binary position. The combination of prisoners who show symptoms after 24 hours will reveal the poisoned bottle by reconstructing the binary number.

### f. **Problem Decomposition**
- **Component:** 1000 Bottles of Wine
  - **Supporting Quote:** "The king has 1000 sweet bottles of wine."
- **Component:** One Poisoned Bottle
  - **Supporting Quote:** "One contains a very bitter poison."
- **Component:** 10 Prisoners
  - **Supporting Quote:** "He has 10 prisoners to test the wine."
- **Component:** Time Constraint
  - **Supporting Quote:** "The poison takes effect exactly 24 hours after consumption."
- **Interaction:** Prisoners consume specific combinations of wine based on binary assignments, and the poisoned bottle is identified after 24 hours.

### g. **Graph of Thought** (describe connections & insights)
- **Key Concepts:**
  - 1000 Bottles
  - 1 Poisoned Bottle
  - 10 Prisoners
  - 24-Hour Time Constraint
- **Relationships:**
  - Each bottle has a binary number assigned.
  - Each prisoner drinks wine corresponding to a "1" in their binary position.
  - After 24 hours, the set of prisoners showing symptoms correlates to the binary number of the poisoned bottle.

### h. **Spatial and Object Analysis**
- **Objects:**
  - 1000 Bottles of Wine
  - 10 Prisoners
- **Spatial Relationships:**
  - Prisoners are assigned bottles based on binary number positions.
  - Each bottle corresponds to a unique combination of prisoners.
- **Dynamics:** 
  - Prisoners consume from their assigned bottles simultaneously. After 24 hours, the poisoned bottle is identified by the combination of prisoners who show symptoms.

### i. **Bayesian Updates**
- No Bayesian updates are necessary as the solution relies solely on binary logic rather than probabilistic inference.

### j. **Ethical Check and Decision-Making Under Uncertainty**
- **Ethical Consideration:** The prisoners are being used to test for poison, which implies significant risk. The solution is ethically problematic but fits the problem's constraints.
- **Decision:** Given the problem's setup, using binary testing minimizes the number of prisoners affected while still identifying the poisoned bottle within the 24-hour time frame.

### k. **Multiple Solutions**
1. **Binary Testing Solution:**
   - **Support:** Maximizes the use of prisoners, allowing identification of the poisoned bottle within 24 hours.
   - **Based On:** "He has 10 prisoners to test the wine," "1000 bottles," and the binary system of assigning bottles.
  
2. **Grouping Strategy:**
   - **Method:** Divide the bottles into 10 groups, have each prisoner drink from one group. If a prisoner shows symptoms, test the bottles in that group further.
   - **Limitation:** This requires more than 24 hours if the group with poison needs further testing.

### l. **Quickest and Easiest Solution**
- **Binary Testing:** This is the most efficient and straightforward method, as it ensures that the poisoned bottle is identified within 24 hours using the exact number of prisoners available.

### m. **Reflection**
- **Review:** The binary testing method efficiently uses the resources provided within the time constraint. No assumptions beyond the given information were made, and the solution adheres to the problem's requirements.
- **Conclusion:** The solution is logically sound and correctly identifies the poisoned bottle within the time frame using the 10 prisoners available.

**Recommendation:**
- The king should use the binary testing method, where each prisoner is assigned to test bottles corresponding to specific binary numbers. This method will identify the poisoned bottle with certainty within the 24-hour time constraint.
\end{lstlisting}
\end{quote}
\noindent\hrulefill

\noindent\textbf{D.17 Zero-Shot OpenAI GPT-4o-mini Response - Score: 0\%}
\begin{quote}
\begin{lstlisting}[basicstyle=\ttfamily\small, breaklines=true]
To identify the poisoned bottle among the 1000 bottles with only 10 prisoners, you can use a binary representation method. Here's how:

1. **Label the Bottles**: Number the bottles from 0 to 999. This means each bottle can be represented as a unique 10-bit binary number (since \(2^{10} = 1024\) which is more than 1000).

2. **Assign Binary Digits to Prisoners**: Each prisoner will represent one bit position in the 10-bit binary numbers. Label the prisoners from 1 to 10.

3. **Prepare the Wine for Testing**: For each bottle, convert its number into a 10-bit binary number. If a bit in this binary number is 1, then that prisoner will drink from that bottle. For instance, if bottle 5 (binary 0000000101) needs to be tested, prisoners 1 and 3 will drink from it.

4. **Administer the Wine**: After 24 hours, observe which prisoners show symptoms of poisoning. The binary pattern of the poisoned bottles can be determined based on which prisoners are affected.

5. **Determine the Poisoned Bottle**: Convert the pattern of affected prisoners back into a 10-bit binary number. This binary number will directly correspond to the poisoned bottle's number.

### Example

Suppose bottle 123 is poisoned. The binary representation of 123 is 1111011. This corresponds to prisoners 1, 2, 3, 4, 5, 6, and 7 drinking from this bottle.

After 24 hours:
- If prisoners 1, 2, 3, 4, 5, 6, and 7 are affected, you get the binary number 1111011.
- Convert this binary number to decimal to identify the poisoned bottle, which is bottle 123.

By using this binary method, you can efficiently pinpoint the poisoned bottle using only 10 prisoners.
\end{lstlisting}
\end{quote}
\noindent\hrulefill

\noindent\textbf{D.17 REAP OpenAI GPT-4o-mini Response - Score: 0\%}
\begin{quote}
\begin{lstlisting}[basicstyle=\ttfamily\small, breaklines=true]
# REAP Results:

**a. Comprehensive Feature List**

1. **Objects and Entities:**
   - "The king has 1000 sweet bottles of wine" - The problem involves 1000 bottles.
   - "one contains a very bitter poison" - Only one of these bottles is poisoned.
   - "He has 10 prisoners to test the wine" - There are 10 prisoners available for testing.

2. **Actions and Processes:**
   - The king needs to "find the poisoned bottle in 24 hours" - The objective is to identify the poisoned bottle within 24 hours.
   - The poison "takes effect exactly 24 hours after consumption" - The poison's effect is delayed by 24 hours.

3. **Temporal Information:**
   - "The poison takes effect exactly 24 hours after consumption" - The delay is precisely 24 hours.
   - The king needs to find the poisoned bottle "in 24 hours for an event" - The finding must occur within a 24-hour timeframe.

4. **Constraints and Conditions:**
   - There are 1000 bottles, and only one is poisoned.
   - The testing must be completed within 24 hours, and the poison's effect will be observable only after 24 hours.

**b. Sequential and Mechanical Process Check**

1. **Sequence of Events:**
   - **Testing:** Prisoners must test the wine bottles, and the results will be visible only after 24 hours.
   - **Observing Results:** After 24 hours, the king will observe which prisoners show symptoms of poisoning.

2. **Mechanical Process:**
   - Each prisoner's reaction to the wine is an indicator of whether they consumed the poisoned bottle.
   - The process involves a binary outcome for each prisoner: either they are poisoned or not.

3. **Key Questions:**
   - Does the sequence of events lock in a particular outcome? Yes, after 24 hours, the symptoms will indicate which bottle was poisoned.
   - If the problem involves a mechanical process, how does each step impact the overall system? The testing process needs to be planned such that the results of each prisoner's condition after 24 hours reveal the poisoned bottle.

**c. Key Insight Check Results**

**Key Insight:**
The poison's effect takes exactly 24 hours. Using binary representations, the king can systematically identify the poisoned bottle by using a combination of prisoners to test different combinations of bottles. 

**d. Rephrased Question**

How can the king identify the poisoned bottle among 1000 bottles using 10 prisoners, considering the poison takes effect exactly 24 hours after consumption?

**e. Known and Deduced Information**

1. "The poison takes effect exactly 24 hours after consumption" - Ensures that symptoms are observed after 24 hours.
2. "He has 10 prisoners to test the wine" - The number of prisoners is key to determining a systematic testing method.

**f. Problem Decomposition**

1. **Binary Representation:**
   - With 10 prisoners, we can represent numbers from 0 to 1023 (2^10 - 1) using binary codes. This range covers all 1000 bottles.

2. **Testing Method:**
   - Each prisoner can be assigned a binary digit (0 or 1) to test bottles.
   - Each bottle can be assigned a unique binary number from 0 to 999.
   - Each prisoner drinks from bottles whose binary representation has a '1' in the corresponding digit position.

3. **Outcome Analysis:**
   - After 24 hours, the combination of sick prisoners will indicate which binary digits were set to 1 for the poisoned bottle.

**g. Graph of Thought**

1. **Nodes:**
   - Bottles (0 to 999)
   - Prisoners (P1 to P10)

2. **Edges:**
   - A bottle is linked to a prisoner if the binary representation of the bottle number has a 1 in the position corresponding to that prisoner.

3. **Connections:**
   - Each prisoner represents a bit in a binary number. The intersection of the set of prisoners who fall ill will reveal the binary number corresponding to the poisoned bottle.

**h. Spatial and Object Analysis**

1. **Objects:**
   - 1000 bottles
   - 10 prisoners

2. **Spatial Relationships:**
   - Each bottle is tested by a combination of prisoners. This combination is determined by the binary representation of the bottle's number.

3. **Movement and Dynamics:**
   - Bottles are distributed among prisoners based on their binary positions.

4. **Constraints:**
   - Testing must be completed in a way that each bottle is uniquely identifiable based on the binary representation of the outcomes.

**i. Bayesian Updates**

1. No Bayesian updates are needed as the approach is based solely on the binary representation and systematic testing.

**j. Ethical Check and Decision-Making Under Uncertainty**

1. **Risks:**
   - The risk is that some prisoners might be poisoned, but this is unavoidable to find the poisoned bottle.

2. **Risk Aversion Principle:**
   - Ensure that the testing method does not exceed the number of prisoners available and accurately identifies the poisoned bottle.

**k. Multiple Solutions**

1. **Binary Encoding:**
   - Assign each bottle a unique binary number and test each bottle according to binary codes.
   - After 24 hours, decode the result from the sick prisoners' combination to determine the poisoned bottle.

2. **Combination Testing:**
   - Test combinations of bottles systematically to identify which set of prisoners gets sick.

3. **Single Bottle Testing:**
   - Test each bottle individually if fewer than 10 bottles, though this is inefficient for 1000 bottles.

**l. Quickest and Easiest Solution**

The quickest and easiest solution based on the problem's explicit details is the binary encoding method:
- Assign each bottle a unique binary number.
- Have each prisoner test bottles according to binary codes.
- After 24 hours, observe which prisoners fall ill and decode the binary result to identify the poisoned bottle.

**m. Reflection**

This solution adheres strictly to the problem's constraints and utilizes binary encoding to systematically identify the poisoned bottle with the 10 prisoners available. No assumptions beyond the given information were made, and all constraints were considered.

**Recommendation:**

The easiest and most efficient method to identify the poisoned bottle is to use binary encoding. Assign a unique binary number to each bottle, have each prisoner drink from bottles corresponding to binary digits, and determine the poisoned bottle based on the prisoners' reactions after 24 hours. This method ensures all 1000 bottles can be tested with 10 prisoners efficiently.

This approach directly addresses the problem's requirements and makes the best use of available resources.
\end{lstlisting}
\end{quote}
\noindent\hrulefill

\noindent\textbf{D.17 Zero-Shot Google Gemini 1.5 Pro Response - Score: 0\%}
\begin{quote}
\begin{lstlisting}[basicstyle=\ttfamily\small, breaklines=true]
Here's how the king can solve this classic puzzle:

**The Strategy**

The king will use the prisoners and the binary system to pinpoint the poisoned bottle.

**The Execution**

1. **Number the bottles:** Label each bottle with a number from 0 to 999.
2. **Convert to binary:** Convert each bottle's number into its binary equivalent (up to 10 digits, since 2^10 = 1024). 
3. **Assign prisoners to digits:** Assign each of the 10 prisoners to a specific digit (from the rightmost to the leftmost) in the binary numbers.
4. **Give wine based on binary:**  
   * If a bottle's binary representation has a '1' in the prisoner's assigned digit, that prisoner drinks a small amount from that bottle.
   * If it has a '0', the prisoner doesn't drink from that bottle.
5. **Observe after 24 hours:**  
   * The prisoners who drank from the poisoned bottle will die.
   * The binary representation of the poisoned bottle will be formed by:
     * A '1' in the digits corresponding to the prisoners who died.
     * A '0' in the digits corresponding to the prisoners who survived.

**Example**

Let's say prisoners 3, 5, and 9 (counting from the right) die. The binary representation of the poisoned bottle would be:

0010001001

Converting this back to decimal gives us bottle number 137.

**Why this works**

Each prisoner essentially represents a specific bit in the binary numbers.  By strategically assigning prisoners to drink from bottles based on their binary representation, the king creates a unique "death pattern" that directly points to the poisoned bottle.
\end{lstlisting}
\end{quote}
\noindent\hrulefill

\noindent\textbf{D.17 REAP Google Gemini 1.5 Pro Response - Score: 0\%}
\begin{quote}
\begin{lstlisting}[basicstyle=\ttfamily\small, breaklines=true]
# REAP Results:

a. **Comprehensive Feature List** (with explicit quotes)

* "A king has 1000 sweet bottles of wine"
* "one contains a very bitter poison" 
* "The poison takes effect exactly 24 hours after consumption" 
* "The king needs to find the poisoned bottle in 24 hours for an event."
* "He has 10 prisoners to test the wine."

b. **Sequential and Mechanical Process Check**

* The poison takes effect 24 hours after consumption. 
* The king has 24 hours to identify the poisoned bottle.
* This creates a time constraint where the testing and identification must be completed within the 24-hour window before the event.

c. **Key Insight Check Results**

* The key insight is that we have 10 prisoners and need to test 1000 bottles in 24 hours, with the poison taking effect in 24 hours. This suggests a need for an efficient testing strategy that utilizes the prisoners to narrow down the poisoned bottle within the time constraint.

d. **Rephrased Question:** (The core question in its simplest form)

How can the king use 10 prisoners to identify 1 poisoned bottle out of 1000 within 24 hours, knowing the poison takes 24 hours to take effect? 

e. **Known and Deduced Information**

* Explicit Facts: 
    * "A king has 1000 sweet bottles of wine" 
    * "one contains a very bitter poison" - There is only one poisoned bottle among the 1000.
    * "The poison takes effect exactly 24 hours after consumption" 
    * "The king needs to find the poisoned bottle in 24 hours for an event." 
    * "He has 10 prisoners to test the wine." 

* Feature-Based Deduction:
    * Each prisoner can only safely test a limited number of bottles within the 24-hour timeframe, as consuming too much could be fatal even if those bottles aren't poisoned
    * The king needs a strategy to systematically test combinations of wine to identify the single poisoned bottle

f. **Problem Decomposition**

* Component/Subcomponent: Poisoned Wine
    * Supporting Quote(s): "one contains a very bitter poison" 

* Component/Subcomponent: Time Constraint
    * Supporting Quote(s): "The poison takes effect exactly 24 hours after consumption", "The king needs to find the poisoned bottle in 24 hours for an event"

* Component/Subcomponent: Testing Resources
    * Supporting Quote(s): "He has 10 prisoners to test the wine"

* Component/Subcomponent: Goal
    * Supporting Quote(s): "The king needs to find the poisoned bottle"

* Interactions:
    * The time constraint limits the testing strategy
    * The number of prisoners limits how many bottles can be tested individually
    * The goal is to identify the single poisoned bottle using the available resources within the time constraint

g. **Graph of Thought** (describe connections & insights)

* Key Concepts:
    * Wine bottles (1000)
    * Poisoned bottle (1)
    * Prisoners (10)
    * Time constraint (24 hours)

* Relationships:
    * Prisoners test wine bottles
    * Poisoned bottle causes effect after 24 hours
    * Time constraint limits testing possibilities

* Insights:
    * Need to devise a testing strategy that utilizes all prisoners efficiently
    * Strategy should allow identification of the poisoned bottle within 24 hours
    * Binary system might be helpful to represent bottles and prisoners

h. **Spatial and Object Analysis**

* Objects: 1000 wine bottles, 1 poisoned bottle, 10 prisoners
* Spatial Relationships: Not explicitly stated, assume bottles are arranged for testing
* Movement and Dynamics: Prisoners consume wine, poison takes effect after 24 hours
* Constraints: 24-hour time limit, limited number of prisoners
* Object Properties: Wine bottles (sweet or poisoned), poison (bitter, takes effect in 24 hours)
* Spatial Implications: Arrangement of bottles might facilitate testing strategy

i. **Bayesian Updates** 

* Initial Belief: Each bottle has an equal probability (1/1000) of being poisoned
* After each test: Update probabilities based on prisoner's reaction (or lack thereof) after 24 hours

j. **Ethical Check and Decision-Making Under Uncertainty**

* Known Information: 1 poisoned bottle, 10 prisoners, 24-hour time limit, poison's effect
* Risks: Prisoners could die if they consume too much of the poisoned wine
* Probability Consideration: Need to minimize the number of bottles each prisoner consumes to reduce risk
* Risk Aversion Principle: Prioritize testing strategies that minimize risk to prisoners
* Ethical and Safety Considerations: Avoid any strategy that could lead to the death of a prisoner
* Decision Recommendation: Choose the strategy that efficiently identifies the poisoned bottle while minimizing risk to the prisoners

k. **Multiple Solutions**

* Solution 1: Binary Coding - Assign each bottle a unique binary code (up to 10 digits since 2^10 = 1024). Each prisoner represents a digit in the binary code. Have each prisoner drink from bottles where their corresponding digit is a '1'. After 24 hours, the binary code formed by the prisoners who show symptoms will identify the poisoned bottle.
* Solution 2: Groups of 100 - Divide the bottles into 10 groups of 100. Assign each prisoner one group. Have each prisoner take a small sip from each bottle in their group. After 24 hours, the prisoner who shows symptoms identifies the group containing the poisoned bottle. Further divide that group into smaller groups for additional testing if needed.

l. **Quickest and Easiest Solution**

The quickest and easiest solution is the **Binary Coding** method

* It allows for the identification of the poisoned bottle in a single round of testing
* It efficiently utilizes all 10 prisoners
* It minimizes the risk to prisoners by having them consume from multiple bottles, reducing the amount each prisoner drinks

m. **Reflection**

* The solutions presented are logically sound and supported by the explicit information in the problem statement
* No assumptions or inferences were made beyond the given information
* The analysis directly answers the question of how to identify the poisoned bottle
* The recommended solution prioritizes the safety of the prisoners while efficiently achieving the goal

**Recommendation:**

The king should use the **Binary Coding** method to identify the poisoned bottle. This strategy is the most efficient, utilizing all prisoners and minimizing risk. It allows for the identification of the poisoned bottle within the 24-hour time constraint.
\end{lstlisting}
\end{quote}
\noindent\hrulefill

\noindent\textbf{D.17 Zero-Shot Claude 3.5 Sonnet Response - Score: 0\%}
\begin{quote}
\begin{lstlisting}[basicstyle=\ttfamily\small, breaklines=true]
This is an interesting logic puzzle. Let's approach it step-by-step:

1) First, we need to realize that we can use the 10 prisoners to create a binary system. With 10 prisoners, we can represent 2^10 = 1024 different possibilities, which is enough to cover our 1000 bottles.

2) We can assign each bottle a unique binary number from 0 to 999 (000000000 to 111110111 in binary).

3) Each prisoner will represent a bit in this 10-digit binary number. 

4) Here's the testing process:
   - For each bottle, if its binary representation has a '1' in the nth position, the nth prisoner drinks from that bottle.
   - This means each prisoner will drink from roughly half of the bottles.

5) After 24 hours:
   - If a prisoner is fine, their bit is 0.
   - If a prisoner dies, their bit is 1.

6) The resulting 10-bit binary number formed by the prisoners' conditions will correspond to the number of the poisoned bottle.

For example:
- If prisoners 1, 4, and 7 die, and the rest live, the binary would be 1001001000.
- This corresponds to bottle number 584 in decimal.

This method allows the king to test all 1000 bottles simultaneously and get the result within 24 hours, using only 10 prisoners.

Would you like me to elaborate on any part of this solution?
\end{lstlisting}
\end{quote}
\noindent\hrulefill

\noindent\textbf{D.17 REAP Claude 3.5 Sonnet Response - Score: 0\%}
\begin{quote}
\begin{lstlisting}[basicstyle=\ttfamily\small, breaklines=true]
# REAP Results:

a. **Comprehensive Feature List** (with explicit quotes):

1. "The king has 1000 sweet bottles of wine"
2. "one contains a very bitter poison"
3. "The poison takes effect exactly 24 hours after consumption"
4. "The king needs to find the poisoned bottle in 24 hours for an event"
5. "He has 10 prisoners to test the wine"

b. **Sequential and Mechanical Process Check**:
The problem describes a sequence of events:
1. Wine consumption
2. 24-hour delay
3. Poison takes effect
This sequence is crucial to understanding the time constraint and the nature of the poison's effect.

c. **Key Insight Check Results**:
A key insight is the 24-hour timeframe mentioned twice: once for the poison's effect and once for the king's deadline. This suggests that only one round of testing is possible within the given constraints.

d. **Rephrased Question**: 
How can the king use 10 prisoners to identify 1 poisoned bottle out of 1000 within 24 hours?

e. **Known and Deduced Information**:
Known:
- "The king has 1000 sweet bottles of wine"
- "one contains a very bitter poison"
- "The poison takes effect exactly 24 hours after consumption"
- "The king needs to find the poisoned bottle in 24 hours for an event"
- "He has 10 prisoners to test the wine"

Deduced:
Deduction: Only one round of testing is possible.
Logical Basis: The poison takes 24 hours to take effect, and the king needs to find the poisoned bottle in 24 hours.
Based on: "The poison takes effect exactly 24 hours after consumption", "The king needs to find the poisoned bottle in 24 hours for an event"

f. **Problem Decomposition**:
Component 1: Wine bottles
Supporting Quote: "The king has 1000 sweet bottles of wine"

Component 2: Poisoned bottle
Supporting Quote: "one contains a very bitter poison"

Component 3: Poison effect time
Supporting Quote: "The poison takes effect exactly 24 hours after consumption"

Component 4: Time constraint
Supporting Quote: "The king needs to find the poisoned bottle in 24 hours for an event"

Component 5: Testing resources
Supporting Quote: "He has 10 prisoners to test the wine"

g. **Graph of Thought**:
Nodes:
1. 1000 wine bottles
2. 1 poisoned bottle
3. 24-hour poison effect
4. 24-hour time limit
5. 10 prisoners

Edges:
1-2: One of the 1000 bottles is poisoned
2-3: The poisoned bottle's effect takes 24 hours
3-4: The poison effect time matches the time limit
4-5: The time limit constrains how the prisoners can be used for testing

h. **Spatial and Object Analysis**:
Objects: 1000 wine bottles, 10 prisoners
The problem doesn't provide explicit spatial information about the arrangement of bottles or prisoners.

i. **Bayesian Updates**:
No new information is provided beyond the initial problem statement, so no Bayesian updates are necessary.

j. **Ethical Check and Decision-Making Under Uncertainty**:
The problem involves potential harm to prisoners, which raises ethical concerns. However, as this is a hypothetical scenario, we'll proceed with analyzing the problem as stated while noting the ethical implications.

k. **Multiple Solutions**:
1. Binary system: Use the 10 prisoners to represent a 10-digit binary number, allowing them to test all 1000 bottles in one round.
2. Random testing: Have each prisoner drink from 100 random bottles, but this doesn't guarantee finding the poisoned bottle.
3. Sequential testing: Have prisoners drink from bottles in sequence, but this would exceed the time limit.

l. **Quickest and Easiest Solution**:
The binary system solution is the quickest and easiest way to guarantee finding the poisoned bottle within the given constraints.

Explanation: By assigning each prisoner to represent a bit in a 10-digit binary number, we can uniquely identify all 1000 bottles (2^10 = 1024). Each prisoner drinks from all bottles that have a 1 in their corresponding bit position. After 24 hours, the pattern of which prisoners are affected will reveal the binary number of the poisoned bottle.

This solution is supported by the following quotes:
"The king has 1000 sweet bottles of wine"
"He has 10 prisoners to test the wine"
"The poison takes effect exactly 24 hours after consumption"
"The king needs to find the poisoned bottle in 24 hours for an event"

m. **Reflection**:
The analysis adheres strictly to the given information and constraints. The binary system solution directly addresses the question of finding the poisoned bottle within 24 hours using the 10 prisoners. No assumptions beyond the given information were made. The ethical implications of using prisoners for such a test were noted but not factored into the solution, as the problem presents this as a given condition.

**Recommendation**:
Based on the given information and constraints, the recommended solution is to use a binary system for testing. Assign each prisoner a position in a 10-digit binary number (2^10 = 1024, which covers all 1000 bottles). Have each prisoner drink from all bottles that have a 1 in their corresponding position. After 24 hours, the pattern of affected prisoners will reveal the binary number of the poisoned bottle.

This solution:
1. Guarantees finding the poisoned bottle within the 24-hour timeframe
2. Uses exactly 10 prisoners as specified
3. Requires only one round of testing, which fits within the poison's 24-hour effect time
4. Can uniquely identify any bottle among the 1000

It's important to note that this solution, while mathematically sound, raises significant ethical concerns about the use of prisoners for potentially lethal tests. In a real-world scenario, such a method would be unethical and illegal.
\end{lstlisting}
\end{quote}
\noindent\hrulefill

\label{subsec:question18}
\subsection{\textbf{\large Question 18: Write a grammatically correct sentence without reusing any letter more than once.}}
\label{sec:question18}

\noindent\hrulefill 

\noindent\textbf{Human Response - Score 100\%:}
\begin{quote}
Dogs bark with me.
\end{quote}

\noindent\hrulefill 

\noindent\textbf{D.18 Zero-Shot OpenAI o1-preview Response - Score: 100\%}
\begin{quote}
\begin{lstlisting}[basicstyle=\ttfamily\small, breaklines=true]
Certainly! Here's a grammatically correct sentence without reusing any letter:

"Mr. Jock, TV quiz PhD, bags few lynx."
\end{lstlisting}
\end{quote}

\noindent\hrulefill 

\noindent\textbf{D.18 REAP OpenAI o1-preview Response - Score: 100\%}
\begin{quote}
\begin{lstlisting}[basicstyle=\ttfamily\small, breaklines=true]
#REAP Results:

a. **Comprehensive Feature List**:

- **Action**: "[Write]"
- **Object**: "[a grammatically correct sentence]"
- **Constraint**: "[without reusing any letter more than once]"

b. **Sequential and Mechanical Process Check**:

- The action "Write" leads to creating "a grammatically correct sentence" under the constraint "[without reusing any letter more than once]".

c. **Key Insight Check Results**:

- **Key Insight**: Utilizing short words with unique letters allows for constructing a sentence that meets the constraint.

d. **Rephrased Question**:

- Can a grammatically correct sentence be written without reusing any letter more than once?

e. **Known and Deduced Information**:

- "[Write]" - Instruction to compose.
- "[a grammatically correct sentence]" - The sentence must follow grammatical rules.
- "[without reusing any letter more than once]" - Each letter can be used only once.
- **Deduction**: The sentence must consist of words that do not share any letters.

f. **Problem Decomposition**:

- **Component 1**: Compose a sentence.
  - **Supporting Quote**: "[Write]"
- **Component 2**: Ensure grammatical correctness.
  - **Supporting Quote**: "[a grammatically correct sentence]"
- **Component 3**: Use unique letters throughout.
  - **Supporting Quote**: "[without reusing any letter more than once]"

g. **Graph of Thought**:

- **Nodes**: Letters -> Words -> Sentence.
- **Connections**: Select words with non-overlapping letters to form a sentence adhering to grammatical rules.

h. **Spatial and Object Analysis**:

- **Objects**:
  - Letters: Building blocks of words.
  - Words: Components of the sentence.
- **Spatial Arrangement**: Letters form words without repeating; words are ordered to create a grammatically correct sentence.

i. **Bayesian Updates**:

- Given the constraints, shorter sentences with fewer letters are more likely to meet the requirements.

j. **Ethical Check and Decision-Making Under Uncertainty**:

- No ethical concerns identified; proceeding with solutions that fulfill the given constraints.

k. **Multiple Solutions**:

1. **Solution 1**:
   - Sentence: "Go!"
   - **Supporting Information**: Uses letters G and O once each.
2. **Solution 2**:
   - Sentence: "Fly!"
   - **Supporting Information**: Uses letters F, L, and Y once each.
3. **Solution 3**:
   - Sentence: "I go."
   - **Supporting Information**: Uses letters I, G, and O once each.

l. **Quickest and Easiest Solution**:

- "Go!" is a grammatically correct sentence that meets the constraint with the fewest letters.

m. **Reflection**:

- The solutions adhere strictly to the constraints.
- No assumptions were made beyond the explicit information.
- The answers directly address the question using only the provided information.

**Recommendation:**

The sentence **"Go!"** fulfills all the given requirements. It is grammatically correct and does not reuse any letter more than once.
\end{lstlisting}
\end{quote}

\noindent\hrulefill 

\noindent\textbf{D.18 Zero-Shot OpenAI o1-mini Response - Score: 100\%}
\begin{quote}
\begin{lstlisting}[basicstyle=\ttfamily\small, breaklines=true]
Sure! Here's a grammatically correct sentence where no letter is used more than once:

**"Jump frog, vex badly."**
\end{lstlisting}
\end{quote}

\noindent\hrulefill 

\noindent\textbf{D.18 REAP OpenAI o1-mini Response - Score: 100\%}
\begin{quote}
\begin{lstlisting}[basicstyle=\ttfamily\small, breaklines=true]
# REAP Results: :

### a. **Comprehensive Feature List**
- "Write a grammatically correct sentence" 
  - **Significance:** The sentence must follow standard grammatical rules.
- "without reusing any letter more than once."
  - **Significance:** Each letter in the sentence must be unique; no letter should appear more than once.

### b. **Sequential and Mechanical Process Check**
- **Sequence of Actions:**
  1. Select words that do not share any common letters.
  2. Arrange these words into a grammatically correct sentence.
- **Influence on Outcome:** Each step must ensure that the letter uniqueness is maintained throughout the sentence.

### c. **Key Insight Check Results**
- **Key Insight:** Crafting a meaningful, grammatically correct sentence without repeating any letters is exceptionally challenging due to the limited number of unique letters required for coherent communication.

### d. **Rephrased Question:**
Create a grammatically correct sentence where no letter is used more than once.

### e. **Known and Deduced Information**

**Explicit Facts:**
- "Write a grammatically correct sentence"
- "without reusing any letter more than once."

**Feature-Based Deduction:**
- **Deduction:** The sentence must consist entirely of unique letters without any repetition.
  - **Logical Basis:** Directly follows from "without reusing any letter more than once."

### f. **Problem Decomposition**
- **Component/Subcomponent:** Letter Selection
  - **Supporting Quote(s):** "without reusing any letter more than once."
  - **Deduction:** Each letter in the sentence must be unique.
- **Component/Subcomponent:** Grammar Construction
  - **Supporting Quote(s):** "Write a grammatically correct sentence."
  - **Deduction:** The sentence must adhere to standard grammatical rules.
- **Component/Subcomponent:** Vocabulary Limitation
  - **Supporting Quote(s):** Entire problem statement.
  - **Deduction:** Limited to using words that collectively use unique letters.

### g. **Graph of Thought**
- **Nodes:** Letters -> Words -> Sentence.
- **Connections:** Select words with non-overlapping letters to form a sentence adhering to grammatical rules.

### h. **Spatial and Object Analysis**
- **Objects:** Letters of the alphabet.
- **Spatial Relationships:** Not applicable in this context.
- **Movement and Dynamics:** N/A
- **Constraints:** Each letter can be used only once.
- **Visualization:** A sentence constructed from non-repeating letters, requiring careful selection of words.

### i. **Bayesian Updates**
- Initial belief: It's possible to create such a sentence.
- Updated belief: It is highly challenging and may require very short or contrived sentences.

### j. **Ethical Check and Decision-Making Under Uncertainty**
- **Assessment:** No ethical concerns involved.
- **Decision:** Proceed to attempt creating the sentence.

### k. **Multiple Solutions**

1. **Solution 1:**
   - "Jump frog vex twyl."
   - **Support:** Each letter is unique, and the sentence follows a basic grammatical structure.
2. **Solution 2:**
   - "Quick jump frog."
   - **Support:** Maintains unique letters with a simple subject-verb-object structure.
3. **Solution 3:**
   - "Blitz frog jump."
   - **Support:** Another variation ensuring all letters are unique and the sentence is grammatically coherent.

*Note: These solutions prioritize letter uniqueness, but some may appear contrived or less meaningful.*

### l. **Quickest and Easiest Solution**
- **Solution:** "Jump frog vex twyl."
- **Reasoning:** This is a short, simple sentence that adheres to the letter uniqueness constraint while maintaining grammatical correctness.

### m. **Reflection**
- The generated sentences meet the requirement of not reusing any letters and are grammatically correct.
- However, the sentences are somewhat nonsensical, indicating that creating a meaningful and longer sentence under these constraints is highly challenging.
- No assumptions were made beyond the explicit instructions.

**Recommendation:**

Based on the analysis, the simplest and safest solution is:

**"Jump frog vex twyl."**

- **Rationale:** This sentence strictly adheres to the requirement of not reusing any letters and maintains grammatical correctness.
- **Considerations:** While the sentence is grammatically correct, it is somewhat contrived. Crafting a more meaningful sentence without repeating any letters may not be feasible given the constraints.
- **Final Note:** If a more meaningful sentence is essential, relaxing the constraint of not repeating letters may be necessary.
\end{lstlisting}
\end{quote}

\noindent\hrulefill 

\noindent\textbf{D.18 Zero-Shot OpenAI GPT-4o Response - Score: 0\%}
\begin{quote}
\begin{lstlisting}[basicstyle=\ttfamily\small, breaklines=true]
Jumpy frog waltzed brisk.
\end{lstlisting}
\end{quote}
\noindent\hrulefill

\noindent\textbf{D.18 REAP OpenAI GPT-4o Response - Score: 100\%}
\begin{quote}
\begin{lstlisting}[basicstyle=\ttfamily\small, breaklines=true]
# REAP Results

### a. Comprehensive Feature List
- **Objects Mentioned**: Sentence
- **Actors or Entities Involved**: None specified
- **Actions Described**: "Write a grammatically correct sentence"
- **Directional, Spatial, Relational, and Sequential Phrases**: None
- **Stated Conditions or Constraints**:
  - "Grammatically correct"
  - "Without reusing any letter more than once"
- **Numerical Values or Quantities**: None specified
- **Temporal Information**: None specified
- **Spatial Information**: None specified
- **Causal Relationships**: None specified
- **Stated Goals or Objectives**: "Write a grammatically correct sentence"

### b. Sequential and Mechanical Process Check
- The process involves creating a sentence.
- The sentence must be grammatically correct.
- Each letter can only be used once.

### c. Key Insight Check Results
- **Key Insight**: The challenge lies in balancing grammatical correctness with the constraint of using each letter only once. This is likely the most difficult aspect of the problem.

### d. Rephrased Question
- "How can one construct a grammatically correct sentence using each letter only once?"

### e. Known and Deduced Information
- **Explicit Fact 1**: "Grammatically correct sentence" - The sentence must follow grammatical rules.
- **Explicit Fact 2**: "Without reusing any letter more than once" - Each letter can only appear once in the sentence.
  
### f. Problem Decomposition
- **Component**: Grammatical correctness
  - **Supporting Quote**: "Grammatically correct"
- **Component**: Letter usage
  - **Supporting Quote**: "Without reusing any letter more than once"

### g. Graph of Thought
- **Nodes**:
  - Grammatical correctness
  - Single-use of letters
- **Edges**:
  - The need to balance both grammatical correctness and unique letter usage to form a valid sentence.

### h. Spatial and Object Analysis
- **Objects**: Letters of the alphabet.
- **Spatial Relationships**: No direct spatial relationships, but the letters must be arranged to form a correct sentence.
  
### i. Bayesian Updates
- No updates needed as we rely only on the explicit problem statement.

### j. Ethical Check and Decision-Making Under Uncertainty
- No significant ethical considerations are involved in this problem.

### k. Multiple Solutions
1. **Solution 1**: "Jump!"
   - **Support**: A grammatically correct sentence; all letters are unique.
2. **Solution 2**: "Dry!"
   - **Support**: Another example of a correct sentence with unique letters.
3. **Solution 3**: "Quiz!"
   - **Support**: Yet another valid sentence, each letter used only once.

### l. Quickest and Easiest Solution
- **Solution**: "Jump!"
  - This sentence is both grammatically correct and adheres to the letter usage rule.

### m. Reflection
- The solution directly addresses the problem's requirements without making any assumptions.
- The approach ensures that each solution is simple and adheres strictly to the given constraints.

---

**Recommendation:**
Based on the analysis, the sentence "Jump!" is a grammatically correct solution that satisfies the condition of not reusing any letter. This solution is straightforward and fully adheres to the problem's requirements.
\end{lstlisting}
\end{quote}
\noindent\hrulefill

\noindent\textbf{D.18 Zero-Shot OpenAI GPT-4o-mini Response - Score: 0\%}
\begin{quote}
\begin{lstlisting}[basicstyle=\ttfamily\small, breaklines=true]
Sure! Here's a sentence that meets your criteria:

"Jumpy wizards vex bold, quaint folk."

This sentence uses each letter of the alphabet only once.
\end{lstlisting}
\end{quote}
\noindent\hrulefill

\noindent\textbf{D.18 REAP OpenAI GPT-4o-mini Response - Score: 100\%}
\begin{quote}
\begin{lstlisting}[basicstyle=\ttfamily\small, breaklines=true]
# REAP Results:

**a. Comprehensive Feature List**

1. **Action Required:** "Write a grammatically correct sentence."
   - **Significance:** The task is to form a sentence that is grammatically correct.
   
2. **Constraint:** "Without reusing any letter more than once."
   - **Significance:** Each letter of the alphabet can only appear once in the sentence.

**b. Sequential and Mechanical Process Check**

- **Sequence of Actions:**
  1. Identify a set of letters that can be used without repetition.
  2. Construct a grammatically correct sentence using these letters.

- **Mechanical Process:**
  - The sentence must be created with the constraint of non-repetition of letters.
  
- **Outcome Consideration:**
  - Each letter in the sentence must be unique.

**c. Key Insight Check Results**

- **Key Insight:** 
  - The constraint of not reusing any letter means the sentence must be composed of a set of letters where each appears exactly once.

**d. Rephrased Question:**

- **Question:** Write a grammatically correct sentence where no letter is used more than once.

**e. Known and Deduced Information**

1. **Explicit Fact:** "Write a grammatically correct sentence."
   - **Relevance:** The sentence must follow standard grammar rules.

2. **Explicit Constraint:** "Without reusing any letter more than once."
   - **Relevance:** Every letter in the sentence must be unique.

**f. Problem Decomposition**

1. **Component/Subcomponent: Sentence Construction**
   - **Description:** Construct a sentence using a unique set of letters.
   - **Supporting Quote(s):** "Write a grammatically correct sentence without reusing any letter more than once."
   - **Deduction (if applicable):** The sentence must use a combination of letters that do not repeat.

**g. Graph of Thought**

- **Nodes (Concepts):**
  1. Sentence
  2. Unique Letters
  3. Grammar Rules

- **Edges (Relationships):**
  1. Sentence -> Unique Letters: The sentence must be composed of unique letters.
  2. Sentence -> Grammar Rules: The sentence must follow grammatical rules.

**h. Spatial and Object Analysis**

- **Objects:** 
  - Letters of the alphabet.
  
- **Spatial Relationships:** 
  - The letters must be arranged to form a grammatically correct sentence.
  
- **Constraints:**
  - Each letter can only be used once.

**i. Bayesian Updates**

- **Update:** The solution must respect the constraint of non-repetition of letters, and the sentence must be grammatically correct.

**j. Ethical Check and Decision-Making Under Uncertainty**

- **Assessment:** There are no severe risks or ethical issues in constructing a sentence with unique letters.

**k. Multiple Solutions**

1. **Solution 1:** "A big fox jumps."
   - **Support:** Contains unique letters and forms a grammatically correct sentence.

2. **Solution 2:** "Quick brown fox jumps."
   - **Support:** Contains unique letters and follows grammatical rules.

3. **Solution 3:** "Jumpy frog's task."
   - **Support:** Meets the requirement of unique letters and grammatical correctness.

**l. Quickest and Easiest Solution**

- **Simplest Solution:** "A big fox jumps."
  - **Reasoning:** This sentence is short, uses unique letters, and is grammatically correct.

**m. Reflection**

- **Check:** All solutions use unique letters and are grammatically correct.
- **Assumptions:** None beyond the explicitly stated constraints and requirements.

**Recommendation:**

The simplest and safest sentence that meets the requirements is: "A big fox jumps." It is grammatically correct and adheres to the constraint of not reusing any letter more than once.
\end{lstlisting}
\end{quote}
\noindent\hrulefill

\noindent\textbf{D.18 Zero-Shot Google Gemini 1.5 Pro Response - Score: 0\%}
\begin{quote}
\begin{lstlisting}[basicstyle=\ttfamily\small, breaklines=true]
The quick brown fox jumps over a lazy dog.
\end{lstlisting}
\end{quote}
\noindent\hrulefill

\noindent\textbf{D.18 REAP Google Gemini 1.5 Pro Response - Score: 0\%}
\begin{quote}
\begin{lstlisting}[basicstyle=\ttfamily\small, breaklines=true]
# REAP Results:

a. **Comprehensive Feature List** (with explicit quotes)

* "Write a grammatically correct sentence"
* "without reusing any letter more than once"

b. **Sequential and Mechanical Process Check**

* The problem describes a single action: "Write a sentence." The constraint is that no letter can be reused.

c. **Key Insight Check Results**

* The key insight is that we need to construct a sentence using only unique letters. This limits the sentence length and word choices.

d. **Rephrased Question:**

* Can you write a grammatically correct sentence using each letter only once?

e. **Known and Deduced Information**

* Explicit Facts:
    * "Write a grammatically correct sentence" - This tells us the sentence must adhere to the rules of grammar.
    * "without reusing any letter more than once" - Each letter can only appear once in the sentence.

f. **Problem Decomposition**

* Component: Grammatically correct sentence
    * Supporting Quote: "Write a grammatically correct sentence"
* Component: Unique letters
    * Supporting Quote: "without reusing any letter more than once"

g. **Graph of Thought**

* Key Concepts: Sentence, Grammar, Unique letters
* Relationships:
    * The sentence must adhere to the rules of grammar
    * The sentence can only use each letter of the alphabet once

h. **Spatial and Object Analysis**

* Not applicable to this problem

i. **Bayesian Updates**

* Not applicable to this problem

j. **Ethical Check and Decision-Making Under Uncertainty**

* Not applicable to this problem

k. **Multiple Solutions**

* It is possible to generate multiple solutions as long as they meet the criteria of being grammatically correct and using each letter only once.
* Example solutions:
    1. "The quick brown fox jumps over the lazy dog."
    2. "Pack my box with five dozen liquor jugs."

l. **Quickest and Easiest Solution**

* The quickest and easiest solution is to provide one of the well-known pangrams (sentences that use every letter of the alphabet at least once) that fulfill the criteria:
* "The quick brown fox jumps over the lazy dog."

m. **Reflection**

* The analysis directly addresses the question asked in the problem.
* The solution provided is based solely on the given information and adheres to the constraints.
* No assumptions or inferences were made beyond the explicit wording of the problem

**Recommendation:**

The sentence "The quick brown fox jumps over the lazy dog." fulfills all the requirements of the problem. It is grammatically correct and uses each letter of the alphabet exactly once. 
\end{lstlisting}
\end{quote}
\noindent\hrulefill

\noindent\textbf{D.18 Zero-Shot Claude 3.5 Sonnet Response - Score: 0\%}
\begin{quote}
\begin{lstlisting}[basicstyle=\ttfamily\small, breaklines=true]
Here's a grammatically correct sentence without reusing any letter more than once:

Sphinx of black quartz, judge my vow.

This sentence is known as a pangram, which means it uses every letter of the alphabet at least once. In this case, each letter is used exactly once, making it a perfect pangram. It's often used to display fonts or test keyboards because it contains all 26 letters of the English alphabet.

Would you like me to explain the meaning of this sentence or provide any additional information about pangrams?
\end{lstlisting}
\end{quote}
\noindent\hrulefill

\noindent\textbf{D.18 REAP Claude 3.5 Sonnet Response - Score: 0\%}
\begin{quote}
\begin{lstlisting}[basicstyle=\ttfamily\small, breaklines=true]
# REAP Results:

a. Comprehensive Feature List (with explicit quotes):
- "Write a grammatically correct sentence"
- "without reusing any letter more than once"

b. Sequential and Mechanical Process Check:
No explicit sequential or mechanical process is described in the problem statement.

c. Key Insight Check Results:
The key insight is that the sentence must be constructed using each letter only once, which severely limits the possible words and sentence structures.

d. Rephrased Question: 
How can we create a grammatically valid sentence using each letter of the alphabet at most once?

e. Known and Deduced Information:
Explicit Facts:
"Write a grammatically correct sentence" - This defines the primary task.
"without reusing any letter more than once" - This provides the main constraint.

Feature-Based Deduction:
Deduction: The sentence must be relatively short.
Logical Basis: Since each letter can only be used once, the sentence length is limited to a maximum of 26 letters (the number of letters in the English alphabet).
Based on: "without reusing any letter more than once"

Deduction: Common words with repeated letters (e.g., "the", "and", "is") cannot be used.
Logical Basis: These words contain repeated letters, which violates the constraint.
Based on: "without reusing any letter more than once"

f. Problem Decomposition:
Component: Grammatically correct sentence
Supporting Quote: "Write a grammatically correct sentence"

Subcomponent: Letter usage constraint
Supporting Quote: "without reusing any letter more than once"

g. Graph of Thought:
Nodes:
1. Sentence construction
2. Letter selection
3. Grammar rules

Edges:
1 -> 2: Sentence construction requires careful letter selection
1 -> 3: Sentence must adhere to grammar rules
2 -> 3: Letter selection impacts ability to follow grammar rules

h. Spatial and Object Analysis:
Not applicable to this abstract language problem.

i. Bayesian Updates:
No new information provided to update beliefs.

j. Ethical Check and Decision-Making Under Uncertainty:
This problem does not involve ethical considerations or significant risks.

k. Multiple Solutions:
1. "Waltz, nymph, for quick jigs vex Bud." 
   This solution uses each letter once and forms a grammatically correct imperative sentence.

2. "Sphinx of black quartz, judge my vow."
   This is another valid solution, using each letter once in a grammatically correct declarative sentence.

3. "Pack my box with five dozen liquor jugs."
   This solution also meets the criteria, forming a grammatically correct imperative sentence using each letter once.

l. Quickest and Easiest Solution:
The quickest and easiest solution is: "Waltz, nymph, for quick jigs vex Bud."

This sentence is grammatically correct, uses each letter of the alphabet only once, and is relatively short and straightforward.

m. Reflection:
The analysis directly addresses the question asked in the problem. We have generated multiple valid solutions that meet all the stated criteria. No assumptions were made beyond the explicit information given in the problem statement. The solutions provided are based solely on the constraints of writing a grammatically correct sentence without reusing any letter more than once.

Recommendation:
Based on the comprehensive analysis and the criteria provided, I recommend using the sentence: "Waltz, nymph, for quick jigs vex Bud." This solution satisfies all the requirements stated in the problem:
1. It is a grammatically correct sentence.
2. It does not reuse any letter more than once.
3. It uses a variety of letters to form a coherent, albeit unusual, sentence.

This recommendation directly answers the question posed in the problem while adhering strictly to the given constraints. No additional assumptions were made, and the solution is based entirely on the explicitly stated requirements.
\end{lstlisting}
\end{quote}
\noindent\hrulefill

\label{subsec:question19}
\subsection{\textbf{\large Question 19: An arrow points up. We rotate it 90 degrees to the left, mirror it along its flat end, and rotate it another 90 degrees to the left. Which direction is it pointing?}}
\label{sec:question19}

\noindent\hrulefill 

\noindent\textbf{Human Response - Score 100\%:}
\begin{quote}
Up
\end{quote}

\noindent\hrulefill 

\noindent\textbf{D.19 Zero-Shot OpenAI o1-preview Response - Score: 100\%}
\begin{quote}
\begin{lstlisting}[basicstyle=\ttfamily\small, breaklines=true]
To solve this problem, we'll carefully apply each transformation step by step to determine the final direction of the arrow.

**Starting Position:**
- The arrow is pointing **up**.

**Step 1: Rotate 90 degrees to the left**
- Rotating 90 degrees to the left is equivalent to a 90-degree **counterclockwise** rotation.
- After this rotation, the arrow, which was pointing up, now points to the **left**.

**Step 2: Mirror along the flat end**
- The "flat end" of the arrow is the end opposite the arrowhead. When the arrow is pointing left, the flat end is on the **right** side.
- Mirroring along the flat end means reflecting the arrow over a line that is perpendicular to the flat end and passes through it.
- This reflection flips the arrow over the vertical line at the flat end, so the arrow that was pointing left now points to the **right**.

**Step 3: Rotate another 90 degrees to the left**
- Again, we rotate the arrow 90 degrees counterclockwise.
- The arrow that was pointing right now rotates to point **up**.

**Final Position:**
- After all transformations, the arrow is pointing **up**.

So, the arrow ends up pointing in the **same direction** as it started.

**Answer: Up**
\end{lstlisting}
\end{quote}

\noindent\hrulefill 

\noindent\textbf{D.19 REAP OpenAI o1-preview Response - Score: 100\%}
\begin{quote}
\begin{lstlisting}[basicstyle=\ttfamily\small, breaklines=true]
# REAP Results:

**a. Comprehensive Feature List**

1. **Object:** "An arrow"
2. **Initial Direction:** "points up"
3. **First Action:** "We rotate it 90 degrees to the left"
4. **Second Action:** "mirror it along its flat end"
5. **Third Action:** "rotate it another 90 degrees to the left"
6. **Question:** "Which direction is it pointing?"

---

**b. Sequential and Mechanical Process Check**

- **Sequence of Actions:**
  1. The arrow starts by pointing up.
  2. **Rotation:** "We rotate it 90 degrees to the left."
  3. **Mirroring:** "mirror it along its flat end."
  4. **Rotation:** "rotate it another 90 degrees to the left."
- **Process Analysis:**
  - Each action modifies the arrow's orientation.
  - Actions are performed in the given order.

---

**c. Key Insight Check Results**

- Performing the sequence of rotations and mirroring brings the arrow back to its original orientation, pointing up.

---

**d. Rephrased Question**

- After performing the given rotations and mirroring, in which direction is the arrow pointing?

---

**e. Known and Deduced Information**

- **Explicit Facts:**
  1. "An arrow points up."
  2. "We rotate it 90 degrees to the left."
  3. "mirror it along its flat end."
  4. "rotate it another 90 degrees to the left."
- **Deductions:**
  - **After first rotation:** The arrow points left.
    - *Based on rotating an object 90 degrees to the left.*
  - **After mirroring:** The arrow points right.
    - *Mirroring along the flat end flips the arrow horizontally.*
  - **After second rotation:** The arrow points up.
    - *Rotating 90 degrees to the left from right-facing returns it to pointing up.*

---

**f. Problem Decomposition**

- **Component 1: Initial Position**
  - **Description:** Arrow pointing up.
  - **Supporting Quote:** "An arrow points up."
- **Component 2: First Rotation**
  - **Description:** Arrow rotated 90 degrees to the left (now points left).
  - **Supporting Quote:** "We rotate it 90 degrees to the left."
- **Component 3: Mirroring**
  - **Description:** Arrow mirrored along its flat end (now points right).
  - **Supporting Quote:** "mirror it along its flat end."
- **Component 4: Second Rotation**
  - **Description:** Arrow rotated another 90 degrees to the left (now points up).
  - **Supporting Quote:** "rotate it another 90 degrees to the left."

---

**g. Graph of Thought**

- **Nodes and Connections:**
  1. **Start:** Arrow pointing up.
     - *Action:* Rotate 90deg left ->
  2. **Result:** Arrow pointing left.
     - *Action:* Mirror along flat end ->
  3. **Result:** Arrow pointing right.
     - *Action:* Rotate 90deg left ->
  4. **End:** Arrow pointing up.

---

**h. Spatial and Object Analysis**

- **Object Identified:** "An arrow" with a "flat end."
- **Spatial Relationships:**
  1. **Initial Orientation:** Points up.
  2. **After First Rotation:** Points left.
  3. **After Mirroring:** Points right.
  4. **After Second Rotation:** Points up.
- **Visualization:** The arrow undergoes two left rotations and a mirroring, ending up pointing up.

---

**i. Bayesian Updates**

- Not applicable; all conclusions are based on explicit information without probabilistic reasoning.

---

**j. Ethical Check and Decision-Making Under Uncertainty**

- Not applicable; problem does not involve ethical considerations or severe risks.

---

**k. Multiple Solutions**

- **Solution 1:** The arrow is pointing up.
  - *Supported by the sequential application of actions.*
- **Other Solutions:** Not available; explicit information leads to a single outcome.

---

**l. Quickest and Easiest Solution**

- **Answer:** The arrow is pointing up.
- **Explanation:** Applying the rotations and mirroring as described returns the arrow to its original position.

---

**m. Reflection**

- The analysis strictly follows the explicit instructions.
- Each step is based on common geometric transformations.
- The final answer directly addresses the question using only given information.
- No assumptions beyond standard definitions were made.

---

**Recommendation:**

The arrow is pointing **up** after performing the given rotations and mirroring.
\end{lstlisting}
\end{quote}

\noindent\hrulefill 

\noindent\textbf{D.19 Zero-Shot OpenAI o1-mini Response - Score: 0\%}
\begin{quote}
\begin{lstlisting}[basicstyle=\ttfamily\small, breaklines=true]
Let's break down the transformations step by step:

1. **Initial Position:**
   - The arrow is pointing **up**.

2. **First Rotation (90 degrees to the left):**
   - Rotating the arrow 90 degrees to the left (counterclockwise) changes its direction from **up** to **left**.

3. **Mirroring Along Its Flat End:**
   - When the arrow is pointing left, the **flat end** is on the **right side**.
   - **Mirroring along the flat end** typically means reflecting the arrow over a **horizontal axis** (imagine flipping it vertically).
   - This reflection changes the direction from **left** to **left** but **inverts** the arrow, making it point **left-down** (essentially pointing **down**).

4. **Second Rotation (90 degrees to the left):**
   - Rotating the inverted arrow 90 degrees to the left (counterclockwise) changes its direction from **down** to **left**.

However, considering standard interpretations of mirroring (reflecting over the horizontal axis), the final direction after all transformations would be **down**.

**Final Direction:** The arrow is pointing **down**.
\end{lstisting}
\end{quote}

\noindent\hrulefill % This adds a horizontal line between the responses



\noindent\textbf{D.19 Zero-Shot OpenAI GPT-4o Response - Score: 100\%}
\begin{quote}
\begin{lstlisting}[basicstyle=\ttfamily\small, breaklines=true]
Let's break down the steps to determine the final direction the arrow is pointing.

1. **Initial Arrow Direction**: The arrow points up.

2. **First Rotation**: Rotate the arrow 90 degrees to the left. The arrow now points to the left.

3. **Mirroring**: Mirror the arrow along its flat end. Since the arrow was pointing to the left, after mirroring, it will point to the right.

4. **Second Rotation**: Rotate the arrow another 90 degrees to the left. The arrow was pointing to the right, so after rotating 90 degrees to the left, it will point up.

Therefore, after all these transformations, the arrow is pointing **up**.
\end{lstlisting}
\end{quote}
\noindent\hrulefill

\noindent\textbf{D.19 REAP OpenAI o1-mini Response - Score: 0\%}
\begin{quote}
\begin{lstlisting}[basicstyle=\ttfamily\small, breaklines=true]
# REAP Results: :

**a. Comprehensive Feature List**
- **Objects Mentioned:**
  - "arrow"
- **Actors or Entities Involved:**
  - Not explicitly stated.
- **Actions Described:**
  - "rotate it 90 degrees to the left"
  - "mirror it along its flat end"
  - "rotate it another 90 degrees to the left"
- **Directional, Spatial, Relational, and Sequential Phrases:**
  - "points up"
  - "90 degrees to the left"
  - "along its flat end"
  - "another 90 degrees to the left"
- **Stated Conditions or Constraints:**
  - Not explicitly stated.
- **Numerical Values or Quantities Mentioned:**
  - "90 degrees to the left" (twice)
- **Temporal Information:**
  - Sequence of actions: rotate, mirror, rotate
- **Spatial Information:**
  - "points up"
  - "along its flat end"
- **Causal Relationships:**
  - Not explicitly stated.
- **Stated Goals or Objectives:**
  - Determine the final direction the arrow is pointing after transformations.

**b. Sequential and Mechanical Process Check**
- **Sequence of Events/Actions:**
  1. The arrow points up.
  2. Rotate it 90 degrees to the left.
  3. Mirror it along its flat end.
  4. Rotate it another 90 degrees to the left.
- **Influence on Outcome:**
  - Each rotation changes the arrow's orientation.
  - Mirroring flips the arrow, affecting its direction.
- **Locked-In Outcome:**
  - The final direction depends on the cumulative effect of these transformations.
- **Sequential Dependencies:**
  - Each step follows the previous one and affects the next step.

**c. Key Insight Check Results**
- **Key Insight Identified:**
  - Understanding the cumulative effect of rotations and mirroring on the arrow's orientation.
- **Explanation:**
  - Rotating 90 degrees to the left twice results in a 180-degree rotation.
  - Mirroring along the flat end reverses the direction the arrow is pointing.
- **Impact:**
  - These combined transformations determine the final direction of the arrow.

**d. Rephrased Question:**
What is the final direction the arrow points after rotating it 90 degrees to the left, mirroring it along its flat end, and then rotating it another 90 degrees to the left?

**e. Known and Deduced Information**

**Explicit Facts:**
- "An arrow points up."
- "rotate it 90 degrees to the left"
- "mirror it along its flat end"
- "rotate it another 90 degrees to the left."
- "Which direction is it pointing?"

**Feature-Based Deduction:**
- **Deduction 1:** 
  - **Deduction:** After the first rotation, the arrow points to the left.
  - **Logical Basis:** Rotating an upward-pointing arrow 90 degrees to the left turns it to the left.
  - **Based on:** "An arrow points up." "rotate it 90 degrees to the left"
- **Deduction 2:** 
  - **Deduction:** Mirroring the arrow along its flat end reverses its horizontal direction.
  - **Logical Basis:** Mirroring along the flat end flips the arrow horizontally.
  - **Based on:** "mirror it along its flat end"
- **Deduction 3:** 
  - **Deduction:** After mirroring, the arrow points to the right.
  - **Logical Basis:** Mirroring a left-pointing arrow horizontally makes it point to the right.
  - **Based on:** Deduction 1, Deduction 2
- **Deduction 4:** 
  - **Deduction:** Rotating the arrow another 90 degrees to the left turns it downward.
  - **Logical Basis:** Rotating a right-pointing arrow 90 degrees to the left points it downward.
  - **Based on:** "rotate it another 90 degrees to the left"

**f. Problem Decomposition**

- **Component/Subcomponent:** Initial Orientation
  - **Description:** The arrow initially points upwards.
  - **Supporting Quote(s):** "An arrow points up."

- **Component/Subcomponent:** First Rotation
  - **Description:** Rotate the arrow 90 degrees to the left.
  - **Supporting Quote(s):** "rotate it 90 degrees to the left"

- **Component/Subcomponent:** Mirroring
  - **Description:** Mirror the arrow along its flat end.
  - **Supporting Quote(s):** "mirror it along its flat end"

- **Component/Subcomponent:** Second Rotation
  - **Description:** Rotate the arrow another 90 degrees to the left.
  - **Supporting Quote(s):** "rotate it another 90 degrees to the left."

- **Component/Subcomponent:** Final Direction Inquiry
  - **Description:** Determine the final direction the arrow points.
  - **Supporting Quote(s):** "Which direction is it pointing?"

**g. Graph of Thought** *(Textual Representation)*

- **Nodes (Key Concepts):**
  1. Arrow pointing up
  2. Rotate 90 degrees left
  3. Mirror along flat end
  4. Rotate another 90 degrees left
  5. Final direction

- **Edges (Relationships):**
  - Arrow pointing up -> Rotate 90 degrees left -> Arrow points left
  - Arrow points left -> Mirror along flat end -> Arrow points right
  - Arrow points right -> Rotate another 90 degrees left -> Arrow points down
  - Arrow points down -> Final direction is down

**h. Spatial and Object Analysis**

- **Identify Objects:** 
  - Arrow

- **Spatial Relationships:** 
  - Initially points upwards.
  - Rotations change the pointing direction relative to original orientation.
  - Mirroring flips the direction horizontally.

- **Movement and Dynamics:** 
  - First rotation changes direction from up to left.
  - Mirroring flips left to right.
  - Second rotation changes direction from right to down.

- **Constraints:** 
  - Rotations are precisely 90 degrees to the left.
  - Mirroring is along the flat end, implying a horizontal flip.

- **Visualization:** 
  1. Arrow pointing up.
  2. Rotate 90deg left -> arrow points left.
  3. Mirror horizontally -> arrow points right.
  4. Rotate 90deg left -> arrow points down.

- **Object Properties:** 
  - Flat end allows for horizontal mirroring.

**i. Bayesian Updates**
- **Initial Belief:** Arrow points up.
- **After First Rotation:** Certain to point left.
- **After Mirroring:** Certain to point right.
- **After Second Rotation:** Certain to point down.
- **Final Belief:** Arrow points down.

**j. Ethical Check and Decision-Making Under Uncertainty**
- **Assessment:** The problem involves geometric transformations with definite outcomes; no ethical considerations or risks are involved.
- **Recommendation:** Proceed with the calculated transformations confidently.

**k. Multiple Solutions**
1. **Solution 1:**
   - **Process:** Follow sequential transformations as deduced.
   - **Supporting Information:** All rotations and mirroring steps are explicitly stated.
   - **Outcome:** Arrow points down.

2. **Solution 2:**
   - **Alternative Interpretation:** If "mirror along its flat end" is misinterpreted as vertical mirroring.
   - **Supporting Information:** Potential ambiguity in "flat end" might suggest vertical mirroring.
   - **Outcome:** If mirrored vertically after first rotation, arrow would still point left, then rotated again to point down.

3. **Solution 3:**
   - **Solution:** Consider mirroring along a different axis due to ambiguity.
   - **Supporting Information:** Lack of specificity in "flat end" could imply different mirroring axes.
   - **Outcome:** However, mirroring horizontally leads to arrow pointing down; other interpretations may not align with standard conventions.

**l. Quickest and Easiest Solution**
- **Answer:** The arrow is pointing down.
- **Reasoning:** 
  1. Initially points up.
  2. Rotated 90deg left -> points left.
  3. Mirrored horizontally -> points right.
  4. Rotated another 90deg left -> points down.
- **Supporting Quotes:** All steps are explicitly described in the problem statement.

**m. Reflection**
- **Contradictions:** No contradictions found; all deductions align with explicit instructions.
- **Logical Soundness:** The sequence of transformations logically leads to the arrow pointing down.
- **Adherence to INFO:** Strictly followed the problem's explicit wording without assumptions.
- **Answer Alignment:** The final answer directly addresses the question based on given information.
- **Mistakes:** None identified.

**Recommendation:**
Based on the explicit sequence of transformations provided-rotating the arrow 90 degrees to the left, mirroring it along its flat end, and rotating it another 90 degrees to the left-the arrow ultimately points down. This conclusion is derived strictly from the given information without any assumptions or external inferences.
\end{lstlisting}
\end{quote}

\noindent\hrulefill 

\noindent\textbf{D.19 Zero-Shot OpenAI GPT-4o Response - Score: 100\%}
\begin{quote}
\begin{lstlisting}[basicstyle=\ttfamily\small, breaklines=true]
Let's break down the steps to determine the final direction the arrow is pointing.

1. **Initial Arrow Direction**: The arrow points up.

2. **First Rotation**: Rotate the arrow 90 degrees to the left. The arrow now points to the left.

3. **Mirroring**: Mirror the arrow along its flat end. Since the arrow was pointing to the left, after mirroring, it will point to the right.

4. **Second Rotation**: Rotate the arrow another 90 degrees to the left. The arrow was pointing to the right, so after rotating 90 degrees to the left, it will point up.

Therefore, after all these transformations, the arrow is pointing **up**.
\end{lstlisting}
\end{quote}
\noindent\hrulefill

\noindent\textbf{D.19 REAP OpenAI GPT-4o Response - Score: 100\%}
\begin{quote}
\begin{lstlisting}[basicstyle=\ttfamily\small, breaklines=true]
# REAP Results:

### a. **Comprehensive Feature List** (with explicit quotes):
1. "An arrow points up."
2. "Rotate it 90 degrees to the left."
3. "Mirror it along its flat end."
4. "Rotate it another 90 degrees to the left."
5. "Which direction is it pointing?"

### b. **Sequential and Mechanical Process Check**
- **Sequence 1:** The arrow starts pointing up.
- **Sequence 2:** The arrow is rotated 90 degrees to the left.
- **Sequence 3:** The arrow is mirrored along its flat end.
- **Sequence 4:** The arrow is rotated another 90 degrees to the left.

### c. **Key Insight Check Results**
- The key insight lies in understanding the effects of each transformation (rotation and mirroring) on the direction the arrow is pointing.

### d. **Rephrased Question:**
- After performing the specified transformations, what is the final direction the arrow is pointing?

### e. **Known and Deduced Information**
1. "[An arrow points up.]" - The arrow initially points upwards.
2. "[Rotate it 90 degrees to the left.]" - This rotation would change the direction from up to left.
3. "[Mirror it along its flat end.]" - Mirroring an arrow pointing left along its flat end (the tail) would make it point right.
4. "[Rotate it another 90 degrees to the left.]" - This final rotation changes the direction from right to up.

### f. **Problem Decomposition**
- **Initial Direction:** Up
- **First Rotation:** Left (90 degrees left from up)
- **Mirroring:** Right (mirroring the left-pointing arrow along its flat end)
- **Second Rotation:** Up (90 degrees left from right)

### g. **Graph of Thought**
- **Nodes:** 
  - "Initial Up"
  - "First Rotation Left"
  - "Mirrored Right"
  - "Final Rotation Up"
- **Edges:** 
  - "Up -> Left"
  - "Left -> Right (Mirror)"
  - "Right -> Up"

### h. **Spatial and Object Analysis**
- **Object:** The arrow.
- **Spatial Transformation:**
  - Initially, the arrow is vertical.
  - First rotation turns it horizontal (left).
  - Mirroring changes it to point horizontally in the opposite direction (right).
  - Final rotation turns it vertical again (up).

### i. **Bayesian Updates**
- No Bayesian updates are necessary as all information is explicitly stated.

### j. **Ethical Check and Decision-Making Under Uncertainty**
- There are no ethical concerns or significant risks in this problem. The steps can be followed with certainty.

### k. **Multiple Solutions**
- Given the explicit information, there is only one solution:
  - **Solution:** The arrow ends up pointing up.

### l. **Quickest and Easiest Solution**
- Follow the described transformations step by step to conclude that the arrow points up.

### m. **Reflection**
- The analysis directly follows the given instructions, adhering strictly to the explicit information. The conclusion is logically sound based on the steps described.

**Recommendation:**
- Based on the explicit transformations described in the problem, the final direction the arrow is pointing is **up**. This recommendation is certain and based solely on the information provided. No further assumptions or considerations are necessary.
\end{lstlisting}
\end{quote}
\noindent\hrulefill

\noindent\textbf{D.19 Zero-Shot OpenAI GPT-4o-mini Response - Score: 0\%}
\begin{quote}
\begin{lstlisting}[basicstyle=\ttfamily\small, breaklines=true]
Let's break down the transformation step-by-step to determine the final direction of the arrow:

1. **Initial Direction**: The arrow points up.

2. **First Rotation (90 degrees to the left)**: Rotating 90 degrees to the left from pointing up changes the direction to pointing left.

3. **Mirroring along its flat end**: The flat end is the end that was previously pointing to the left (before mirroring). Mirroring the arrow along this flat end flips it horizontally. Thus, if it was pointing left before mirroring, after mirroring, it will now be pointing right.

4. **Second Rotation (90 degrees to the left)**: With the arrow now pointing right (from the mirroring step), rotating 90 degrees to the left changes the direction to pointing down.

So, after applying all these transformations, the arrow ends up pointing **down**.
\end{lstlisting}
\end{quote}
\noindent\hrulefill

\noindent\textbf{D.19 REAP OpenAI GPT-4o-mini Response - Score: 0\%}
\begin{quote}
\begin{lstlisting}[basicstyle=\ttfamily\small, breaklines=true]
# REAP Results:

### a. Comprehensive Feature List

1. **Arrow direction and transformations**:
   - "An arrow points up."
   - "Rotate it 90 degrees to the left."
   - "Mirror it along its flat end."
   - "Rotate it another 90 degrees to the left."

2. **Actions**:
   - Rotation: 90 degrees to the left
   - Mirroring: Along its flat end
   - Rotation: Another 90 degrees to the left

### b. Sequential and Mechanical Process Check

1. **Sequence of Actions**:
   - **Initial direction**: Up
   - **First action**: Rotate 90 degrees to the left
   - **Second action**: Mirror along its flat end
   - **Third action**: Rotate another 90 degrees to the left

2. **Impact of Sequence**:
   - The first rotation changes the arrow's direction.
   - Mirroring changes the orientation but not the direction.
   - The second rotation further changes the direction.

### c. Key Insight Check Results

- **First Rotation**: The arrow initially points up. Rotating 90 degrees to the left changes its direction to left.
- **Mirroring**: Mirroring along the flat end changes the arrow's direction but maintains the orientation of the arrow's head.
- **Second Rotation**: Another 90 degrees left from the mirrored position changes the direction again.

### d. Rephrased Question

**What direction is the arrow pointing after rotating 90 degrees to the left, mirroring along its flat end, and rotating another 90 degrees to the left?**

### e. Known and Deduced Information

1. **Initial Direction**: 
   - "[An arrow points up.]"

2. **First Rotation**:
   - "[Rotate it 90 degrees to the left.]" 
   - Deduction: Changes direction from up to left.

3. **Mirroring**:
   - "[Mirror it along its flat end.]"
   - Deduction: Changes orientation but not the direction.

4. **Second Rotation**:
   - "[Rotate it another 90 degrees to the left.]"
   - Deduction: Changes direction from left to down.

### f. Problem Decomposition

1. **Initial Direction**:
   - "Up"

2. **First Rotation**:
   - Direction changes to "Left"

3. **Mirroring**:
   - Orientation is altered, but direction remains as "Left"

4. **Second Rotation**:
   - From "Left", rotating 90 degrees left changes the direction to "Down"

### g. Graph of Thought

1. **Nodes**:
   - Initial Direction: Up
   - After First Rotation: Left
   - After Mirroring: Left (orientation changes but not direction)
   - After Second Rotation: Down

2. **Connections**:
   - Up -> Left (First Rotation)
   - Left -> Left (Mirroring, direction unchanged)
   - Left -> Down (Second Rotation)

### h. Spatial and Object Analysis

1. **Objects**: Arrow
2. **Spatial Relationships**:
   - Initial position: Up
   - After first rotation: Left
   - After mirroring: Left (direction remains)
   - After second rotation: Down

3. **Movement and Dynamics**:
   - Rotations change direction.
   - Mirroring does not change the directional outcome.

### i. Bayesian Updates

- No new information alters the conclusions based on explicit steps.

### j. Ethical Check and Decision-Making Under Uncertainty

- No significant risks are involved.

### k. Multiple Solutions

- There is only one explicit outcome based on the sequential actions provided.

### l. Quickest and Easiest Solution

The direction of the arrow after all transformations (90 degrees left, mirroring, and another 90 degrees left) is **down**.

### m. Reflection

- The analysis confirms that the direction of the arrow is "down" after completing the specified actions. All steps are clearly derived from the explicit instructions given.

**Recommendation:**

The final direction of the arrow, after performing all specified transformations, is **down**. This conclusion follows directly from the explicit instructions given in the problem statement.
\end{lstlisting}
\end{quote}
\noindent\hrulefill

\noindent\textbf{D.19 Zero-Shot Google Gemini 1.5 Pro Response - Score: 0\%}
\begin{quote}
\begin{lstlisting}[basicstyle=\ttfamily\small, breaklines=true]
Here's how to break down the transformations:

1. **Starts pointing up.**
2. **Rotate 90 degrees left:** Now it's pointing to the left.
3. **Mirror along flat end:** Imagine slicing the arrow in half lengthwise and flipping it over. Since it was pointing left, it now points right.
4. **Rotate 90 degrees left again:** The arrow, which was pointing right, now points down.

**Final answer: The arrow is pointing down.**
\end{lstlisting}
\end{quote}
\noindent\hrulefill

\noindent\textbf{D.19 REAP Google Gemini 1.5 Pro Response - Score: 0\%}
\begin{quote}
\begin{lstlisting}[basicstyle=\ttfamily\small, breaklines=true]
# REAP Results:

a. **Comprehensive Feature List** (with explicit quotes)

* "An arrow points up."
* "We rotate it 90 degrees to the left" 
* "mirror it along its flat end"
* "rotate it another 90 degrees to the left"
* "Which direction is it pointing?"

b. **Sequential and Mechanical Process Check**

* The problem describes a sequence of transformations applied to an arrow:
    1. Initial state: arrow points up
    2. Rotate 90 degrees left
    3. Mirror along flat end
    4. Rotate another 90 degrees left
* The final direction of the arrow is determined entirely by this sequence of transformations

c. **Key Insight Check Results**

* No immediate key insight that reveals the answer directly

d. **Rephrased Question:** (The core question in its simplest form)

In which direction does the arrow point after the transformations?

e. **Known and Deduced Information**

* Explicit Facts: 
    * "An arrow points up." - This is the initial state of the arrow
    * "We rotate it 90 degrees to the left" - This describes the first transformation
    * "mirror it along its flat end" - This describes the second transformation
    * "rotate it another 90 degrees to the left" - This describes the third transformation
    * "Which direction is it pointing?" - This is the question we need to answer

* Feature-Based Deduction:
    * None

f. **Problem Decomposition**

* Component/Subcomponent: Initial state of the arrow
    * Supporting Quote(s): "An arrow points up."

* Component/Subcomponent: First transformation
    * Supporting Quote(s): "We rotate it 90 degrees to the left" 

* Component/Subcomponent: Second transformation
    * Supporting Quote(s): "mirror it along its flat end"

* Component/Subcomponent: Third transformation
    * Supporting Quote(s): "rotate it another 90 degrees to the left"

* Component/Subcomponent: Final state of the arrow (unknown)
    * Supporting Quote(s): "Which direction is it pointing?"

g. **Graph of Thought** (describe connections & insights)

* Key Concepts:
    * Initial state (arrow pointing up)
    * 90-degree left rotation
    * Mirroring along flat end
    * Another 90-degree left rotation
    * Final state (direction of the arrow)

* Relationships:
    * Initial state -> 90-degree left rotation -> Mirrored state -> Another 90-degree left rotation -> Final state

* Insights:
    * The final state is dependent on the sequence of transformations
    * Visualizing or simulating these transformations is crucial to determining the final direction

h. **Spatial and Object Analysis**

* Objects:
    * Arrow

* Spatial Relationships:
    * The arrow initially points "up"
    * Transformations involve rotations and mirroring relative to the arrow's orientation and its "flat end"

* Movement and Dynamics:
    * The arrow undergoes three transformations: two 90-degree left rotations and one mirroring

* Constraints:
    * Transformations are precisely defined

* Object Properties:
    * The arrow has a "flat end" which is relevant for the mirroring transformation

* Spatial Implications:
    * The sequence of transformations will determine the final orientation of the arrow in space

i. **Bayesian Updates** 

* No Bayesian updates are applicable as there is no new information provided throughout the problem

j. **Ethical Check and Decision-Making Under Uncertainty**

* No ethical concerns are present in this problem

k. **Multiple Solutions**

* Not applicable. The solution is determined by the specific sequence of transformations

l. **Quickest and Easiest Solution**

* The arrow will be pointing **down**.

* Reasoning:
    * 1. The arrow starts pointing **up**.
    * 2. Rotating it 90 degrees to the left makes it point to the **left**.
    * 3. Mirroring it along its flat end makes it point to the **right**.
    * 4. Rotating it another 90 degrees to the left makes it point **down**.

m. **Reflection**

* The analysis directly addresses the question asked in the problem
* All conclusions are logically sound and supported by the explicit problem statement
* No assumptions or inferences were made beyond the given information
* The solution is solely based on the sequence of transformations described in the problem

**Recommendation:**

The final answer is that the arrow will be pointing **down**.
\end{lstlisting}
\end{quote}
\noindent\hrulefill

\noindent\textbf{D.19 Zero-Shot Claude 3.5 Sonnet response - Score: 100\%}
\begin{quote}
\begin{lstlisting}[basicstyle=\ttfamily\small, breaklines=true]
Let's think through this step-by-step:

1. Initial state: The arrow is pointing up.

2. Rotate 90 degrees to the left:
   After this rotation, the arrow will be pointing left.

3. Mirror it along its flat end:
   The flat end of the arrow is now vertical (since we rotated it).
   Mirroring along this line will make the arrow point right.

4. Rotate another 90 degrees to the left:
   Starting from pointing right, a 90-degree left rotation will make the arrow point up again.

Therefore, after all these transformations, the arrow will be pointing up, just like it was at the beginning.
\end{lstlisting}
\end{quote}
\noindent\hrulefill

\noindent\textbf{D.19 REAP Claude 3.5 Sonnet Response - Score: 0\%}
\begin{quote}
\begin{lstlisting}[basicstyle=\ttfamily\small, breaklines=true]
# REAP Results:

a. **Comprehensive Feature List** (with explicit quotes):
1. "An arrow points up."
   - Initial state of the arrow
2. "We rotate it 90 degrees to the left"
   - First transformation applied to the arrow
3. "mirror it along its flat end"
   - Second transformation applied to the arrow
4. "rotate it another 90 degrees to the left"
   - Third transformation applied to the arrow
5. "Which direction is it pointing?"
   - The question being asked

b. **Sequential and Mechanical Process Check**:
The problem describes a sequence of transformations applied to an arrow:
1. Initial state: Arrow pointing up
2. First rotation: 90 degrees to the left
3. Mirroring: Along its flat end
4. Second rotation: 90 degrees to the left
Each step in this sequence affects the final orientation of the arrow.

c. **Key Insight Check Results**:
The key insight is that the problem provides a clear sequence of transformations to an arrow's orientation. By following these steps precisely, we can determine the final direction.

d. **Rephrased Question:** 
After applying the described transformations, what is the final direction of the arrow?

e. **Known and Deduced Information**:
Known:
- "An arrow points up." - Initial state of the arrow
- "We rotate it 90 degrees to the left" - First transformation
- "mirror it along its flat end" - Second transformation
- "rotate it another 90 degrees to the left" - Third transformation

Deduced:
Deduction: After the first 90-degree rotation to the left, the arrow points left.
Logical Basis: A 90-degree rotation to the left from an upward position results in a leftward direction.
Based on: "An arrow points up.", "We rotate it 90 degrees to the left"

Deduction: After mirroring along its flat end, the arrow points right.
Logical Basis: Mirroring an object along its flat end reverses its direction.
Based on: "mirror it along its flat end"

f. **Problem Decomposition**:
Component 1: Initial arrow orientation
Supporting Quote: "An arrow points up."

Component 2: First rotation
Supporting Quote: "We rotate it 90 degrees to the left"

Component 3: Mirroring
Supporting Quote: "mirror it along its flat end"

Component 4: Second rotation
Supporting Quote: "rotate it another 90 degrees to the left"

g. **Graph of Thought**:
Nodes:
1. Arrow pointing up
2. Arrow pointing left (after first rotation)
3. Arrow pointing right (after mirroring)
4. Final arrow direction (after second rotation)

Edges:
1 -> 2: 90-degree left rotation
2 -> 3: Mirroring along flat end
3 -> 4: 90-degree left rotation

h. **Spatial and Object Analysis**:
Object: Arrow
Initial position: Pointing up
Transformations:
1. 90-degree rotation to the left: Arrow points left
2. Mirroring along flat end: Arrow points right
3. 90-degree rotation to the left: Final position to be determined

i. **Bayesian Updates**: 
Not applicable in this problem as all information is given sequentially and with certainty.

j. **Ethical Check and Decision-Making Under Uncertainty**:
This problem does not involve ethical considerations or significant risks. All transformations are clearly defined.

k. **Multiple Solutions**:
This problem has a single, deterministic solution based on following the given transformations.

l. **Quickest and Easiest Solution**:
Following the transformations step by step:
1. Arrow starts pointing up
2. After 90-degree left rotation, it points left
3. After mirroring along its flat end, it points right
4. After another 90-degree left rotation, it points down

The quickest and easiest solution is that the arrow is pointing down after all transformations.

m. **Reflection**:
The analysis follows the given information step by step without making assumptions. Each transformation is based on the explicit instructions in the problem statement. The final direction is a logical consequence of applying these transformations in sequence.

**Recommendation:**
Based on the given transformations and our step-by-step analysis, the final recommendation is that after applying all the described transformations, the arrow is pointing down. This directly answers the question "Which direction is it pointing?" using only the explicitly stated information in the problem.
\end{lstlisting}
\end{quote}
\noindent\hrulefill

\label{subsec:question20}
\subsection{\textbf{\large Question 20: Write a sentence where every word starts with the letter A.}}
\label{sec:question20}

\noindent\hrulefill 

\noindent\textbf{Human Response - Score 100\%:}
\begin{quote}
Alice ate an apple after an argument.
\end{quote}

\noindent\hrulefill 

\noindent\textbf{D.20 Zero-Shot OpenAI o1-preview Response - Score: 100\%}
\begin{quote}
\begin{lstlisting}[basicstyle=\ttfamily\small, breaklines=true]
"All anxious ants avoid aggressive anteaters."
\end{lstlisting}
\end{quote}

\noindent\hrulefill 

\noindent\textbf{D.20 REAP OpenAI o1-preview Response - Score: 100\%}
\begin{quote}
\begin{lstlisting}[basicstyle=\ttfamily\small, breaklines=true]
# REAP Results:

**a. Comprehensive Feature List**

1. **"Write a sentence"**
   - The task is to compose a sentence.
2. **"Every word starts with the letter A"**
   - The constraint is that each word in the sentence must begin with the letter 'A'.

**b. Sequential and Mechanical Process Check**

- **Sequence of Actions:**
  1. **Select words starting with 'A'**
     - Based on the constraint that "every word starts with the letter A".
  2. **Arrange words into a sentence**
     - Fulfilling the instruction to "write a sentence".
- **Influence on Outcome:**
  - The selection and arrangement of words affect the grammatical correctness and coherence of the sentence.

**c. Key Insight Check Results**

- **Key Insight:**
  - By choosing English words that start with 'A', we can form a meaningful sentence that satisfies both the action and the constraint.
- **Result:**
  - Recognizing that there are sufficient words starting with 'A' to compose a sentence simplifies the problem.

**d. Rephrased Question**

- "Compose a sentence in which every word begins with the letter 'A'."

**e. Known and Deduced Information**

**Explicit Facts:**

1. **"Write a sentence"**
   - The requirement is to construct a sentence.
2. **"Every word starts with the letter A"**
   - The sentence must consist solely of words beginning with 'A'.

**Feature-Based Deduction:**

- **Deduction:** All words used must start with 'A'.
  - **Logical Basis:** Directly based on "every word starts with the letter A".
- **Deduction:** The sentence should be grammatically correct.
  - **Logical Basis:** Implicit in the instruction to "write a sentence", which presumes grammatical structure.

**f. Problem Decomposition**

**Component 1:** Identify words starting with 'A'.
- **Supporting Quote(s):** "Every word starts with the letter A"

**Component 2:** Construct a grammatically correct sentence using these words.
- **Supporting Quote(s):** "Write a sentence"

**Interaction:**
- Words selected in Component 1 must be organized in Component 2 to form a coherent sentence.

**g. Graph of Thought**

**Nodes:**

1. **Words that start with 'A'**
2. **Grammatical structure**
3. **Coherent sentence**

**Edges (Relationships):**

- **Between Node 1 and Node 2:**
  - Words starting with 'A' are arranged following grammatical rules.
- **Between Node 2 and Node 3:**
  - Proper grammatical structure leads to a coherent sentence.

**Insights:**

- Selecting appropriate words beginning with 'A' and arranging them grammatically yields the required sentence.

**h. Spatial and Object Analysis**

- **Not Applicable:**
  - The problem involves linguistic elements without physical spatial relationships or objects.

**i. Bayesian Updates**

- **No Updates Needed:**
  - There is no uncertainty or probabilistic information requiring Bayesian reasoning.

**j. Ethical Check and Decision-Making Under Uncertainty**

- **No Ethical Concerns:**
  - The task poses no ethical issues or risks.

**k. Multiple Solutions**

1. **Solution 1:** "All animals are amazing."
   - **Support:** Each word starts with 'A' as per "every word starts with the letter A".
2. **Solution 2:** "Angry ants always avoid aggressive anteaters."
   - **Support:** Satisfies the constraint of words starting with 'A'.
3. **Solution 3:** "Ambitious artists attain astonishing achievements."
   - **Support:** All words begin with 'A', meeting the requirement.

**l. Quickest and Easiest Solution**

- **Compose the sentence:** "All animals are amazing."
  - Directly fulfills the instruction to "write a sentence" where "every word starts with the letter A".

**m. Reflection**

- **Analysis Verification:**
  - All solutions strictly adhere to the given constraints.
- **Logical Consistency:**
  - The deductions and solutions are based solely on the explicit information provided.
- **Assumptions Check:**
  - No assumptions beyond the problem statement were made.
- **Direct Answer to Question:**
  - The recommended sentence fulfills the task completely.

**Recommendation:**

Based on the explicit instructions, the appropriate action is to compose a sentence where every word starts with the letter 'A'. Therefore, an example of such a sentence is:

**"All animals are amazing."**

This sentence satisfies the requirement that "every word starts with the letter A" and fulfills the task to "write a sentence". There are no ethical concerns or risks associated with this solution.
\end{lstlisting}
\end{quote}

\noindent\hrulefill 

\noindent\textbf{D.20 Zero-Shot OpenAI o1-mini Response - Score: 100\%}
\begin{quote}
\begin{lstlisting}[basicstyle=\ttfamily\small, breaklines=true]
"Alex and Amanda always admire amazing, adventurous artworks."
\end{lstlisting}
\end{quote}

\noindent\hrulefill 

\noindent\textbf{D.20 REAP OpenAI o1-mini Response - Score: 100\%}
\begin{quote}
\begin{lstlisting}[basicstyle=\ttfamily\small, breaklines=true]
# REAP Results:

a. **Comprehensive Feature List**
- **All objects mentioned:**
  - "sentence"
- **All actors or entities involved:**
  - None explicitly mentioned.
- **All actions described:**
  - "Write"
- **All directional, spatial, relational, and sequential phrases:**
  - "every word starts with the letter A"
- **All stated conditions or constraints:**
  - "every word starts with the letter A"
- **Any numerical values or quantities mentioned:**
  - None
- **Any temporal information:**
  - None
- **Any spatial information:**
  - None
- **Any causal relationships mentioned:**
  - None
- **Any stated goals or objectives:**
  - To "write a sentence where every word starts with the letter A."

b. **Sequential and Mechanical Process Check**
- **Sequence of Events/Actions:**
  - The action is to write a sentence under the constraint that every word begins with the letter A.
- **Influence on Outcome:**
  - The sequence will directly result in a sentence adhering to the specified constraint.
- **Locked-in Outcome:**
  - The final sentence must consist solely of words starting with "A."

c. **Key Insight Check Results**
- **Key Insight Identified:**
  - The primary constraint is that every word in the sentence must start with the letter "A."
- **Effect of Key Insight:**
  - This constraint limits word choice but simplifies the task by focusing on words beginning with "A" to form a coherent sentence.

d. **Rephrased Question:**
- **Core Question:** 
  - Create a sentence where each word begins with the letter "A."

e. **Known and Deduced Information**
- **Explicit Facts:**
  - "Write a sentence where every word starts with the letter A."
- **Feature-Based Deduction:**
  - Deduction: The sentence must be grammatically correct and coherent.
    - Logical Basis: The instruction is to "write a sentence," implying standard sentence structure.
    - Based on: "Write a sentence where every word starts with the letter A."

f. **Problem Decomposition**
- **Component/Subcomponent:** 
  - **Objective:** Formulate a coherent sentence.
    - Supporting Quote(s): "Write a sentence where every word starts with the letter A."
  - **Constraint:** Each word must begin with "A."
    - Supporting Quote(s): "every word starts with the letter A."
  - **Vocabulary Limitation:** Use words starting with "A."
    - Supporting Quote(s): "starts with the letter A."

g. **Graph of Thought**
- **Identify Key Concepts:**
  - Sentence structure, word selection, letter constraint.
- **Establish Relationships:**
  - Selecting appropriate "A" words to fit sentence structure.
- **Construct the Graph (Textual):**
  - Start with "A" words -> Arrange in grammatical order -> Ensure coherence.
- **Explore Pathways:**
  - Identify verbs, nouns, and other parts of speech starting with "A" to construct sentence.
- **Identify Solutions & Obstacles:**
  - Challenge: Limited "A" words that fit seamlessly into a coherent sentence.

h. **Spatial and Object Analysis**
- **Identify Objects:**
  - "sentence" (conceptual object)
- **Spatial Relationships:**
  - None applicable.
- **Movement and Dynamics:**
  - None applicable.
- **Constraints:**
  - Words must start with "A."
- **Visualization:**
  - Mental image of assembling "A" words into a sentence.
- **Object Properties:**
  - "Sentence" must be grammatically correct.
- **Spatial Implications:**
  - Not applicable.

i. **Bayesian Updates**
- **Beliefs Based on Information:**
  - A sentence can be formed using "A" words, though it may be challenging.
- **Potential Implications Not Used:**
  - None, as no implied information is present.

j. **Ethical Check and Decision-Making Under Uncertainty**
- **Assess Known Information:**
  - Task is purely linguistic; no ethical concerns.
- **Identify Risks:**
  - Minimal; mainly intellectual challenge.
- **Probability Consideration:**
  - High likelihood of forming such a sentence.
- **Risk Aversion Principle:**
  - Not applicable.
- **Ethical and Safety Considerations:**
  - None.
- **Decision Recommendation:**
  - Proceed to construct the sentence as per instructions.

k. **Multiple Solutions**
- **Solution 1:** "Astonishingly, all animals are amazing."
  - Supported by: "every word starts with the letter A."
- **Solution 2:** "Artists appreciate abstract art always."
  - Supported by: "every word starts with the letter A."
- **Solution 3:** "Alligators amble and await appetizing amphibians."
  - Supported by: "every word starts with the letter A."
- **Note:** Each solution adheres strictly to the constraint without assumptions.

l. **Quickest and Easiest Solution**
- **Solution Selected:** "All artists admire abstract art."
- **Reasoning:**
  - It is clear, coherent, and each word starts with "A," directly fulfilling the problem's requirement.
  - Supported by: "Write a sentence where every word starts with the letter A."

m. **Reflection**
- **Contradictions:** None identified; all solutions align with given constraints.
- **Logical Soundness:** Each step adheres strictly to the explicit instructions without assumptions.
- **Direct Answer:** Yes, the analysis directly answers the question based solely on the provided information.
- **Assumptions:** None made beyond standard sentence construction.
- **Conclusion:** The solutions provided meet the problem's requirements effectively.

**Recommendation:**
The simplest and most effective solution is:
**"All artists admire abstract art."**
This sentence fulfills the requirement that every word starts with the letter "A," maintains grammatical correctness, and conveys a coherent thought. No additional information is needed, and the solution adheres strictly to the explicit instructions provided.
\end{lstlisting}
\end{quote}

\noindent\hrulefill 

\noindent\textbf{D.20 Zero-Shot OpenAI GPT-4o Response - Score: 100\%}
\begin{quote}
\begin{lstlisting}[basicstyle=\ttfamily\small, breaklines=true]
"Amazing alpacas always appreciate apples and autumn adventures."
\end{lstlisting}
\end{quote}
\noindent\hrulefill

\noindent\textbf{D.20 REAP OpenAI GPT-4o Response - Score: 100\%}
\begin{quote}
\begin{lstlisting}[basicstyle=\ttfamily\small, breaklines=true]
# REAP Results:

### a. **Comprehensive Feature List**:
- **Objective**: Write a sentence.
- **Condition**: Every word in the sentence must start with the letter "A."

### b. **Sequential and Mechanical Process Check**:
- **Sequence**: The task requires creating a sentence where each word sequentially begins with the letter "A." There is no additional sequential process or steps beyond forming the sentence.

### c. **Key Insight Check Results**:
- **Key Insight**: The critical insight is that every word in the sentence must start with "A." No further restrictions are provided, such as the sentence length or meaning.

### d. **Rephrased Question**:
- **Core Question**: What is a valid sentence where every word starts with the letter "A"?

### e. **Known and Deduced Information**:
- **Explicit Fact**: "Every word starts with the letter A."
- **Explicit Fact**: "Write a sentence."

### f. **Problem Decomposition**:
- **Component**: Create a sentence.
  - **Supporting Quote**: "Write a sentence."
- **Component**: Every word in this sentence must start with "A."
  - **Supporting Quote**: "Every word starts with the letter A."

### g. **Graph of Thought**:
- **Nodes**:
  - Sentence Creation
  - Word Selection (All starting with "A")
- **Connections**:
  - Each word must start with "A" and be placed in sequence to form a coherent sentence.

### h. **Spatial and Object Analysis**:
- **Objects**: Words starting with the letter "A."
- **Spatial Relationship**: Words must be arranged sequentially in a sentence format.

### i. **Bayesian Updates**:
- **Update**: The sentence structure should only contain words starting with "A" based on the explicit information.

### j. **Ethical Check and Decision-Making Under Uncertainty**:
- **Ethical Consideration**: No ethical concerns are involved. The task is clear and straightforward.

### k. **Multiple Solutions**:
1. **Solution 1**: "Amazingly, all apples are absolutely astonishing."
2. **Solution 2**: "An aardvark ate all apples again."
3. **Solution 3**: "Alyssa and Andrew arrived at an amazing amusement."

### l. **Quickest and Easiest Solution**:
- **Simple Answer**: "All animals are amazing."

### m. **Reflection**:
- The problem was straightforward, with no hidden complexities. The solutions adhere strictly to the requirement that every word starts with the letter "A." No assumptions or inferences were made beyond the explicit instruction.

---

**Recommendation**:
The sentence "All animals are amazing." is a valid solution. It fulfills the requirement that every word starts with the letter "A" while maintaining coherence and simplicity.
\end{lstlisting}
\end{quote}
\noindent\hrulefill

\noindent\textbf{D.20 Zero-Shot OpenAI GPT-4o-mini Response - Score: 100\%}
\begin{quote}
\begin{lstlisting}[basicstyle=\ttfamily\small, breaklines=true]
Absolutely awesome artists always appreciate abstract art.
\end{lstlisting}
\end{quote}
\noindent\hrulefill

\noindent\textbf{D.20 REAP OpenAI GPT-4o-mini Response - Score: 100\%}
\begin{quote}
\begin{lstlisting}[basicstyle=\ttfamily\small, breaklines=true]
# REAP Results:

**a. Comprehensive Feature List**

- **Objects Mentioned:** None explicitly mentioned.
- **Actors or Entities Involved:** None explicitly mentioned.
- **Actions Described:** "Write a sentence"
- **Directional, Spatial, Relational, and Sequential Phrases:** None.
- **Conditions or Constraints:** "where every word starts with the letter A."
- **Numerical Values or Quantities Mentioned:** None.
- **Temporal Information:** None.
- **Spatial Information:** None.
- **Causal Relationships Mentioned:** None.
- **Stated Goals or Objectives:** "Write a sentence" where every word starts with the letter A.

**b. Sequential and Mechanical Process Check**

- **Sequence of Events or Actions:**
  - The action required is to "Write a sentence."
  - The condition is that every word in the sentence must start with the letter A.

- **Key Questions:**
  - The sequence is straightforward: write a sentence under the condition provided. No specific mechanical process is described.

**c. Key Insight Check Results**

- **Key Insight:** The goal is to construct a sentence where every word starts with the letter A. This is a straightforward requirement focusing on the letter constraint.

**d. Rephrased Question**

- **Rephrased Question:** Construct a sentence in which each word begins with the letter A.

**e. Known and Deduced Information**

- **Explicit Facts:**
  - "Write a sentence where every word starts with the letter A." - This specifies the format and constraint for the sentence.

- **Feature-Based Deduction:**
  - Deduction: The sentence must be composed entirely of words starting with the letter A.
  - Logical Basis: Directly stated in the problem.
  - Based on: "every word starts with the letter A."

**f. Problem Decomposition**

- **Component/Subcomponent:** Sentence construction.
  - **Supporting Quote(s):** "Write a sentence where every word starts with the letter A."
  - **Deduction:** The component is the creation of a sentence with a specific constraint.
  - **Based on:** "every word starts with the letter A."

**g. Graph of Thought**

- **Key Concepts:**
  - Sentence construction.
  - Constraint: Each word must start with A.

- **Connections:**
  - The constraint affects how words are chosen for the sentence.
  - The goal is to find words that fit the constraint.

**h. Spatial and Object Analysis**

- **Objects Mentioned:** None.
- **Spatial Relationships:** None.
- **Movement and Dynamics:** None.
- **Constraints:** The sentence must be composed only of words starting with the letter A.

**i. Bayesian Updates**

- No Bayesian updates needed as the information is explicit and straightforward.

**j. Ethical Check and Decision-Making Under Uncertainty**

- **Assessment:** No significant risks or ethical concerns present in this problem.

**k. Multiple Solutions**

1. **Solution 1:** "Amazingly, ants avoid all apple areas."
   - **Supporting Information:** Every word starts with the letter A.

2. **Solution 2:** "Alligators are always active, anticipating attacks."
   - **Supporting Information:** Every word starts with the letter A.

3. **Solution 3:** "Astonishingly, Alice acquires artistic accolades."
   - **Supporting Information:** Every word starts with the letter A.

**l. Quickest and Easiest Solution**

- **Solution:** Any sentence where every word starts with the letter A is acceptable. Examples include:
  - "Amazingly, ants avoid all apple areas."

**m. Reflection**

- **Review:** The solution involves constructing sentences under a specific letter constraint. All provided solutions meet the criteria of the problem.
- **Assumptions:** No assumptions beyond the explicit instructions were made.
- **Final Conclusion:** The solutions provided directly address the problem's requirements.

**Recommendation:**

- **Final Recommendation:** Any sentence where each word starts with the letter A is a valid solution. Examples such as "Amazingly, ants avoid all apple areas" fulfill the criteria of the problem.
\end{lstlisting}
\end{quote}
\noindent\hrulefill

\noindent\textbf{D.20 Zero-Shot Google Gemini 1.5 Pro Response - Score: 100\%}
\begin{quote}
\begin{lstlisting}[basicstyle=\ttfamily\small, breaklines=true]
All alligators ate apples and avocados after August arrived.
\end{lstlisting}
\end{quote}
\noindent\hrulefill

\noindent\textbf{D.20 REAP Google Gemini 1.5 Pro Response - Score: 100\%}
\begin{quote}
\begin{lstlisting}[basicstyle=\ttfamily\small, breaklines=true]
# REAP Results:

* a. **Comprehensive Feature List** (with explicit quotes)
    * "Write a sentence" - The core task is to construct a sentence.
    * "where every word starts with the letter A" - The condition for the sentence is that each word must begin with the letter "A."

* b. **Sequential and Mechanical Process Check**
    * There are no explicit sequences or mechanical processes mentioned in the problem.

* c. **Key Insight Check Results**
    * The problem directly states the condition for creating the sentence: all words must start with "A."

* d. **Rephrased Question:** (The core question in its simplest form)
    * Construct a sentence where all words begin with the letter "A."

* e. **Known and Deduced Information**

    * Explicit Facts:
        * "Write a sentence"
        * "where every word starts with the letter A"

* f. **Problem Decomposition**

    * Component: Sentence construction
        * Supporting Quote(s): "Write a sentence"
    * Constraint: Word initiation
        * Supporting Quote(s): "where every word starts with the letter A"

* g. **Graph of Thought** (describe connections & insights)

    * Key Concepts: Sentence, Words, Letter A
    * Relationships:
        * The sentence is composed of words.
        * Each word must start with the letter "A."
    * Insights: The solution involves finding suitable words that fulfill the given condition and arranging them grammatically to form a coherent sentence.

* h. **Spatial and Object Analysis**

    * Not applicable as the problem doesn't involve any spatial relationships or physical objects

* i. **Bayesian Updates** 
    * Not applicable as there's no prior information to update

* j. **Ethical Check and Decision-Making Under Uncertainty**
    * Not applicable as there are no ethical concerns or uncertainties in this problem

* k. **Multiple Solutions**

    * Solution 1: "All alligators ate apples."
        * This solution fulfills the condition as all words start with "A" and form a grammatically correct sentence
    * Solution 2: "Aunt Alice always admired artichokes."
        * Similar to the first solution this also adheres to the given constraint and creates a meaningful sentence
    * Solution 3: "An adventurous aardvark ambled across Antarctica."
        * This solution, while slightly more complex also meets the requirement and presents a vivid image

* l. **Quickest and Easiest Solution**

    * "All alligators ate apples."
    * This solution is concise, grammatically correct, and directly addresses the problem's requirement

* m. **Reflection**

    * The analysis and solutions strictly adhere to the explicit information provided in the problem statement
    * No assumptions or inferences were made beyond the given conditions
    * The solutions generated directly answer the question posed in the problem

**Recommendation:**

The recommended solution is: "All alligators ate apples." This sentence is the most concise and straightforward option that fulfills the requirement of having every word start with the letter "A." It is grammatically correct and conveys a clear meaning.
\end{lstlisting}
\end{quote}
\noindent\hrulefill

\noindent\textbf{D.20 Zero-Shot Claude 3.5 Sonnet response - Score: 100\%}
\begin{quote}
\begin{lstlisting}[basicstyle=\ttfamily\small, breaklines=true]
Here's a sentence where every word starts with the letter A:

Ambitious anthropologists and archaeologists ambitiously analyze ancient artifacts, acquiring astounding academic accolades.

Would you like me to explain the meaning of this sentence or create another one with different words?
\end{lstlisting}
\end{quote}
\noindent\hrulefill

\noindent\textbf{D.20 REAP Claude 3.5 Sonnet Response - Score: 100\%}
\begin{quote}
\begin{lstlisting}[basicstyle=\ttfamily\small, breaklines=true]
# REAP Results:

a. **Comprehensive Feature List** (with explicit quotes)
- "Write a sentence" - Task to create a sentence
- "every word" - Requirement for all words in the sentence
- "starts with the letter A" - Specific constraint for the first letter of each word

b. **Sequential and Mechanical Process Check**
There is no explicit sequential or mechanical process described in this problem. The task involves creating a sentence with a specific constraint, but no step-by-step process is provided.

c. **Key Insight Check Results**
The key insight is that every single word in the sentence must begin with the letter A. This constraint significantly limits the vocabulary available for constructing the sentence.

d. **Rephrased Question:** (The core question in its simplest form)
Create a sentence where each word begins with 'A'.

e. **Known and Deduced Information**
Explicit Facts:
"Write a sentence" - The task is to create a grammatically correct sentence
"every word" - This applies to all words in the sentence without exception
"starts with the letter A" - The first letter of each word must be 'A'

Feature-Based Deduction:
Deduction: The sentence must contain at least one word.
Logical Basis: A sentence, by definition, must contain at least one word.
Based on: "Write a sentence"

f. **Problem Decomposition**
Component: Sentence Creation
Supporting Quote: "Write a sentence"
Component: Word Selection
Supporting Quote: "every word starts with the letter A"
Deduction: Each word chosen must begin with 'A'
Based on: "every word starts with the letter A"

g. **Graph of Thought** (describe connections & insights)
Node: Sentence
Edge: Composed of -> Words
Node: Words
Edge: Constrained by -> 'A' requirement
Node: 'A' requirement
Edge: Applies to -> Every word

This graph structure highlights the core components of the problem and their relationships. The sentence is composed of words, and every word is constrained by the 'A' requirement.

h. **Spatial and Object Analysis**
This problem does not involve explicit spatial relationships or physical objects, so a spatial analysis is not applicable.

i. **Bayesian Updates** 
There is no new information provided beyond the initial problem statement, so no Bayesian updates are necessary.

j. **Ethical Check and Decision-Making Under Uncertainty**
This problem does not involve any ethical considerations or significant risks. The task is purely linguistic and creative in nature.

k. **Multiple Solutions**
1. "Angry ants attack another ant."
2. "Amazing artists always attempt ambitious artwork."
3. "Awkward antelopes amble around arid areas."

Each solution satisfies the requirement of using only words that start with the letter 'A' while forming a grammatically correct sentence.

l. **Quickest and Easiest Solution**
"Angry ants attack another ant."

This solution is quick and easy to generate while meeting all the stated requirements. It uses simple, common words that all start with 'A' and forms a grammatically correct sentence.

m. **Reflection**
The analysis adheres strictly to the given information without making assumptions beyond the stated requirements. The solutions provided directly answer the question by creating sentences where every word starts with 'A'. No contradictions or logical inconsistencies were found in the analysis.

**Recommendation:**
Based on the given information and requirements, I recommend using the sentence: "Angry ants attack another ant." This sentence fulfills all the stated criteria:
1. It is a complete, grammatically correct sentence.
2. Every word in the sentence starts with the letter 'A'.
3. It is simple and easy to understand.

This solution directly addresses the task of writing a sentence where every word starts with the letter A, without making any assumptions beyond the explicitly stated requirements.
\end{lstlisting}
\end{quote}
\noindent\hrulefill

\label{subsec:question21}
\subsection{\textbf{\large Question 21: Sally (a girl) has 3 brothers. Each brother has 2 sisters. How many sisters does Sally have?}}
\label{sec:question21}

\noindent\hrulefill 

\noindent\textbf{Human Response - Score 100\%:}
\begin{quote}
One
\end{quote}
\noindent\hrulefill 

\noindent\textbf{D.21 Zero-Shot OpenAI o1-preview Response - Score: 100\%}
\begin{quote}
\begin{lstlisting}[basicstyle=\ttfamily\small, breaklines=true]
To determine the number of sisters Sally has, let's analyze the family structure based on the information provided.

**Information Given:**
1. Sally has **3 brothers**.
2. **Each brother has 2 sisters**.

**Understanding the Family Structure:**

- **Sally's Brothers:** Since Sally has 3 brothers, there are 3 boys in the family.
  
- **Sisters of the Brothers:** Each brother has 2 sisters. Since all the siblings are from the same family, the sisters of the brothers are the girls in the family.

- **Calculating the Number of Sisters:**
  - The brothers, being in the same family, share the same sisters.
  - Therefore, the 2 sisters that each brother has are the same people.
  - Since there are 2 sisters for each brother, there are exactly 2 girls in the family.

- **Identifying Sally's Sisters:**
  - One of the sisters is Sally herself.
  - Therefore, the other sister must be Sally's sister.

**Conclusion:**

- **Total Sisters in the Family:** 2 (Sally and her sister).
- **Number of Sisters Sally Has:** 1 sister (excluding herself).

**Answer:**

1
\end{lstlisting}
\end{quote}
\noindent\hrulefill 

\noindent\textbf{D.21 REAP OpenAI o1-preview Response - Score: 100\%}
\begin{quote}
\begin{lstlisting}[basicstyle=\ttfamily\small, breaklines=true]
# REAP Results:

**a. Comprehensive Feature List**

- **"Sally (a girl) has 3 brothers."**
  - Sally is a girl.
  - Sally has 3 brothers.
- **"Each brother has 2 sisters."**
  - Each brother has 2 sisters.
- **"How many sisters does Sally have?"**
  - The question asks for the number of sisters Sally has.

**b. Sequential and Mechanical Process Check**

- **Sequence of relationships:**
  - Sally has 3 brothers.
  - Each brother has 2 sisters.
- **Analysis:**
  - The brothers' sisters include Sally and possibly others.
  - Determining if there are additional sisters besides Sally.

**c. Key Insight Check Results**

- **Key Insight:**
  - Since each brother has 2 sisters and Sally is one sister, there must be another sister.
  - This suggests Sally has 1 sister.

**d. Rephrased Question**

- "What is the number of sisters that Sally has?"

**e. Known and Deduced Information**

- **Explicit Facts:**
  - **"Sally (a girl) has 3 brothers."**
    - Sally has 3 brothers.
  - **"Each brother has 2 sisters."**
    - Each brother has 2 sisters.
- **Deductions:**
  - **Deduction:** Sally is one of the sisters to her brothers.
    - **Logical Basis:** As Sally is a girl and a sibling to her brothers.
    - **Based on:** **"Sally (a girl) has 3 brothers."**
  - **Deduction:** There must be another sister besides Sally.
    - **Logical Basis:** Since each brother has 2 sisters and Sally is one, there must be one more sister.
    - **Based on:** **"Each brother has 2 sisters."**, **"Sally (a girl) has 3 brothers."**

**f. Problem Decomposition**

- **Component:** Sally has 3 brothers.
  - **Supporting Quote:** **"Sally (a girl) has 3 brothers."**
- **Component:** Each brother has 2 sisters.
  - **Supporting Quote:** **"Each brother has 2 sisters."**
- **Interaction:**
  - The brothers share the same sisters: Sally and one additional sister.
  - **Deduction:** Sally has 1 sister.
    - **Based on:** The brothers' 2 sisters are Sally and the other sister.

**g. Graph of Thought**

- **Nodes:**
  - Sally
  - Sister (unknown name)
  - Brother 1
  - Brother 2
  - Brother 3
- **Edges:**
  - Sally is sister to Brothers 1, 2, and 3.
  - Sister is sister to Brothers 1, 2, and 3.
  - Sally and Sister are sisters to each other.
- **Insights:**
  - All siblings are related.
  - The brothers' two sisters are Sally and the other sister.

**h. Spatial and Object Analysis**

- **Objects:**
  - Sally
  - Three brothers
  - One additional sister
- **Spatial Relationships:**
  - Family members related by sibling relationships.
- **Visualization:**
  - A family with five children: 3 brothers and 2 sisters.

**i. Bayesian Updates**

- **Updated Belief:** Based on the explicit information, it is certain that Sally has 1 sister.

**j. Ethical Check and Decision-Making Under Uncertainty**

- **Assessment:** There are no ethical concerns or severe risks involved in this problem.

**k. Multiple Solutions**

- **Solution 1:** Sally has 1 sister.
  - **Supported by:** Deductions from the explicit information.
- **Solution 2:** Sally has no sisters.
  - **Rejection Basis:** Contradicts **"Each brother has 2 sisters."**
- **Solution 3:** Sally has more than 1 sister.
  - **Rejection Basis:** No explicit information supports additional sisters beyond one.

**l. Quickest and Easiest Solution**

- **Answer:** Sally has 1 sister.
  - **Reasoning:** Since each brother has 2 sisters and one of them is Sally, there must be one more sister.

**m. Reflection**

- **Consistency Check:** All deductions are directly based on explicit statements.
- **No Contradictions Found:** The conclusion aligns with the given information.
- **Assumptions:** None made beyond the explicit statements.
- **Answer Validity:** The answer directly addresses the question using the provided information.

---

**Recommendation:**

Based on the explicit information provided, **Sally has 1 sister**. This conclusion is reached by noting that:

- Sally has 3 brothers.
- Each brother has 2 sisters.
- Since Sally is one sister, there must be one additional sister to fulfill the condition that each brother has 2 sisters.

Therefore, the number of sisters Sally has is **1**.
\end{lstlisting}
\end{quote}

\noindent\hrulefill 

\noindent\textbf{D.21 Zero-Shot OpenAI o1-mini Response - Score: 100\%}
\begin{quote}
\begin{lstlisting}[basicstyle=\ttfamily\small, breaklines=true]
To solve the problem, let's break it down step by step:

1. **Siblings Overview:**
   - **Brothers:** Sally has **3 brothers**.
   - **Sisters:** Each of these 3 brothers has **2 sisters**.

2. **Understanding the Sisters:**
   - Since all 3 brothers share the same sisters, the total number of sisters is the same for each brother.
   - This means there are **2 sisters** in total.

3. **Identifying Sally's Position:**
   - Sally is one of the sisters.
   - Therefore, the other sister is Sally's **only sister**.

**Conclusion:** Sally has **1 sister**.
\end{lstlisting}
\end{quote}

\noindent\hrulefill 

\noindent\textbf{D.21 Zero-Shot OpenAI GPT-4o Response - Score: 100\%}
\begin{quote}
\begin{lstlisting}[basicstyle=\ttfamily\small, breaklines=true]
# REAP Results:

a. **Comprehensive Feature List** (with explicit quotes)

- **Objects Mentioned:**
  - "Sally" - a girl.
  - "3 brothers" - referring to Sally's brothers.

- **Actors or Entities Involved:**
  - "Sally" - a girl.
  - "brothers" - Sally's brothers.

- **Actions Described:**
  - "has" - possession.

- **Directional, Spatial, Relational, and Sequential Phrases:**
  - "Each brother has 2 sisters."

- **Stated Conditions or Constraints:**
  - "Sally (a girl) has 3 brothers."
  - "Each brother has 2 sisters."

- **Numerical Values or Quantities Mentioned:**
  - "3 brothers"
  - "2 sisters" per brother.

- **Temporal Information:**
  - None provided.

- **Spatial Information:**
  - None provided.

- **Causal Relationships:**
  - None provided.

- **Stated Goals or Objectives:**
  - "How many sisters does Sally have?"

- **Potential Implications:**
  - The information may imply that Sally is one of the sisters each brother has.

b. **Sequential and Mechanical Process Check**

- **Sequence of Events or Actions:**
  1. Sally has 3 brothers.
  2. Each brother has 2 sisters.

- **Influence on Outcome:**
  - Determining the number of sisters Sally has based on the number of brothers and the number of sisters per brother.

- **Key Questions:**
  - Does the sequence lock in a particular outcome?
    - Yes, it allows for the calculation of the total number of sisters.
  
- **Mechanical Process Influence:**
  - Each statement directly influences the count of sisters by establishing the relationships between siblings.

c. **Key Insight Check Results**

- **Potential Key Insight:**
  - Recognizing that Sally herself is one of the two sisters each brother has allows for determining the total number of sisters.

d. **Rephrased Question:**

- "Given that Sally has 3 brothers and each brother has 2 sisters, how many sisters does Sally have?"

e. **Known and Deduced Information**

**Explicit Facts:**
- "Sally (a girl) has 3 brothers."
- "Each brother has 2 sisters."
- "How many sisters does Sally have?"

**Feature-Based Deduction:**
- **Deduction:** Sally is one of the sisters that each brother has.
  - **Logical Basis:** Since "each brother has 2 sisters," and Sally is a sister to her brothers, she occupies one of those two sister roles.
  - **Based on:** "Each brother has 2 sisters."

- **Deduction:** There is one additional sister besides Sally.
  - **Logical Basis:** If each brother has 2 sisters and Sally is one, it implies the existence of one more sister.
  - **Based on:** "Each brother has 2 sisters."

f. **Problem Decomposition**

- **Component/Subcomponent: Number of Brothers**
  - **Description:** Sally has 3 brothers.
  - **Supporting Quote(s):** "Sally (a girl) has 3 brothers."
  
- **Component/Subcomponent: Number of Sisters per Brother**
  - **Description:** Each brother has 2 sisters.
  - **Supporting Quote(s):** "Each brother has 2 sisters."

- **Component/Subcomponent: Identifying Sally as One Sister**
  - **Description:** Sally is one of the two sisters each brother has.
  - **Supporting Quote(s):** "Sally has 3 brothers." and "Each brother has 2 sisters."

g. **Graph of Thought**

- **Nodes:**
  - Sally
  - 3 brothers
  - 2 sisters per brother

- **Relationships:**
  - Sally has 3 brothers.
  - Each brother has 2 sisters, which includes Sally and one other sister.

- **Pathways:**
  - From brothers to sisters: Each brother's 2 sisters include Sally, indicating a shared sister.

h. **Spatial and Object Analysis**

- **Objects:**
  - Sally
  - 3 brothers

- **Spatial Relationships:**
  - Sibling relationships are established; no spatial data provided.

- **Movement and Dynamics:**
  - None.

- **Constraints:**
  - Each brother has exactly 2 sisters.

- **Object Properties:**
  - Sally is identified as one sister.
  - Each brother has 2 sisters, implying Sally plus one more.

i. **Bayesian Thinking**

- **Updates Based on Explicit Information:**
  - Initially, each of the 3 brothers has 2 sisters.
  - Knowing Sally is one sister confirms the existence of one additional sister.

j. **Ethical Check and Decision-Making Under Uncertainty**

- **Assessment:**
  - The problem is purely mathematical with no ethical implications.

k. **Multiple Solutions**

- **Solution 1:**
  - Sally is one of the 2 sisters for each brother.
  - **Supporting Information:** "Each brother has 2 sisters" and Sally being one of them.
  - **Conclusion:** Total sisters = 2 (Sally + one more sister).

- **Solution 2:**
  - Misinterpretation where each brother's sisters are different, leading to a higher count.
  - **Note:** This contradicts the deduction that sisters are shared among brothers.

- **Solution 3:**
  - Recognizing that all brothers share the same sisters, resulting in a total of 2 sisters.
  - **Conclusion:** Consistent with Solution 1.

- **Valid Solution:**
  - Only one valid solution aligns with explicit information: Sally has one sister.

l. **Quickest and Easiest Solution**

- **Based on explicit information:**
  - Each brother has 2 sisters, with Sally being one of them.
  - **Conclusion:** Sally has one sister.

m. **Reflection**

- All deductions are strictly based on the explicit information provided.
- No assumptions beyond the stated facts were made.
- The analysis directly addresses the question with logical certainty.
- The simplicity of the problem restricts interpretations to a single valid solution, ensuring clarity and correctness.

**Recommendation:**

Based on the explicit information provided, Sally has **one sister**. This conclusion is drawn from the fact that each of Sally's 3 brothers has 2 sisters, one of whom is Sally herself, indicating the existence of one additional sister.
\end{lstlisting}
\end{quote}

\noindent\hrulefill 

\noindent\textbf{D.21 Zero-Shot OpenAI GPT-4o Response - Score: 100\%}
\begin{quote}
\begin{lstlisting}[basicstyle=\ttfamily\small, breaklines=true]
The problem states that Sally has 3 brothers, and each brother has 2 sisters.

Since all of Sally's brothers share the same sisters, and Sally is one of them, the number of sisters Sally has is the same as the number of sisters each brother has. 

Each brother has 2 sisters, and one of them is Sally herself. Therefore, Sally has 1 sister.

So, Sally has **1 sister**.
\end{lstlisting}
\end{quote}
\noindent\hrulefill

\noindent\textbf{D.21 REAP OpenAI GPT-4o Response - Score: 100\%}
\begin{quote}
\begin{lstlisting}[basicstyle=\ttfamily\small, breaklines=true]
# REAP Results:

### a. **Comprehensive Feature List** (with explicit quotes)

- "Sally (a girl) has 3 brothers." 
- "Each brother has 2 sisters."

### b. **Sequential and Mechanical Process Check**

- **Sequence of relationships:** Sally is a girl who has three brothers. Each of these brothers has two sisters.

### c. **Key Insight Check Results**

- The key insight is that the problem explicitly states that each of Sally's brothers has two sisters. Since Sally is one of the sisters, the number of sisters must be analyzed in relation to this information.

### d. **Rephrased Question:**  
How many sisters does Sally have, based on the relationships described in the problem?

### e. **Known and Deduced Information**

- "[Sally (a girl) has 3 brothers.]" - Sally is a girl with three brothers.
- "[Each brother has 2 sisters.]" - Each brother has two sisters. Since all the brothers share the same family, the sisters they have must be the same.

### f. **Problem Decomposition**

- **Component:** Sally - A girl with three brothers.
  - **Supporting Quote:** "[Sally (a girl) has 3 brothers.]"
- **Component:** Brothers - Three in number, each with two sisters.
  - **Supporting Quote:** "[Each brother has 2 sisters.]"
  - **Deduction:** Since there are three brothers and they all have the same sisters, the total number of sisters in the family is the same for each brother. Sally, being one of them, is included in this count.

### g. **Graph of Thought**

- **Nodes:** 
  - Sally
  - Three Brothers
  - Two Sisters
- **Connections:**
  - Each brother shares the same two sisters.
  - Sally is one of those two sisters.

### h. **Spatial and Object Analysis**

- **Objects:** Sally, three brothers, two sisters.
- **Spatial Relationships:** The problem involves family relationships, not spatial ones. Each brother has two sisters. Sally must be one of these two sisters.

### i. **Bayesian Updates**

- No new information to update. The problem statement provides all needed information.

### j. **Ethical Check and Decision-Making Under Uncertainty**

- No ethical concerns or significant risks in this problem.

### k. **Multiple Solutions**

1. **Solution 1:** 
   - **Supporting Quote:** "[Each brother has 2 sisters.]"
   - **Conclusion:** Since each brother has two sisters and they all share the same sisters, Sally is one of them. Therefore, the total number of sisters is two, and Sally has one sister besides herself.

2. **Solution 2:** 
   - **Alternative Interpretation:** If one were to mistakenly assume that each brother has a different set of two sisters (which isn't stated explicitly), one might incorrectly conclude there are more sisters. However, this contradicts the literal interpretation rule.

### l. **Quickest and Easiest Solution**

- The simplest and most accurate solution is that Sally has **one sister** besides herself. This is based on the explicit information that each brother has two sisters, and all three brothers must share the same two sisters.

### m. **Reflection**

- The analysis is consistent with the problem's wording. No assumptions were made beyond the explicit statements. The interpretation that each brother has the same set of two sisters, one of whom is Sally, leads to the conclusion that Sally has one sister.

**Recommendation:**

- **Final Answer:** Sally has **one sister**.
  
This conclusion is directly based on the explicit information provided in the problem statement. No further assumptions are necessary.
\end{lstlisting}
\end{quote}
\noindent\hrulefill

\noindent\textbf{D.21 Zero-Shot OpenAI GPT-4o-mini Response - Score: 100\%}
\begin{quote}
\begin{lstlisting}[basicstyle=\ttfamily\small, breaklines=true]
To solve the problem, let's analyze the given information step by step.

1. **Sally has 3 brothers.** 

2. **Each of her brothers has 2 sisters.** 

Since each brother has 2 sisters and Sally is one of them, the number of sisters each brother has includes Sally and one other sister. This means that there must be exactly one other sister in addition to Sally.

Thus, the total number of sisters Sally has is:

\[ \text{Sally herself} + \text{1 other sister} = 2 \text{ sisters} \]

Therefore, Sally has **1 sister**.
\end{lstlisting}
\end{quote}
\noindent\hrulefill

\noindent\textbf{D.21 REAP OpenAI GPT-4o-mini Response - Score: 100\%}
\begin{quote}
\begin{lstlisting}[basicstyle=\ttfamily\small, breaklines=true]
# REAP Results:

### a. Comprehensive Feature List:
1. **Objects/Entities:**
   - "Sally (a girl)"
   - "3 brothers"
   - "Each brother"
   - "2 sisters"

2. **Actions/Conditions:**
   - Sally has 3 brothers.
   - Each brother has 2 sisters.

3. **Directional/Spatial/Relational/Sequential:**
   - "Each brother has 2 sisters" - This indicates that the number of sisters is the same for each brother.

4. **Numerical Values:**
   - Sally has 3 brothers.
   - Each brother has 2 sisters.

5. **Temporal Information:**
   - No temporal information provided.

6. **Spatial Information:**
   - No spatial information provided.

7. **Causal Relationships:**
   - No explicit causal relationships provided.

8. **Goals/Objectives:**
   - To determine how many sisters Sally has.

### b. Sequential and Mechanical Process Check:
- The problem involves counting siblings with given constraints. The sequence is:
  1. Identify the number of brothers Sally has.
  2. Determine how many sisters each brother has.
  3. Apply this information to find out how many sisters Sally has.

### c. Key Insight Check Results:
- Key Insight: Each brother has 2 sisters, and Sally is a girl, so one of the 2 sisters must be Sally herself.
- This means the total number of sisters each brother has (2) includes Sally. Therefore, Sally has only 1 additional sister.

### d. Rephrased Question:
- How many sisters does Sally have?

### e. Known and Deduced Information:
1. **Explicit Quotes:**
   - "Sally (a girl) has 3 brothers."
   - "Each brother has 2 sisters."

2. **Feature-Based Deduction:**
   - Deduction: Sally has 1 sister.
   - Logical Basis: Since each brother has 2 sisters, and one of these sisters is Sally, there must be one additional sister.

### f. Problem Decomposition:
1. **Component:** Sally's siblings
   - Supporting Quote: "Sally (a girl) has 3 brothers."
   - Deduction: Sally is one of the sisters counted by each brother.

2. **Component:** Sisters of each brother
   - Supporting Quote: "Each brother has 2 sisters."
   - Deduction: Since Sally is already one of these sisters, there must be exactly one other sister.

### g. Graph of Thought:
- **Nodes:**
  1. Sally
  2. Sally's 3 brothers
  3. Each brother's 2 sisters

- **Edges/Connections:**
  1. Sally -> Sally's 3 brothers
  2. Each brother -> 2 sisters
  3. Sally is one of the 2 sisters for each brother.

- **Pathways:**
  1. Sally -> Each brother's 2 sisters: One of these sisters is Sally. The other sister is not explicitly stated but must be 1 additional sister.

### h. Spatial and Object Analysis:
- No spatial analysis is needed as the problem is based on relationships rather than physical arrangement.

### i. Bayesian Updates:
- No Bayesian updates are necessary as the problem is straightforward and contains no probabilities or uncertainties.

### j. Ethical Check and Decision-Making Under Uncertainty:
- No severe risks or ethical issues are involved.

### k. Multiple Solutions:
- Based on the explicit information, there is only one solution:
  - Sally has 1 sister.

### l. Quickest and Easiest Solution:
- **Answer:** Sally has 1 sister.
- **Reasoning:** Since each brother has 2 sisters, and one of these is Sally, the only other sister must be Sally's additional sister.

### m. Reflection:
- The solution directly follows from the given information without assumptions. The interpretation of each brother having 2 sisters, including Sally, logically leads to Sally having exactly 1 additional sister. No contradictions are present.

**Recommendation:**

Sally has **1 sister** based on the provided information and the reasoning applied.
\end{lstlisting}
\end{quote}
\noindent\hrulefill

\noindent\textbf{D.21 Zero-Shot Google Gemini 1.5 Pro Response - Score: 100\%}
\begin{quote}
\begin{lstlisting}[basicstyle=\ttfamily\small, breaklines=true]
Sally has **1** sister. 

This is a classic word puzzle that plays on how we interpret relationships. All three brothers share the same two sisters: Sally and one other girl.
\end{lstlisting}
\end{quote}
\noindent\hrulefill

\noindent\textbf{D.21 REAP Google Gemini 1.5 Pro Response - Score: 100\%}
\begin{quote}
\begin{lstlisting}[basicstyle=\ttfamily\small, breaklines=true]
# REAP Results:

### a. Comprehensive Feature List

* "Sally (a girl)" - Sally is explicitly identified as a girl.
* "has 3 brothers" - Sally has three brothers.
* "Each brother has 2 sisters" - Every one of Sally's brothers has two sisters.
* "How many sisters does Sally have?" - The question asks specifically about the number of Sally's sisters.

### b. Sequential and Mechanical Process Check

There is no sequential or mechanical process described in this problem.

### c. Key Insight Check Results

* The key insight is that Sally is included in the count of sisters that each brother has. 
* This leads to the immediate answer that Sally has 1 sister.

### d. Rephrased Question

How many sisters does Sally have, considering she is one of her brothers' sisters?

### e. Known and Deduced Information

* Explicit Facts:
    * "Sally (a girl)" - Sally is female.
    * "has 3 brothers" 
    * "Each brother has 2 sisters" 

* Feature-Based Deduction:
    * Deduction: Sally is one of the two sisters that each of her brothers has.
    * Logical Basis: The problem states each brother has two sisters, and Sally is a girl and thus a sister.
    * Based on: "Sally (a girl)", "has 3 brothers", "Each brother has 2 sisters"

### f. Problem Decomposition

* Component: Sally's Siblings
    * Supporting Quote(s): "Sally (a girl) has 3 brothers" 
* Component: Each Brother's Sisters
    * Supporting Quote(s): "Each brother has 2 sisters"
* Component: The Question about Sally's Sisters
    * Supporting Quote(s): "How many sisters does Sally have?"

### g. Graph of Thought

* Key Concepts: Sally, Brothers, Sisters
* Relationships:
    * Sally is a sister.
    * Sally has brothers.
    * Each brother has two sisters (including Sally).
* Insights: The connection between "Each brother has two sisters" and Sally being a girl leads to the solution.

### h. Spatial and Object Analysis

Not applicable to this problem.

### i. Bayesian Updates

Not applicable to this problem

### j. Ethical Check and Decision-Making Under Uncertainty

Not applicable to this problem

### k. Multiple Solutions

* There is only one solution supported by the explicit information in the problem statement

### l. Quickest and Easiest Solution

Sally has 1 sister.

### m. Reflection

The analysis directly answers the question asked in the problem. No assumptions or inferences were made beyond the explicit wording of the problem. The conclusion is logically sound and supported by the evidence from the problem statement.

**Recommendation:**

The answer is 1. 
\end{lstlisting}
\end{quote}
\noindent\hrulefill

\noindent\textbf{D.21 Zero-Shot Claude 3.5 Sonnet response - Score: 100\%}
\begin{quote}
\begin{lstlisting}[basicstyle=\ttfamily\small, breaklines=true]
Let's think through this step-by-step:

1. We know Sally has 3 brothers.

2. We're told that each brother has 2 sisters.

3. We know Sally is one of these sisters, as she's a sister to all 3 brothers.

4. If Sally is one of the two sisters that each brother has, there must be one other sister besides Sally.

5. This other sister would be a sister to all 3 brothers as well, just like Sally.

Therefore, Sally has 1 sister.

To double-check:
- Sally + her 1 sister = 2 sisters total
- This matches the statement that each brother has 2 sisters

So, the answer is: Sally has 1 sister.
\end{lstlisting}
\end{quote}
\noindent\hrulefill

\noindent\textbf{D.21 REAP Claude 3.5 Sonnet Response - Score: 100\%}
\begin{quote}
\begin{lstlisting}[basicstyle=\ttfamily\small, breaklines=true]
# REAP Results:

a. **Comprehensive Feature List** (with explicit quotes):
- "Sally (a girl)"
- "Sally has 3 brothers"
- "Each brother has 2 sisters"
- The question asked: "How many sisters does Sally have?"

b. **Sequential and Mechanical Process Check**:
There is no explicit sequential or mechanical process described in this problem.

c. **Key Insight Check Results**:
The key insight here is that the problem provides information about Sally's brothers and their sisters, which directly relates to Sally's sisters. The statement "Each brother has 2 sisters" is crucial for determining Sally's sisters.

d. **Rephrased Question:** How many sisters does Sally have?

e. **Known and Deduced Information**:
Explicit Facts:
- "Sally (a girl)" - This establishes Sally as one of the sisters in the family.
- "Sally has 3 brothers" - This gives us the number of brothers in the family.
- "Each brother has 2 sisters" - This tells us how many sisters each brother has.

Feature-Based Deduction:
Deduction: Sally has 1 sister (besides herself).
Logical Basis: Since each brother has 2 sisters, and Sally is one of those sisters, there must be one other sister to make the total of 2.
Based on: "Sally (a girl)", "Sally has 3 brothers", "Each brother has 2 sisters"

f. **Problem Decomposition**:
Component 1: Sally's family structure
Supporting Quote: "Sally (a girl) has 3 brothers"
Component 2: Number of sisters for each brother
Supporting Quote: "Each brother has 2 sisters"

g. **Graph of Thought**:
Nodes:
- Sally (girl)
- 3 brothers
- 2 sisters per brother

Connections:
- Sally is connected to 3 brothers
- Each brother is connected to 2 sisters (including Sally)

h. **Spatial and Object Analysis**:
Not applicable for this problem as no spatial relationships are described.

i. **Bayesian Updates**:
No new information is provided that would require updating our initial understanding.

j. **Ethical Check and Decision-Making Under Uncertainty**:
This problem does not involve any ethical considerations or severe risks.

k. **Multiple Solutions**:
Only one solution is possible based on the given information.

l. **Quickest and Easiest Solution**:
Sally has 1 sister (besides herself).
Reasoning: The problem states that "Each brother has 2 sisters". Since Sally is one of these sisters, and she has 3 brothers, there must be one other sister to make the total of 2 sisters for each brother.

m. **Reflection**:
The analysis directly addresses the question asked using only the information provided in the problem statement. No assumptions beyond the given information were made. The conclusion is logically sound and supported by the explicit statements in the problem.

**Recommendation**:
Based on the given information, the recommendation is that Sally has 1 sister (besides herself). This conclusion is drawn directly from the facts stated in the problem: Sally is a girl, she has 3 brothers, and each brother has 2 sisters. Since Sally is one of these sisters, there must be one other sister to make the total of 2 sisters for each brother. This answer directly addresses the question asked while using only the explicitly stated information in the problem.
\end{lstlisting}
\end{quote}
\noindent\hrulefill

\label{subsec:question22}
\subsection{\textbf{\large Question 22: I'm in London and facing west, is Edinburgh to my left or my right?}}
\label{sec:question22}

\noindent\hrulefill 

\noindent\textbf{Human Response - Score 100\%:}
\begin{quote}
Right.
\end{quote}

\noindent\hrulefill 

\noindent\textbf{D.22 Zero-Shot OpenAI o1-preview Response - Score: 100\%}
\begin{quote}
\begin{lstlisting}[basicstyle=\ttfamily\small, breaklines=true]
When you're in London facing west, Edinburgh is to your **right**. Edinburgh is located north of London, so if you're facing west, north (and thus Edinburgh) would be to your right.
\end{lstlisting}
\end{quote}

\noindent\hrulefill 

\noindent\textbf{D.22 REAP OpenAI o1-preview Response - Score: 100\%}
\begin{quote}
\begin{lstlisting}[basicstyle=\ttfamily\small, breaklines=true]
# REAP Results:

*a. Comprehensive Feature List* (with explicit quotes)

- **Objects Mentioned:**
  - "London"
  - "Edinburgh"

- **Actions Described:**
  - "I'm in London and facing west"

- **Directional, Spatial, Relational, and Sequential Phrases:**
  - "in London"
  - "facing west"

- **Stated Goals or Objectives:**
  - "is Edinburgh to my left or my right?"

*b. Sequential and Mechanical Process Check*

- **No sequences or mechanical processes** are described in the problem statement.

*c. Key Insight Check Results*

- **Key Insight:** Edinburgh is north of London; when facing west, north is to the right.

*d. Rephrased Question:*

- When standing in London facing west, is Edinburgh to the left or right?

*e. Known and Deduced Information*

- **Explicit Facts:**
  - "I'm in London and facing west"
  - "Is Edinburgh to my left or my right?"

- **Deductions:**
  - **Deduction:** Edinburgh is north of London.
    - *Logical Basis:* Common geographical knowledge of the United Kingdom.
  - **Deduction:** When facing west, north is to your right and south is to your left.
    - *Logical Basis:* Understanding of cardinal directions.

*f. Problem Decomposition*

- **Component:** Person's Location
  - *Description:* The person is "in London."
  - *Supporting Quote:* "I'm in London"

- **Component:** Person's Orientation
  - *Description:* The person is "facing west."
  - *Supporting Quote:* "facing west"

- **Component:** Position of Edinburgh Relative to London
  - *Description:* "Edinburgh is north of London."
  - *Supporting Knowledge:* General knowledge of UK geography.

- **Interaction:**
  - When facing west in London, north (Edinburgh) is to the right, and south is to the left.

*g. Graph of Thought*

- **Nodes:**
  - *Node 1:* London (current position)
  - *Node 2:* Facing West (current orientation)
  - *Node 3:* Edinburgh is north of London
  - *Node 4:* Cardinal directions relative to facing west

- **Edges:**
  - Edge from Node 1 to Node 3: "Edinburgh is north of London."
  - Edge from Node 2 to Node 4: "When facing west, north is to the right."
  - Connection between Node 3 and Node 4 leads to the conclusion that "Edinburgh is to the right."

*h. Spatial and Object Analysis*

- **Objects:**
  - The person in London
  - Edinburgh

- **Spatial Relationships:**
  - Edinburgh is north of London.

- **Orientation and Directions:**
  - When facing west:
    - Right is north.
    - Left is south.

- **Conclusion:**
  - **Edinburgh is to the right** when facing west from London.

*i. Bayesian Updates*

- **No updates necessary;** all information is based on certain knowledge of geography and cardinal directions.

*j. Ethical Check and Decision-Making Under Uncertainty*

- **No ethical concerns** or significant risks are associated with answering this question.

*k. Multiple Solutions*

1. **Solution 1:** Edinburgh is to your right.
   - *Based on:* Known positions of London and Edinburgh and understanding of cardinal directions.

2. **Solution 2:** If unaware of geographical positions, it's unclear whether Edinburgh is to the left or right.
   - *Based on:* Lack of explicit information in the problem statement about the relative locations.

3. **Solution 3:** Without common geographical knowledge, cannot determine the direction.
   - *Based on:* Strict interpretation without external information.

*l. Quickest and Easiest Solution*

- **Answer:** Edinburgh is to your right.

*m. Reflection*

- The conclusion is logically sound and directly answers the question.
- All deductions are based on explicit statements and universally accepted geographical knowledge.
- There are no contradictions or overlooked information in the analysis.

---

**Recommendation:**

When you're in London facing west, **Edinburgh is to your right**.
\end{lstlisting}
\end{quote}

\noindent\hrulefill 

\noindent\textbf{D.22 Zero-Shot OpenAI o1-mini Response - Score: 0\%}
\begin{quote}
\begin{lstlisting}[basicstyle=\ttfamily\small, breaklines=true]
If you're in London and facing west, Edinburgh would be to your **left**. Here's why:

- **Geographical Position**: Edinburgh is located to the northwest of London.
- **Facing West**: When you face west, north is to your right and south is to your left. Since Edinburgh is northwest of London, it slightly tilts towards the north side of your left.

So, from your vantage point facing west in London, Edinburgh would appear to your left.
\end{lstlisting}
\end{quote}

\noindent\hrulefill 

\noindent\textbf{D.22 REAP OpenAI o1-mini Response - Score: 100\%}
\begin{quote}
\begin{lstlisting}[basicstyle=\ttfamily\small, breaklines=true]
# REAP Results: 

**a. Comprehensive Feature List**
- "I'm in London" 
  - Signifies the starting location: London.
- "facing west"
  - Indicates the direction the person is oriented towards.
- "is Edinburgh to my left or my right?"
  - The core question determining the relative position of Edinburgh from the person's current orientation.

**b. Sequential and Mechanical Process Check**
- Sequence of actions:
  1. The person is in London.
  2. The person faces west.
  3. The person seeks to determine if Edinburgh is to their left or right.
- Influence on outcome:
  - The orientation (facing west) directly affects the perception of left and right regarding Edinburgh's location.

**c. Key Insight Check Results**
- Key Insight:
  - Understanding the cardinal directions relative to the person's orientation is crucial.
  - London's geographical position relative to Edinburgh can determine the direction.
- Explanation:
  - London is south of Edinburgh. When facing west from London, north (where Edinburgh lies) would be to the person's right.

**d. Rephrased Question:**
- Is Edinburgh located on the left or right side when facing west from London?

**e. Known and Deduced Information**

*Explicit Facts:*
- "I'm in London"
  - Indicates the person's current location.
- "facing west"
  - Specifies the person's current orientation.
- "is Edinburgh to my left or my right?"
  - The question to be answered.

*Feature-Based Deduction:*
- Deduction: Edinburgh is to the person's right when facing west from London.
  - Logical Basis: London is south of Edinburgh; facing west, north (Edinburgh's direction) is to the right.
  - Based on: "I'm in London", "facing west", "is Edinburgh to my left or my right?"

**f. Problem Decomposition**

- **Component/Subcomponent:** Current Location
  - **Description:** The person is located in London.
  - **Supporting Quote(s):** "I'm in London"
  
- **Component/Subcomponent:** Current Orientation
  - **Description:** The person is facing west.
  - **Supporting Quote(s):** "facing west"
  
- **Component/Subcomponent:** Relative Direction Query
  - **Description:** Determining if Edinburgh is to the left or right of the person's current orientation.
  - **Supporting Quote(s):** "is Edinburgh to my left or my right?"

**g. Graph of Thought**
- **Nodes:**
  - London (current location)
  - Facing West (current orientation)
  - Edinburgh (target location)
  - Left Direction
  - Right Direction
- **Edges:**
  - London -> Facing West (current state)
  - Facing West -> Left Direction
  - Facing West -> Right Direction
  - London -> Edinburgh (geographical relationship)
- **Connections & Insights:**
  - Edinburgh is geographically north of London.
  - When facing west, north is to the right.

**h. Spatial and Object Analysis**
- **Identify Objects:** 
  - London
  - Edinburgh
  - Person (positioned in London, facing west)
- **Spatial Relationships:**
  - London is south of Edinburgh.
  - The person is facing west from London.
- **Movement and Dynamics:** 
  - Not applicable as there is no movement described.
- **Constraints:** 
  - The person's orientation is fixed to west.
- **Visualization:** 
  - Imagining London on a map, facing west would have north (Edinburgh's direction) to the person's right.
- **Object Properties:** 
  - London and Edinburgh are major cities in the United Kingdom.
- **Spatial Implications:** 
  - The direction faced affects the left and right orientation relative to Edinburgh's position.

**i. Bayesian Updates**
- Initial belief: Uncertain of Edinburgh's position relative to facing west in London.
- Updated belief: Based on geographical positions, Edinburgh is to the right when facing west from London.
- Note: No additional implied information is used; updates are based solely on explicit geographical facts.

**j. Ethical Check and Decision-Making Under Uncertainty**
- **Assess Known Information:** 
  - Geographical locations of London and Edinburgh.
  - Person's orientation towards west.
- **Identify Risks:** 
  - Minimal, as the question pertains to directional orientation.
- **Probability Consideration:** 
  - High confidence in geographical data.
- **Risk Aversion Principle:** 
  - N/A, as the decision involves directional information without severe consequences.
- **Ethical and Safety Considerations:** 
  - Ensuring accurate directional guidance.
- **Decision Recommendation:** 
  - Provide clear directional information based on known geography.

**k. Multiple Solutions**
1. **Solution 1:** 
   - Edinburgh is to the right when facing west from London.
   - **Support:** Based on London's southern position relative to Edinburgh and cardinal directions.
2. **Solution 2:**
   - If considering only pure westward orientation without geographical context, Edinburgh could be perceived as intermediate between left and right.
   - **Support:** Lack of explicit directional bearings beyond west-oriented posture.
3. **Solution 3:**
   - Insufficient information to determine if there are landmarks or other factors influencing perceived direction.
   - **Support:** The problem only mentions general locations and orientation.

**l. Quickest and Easiest Solution**
- **Answer:** Edinburgh is to your right when you are in London and facing west.
- **Reasoning:** Geographically, Edinburgh lies north of London. When facing west, north is to the right side.

**m. Reflection**
- The analysis remains consistent with known geographical facts without introducing assumptions.
- All conclusions are directly supported by explicit information regarding locations and cardinal directions.
- The final answer accurately addresses the core question based on the provided information.

**Recommendation:**
Edinburgh is to your right when you are in London and facing west. This conclusion is based on the geographical positions of London and Edinburgh, with Edinburgh located to the north of London. Facing west from London means that north (and thus Edinburgh) lies to your right.
\end{lstlisting}
\end{quote}

\noindent\hrulefill 

\noindent\textbf{D.22 Zero-Shot OpenAI GPT-4o Response - Score: 100\%}
\begin{quote}
\begin{lstlisting}[basicstyle=\ttfamily\small, breaklines=true]
If you're in London facing west, Edinburgh would be to your right. 

Edinburgh is located north of London, so when facing west (which means your left is south and your right is north), Edinburgh would be on your right-hand side.
\end{lstlisting}
\end{quote}
\noindent\hrulefill

\noindent\textbf{D.22 REAP OpenAI GPT-4o Response - Score: 100\%}
\begin{quote}
\begin{lstlisting}[basicstyle=\ttfamily\small, breaklines=true]
# REAP Results:

## a. **Comprehensive Feature List** (with explicit quotes)

1. **Location**: "I'm in London"
   - Significance: Establishes the current location of the subject.
   
2. **Orientation**: "facing west"
   - Significance: Establishes the direction the subject is currently facing.
   
3. **Other Location**: "Edinburgh"
   - Significance: Identifies the location being considered in relation to London.
   
4. **Question**: "is Edinburgh to my left or my right?"
   - Significance: The core inquiry asking for the relative position of Edinburgh based on the current orientation.

## b. **Sequential and Mechanical Process Check**

- Sequence: The person is facing west in London and is trying to determine whether Edinburgh is to their left or right. 
- The direction they are facing directly impacts the answer, as Edinburgh's location relative to London must be considered based on this orientation.

## c. **Key Insight Check Results**

- **Key Insight**: The relative position of Edinburgh from London depends entirely on the fact that the person is facing west. Edinburgh is north of London, which means when facing west, north would be to the right.

## d. **Rephrased Question:** 

- "Which direction is Edinburgh relative to London when I am facing west?"

## e. **Known and Deduced Information**

- "[I'm in London]" - The subject's location is established as London.
- "[facing west]" - The subject is oriented facing west.
- "[is Edinburgh to my left or my right?]" - The core question asking for the relative direction of Edinburgh when facing west.
  
**Deductions:**
- **Deduction:** Edinburgh is north of London.
  - **Logical Basis:** Edinburgh is geographically north of London.
  - **Based on:** "[I'm in London]", "[Edinburgh]"
- **Deduction:** When facing west, north is to the right.
  - **Logical Basis:** Cardinal directions dictate that when one faces west, north is to the right.
  - **Based on:** "[facing west]"

## f. **Problem Decomposition**

1. **Component**: Subject's location is London.
   - **Supporting Quote(s)**: "[I'm in London]"
   
2. **Component**: Subject is facing west.
   - **Supporting Quote(s)**: "[facing west]"
   
3. **Component**: Edinburgh's location is north of London.
   - **Supporting Quote(s)**: "[Edinburgh]"
   - **Deduction:** Based on geographical knowledge.

4. **Component**: Determining if Edinburgh is to the left or right.
   - **Supporting Quote(s)**: "[is Edinburgh to my left or my right?]"

## g. **Graph of Thought** (describe connections & insights)

- **Nodes**: 
  - London
  - West (facing direction)
  - Edinburgh (north of London)

- **Connections**: 
  - London -> Facing West -> North (right side) -> Edinburgh.

## h. **Spatial and Object Analysis**

- **Objects**: London, Edinburgh.
- **Spatial Relationship**: Edinburgh is geographically north of London.
- **Orientation Consideration**: When facing west, north would be to the right, placing Edinburgh to the right.

## i. **Bayesian Updates**

- No Bayesian updates are required as the problem is straightforward with known information.

## j. **Ethical Check and Decision-Making Under Uncertainty**

- No ethical concerns or severe risks are associated with this problem.

## k. **Multiple Solutions**

1. **Solution 1**: Edinburgh is to the right when facing west in London.
   - **Supporting Information**: Based on the known geographical position of Edinburgh relative to London and the orientation westward.
  
2. **Solution 2**: If the orientation were different (e.g., facing east), Edinburgh would be to the left.
   - **Hypothetical Supporting Information**: This would be true if the orientation changed, but it's not the case here.

3. **Solution 3**: If the question asks for a different location instead of Edinburgh, the answer would vary.
   - **Hypothetical Supporting Information**: Again, this is hypothetical, and based on the current setup, it doesn't apply.

## l. **Quickest and Easiest Solution**

- **Answer**: Edinburgh is to the right when facing west in London.
  - **Reasoning**: Given the subject's position in London, facing west means that north (and therefore Edinburgh) is to the right.

## m. **Reflection**

- The conclusion is logically sound and directly answers the question posed. No assumptions beyond the explicit information given were made. The direction north being to the right when facing west is a simple application of geographical knowledge.

# **Recommendation:**

Based on the explicit information and logical deductions, the correct and simplest answer is: **Edinburgh is to your right when you are facing west in London.**
\end{lstlisting}
\end{quote}
\noindent\hrulefill

\noindent\textbf{D.22 Zero-Shot OpenAI GPT-4o-mini Response - Score: 0\%}
\begin{quote}
\begin{lstlisting}[basicstyle=\ttfamily\small, breaklines=true]
If you're in London facing west, Edinburgh would be to your left. Edinburgh is located to the north of London, so facing west means you'd have to turn left to face north.
\end{lstlisting}
\end{quote}
\noindent\hrulefill

\noindent\textbf{D.22 REAP OpenAI GPT-4o-mini Response - Score: 0\%}
\begin{quote}
\begin{lstlisting}[basicstyle=\ttfamily\small, breaklines=true]
# REAP Results

## a. Comprehensive Feature List
1. **Location**: "I'm in London" - Indicates the starting location.
2. **Facing Direction**: "facing west" - Specifies the direction the person is facing.
3. **Comparative Location**: "is Edinburgh to my left or my right?" - Question about the relative position of Edinburgh.

## b. Sequential and Mechanical Process Check
There is no sequence of events or mechanical process described. The problem involves understanding directional relationships based on a given facing direction and comparing it to another location.

## c. Key Insight Check Results
**Key Insight**: To determine if Edinburgh is to the left or right, we need to know the relative positions of London and Edinburgh and apply the direction facing.

## d. Rephrased Question
Is Edinburgh to the left or right when facing west from London?

## e. Known and Deduced Information
1. **Known Fact**: London is in the south of the United Kingdom.
2. **Known Fact**: Edinburgh is in the north of the United Kingdom.
3. **Facing Direction**: The person is facing west.

**Deduction**: 
- Since London is to the south of Edinburgh, if one is facing west from London, Edinburgh will be in the northern direction, which is to the left.

## f. Problem Decomposition
1. **Component**: Facing Direction
   - **Supporting Quote**: "facing west"
   - **Deduction**: When facing west, north is to the left.
2. **Component**: Relative Location of Edinburgh
   - **Supporting Quote**: Edinburgh is to the north of London.
   - **Deduction**: When facing west, north (where Edinburgh is) will be to the left.

## g. Graph of Thought
**Nodes**:
- London
- Edinburgh
- Direction facing (west)

**Edges**:
- London (south) -> Edinburgh (north)
- Facing west -> Left is north

**Pathways**:
- Facing west from London means that north (where Edinburgh is) will be to the left.

## h. Spatial and Object Analysis
**Objects**:
- London
- Edinburgh

**Spatial Relationships**:
- Edinburgh is to the north of London.

**Movement and Dynamics**:
- Facing west, north is directly to the left.

**Constraints**:
- The problem does not provide constraints beyond directional and spatial relationships.

## i. Bayesian Updates
- No new information was provided that alters the understanding of the relative positions of London and Edinburgh.

## j. Ethical Check and Decision-Making Under Uncertainty
- No severe risks or ethical concerns are associated with the problem.

## k. Multiple Solutions
1. **Solution**: Edinburgh is to the left when facing west from London.
   - **Supporting Information**: Edinburgh is to the north of London, and facing west means north is to the left.
   
2. **Alternative Solution**: There are no other solutions based on the given information.

## l. Quickest and Easiest Solution
**Solution**: Edinburgh is to the left when facing west from London.

**Reasoning**: Since Edinburgh is located to the north of London, and facing west means that north is to the left.

## m. Reflection
The analysis correctly adheres to the explicit information provided in the problem. No additional assumptions were made, and all conclusions were drawn based on the stated directional relationships.

**Recommendation**:
Edinburgh is to the left when facing west from London. This conclusion is based solely on the explicit geographical information and the direction specified.
\end{lstlisting}
\end{quote}
\noindent\hrulefill

\noindent\textbf{D.22 Zero-Shot Google Gemini 1.5 Pro Response - Score: 0\%}
\begin{quote}
\begin{lstlisting}[basicstyle=\ttfamily\small, breaklines=true]
Edinburgh would be to your **left**. 

Here's why:

* **London's Location:** London is in the southeastern part of the island of Great Britain.
* **Edinburgh's Location:** Edinburgh is in the southeastern part of Scotland, which is in the northern part of the island of Great Britain.
* **Facing West:** If you're in London and facing west, you're looking towards the central part of the island.  Since Edinburgh is further north, it would be on your left side.
\end{lstlisting}
\end{quote}
\noindent\hrulefill

\noindent\textbf{D.22 REAP Google Gemini 1.5 Pro Response - Score: 0\%}
\begin{quote}
\begin{lstlisting}[basicstyle=\ttfamily\small, breaklines=true]
# REAP Results:

a. **Comprehensive Feature List** (with explicit quotes)

- "I'm in London" 
- "facing west" 
- "is Edinburgh to my left or my right?"

b. **Sequential and Mechanical Process Check**

- No explicit sequential or mechanical processes are described.

c. **Key Insight Check Results**

- The key insight is the spatial relationship between London, the direction the person is facing (west), and the location of Edinburgh.

d. **Rephrased Question:**

- Is Edinburgh to the left or right of someone facing west in London?

e. **Known and Deduced Information**

* Explicit Facts:
    - "I'm in London" - The person is located in London.
    - "facing west" - The person is oriented towards the west.
    - "is Edinburgh to my left or my right?" - The question is about the relative position of Edinburgh.

* Feature-Based Deduction:
    - Deduction: Edinburgh is located to the north of London - Logical Basis: Common geographical knowledge, Edinburgh is in Scotland, which is north of England where London is located. 

f. **Problem Decomposition**

* Component/Subcomponent: Person's location
    * Supporting Quote(s): "I'm in London"

* Component/Subcomponent: Person's orientation
    * Supporting Quote(s): "facing west"

* Component/Subcomponent: Relative position of Edinburgh
    * Supporting Quote(s): "is Edinburgh to my left or my right?"

* Component/Subcomponent: Geographical relationship between London and Edinburgh
    * Deduction: Edinburgh is located to the north of London
    * Based on: Common geographical knowledge

g. **Graph of Thought**

* Key Concepts:
    - Person
    - London
    - Edinburgh
    - West (facing direction)
    - Left/Right (relative position)

* Relationships:
    - Person is in London
    - Person is facing west
    - Edinburgh is north of London

* Textual Graph:
    - Person -- is in --> London
    - Person -- is facing --> West
    - Edinburgh -- is north of --> London

* Insights:
    - If a person faces west, north is to their left.
    - Since Edinburgh is north of London, it will be on the person's left.

h. **Spatial and Object Analysis**

* Objects:
    - Person
    - London
    - Edinburgh

* Spatial Relationships:
    - Person is in London
    - Edinburgh is north of London

* Visualization:
    - Imagine a map with London in the center.
    - The person is standing in London, looking towards the west.
    - Edinburgh is located somewhere above (north of) London on this map.

* Spatial Implications:
    - When facing west, anything located to the north will be on the left side.

i. **Bayesian Updates**

- No Bayesian updates are needed as the information is straightforward and based on common geographical knowledge.

j. **Ethical Check and Decision-Making Under Uncertainty**

- No ethical concerns or uncertainties are involved in this problem.

k. **Multiple Solutions**

- There is only one solution based on the given information and geographical facts.

l. **Quickest and Easiest Solution**

- Edinburgh is to my **left**.

* Reasoning:
    * I am in London facing west.
    * Edinburgh is located to the north of London.
    * When facing west, north is to the left.

m. **Reflection**

- The analysis is entirely based on the explicit information provided in the problem statement and common geographical knowledge.
- The conclusion directly answers the question asked.
- No assumptions or inferences beyond the given information were made.
\end{lstlisting}
\end{quote}
\noindent\hrulefill

\noindent\textbf{D.22 Zero-Shot Claude 3.5 Sonnet response - Score: 100\%}
\begin{quote}
\begin{lstlisting}[basicstyle=\ttfamily\small, breaklines=true]
Edinburgh would be to your right.

To understand this, let's consider the geography:

1. You're in London facing west.
2. Edinburgh is located in Scotland, which is north of England.
3. When facing west, north is to your right.

Therefore, Edinburgh, being north of London, would be to your right when you're facing west in London.

Would you like me to explain more about the relative positions of UK cities or cardinal directions?
\end{lstlisting}
\end{quote}
\noindent\hrulefill

\noindent\textbf{D.22 REAP Claude 3.5 Sonnet Response - Score: 0\%}
\begin{quote}
\begin{lstlisting}[basicstyle=\ttfamily\small, breaklines=true]
#"REAP Results:"

a. **Comprehensive Feature List** (with explicit quotes)

1. "I'm in London" - Location of the person
2. "facing west" - Direction the person is facing
3. "is Edinburgh to my left or my right?" - Question about the relative position of Edinburgh

b. **Sequential and Mechanical Process Check**
There is no explicit sequential or mechanical process described in this problem.

c. **Key Insight Check Results**
No immediate key insight that provides a direct answer based solely on the given information. The problem does not provide enough explicit information about the relative position of Edinburgh to London.

d. **Rephrased Question:** 
What is the relative position of Edinburgh to a person in London facing west?

e. **Known and Deduced Information**

Explicit Facts:
"I'm in London" - States the location of the person
"facing west" - Indicates the direction the person is facing

Feature-Based Deduction:
No certain deductions can be made about Edinburgh's position relative to London based solely on the given information.

f. **Problem Decomposition**

Component: Person's location
Supporting Quote: "I'm in London"

Component: Person's orientation
Supporting Quote: "facing west"

Component: Edinburgh's relative position
Supporting Quote: "is Edinburgh to my left or my right?"

g. **Graph of Thought**
Node 1: Person in London
Node 2: Facing West
Node 3: Edinburgh (unknown position)

Connection between Node 1 and Node 2: Person in London is facing West
No explicit connections can be made to Node 3 (Edinburgh) based on the given information.

h. **Spatial and Object Analysis**
Objects: Person, London, Edinburgh
Spatial Relationships: 
- The person is in London
- The person is facing west
- Edinburgh's position relative to the person is not explicitly stated

i. **Bayesian Updates**
No new information is provided to update our understanding beyond the initial problem statement.

j. **Ethical Check and Decision-Making Under Uncertainty**
This problem does not involve any apparent ethical concerns or severe risks. The uncertainty is about geographical positions and does not impact safety or ethical considerations.

k. **Multiple Solutions**
Based on the explicitly stated information, we cannot generate multiple solutions as the problem lacks crucial information about Edinburgh's position relative to London.

l. **Quickest and Easiest Solution**
Based ONLY on the explicit information given in the problem statement, the quickest and easiest solution is to state that we do not have enough information to determine whether Edinburgh is to the left or right of a person facing west in London. The problem does not provide any explicit information about Edinburgh's location relative to London.

m. **Reflection**
Upon review, our analysis adheres strictly to the given information without making assumptions. We have correctly identified that the problem lacks crucial information needed to determine Edinburgh's position relative to a person facing west in London. Our approach directly addresses the question asked while acknowledging the limitations of the provided information.

**Recommendation:**
Based on the explicitly stated information in the problem, I cannot determine whether Edinburgh is to your left or right when you're facing west in London. The problem does not provide sufficient information about Edinburgh's geographical position relative to London. To answer this question accurately, we would need additional information about the relative positions of London and Edinburgh.
\end{lstlisting}
\end{quote}
\noindent\hrulefill

\label{subsec:question23}
\subsection{\textbf{\large Question 23: Count the number of occurrences of the letter 'L' in the word 'LOLLAPALOOZA'.}}
\label{sec:question23}

\noindent\hrulefill 

\noindent\textbf{Human Response - Score 100\%:}
\begin{quote}
Four
\end{quote}

\noindent\hrulefill 

\noindent\textbf{D.23 Zero-Shot OpenAI o1-preview Response - Score: 100\%}
\begin{quote}
\begin{lstlisting}[basicstyle=\ttfamily\small, breaklines=true]
The word "LOLLAPALOOZA" consists of 12 letters. To find the number of times the letter 'L' occurs, we'll examine each letter in the word:

1. **L**
2. O
3. **L**
4. **L**
5. A
6. P
7. A
8. **L**
9. O
10. O
11. Z
12. A

Counting the positions where 'L' appears (positions 1, 3, 4, and 8), we find that the letter 'L' occurs **4 times** in "LOLLAPALOOZA".

**Answer:** 4
\end{lstlisting}
\end{quote}

\noindent\hrulefill 

\noindent\textbf{D.23 REAP OpenAI o1-preview Response - Score: 100\%}
\begin{quote}
\begin{lstlisting}[basicstyle=\ttfamily\small, breaklines=true]
# REAP Results:

**a. Comprehensive Feature List (with explicit quotes)**

- **Objects mentioned:**
  - "the letter 'L'"
  - "the word 'LOLLAPALOOZA'"

- **Actions described:**
  - "Count the number of occurrences of the letter 'L' in the word 'LOLLAPALOOZA'"

- **Actors or entities involved:**
  - (Implicitly) The person performing the count

- **Directional, spatial, relational, and sequential phrases:**
  - None explicitly stated

- **Stated conditions or constraints:**
  - The task is to "Count the number of occurrences of the letter 'L' in the word 'LOLLAPALOOZA'"

- **Numerical values or quantities mentioned:**
  - None explicitly provided

- **Temporal information:**
  - None

- **Spatial information:**
  - None explicitly provided

- **Causal relationships:**
  - None

- **Stated goals or objectives:**
  - To determine how many times the letter 'L' appears in 'LOLLAPALOOZA'

**b. Sequential and Mechanical Process Check**

- **Sequence of actions:**
  1. Examine the word "LOLLAPALOOZA"
  2. Identify each occurrence of the letter 'L'
  3. Count the total number of 'L's found

- **Does the sequence imply a certain outcome?**
  - Yes, following these steps will result in the total count of 'L's in the word

**c. Key Insight Check Results**

- **Key Insight:**
  - The word "LOLLAPALOOZA" contains multiple instances of the letter 'L', and careful examination is needed to count them all.

- **Immediate Answer:**
  - Not yet determined; counting is required.

**d. Rephrased Question:**

- "How many times does the letter 'L' appear in the word 'LOLLAPALOOZA'?"

**e. Known and Deduced Information**

- **Explicit Facts:**
  - "Count the number of occurrences of the letter 'L' in the word 'LOLLAPALOOZA'."

- **Deductions:**
  - The task involves identifying and counting each 'L' in the provided word.

**f. Problem Decomposition**

- **Component/Subcomponent:**
  - **Word to analyze:** "LOLLAPALOOZA"
    - **Supporting Quote:** "the word 'LOLLAPALOOZA'"
  - **Letter to count:** 'L'
    - **Supporting Quote:** "the letter 'L'"
  - **Task:** Counting occurrences
    - **Supporting Quote:** "Count the number of occurrences of the letter 'L' in the word 'LOLLAPALOOZA'"

**g. Graph of Thought (describe connections & insights)**

- **Nodes:**
  - 'L'
  - 'O'
  - 'A'
  - 'P'
  - 'Z'

- **Connections:**
  - The letters are arranged sequentially in the word "LOLLAPALOOZA".
  - The letter 'L' appears multiple times within this sequence.

- **Insights:**
  - Identifying the positions of 'L' can aid in accurate counting.

**h. Spatial and Object Analysis**

- **Objects:**
  - Individual letters in "LOLLAPALOOZA"

- **Spatial Relationships:**
  - Letters are ordered sequentially from first to last in the word.

- **Visualization:**
  - Mapping each letter to its position in the word:
    1. L
    2. O
    3. L
    4. L
    5. A
    6. P
    7. A
    8. L
    9. O
    10. O
    11. Z
    12. A

**i. Bayesian Updates**

- **Not applicable:** The task is deterministic with explicit information; no probability updates are necessary.

**j. Ethical Check and Decision-Making Under Uncertainty**

- **Assessment:**
  - No ethical concerns or severe risks are associated with this task.

**k. Multiple Solutions**

- **Solution 1:**
  - Count each 'L' in "LOLLAPALOOZA" by examining each letter sequentially.
- **Solution 2:**
  - Use a tally method to mark each occurrence of 'L' upon finding it.
- **Solution 3:**
  - Utilize a computational approach (e.g., programming) to count 'L's.

**l. Quickest and Easiest Solution**

- **Direct Count:**
  - Examine the word letter by letter and count the 'L's.

**m. Reflection**

- **Verification:**
  - Ensure all instances of 'L' have been counted and none are missed or double-counted.
- **Conclusion:**
  - The count is based solely on the explicit information provided.

---

**Recommendation:**

After analyzing the word "LOLLAPALOOZA" by examining each letter individually:

1. **List the letters in order:**

   1. L
   2. O
   3. L
   4. L
   5. A
   6. P
   7. A
   8. L
   9. O
   10. O
   11. Z
   12. A

2. **Identify and count the occurrences of 'L':**

   - Position 1: **L**
   - Position 3: **L**
   - Position 4: **L**
   - Position 8: **L**

3. **Total number of 'L's:**

   - **4**

**Final Answer:**

There are **4** occurrences of the letter 'L' in the word "LOLLAPALOOZA".
\end{lstlisting}
\end{quote}

\noindent\hrulefill 

\noindent\textbf{D.23 Zero-Shot OpenAI o1-mini Response - Score: 100\%}
\begin{quote}
\begin{lstlisting}[basicstyle=\ttfamily\small, breaklines=true]
The letter **'L'** appears **4 times** in the word **'LOLLAPALOOZA'**.

**Breakdown:**

1. **L**OLLAPALOOZA
2. L**O**LLAPALOOZA
3. LO**L**LAPALOOZA
4. LOL**L**APALOOZA
5. LOLLAPALOOZA (No 'L' here)
6. LOLL**A**PALOOZA (No 'L' here)
7. LOLLAP**A**LOOZA (No 'L' here)
8. LOLLAPA**L**OOZA
9. LOLLAPAL**O**OZA (No 'L' here)
10. LOLLAPALO**O**ZA (No 'L' here)
11. LOLLAPALOO**Z**A (No 'L' here)
12. LOLLAPALOOZ**A** (No 'L' here)

So, the positions with 'L' are 1st, 3rd, 4th, and 8th.
\end{lstlisting}
\end{quote}

\noindent\hrulefill 

\noindent\textbf{D.23 REAP OpenAI o1-mini Response - Score: 100\%}
\begin{quote}
\begin{lstlisting}[basicstyle=\ttfamily\small, breaklines=true]
# REAP Results:

### a. **Comprehensive Feature List**
- **Object Mentioned:** "the letter 'L'"
- **Word Analyzed:** "'LOLLAPALOOZA'"
- **Action Described:** "Count the number of occurrences of"
- **Letters in Word:** 'L', 'O', 'L', 'L', 'A', 'P', 'A', 'L', 'O', 'O', 'Z', 'A'

### b. **Sequential and Mechanical Process Check**
- **Sequence of Actions:** 
  1. Identify each letter in the word 'LOLLAPALOOZA'.
  2. Count each occurrence of the letter 'L'.
- **Impact on Outcome:** Following this sequence ensures an accurate count of 'L's.

### c. **Key Insight Check Results**
- **Key Insight Identified:** The problem requires a literal count of the letter 'L' in the provided word without any additional interpretation or context.

### d. **Rephrased Question:**
What is the total number of times the letter 'L' appears in the word 'LOLLAPALOOZA'?

### e. **Known and Deduced Information**
- **Explicit Facts:**
  - "the letter 'L'"
  - "'LOLLAPALOOZA'"
  - "Count the number of occurrences of"
- **Feature-Based Deduction:**
  - **Deduction:** Each 'L' in the word should be counted individually.
    - **Logical Basis:** The instruction explicitly asks to "count the number of occurrences."

### f. **Problem Decomposition**
- **Component:** Identification of Letters
  - **Supporting Quote(s):** "the letter 'L'", "'LOLLAPALOOZA'"
- **Component:** Counting Occurrences
  - **Supporting Quote(s):** "Count the number of occurrences of"

### g. **Graph of Thought**
- **Nodes:**
  - Letter 'L'
  - Word 'LOLLAPALOOZA'
- **Connections:**
  - The letter 'L' is a component of the word 'LOLLAPALOOZA'.
  - The task is to count how many times 'L' appears within the word.

### h. **Spatial and Object Analysis**
- **Objects Identified:** The word 'LOLLAPALOOZA'
- **Spatial Relationships:** The arrangement of letters within the word affects the count of 'L's.
- **Visualization:** L-O-L-L-A-P-A-L-O-O-Z-A

### i. **Bayesian Updates**
- **Belief Update:** Initial belief is that counting letters directly will yield the correct number of 'L's.
- **New Information:** Confirmation that no hidden or implied rules are present; the count is straightforward.

### j. **Ethical Check and Decision-Making Under Uncertainty**
- **Assessment:** The task involves a simple letter count with no ethical implications.
- **Decision:** Proceed with counting as per instructions.

### k. **Multiple Solutions**
1. **Manual Counting:**
   - **Support:** Directly follows "Count the number of occurrences of the letter 'L'".
2. **Visual Inspection:**
   - **Support:** Reading the word and noting each 'L' visually.
3. **Using a Counter Tool:**
   - **Support:** Employing a tally counter to mark each 'L' as it is identified.

### l. **Quickest and Easiest Solution**
Manually count each 'L' in 'LOLLAPALOOZA' by reading the word from left to right and tallying each occurrence.

### m. **Reflection**
All steps adhere strictly to the explicit information provided. No assumptions or inferences beyond the literal instructions have been made. The problem is straightforward, and the analysis confirms that a direct counting approach is appropriate.

---

**Recommendation:**

Based on the provided information, the most straightforward and reliable method is to manually count each 'L' in the word 'LOLLAPALOOZA'. This approach directly follows the instructions without introducing any unnecessary complexity.

**Counting Result:**
- 'L' appears **4** times in 'LOLLAPALOOZA'.
\end{lstlisting}
\end{quote}

\noindent\hrulefill 

\noindent\textbf{D.23 Zero-Shot OpenAI GPT-4o Response - Score: 100\%}
\begin{quote}
\begin{lstlisting}[basicstyle=\ttfamily\small, breaklines=true]
The letter 'L' appears 4 times in the word "LOLLAPALOOZA."
\end{lstlisting}
\end{quote}
\noindent\hrulefill

\noindent\textbf{D.23 REAP OpenAI GPT-4o Response - Score: 100\%}
\begin{quote}
\begin{lstlisting}[basicstyle=\ttfamily\small, breaklines=true]
# REAP Results:

**a. Comprehensive Feature List:**
   - **Objects Mentioned:**
     - "The letter 'L'"
     - "The word 'LOLLAPALOOZA'"
   - **Actions Described:**
     - "Count the number of occurrences of the letter 'L'"
   - **Directional/Relational Phrases:**
     - None explicitly mentioned
   - **Numerical Values:**
     - None explicitly mentioned, but the concept of counting implies a numerical result.
   - **Temporal Information:**
     - None explicitly mentioned
   - **Spatial Information:**
     - None explicitly mentioned
   - **Causal Relationships:**
     - None explicitly mentioned
   - **Stated Goals/Objectives:**
     - "Count the number of occurrences of the letter 'L'"

**b. Sequential and Mechanical Process Check:**
   - **Sequence of Actions:**
     - Identify the letter 'L' in the word "LOLLAPALOOZA".
     - Count each occurrence.

**c. Key Insight Check Results:**
   - The key insight is straightforward: the task is to count each 'L' in the word "LOLLAPALOOZA".

**d. Rephrased Question:**
   - How many times does the letter 'L' appear in the word "LOLLAPALOOZA"?

**e. Known and Deduced Information:**
   - "[The letter 'L']" appears multiple times in the word "LOLLAPALOOZA."
   - Deduction: The number of 'L's is exactly as many times as it appears upon inspection. No hidden or implied occurrences exist.

**f. Problem Decomposition:**
   - **Component 1: The word "LOLLAPALOOZA"**
     - Supporting Quote(s): "The word 'LOLLAPALOOZA'"
   - **Component 2: The task of counting**
     - Supporting Quote(s): "Count the number of occurrences of the letter 'L'"

**g. Graph of Thought:**
   - **Nodes:**
     - "LOLLAPALOOZA"
     - "Letter 'L'"
   - **Edges:**
     - "Occurrences of 'L' in 'LOLLAPALOOZA'"
   - **Pathways:**
     - Count the occurrences of 'L' as it appears in sequence in the word.

**h. Spatial and Object Analysis:**
   - **Objects Identified:**
     - The letter 'L'
     - The word "LOLLAPALOOZA"
   - **Spatial Relationships:**
     - The letter 'L' appears multiple times within the spatial arrangement of the word "LOLLAPALOOZA".
   - **Movement/Dynamics:**
     - Static analysis, no movement involved.

**i. Bayesian Updates:**
   - No updates are necessary as the problem is deterministic and involves counting explicitly stated occurrences.

**j. Ethical Check and Decision-Making Under Uncertainty:**
   - There are no ethical concerns or uncertainties in this problem. The task is simple and straightforward.

**k. Multiple Solutions:**
   - **Solution 1:** Count the 'L's manually in "LOLLAPALOOZA."
   - **Solution 2:** Verify the count by scanning the word from left to right.
   - **Solution 3:** Use a computer program to count the occurrences of 'L'.

**l. Quickest and Easiest Solution:**
   - **Manual Counting:** Count each 'L' in "LOLLAPALOOZA" by visually inspecting the word.

**m. Reflection:**
   - The problem asks for a literal count of the letter 'L' in a given word. The interpretation and solution are straightforward, with no room for alternative interpretations. The count is accurate based solely on the explicit information provided.

**Recommendation:**
   - The letter 'L' appears **four** times in the word "LOLLAPALOOZA." This conclusion is based on the literal interpretation and manual counting of the letters within the word. No additional complexities or ethical considerations are involved.
\end{lstlisting}
\end{quote}
\noindent\hrulefill

\noindent\textbf{D.23 Zero-Shot OpenAI GPT-4o-mini Response - Score: 100\%}
\begin{quote}
\begin{lstlisting}[basicstyle=\ttfamily\small, breaklines=true]
To count the number of occurrences of the letter 'L' in the word 'LOLLAPALOOZA', let's examine the word carefully:

- L
- O
- L
- L
- A
- P
- A
- L
- O
- O
- Z
- A

Counting the 'L's:

1. First 'L'
2. Second 'L'
3. Third 'L'
4. Fourth 'L'

There are 4 occurrences of the letter 'L' in the word 'LOLLAPALOOZA'.
\end{lstlisting}
\end{quote}
\noindent\hrulefill

\noindent\textbf{D.23 REAP OpenAI GPT-4o-mini Response - Score: 100\%}
\begin{quote}
\begin{lstlisting}[basicstyle=\ttfamily\small, breaklines=true]
# REAP Results:

**a. Comprehensive Feature List:**

- **Word in Question:** "LOLLAPALOOZA"
- **Letter to Count:** 'L'

**b. Sequential and Mechanical Process Check:**

- **Sequence of Actions:** 
  1. Identify the occurrences of 'L' in "LOLLAPALOOZA".
  2. Count the total number of times 'L' appears.

- **Key Question:** Does the process of counting 'L' in "LOLLAPALOOZA" yield a specific number of occurrences?

**c. Key Insight Check Results:**

- **Key Insight:** The word "LOLLAPALOOZA" contains multiple instances of the letter 'L'.
- **Direct Answer:** The exact number of occurrences of 'L' can be determined by counting.

**d. Rephrased Question:**

How many times does the letter 'L' appear in the word "LOLLAPALOOZA"?

**e. Known and Deduced Information:**

- **Exact Quotes:**
  - Word: "LOLLAPALOOZA"
  - Letter: 'L'

- **Feature-Based Deduction:**
  - Deduction: The number of 'L's in "LOLLAPALOOZA" can be determined by counting each occurrence of 'L'.
  - Logical Basis: The task is to count occurrences, which is a direct counting task.
  - Based on: "Count the number of occurrences of the letter 'L' in the word 'LOLLAPALOOZA'."

**f. Problem Decomposition:**

- **Component:** Word "LOLLAPALOOZA"
  - **Supporting Quote:** "LOLLAPALOOZA"
  - **Deduction:** The word contains several letters, including 'L'.
  
- **Component:** Counting 'L'
  - **Supporting Quote:** "Count the number of occurrences of the letter 'L'."
  - **Deduction:** The task is to identify and count 'L' from the given word.

**g. Graph of Thought:**

- **Nodes:**
  - Word: "LOLLAPALOOZA"
  - Letter to Count: 'L'
  - Action: Counting

- **Edges:**
  - Word contains multiple letters.
  - Letter 'L' is present in the word.
  - Action: Count each 'L' in the word.

- **Pathways:**
  - Identify all 'L's in "LOLLAPALOOZA".
  - Count the occurrences.

**h. Spatial and Object Analysis:**

- **Objects:**
  - The word "LOLLAPALOOZA".
  - The letter 'L'.

- **Spatial Relationships:**
  - The letter 'L' appears at specific positions within the word.

- **Movement and Dynamics:**
  - No movement, only counting occurrences.

- **Constraints:**
  - Counting should be accurate.

**i. Bayesian Updates:**

- **Update:** No new information is introduced that changes the counting process.

**j. Ethical Check and Decision-Making Under Uncertainty:**

- **Assessment:** The problem involves a simple counting task with no severe risks.

**k. Multiple Solutions:**

- **Solution 1:** Count each occurrence of 'L' in "LOLLAPALOOZA".
  - **Support:** Direct counting.

- **Solution 2:** Use a text analysis tool to count occurrences.
  - **Support:** The task can be automated.

- **Solution 3:** Manually count each 'L' in the word.

**l. Quickest and Easiest Solution:**

- **Solution:** Count the number of occurrences of 'L' in "LOLLAPALOOZA".

**m. Reflection:**

- **Review:** All steps followed are based on explicit information in the problem.
- **Contradictions:** None identified.
- **Assumptions:** None made beyond the explicit task of counting.

**Recommendation:**

The number of occurrences of the letter 'L' in the word "LOLLAPALOOZA" is determined by counting each 'L'. The letter 'L' appears **4 times** in the word.
\end{lstlisting}
\end{quote}
\noindent\hrulefill

\noindent\textbf{D.23 Zero-Shot Google Gemini 1.5 Pro Response - Score: 100\%}
\begin{quote}
\begin{lstlisting}[basicstyle=\ttfamily\small, breaklines=true]
The letter 'L' appears **4** times in the word 'LOLLAPALOOZA'.
\end{lstlisting}
\end{quote}
\noindent\hrulefill

\noindent\textbf{D.23 REAP Google Gemini 1.5 Pro Response - Score: 100\%}
\begin{quote}
\begin{lstlisting}[basicstyle=\ttfamily\small, breaklines=true]
# REAP Results:

a. **Comprehensive Feature List** (with explicit quotes)

* "Count the number of occurrences of the letter 'L' in the word 'LOLLAPALOOZA'."

b. **Sequential and Mechanical Process Check**

* There is no sequential or mechanical process mentioned.

c. **Key Insight Check Results**

* The problem directly asks us to count a specific letter in a given word. This is a straightforward counting task.

d. **Rephrased Question:**

* How many times does the letter 'L' appear in 'LOLLAPALOOZA'?

e. **Known and Deduced Information**

* Explicit Facts:
    * "Count the number of occurrences of the letter 'L' in the word 'LOLLAPALOOZA'." - This is the core instruction of the problem.

* Feature-Based Deduction:
    * None

f. **Problem Decomposition**

* Component: Counting the occurrences of 'L'
    * Supporting Quote(s): "Count the number of occurrences of the letter 'L' in the word 'LOLLAPALOOZA'."

g. **Graph of Thought**

* Key Concepts: Letter 'L', Word 'LOLLAPALOOZA', Counting
* Relationships: We need to count how many times 'L' appears within the given word
* Insights: This is a simple counting task within a defined string

h. **Spatial and Object Analysis**

* Objects: Letter 'L', Word 'LOLLAPALOOZA'
* Spatial Relationships: 'L' is located within the word 'LOLLAPALOOZA' at various positions
* Spatial Implications: The spatial arrangement of 'L' within the word is crucial for accurate counting

i. **Bayesian Updates**

* No Bayesian updates are applicable as there is no prior information or new evidence to consider

j. **Ethical Check and Decision-Making Under Uncertainty**

* There are no ethical concerns or uncertainties in this problem

k. **Multiple Solutions**

* There is only one straightforward solution: count the occurrences of 'L' in the given word

l. **Quickest and Easiest Solution**

* Count the number of times the letter 'L' appears in the word 'LOLLAPALOOZA'.
* 'L' appears 4 times in the word.
* Therefore, the answer is 4

m. **Reflection**

* The analysis directly addresses the question asked in the problem.
* All conclusions are based solely on the explicitly stated information.
* No assumptions or inferences were made beyond the given text.
* The solution is straightforward and logically sound

**Recommendation:**

* The final answer is 4. The letter 'L' appears 4 times in the word 'LOLLAPALOOZA'.
\end{lstlisting}
\end{quote}
\noindent\hrulefill

\noindent\textbf{D.23 Zero-Shot Claude 3.5 Sonnet response - Score: 100\%}
\begin{quote}
\begin{lstlisting}[basicstyle=\ttfamily\small, breaklines=true]
Let's count the occurrences of 'L' in 'LOLLAPALOOZA':

1. L
2. L
3. L
4. L

There are 4 occurrences of the letter 'L' in the word 'LOLLAPALOOZA'.
\end{lstlisting}
\end{quote}
\noindent\hrulefill

\noindent\textbf{D.23 REAP Claude 3.5 Sonnet Response - Score: 100\%}
\begin{quote}
\begin{lstlisting}[basicstyle=\ttfamily\small, breaklines=true]
#REAP Results:

a. Comprehensive Feature List (with explicit quotes):
- "Count the number of occurrences"
- "letter 'L'"
- "word 'LOLLAPALOOZA'"

b. Sequential and Mechanical Process Check:
No explicit sequential or mechanical process is described in the problem statement.

c. Key Insight Check Results:
The key insight is that we need to count a specific letter in a given word. This is directly stated in the problem and doesn't require complex analysis.

d. Rephrased Question:
How many times does the letter 'L' appear in 'LOLLAPALOOZA'?

e. Known and Deduced Information:
Explicit Facts:
- "Count the number of occurrences" - We need to perform a counting operation
- "letter 'L'" - The specific letter we're counting
- "word 'LOLLAPALOOZA'" - The word in which we're counting the letter

Feature-Based Deduction:
Deduction: The count will be a non-negative integer.
Logical Basis: Counting occurrences of a letter in a word always results in a non-negative integer.
Based on: "Count the number of occurrences", "letter 'L'", "word 'LOLLAPALOOZA'"

f. Problem Decomposition:
Component: Counting task
Supporting Quote: "Count the number of occurrences"

Component: Target letter
Supporting Quote: "letter 'L'"

Component: Source word
Supporting Quote: "word 'LOLLAPALOOZA'"

g. Graph of Thought:
Node 1: Letter 'L'
Node 2: Word 'LOLLAPALOOZA'
Edge: Count occurrences of Node 1 in Node 2

h. Spatial and Object Analysis:
Objects:
1. Letter 'L'
2. Word 'LOLLAPALOOZA'

Spatial Relationship: The letter 'L' appears multiple times within the word 'LOLLAPALOOZA'.

i. Bayesian Updates:
No new information is provided beyond the initial problem statement, so no Bayesian updates are necessary.

j. Ethical Check and Decision-Making Under Uncertainty:
This problem does not involve any ethical considerations or significant risks. It's a straightforward counting task.

k. Multiple Solutions:
1. Manual counting: Go through the word 'LOLLAPALOOZA' letter by letter and count the 'L's.
2. Use of a programming function: If implemented in a programming language, use a string counting function to count the occurrences of 'L'.
3. Visual grouping: Visually group the 'L's in the word and then count the groups.

l. Quickest and Easiest Solution:
The quickest and easiest solution is to manually count the occurrences of 'L' in 'LOLLAPALOOZA'.

m. Reflection:
The analysis directly addresses the question asked in the problem. All conclusions are based solely on the given information without any assumptions. The task is straightforward and doesn't require complex reasoning or additional information.

Recommendation:
Based on the given information and analysis, I recommend manually counting the occurrences of the letter 'L' in the word 'LOLLAPALOOZA'. This method is quick, simple, and directly answers the question. The result of this count is 4. The letter 'L' appears four times in the word 'LOLLAPALOOZA'.
\end{lstlisting}
\end{quote}
\noindent\hrulefill

\label{subsec:question24}
\subsection{\textbf{\large Question 24: How many pairs of twins do you need in a room for there to be at least a 50\% chance that two people have the same birthday?}}
\label{sec:question24}

\noindent\hrulefill 

\noindent\textbf{Human Response - Score 100\%:}
\begin{quote}
One
\end{quote}

\noindent\hrulefill 

\noindent\textbf{D.24 Zero-Shot OpenAI o1-preview Response - Score: 0\%}
\begin{quote}
\begin{lstlisting}[basicstyle=\ttfamily\small, breaklines=true]
To solve this problem, we need to determine the minimum number of twin pairs required in a room so that there's at least a 50% chance that two people (outside of their own twin pair) share the same birthday.

Here's how we approach it:

1. **Understand the Problem Structure:**
   - There are \( n \) pairs of twins, making \( 2n \) people in total.
   - Each twin pair shares the same birthday.
   - We need to find \( n \) such that the probability of at least two people (from different twin pairs) sharing a birthday is at least 50%.

2. **Simplify the Problem:**
   - Since each twin pair shares a birthday, we effectively have \( n \) unique birthdays to consider.
   - This reduces the problem to the classic "birthday problem" with \( n \) individuals.

3. **Apply the Birthday Problem Solution:**
   - In the standard birthday problem, it's well-known that with 23 people, there's a greater than 50% chance that at least two people share a birthday.
   - Therefore, with \( n = 23 \) twin pairs (46 people), the probability that at least two twin pairs share a birthday exceeds 50%.

4. **Conclusion:**
   - The minimum number of twin pairs required is **23**.

**Answer:** 23
\end{lstlisting}
\end{quote}

\noindent\hrulefill 

\noindent\textbf{D.24 REAP OpenAI o1-preview Response - Score: 0\%}
\begin{quote}
\begin{lstlisting}[basicstyle=\ttfamily\small, breaklines=true]
# REAP Results:

**a. Comprehensive Feature List (with explicit quotes)**

- **Objects Mentioned:**
  - "pairs of twins"
  - "a room"
  - "two people"
  - "the same birthday"

- **Actors/Entities Involved:**
  - "you"
  - "pairs of twins"
  - "two people"

- **Actions Described:**
  - "do you need"
  - "have the same birthday"

- **Conditions or Constraints:**
  - "How many pairs of twins do you need in a room for there to be at least a 50% chance"
  - "at least a 50% chance"
  - "two people have the same birthday"

- **Numerical Values or Quantities:**
  - "at least a 50% chance"

- **Temporal Information:**
  - "the same birthday"

- **Spatial Information:**
  - "in a room"

- **Stated Goal or Objective:**
  - "How many pairs of twins do you need in a room for there to be at least a 50% chance that two people have the same birthday?"

**b. Sequential and Mechanical Process Check**

- **Sequence of Events:**
  - Increasing the number of "pairs of twins" in "a room" affects the "chance that two people have the same birthday."

- **Influence on Outcome:**
  - As more pairs of twins are added, the probability of "two people" sharing "the same birthday" changes.

- **Logical Sequence:**
  - Need to determine how the number of twin pairs relates to reaching "at least a 50% chance" of shared birthdays.

**c. Key Insight Check Results**

- **Key Insight:**
  - Each "pair of twins" shares "the same birthday," effectively reducing the number of unique birthdays compared to the number of people.
  - The problem parallels the classic birthday problem but with twin pairs acting as single units with shared birthdays.

**d. Rephrased Question**

- What is the minimum number of twin pairs needed in a room so that there is at least a 50% chance that two people share the same birthday?

**e. Known and Deduced Information**

- **Explicit Facts:**
  - "pairs of twins" - Groups of two people.
  - "two people have the same birthday" - The event of interest.
  - "at least a 50% chance" - The required probability threshold.
  - "in a room" - All individuals are in the same location.

- **Valid Deductions:**
  - Twins share the same birthday. (Common understanding of "twins.")
  - Total number of people is twice the number of twin pairs.
  - Number of unique birthdays is at most equal to the number of twin pairs.

**f. Problem Decomposition**

- **Component:** Calculate the probability that at least two people share the same birthday among 2n individuals arranged as n twin pairs.
  - **Supporting Quote:** "How many pairs of twins do you need in a room for there to be at least a 50% chance that two people have the same birthday?"
- **Component:** Understand that each twin pair contributes one unique birthday.
  - **Deduction Based on:** Twins share the same birthday.
- **Component:** Apply the classic birthday problem to the twin pairs.

**g. Graph of Thought**

- **Nodes:**
  - Twin Pairs (n)
  - Total People (2n)
  - Unique Birthdays (n)
  - Probability of Shared Birthday

- **Connections:**
  - Each Twin Pair -> One Unique Birthday
  - More Twin Pairs -> Higher Probability of Shared Birthdays among pairs
  - Probability depends on the number of unique birthdays (n)

**h. Spatial and Object Analysis**

- **Objects Identified:**
  - Twin pairs (each consisting of two people with the same birthday)
  - Room (location where all individuals are present)

- **Spatial Relationships:**
  - All twin pairs are together "in a room," allowing for the possibility of shared birthdays among any "two people."

- **Object Properties:**
  - Twins: Share the same birthday.

**i. Bayesian Updates**

- **Knowledge Update:** 
  - As the number of twin pairs increases, the prior probability of shared birthdays among any two people increases accordingly.

**j. Ethical Check and Decision-Making Under Uncertainty**

- **Assessment:**
  - No ethical concerns or severe risks are associated with the problem.
  - Safe to proceed with calculations based on given information.

**k. Multiple Solutions**

1. **Solution Using Twin Pairs as Single Units:**
   - Treat each twin pair as one individual with a unique birthday.
   - Apply the classic birthday problem formula with n individuals.

2. **Solution Considering All Individuals:**
   - Account for the fact that there are 2n people but only n unique birthdays.
   - Adjust the birthday problem calculations accordingly.

3. **Alternative Approach:**
   - Use simulation or probabilistic methods to empirically determine the number of twin pairs needed to reach the 50% threshold.

**l. Quickest and Easiest Solution**

- **Applying the Classic Birthday Problem:**
  - In the standard problem, 23 individuals are needed for a 50% chance that two share a birthday.
  - Since each twin pair shares a birthday, treat twin pairs as single units.
  - Therefore, **23 twin pairs** are needed to reach the 50% probability threshold.

**m. Reflection**

- **Consistency Check:**
  - The approach aligns with the classic birthday problem.
  - No assumptions beyond common knowledge about twins were made.
  - The solution directly answers the question using only the provided information.

---

**Recommendation:**

You need **23 pairs of twins** in a room to have at least a 50% chance that two people share the same birthday.
\end{lstlisting}
\end{quote}

\noindent\hrulefill 

\noindent\textbf{D.24 Zero-Shot OpenAI o1-mini Response - Score: 0\%}
\begin{quote}
\begin{lstlisting}[basicstyle=\ttfamily\small, breaklines=true]
To determine how many pairs of twins are needed in a room to have at least a 50% chance that two people share the same birthday, we can break down the problem as follows:

### Understanding the Problem

1. **Twin Pairs:** Each pair of twins shares the same birthday. So, if there are \( n \) twin pairs, there are \( 2n \) individuals in the room.
2. **Birthday Assignment:** Each twin pair has one unique birthday (shared by both twins). These birthdays are assumed to be uniformly and independently distributed across 365 days (ignoring leap years).

### Simplifying the Problem

Since each twin pair shares a birthday, the problem reduces to finding the number of independent birthday assignments (\( n \)) needed such that the probability of at least two pairs sharing the same birthday is at least 50%.

This scenario is analogous to the classic **birthday problem**, where the goal is to find the number of people needed in a room for there to be a 50% chance that at least two people share the same birthday. In the standard problem, the answer is **23 people**.

### Applying to Twin Pairs

In our case:

- **Number of "Independent" Birthdays:** \( n \) pairs of twins correspond to \( n \) independent birthday assignments.
- **Probability Calculation:** We use the same logic as the standard birthday problem. The probability that no two pairs share the same birthday is:

  \[
  P(\text{no shared birthdays}) = \prod_{k=0}^{n-1} \left(1 - \frac{k}{365}\right)
  \]

  We seek the smallest \( n \) such that:

  \[
  1 - P(\text{no shared birthdays}) \geq 0.5
  \]

### Conclusion

For the probability to exceed 50%, **23 twin pairs** are needed. This mirrors the classic birthday problem's result because we're essentially dealing with 23 independent birthday assignments rather than 23 individual people.

**Answer:**  
A total of 23 twin pairs are needed to reach a 50% probability of a shared birthday.
\end{lstlisting}
\end{quote}

\noindent\hrulefill 

\noindent\textbf{D.24 REAP OpenAI o1-mini Response - Score: 100\%}
\begin{quote}
\begin{lstlisting}[basicstyle=\ttfamily\small, breaklines=true]
# REAP Results:

### a. **Comprehensive Feature List**

- **Objects Mentioned:**
  - "pairs of twins"
  - "room"
  - "two people"
  - "same birthday"

- **Actors or Entities Involved:**
  - "twins"
  - "people"

- **Actions Described:**
  - "need in a room"
  - "have the same birthday"

- **Directional, Spatial, Relational, and Sequential Phrases:**
  - "in a room"
  - "two people"

- **Stated Conditions or Constraints:**
  - "at least a 50% chance"
  - "two people have the same birthday"

- **Numerical Values or Quantities Mentioned:**
  - "50% chance"

- **Temporal Information:**
  - None explicitly stated.

- **Spatial Information:**
  - "in a room"

- **Causal Relationships:**
  - None explicitly stated.

- **Stated Goals or Objectives:**
  - Determine the number of "pairs of twins" needed to achieve "at least a 50% chance that two people have the same birthday."

### b. **Sequential and Mechanical Process Check**

- **Sequence of Events or Actions:**
  - Introduce pairs of twins into a room.
  - Assess the probability that two people share the same birthday.

- **Influence on Outcome:**
  - Each pair of twins introduces two people with the same birthday, directly affecting the probability.

- **Key Questions Addressed:**
  - How each additional pair of twins impacts the overall probability of shared birthdays.
  - Whether the introduction of twins changes the standard birthday problem dynamics.

### c. **Key Insight Check Results**

- **Potential Key Insight:**
  - Each pair of twins guarantees at least one shared birthday within the room.

- **Explanation:**
  - Since twins share the same birthday, introducing one pair already ensures that two people in the room have the same birthday, achieving a 100% probability.

### d. **Rephrased Question:**

How many pairs of twins must be present in a room to ensure there is at least a 50% probability that two individuals share the same birthday?

### e. **Known and Deduced Information**

- **Explicit Facts:**
  - "pairs of twins"
  - "in a room"
  - "two people have the same birthday"
  - "at least a 50% chance"

- **Feature-Based Deduction:**
  - **Deduction:** One pair of twins ensures that two people in the room share the same birthday.
    - **Logical Basis:** Twins inherently share the same birthday.
    - **Based on:** "pairs of twins", "two people have the same birthday"

### f. **Problem Decomposition**

- **Component/Subcomponent:** Introduction of twin pairs into a room.
  - **Supporting Quote(s):** "pairs of twins do you need in a room"

- **Component/Subcomponent:** Probability assessment of shared birthdays.
  - **Supporting Quote(s):** "at least a 50% chance that two people have the same birthday"

- **Interaction Between Components:**
  - Each twin pair contributes to the probability of sharing birthdays by introducing duplicate birthdays into the room.

### g. **Graph of Thought**

- **Nodes:**
  - Twins
  - Room
  - Shared Birthday
  - Probability (greater than or equal to 50%)

- **Connections:**
  - Twins -> Shared Birthday (Twins share the same birthday)
  - Shared Birthday -> Probability (Each shared birthday increases the probability of at least two people sharing a birthday)
  - Twins -> Room (Twins are introduced into the room)

- **Insights:**
  - Introducing twins directly impacts the shared birthday probability by ensuring duplicates.

### h. **Spatial and Object Analysis**

- **Objects Identified:**
  - Pairs of twins
  - Room
  - People

- **Spatial Relationships:**
  - Twins are present within the confines of the room.

- **Movement and Dynamics:**
  - Static presence of twins; no movement described.

- **Constraints:**
  - Limited to the number of twin pairs introduced.
  - Focus on birthdays, not other attributes.

- **Visualization:**
  - Imagine a room where each introduced pair of twins adds two individuals sharing the same birthday.

- **Object Properties:**
  - Twins share an identical birthday.

- **Spatial Implications:**
  - The room's population increases by two with each twin pair, directly affecting the probability calculations.

### i. **Bayesian Updates**

- **Initial Belief:** The standard birthday problem requires 23 individuals for a greater than or equal to 50% probability of shared birthdays.
  
- **New Information:** Each pair of twins adds two people with a guaranteed shared birthday.
  
- **Updated Belief:** The presence of twins alters the probability dynamics, potentially requiring fewer twin pairs to achieve the 50% threshold.

### j. **Ethical Check and Decision-Making Under Uncertainty**

- **Assess Known Information:** Introducing twin pairs guarantees shared birthdays.
  
- **Identify Risks:** Minimal ethical concerns; the problem is theoretical.
  
- **Probability Consideration:** Achieving at least a 50% chance is straightforward with twin pairs.
  
- **Risk Aversion Principle:** N/A as the scenario poses no significant risks.
  
- **Ethical and Safety Considerations:** N/A.
  
- **Decision Recommendation:** Utilize the deterministic nature of twin births to determine the required number of pairs.

### k. **Multiple Solutions**

1. **Solution 1:**
   - **Approach:** Recognize that one pair of twins ensures two people share a birthday, meeting the greater than or equal to 50% requirement.
   - **Supporting Information:** "pairs of twins", "two people have the same birthday"

2. **Solution 2:**
   - **Approach:** Calculate based on probability, considering each twin pair increases the chance exponentially, but since each pair already guarantees a shared birthday, the minimum is one pair.
   - **Supporting Information:** "at least a 50% chance", "pairs of twins"

3. **Solution 3:**
   - **Approach:** Compare to the standard birthday problem and adjust for guaranteed duplicates introduced by twins, concluding fewer twin pairs are needed, specifically one pair.
   - **Supporting Information:** "pairs of twins", "two people have the same birthday"

- **Note:** All solutions converge to the same conclusion due to the deterministic nature of twins sharing birthdays.

### l. **Quickest and Easiest Solution**

- **Solution:** Only one pair of twins is needed in the room to ensure at least a 50% chance that two people share the same birthday.

- **Reasoning:** Since twins share the same birthday, introducing one pair guarantees that two individuals in the room have identical birthdays, achieving a 100% probability, which exceeds the 50% threshold.

- **Supporting Quotes:** 
  - "pairs of twins do you need in a room"
  - "at least a 50% chance that two people have the same birthday"

### m. **Reflection**

- **Contradictions:** None identified; the deterministic nature of twins sharing birthdays aligns with the conclusion.

- **Logical Soundness:** The analysis logically deduces that one twin pair suffices based on explicit information.

- **Assumptions:** None beyond what is explicitly stated.

- **Direct Answering:** The analysis directly addresses the question using the given information.

- **Information Sufficiency:** The problem provides enough information to determine that one pair of twins meets the requirement.

### **Recommendation:**

Introduce **one pair of twins** into the room to ensure there is at least a 50% chance that two people share the same birthday. Since twins inherently share a birthday, a single pair guarantees that two individuals in the room have identical birthdays, achieving a 100% probability, which satisfies and exceeds the 50% requirement.
\end{lstlisting}
\end{quote}

\noindent\hrulefill 

\noindent\textbf{D.24 Zero-Shot OpenAI GPT-4o Response - Score: 0\%}
\begin{quote}
\begin{lstlisting}[basicstyle=\ttfamily\small, breaklines=true]
This problem is a variation of the classic "birthday problem," but instead of just individuals, it involves pairs of twins. To find out how many pairs of twins are needed for at least a 50% chance that two people have the same birthday, we can follow a similar logic as the original birthday problem but considering pairs of twins.

### Assumptions:
1. Each twin in a pair shares the same birthday.
2. There are 365 possible birthdays (ignoring leap years).
3. Birthdays are uniformly distributed.

### Step-by-Step Solution:
1. **Probability that no two pairs share the same birthday:**
   - The first pair of twins can have any birthday.
   - The second pair of twins must have a different birthday from the first pair, giving them \(364/365\) possible birthdays.
   - The third pair of twins must have a different birthday from the first two pairs, giving them \(363/365\) possible birthdays, and so on.

   The probability that all \(n\) pairs of twins have different birthdays is:

   \[
   P(\text{no shared birthday}) = \frac{365}{365} \times \frac{364}{365} \times \frac{363}{365} \times \cdots \times \frac{365 - (n-1)}{365}
   \]

   This simplifies to:

   \[
   P(\text{no shared birthday}) = \frac{365!}{(365-n)! \times 365^n}
   \]

2. **Probability that at least two pairs share the same birthday:**
   - This is simply \(1 - P(\text{no shared birthday})\).

3. **Finding the number of pairs for a 50% chance:**
   - We need to solve for \(n\) in the equation:

   \[
   1 - \frac{365!}{(365-n)! \times 365^n} \geq 0.5
   \]

Given the complexity of solving this equation directly, it's easier to use an approximation or numerical methods to estimate \(n\).

### Calculation:
To simplify, we use an approximation formula based on the birthday problem. The classic result gives that the number \( n \) of people needed for a 50% chance is approximately:

\[
n \approx \sqrt{2 \times 365 \times \ln(2)} \approx 23 \text{ people}
\]

Since we are dealing with twins, each pair counts as one unit, so the number of pairs needed is half of this number. However, because each pair is treated as a single entity for the probability, we would actually look for when the probability reaches 50%.

Through calculations or a table of values, it turns out that the required number of pairs for a 50% chance is around 14 pairs.

### Conclusion:
You need approximately **14 pairs of twins** in a room for there to be at least a 50% chance that two people (from different pairs) have the same birthday.
\end{lstlisting}
\end{quote}
\noindent\hrulefill

\noindent\textbf{D.24 REAP OpenAI GPT-4o Response - Score: 0\%}
\begin{quote}
\begin{lstlisting}[basicstyle=\ttfamily\small, breaklines=true]
# REAP Results:

### a. **Comprehensive Feature List** (with explicit quotes)
1. "How many pairs of twins do you need in a room..."
2. "...for there to be at least a 50% chance..."
3. "...that two people have the same birthday?"

### b. **Sequential and Mechanical Process Check**
- The problem asks for a number of pairs of twins required to reach a probability of 50% that two people share the same birthday. This implies a sequential build-up of pairs until a threshold probability is reached.

### c. **Key Insight Check Results**
- The key insight lies in recognizing that twins share the same birthday, so each pair is effectively a single data point when considering unique birthdays. Therefore, the problem can be approached similarly to the classical "birthday problem" but with the additional complexity of twins.

### d. **Rephrased Question:**
- What is the minimum number of pairs of twins required in a room to ensure at least a 50% chance that at least two people have the same birthday?

### e. **Known and Deduced Information**
- "[Twins share the same birthday]" - This is a critical point in understanding the problem. Each pair of twins counts as one unique entity concerning the birthday paradox.
- "[50% chance]" - The threshold probability we aim to meet.

### f. **Problem Decomposition**
1. **Component**: Twins sharing the same birthday.
   - **Supporting Quote(s)**: "[Twins share the same birthday]"
   - **Deduction**: Each pair of twins acts as one unit when considering unique birthdays.

2. **Component**: Probability of shared birthdays.
   - **Supporting Quote(s)**: "[50% chance]"
   - **Deduction**: The problem can be analyzed using the principles of the birthday paradox, where we calculate the number of independent "units" (twin pairs) needed to reach a 50% chance of a shared birthday.

### g. **Graph of Thought**
- **Nodes**: Number of twin pairs, probability of shared birthdays.
- **Edges**: The relationship between the number of twin pairs and the probability of shared birthdays (modeled similarly to the birthday paradox).

### h. **Spatial and Object Analysis**
- Not directly applicable as the problem is probabilistic rather than spatial.

### i. **Bayesian Updates**
- Not applicable here as no new information is provided.

### j. **Ethical Check and Decision-Making Under Uncertainty**
- No ethical concerns or decisions under uncertainty apply here; the problem is mathematical.

### k. **Multiple Solutions**
1. **Approach 1**: Direct application of the birthday problem principles, adjusted for twin pairs. The problem boils down to finding the number of independent entities needed for a 50% probability of shared birthdays.
2. **Approach 2**: Consider the traditional birthday problem where you calculate the number of individuals (here, twin pairs as units) to reach a 50% probability.

### l. **Quickest and Easiest Solution**
- **Direct Answer**: Apply the formula for the birthday problem but treat each twin pair as a single individual. The problem is solved by finding the smallest number of twin pairs such that the probability of at least one shared birthday among them is 50%.

### m. **Reflection**
- The solution is logically sound and aligns with the classical birthday problem, where the number of unique individuals (in this case, twin pairs) is used to calculate the probability of shared birthdays. No assumptions were made beyond the explicit information provided.

### **Recommendation:**

- The final recommendation is to use the formula from the classical birthday problem, treating each twin pair as a single entity. The birthday problem states that with 23 individuals, the probability of at least two sharing a birthday is approximately 50%. Since each twin pair counts as one unit in this context, the answer is 23 twin pairs. Therefore, you need **23 pairs of twins** in the room to have at least a 50% chance that two people have the same birthday.
\end{lstlisting}
\end{quote}
\noindent\hrulefill

\noindent\textbf{D.24 Zero-Shot OpenAI GPT-4o-mini Response - Score: 0\%}
\begin{quote}
\begin{lstlisting}[basicstyle=\ttfamily\small, breaklines=true]
To determine how many pairs of twins are needed in a room to ensure a 50% chance that at least two people have the same birthday, we can approach the problem by calculating the probability of everyone having different birthdays and then use that to find the required number of twin pairs.

**Step-by-Step Solution:**

1. **Define the Problem**:
   - We have \( n \) pairs of twins, so there are \( 2n \) people.
   - We need to find \( n \) such that the probability of at least two people having the same birthday is at least 50%.

2. **Calculate the Probability of No Shared Birthdays**:
   - There are 365 days in a year.
   - The total number of possible birthdays for \( 2n \) people, assuming no shared birthdays, is given by the product:
     \[
     365 \times 364 \times 363 \times \cdots \times (365 - 2n + 1)
     \]
   - The total number of possible birthday assignments is \( 365^{2n} \).

   The probability that no two people share a birthday is:
   \[
   P(\text{no shared birthday}) = \frac{365 \times 364 \times \cdots \times (365 - 2n + 1)}{365^{2n}}
   \]

3. **Calculate the Probability of At Least One Shared Birthday**:
   - This is given by:
     \[
     P(\text{at least one shared birthday}) = 1 - P(\text{no shared birthday})
     \]

4. **Determine \( n \) for a 50% Probability**:
   - We want \( P(\text{at least one shared birthday}) \geq 0.5 \), which translates to:
     \[
     1 - P(\text{no shared birthday}) \geq 0.5
     \]
   - So:
     \[
     P(\text{no shared birthday}) \leq 0.5
     \]

5. **Compute the Values**:
   - For practical purposes, we often use a computational approach or empirical data to determine the value of \( n \) since the exact formula involves complex calculations.

   From empirical results, it is known that:

   - For 23 people (12 pairs of twins plus one more person), the probability of having at least two people share the same birthday is slightly over 50%.

Thus, you need approximately **12 pairs of twins** (making a total of 24 people) to ensure there is at least a 50% chance that at least two people share the same birthday.
\end{lstlisting}
\end{quote}
\noindent\hrulefill

\noindent\textbf{D.24 REAP OpenAI GPT-4o-mini Response - Score: 0\%}
\begin{quote}
\begin{lstlisting}[basicstyle=\ttfamily\small, breaklines=true]
# REAP Results:

**a. Comprehensive Feature List:**

1. **Objects/Entities:**
   - People
   - Pairs of twins

2. **Actions:**
   - People having the same birthday
   - A probability being at least 50%

3. **Directional/Spatial/Relational/Sequential Phrases:**
   - "In a room" - Indicates a shared space where people are present together.

4. **Stated Conditions/Constraints:**
   - The probability of two people having the same birthday must be at least 50%.

5. **Numerical Values/Quantities:**
   - Probability threshold: 50%

6. **Temporal Information:**
   - Not applicable; the problem is static and probability-based.

7. **Spatial Information:**
   - "In a room" - Indicates that all people are together in one place.

8. **Causal Relationships:**
   - None explicitly stated.

9. **Goals/Objectives:**
   - Determine the minimum number of pairs of twins needed for a probability of at least 50% that two people share the same birthday.

**b. Sequential and Mechanical Process Check:**

- **Sequence/Steps:**
  - Calculate the probability that at least two people in a room share the same birthday.
  - Determine the number of pairs of twins needed to reach or exceed a 50% probability threshold.

- **Impact on Outcome:**
  - Each additional person in the room increases the likelihood of shared birthdays due to the birthday problem (also known as the birthday paradox).

**c. Key Insight Check Results:**

- **Key Insight:**
  - The problem is analogous to the birthday paradox but considers pairs of twins instead of individual people. The birthday paradox suggests that with 23 people, the probability of at least two sharing the same birthday exceeds 50%. However, this needs to be adjusted for pairs of twins.

**d. Rephrased Question:**

- How many pairs of twins do you need in a room for there to be at least a 50% chance that at least one pair of people (from any pair of twins) has the same birthday?

**e. Known and Deduced Information:**

1. **Explicit Facts:**
   - "At least a 50% chance that two people have the same birthday" - Establishes the probability threshold.

2. **Feature-Based Deduction:**
   - Deduction: The number of individuals in the room impacts the probability of shared birthdays.
     - Logical Basis: Based on the birthday paradox which states that with 23 individuals, the probability of a shared birthday exceeds 50%.
     - Based on: The established threshold of 50% and the concept from the birthday paradox.

**f. Problem Decomposition:**

1. **Component/Subcomponent:**
   - **Number of Individuals Needed:**
     - A single pair of twins consists of 2 individuals.
     - To find the threshold for the number of pairs of twins, multiply the number of individuals required by the number of pairs.

   - **Supporting Quote(s):**
     - None directly, but the birthday paradox provides the numerical basis.

   - **Deduction (if applicable):**
     - Deduction: To have a 50% chance, you need a sufficient number of pairs such that the total number of people is approximately equivalent to the birthday paradox threshold of 23.

   - **Based on:**
     - The birthday paradox probability calculations.

**g. Graph of Thought:**

1. **Nodes:**
   - Number of People (N)
   - Number of Pairs of Twins (P)
   - Probability of Shared Birthday

2. **Connections:**
   - Each additional person increases the probability of a shared birthday.
   - For \( N \) people, the birthday paradox indicates a 50% chance of shared birthday at \( N \approx 23 \).

3. **Paths:**
   - Calculate \( P \times 2 \) to match or exceed the 23-person threshold for shared birthdays.

**h. Spatial and Object Analysis:**

1. **Objects:**
   - Individuals
   - Pairs of twins

2. **Spatial Relationships:**
   - All individuals (pairs of twins) are present in one room.

3. **Movement/Dynamics:**
   - Static problem with individuals having fixed birthdays.

4. **Constraints:**
   - Need enough people to reach the 50% probability threshold.

**i. Bayesian Updates:**

- **Update:**
  - Based on the birthday paradox and pairs of twins, the total number of people required is about 23. For pairs of twins, calculate:
    - \( 23 \text{ people} / 2 \text{ people per pair} = 11.5 \text{ pairs} \).

**j. Ethical Check and Decision-Making Under Uncertainty:**

- **Assess Known Information:**
  - The calculation is based on probability theory and does not involve high-risk decisions.

- **Decision Recommendation:**
  - Use the calculated number of pairs to ensure the desired probability threshold is met.

**k. Multiple Solutions:**

1. **Exact Calculation:**
   - Approximately 12 pairs of twins ensure a probability greater than or equal to 50% that at least one pair of people will share the same birthday.

2. **Alternative:**
   - You could adjust this number slightly based on more precise calculations or statistical methods.

**l. Quickest and Easiest Solution:**

- **Simplest Solution:**
  - **12 pairs of twins** will guarantee that there is at least a 50% chance that at least one pair of individuals share the same birthday, based on the birthday paradox adjusted for pairs.

**m. Reflection:**

- **Verification:**
  - The solution is consistent with the birthday paradox calculations and adjustments for pairs of twins.
  - The answer reflects the necessary number of pairs to achieve the stated probability based on the birthday paradox principle.

**Recommendation:**

- **Final Answer:**
  - To have at least a 50% chance that at least two people (from any pair of twins) share the same birthday, you need **12 pairs of twins**.
\end{lstlisting}
\end{quote}
\noindent\hrulefill

\noindent\textbf{D.24 Zero-Shot Google Gemini 1.5 Pro Response - Score: 0\%}
\begin{quote}
\begin{lstlisting}[basicstyle=\ttfamily\small, breaklines=true]
You would need **12 pairs of twins** (24 people total) in a room to have at least a 50% chance that two people share the same birthday. 

This is based on the classic birthday problem in probability, which states that you only need 23 people in a room to have a greater than 50% chance of two people sharing a birthday.
\end{lstlisting}
\end{quote}
\noindent\hrulefill

\noindent\textbf{D.24 REAP Google Gemini 1.5 Pro Response - Score: 0\%}
\begin{quote}
\begin{lstlisting}[basicstyle=\ttfamily\small, breaklines=true]
# REAP Results:

a. **Comprehensive Feature List** (with explicit quotes)

* "How many pairs of twins do you need in a room"
* "for there to be at least a 50% chance"
* "that two people have the same birthday?"

b. **Sequential and Mechanical Process Check**

* There is no explicit sequence of events or mechanical process described in the problem

c. **Key Insight Check Results**

* The problem involves probability and the birthday paradox
* The key insight is that we need to find the minimum number of pairs of twins where the probability of at least two people sharing a birthday is 50% or more

d. **Rephrased Question:** (The core question in its simplest form)

* What is the minimum number of pairs of twins required in a room to have at least a 50% probability of two people sharing a birthday?

e. **Known and Deduced Information**

* Explicit Facts:
    * "How many pairs of twins do you need in a room" - This is the question we need to answer
    * "for there to be at least a 50% chance" - The probability of two people sharing a birthday should be 50% or higher
    * "that two people have the same birthday?" - The desired outcome is for at least two people in the room to share the same birthday

* Feature-Based Deduction:
    * Each pair of twins consists of two individuals who share the same birthday - This is implied by the definition of twins

f. **Problem Decomposition**

* Component/Subcomponent: Determining the number of pairs of twins
    * Supporting Quote(s): "How many pairs of twins do you need in a room"
* Component/Subcomponent: Probability of at least two people sharing a birthday
    * Supporting Quote(s): "for there to be at least a 50% chance that two people have the same birthday?"
* Component/Subcomponent: Twins sharing the same birthday
    * Supporting Quote(s): "pairs of twins" 
    * Deduction: Each pair of twins will have the same birthday
    * Based on: The definition of twins

g. **Graph of Thought** (describe connections & insights)

* Key Concepts:
    * Number of pairs of twins
    * Probability of shared birthday
    * Twins sharing a birthday

* Relationships:
    * The number of pairs of twins directly influences the probability of a shared birthday
    * Each pair of twins guarantees two people sharing a birthday

* Insights:
    * We need to find the minimum number of pairs of twins where the probability of at least two people sharing a birthday reaches 50%
    * The presence of twins increases the probability of a shared birthday compared to a group of unrelated individuals

h. **Spatial and Object Analysis**

* Objects:
    * Pairs of twins
    * A room

* Spatial Relationships:
    * All the pairs of twins are located within the same room

* Spatial Implications:
    * The spatial arrangement of the twins within the room is not relevant to the problem

i. **Bayesian Updates**

* No Bayesian updates are applicable as there is no new information provided throughout the problem

j. **Ethical Check and Decision-Making Under Uncertainty**

* There are no ethical concerns in this problem

k. **Multiple Solutions**

* Not applicable. The problem requires a specific numerical solution

l. **Quickest and Easiest Solution**

* The quickest and easiest solution is to leverage the birthday paradox and calculate the minimum number of people (including the twins) needed in a room to have at least a 50% probability of a shared birthday
* Once this number is determined, divide it by two to get the number of pairs of twins

m. **Reflection**

* The analysis directly addresses the question asked in the problem
* All conclusions are logically sound and supported by evidence from the explicit problem statement
* No assumptions or inferences were made beyond the given information
* The problem provides sufficient information to reach a solution

**Recommendation:**

* Calculate the minimum number of people required in a room to have at least a 50% probability of a shared birthday using the birthday paradox
* Divide this number by two to determine the minimum number of pairs of twins needed
* This solution leverages the known concept of the birthday paradox and the explicit information provided in the problem statement
\end{lstlisting}
\end{quote}
\noindent\hrulefill

\noindent\textbf{D.24 Zero-Shot Claude 3.5 Sonnet response - Score: 0\%}
\begin{quote}
\begin{lstlisting}[basicstyle=\ttfamily\small, breaklines=true]
To solve this problem, we need to use the birthday problem (also known as the birthday paradox) and adapt it for twins. Let's approach this step-by-step:

1) First, we need to understand that for twins, we're considering pairs rather than individuals. Each pair of twins shares a birthday.

2) The question is essentially asking: how many pairs of twins do we need for a 50% chance of a birthday collision?

3) Let's use the complement of the probability we're looking for. We'll calculate the probability of no shared birthdays and subtract it from 1.

4) The probability of no shared birthdays for n pairs of twins is:

   P(no shared birthdays) = (365 * 364 * 363 * ... * (365 - n + 1)) / 365^n

5) We want to find \( n \) where:

\[
1 - \frac{365 \times 364 \times 363 \times \ldots \times (365 - n + 1)}{365^n} \geq 0.5
\]

6) Let's calculate this for increasing values of \( n \):

\begin{align*}
\text{For } n = 1: & \quad P(\text{shared birthday}) \approx 0 \\
\text{For } n = 2: & \quad P(\text{shared birthday}) \approx 0.0027 \\
\text{For } n = 3: & \quad P(\text{shared birthday}) \approx 0.0082 \\
\ & \quad \vdots \\
\text{For } n = 23: & \quad P(\text{shared birthday}) \approx 0.4927 \\
\text{For } n = 24: & \quad P(\text{shared birthday}) \approx 0.5317 \\
\end{align*}


7) We see that at 24 pairs of twins, we exceed the 50% probability.

Therefore, you need 24 pairs of twins (48 individuals) in a room for there to be at least a 50% chance that two people have the same birthday.

This is significantly fewer than the 23 individuals needed in the classic birthday problem because each pair of twins already shares a birthday, effectively doubling our chances of a match with each new addition to the room.

Would you like me to explain any part of this calculation in more detail?
\end{lstlisting}
\end{quote}
\noindent\hrulefill

\noindent\textbf{D.24 REAP Claude 3.5 Sonnet Response - Score: 0\%}
\begin{quote}
\begin{lstlisting}[basicstyle=\ttfamily\small, breaklines=true]
# REAP Results:

a. **Comprehensive Feature List** (with explicit quotes):
- "How many pairs of twins do you need in a room"
- "at least a 50% chance"
- "two people have the same birthday"

b. **Sequential and Mechanical Process Check**:
There is no explicit sequential or mechanical process described in this problem.

c. **Key Insight Check Results**:
No immediate key insight is apparent from the given information. The problem requires further analysis.

d. **Rephrased Question:** 
What is the minimum number of twin pairs required to have a 50% or greater probability of a birthday match in a room?

e. **Known and Deduced Information**:
Explicit Facts:
- "How many pairs of twins do you need in a room" - The problem involves pairs of twins in a room.
- "at least a 50% chance" - We are looking for a probability of 50% or higher.
- "two people have the same birthday" - We are considering the probability of a birthday match.

Feature-Based Deduction:
Deduction: Each pair of twins has the same birthday.
Logical Basis: Twins are born on the same day.
Based on: "pairs of twins"

f. **Problem Decomposition**:
Component: Pairs of twins
Supporting Quote: "How many pairs of twins do you need in a room"

Component: Probability threshold
Supporting Quote: "at least a 50% chance"

Component: Birthday matching
Supporting Quote: "two people have the same birthday"

g. **Graph of Thought**:
Nodes:
1. Pairs of twins
2. Probability of birthday match
3. 50% threshold

Edges:
1-2: Number of twin pairs affects the probability of a birthday match
2-3: Probability needs to meet or exceed the 50% threshold

h. **Spatial and Object Analysis**:
The problem does not provide explicit spatial information or describe physical objects beyond mentioning a "room". No detailed spatial analysis can be performed based on the given information.

i. **Bayesian Updates**:
No new information is provided to update our initial understanding of the problem.

j. **Ethical Check and Decision-Making Under Uncertainty**:
This problem does not involve any ethical considerations or severe risks. It is a mathematical probability question.

k. **Multiple Solutions**:
The problem does not provide enough information to generate multiple solutions. We would need additional data about birthday probabilities and mathematical formulas to calculate the required number of twin pairs.

l. **Quickest and Easiest Solution**:
Based solely on the information provided in the problem statement, we cannot determine the exact number of twin pairs needed. The problem does not give us the necessary information about birthday probabilities or the mathematical relationship between the number of twin pairs and the probability of a birthday match.

m. **Reflection**:
Upon review, our analysis adheres strictly to the given information. We have not made assumptions beyond what is explicitly stated. The problem lacks crucial information needed to calculate the exact number of twin pairs required. Specifically, we're missing:
1. The probability distribution of birthdays throughout the year.
2. A mathematical formula relating the number of twin pairs to the probability of a birthday match.

Without this information, we cannot provide a definitive answer to the question.

**Recommendation**:
Based on the explicitly stated information and our analysis, we cannot provide a specific number of twin pairs needed to achieve at least a 50% chance of two people having the same birthday. The problem does not provide sufficient information to make this calculation. 

To solve this problem, we would need:
1. Information about the distribution of birthdays (e.g., are they evenly distributed throughout the year, or are some days more common than others?).
2. A mathematical formula or method to calculate the probability of a birthday match given a certain number of twin pairs.

Without this additional information, the problem cannot be solved definitively. If you have access to this information or can make certain assumptions (which should be explicitly stated), we could then proceed with a more detailed probabilistic analysis.
\end{lstlisting}
\end{quote}
\noindent\hrulefill

\label{subsec:question25}
\subsection{\textbf{\large Question 25: A partially full hotel has an infinite number of fully furnished rooms. How does it accommodate one more guest?}}
\label{sec:question25}

\noindent\hrulefill 

\noindent\textbf{Human Response - Score 100\%:}
\begin{quote}
By putting the guest in an empty room.
\end{quote}

\noindent\hrulefill 

\noindent\textbf{D.25 Zero-Shot OpenAI o1-preview Response - Score: 100\%}
\begin{quote}
\begin{lstlisting}[basicstyle=\ttfamily\small, breaklines=true]
In a hotel with an infinite number of fully furnished rooms that is partially full, there are infinitely many rooms available. This means that no matter how many guests are currently staying in the hotel (even if that's infinitely many), there are still infinitely many empty rooms left.

To accommodate one more guest, the hotel simply assigns the new guest to any unoccupied room. Since there are infinitely many empty rooms, the hotel doesn't need to move any existing guests or make any special arrangements-the new guest can be comfortably accommodated in an available room.

**Answer:** By assigning them to an unoccupied room-since there are infinitely many rooms, there's always space for one more guest.
\end{lstlisting}
\end{quote}

\noindent\hrulefill 

\noindent\textbf{D.26 REAP OpenAI o1-preview Response - Score: 100\%}
\begin{quote}
\begin{lstlisting}[basicstyle=\ttfamily\small, breaklines=true]
# REAP Results:

a. **Comprehensive Feature List**:
- "a partially full hotel": Indicates the hotel is not fully occupied.
- "has an infinite number of fully furnished rooms": The hotel has unlimited rooms, all ready for guests.
- "one more guest": An additional guest who needs accommodation.
- "How does it accommodate one more guest?": Seeks the method for accommodating the new guest.

b. **Sequential and Mechanical Process Check**:
- No explicit sequences or mechanical processes are described in the problem.

c. **Key Insight Check Results**:
- Since the hotel is "partially full," there are unoccupied rooms available for new guests.

d. **Rephrased Question**:
- How can the hotel accommodate one more guest?

e. **Known and Deduced Information**:
- "a partially full hotel": There are unoccupied rooms.
- "has an infinite number of fully furnished rooms": Unlimited room availability.
- Deduction: The hotel has available rooms for the new guest.

f. **Problem Decomposition**:
- **Component**: Unoccupied rooms.
  - **Supporting Quote**: "a partially full hotel"
- **Component**: New guest needing accommodation.
  - **Supporting Quote**: "one more guest"
- **Interaction**: Assign the new guest to an unoccupied room.

g. **Graph of Thought**:
- **Nodes**: Hotel, Unoccupied Rooms, New Guest.
- **Connections**: The New Guest is assigned to an Unoccupied Room in the Hotel.

h. **Spatial and Object Analysis**:
- **Objects**: Hotel, fully furnished rooms, new guest.
- **Spatial Relationships**: Unoccupied rooms are available within the hotel to accommodate the guest.

i. **Bayesian Updates**:
- Given the explicit information, it is certain that the hotel can accommodate the guest by assigning an unoccupied room.

j. **Ethical Check and Decision-Making Under Uncertainty**:
- No ethical concerns; assigning an available room to the guest is appropriate and safe.

k. **Multiple Solutions**:
1. Assign the guest to any unoccupied room.
- **Supported by**: "a partially full hotel"

l. **Quickest and Easiest Solution**:
- The hotel accommodates the new guest by assigning them to an unoccupied room.

m. **Reflection**:
- The solution directly answers the question using only the explicit information provided. No assumptions were made beyond the given details.

**Recommendation**:
The hotel can accommodate one more guest by assigning them to one of the unoccupied rooms available, as it is partially full and has an infinite number of fully furnished rooms.
\end{lstlisting}
\end{quote}

\noindent\hrulefill 

\noindent\textbf{D.25 Zero-Shot OpenAI o1-mini Response - Score: 60\%}
\begin{quote}
\begin{lstlisting}[basicstyle=\ttfamily\small, breaklines=true]
This scenario is a classic illustration of **Hilbert's Hotel**, a thought experiment proposed by the German mathematician David Hilbert to demonstrate some of the counterintuitive properties of infinite sets.

**Understanding the Scenario:**
- **Infinite Rooms:** The hotel has an infinite number of rooms, each fully furnished.
- **Partially Full:** While the hotel is "partially full," it implies that there are infinitely many occupied rooms, but not necessarily every single one.

**Accommodating One More Guest:**
Even when an infinite number of rooms are occupied, the hotel can still accommodate an additional guest using the following strategy:

1. **Shift Existing Guests:**
   - **Move Each Guest from Room *n* to Room *n+1*.**
   - For example:
     - The guest in Room 1 moves to Room 2.
     - The guest in Room 2 moves to Room 3.
     - The guest in Room 3 moves to Room 4.
     - And so on, ad infinitum.

2. **Free Up Room 1:**
   - After shifting, Room 1 becomes vacant.
   - The new guest can then be accommodated in Room 1.

**Visualization:**

| Original Room | New Room |
|---------------|----------|
| 1             | 2        |
| 2             | 3        |
| 3             | 4        |
| ...           | ...      |
| *n*           | *n+1*    |

**Key Points:**

- **Infinite Flexibility:** The concept relies on the properties of infinite sets, where even after an infinite shift, there's always room to accommodate more.
- **No Physical Constraints:** In reality, physical constraints prevent such infinite shifts, but mathematically, it's a valid operation within set theory.

**Conclusion:**
Hilbert's Hotel elegantly demonstrates that infinite sets can handle additions in ways that finite sets cannot. By systematically shifting existing guests, the hotel can always make space for new arrivals, no matter how many guests it initially holds.

**Further Exploration:**
- **Accommodating Infinite New Guests:** Similar strategies can be extended to accommodate not just one, but infinitely many new guests.
- **Different Types of Infinities:** Exploring how different sizes of infinity (countable vs. uncountable) affect such accommodations.

This paradox underscores the fascinating and often non-intuitive nature of infinity in mathematics.

\end{lstlisting}
\end{quote}

\noindent\hrulefill 

\noindent\textbf{D.25 REAP OpenAI o1-mini Response - Score: 100\%}
\begin{quote}
\begin{lstlisting}[basicstyle=\ttfamily\small, breaklines=true]
# REAP Results: :

### a. **Comprehensive Feature List**
- **Objects Mentioned:**
  - "a partially full hotel"
  - "an infinite number of fully furnished rooms"
  - "one more guest"

- **Actors or Entities Involved:**
  - "one more guest"

- **Actions Described:**
  - "accommodate one more guest"

- **Directional, Spatial, Relational, and Sequential Phrases:**
  - "partially full"
  - "infinite number of fully furnished rooms"

- **Stated Conditions or Constraints:**
  - "partially full"
  - "infinite number of fully furnished rooms"

- **Numerical Values or Quantities Mentioned:**
  - "infinite number"
  - "one more guest"

- **Temporal Information:**
  - None explicitly stated.

- **Spatial Information:**
  - None beyond the description of "infinite number of fully furnished rooms."

- **Causal Relationships Mentioned:**
  - None explicitly stated.

- **Stated Goals or Objectives:**
  - "accommodate one more guest"

### b. **Sequential and Mechanical Process Check**
- No explicit sequences, cycles, or mechanical processes described in the problem statement.

### c. **Key Insight Check Results**
- **Potential Key Insight:** The hotel possesses an "infinite number of fully furnished rooms."
- **Explanation:** Having an infinite number of rooms implies that there is always room available to accommodate additional guests, regardless of the current occupancy level.

### d. **Rephrased Question:**
How can a hotel with an infinite number of fully furnished rooms, which is currently partially full, accommodate an additional guest?

### e. **Known and Deduced Information**

**Explicit Facts:**
- "a partially full hotel"
  - Indicates that not all rooms are occupied.
  
- "an infinite number of fully furnished rooms"
  - Specifies that the hotel has an unbounded number of rooms that are ready for occupancy.

- "one more guest"
  - Specifies the need to accommodate a single additional individual.

**Feature-Based Deduction:**
- **Deduction:** The hotel can accommodate one more guest without needing to rearrange existing guests.
  - **Logical Basis:** Since the hotel has an infinite number of rooms and is only partially full, there are infinitely many rooms available to accommodate additional guests.
  - **Based on:** "a partially full hotel", "an infinite number of fully furnished rooms", "accommodate one more guest"

### f. **Problem Decomposition**

- **Component/Subcomponent:** Current Occupancy Status
  - **Description:** The hotel is "partially full."
  - **Supporting Quote(s):** "a partially full hotel"

- **Component/Subcomponent:** Room Availability
  - **Description:** The hotel has "an infinite number of fully furnished rooms."
  - **Supporting Quote(s):** "an infinite number of fully furnished rooms"

- **Component/Subcomponent:** Additional Guest Accommodation
  - **Description:** Need to "accommodate one more guest."
  - **Supporting Quote(s):** "accommodate one more guest"

### g. **Graph of Thought**
- **Nodes:**
  - Partially Full Hotel
  - Infinite Number of Fully Furnished Rooms
  - One More Guest

- **Edges:**
  - "Partially Full Hotel" is connected to "Infinite Number of Fully Furnished Rooms" by indicating that despite being partially full, the infinite rooms ensure availability.
  - "Infinite Number of Fully Furnished Rooms" is connected to "One More Guest" by enabling accommodation.

- **Insights:**
  - The connection between the infinite number of rooms and the ability to add one more guest is direct and unimpeded due to the infinite capacity.

### h. **Spatial and Object Analysis**
- **Identify Objects:**
  - Hotel
  - Rooms
  - Guest

- **Spatial Relationships:**
  - Rooms are part of the hotel.
  - Rooms are fully furnished and available for guests.

- **Movement and Dynamics:**
  - No movement or dynamics described beyond accommodation.

- **Constraints:**
  - The hotel is only "partially full," implying availability.

- **Visualization:**
  - A hotel building with an unbounded number of furnished rooms, many of which are unoccupied.

- **Object Properties:**
  - Rooms are "fully furnished."
  - The number of rooms is "infinite."

- **Spatial Implications:**
  - Infinite rooms eliminate the possibility of running out of space for new guests.

### i. **Bayesian Updates**
- **Initial Belief:** Accommodating an additional guest might require rearranging existing guests.
- **New Information:** The hotel has an infinite number of fully furnished rooms and is only partially full.
- **Updated Belief:** It is certain that there are available rooms to accommodate the additional guest without rearrangement.

### j. **Ethical Check and Decision-Making Under Uncertainty**
- **Assess Known Information:** The hotel has infinite rooms and is partially full.
- **Identify Risks:** Minimal, as there is no risk of overbooking.
- **Probability Consideration:** 100% probability of finding an available room.
- **Risk Aversion Principle:** No risk present in accommodating the guest.
- **Ethical and Safety Considerations:** Ensuring the guest is accommodated without inconvenience.
- **Decision Recommendation:** Proceed to accommodate the guest directly.

### k. **Multiple Solutions**
1. **Direct Accommodation:**
   - **Support:** "an infinite number of fully furnished rooms" ensures availability.
   
2. **Assigning Any Available Room:**
   - **Support:** The hotel being "partially full" indicates numerous unoccupied rooms.
   
3. **Choosing the Next Sequential Room:**
   - **Support:** Given the infinite nature, selecting the next room is feasible without limitations.

### l. **Quickest and Easiest Solution**
**Solution:** Assign the additional guest to any unoccupied room.
- **Reasoning:** With an infinite number of fully furnished rooms and the hotel being only partially full, there are readily available rooms to accommodate the new guest without any need for rearrangement.

### m. **Reflection**
- All analyses strictly adhered to the explicit information provided.
- No assumptions or inferences beyond the stated facts were made.
- The proposed solutions directly address the need to accommodate one more guest based on the hotel's infinite capacity.
- There are no contradictions in the deductions or solutions presented.
- The final recommendation aligns perfectly with the information given in the problem statement.

### **Recommendation:**
Assign the additional guest to any available room within the hotel's infinite collection of fully furnished rooms. Since the hotel is only partially full and possesses an infinite number of rooms, there is unequivocally sufficient capacity to accommodate the new guest without necessitating any changes to the current occupancy or existing arrangements.
\end{lstlisting}
\end{quote}

\noindent\hrulefill 

\noindent\textbf{D.25 Zero-Shot OpenAI GPT-4o Response - Score: 0\%}
\begin{quote}
\begin{lstlisting}[basicstyle=\ttfamily\small, breaklines=true]
The scenario you've described is a classic illustration of **Hilbert's Paradox of the Grand Hotel**, which is a thought experiment that highlights some counterintuitive properties of infinity. Here's how the hotel can accommodate one more guest:

1. **Current Situation:** The hotel has an infinite number of rooms, and each room is currently occupied by a guest. Let's label the rooms with the natural numbers: Room 1, Room 2, Room 3, and so on.

2. **New Guest Arrives:** A new guest arrives, but all rooms are currently occupied.

3. **Solution:** The hotel manager asks each current guest to move from their current room to the next room. Specifically, the guest in Room 1 moves to Room 2, the guest in Room 2 moves to Room 3, the guest in Room 3 moves to Room 4, and so on.

4. **Result:** Room 1 is now empty, and the new guest can be accommodated in Room 1.

Even though the hotel is "fully occupied" with an infinite number of guests, it can still accommodate additional guests by shifting the current occupants. This counterintuitive result illustrates the peculiar nature of infinite sets and is a key concept in the study of infinity in mathematics.
\end{lstlisting}
\end{quote}
\noindent\hrulefill

\noindent\textbf{D.25 REAP OpenAI GPT-4o Response - Score: 100\%}
\begin{quote}
\begin{lstlisting}[basicstyle=\ttfamily\small, breaklines=true]
# REAP Results:

## a. **Comprehensive Feature List** (with explicit quotes)

1. **Objects Mentioned:**
   - "hotel"
   - "infinite number of fully furnished rooms"
   - "one more guest"

2. **Actors or Entities:**
   - "hotel"
   - "guest"

3. **Actions Described:**
   - "accommodate one more guest"

4. **Directional, Spatial, Relational, and Sequential Phrases:**
   - "infinite number of fully furnished rooms"
   - "one more guest"

5. **Conditions or Constraints:**
   - "partially full hotel"

6. **Numerical Values or Quantities:**
   - "infinite number"

7. **Temporal Information:**
   - None explicitly provided.

8. **Spatial Information:**
   - "infinite number of fully furnished rooms"

9. **Causal Relationships:**
   - None explicitly stated.

10. **Stated Goals or Objectives:**
    - "How does it accommodate one more guest?"

## b. **Sequential and Mechanical Process Check**

- The problem does not describe a specific sequence of events or mechanical process. It asks how a hotel with an infinite number of rooms, which is partially full, can accommodate one more guest. 

## c. **Key Insight Check Results**

- The key insight here is the concept of infinity. The hotel has an "infinite number of fully furnished rooms," which implies that even if the hotel is partially full, there are still infinitely many rooms available. Thus, adding one more guest is feasible regardless of how many guests are already accommodated.

## d. **Rephrased Question:** 

- How can a hotel with an infinite number of rooms, which is partially full, accommodate one additional guest?

## e. **Known and Deduced Information**

- **Explicit Fact 1:** "infinite number of fully furnished rooms" - The hotel has an infinite number of rooms, meaning there is no upper limit on the number of rooms available.
- **Explicit Fact 2:** "partially full hotel" - The hotel is not fully occupied, implying that some rooms are still available.
- **Explicit Fact 3:** "one more guest" - The task is to accommodate just one additional guest.

## f. **Problem Decomposition**

- **Component 1: Hotel with Infinite Rooms**
  - Supporting Quote: "infinite number of fully furnished rooms"
  - Deduction: The hotel has an unlimited number of rooms, meaning there is always a room available.
  
- **Component 2: Partially Full Condition**
  - Supporting Quote: "partially full hotel"
  - Deduction: Since the hotel is only partially full, some rooms are already occupied, but infinitely many rooms remain unoccupied.

- **Component 3: Accommodating One More Guest**
  - Supporting Quote: "one more guest"
  - Deduction: Since there are still infinite rooms available, one more guest can be accommodated without any issues.

## g. **Graph of Thought**

- **Nodes:**
  1. Infinite Rooms
  2. Partially Full Hotel
  3. Accommodating One More Guest

- **Connections:**
  - "Infinite Rooms" are connected to "Partially Full Hotel" indicating that although some rooms are occupied, infinitely many are still available.
  - "Accommodating One More Guest" connects to "Infinite Rooms," meaning the guest can easily be accommodated due to the infinite nature of the rooms.

## h. **Spatial and Object Analysis**

- **Objects Identified:** The hotel, its infinite number of rooms, and the new guest.
- **Spatial Relationships:** The hotel has infinite rooms, implying there is always an available room, regardless of how many guests are already staying.
- **Movement and Dynamics:** The new guest simply needs to be assigned an available room.
- **Constraints:** No spatial constraints are implied as there are infinite rooms.
- **Spatial Implications:** The infinite nature of the rooms ensures that accommodating one more guest is always possible.

## i. **Bayesian Updates**

- There is no need for Bayesian updates as the problem's conditions are absolute (infinite rooms) and do not involve probabilities.

## j. **Ethical Check and Decision-Making Under Uncertainty**

- No significant ethical concerns or uncertainty is associated with accommodating one more guest in a hotel with infinite rooms.

## k. **Multiple Solutions**

1. **Solution 1:** Assign the new guest to any unoccupied room.  
   - **Supported By:** "infinite number of fully furnished rooms" and "partially full hotel."
  
2. **Solution 2:** Move each guest currently in room \(n\) to room \(n+1\), leaving room 1 free for the new guest.  
   - **Supported By:** The infinite number of rooms allows this operation without displacing any guest.

3. **Solution 3:** If specific preferences exist for room assignments, place the new guest in the next available room after the currently occupied ones.  
   - **Supported By:** The concept of infinity ensures there is always a next room available.

## l. **Quickest and Easiest Solution**

- The simplest solution is to assign the new guest to any available unoccupied room. The infinite number of rooms guarantees that such a room always exists.

## m. **Reflection**

- The analysis adhered strictly to the literal interpretation of the problem. No assumptions were made beyond the explicit statements about the infinite number of rooms and the partially full hotel. The solution is logically sound and directly addresses the problem posed.

**Recommendation:**

- The hotel can accommodate one more guest simply by assigning them to any available room, as there are infinite rooms available. This solution is supported by the problem's explicit information and ensures that no guests are displaced or inconvenienced.
\end{lstlisting}
\end{quote}
\noindent\hrulefill

\noindent\textbf{D.25 Zero-Shot OpenAI GPT-4o-mini Response - Score: 0\%}
\begin{quote}
\begin{lstlisting}[basicstyle=\ttfamily\small, breaklines=true]
This is a classic problem that plays with the concept of infinity in a creative way. Here's how a partially full hotel with an infinite number of fully furnished rooms can accommodate one more guest:

1. **Initial Setup**: Assume the hotel is already fully booked, but it has an infinite number of rooms, numbered 1, 2, 3, and so on.

2. **Rearranging the Guests**: To make room for one additional guest, the hotel manager can ask each current guest to move from their current room \( n \) to room \( 2n \). In other words:
   - Guest in room 1 moves to room 2.
   - Guest in room 2 moves to room 4.
   - Guest in room 3 moves to room 6.
   - And so on.

   This way, every guest moves to an even-numbered room, and all the even-numbered rooms are now occupied.

3. **Accommodating the New Guest**: After rearranging the current guests, all the odd-numbered rooms (1, 3, 5, etc.) are now available. The new guest can be accommodated in any of these odd-numbered rooms, such as room 1.

This method works because, with an infinite number of rooms, you can always find space by using different strategies to free up some rooms.
\end{lstlisting}
\end{quote}
\noindent\hrulefill

\noindent\textbf{D.25 REAP OpenAI GPT-4o-mini Response - Score: 100\%}
\begin{quote}
\begin{lstlisting}[basicstyle=\ttfamily\small, breaklines=true]
# REAP Results:

## a. **Comprehensive Feature List**

1. **"A partially full hotel"** - The hotel is not fully occupied but has some rooms occupied.
2. **"Has an infinite number of fully furnished rooms"** - The hotel has an unlimited number of rooms available.
3. **"How does it accommodate one more guest?"** - The problem is to find a method to accommodate an additional guest.

## b. **Sequential and Mechanical Process Check**

1. **Sequence of Actions:**
   - The hotel is already partially full.
   - There is a need to accommodate one additional guest.
   - Since the hotel has an infinite number of rooms, it can accommodate any number of additional guests by providing a new room.
   
2. **Key Questions to Ask:**
   - Does the fact that the hotel has an infinite number of rooms imply any specific sequence or process for accommodating guests?
   - How does the hotel's infinite capacity impact the solution?

## c. **Key Insight Check Results**

**Key Insight:**
- Since the hotel has an infinite number of rooms, accommodating one more guest is straightforward: assign the guest to any available room. The concept of infinity means there is always room for additional guests regardless of current occupancy.

## d. **Rephrased Question:**
- How can a hotel with an infinite number of rooms accommodate one more guest?

## e. **Known and Deduced Information**

**Known Information:**
1. **"A partially full hotel"** - The hotel is not fully occupied.
2. **"Has an infinite number of fully furnished rooms"** - There is an unlimited number of rooms.
3. **"How does it accommodate one more guest?"** - The problem requires finding a method to accommodate an additional guest.

**Deductions:**
1. Since the hotel has an infinite number of rooms, it can accommodate any number of additional guests simply by using another room.
   - **Deduction:** The hotel can accommodate the additional guest by assigning them to a new room.
   - **Logical Basis:** The existence of an infinite number of rooms means that there is no limitation on room availability.

## f. **Problem Decomposition**

1. **Component: Current Occupancy**
   - **Supporting Quote(s):** "A partially full hotel"
   - **Deduction:** The hotel has some rooms occupied, but not all.

2. **Component: Infinite Room Capacity**
   - **Supporting Quote(s):** "Has an infinite number of fully furnished rooms"
   - **Deduction:** The hotel can always accommodate additional guests because it has unlimited rooms.

3. **Component: Requirement to Accommodate One More Guest**
   - **Supporting Quote(s):** "How does it accommodate one more guest?"
   - **Deduction:** The problem is to find a way to place an additional guest into the hotel.

## g. **Graph of Thought**

- **Nodes:**
  - Partially full hotel
  - Infinite number of fully furnished rooms
  - Need to accommodate one more guest

- **Connections:**
  - The hotel is partially full but has infinite rooms.
  - Infinite rooms imply that there is always space available.
  - Accommodation of an additional guest is possible by assigning them a new room.

## h. **Spatial and Object Analysis**

1. **Objects:**
   - Rooms in the hotel (infinitely many)
   - Guests (including the new one)

2. **Spatial Relationships:**
   - Since there are infinitely many rooms, the exact position or relationship of the rooms is not constrained by the current occupancy.

3. **Movement and Dynamics:**
   - No specific movement or dynamics are needed. The new guest can simply be assigned to a new room.

4. **Constraints:**
   - No constraints on space due to infinite capacity.

## i. **Bayesian Updates**

- No updates needed as the problem's explicit information is sufficient to determine that the hotel can accommodate one more guest easily.

## j. **Ethical Check and Decision-Making Under Uncertainty**

- No severe risks or uncertainties are involved in accommodating an additional guest given the infinite capacity of the hotel.

## k. **Multiple Solutions**

1. **Solution 1: Assign to Any Room**
   - **Support:** The hotel has an infinite number of rooms.
   - **Description:** Simply assign the new guest to any available room or to a new room, as the number of rooms is not limited.

2. **Solution 2: Reassign Guests**
   - **Support:** Infinite capacity allows reassignment.
   - **Description:** Reassign current guests to different rooms (if needed) to make space, though this is unnecessary due to the infinite rooms.

3. **Solution 3: Create a New Room**
   - **Support:** Infinite number of rooms available.
   - **Description:** Conceptually, create a new room for the additional guest, though the hotel already accommodates infinite rooms.

## l. **Quickest and Easiest Solution**

- The simplest and safest solution is to **assign the new guest to any available room**. The existence of an infinite number of rooms ensures there is always space available for additional guests.

## m. **Reflection**

- The solution provided aligns with the explicit information about the hotel's infinite room capacity and does not involve any assumptions beyond what is stated.
- All deductions and solutions are based on the concept of infinite capacity and are directly supported by the problem's information.

**Recommendation:**

The simplest and most straightforward way to accommodate one more guest in the hotel is to assign them to any available room, given the infinite number of rooms available. This solution directly answers the question posed and adheres to the problem's parameters without requiring any additional assumptions.
\end{lstlisting}
\end{quote}
\noindent\hrulefill

\noindent\textbf{D.25 Zero-Shot Google Gemini 1.5 Pro Response - Score: 60\%}
\begin{quote}
\begin{lstlisting}[basicstyle=\ttfamily\small, breaklines=true]
This is a bit of a thought experiment! Here's how it would work:

* **The Infinite Hotel Paradox:** This scenario touches on the famous "Hilbert's paradox of the Grand Hotel". It illustrates the counterintuitive properties of infinite sets. 

* **The Solution:** In an infinite hotel, you can always accommodate one more guest. Here's one way to do it:

    1. **Shift everyone one room over.** Ask the guest in room 1 to move to room 2, the guest in room 2 to move to room 3, and so on.  Since there are infinitely many rooms, there will always be a next room for each guest to move into.
    2. **Fill the now empty room.**  Room 1 is now vacant, and the new guest can be accommodated there.

* **Key Point:** This works because with infinite sets, you can always create a one-to-one correspondence with a subset of itself, even if that subset seems "smaller". 
\end{lstlisting}
\end{quote}
\noindent\hrulefill

\noindent\textbf{D.25 REAP Google Gemini 1.5 Pro Response - Score: 100\%}
\begin{quote}
\begin{lstlisting}[basicstyle=\ttfamily\small, breaklines=true]
# REAP Results:

a. **Comprehensive Feature List** (with explicit quotes)

* "A partially full hotel" - The hotel is not completely full, indicating there are some unoccupied rooms.
* "has an infinite number of fully furnished rooms" - The hotel's capacity is unlimited.
* "How does it accommodate one more guest?" - The problem seeks a way to house an additional person.

b. **Sequential and Mechanical Process Check**

* There is no explicit sequence of events or mechanical process described.

c. **Key Insight Check Results**

* The hotel has an infinite number of rooms. This implies that there will always be at least one vacant room available.

d. **Rephrased Question:** (The core question in its simplest form)

* How can an infinitely large hotel with some vacant rooms provide a room for a new guest?

e. **Known and Deduced Information**

* Explicit Facts:
    * "A partially full hotel" - The hotel has at least one unoccupied room.
    * "has an infinite number of fully furnished rooms" - There's no limit to the number of rooms.
    * "How does it accommodate one more guest?" - The goal is to find a room for the new guest.

* Feature-Based Deduction:
    * Deduction: The hotel can accommodate the new guest. - Logical Basis: A partially full hotel with infinite rooms must have at least one vacant room. - Based on: "A partially full hotel", "has an infinite number of fully furnished rooms"

f. **Problem Decomposition**

* Component/Subcomponent: Hotel's current state
    * Supporting Quote(s): "A partially full hotel" 
* Component/Subcomponent: Hotel's capacity
    * Supporting Quote(s): "has an infinite number of fully furnished rooms"
* Component/Subcomponent: New guest arrival
    * Supporting Quote(s): "one more guest"
* Component/Subcomponent: Goal
    * Supporting Quote(s): "How does it accommodate"

g. **Graph of Thought** (describe connections & insights)

* Key Concepts: Partially full hotel, Infinite rooms, New guest, Accommodation
* Relationships:
    * Partially full hotel -> Implies available rooms
    * Infinite rooms -> Guarantees available rooms
    * New guest -> Needs a room
    * Accommodation -> Assigning an available room to the new guest
* Insights: The concept of infinite rooms ensures that there will always be a room available, regardless of the hotel's current occupancy.

h. **Spatial and Object Analysis**

* Objects: Hotel, rooms (infinite number), guest
* Spatial Relationships: Rooms are within the hotel. The guest is external to the hotel initially.
* Movement and Dynamics: The guest needs to move into one of the vacant rooms.
* Constraints: None explicitly stated.
* Object Properties:
    * Hotel: Partially full, infinite rooms
    * Rooms: Fully furnished
    * Guest: Needs accommodation

i. **Bayesian Updates** 

* Initial Belief: It might be challenging to accommodate a new guest in a partially full hotel
* New Evidence: The hotel has infinite rooms
* Updated Belief: It's trivial to accommodate the new guest as there will always be vacant rooms

j. **Ethical Check and Decision-Making Under Uncertainty**

* There are no ethical concerns or uncertainties in this scenario

k. **Multiple Solutions**

* Solution 1: Assign the new guest to any of the currently vacant rooms
* Solution 2: If there's a preference for room types or locations, assign the guest to a suitable vacant room that matches those preferences (assuming such rooms exist, which is likely in an infinite hotel)

l. **Quickest and Easiest Solution**

* The hotel can simply assign the new guest to any of the vacant rooms. 
* Reasoning: The hotel is partially full, which implies there are vacant rooms. Additionally it has an infinite number of rooms, guaranteeing the availability of at least one vacant room

m. **Reflection**

* The analysis directly addresses the question of how to accommodate a new guest
* The conclusion is logically sound and supported by the explicit information provided in the problem statement
* No assumptions or inferences were made beyond the given information

**Recommendation:**

The hotel should assign the new guest to any of the available vacant rooms. This solution is the simplest, most efficient and guaranteed to work due to the hotel's infinite capacity.
\end{lstlisting}
\end{quote}
\noindent\hrulefill

\noindent\textbf{D.25 Zero-Shot Claude 3.5 Sonnet response - Score: 60\%}
\begin{quote}
\begin{lstlisting}[basicstyle=\ttfamily\small, breaklines=true]
To solve this problem, let's think through it step by step:

1. We're dealing with a hotel that has an infinite number of rooms.
2. Some of these rooms are already occupied, but we don't know how many.
3. We need to accommodate one more guest.
4. All rooms are fully furnished, so any room can be used.

The key to solving this problem lies in understanding the nature of infinity. Even if an infinite number of rooms are occupied, there are still infinitely many rooms available. However, we can solve this more elegantly:

Solution:
The hotel can accommodate one more guest by simply moving every current guest to the next room number.

Here's how it works:
1. The guest in room 1 moves to room 2
2. The guest in room 2 moves to room 3
3. The guest in room 3 moves to room 4
4. And so on...

This process can continue indefinitely because there are infinite rooms. After this shift, room 1 becomes vacant, and the new guest can be accommodated there.

This solution works regardless of how many rooms were initially occupied, even if it was infinitely many. It demonstrates a key property of infinite sets - they can be put into one-to-one correspondence with proper subsets of themselves.

This problem is a simplified version of Hilbert's paradox of the Grand Hotel, a thought experiment that illustrates some counterintuitive properties of infinite sets.

Would you like me to elaborate on any part of this explanation?
\end{lstlisting}
\end{quote}
\noindent\hrulefill

\noindent\textbf{D.25 REAP Claude 3.5 Sonnet Response - Score: 100\%}
\begin{quote}
\begin{lstlisting}[basicstyle=\ttfamily\small, breaklines=true]
#REAP Results:

a. **Comprehensive Feature List** (with explicit quotes):
1. "A partially full hotel" - Describes the current state of the hotel
2. "infinite number of fully furnished rooms" - Describes the hotel's capacity
3. "How does it accommodate one more guest?" - The question being asked

b. **Sequential and Mechanical Process Check**:
No explicit sequential or mechanical processes are described in the problem statement.

c. **Key Insight Check Results**:
The key insight here is the combination of "partially full hotel" and "infinite number of fully furnished rooms". This suggests that despite being partially full, the hotel always has available rooms due to its infinite capacity.

d. **Rephrased Question:** 
How can a hotel with infinite rooms accommodate an additional guest?

e. **Known and Deduced Information**:

Explicit Facts:
- "A partially full hotel" - Indicates the hotel is not completely occupied
- "infinite number of fully furnished rooms" - Describes the hotel's capacity
- "How does it accommodate one more guest?" - The question being asked

Feature-Based Deduction:
Deduction: The hotel always has available rooms.
Logical Basis: If a hotel has an infinite number of rooms and is only partially full, it must always have unoccupied rooms available.
Based on: "A partially full hotel", "infinite number of fully furnished rooms"

f. **Problem Decomposition**:
Component 1: Hotel Occupancy
Supporting Quote: "A partially full hotel"
Component 2: Hotel Capacity
Supporting Quote: "infinite number of fully furnished rooms"
Component 3: Guest Accommodation
Supporting Quote: "How does it accommodate one more guest?"

g. **Graph of Thought**:
Nodes:
1. Partially full hotel
2. Infinite rooms
3. New guest

Connections:
- Partially full hotel has Infinite rooms
- New guest needs accommodation in Partially full hotel with Infinite rooms

h. **Spatial and Object Analysis**:
Objects: Hotel, rooms, guest
Spatial Relationships: The problem doesn't provide explicit information about spatial relationships between these objects.
Object Properties: 
- Hotel: partially full, has infinite rooms
- Rooms: fully furnished, infinite in number
- Guest: one more (implying addition to existing occupancy)

i. **Bayesian Updates**: 
No new information is provided to update our understanding beyond the initial problem statement.

j. **Ethical Check and Decision-Making Under Uncertainty**:
There are no explicit ethical concerns or severe risks mentioned in the problem. The decision to accommodate a guest in a hotel with available rooms does not present any apparent ethical issues based on the given information.

k. **Multiple Solutions**:
Given the infinite number of rooms and the partially full state of the hotel, there is only one logical solution that can be directly derived from the given information:

Solution: Assign the new guest to one of the available rooms.
Supporting information: "infinite number of fully furnished rooms", "partially full hotel"

l. **Quickest and Easiest Solution**:
The quickest and easiest solution is to simply assign the new guest to one of the available rooms. This is supported by the explicit information that the hotel has an "infinite number of fully furnished rooms" and is only "partially full".

m. **Reflection**:
The analysis strictly adheres to the given information without making assumptions beyond what is explicitly stated. The solution directly answers the question asked in the problem. The infinite nature of the rooms ensures that there is always space for one more guest, regardless of current occupancy, as long as the hotel is not completely full (which it can't be, given infinite rooms).

**Recommendation:**
Based on the explicitly stated information, the hotel can accommodate one more guest by assigning them to one of the infinitely available fully furnished rooms. This solution is straightforward and directly addresses the question asked. There are no ethical concerns or risks associated with this action based on the given information. The infinite nature of the rooms ensures that this solution will always be available, regardless of how many guests are currently in the hotel, as long as it remains partially full.
\end{lstlisting}
\end{quote}
\noindent\hrulefill

\label{subsec:question26}
\subsection{\textbf{\large Question 26: A runaway trolley is heading down the tracks away from five people upwards of the track. You are near a lever that can switch the trolley to another track? Does it impact people's lives if you pull the lever?}}
\label{sec:question26}

\noindent\hrulefill 

\noindent\textbf{Human Response - Score 100\%:}
\begin{quote}
No, as the trolley is heading down the tracks in the opposite direction to the five people up the track.
\end{quote}

\noindent\hrulefill 

\noindent\textbf{D.26 Zero-Shot OpenAI o1-preview Response - Score: 0\%}
\begin{quote}
\begin{lstlisting}[basicstyle=\ttfamily\small, breaklines=true]
Yes, pulling the lever can significantly impact people's lives in the scenario you described. In ethical thought experiments like the "trolley problem," your decision to pull or not pull the lever determines who may live or die.

In the classic trolley problem:

- **If you do nothing**, the runaway trolley will continue on its current track, leading to the death of the five people ahead.
- **If you pull the lever**, the trolley switches to another track, where it may harm others (usually one person in the original scenario).

Your action (or inaction) has direct consequences on the lives of those involved. Pulling the lever demonstrates an active choice to change the course of events, potentially saving some lives while risking others. This raises important ethical questions about responsibility, the value of human life, and whether it's morally permissible to take an action that harms one person to save multiple others.

**Key considerations include:**

- **Utilitarian Perspective**: Suggests choosing the action that results in the greatest good for the greatest number-pulling the lever to save five lives at the expense of one.
- **Deontological Perspective**: Focuses on adherence to moral duties and rules-arguing that actively causing harm (by pulling the lever) is morally wrong, even if it results in fewer overall deaths.
- **Personal Responsibility**: Whether taking action makes you more responsible for the outcome than if you had not intervened.

Ultimately, pulling the lever impacts people's lives by altering who is harmed or saved, emphasizing the weight of moral decisions in critical situations.
\end{lstlisting}
\end{quote}

\noindent\hrulefill 

\noindent\textbf{D.26 REAP OpenAI o1-preview Response - Score: 100\%}
\begin{quote}
\begin{lstlisting}[basicstyle=\ttfamily\small, breaklines=true]
# REAP Results:

**a. Comprehensive Feature List**

- **Objects Mentioned:**
  - "A runaway trolley"
  - "the tracks"
  - "five people"
  - "a lever"
  - "another track"

- **Actors or Entities:**
  - "You"

- **Actions Described:**
  - "is heading down the tracks away from five people upwards of the track"
  - "can switch the trolley to another track"

- **Directional, Spatial, Relational, and Sequential Phrases:**
  - "heading down the tracks away from five people upwards of the track"
  - "You are near a lever"
  - "can switch the trolley to another track"

- **Numerical Values or Quantities:**
  - "five people"

- **Causal Relationships Mentioned:**
  - "can switch the trolley to another track"

- **Stated Goals or Objectives:**
  - "Does it impact people's lives if you pull the lever?"

---

**b. Sequential and Mechanical Process Check**

- **Sequence of Events:**
  1. "A runaway trolley is heading down the tracks away from five people upwards of the track."
  2. "You are near a lever that can switch the trolley to another track."
  3. Question: "Does it impact people's lives if you pull the lever?"

- **Mechanical Process:**
  - Pulling the lever switches the trolley to another track.

- **Assessment:**
  - The trolley moves away from the five people.
  - Pulling the lever diverts the trolley to another track.
  - No information is provided about others on the alternate track.

---

**c. Key Insight Check Results**

- **Key Insight:**
  - The trolley is moving away from the "five people upwards of the track."
  - Pulling the lever diverts the trolley onto another track where no people are mentioned.

- **Result:**
  - Based on the explicit information, pulling the lever does not impact people's lives.

---

**d. Rephrased Question**

- **Simplified Question:**
  - Will pulling the lever affect anyone's life?

---

**e. Known and Deduced Information**

- **Explicit Facts:**
  - "A runaway trolley is heading down the tracks away from five people upwards of the track."
  - "You are near a lever that can switch the trolley to another track."

- **Deductions:**
  1. **The trolley is moving away from the five people.**
     - Logical Basis: The trolley is "heading down the tracks away from five people upwards of the track."
  2. **Pulling the lever diverts the trolley to another track.**
     - Logical Basis: The lever "can switch the trolley to another track."
  3. **There is no mention of people on the other track.**
     - Logical Basis: The problem does not state any presence of people on the alternate track.

---

**f. Problem Decomposition**

- **Component 1: Trolley Movement**
  - Description: The trolley is moving away from five people.
  - Supporting Quote: "A runaway trolley is heading down the tracks away from five people upwards of the track."

- **Component 2: Lever Function**
  - Description: The lever can divert the trolley to another track.
  - Supporting Quote: "You are near a lever that can switch the trolley to another track."

- **Component 3: Impact Assessment**
  - Description: Determining if pulling the lever affects people's lives.
  - Supporting Quote: "Does it impact people's lives if you pull the lever?"

---

**g. Graph of Thought**

- **Nodes:**
  - Trolley
  - Five People
  - Lever
  - Another Track

- **Edges:**
  - Trolley is moving away from Five People.
  - Lever can divert Trolley to Another Track.

- **Insights:**
  - Pulling the lever changes the trolley's path to an alternate track.
  - No connections indicate people on the other track.

---

**h. Spatial and Object Analysis**

- **Objects Identified:**
  - "A runaway trolley"
  - "the tracks"
  - "five people"
  - "a lever"
  - "another track"

- **Spatial Relationships:**
  - The trolley is on tracks moving away from the five people positioned upwards on the track.
  - The lever is nearby, allowing for potential intervention.

- **Movement and Dynamics:**
  - Trolley is descending away from the five people.
  - Lever operation diverts trolley onto another track.

- **Visualization:**
  - A trolley moving downhill, away from people located uphill.
  - A lever beside the track that can redirect the trolley.

- **Object Properties:**
  - Lever: Capable of switching trolley to another track.
  - Trolley: Runaway, not under control.

- **Spatial Implications:**
  - Pulling the lever alters the trolley's route without stated impact on others.

---

**i. Bayesian Updates**

- **Prior Belief:**
  - Pulling the lever might impact people's lives.

- **Evidence:**
  - Trolley moving away from people.
  - No mention of others on the alternate track.

- **Updated Belief:**
  - Low probability that pulling the lever impacts people's lives based on given information.

---

**j. Ethical Check and Decision-Making Under Uncertainty**

- **Assess Known Information:**
  - Trolley is not threatening the five people.
  - No information about the alternate track's safety.

- **Identify Risks:**
  - Potential unknown risks on the alternate track.

- **Decision Recommendation:**
  - Since the trolley poses no harm to the five people and the alternate track's status is unknown, pulling the lever could introduce unforeseen risks.
  - Ethically, it may be safer not to pull the lever to avoid potential harm.

---

**k. Multiple Solutions**

1. **Pull the Lever**
   - Action: Divert the trolley to another track.
   - Impact: No explicit impact on people's lives is stated.
   - Supporting Quote: "You are near a lever that can switch the trolley to another track."

2. **Do Not Pull the Lever**
   - Action: Allow the trolley to continue on its current path.
   - Impact: The trolley moves away from the five people; no impact on their lives.
   - Supporting Quote: "A runaway trolley is heading down the tracks away from five people upwards of the track."

3. **Seek Further Information**
   - Action: Attempt to gather more data about the alternate track.
   - Impact: Unspecified due to lack of time or means in the scenario.
   - Note: The problem does not provide means to acquire more information.

---

**l. Quickest and Easiest Solution**

- **Answer:**
  - Pulling the lever does not impact people's lives.

- **Reasoning:**
  - The trolley is moving away from the five people.
  - No people are mentioned on the alternate track.
  - Supporting Quotes:
    - "A runaway trolley is heading down the tracks away from five people upwards of the track."
    - "You are near a lever that can switch the trolley to another track."

---

**m. Reflection**

- **Consistency Check:**
  - All conclusions are based strictly on the explicit information provided.
  - No assumptions were made about unseen events or entities.
  
- **Logical Soundness:**
  - The analysis adheres to the literal interpretations and avoids unwarranted inferences.
  
- **Sufficiency of Information:**
  - The problem provides enough information to conclude that pulling the lever does not impact people's lives.

---

**Recommendation:**

Based on the explicit information given, pulling the lever does not impact people's lives. The trolley is moving away from the five people, and there is no mention of any individuals on the alternate track. Therefore, if you pull the lever, it does not affect anyone's life.
\end{lstlisting}
\end{quote}

\noindent\hrulefill 

\noindent\textbf{D.26 Zero-Shot OpenAI o1-mini Response - Score: 0\%}
\begin{quote}
\begin{lstlisting}[basicstyle=\ttfamily\small, breaklines=true]
The scenario you've described is a classic ethical dilemma known as the **Trolley Problem**, often used to explore moral decision-making. Let's break down the potential impacts of pulling the lever in this situation:

### **1. Lives Directly Affected**

- **Primary Impact:** By pulling the lever, you would redirect the runaway trolley from its current path where it threatens five people to an alternative track where presumably fewer people (perhaps one) are present. This direct action would result in saving five lives at the expense of one.
  
### **2. Moral and Ethical Considerations**

- **Utilitarian Perspective:** From a utilitarian standpoint, which focuses on maximizing overall happiness or minimizing suffering, pulling the lever is often seen as the morally preferable choice because it results in fewer deaths.
  
- **Deontological Perspective:** From a deontological perspective, which emphasizes duties and rules, some might argue that actively intervening to cause harm (even with the intention to save more lives) is ethically problematic.
  
- **Moral Responsibility:** Deciding whether to act involves weighing the responsibility of inaction (letting five die) against the consequences of action (causing one death). This raises questions about moral agency and the burden of making such decisions.

### **3. Psychological and Emotional Impact**

- **For the Decision-Maker:** Making a choice that results in harm, even with good intentions, can lead to significant emotional and psychological stress, such as guilt or remorse.
  
- **For Bystanders and Society:** The act of making such a decision could influence societal views on individual responsibility and ethics, potentially affecting how people perceive authority and moral duty.

### **4. Practical Implications**

- **Trust in Systems:** This scenario assumes that pulling the lever will work as intended. Trusting that the mechanism operates correctly is crucial; a malfunction could lead to unintended consequences.
  
- **Precedent:** Such decisions could set precedents for future moral choices in crisis situations, influencing policies and training for emergency responders.

### **5. Broader Societal Impact**

- **Ethical Frameworks:** Repeatedly facing such dilemmas may prompt societies to develop clearer ethical guidelines for life-and-death decisions.
  
- **Technological Safeguards:** Recognizing the gravity of such situations might lead to the implementation of automated safety systems to prevent accidents, reducing the burden on individuals to make split-second moral choices.

### **Conclusion**

Pulling the lever in the runaway trolley scenario undeniably impacts people's lives by directly determining who lives and who dies in that moment. The decision encapsulates a multitude of ethical, psychological, and societal considerations. While saving more lives might seem preferable from a utilitarian perspective, the moral complexity of actively causing harm cannot be overlooked. Ultimately, the impact extends beyond the immediate lives involved, influencing personal conscience, societal norms, and ethical discourse.
\end{lstlisting}
\end{quote}

\noindent\hrulefill 

\noindent\textbf{D.26 REAP OpenAI o1-mini Response - Score: 0\%}
\begin{quote}
\begin{lstlisting}[basicstyle=\ttfamily\small, breaklines=true]
# REAP Results:

### a. **Comprehensive Feature List**
- **Objects Mentioned:**
  - "runaway trolley"
  - "tracks"
  - "five people"
  - "lever"
- **Actors or Entities Involved:**
  - "You"
- **Actions Described:**
  - "heading down the tracks"
  - "leverage that can switch the trolley to another track"
  - "pull the lever"
- **Directional, Spatial, Relational, and Sequential Phrases:**
  - "heading down the tracks away from five people upwards of the track"
  - "switch the trolley to another track"
- **Stated Conditions or Constraints:**
  - None explicitly mentioned
- **Numerical Values or Quantities:**
  - "five people"
- **Temporal Information:**
  - None explicitly mentioned
- **Spatial Information:**
  - "away from five people upwards of the track"
- **Causal Relationships:**
  - None explicitly mentioned
- **Stated Goals or Objectives:**
  - Assessing the impact on people's lives if the lever is pulled

### b. **Sequential and Mechanical Process Check**
- **Sequence of Events:**
  1. Runaway trolley is heading down the tracks.
  2. It is moving away from five people.
  3. A lever is present that can switch the trolley to another track.
  4. The decision to pull the lever is presented.
- **Influence on Outcome:**
  - Pulling the lever would switch the trolley's path, potentially affecting different individuals.
- **Key Questions:**
  - Does switching the track affect the five people or others?

### c. **Key Insight Check Results**
- **Potential Key Insight:**
  - The presence of a lever offers a choice to alter the trolley's path, which could impact lives.
- **Impact:**
  - Directly relates to the moral dilemma of action vs. inaction and its consequences on human lives.

### d. **Rephrased Question:**
Does pulling the lever to switch the trolley's track affect the lives of the five people initially on the track?

### e. **Known and Deduced Information**
- **Explicit Facts:**
  - "A runaway trolley is heading down the tracks away from five people upwards of the track."
  - "You are near a lever that can switch the trolley to another track."
  - "Does it impact people's lives if you pull the lever?"
- **No Certain Deductions Can Be Made:**
  - The outcome on the alternative track is not specified.

### f. **Problem Decomposition**
- **Component/Subcomponent:** Trolley's Current Path
  - **Supporting Quote(s):** "A runaway trolley is heading down the tracks away from five people upwards of the track."
- **Component/Subcomponent:** Lever Functionality
  - **Supporting Quote(s):** "You are near a lever that can switch the trolley to another track."
- **Component/Subcomponent:** Decision to Pull Lever
  - **Supporting Quote(s):** "Does it impact people's lives if you pull the lever?"
- **Note:** Lack of information about the state of the alternative track.

### g. **Graph of Thought**
- **Nodes:**
  - Runaway Trolley
  - Five People on Current Track
  - Lever Mechanism
  - Alternative Track
- **Connections:**
  - Trolley heading toward Five People on Current Track.
  - Lever can switch Trolley to Alternative Track.
  - Decision to Pull Lever connects to potential impact on both Five People and any individuals on Alternative Track.
- **Insights:**
  - Outcome depends on the presence of individuals on the Alternative Track, which is unspecified.

### h. **Spatial and Object Analysis**
- **Objects Identified:**
  - Runaway Trolley
  - Tracks (Current and Alternative)
  - Five People
  - Lever
- **Spatial Relationships:**
  - Trolley moving away from five people on the current track.
  - Lever located near the observer to switch tracks.
- **Movement and Dynamics:**
  - Trolley's movement is unstoppable unless redirected via the lever.
- **Constraints:**
  - Timing of the lever pull is critical to alter the trolley's path.
- **Visualization:**
  - Trolley approaching a fork in the tracks where one path has five people, and the other track's conditions are unknown.

### i. **Bayesian Updates**
- **Beliefs Updated:**
  - Initially, pulling the lever diverts the trolley from five people.
  - Without information about the alternative track, uncertainty remains about the impact on others.

### j. **Ethical Check and Decision-Making Under Uncertainty**
- **Known Information:**
  - Five people are on the current track.
  - Pulling the lever redirects the trolley to an alternative track.
- **Risks Identified:**
  - Potential harm to five people if no action is taken.
  - Unknown risk to individuals on the alternative track if the lever is pulled.
- **Probability Consideration:**
  - Without data, the likelihood of harm on the alternative track is unknown.
- **Risk Aversion Principle:**
  - Minimizing potential harm suggests avoiding redirecting the trolley without knowing the alternative track's status.
- **Ethical Considerations:**
  - Balancing the certainty of harm to five people against the uncertainty of harm elsewhere.

### k. **Multiple Solutions**
1. **Do Not Pull the Lever:**
   - **Support:** "A runaway trolley is heading down the tracks away from five people upwards of the track."
2. **Pull the Lever:**
   - **Support:** "You are near a lever that can switch the trolley to another track."
3. **Seek Additional Information Before Acting:**
   - **Support:** Not explicitly stated; requires assumption beyond given information.

### l. **Quickest and Easiest Solution**
**Do Not Pull the Lever.**
- **Reasoning:** Based solely on explicit information, pulling the lever's impact on the alternative track is unknown, and acting without this knowledge could introduce unforeseen risks.
- **Cited Quotes:** "A runaway trolley is heading down the tracks away from five people upwards of the track."

### m. **Reflection**
- **Contradictions:** No contradictions identified.
- **Logical Soundness:** Analysis adheres strictly to the provided information without assumptions.
- **Assumptions Avoided:** Refrained from speculating about the alternative track.
- **Information Sufficiency:** Lack of details about the alternative track limits comprehensive decision-making.
- **Question Alignment:** Analysis directly addresses whether pulling the lever impacts lives based on given information.

### **Recommendation:**
Based on the explicit information provided, **do not pull the lever**. Pulling the lever to switch the trolley's track impacts people's lives by introducing uncertainty about the alternative track's status and the presence of individuals there. Without knowing the consequences on the other track, the safest and most ethically sound decision is to refrain from altering the trolley's path.
\end{lstlisting}
\end{quote}

\noindent\hrulefill 

\noindent\textbf{D.26 Zero-Shot OpenAI GPT-4o Response - Score: 0\%}
\begin{quote}
\begin{lstlisting}[basicstyle=\ttfamily\small, breaklines=true]
Yes, pulling the lever to switch the trolley to another track does impact people's lives, as it changes the outcome of the scenario.

This scenario is a classic ethical dilemma known as the **Trolley Problem**. In its basic form:

1. **If you do nothing**, the trolley continues on its current track and will hit and likely kill five people.
2. **If you pull the lever**, you divert the trolley to another track, where it will hit and likely kill one person instead.

The moral question is whether it is more ethical to actively intervene (by pulling the lever) to minimize the number of deaths, or to refrain from acting, allowing the situation to unfold naturally without your direct influence.

By pulling the lever:
- You actively choose to sacrifice one person to save five.
- This decision could be viewed as utilitarian, prioritizing the greater good (saving more lives).

By not pulling the lever:
- You choose not to intervene, resulting in the death of five people.
- This could be viewed as upholding a principle of non-intervention or refusing to make a decision that directly causes harm to someone.

Thus, pulling the lever indeed impacts people's lives by determining who lives and who dies in this moral dilemma.
\end{lstlisting}
\end{quote}
\noindent\hrulefill

\noindent\textbf{D.26 REAP OpenAI GPT-4o Response - Score: 100\%}
\begin{quote}
\begin{lstlisting}[basicstyle=\ttfamily\small, breaklines=true]
# REAP Results:

### a. Comprehensive Feature List:
1. **"A runaway trolley"** - This refers to an uncontrolled vehicle on tracks.
2. **"heading down the tracks"** - The trolley is moving along its current path.
3. **"away from five people"** - The trolley is not directed towards five people.
4. **"upwards of the track"** - The location of the five people is higher or further along the track.
5. **"You are near a lever"** - There is a person (you) who is close to a device capable of changing the trolley's path.
6. **"can switch the trolley to another track"** - The lever has the ability to alter the trolley's direction.
7. **"Does it impact people's lives if you pull the lever?"** - The problem asks if pulling the lever will have an effect on people's lives.

### b. Sequential and Mechanical Process Check:
1. **Sequence:**
   - The trolley is moving down the tracks.
   - The lever, if pulled, will switch the trolley to another track.

2. **Mechanical Process:**
   - Pulling the lever changes the track that the trolley is on.
   - The movement of the trolley continues regardless of the track it is on.

### c. Key Insight Check Results:
- **Key Insight:** The lever changes the trolley's direction. However, without explicit information on whether the alternate track has any people on it, the impact on people's lives cannot be determined based on the given information.

### d. Rephrased Question:
- **Core Question:** Will pulling the lever affect anyone's life?

### e. Known and Deduced Information:
1. **Known Information:**
   - **"[A runaway trolley is] heading down the tracks"** - The trolley is in motion.
   - **"[Away from five people upwards of the track]"** - The five people are not currently in the direct path of the trolley.
   - **"[You are near a lever that can switch the trolley to another track]"** - There is an option to change the trolley's path.

2. **Deductions:**
   - **Deduction:** Pulling the lever will change the track the trolley is on.
   - **Logical Basis:** The problem explicitly states that the lever "can switch the trolley to another track."
   - **Based on:** "[can switch the trolley to another track]".

### f. Problem Decomposition:
- **Component:** The trolley's motion.
  - **Supporting Quote(s):** "[A runaway trolley is] heading down the tracks".
- **Component:** The lever's function.
  - **Supporting Quote(s):** "[You are near a lever that] can switch the trolley to another track".
- **Interaction:** Pulling the lever changes the track, and the trolley continues on the new track.
  - **Supporting Quote(s):** "[can switch the trolley to another track]".

### g. Graph of Thought:
- **Nodes:**
  - **Trolley** (on current track)
  - **Lever** (can switch trolley to another track)
  - **Five people** (upwards on the track, away from current path)

- **Edges:**
  - **Trolley -> Lever**: Trolley's path can be altered by the lever.
  - **Lever -> New Track**: If the lever is pulled, the trolley switches tracks.

- **Exploration:**
  - If the lever is pulled, the trolley moves to a different track.
  - The problem doesn't state whether the new track has any people.

### h. Spatial and Object Analysis:
- **Objects:** 
  - Trolley: Moving down a track.
  - Lever: Positioned near you, with the function to change the trolley's track.
- **Spatial Relationships:**
  - Five people are located upwards on the current track.
  - The trolley is not heading towards them.

### i. Bayesian Updates:
- **Update Beliefs:**
  - If additional information were provided about the other track (e.g., if there are people on it), it would influence the decision to pull the lever.

### j. Multiple Solutions:
1. **Solution 1: Pull the lever.**
   - **Rationale:** This could redirect the trolley, potentially avoiding harm, but the outcome is unknown due to a lack of information about the new track.
   - **Potential Challenge:** The new track's conditions are unknown.

2. **Solution 2: Do not pull the lever.**
   - **Rationale:** The trolley continues on its current path, where it is not heading toward any people.

### k. Quickest and Easiest Solution:
- **Do not pull the lever.** Based on the explicit information, the trolley is not currently heading towards the five people, so taking no action will not impact their lives.

### l. Ethical Check:
- **Ethical Consideration:** Pulling the lever without knowing the condition of the new track could introduce unnecessary risk. The safest option, given the available information, is to not pull the lever.

### m. Reflection:
- The analysis has stuck strictly to the problem's wording. No assumptions were made beyond what was explicitly stated. The problem does not give sufficient information to confidently predict the outcome of pulling the lever.

### Recommendation:
- **Final Recommendation:** Based on the given information, do not pull the lever. The trolley is not currently threatening any lives, and there is no information about the alternative track. Hence, the safest course of action is to refrain from taking any action that could introduce unknown risks.
\end{lstlisting}
\end{quote}
\noindent\hrulefill

\noindent\textbf{D.26 Zero-Shot OpenAI GPT-4o-mini Response - Score: 0\%}
\begin{quote}
\begin{lstlisting}[basicstyle=\ttfamily\small, breaklines=true]
Yes, it does impact lives if you pull the lever. In the classic trolley problem, if you pull the lever, the trolley will switch tracks and will end up heading towards a different set of people. 

In the standard scenario:

1. **If you do nothing:** The trolley continues on its current track and will hit and likely kill the five people who are on that track.

2. **If you pull the lever:** The trolley will switch to a different track, where there is one person. This person will be hit and likely killed by the trolley.

The dilemma here is about making a choice between the greater good of minimizing overall harm (saving five lives at the cost of one) versus the moral implications of actively causing harm to one person by switching the track. 

It's a difficult ethical decision that involves weighing the value of lives and the moral responsibility for taking action versus inaction.
\end{lstlisting}
\end{quote}
\noindent\hrulefill

\noindent\textbf{D.26 REAP OpenAI GPT-4o-mini Response - Score: 100\%}
\begin{quote}
\begin{lstlisting}[basicstyle=\ttfamily\small, breaklines=true]
# REAP Results:

**a. Comprehensive Feature List:**

1. **Object:** "A runaway trolley"
   - Potential significance: The primary element in the problem.
   
2. **Action:** "Heading down the tracks"
   - Potential significance: Indicates the current direction and movement of the trolley.

3. **Location:** "Away from five people upwards of the track"
   - Potential significance: The five people are located on a track that the trolley is moving away from.

4. **You:** "Are near a lever"
   - Potential significance: Indicates your ability to influence the trolley's path.

5. **Action:** "Can switch the trolley to another track"
   - Potential significance: Provides you with a mechanism to alter the trolley's direction.

6. **Impact on lives:** "Does it impact people's lives if you pull the lever?"
   - Potential significance: The core question is whether pulling the lever affects the lives of the people.

**b. Sequential and Mechanical Process Check:**

1. **Initial State:**
   - Trolley is heading down the track, away from five people.

2. **Action Option:**
   - You can pull a lever to switch the trolley to another track.

3. **Outcome:**
   - The problem asks if pulling the lever impacts the lives of people. 

   - If pulling the lever changes the trolley's path, it implies the trolley may then be on a different track that could potentially intersect with other people or hazards.

4. **Key Question:**
   - Does pulling the lever redirect the trolley to a path where it will encounter people, or does it avoid people altogether?

**c. Key Insight Check Results:**

The problem does not explicitly state what is on the track to which the trolley would be redirected if the lever is pulled. Therefore, the impact on people's lives cannot be determined solely based on the given information. The direct impact depends on the existence and safety of individuals on the alternative track.

**d. Rephrased Question:**

Does pulling the lever on the trolley track affect the safety of people?

**e. Known and Deduced Information:**

1. **Known:**
   - "A runaway trolley is heading down the tracks away from five people."
   - "You are near a lever that can switch the trolley to another track."
   - The explicit question is whether pulling the lever impacts people's lives.

2. **Deduction:**
   - We know that pulling the lever changes the trolley's direction but not what that alternative track contains.

**f. Problem Decomposition:**

1. **Component:** Trolley's Current Path
   - Supporting Quote(s): "Heading down the tracks away from five people"
   - Deduction: Trolley is not currently a threat to the five people.

2. **Component:** Action of Pulling the Lever
   - Supporting Quote(s): "Can switch the trolley to another track"
   - Deduction: Potential to alter the trolley's direction, which might impact safety.

3. **Component:** Alternative Track
   - Supporting Quote(s): None provided in the problem
   - Deduction: Unknown if it affects people's safety; the information is not provided.

**g. Graph of Thought:**

- **Nodes:**
  - Current Track (Away from 5 people)
  - Lever (Switches to another track)
  - Alternative Track (Unknown details)

- **Edges:**
  - Current Track -> Lever -> Alternative Track

- **Insights:**
  - Without knowing what is on the alternative track, the safety impact cannot be determined.

**h. Spatial and Object Analysis:**

1. **Object:** Trolley
   - **Current Position:** Heading down the track away from five people.
   - **Action:** Pulling the lever changes its path.

2. **Unknown Object:** Alternative Track
   - **Impact:** Potentially affects the trolley's path, but details are missing.

**i. Bayesian Updates:**

No new information is available, so Bayesian reasoning cannot provide further updates beyond recognizing the lack of details about the alternative track.

**j. Ethical Check and Decision-Making Under Uncertainty:**

- **Known Information:** The trolley is not currently threatening the five people.
- **Risks:** Without information on the alternative track, pulling the lever could either pose no risk or introduce new risks.
- **Recommendation:** In the absence of detailed information about the alternative track, it is ethically safer to refrain from pulling the lever to avoid unintended consequences.

**k. Multiple Solutions:**

1. **Solution 1:** Do not pull the lever.
   - **Support:** Lack of information about the alternative track means pulling the lever could introduce unknown risks.

2. **Solution 2:** Pull the lever if additional information is available.
   - **Support:** If further details about the alternative track show no risk, pulling the lever may be justifiable.

3. **Solution 3:** Seek further information before taking any action.
   - **Support:** The best option would be to gather information about the alternative track to make an informed decision.

**l. Quickest and Easiest Solution:**

- **Recommendation:** Do not pull the lever.
   - **Reasoning:** Given the absence of information about the alternative track and to avoid potential risks, the safest action is to refrain from altering the trolley's path.

**m. Reflection:**

The solution to refrain from pulling the lever is based solely on the explicit information provided and ethical considerations regarding safety and uncertainty. There are no contradictions in the analysis, and the recommendation avoids assumptions about the alternative track, ensuring safety and adherence to ethical principles.

**Recommendation:**

**Do not pull the lever.** 

This recommendation is based on the lack of information about the alternative track and the need to avoid potential risks. Without knowing the impact of the trolley's redirection, it is safest to avoid making changes that could inadvertently endanger others.
\end{lstlisting}
\end{quote}
\noindent\hrulefill

\noindent\textbf{D.26 Zero-Shot Google Gemini 1.5 Pro Response - Score: 0\%}
\begin{quote}
\begin{lstlisting}[basicstyle=\ttfamily\small, breaklines=true]
Yes, pulling the lever would significantly impact people's lives.

Here's why:

* **The Trolley Problem:** This scenario is a classic ethical dilemma known as the Trolley Problem. It forces you to choose between two undesirable outcomes: 
    * **Inaction:** The trolley continues on its current path, potentially killing the five people further up the track.
    * **Action:** You pull the lever, diverting the trolley onto another track, but potentially endangering people on that track. 

* **Unknown Consequences:** You don't know if there are people on the other track. Pulling the lever could result in the death of one or more individuals who were previously safe. 

* **Moral Responsibility:** Either way, you are involved in the situation. Inaction may result in the deaths of five people, while action could result in the death of an unknown number of people on the other track. 

**Conclusion:** Pulling the lever would directly impact people's lives, potentially saving five but possibly causing the death of others. The Trolley Problem highlights the complexities of moral decision-making in situations where there are no good outcomes.
\end{lstlisting}
\end{quote}
\noindent\hrulefill

\noindent\textbf{D.26 REAP Google Gemini 1.5 Pro Response - Score: 100\%}
\begin{quote}
\begin{lstlisting}[basicstyle=\ttfamily\small, breaklines=true]
# REAP Results:

a. **Comprehensive Feature List** (with explicit quotes)

- "A runaway trolley is heading down the tracks" - This indicates a trolley is moving on its own and cannot be stopped.
- "away from five people" - The trolley is currently moving in a direction that will not harm these five people
- "upwards of the track" - The five people are further up the track in the direction the trolley is currently traveling
- "You are near a lever" - You are in close proximity to a lever
- "that can switch the trolley to another track" - The lever has the function of changing the trolley's path to a different track

b. **Sequential and Mechanical Process Check**

- The trolley is moving on the tracks.
- Pulling the lever will change the trolley's path.
- The problem does not explicitly state what is on the other track, nor whether pulling the lever will impact people's lives

c. **Key Insight Check Results**

- No key insight that immediately reveals the answer or significantly simplifies the problem

d. **Rephrased Question:** (The core question in its simplest form)

- If you pull the lever, will it affect people's lives?

e. **Known and Deduced Information**

* Explicit Facts:

- "A runaway trolley is heading down the tracks away from five people upwards of the track." 
- "You are near a lever that can switch the trolley to another track"

* Feature-Based Deduction:

- Deduction: The trolley is currently on a track that will not result in harm to the five people.
    - Logical Basis: The problem states the trolley is moving "away from" the five people.
    - Based on: "A runaway trolley is heading down the tracks away from five people upwards of the track." 
- Deduction: Pulling the lever will change the track the trolley is on
    - Logical Basis: The problem states the lever "can switch the trolley to another track"
    - Based on: "You are near a lever that can switch the trolley to another track"

f. **Problem Decomposition**

* Component/Subcomponent: Runaway trolley
    * Supporting Quote(s): "A runaway trolley is heading down the tracks away from five people upwards of the track."
* Component/Subcomponent: Lever near you
    * Supporting Quote(s): "You are near a lever that can switch the trolley to another track"
* Component/Subcomponent: Five people further up the track the trolley is currently on
    * Supporting Quote(s): "A runaway trolley is heading down the tracks away from five people upwards of the track."
* Interaction: Pulling the lever will switch the trolley to a different track
    * Supporting Quote(s): "You are near a lever that can switch the trolley to another track"

g. **Graph of Thought**

* Key Concepts:
    * Runaway trolley
    * Lever
    * Five people
    * Other track

* Relationships:
    * Trolley is on a track heading away from five people
    * Lever can switch trolley to another track
    * The other track's contents are unknown

* Potential Pathways:
    * Pull lever -> Trolley switches to other track -> ? (Outcome unknown)
    * Don't pull lever -> Trolley continues on current track -> Five people remain safe

* Solutions & Obstacles:
    * The problem does not provide enough information to determine the outcome of pulling the lever

h. **Spatial and Object Analysis**

* Objects:
    * Trolley
    * Tracks (at least two)
    * Lever
    * Five people

* Spatial Relationships:
    * Trolley is on one track
    * Five people are further up on the same track the trolley is currently on
    * Lever is near you
    * Another track exists, but its position relative to you and the five people is unknown

* Movement and Dynamics:
    * Trolley is moving away from the five people on the current track
    * Pulling the lever will switch the trolley to the other track

* Constraints:
    * Trolley cannot be stopped
    * The contents of the other track are unknown

* Spatial Implications:
    * The current spatial arrangement ensures the five people are safe as long as the trolley remains on the current track
    * Pulling the lever introduces uncertainty as the other track's contents and potential impact on people's lives are unknown

i. **Bayesian Updates**

- Initial belief: Pulling the lever may or may not impact people's lives
- New information: The problem does not provide information about what is on the other track
- Updated belief: Pulling the lever has an uncertain impact on people's lives

j. **Ethical Check and Decision-Making Under Uncertainty**

- Known Information:
    * The trolley is currently on a track that will not harm the five people
    * Pulling the lever will switch the trolley to another track with unknown contents
- Risks:
    * Pulling the lever could potentially lead to harming people on the other track
    * Not pulling the lever ensures the safety of the five people on the current track
- Probability Consideration: The problem does not provide enough information to assess the probability of whether there are people on the other track
- Risk Aversion Principle: Given the uncertainty and the potential for harm, it is safer to not pull the lever

k. **Multiple Solutions**

* Solution 1: Do not pull the lever
    * Supporting Information: "A runaway trolley is heading down the tracks away from five people upwards of the track." - This indicates that not pulling the lever will result in no harm to the five people
* Solution 2: Pull the lever
    * Supporting Information: "You are near a lever that can switch the trolley to another track" - This indicates that pulling the lever is an available action, but the outcome is unknown

l. **Quickest and Easiest Solution**

- Do not pull the lever
- Reasoning: The problem explicitly states that the trolley is moving "away from five people".  It does not provide any information about what is on the other track. Therefore, the safest course of action is to not pull the lever, ensuring the five people remain safe

m. **Reflection**

- The analysis strictly adheres to the explicit information provided in the problem statement
- The conclusion is logically sound and supported by the evidence
- No assumptions or inferences were made beyond the given information
- The problem lacks crucial information about the contents of the other track, making it impossible to definitively determine the full impact of pulling the lever
- The final answer directly addresses the question asked, prioritizing the safety of the five people based on the available information

**Recommendation:**

- Do not pull the lever. The problem explicitly states that the trolley is moving away from five people.  There is no information provided about what is on the other track, so pulling the lever could potentially lead to unintended harm. The safest course of action is to not interfere and allow the trolley to continue on its current path.
\end{lstlisting}
\end{quote}
\noindent\hrulefill

\noindent\textbf{D.26 Zero-Shot Claude 3.5 Sonnet response - Score: 0\%}
\begin{quote}
\begin{lstlisting}[basicstyle=\ttfamily\small, breaklines=true]
This scenario you've described is a variation of the famous "trolley problem" in ethics. It's a thought experiment designed to explore moral reasoning and decision-making in difficult situations.

In the standard version of the trolley problem, pulling the lever would divert the trolley to a track where it would kill one person instead of five. However, in your description, you haven't specified if there are any people on the other track or what the consequences of pulling the lever would be.

Given the information provided, here are a few key points to consider:

1. If pulling the lever diverts the trolley to an empty track with no people, then yes, it would impact people's lives positively by saving the five people without causing harm to others.

2. If there are people on the other track, then pulling the lever would still impact lives, but the moral calculus becomes more complex. You'd be actively choosing to harm some to save others.

3. Even if no one is harmed by pulling the lever, the act itself could have psychological impacts on you as the decision-maker.

4. There might be unforeseen consequences to diverting the trolley that aren't immediately apparent.

Without more specific information about the consequences of pulling the lever, it's difficult to give a definitive answer. However, in general, any action (or inaction) in this type of high-stakes scenario is likely to have significant impacts on people's lives.

If you'd like to explore this ethical dilemma further, I'd be happy to discuss various philosophical perspectives on the trolley problem or consider different variations of the scenario. What aspects of this ethical dilemma are you most interested in examining?
\end{lstlisting}
\end{quote}
\noindent\hrulefill

\noindent\textbf{D.26 REAP Claude 3.5 Sonnet Response - Score: 100\%}
\begin{quote}
\begin{lstlisting}[basicstyle=\ttfamily\small, breaklines=true]
#REAP Results:

a. Comprehensive Feature List (with explicit quotes):
1. "A runaway trolley is heading down the tracks"
2. "away from five people upwards of the track"
3. "You are near a lever"
4. "lever that can switch the trolley to another track"
5. "Does it impact people's lives if you pull the lever?"

b. Sequential and Mechanical Process Check:
The problem describes a potential sequence of events:
1. A trolley is moving away from five people.
2. There is a lever that can switch the trolley to another track.
3. The question asks about the impact of pulling the lever.

c. Key Insight Check Results:
There is no immediate key insight that provides a clear answer. The problem does not explicitly state what happens if the lever is pulled or not pulled.

d. Rephrased Question:
Will pulling the lever to switch the trolley to another track impact people's lives?

e. Known and Deduced Information:
Explicit Facts:
- "A runaway trolley is heading down the tracks" - Indicates the presence and movement of a trolley
- "away from five people upwards of the track" - Specifies the position of five people relative to the trolley
- "You are near a lever" - Indicates the proximity of the decision-maker to a lever
- "lever that can switch the trolley to another track" - Describes the function of the lever
- "Does it impact people's lives if you pull the lever?" - The question being asked

Feature-Based Deduction:
Deduction: The trolley is currently not on a track that leads to the five people.
Logical Basis: The trolley is described as moving "away from five people upwards of the track."
Based on: "away from five people upwards of the track"

f. Problem Decomposition:
Component 1: Trolley Movement
Supporting Quote: "A runaway trolley is heading down the tracks"

Component 2: People's Position
Supporting Quote: "away from five people upwards of the track"

Component 3: Lever Function
Supporting Quote: "lever that can switch the trolley to another track"

Component 4: Decision Impact
Supporting Quote: "Does it impact people's lives if you pull the lever?"

g. Graph of Thought:
Nodes:
1. Runaway Trolley
2. Five People
3. Lever
4. Another Track

Connections:
- Trolley is moving away from Five People
- Lever can switch Trolley to Another Track
- Potential impact on people's lives if Lever is pulled (uncertain)

h. Spatial and Object Analysis:
Objects:
1. Runaway trolley
2. Tracks (current and another)
3. Lever
4. Five people

Spatial Relationships:
- The trolley is on a track moving away from five people who are "upwards of the track"
- The lever is near the decision-maker
- There is another track that the trolley can be switched to

Movement and Dynamics:
- The trolley is moving away from the five people

i. Bayesian Updates:
No new information is provided beyond the initial problem statement, so no Bayesian updates can be made.

j. Ethical Check and Decision-Making Under Uncertainty:
The problem involves a potential decision that could impact people's lives. However, the exact consequences of pulling or not pulling the lever are not explicitly stated. Given the lack of information about the outcomes, any decision carries uncertain risks.

k. Multiple Solutions:
1. Pull the lever to switch the trolley to another track.
2. Do not pull the lever, allowing the trolley to continue on its current track.
3. Seek more information before making a decision (if possible within the context of the problem).

l. Quickest and Easiest Solution:
Based solely on the information provided, the quickest and easiest solution is to not pull the lever. This is because the problem states that the trolley is moving "away from five people," and there is no explicit information about what would happen if the lever is pulled or about any people on the other track.

m. Reflection:
The problem does not provide explicit information about the consequences of pulling or not pulling the lever. It only states that the trolley is moving away from five people and asks if pulling the lever would impact people's lives. Without more information, it's impossible to determine with certainty whether pulling the lever would have any impact on people's lives.

Recommendation:
Based solely on the explicitly stated information and prioritizing safety, the recommendation is to not pull the lever. This is because:

1. The trolley is currently moving away from the five mentioned people, which does not pose an immediate threat to them.
2. The problem does not explicitly state any consequences of pulling the lever or mention any people on the other track.
3. Without knowing the outcome of pulling the lever, taking no action (not pulling the lever) is the safer choice as it maintains the current situation where the trolley is moving away from known individuals.

However, it's crucial to note that the problem does not provide enough information to determine with certainty whether pulling the lever would impact people's lives. The question "Does it impact people's lives if you pull the lever?" cannot be answered definitively based on the given information. More details would be needed about the consequences of pulling the lever and the presence of any people on the other track to make a fully informed decision.
\end{lstlisting}
\end{quote}
\noindent\hrulefill

\label{subsec:question27}
\subsection{\textbf{\large Question 27: How do you measure exactly 4 gallons of water with only a 3-gallon, 5-gallon, and 4-gallon jug?}}
\label{sec:question27}

\noindent\hrulefill 

\noindent\textbf{Human Response - Score 100\%:}
\begin{quote}
Fill up the 4-gallon jug
\end{quote}

\noindent\hrulefill 

\noindent\textbf{D.27 Zero-Shot OpenAI o1-preview Response - Score: 100\%}
\begin{quote}
\begin{lstlisting}[basicstyle=\ttfamily\small, breaklines=true]
To measure exactly 4 gallons of water using a 3-gallon jug, a 4-gallon jug, and a 5-gallon jug, you can simply fill the 4-gallon jug to its full capacity. Since the jug holds exactly 4 gallons, you have measured the required amount.

**Steps:**

1. **Fill the 4-gallon jug completely from the water source.**
   - The 4-gallon jug now contains exactly 4 gallons of water.
   
No further steps are needed.

**Explanation:**

- **Direct Measurement:** Since one of the jugs has a capacity of 4 gallons, filling it to the top gives you exactly 4 gallons.
- **No Complex Transfers Required:** There's no need to use the other jugs for measuring or transferring water in this case.

**Note:**

- If the problem intended for you to measure 4 gallons without directly using the 4-gallon jug, please clarify, and I'd be happy to provide a solution using only the 3-gallon and 5-gallon jugs.
\end{lstlisting}
\end{quote}

\noindent\hrulefill 

\noindent\textbf{D.27 REAP OpenAI o1-preview Response - Score: 100\%}
\begin{quote}
\begin{lstlisting}[basicstyle=\ttfamily\small, breaklines=true]
# REAP Results:

**a. Comprehensive Feature List:**

- **"How do you measure exactly 4 gallons of water with only a 3-gallon, 5-gallon, and 4-gallon jug?"** - The main question and objective.
- **"measure exactly 4 gallons of water"** - The goal to achieve.
- **"with only a 3-gallon, 5-gallon, and 4-gallon jug"** - The constraints specifying the tools available.
- **"a 3-gallon ... jug"** - An available jug with a capacity of 3 gallons.
- **"5-gallon ... jug"** - An available jug with a capacity of 5 gallons.
- **"4-gallon jug"** - An available jug with a capacity of 4 gallons, matching the desired measurement.

**b. Sequential and Mechanical Process Check:**

- There are no sequences or mechanical processes explicitly described in the problem statement that influence the outcome.

**c. Key Insight Check Results:**

- **Key Insight:** Since there is a **"4-gallon jug"** available, filling it to capacity will **"measure exactly 4 gallons of water"**, which directly satisfies the objective.

**d. Rephrased Question:**

- How can you measure exactly 4 gallons of water using only a 3-gallon jug, a 5-gallon jug, and a 4-gallon jug?

**e. Known and Deduced Information:**

- **Explicit Facts:**
  - **"a 3-gallon ... jug"** - A jug that holds up to 3 gallons.
  - **"5-gallon ... jug"** - A jug that holds up to 5 gallons.
  - **"4-gallon jug"** - A jug that holds up to 4 gallons.
  - **"measure exactly 4 gallons of water"** - The objective.
  - **"with only a 3-gallon, 5-gallon, and 4-gallon jug"** - The tools available.

- **Deduction:**
  - **Filling the "4-gallon jug" to capacity measures exactly 4 gallons of water.**
    - **Logical Basis:** A jug filled to its full capacity holds an amount equal to its capacity.
    - **Based on:** **"4-gallon jug"**, **"measure exactly 4 gallons of water"**.

**f. Problem Decomposition:**

- **Component:** Measuring exactly 4 gallons of water.
  - **Supporting Quote:** **"measure exactly 4 gallons of water"**
- **Subcomponent:** Utilizing available jugs.
  - **Supporting Quote:** **"with only a 3-gallon, 5-gallon, and 4-gallon jug"**
- **Interaction:** Use the **"4-gallon jug"** to measure the required amount.
  - **Deduction:** Filling the **"4-gallon jug"** provides exactly 4 gallons.
  - **Based on:** **"4-gallon jug"**, **"measure exactly 4 gallons of water"**

**g. Graph of Thought:**

- **Nodes:**
  - **Node 1:** 3-gallon jug
  - **Node 2:** 4-gallon jug
  - **Node 3:** 5-gallon jug
  - **Node 4:** Measure exactly 4 gallons of water
- **Edges:**
  - **Edge from Node 2 to Node 4:** Filling the **"4-gallon jug"** achieves the objective.
- **Insights:**
  - Directly filling the **"4-gallon jug"** satisfies the measurement requirement.

**h. Spatial and Object Analysis:**

- **Objects Identified:**
  - **"3-gallon jug"** - Capacity: 3 gallons.
  - **"4-gallon jug"** - Capacity: 4 gallons.
  - **"5-gallon jug"** - Capacity: 5 gallons.
- **Spatial Relationships:**
  - Not explicitly stated.
- **Object Properties:**
  - Each jug holds water up to its stated capacity.
- **Spatial Implications:**
  - Filling the **"4-gallon jug"** to its brim measures exactly 4 gallons of water.

**i. Bayesian Updates:**

- **Current Belief:** Filling the **"4-gallon jug"** provides exactly 4 gallons.
- **Update:** No new information alters this belief; the solution remains certain based on explicit information.

**j. Ethical Check and Decision-Making Under Uncertainty:**

- **Assessment:** No ethical concerns or severe risks are associated with the proposed solution.
- **Decision:** Proceed with filling the **"4-gallon jug"** to measure 4 gallons.

**k. Multiple Solutions:**

1. **Solution 1:**
   - **Action:** Fill the **"4-gallon jug"** to its full capacity.
   - **Supporting Quote:** **"4-gallon jug"**, **"measure exactly 4 gallons of water"**
2. **Solution 2:**
   - **No additional solutions** can be provided based solely on the explicit information without making assumptions beyond what is stated.

**l. Quickest and Easiest Solution:**

- **Fill the "4-gallon jug" to capacity to measure exactly 4 gallons of water.**
  - **Supporting Quote:** **"4-gallon jug"**, **"measure exactly 4 gallons of water"**

**m. Reflection:**

- The proposed solution directly addresses the question using only the explicitly provided information.
- No assumptions or inferences beyond the problem statement were made.
- The conclusion is logically sound and complies with the literal and strict interpretation rules.
- There are no contradictions with the known or deduced information.

---

**Recommendation:**

You can measure exactly 4 gallons of water by simply filling the **"4-gallon jug"** to its full capacity using only the **3-gallon**, **5-gallon**, and **4-gallon** jugs provided.
\end{lstlisting}
\end{quote}

\noindent\hrulefill 

\noindent\textbf{D.27 Zero-Shot OpenAI o1-mini Response - Score: 100\%}
\begin{quote}
\begin{lstlisting}[basicstyle=\ttfamily\small, breaklines=true]
If you have a **3-gallon**, **4-gallon**, and **5-gallon** jug and need to measure exactly **4 gallons** of water, the solution is straightforward:

### **Simply Use the 4-Gallon Jug**

1. **Fill the 4-gallon jug** directly from your water source.
   
   - **Result:** You now have exactly 4 gallons of water in the 4-gallon jug.

### **Alternative Method Using 3-Gallon and 5-Gallon Jugs**

If, for some reason, you prefer not to use the 4-gallon jug directly or want to verify the measurement using the other jugs, here's how you can do it:

1. **Fill the 5-gallon jug** completely.
   
   - **5-gallon jug:** 5 gallons
   - **3-gallon jug:** 0 gallons
   - **4-gallon jug:** 0 gallons

2. **Pour water from the 5-gallon jug into the 3-gallon jug** until the 3-gallon jug is full.
   
   - **5-gallon jug:** 2 gallons (5 - 3)
   - **3-gallon jug:** 3 gallons
   - **4-gallon jug:** 0 gallons

3. **Empty the 3-gallon jug** back into the water source or another container.
   
   - **5-gallon jug:** 2 gallons
   - **3-gallon jug:** 0 gallons
   - **4-gallon jug:** 0 gallons

4. **Pour the remaining 2 gallons from the 5-gallon jug into the 3-gallon jug.**
   
   - **5-gallon jug:** 0 gallons
   - **3-gallon jug:** 2 gallons
   - **4-gallon jug:** 0 gallons

5. **Fill the 5-gallon jug again** to its full capacity.
   
   - **5-gallon jug:** 5 gallons
   - **3-gallon jug:** 2 gallons
   - **4-gallon jug:** 0 gallons

6. **Pour water from the 5-gallon jug into the 3-gallon jug** until the 3-gallon jug is full. Since the 3-gallon jug already has 2 gallons, it will take only 1 more gallon.
   
   - **5-gallon jug:** 4 gallons (5 - 1)
   - **3-gallon jug:** 3 gallons
   - **4-gallon jug:** 0 gallons

Now, you have exactly **4 gallons** of water remaining in the **5-gallon jug**.

### **Summary**

- **Direct Method:** Use the 4-gallon jug to measure exactly 4 gallons.
- **Alternative Method:** Utilize the 3-gallon and 5-gallon jugs through a series of pours to arrive at 4 gallons in the 5-gallon jug.

Choose the method that best fits your situation or constraints!
\end{lstlisting}
\end{quote}

\noindent\hrulefill 

\noindent\textbf{D.27 REAP OpenAI o1-mini Response - Score: 100\%}
\begin{quote}
\begin{lstlisting}[basicstyle=\ttfamily\small, breaklines=true]
# REAP Results:

### a. **Comprehensive Feature List** (with explicit quotes)

- **Objects Mentioned:**
  - "3-gallon jug"
  - "5-gallon jug"
  - "4-gallon jug"

- **Actors or Entities Involved:**
  - "you" (implied as the person performing the measurement)

- **Actions Described:**
  - "measure exactly 4 gallons of water"

- **Directional, Spatial, Relational, and Sequential Phrases:**
  - N/A (No explicit directional or spatial relationships mentioned)

- **Conditions or Constraints:**
  - "only a 3-gallon, 5-gallon, and 4-gallon jug"

- **Numerical Values or Quantities:**
  - "3-gallon"
  - "5-gallon"
  - "4-gallon"
  - "exactly 4 gallons"

- **Temporal Information:**
  - N/A (No temporal details provided)

- **Spatial Information:**
  - N/A (No specific locations or directions mentioned)

- **Causal Relationships:**
  - N/A (No causal statements provided)

- **Stated Goals or Objectives:**
  - "measure exactly 4 gallons of water"

### b. **Sequential and Mechanical Process Check**

- **Sequence of Events or Steps:**
  - The problem requires determining a sequence of actions to measure exactly 4 gallons using the given jugs.

- **Influence on Outcome:**
  - The order in which the jugs are filled, emptied, and transferred directly affects the ability to achieve the 4-gallon measurement.

- **Locked-In Outcomes:**
  - Certain sequences will lead to the desired 4-gallon measurement, while others may not.

- **Logical Flow:**
  - Each step (filling, transferring, emptying) must logically lead towards isolating exactly 4 gallons.

### c. **Key Insight Check Results**

- **Key Insight Identified:**
  - Having a 4-gallon jug allows for the direct measurement of 4 gallons without the need for complex transfers.

- **Explanation:**
  - Since a 4-gallon jug is available, filling it to capacity immediately provides the required 4 gallons.

- **Immediate Answer:**
  - Fill the 4-gallon jug to its full capacity.

### d. **Rephrased Question:**
How can exactly 4 gallons of water be measured using only jugs of 3, 4, and 5 gallons?

### e. **Known and Deduced Information**

**Explicit Facts:**
- "3-gallon jug"
- "5-gallon jug"
- "4-gallon jug"
- "measure exactly 4 gallons of water"
- "only a 3-gallon, 5-gallon, and 4-gallon jug"

**Feature-Based Deduction:**
- **Deduction:** The 4-gallon jug can be used to directly measure 4 gallons of water.
  - **Logical Basis:** The problem includes a 4-gallon jug and asks to measure exactly 4 gallons.
  - **Based on:** "4-gallon jug", "measure exactly 4 gallons of water"

### f. **Problem Decomposition**

- **Component/Subcomponent:** Available Jugs
  - **Supporting Quote(s):** "3-gallon jug", "5-gallon jug", "4-gallon jug"
  - **Deduction:** These are the only tools available for measuring water.

- **Component/Subcomponent:** Measurement Goal
  - **Supporting Quote(s):** "measure exactly 4 gallons of water"
  - **Deduction:** The objective is to obtain precisely 4 gallons.

- **Component/Subcomponent:** Method of Measurement
  - **Supporting Quote(s):** "only a 3-gallon, 5-gallon, and 4-gallon jug"
  - **Deduction:** Measurement must be achieved using these three jugs exclusively.

### g. **Graph of Thought** (describe connections & insights)

- **Nodes:**
  - 3-gallon jug
  - 4-gallon jug
  - 5-gallon jug
  - Measurement goal (4 gallons)

- **Edges:**
  - 4-gallon jug directly connects to measurement goal, indicating a straightforward solution.
  - 3-gallon and 5-gallon jugs are present but not necessary for achieving the goal when a 4-gallon jug is available.

- **Insights:**
  - The presence of a 4-gallon jug simplifies the problem, making the use of the other jugs redundant for this specific measurement.

### h. **Spatial and Object Analysis**

- **Objects Identified:**
  - 3-gallon jug
  - 5-gallon jug
  - 4-gallon jug

- **Spatial Relationships:**
  - No specific spatial arrangement is provided.

- **Movement and Dynamics:**
  - Water can be filled into and poured between these jugs, but since the 4-gallon jug matches the measurement goal, direct filling is sufficient.

- **Constraints:**
  - Only the three specified jugs can be used.

- **Visualization:**
  - Imagine three separate jugs with their respective capacities. Filling the 4-gallon jug to its brim yields exactly 4 gallons.

- **Spatial Implications:**
  - Direct access to the 4-gallon capacity eliminates the need for transferring water between jugs.

### i. **Bayesian Updates**

- **Initial Belief:** Complex transfers between jugs are needed to measure 4 gallons.
- **New Information:** A 4-gallon jug is available.
- **Updated Belief:** The measurement can be achieved simply by filling the 4-gallon jug.

### j. **Ethical Check and Decision-Making Under Uncertainty**

- **Assessment of Information:** The problem is mathematical with no ethical implications.
- **Risks Identified:** None.
- **Decision-Making:** Proceed with the simplest and direct solution using the available jug.

### k. **Multiple Solutions**

1. **Solution 1: Direct Use of 4-Gallon Jug**
   - **Action:** Fill the 4-gallon jug to its full capacity.
   - **Supporting Information:** "4-gallon jug", "measure exactly 4 gallons of water"
   - **Effect:** Directly achieves the measurement without additional steps.

2. **Solution 2: Using 5-Gallon and 3-Gallon Jugs**
   - **Action:**
     1. Fill the 5-gallon jug.
     2. Pour water from the 5-gallon jug into the 3-gallon jug until the 3-gallon jug is full.
     3. This leaves exactly 2 gallons in the 5-gallon jug.
     4. Empty the 3-gallon jug and pour the remaining 2 gallons from the 5-gallon jug into the 3-gallon jug.
     5. Fill the 5-gallon jug again.
     6. Pour water from the 5-gallon jug into the 3-gallon jug until it is full.
     7. This results in exactly 4 gallons in the 5-gallon jug.
   - **Supporting Information:** "5-gallon jug", "3-gallon jug"

3. **Solution 3: Using 4-Gallon and 3-Gallon Jugs**
   - **Action:** Fill the 4-gallon jug and transfer to the 3-gallon jug until the 3-gallon jug is full, leaving exactly 1 gallon in the 4-gallon jug, then empty the 3-gallon jug and transfer the remaining 1 gallon back to the 3-gallon jug. Finally, fill the 4-gallon jug again.
   - **Supporting Information:** "4-gallon jug", "3-gallon jug"

**Note:** While multiple solutions exist, Solution 1 is the most straightforward given the available jugs.

### l. **Quickest and Easiest Solution**

**Solution:** Fill the 4-gallon jug to its full capacity.

**Reasoning:**
- The problem includes a "4-gallon jug" and the goal is to "measure exactly 4 gallons of water."
- No additional steps are required, making this the simplest and most direct method.

### m. **Reflection**

- **Contradictions:** No contradictions identified; the 4-gallon jug directly fulfills the measurement requirement.
- **Logical Soundness:** The solution is logically sound as it adheres strictly to the given information without making unnecessary assumptions.
- **Adherence to Instructions:** The analysis sticks to the explicit information provided, avoiding inferences beyond what is stated.
- **Completeness:** The problem is fully addressed with a clear and direct solution.
- **Relevance:** All parts of the analysis directly contribute to solving the measurement problem.

# Recommendation:

**Recommendation:**

To measure exactly 4 gallons of water, simply use the 4-gallon jug by filling it to its full capacity. This approach directly utilizes the available equipment without the need for additional steps or transfers, ensuring the measurement is achieved efficiently and accurately.
\end{lstlisting}
\end{quote}

\noindent\hrulefill 

\noindent\textbf{D.27 Zero-Shot OpenAI GPT-4o Response - Score: 0\%}
\begin{quote}
\begin{lstlisting}[basicstyle=\ttfamily\small, breaklines=true]
To measure exactly 4 gallons of water using a 3-gallon jug, a 5-gallon jug, and a 4-gallon jug, you can follow these steps:

1. **Fill the 5-gallon jug completely** to its 5-gallon capacity.
2. **Pour water from the 5-gallon jug into the 4-gallon jug** until the 4-gallon jug is full. This leaves you with 1 gallon of water in the 5-gallon jug.
3. **Empty the 4-gallon jug** to free it up for more water.
4. **Pour the remaining 1 gallon from the 5-gallon jug into the 4-gallon jug**. Now, the 4-gallon jug has 1 gallon in it.
5. **Fill the 5-gallon jug again** to its 5-gallon capacity.
6. **Pour water from the 5-gallon jug into the 4-gallon jug** until the 4-gallon jug is full. Since the 4-gallon jug already contains 1 gallon, you'll be able to pour 3 more gallons into it, which empties exactly 3 gallons from the 5-gallon jug.
7. The remaining water in the 5-gallon jug is exactly 2 gallons.
8. **Empty the 4-gallon jug again**.
9. **Pour the 2 gallons from the 5-gallon jug into the 4-gallon jug**.
10. **Fill the 5-gallon jug completely again**.
11. **Pour water from the 5-gallon jug into the 4-gallon jug** until the 4-gallon jug is full. Since the 4-gallon jug already contains 2 gallons, you'll only be able to pour 2 more gallons into it, which leaves you with exactly 3 gallons in the 5-gallon jug.

Now, you have 3 gallons in the 5-gallon jug and 4 gallons in the 4-gallon jug, and you are left with an empty 3-gallon jug.
\end{lstlisting}
\end{quote}
\noindent\hrulefill

\noindent\textbf{D.27 REAP OpenAI GPT-4o Response - Score: 100\%}
\begin{quote}
\begin{lstlisting}[basicstyle=\ttfamily\small, breaklines=true]
# REAP Results:

### a. **Comprehensive Feature List** (with explicit quotes)

1. "3-gallon jug" - An object that can hold exactly 3 gallons of water.
2. "5-gallon jug" - An object that can hold exactly 5 gallons of water.
3. "4-gallon jug" - An object that can hold exactly 4 gallons of water.
4. "Measure exactly 4 gallons of water" - The goal is to obtain exactly 4 gallons of water.

### b. **Sequential and Mechanical Process Check**

- There are no explicitly described sequences in the problem, but any sequence involving the transfer of water between the jugs must result in exactly 4 gallons being measured.
- The key is to use the jugs' different capacities to transfer water back and forth until exactly 4 gallons remain in one of the jugs.

### c. **Key Insight Check Results**

The key insight is the existence of a 4-gallon jug. Since the goal is to measure exactly 4 gallons, having a 4-gallon jug means that filling this jug completely will meet the goal.

### d. **Rephrased Question:**

How can you fill the 4-gallon jug with exactly 4 gallons of water using the 3-gallon, 5-gallon, and 4-gallon jugs?

### e. **Known and Deduced Information**

1. "[3-gallon jug]" - Holds exactly 3 gallons.
2. "[5-gallon jug]" - Holds exactly 5 gallons.
3. "[4-gallon jug]" - Holds exactly 4 gallons.

**Deduction**: The 4-gallon jug, when full, contains exactly 4 gallons, which satisfies the goal.

### f. **Problem Decomposition**

- **Component:** Measuring exactly 4 gallons of water.
  - **Supporting Quote(s):** "Measure exactly 4 gallons of water."
- **Component:** Using the 4-gallon jug.
  - **Supporting Quote(s):** "4-gallon jug."
- **Component:** Using the 3-gallon jug and the 5-gallon jug.
  - **Supporting Quote(s):** "3-gallon jug," "5-gallon jug."

### g. **Graph of Thought** 

**Nodes:**
1. 3-gallon jug
2. 5-gallon jug
3. 4-gallon jug
4. Measure 4 gallons

**Connections:**
- The 4-gallon jug directly connects to the goal because filling it achieves the objective.
- The 3-gallon and 5-gallon jugs serve as tools to help manipulate the amount of water.

### h. **Spatial and Object Analysis**

- **Objects:** 3-gallon jug, 5-gallon jug, 4-gallon jug.
- **Spatial Relationships:** Water can be transferred between the jugs, filling them completely or partially.
- **Movements:** Water can be poured between the jugs to achieve the desired quantity.
- **Constraints:** The jugs cannot hold more than their stated capacities.

### i. **Bayesian Updates** 

No new information is introduced, so there is no need for Bayesian updates.

### j. **Ethical Check and Decision-Making Under Uncertainty**

No ethical issues or significant risks are associated with the problem, so no special considerations are required.

### k. **Multiple Solutions**

1. **Solution 1 (Using only the 4-gallon jug):**
   - Fill the 4-gallon jug directly to achieve exactly 4 gallons.
   - **Supporting Quotes:** "4-gallon jug," "Measure exactly 4 gallons of water."

2. **Solution 2 (Using all three jugs):**
   - Fill the 5-gallon jug completely.
   - Pour from the 5-gallon jug into the 3-gallon jug, leaving exactly 2 gallons in the 5-gallon jug.
   - Pour the 2 gallons from the 5-gallon jug into the empty 4-gallon jug, then fill the 5-gallon jug again.
   - Pour water from the 5-gallon jug into the 4-gallon jug until it is full, leaving exactly 3 gallons in the 5-gallon jug.
   - The 4-gallon jug now has exactly 4 gallons of water.
   - **Supporting Quotes:** "3-gallon jug," "5-gallon jug," "4-gallon jug," "Measure exactly 4 gallons of water."

3. **Solution 3 (Using just the 5-gallon and 3-gallon jugs):**
   - Fill the 5-gallon jug completely.
   - Pour water into the 3-gallon jug until it is full, leaving 2 gallons in the 5-gallon jug.
   - Empty the 3-gallon jug.
   - Pour the remaining 2 gallons from the 5-gallon jug into the 3-gallon jug.
   - Fill the 5-gallon jug again.
   - Pour from the 5-gallon jug into the 3-gallon jug until the 3-gallon jug is full, leaving exactly 4 gallons in the 5-gallon jug.
   - **Supporting Quotes:** "3-gallon jug," "5-gallon jug."

### l. **Quickest and Easiest Solution**

The quickest and easiest solution is to fill the 4-gallon jug directly, as it holds exactly 4 gallons of water.

### m. **Reflection**

The problem can be solved directly by filling the 4-gallon jug, which is the simplest and most straightforward solution. The more complex solutions involving the 3-gallon and 5-gallon jugs are unnecessary but can be considered as alternative methods.

**Recommendation:**

Fill the 4-gallon jug directly to measure exactly 4 gallons of water. This solution is the most straightforward and adheres strictly to the explicit information provided.
\end{lstlisting}
\end{quote}
\noindent\hrulefill

\noindent\textbf{D.27 Zero-Shot OpenAI GPT-4o-mini Response - Score: 0\%}
\begin{quote}
\begin{lstlisting}[basicstyle=\ttfamily\small, breaklines=true]
To measure exactly 4 gallons of water using a 3-gallon, 5-gallon, and 4-gallon jug, follow these steps:

1. **Fill the 5-gallon jug completely.**  
   Now you have 5 gallons in the 5-gallon jug, 0 gallons in the 4-gallon jug, and 0 gallons in the 3-gallon jug.

2. **Pour water from the 5-gallon jug into the 4-gallon jug until the 4-gallon jug is full.**  
   You will pour 4 gallons into the 4-gallon jug, leaving you with 1 gallon remaining in the 5-gallon jug.  
   Now you have 1 gallon in the 5-gallon jug, 4 gallons in the 4-gallon jug, and 0 gallons in the 3-gallon jug.

3. **Empty the 4-gallon jug.**  
   Pour out the water from the 4-gallon jug, so you now have 0 gallons in the 4-gallon jug.  
   You still have 1 gallon in the 5-gallon jug.

4. **Pour the 1 gallon of water from the 5-gallon jug into the 4-gallon jug.**  
   Now you have 0 gallons in the 5-gallon jug, 1 gallon in the 4-gallon jug, and 0 gallons in the 3-gallon jug.

5. **Fill the 5-gallon jug completely again.**  
   Now you have 5 gallons in the 5-gallon jug, 1 gallon in the 4-gallon jug, and 0 gallons in the 3-gallon jug.

6. **Pour water from the 5-gallon jug into the 4-gallon jug until the 4-gallon jug is full.**  
   The 4-gallon jug already has 1 gallon in it, so you can only add 3 more gallons to fill it up.  
   Pour 3 gallons from the 5-gallon jug into the 4-gallon jug.  
   Now you have 2 gallons remaining in the 5-gallon jug, and the 4-gallon jug is full (4 gallons).

You have now successfully measured exactly 4 gallons of water in the 4-gallon jug.
\end{lstlisting}
\end{quote}
\noindent\hrulefill

\noindent\textbf{D.27 REAP OpenAI GPT-4o-mini Response - Score: 100\%}
\begin{quote}
\begin{lstlisting}[basicstyle=\ttfamily\small, breaklines=true]
# REAP Results:

**a. Comprehensive Feature List**

1. **Objects Mentioned:**
   - **3-gallon jug**: "[3-gallon jug]"
   - **5-gallon jug**: "[5-gallon jug]"
   - **4-gallon jug**: "[4-gallon jug]"

2. **Actions Described:**
   - **Measuring water**: The task is to measure exactly 4 gallons of water.

3. **Directional, Spatial, Relational, and Sequential Phrases:**
   - No specific spatial or directional instructions provided in the problem statement.

4. **Stated Conditions or Constraints:**
   - **Jugs' capacities**: 3 gallons, 5 gallons, and 4 gallons.

5. **Numerical Values or Quantities Mentioned:**
   - **Capacity of jugs**: 3 gallons, 5 gallons, 4 gallons.
   - **Amount to measure**: 4 gallons.

6. **Temporal Information:**
   - No time-related information provided.

7. **Causal Relationships:**
   - No explicit causal relationships described.

8. **Stated Goals or Objectives:**
   - Measure exactly 4 gallons of water.

**b. Sequential and Mechanical Process Check**

- The task involves measuring an exact quantity of water using given jugs. Since no specific sequence of actions is described, we need to determine a valid sequence to achieve the goal.
- The process should consider how to achieve exactly 4 gallons using the 3-gallon, 5-gallon, and 4-gallon jugs.

**Key Questions:**
  - Is there a specific sequence that guarantees measuring exactly 4 gallons?
  - How does each action affect the state of the jugs?

**c. Key Insight Check Results**

- The problem is about measuring exactly 4 gallons using given jug capacities.
- The 4-gallon jug already matches the target quantity, suggesting that the solution may be directly using the 4-gallon jug.

**d. Rephrased Question:**
How can you use a 3-gallon jug, a 5-gallon jug, and a 4-gallon jug to measure exactly 4 gallons of water?

**e. Known and Deduced Information**

1. **Known Information:**
   - "3-gallon jug"
   - "5-gallon jug"
   - "4-gallon jug"
   - "Measure exactly 4 gallons of water"

2. **Deduction:**
   - Deduction: You can use the 4-gallon jug directly to measure exactly 4 gallons.
   - Logical Basis: The capacity of the 4-gallon jug matches the amount you need to measure.
   - Based on: "4-gallon jug" and the goal to measure "exactly 4 gallons of water".

**f. Problem Decomposition**

1. **Component/Subcomponent:**
   - **Direct Measurement**: Use the 4-gallon jug to measure exactly 4 gallons.
   - **Supporting Quote(s):** "[4-gallon jug]" and "Measure exactly 4 gallons of water."

**g. Graph of Thought**

1. **Nodes:**
   - **3-gallon jug**
   - **5-gallon jug**
   - **4-gallon jug**
   - **Goal: Measure 4 gallons**

2. **Connections:**
   - **4-gallon jug** directly satisfies the goal of measuring 4 gallons.
   - **3-gallon jug** and **5-gallon jug** are not needed for this specific task as the 4-gallon jug already meets the requirement.

**h. Spatial and Object Analysis**

1. **Objects:**
   - **3-gallon jug**
   - **5-gallon jug**
   - **4-gallon jug**

2. **Spatial Relationships:**
   - Not explicitly provided.

3. **Movement and Dynamics:**
   - The 4-gallon jug contains exactly 4 gallons, so no additional movement or transfer is required.

4. **Constraints:**
   - The problem does not provide constraints that affect the direct measurement.

**i. Bayesian Updates**

- No additional information is implied; thus, no Bayesian updates are needed. The use of the 4-gallon jug is directly valid based on explicit information.

**j. Ethical Check and Decision-Making Under Uncertainty**

- There are no ethical issues or severe risks associated with this problem. The solution involves using the 4-gallon jug directly.

**k. Multiple Solutions**

1. **Solution 1:**
   - **Use the 4-gallon jug directly**. This is the simplest solution.
   - **Supporting Information:** The 4-gallon jug has the exact capacity needed.

2. **Solution 2:**
   - **Using the 3-gallon and 5-gallon jug together**: This is unnecessary as the 4-gallon jug alone achieves the goal.

**l. Quickest and Easiest Solution**

- **Use the 4-gallon jug directly** to measure exactly 4 gallons. This is the simplest and most straightforward solution based on the given information.

**m. Reflection**

- The solution to measure exactly 4 gallons of water is straightforward: use the 4-gallon jug directly. The problem does not require the use of other jugs, as the 4-gallon jug already provides the exact measurement needed.

**Recommendation:**

* Use the 4-gallon jug to measure exactly 4 gallons of water. This solution directly meets the problem's requirements with the simplest approach, as explicitly stated in the problem.
\end{lstlisting}
\end{quote}
\noindent\hrulefill

\noindent\textbf{D.27 Zero-Shot Google Gemini 1.5 Pro Response - Score: 0\%}
\begin{quote}
\begin{lstlisting}[basicstyle=\ttfamily\small, breaklines=true]
Here's how to measure exactly 4 gallons of water using only a 3-gallon, 5-gallon, and 4-gallon jug:

1. **Fill the 5-gallon jug completely.**
2. **Pour the water from the 5-gallon jug into the 3-gallon jug until it's full.** This leaves 2 gallons in the 5-gallon jug.
3. **Empty the 3-gallon jug.**
4. **Pour the 2 gallons from the 5-gallon jug into the 3-gallon jug.**
5. **Fill the 5-gallon jug completely again.**
6. **Carefully pour water from the 5-gallon jug into the 3-gallon jug (which already has 2 gallons in it) until the 3-gallon jug is full.**  This will use 1 gallon from the 5-gallon jug, leaving exactly 4 gallons in the 5-gallon jug.

**You now have exactly 4 gallons of water in the 5-gallon jug.**
\end{lstlisting}
\end{quote}
\noindent\hrulefill

\noindent\textbf{D.27 REAP Google Gemini 1.5 Pro Response - Score: 100\%}
\begin{quote}
\begin{lstlisting}[basicstyle=\ttfamily\small, breaklines=true]
# REAP Results:

a. **Comprehensive Feature List** (with explicit quotes)

* "measure exactly 4 gallons of water" 
* "3-gallon jug"
* "5-gallon jug"
* "4-gallon jug" 

b. **Sequential and Mechanical Process Check**

The problem involves a sequence of filling, emptying, and transferring water between jugs to reach the desired 4-gallon measurement. The specific steps and their order will determine the final outcome. 

c. **Key Insight Check Results**

* The presence of a 4-gallon jug immediately suggests a potential direct solution: simply fill the 4-gallon jug completely. However, the problem might be testing our ability to utilize all the given jugs.

d. **Rephrased Question:** (The core question in its simplest form)

What steps should be taken to have exactly 4 gallons of water in one of the jugs?

e. **Known and Deduced Information**

* Explicit Facts: 
    * "3-gallon jug" 
    * "5-gallon jug"
    * "4-gallon jug"

* Feature-Based Deduction:
    * We can fill any of the jugs completely.
    * We can empty any of the jugs completely.
    * We can transfer water from one jug to another until either the first jug is empty or the second jug is full.

f. **Problem Decomposition**

* Component: Measuring exactly 4 gallons
    * Supporting Quote(s): "measure exactly 4 gallons of water"
* Component: 3-gallon jug
    * Supporting Quote(s): "3-gallon jug"
* Component: 5-gallon jug
    * Supporting Quote(s): "5-gallon jug"
* Component: 4-gallon jug
    * Supporting Quote(s): "4-gallon jug"
* Interaction: Filling jugs
    * Supporting Quote(s): Implied by the need to "measure exactly 4 gallons of water"
* Interaction: Emptying jugs
    * Supporting Quote(s): Implied by the need to "measure exactly 4 gallons of water"
* Interaction: Transferring water between jugs
    * Supporting Quote(s): Implied by the need to "measure exactly 4 gallons of water"

g. **Graph of Thought** (describe connections & insights)

* Key Concepts:
    * 3-gallon jug
    * 5-gallon jug
    * 4-gallon jug
    * Target measurement (4 gallons)
* Relationships:
    * Jugs can be filled, emptied, and used to transfer water to each other
    * The target measurement needs to be achieved using these jugs and actions
* Insights:
    * The 4-gallon jug provides a direct solution, but other solutions might exist utilizing all jugs
    * We need to find a sequence of filling, emptying, and transferring that results in exactly 4 gallons in one of the jugs

h. **Spatial and Object Analysis**

* Objects:
    * 3-gallon jug
    * 5-gallon jug
    * 4-gallon jug
* Spatial Relationships:
    * The jugs are separate entities
* Movement and Dynamics:
    * Water can be transferred between the jugs
* Constraints:
    * The jugs have fixed capacities (3, 5, and 4 gallons)
* Object Properties:
    * Each jug has a specific capacity

i. **Bayesian Updates** 

No Bayesian updates are applicable as there is no new information provided throughout the problem-solving process

j. **Ethical Check and Decision-Making Under Uncertainty**

There are no ethical concerns or uncertainties in this problem.

k. **Multiple Solutions**

* **Solution 1: Direct Solution**
    * Fill the 4-gallon jug completely.
    * This directly achieves the goal of having exactly 4 gallons of water.
* **Solution 2: Using all jugs**
    * Fill the 5-gallon jug completely
    * Pour water from the 5-gallon jug into the 3-gallon jug until it's full. This leaves 2 gallons in the 5-gallon jug
    * Empty the 3-gallon jug
    * Pour the 2 gallons from the 5-gallon jug into the 3-gallon jug
    * Fill the 5-gallon jug completely again
    * Carefully pour water from the 5-gallon jug into the 3-gallon jug (which already has 2 gallons) until the 3-gallon jug is full. This will use 1 gallon from the 5-gallon jug, leaving exactly 4 gallons in the 5-gallon jug

l. **Quickest and Easiest Solution**

The quickest and easiest solution is to fill the 4-gallon jug completely. This directly achieves the goal without any additional steps

m. **Reflection**

The analysis directly addresses the question of how to measure exactly 4 gallons of water. All conclusions are based solely on the explicit information provided in the problem statement. No assumptions or inferences were made beyond what was directly stated

**Recommendation:**

Fill the 4-gallon jug completely to measure exactly 4 gallons of water. This is the most straightforward and efficient solution based on the given information.
\end{lstlisting}
\end{quote}
\noindent\hrulefill

\noindent\textbf{D.27 Zero-Shot Claude 3.5 Sonnet response - Score: 0\%}
\begin{quote}
\begin{lstlisting}[basicstyle=\ttfamily\small, breaklines=true]
Let's think through this step-by-step:

1. First, we need to realize that we don't actually need to use the 4-gallon jug to measure 4 gallons, as that would make the problem trivial. We'll focus on using the 3-gallon and 5-gallon jugs to measure 4 gallons.

2. Let's start by filling the 5-gallon jug completely.

3. From the full 5-gallon jug, we'll pour water into the 3-gallon jug until it's full. This leaves 2 gallons in the 5-gallon jug.

4. Empty the 3-gallon jug.

5. Pour the remaining 2 gallons from the 5-gallon jug into the empty 3-gallon jug.

6. Fill the 5-gallon jug again.

7. Carefully pour from the 5-gallon jug into the 3-gallon jug, which already has 2 gallons in it. You'll only be able to add 1 gallon before the 3-gallon jug is full.

8. After this pour, the 5-gallon jug will have exactly 4 gallons left in it.

So the sequence is:
1. Fill 5-gallon jug (5 gallons)
2. Pour from 5-gallon to 3-gallon until 3-gallon is full (2 gallons in 5-gallon, 3 gallons in 3-gallon)
3. Empty 3-gallon jug (2 gallons in 5-gallon, 0 gallons in 3-gallon)
4. Pour 2 gallons from 5-gallon to 3-gallon (0 gallons in 5-gallon, 2 gallons in 3-gallon)
5. Fill 5-gallon jug again (5 gallons in 5-gallon, 2 gallons in 3-gallon)
6. Pour from 5-gallon to 3-gallon until full (4 gallons in 5-gallon, 3 gallons in 3-gallon)

Now you have exactly 4 gallons in the 5-gallon jug.
\end{lstlisting}
\end{quote}
\noindent\hrulefill

\noindent\textbf{D.27 REAP Claude 3.5 Sonnet Response - Score: 100\%}
\begin{quote}
\begin{lstlisting}[basicstyle=\ttfamily\small, breaklines=true]
# REAP Results:

a. Comprehensive Feature List (with explicit quotes):
- "3-gallon" jug
- "5-gallon" jug
- "4-gallon" jug
- "measure exactly 4 gallons of water"

b. Sequential and Mechanical Process Check:
No explicit sequential or mechanical process is described in the problem statement.

c. Key Insight Check Results:
A key insight is that we already have a 4-gallon jug mentioned in the problem statement. This could potentially lead to an immediate solution.

d. Rephrased Question: How to measure 4 gallons of water using given jugs?

e. Known and Deduced Information:
Explicit Facts:
"3-gallon" - A jug with a capacity of 3 gallons
"5-gallon" - A jug with a capacity of 5 gallons
"4-gallon" - A jug with a capacity of 4 gallons
"measure exactly 4 gallons of water" - The goal of the problem

Feature-Based Deduction:
Deduction: The 4-gallon jug can be used to measure exactly 4 gallons of water.
Logical Basis: The problem asks to measure exactly 4 gallons, and we have a jug with exactly that capacity.
Based on: "measure exactly 4 gallons of water", "4-gallon" jug

f. Problem Decomposition:
Component: Measuring device
Supporting Quote: "3-gallon, 5-gallon, and 4-gallon jug"
Component: Measurement goal
Supporting Quote: "measure exactly 4 gallons of water"

g. Graph of Thought:
Nodes:
1. 3-gallon jug
2. 5-gallon jug
3. 4-gallon jug
4. 4 gallons of water (goal)

Edges:
3-gallon jug <-> 4 gallons of water (goal)
5-gallon jug <-> 4 gallons of water (goal)
4-gallon jug <-> 4 gallons of water (goal)

The 4-gallon jug has a direct connection to the goal.

h. Spatial and Object Analysis:
Objects: 3-gallon jug, 5-gallon jug, 4-gallon jug
Properties: Each jug has a specific capacity as stated
No other spatial relationships or movements are explicitly mentioned in the problem statement.

i. Bayesian Updates:
The presence of a 4-gallon jug strongly suggests a straightforward solution.

j. Ethical Check and Decision-Making Under Uncertainty:
No significant ethical concerns or severe risks are apparent in this problem. The task involves measuring water, which does not present any obvious dangers.

k. Multiple Solutions:
1. Use the 4-gallon jug to measure exactly 4 gallons of water.
   Based on: "4-gallon" jug, "measure exactly 4 gallons of water"

2. Fill the 5-gallon jug completely, then pour water from it into the 3-gallon jug until the 3-gallon jug is full. The remaining water in the 5-gallon jug will be 2 gallons. Empty the 3-gallon jug and pour the 2 gallons from the 5-gallon jug into it. Fill the 5-gallon jug again and pour water from it into the 3-gallon jug until it's full (which requires 1 gallon). The 5-gallon jug will now contain exactly 4 gallons.
   Based on: "3-gallon", "5-gallon" jugs, "measure exactly 4 gallons of water"

3. Fill the 3-gallon jug, pour it into the 5-gallon jug. Repeat this process, filling the 3-gallon jug again and pouring it into the 5-gallon jug until the 5-gallon jug is full (which happens after 1 gallon from the second fill of the 3-gallon jug). The 3-gallon jug will now contain exactly 2 gallons. Empty the 5-gallon jug, pour the 2 gallons from the 3-gallon jug into the 5-gallon jug, then fill the 3-gallon jug once more and pour it into the 5-gallon jug. The 5-gallon jug will now contain exactly 4 gallons.
   Based on: "3-gallon", "5-gallon" jugs, "measure exactly 4 gallons of water"

l. Quickest and Easiest Solution:
The simplest solution is to use the 4-gallon jug to measure exactly 4 gallons of water. This directly answers the question using the explicit information given in the problem statement.

m. Reflection:
The analysis adheres strictly to the information provided in the problem statement. No assumptions beyond the given information were made. The presence of the 4-gallon jug makes the solution straightforward, although the problem might have been intended to be solved without using this jug. However, since the problem doesn't explicitly state that the 4-gallon jug can't be used, using it is the simplest solution based on the given information.

Recommendation:
Based on the explicit information provided and adhering to the principle of choosing the simplest solution, I recommend using the 4-gallon jug to measure exactly 4 gallons of water. This solution directly answers the question asked, uses only the tools mentioned in the problem, and doesn't require any complex steps or calculations. It's also the safest and most straightforward approach, minimizing any potential for error or misunderstanding.

If there's any additional constraint or information not mentioned in the original problem statement (such as not being allowed to use the 4-gallon jug directly), this recommendation would need to be revised. In that case, either of the alternative solutions using only the 3-gallon and 5-gallon jugs could be employed.
\end{lstlisting}
\end{quote}
\noindent\hrulefill

\label{subsec:question28}
\subsection{\textbf{\large Question 28: A 2kg tree grows in a planted pot with 10kg of soil. When the tree grows to 3kg, how much soil is left?}}
\label{sec:question28}

\noindent\hrulefill 

\noindent\textbf{Human Response - Score 100\%:}
\begin{quote}
There is still 10kg of soil. Trees are primarily composed of carbon dioxide and water, not soil.
\end{quote}

\noindent\hrulefill 

\noindent\textbf{D.28 Zero-Shot OpenAI o1-preview Response - Score: 100\%}
\begin{quote}
\begin{lstlisting}[basicstyle=\ttfamily\small, breaklines=true]
When a tree grows, its increase in mass comes primarily from carbon dioxide absorbed from the air during photosynthesis, and water taken up from the soil. The minerals and nutrients that plants absorb from the soil contribute only a tiny fraction to their overall mass.

In the 17th century, Van Helmont conducted a famous experiment where he grew a willow tree in a pot of soil. After several years, he found that the tree had gained a significant amount of mass, but the soil's mass had changed very little. This demonstrated that the tree's mass did not come from the soil.

Therefore, when the tree grows from 2kg to 3kg, the soil loses only a negligible amount of mass due to nutrient uptake. The soil's mass remains approximately the same.

**Answer:** Approximately 10kg-the soil's mass remains nearly unchanged as the tree grows.
\end{lstlisting}
\end{quote}
\noindent\hrulefill 

\noindent\textbf{D.28 REAP OpenAI o1-preview Response - Score: 100\%}
\begin{quote}
\begin{lstlisting}[basicstyle=\ttfamily\small, breaklines=true]
# REAP Results:

**a. Comprehensive Feature List (with explicit quotes)**

1. **"A 2kg tree grows in a planted pot with 10kg of soil."**
   - Objects:
     - "2kg tree"
     - "planted pot"
     - "10kg of soil"
   - Actions:
     - "grows"
   - Spatial/Relational Phrases:
     - "in a planted pot with 10kg of soil"

2. **"When the tree grows to 3kg, how much soil is left?"**
   - Objects:
     - "tree"
     - "soil"
   - Actions:
     - "grows to 3kg"
   - Temporal Information:
     - "When the tree grows to 3kg"
   - Goal/Objective:
     - "how much soil is left?"

---

**b. Sequential and Mechanical Process Check**

- **Sequence of Events:**
  1. **Initial State:** "A 2kg tree grows in a planted pot with 10kg of soil."
  2. **Growth Event:** "When the tree grows to 3kg"
  3. **Question:** "how much soil is left?"

- **Analysis:**
  - There is no explicit description of a mechanical process affecting the soil's mass.
  - No steps indicate that soil mass changes during the tree's growth.

---

**c. Key Insight Check Results**

- **Key Insight:** The problem does not explicitly state that the soil's mass decreases when the tree grows.
- **Conclusion:** Without explicit information about soil mass changing, we conclude that the soil remains at **10kg**.

---

**d. Rephrased Question**

- **Simplified Core Question:** When a 2kg tree grows to 3kg in a pot containing 10kg of soil, what is the remaining mass of the soil?

---

**e. Known and Deduced Information**

- **Explicit Facts:**
  1. **"A 2kg tree grows in a planted pot with 10kg of soil."**
  2. **"When the tree grows to 3kg, how much soil is left?"**

- **Deductions:**
  - The tree's mass increases from 2kg to 3kg.
    - **Logical Basis:** The difference in mass is 1kg (3kg - 2kg).
    - **Based on:** "A 2kg tree grows..." and "When the tree grows to 3kg"
  - There is no explicit statement that the soil's mass changes.
    - **Logical Basis:** No information is provided about soil mass decreasing.
    - **Based on:** Lack of statements regarding soil mass change.

---

**f. Problem Decomposition**

1. **Component:** Initial State
   - **Description:** A tree weighing 2kg is in a pot with 10kg of soil.
   - **Supporting Quote:** "A 2kg tree grows in a planted pot with 10kg of soil."

2. **Component:** Tree Growth
   - **Description:** The tree grows until it weighs 3kg.
   - **Supporting Quote:** "When the tree grows to 3kg"

3. **Component:** Soil Mass Inquiry
   - **Description:** Determining the remaining soil mass after the tree grows.
   - **Supporting Quote:** "how much soil is left?"

- **Interaction Analysis:**
  - No explicit interaction indicates that the soil mass decreases as the tree grows.
  - No deductions can be made about soil mass change without assumptions.

---

**g. Graph of Thought**

- **Nodes:**
  - **[2kg Tree]**
  - **[Planted Pot with 10kg of Soil]**
  - **[Tree Grows to 3kg]**
  - **[Soil Mass Left]**

- **Edges:**
  - **[2kg Tree]** is **in** **[Planted Pot with 10kg of Soil]**
  - **[2kg Tree]** **grows to** **[3kg]**
  - **Questioning** **[Soil Mass Left]** after growth

- **Insights:**
  - No edges indicate soil mass decreasing.
  - The graph shows relationships but no explicit changes to soil mass.

---

**h. Spatial and Object Analysis**

- **Objects Identified:**
  - **Tree**
    - Initial mass: 2kg
    - Final mass: 3kg
  - **Soil**
    - Mass: 10kg
  - **Planted Pot**

- **Spatial Relationships:**
  - The tree is planted **in** the soil within the pot.

- **Object Properties:**
  - **Tree:** Mass increases by 1kg.
  - **Soil:** Mass is not stated to change.

- **Analysis:**
  - No explicit information on soil mass changing due to the tree's growth.

---

**i. Bayesian Updates**

- No new explicit information provided to update initial understanding.
- Belief remains that soil mass is unchanged based on lack of explicit statements.

---

**j. Ethical Check and Decision-Making Under Uncertainty**

- **Assessment:**
  - No ethical issues or severe risks are associated with the problem.
  - Uncertainty about soil mass change exists due to lack of explicit information.

- **Decision:**
  - Without explicit information on soil mass change, avoid assuming a decrease.
  - Proceed with the information given.

---

**k. Multiple Solutions**

1. **Solution 1: Soil Mass Remains Unchanged**
   - **Answer:** 10kg of soil is left.
   - **Supporting Information:** No explicit statement that soil mass decreases.

2. **Solution 2: Insufficient Information**
   - **Answer:** Cannot determine soil mass without additional information.
   - **Supporting Information:** Lack of details on soil mass change.

3. **Solution 3: Not Applicable**
   - **Reason:** Any further solutions would require assumptions beyond explicit information.

---

**l. Quickest and Easiest Solution**

- **Answer:** 10kg of soil is left.
- **Explanation:** No explicit information indicates that the soil's mass changes when the tree grows.

---

**m. Reflection**

- **Review:**
  - All conclusions are based solely on explicit statements.
  - No assumptions were made about soil mass decreasing.
  - The analysis strictly adheres to the given information.

- **Considerations:**
  - While it's common for plants to derive mass from soil nutrients, the problem does not state this.
  - Given the strict interpretation rules, we cannot infer soil mass change.

---

**Recommendation:**

Based on the explicit information provided, when the tree grows from 2kg to 3kg in a planted pot with 10kg of soil, there is **10kg of soil left**. The problem does not state any change in the soil's mass; therefore, we conclude that the soil's mass remains the same.
\end{lstlisting}
\end{quote}

\noindent\hrulefill 

\noindent\textbf{D.28 Zero-Shot OpenAI o1-mini Response - Score: 0\%}
\begin{quote}
\begin{lstlisting}[basicstyle=\ttfamily\small, breaklines=true]
To determine how much soil is left when the tree grows from 2kg to 3kg, we can make the following assumptions:

1. **Initial Setup:**
   - **Tree Weight:** 2kg
   - **Soil Weight:** 10kg
   - **Total Weight:** 2kg (tree) + 10kg (soil) = 12kg

2. **After Growth:**
   - **New Tree Weight:** 3kg

3. **Assumption:**
   - The total weight of the system (tree + soil) remains constant at **12kg**. This implies that any increase in the tree's weight comes at the expense of the soil.

4. **Calculating Remaining Soil:**
   - **Increase in Tree Weight:** 3kg (new) - 2kg (initial) = 1kg
   - **Remaining Soil:** 10kg (initial) - 1kg (used for growth) = **9kg**

**Conclusion:** When the tree grows to 3kg, **9kg of soil** is left in the pot.
\end{lstlisting}
\end{quote}

\noindent\hrulefill 

\noindent\textbf{D.28 REAP OpenAI o1-mini Response - Score: 100\%}
\begin{quote}
\begin{lstlisting}[basicstyle=\ttfamily\small, breaklines=true]
# REAP Results: :

## a. **Comprehensive Feature List**

- **Objects Mentioned:**
  - "tree"
  - "planted pot"
  - "soil"

- **Actors or Entities Involved:**
  - "A 2kg tree"
  
- **Actions Described:**
  - "grows in a planted pot"

- **Directional, Spatial, Relational, and Sequential Phrases:**
  - "grows in a planted pot"
  - "with 10kg of soil"
  - "When the tree grows to 3kg"

- **Conditions or Constraints:**
  - "A 2kg tree grows"
  - "with 10kg of soil"
  - "when the tree grows to 3kg"

- **Numerical Values or Quantities:**
  - "2kg"
  - "10kg"
  - "3kg"

- **Temporal Information:**
  - The growth of the tree from 2kg to 3kg implies a time progression.

- **Spatial Information:**
  - The tree is "planted" in a "pot" containing "10kg of soil."

- **Stated Goals or Objectives:**
  - "how much soil is left?"

## b. **Sequential and Mechanical Process Check**

- **Sequence of Events:**
  1. A 2kg tree is planted in a pot containing 10kg of soil.
  2. The tree grows to 3kg.
  3. The question asks for the remaining amount of soil.

- **Influence on Outcome:**
  - The tree's growth from 2kg to 3kg suggests an increase in the tree's mass by 1kg.

- **Logical Dependencies:**
  - The final soil amount depends on any changes caused by the tree's growth.

## c. **Key Insight Check Results**

- **Key Insight Identified:**
  - The problem provides masses of the tree and soil but does not indicate any loss or addition of soil due to the tree's growth.

- **Implication:**
  - Without explicit information about soil displacement or loss, the amount of soil remains unchanged.

## d. **Rephrased Question:**

When a tree's mass increases from 2kg to 3kg in a pot containing 10kg of soil, what is the remaining mass of the soil?

## e. **Known and Deduced Information**

### Explicit Facts:

- "A 2kg tree"
  - Relevance: Initial mass of the tree.

- "grows in a planted pot"
  - Relevance: The tree is contained within a pot.

- "with 10kg of soil."
  - Relevance: Initial mass of the soil in the pot.

- "When the tree grows to 3kg,"
  - Relevance: Final mass of the tree after growth.

- "how much soil is left?"
  - Relevance: The question seeks the remaining mass of the soil after the tree's growth.

### Feature-Based Deduction:

- **Deduction:** The mass of the soil remains unchanged at 10kg.
  - **Logical Basis:** The problem does not state any loss or addition of soil due to the tree's growth.
  - **Based on:** 
    - "with 10kg of soil."
    - "When the tree grows to 3kg, how much soil is left?"

## f. **Problem Decomposition**

- **Component/Subcomponent:** Initial Conditions
  - **Description:** A tree with an initial mass of 2kg is planted in a pot containing 10kg of soil.
  - **Supporting Quote(s):** "A 2kg tree grows in a planted pot with 10kg of soil."

- **Component/Subcomponent:** Growth Event
  - **Description:** The tree grows from 2kg to 3kg.
  - **Supporting Quote(s):** "When the tree grows to 3kg."

- **Component/Subcomponent:** Inquiry
  - **Description:** Determine the remaining mass of the soil after the tree's growth.
  - **Supporting Quote(s):** "how much soil is left?"

## g. **Graph of Thought**

- **Nodes:**
  - Tree (initial mass: 2kg)
  - Soil (mass: 10kg)
  - Growth Event (tree mass increases to 3kg)
  - Remaining Soil Inquiry

- **Edges:**
  - Tree and Soil are both contained within the Planted Pot.
  - Growth Event affects the Tree's mass.
  - Inquiry is dependent on the state of Soil after Growth Event.

- **Insights:**
  - The Growth Event changes only the Tree's mass.
  - No information implies Soil mass changes.

## h. **Spatial and Object Analysis**

- **Objects Identified:**
  - **Tree:**
    - Initial Mass: 2kg
    - Final Mass: 3kg
  - **Soil:**
    - Mass: 10kg
  - **Planted Pot:**
    - Contains both Tree and Soil

- **Spatial Relationships:**
  - The Tree is planted within the Pot.
  - The Soil is also within the same Pot.

- **Movement and Dynamics:**
  - The Tree grows in mass; no movement or displacement of Soil is described.

- **Constraints:**
  - No physical changes to Soil are mentioned despite the Tree's growth.

## i. **Bayesian Updates**

- **Belief Revision:**
  - Initially, one might assume that the Tree's growth could affect Soil mass.
  - Given no explicit information, the belief remains that Soil mass is unchanged.

## j. **Ethical Check and Decision-Making Under Uncertainty**

- **Assessment of Information:**
  - All information pertains to masses; no ethical considerations are relevant.

## k. **Multiple Solutions**

Given the strict interpretation rules, there is only one viable solution based on the explicit information provided.

## l. **Quickest and Easiest Solution**

The mass of the soil remains **10kg**.

- **Reasoning:**
  - The problem states the initial soil mass as 10kg and does not mention any loss or addition.
  - The Tree's growth from 2kg to 3kg does not imply any change to the soil mass.

## m. **Reflection**

The analysis strictly adhered to the explicit information provided. No assumptions were made beyond the stated masses of the Tree and Soil. The conclusion that the soil remains at 10kg is directly supported by the problem's wording.

## **Recommendation:**

The remaining mass of the soil is **10kg**. This conclusion is based solely on the information provided, which specifies that the Tree grows in a pot containing 10kg of soil without indicating any loss or change to the soil mass due to the Tree's growth.
\end{lstlisting}
\end{quote}

\noindent\hrulefill 

\noindent\textbf{D.28 Zero-Shot OpenAI GPT-4o Response - Score: 100\%}
\begin{quote}
\begin{lstlisting}[basicstyle=\ttfamily\small, breaklines=true]
The mass of the soil left in the pot can be determined by understanding that the tree gains mass by absorbing water and nutrients from the soil and the environment (air, sunlight, etc.). However, the actual soil mass is not entirely converted into the tree's mass; only a small fraction of the soil's nutrients is absorbed, while the majority of the tree's growth comes from carbon dioxide from the air.

Given this, the mass of the soil will decrease very slightly due to nutrient absorption, but for practical purposes in this type of problem, it's typically assumed that the soil's mass remains nearly the same, or any mass loss is negligible.

So, if the tree grows from 2 kg to 3 kg:

- Initial mass of soil: 10 kg
- Change in tree mass: 1 kg (from 2 kg to 3 kg)

Assuming the soil mass loss is negligible, the soil will still be approximately 10 kg.
\end{lstlisting}
\end{quote}
\noindent\hrulefill

\noindent\textbf{D.28 REAP OpenAI GPT-4o Response - Score: 100\%}
\begin{quote}
\begin{lstlisting}[basicstyle=\ttfamily\small, breaklines=true]
# REAP Results:

### a. **Comprehensive Feature List** (with explicit quotes):

1. **Object: "tree"** - A tree that has an initial weight and grows over time.
2. **Object: "planted pot"** - A pot in which the tree is planted.
3. **Object: "soil"** - The soil inside the planted pot.
4. **Quantity: "2kg tree"** - The initial weight of the tree is 2kg.
5. **Quantity: "10kg of soil"** - The initial amount of soil in the pot is 10kg.
6. **Action: "grows"** - The tree increases in weight as it grows.
7. **Quantity: "3kg"** - The final weight of the tree after growth is 3kg.
8. **Question: "how much soil is left?"** - The problem is asking for the amount of soil remaining after the tree has grown.

### b. **Sequential and Mechanical Process Check:**

- **Sequence Identified:** The tree starts at 2kg and grows to 3kg.
- **Mechanical Process:** Growth of the tree involves an increase in its weight.

### c. **Key Insight Check Results:**

The problem only explicitly mentions the growth in the tree's weight and does not provide any information regarding how this growth might affect the amount of soil. There is no explicit statement that the growth of the tree results in any soil being removed, reduced, or otherwise affected.

### d. **Rephrased Question:** 

How much soil remains in the pot after the tree grows to a weight of 3kg?

### e. **Known and Deduced Information:**

- **Explicit Fact 1:** "A 2kg tree grows in a planted pot with 10kg of soil."
- **Explicit Fact 2:** "When the tree grows to 3kg."
- **Explicit Fact 3:** "How much soil is left?"

### f. **Problem Decomposition:**

- **Component:** Tree weight before and after growth
  - **Supporting Quote(s):** "2kg tree", "When the tree grows to 3kg."
- **Component:** Soil in the pot
  - **Supporting Quote(s):** "10kg of soil", "how much soil is left?"
- **Interaction:** The problem does not explicitly state any interaction between the tree's growth and the soil quantity.

### g. **Graph of Thought:**

- **Nodes:**
  1. Tree (initial weight: 2kg, final weight: 3kg)
  2. Soil (initial quantity: 10kg)
  3. Growth (increase in tree weight)
- **Edges:**
  - "Growth" connects "Tree (2kg)" to "Tree (3kg)".
  - No explicit connection provided between "Growth" and "Soil".

### h. **Spatial and Object Analysis:**

- **Objects Identified:** Tree, Soil, Planted Pot
- **Spatial Relationship:** The tree is planted in the pot with the soil.
- **Movement/Dynamics:** The tree grows from 2kg to 3kg, but there is no stated movement or change in the soil.

### i. **Bayesian Updates:**

There is no new information provided that changes the initial belief about the soil quantity. The problem doesn't suggest that the soil amount changes due to the tree's growth.

### j. **Ethical Check and Decision-Making Under Uncertainty:**

There are no ethical or severe risk considerations in this problem, as it pertains to the growth of a tree in a pot.

### k. **Multiple Solutions:**

1. **Solution 1:** 
   - **Supporting Information:** The problem does not state any effect on the soil due to the tree's growth.
   - **Conclusion:** The amount of soil remains the same, i.e., 10kg.

2. **Solution 2:**
   - **Supporting Information:** If we consider the possibility of soil being used for the tree's growth (not explicitly stated), the amount of soil could decrease slightly.
   - **Conclusion:** However, since this is not explicitly stated, this is a potential implication and not a certain deduction.

3. **Solution 3:**
   - **Supporting Information:** The soil might have settled or compacted slightly, but this is not stated in the problem.
   - **Conclusion:** Therefore, this cannot be considered.

### l. **Quickest and Easiest Solution:**

**The simplest and most straightforward answer based solely on the explicit information provided in the problem is:**

- The soil remains at **10kg** since there is no explicit information stating otherwise.

### m. **Reflection:**

- The problem explicitly mentions the growth of the tree and the initial amount of soil but does not link the tree's growth to any change in the soil quantity. Thus, assuming the soil quantity remains unchanged is the only conclusion supported by the given information.

---

**Recommendation:**

The amount of soil left after the tree grows to 3kg is **10kg**. This conclusion is based solely on the explicit information provided in the problem statement, without any additional assumptions.
\end{lstlisting}
\end{quote}
\noindent\hrulefill

\noindent\textbf{D.28 Zero-Shot OpenAI GPT-4o-mini Response - Score: 100\%}
\begin{quote}
\begin{lstlisting}[basicstyle=\ttfamily\small, breaklines=true]
The amount of soil in the pot remains unchanged regardless of the weight of the tree. The soil weight does not change as the tree grows. Therefore, if the pot originally had 10 kg of soil, it will still have 10 kg of soil even after the tree grows to 3 kg.

So, the amount of soil left is **10 kg**.
\end{lstlisting}
\end{quote}
\noindent\hrulefill

\noindent\textbf{D.28 REAP OpenAI GPT-4o-mini Response - Score: 100\%}
\begin{quote}
\begin{lstlisting}[basicstyle=\ttfamily\small, breaklines=true]
# REAP Results:

**a. Comprehensive Feature List**

1. **Objects Mentioned:**
   - "A 2kg tree"
   - "A planted pot"
   - "10kg of soil"

2. **Actors/Entities Involved:**
   - The tree
   - The soil

3. **Actions Described:**
   - "Tree grows" 
   - "Tree grows to 3kg"

4. **Directional, Spatial, Relational, and Sequential Phrases:**
   - "In a planted pot" - describes the location of the tree and soil.
   - "When the tree grows to 3kg" - indicates a change in the weight of the tree.

5. **Stated Conditions or Constraints:**
   - The initial weight of the tree: 2kg
   - The initial weight of the soil: 10kg
   - The final weight of the tree: 3kg

6. **Numerical Values or Quantities Mentioned:**
   - Initial weight of the tree: 2kg
   - Initial weight of the soil: 10kg
   - Final weight of the tree: 3kg

7. **Temporal Information:**
   - The weight of the tree changes over time (from 2kg to 3kg).

8. **Spatial Information:**
   - The tree and soil are in a planted pot.

9. **Causal Relationships Mentioned:**
   - No direct causal relationships are stated about how the change in the tree's weight affects the soil.

10. **Stated Goals or Objectives:**
    - Determine how much soil is left when the tree grows to 3kg.

**b. Sequential and Mechanical Process Check**

- **Sequence of Events:**
  1. Initial state: Tree weighs 2kg, soil weighs 10kg.
  2. Tree grows to 3kg.

- **Impact of Sequence:**
  - The sequence indicates that the tree's weight increases from 2kg to 3kg.
  - No explicit statement is given about the soil's change directly tied to the tree's growth.

**Key Questions:**
- Does the change in the tree's weight imply a change in soil weight?
  - The problem does not state explicitly how the soil weight is affected by the tree's weight.

**c. Key Insight Check Results**

- **Key Insight:**
  - There is no explicit information on how the growth of the tree affects the soil. Thus, based on the provided information, the weight of the soil remains unchanged.

**d. Rephrased Question**

- How much soil is left when the tree's weight increases to 3kg?

**e. Known and Deduced Information**

1. **Explicit Facts:**
   - "A 2kg tree" - Initial weight of the tree.
   - "10kg of soil" - Initial weight of the soil.
   - "Tree grows to 3kg" - Final weight of the tree.

2. **Feature-Based Deduction:**
   - Deduction: The soil's weight is not explicitly stated to change based on the tree's growth.
   - Logical Basis: There is no direct information provided that links the change in tree weight to a change in soil weight.
   - Based on: "[Tree grows to 3kg]", "[10kg of soil]".

**f. Problem Decomposition**

1. **Component: Initial State**
   - Description: The tree weighs 2kg, and the soil weighs 10kg.
   - Supporting Quote: "[A 2kg tree]", "[10kg of soil]".

2. **Component: Final State**
   - Description: The tree weighs 3kg.
   - Supporting Quote: "[Tree grows to 3kg]".

3. **Interaction: Effect on Soil**
   - Description: No explicit effect on soil weight is mentioned.
   - Supporting Quote: No quote provides direct information about soil weight change.

**g. Graph of Thought**

- **Nodes:**
  - Tree initial weight: 2kg
  - Soil initial weight: 10kg
  - Tree final weight: 3kg

- **Edges:**
  - Tree's weight increase (2kg to 3kg) does not explicitly affect the soil's weight.

**h. Spatial and Object Analysis**

- **Objects Identified:**
  - Tree
  - Soil
  - Planted pot (not directly relevant to the weight change)

- **Spatial Relationships:**
  - The tree and soil are in the same pot, but there's no explicit statement on their interaction affecting each other's weight.

- **Movement and Dynamics:**
  - The tree's weight increases, but the soil's weight is not stated to change.

- **Constraints:**
  - No specific constraints affecting the soil's weight are mentioned.

**i. Bayesian Updates**

- No new information implies a change in soil weight. The soil's weight is assumed to remain at 10kg.

**j. Ethical Check and Decision-Making Under Uncertainty**

- **Risks:** None identified.
- **Decision:** Based on explicit information, no reason to assume a change in soil weight.

**k. Multiple Solutions**

- **Solution 1:** The soil weight remains 10kg (assuming no change based on the given information).

**l. Quickest and Easiest Solution**

- The quickest solution is that the soil weight remains unchanged at 10kg.

**m. Reflection**

- No contradictions identified with the given information. The solution adheres strictly to the stated facts without making assumptions.

**Recommendation:**

Based on the explicit information provided and the lack of any statement linking tree growth to soil weight change, the simplest and safest solution is:

**The amount of soil left in the pot is 10kg.**
\end{lstlisting}
\end{quote}
\noindent\hrulefill

\noindent\textbf{D.28 Zero-Shot Google Gemini 1.5 Pro Response - Score: 100\%}
\begin{quote}
\begin{lstlisting}[basicstyle=\ttfamily\small, breaklines=true]
There will still be 10kg of soil left. The tree growing doesn't consume the soil, it gains mass from other sources (like water, sunlight, and nutrients from the air).
\end{lstlisting}
\end{quote}
\noindent\hrulefill

\noindent\textbf{D.28 REAP Google Gemini 1.5 Pro Response - Score: 100\%}
\begin{quote}
\begin{lstlisting}[basicstyle=\ttfamily\small, breaklines=true]
# REAP Results:

a. **Comprehensive Feature List** (with explicit quotes)

* "A 2kg tree" - The initial weight of the tree.
* "grows in a planted pot" - The tree is located within a pot.
* "with 10kg of soil" - The initial weight of the soil in the pot.
* "When the tree grows to 3kg" - The tree increases in weight to 3kg.
* "how much soil is left?" - The question asks for the remaining amount of soil.

b. **Sequential and Mechanical Process Check**

* The tree grows from 2kg to 3kg. This is the main sequence.
* There is no explicit information about any change in the soil.

c. **Key Insight Check Results**

* There is no direct statement of the answer or any constraint that eliminates solutions.
* The key insight is that the problem focuses on the relationship between the tree's growth and the soil, but doesn't explicitly state any interaction between them.

d. **Rephrased Question:** (The core question in its simplest form)

* If a tree in a pot increases in weight, does the amount of soil in the pot change?

e. **Known and Deduced Information**

* Explicit Facts: 
    * "A 2kg tree" 
    * "grows in a planted pot" 
    * "with 10kg of soil" 
    * "When the tree grows to 3kg" 
    * "how much soil is left?"

* Feature-Based Deduction:
    * None

f. **Problem Decomposition**

* Component/Subcomponent: Tree growth
    * Supporting Quote(s): "A 2kg tree grows... to 3kg"

* Component/Subcomponent: Soil amount
    * Supporting Quote(s): "with 10kg of soil"
    * "how much soil is left?"

* Interaction between components:
    * No explicit interaction is stated in the problem

g. **Graph of Thought** (describe connections & insights)

* Key Concepts: Tree, Soil, Growth, Weight
* Relationships:
    * Tree -> Growth (The tree increases in weight)
    * Soil -> Weight (The soil has an initial weight)
    * Potential connection: Tree Growth -> Soil Weight (This connection is NOT explicitly stated)
* Insights: The problem doesn't explicitly link the tree's growth to any change in the soil.

h. **Spatial and Object Analysis**

* Objects: Tree, Pot, Soil
* Spatial Relationships:
    * Tree is in the Pot
    * Soil is in the Pot
* Movement and Dynamics:
    * The tree grows (increases in weight)
* Constraints: None explicitly stated
* Object Properties:
    * Tree: Initial weight 2kg, Grows to 3kg
    * Soil: Initial weight 10kg
* Spatial Implications: None explicitly stated

i. **Bayesian Updates** 

* No new information is provided to update any beliefs.

j. **Ethical Check and Decision-Making Under Uncertainty**

* There are no ethical concerns in this problem

k. **Multiple Solutions**

* Solution 1: The soil amount remains the same.
    * Supporting Quote: The problem doesn't state any change in the soil.

* Solution 2: The soil amount decreases.
    * This solution would require assumptions about the tree using soil nutrients for growth, which is not explicitly stated

* Solution 3: The soil amount increases
    * This solution would require assumptions about external factors adding to the soil, which is not explicitly stated

l. **Quickest and Easiest Solution**

* The amount of soil left is 10kg
* Reasoning: The problem states the initial amount of soil as 10kg and does not mention any changes to the soil. The tree's growth is a separate piece of information not explicitly linked to the soil's quantity

m. **Reflection**

* The analysis strictly adheres to the explicit information provided
* The conclusion is logically sound based on the given information
* No assumptions or inferences were made beyond the literal wording of the problem
* The problem lacks crucial information about any interaction between the tree's growth and the soil
* The answer directly addresses the question asked, reflecting the lack of stated effect on the soil

**Recommendation:**

* Based on the explicit information provided in the problem, the amount of soil left is 10kg. The problem does not state any causal relationship or interaction between the tree's growth and the amount of soil.
\end{lstlisting}
\end{quote}
\noindent\hrulefill

\noindent\textbf{D.28 Zero-Shot Claude 3.5 Sonnet Response - Score: 0\%}
\begin{quote}
\begin{lstlisting}[basicstyle=\ttfamily\small, breaklines=true]
Let's approach this step-by-step:

1. Initial conditions:
   - Tree mass: 2 kg
   - Soil mass: 10 kg

2. Final condition:
   - Tree mass: 3 kg

3. Calculate the mass gained by the tree:
   3 kg - 2 kg = 1 kg

4. Assume that the mass gained by the tree comes from the soil.

5. Calculate the remaining soil:
   Initial soil mass - Mass gained by tree
   10 kg - 1 kg = 9 kg

Therefore, when the tree grows from 2 kg to 3 kg, there will be 9 kg of soil left in the pot.

Would you like me to explain any part of this calculation in more detail?
\end{lstlisting}
\end{quote}
\noindent\hrulefill

\noindent\textbf{D.28 REAP Claude 3.5 Sonnet Response - Score: 0\%}
\begin{quote}
\begin{lstlisting}[basicstyle=\ttfamily\small, breaklines=true]
# REAP Results:

a. **Comprehensive Feature List** (with explicit quotes):
- "A 2kg tree grows in a planted pot with 10kg of soil."
- "When the tree grows to 3kg"
- "how much soil is left?"

b. **Sequential and Mechanical Process Check:**
The problem describes a simple growth process:
1. Initial state: 2kg tree in 10kg soil
2. Final state: 3kg tree, unknown amount of soil
There is no explicit description of any mechanical process affecting the soil during tree growth.

c. **Key Insight Check Results:**
No key insight is immediately apparent. The problem does not explicitly state any relationship between tree growth and soil quantity.

d. **Rephrased Question:** 
How much soil remains when a 2kg tree grows to 3kg in a pot that initially contained 10kg of soil?

e. **Known and Deduced Information:**
Explicit Facts:
- "A 2kg tree grows in a planted pot with 10kg of soil."
- "When the tree grows to 3kg"

No certain deductions can be made based on the given information. The problem does not explicitly state how tree growth affects soil quantity.

f. **Problem Decomposition:**
Component 1: Initial tree mass
Supporting Quote: "A 2kg tree"

Component 2: Initial soil mass
Supporting Quote: "10kg of soil"

Component 3: Final tree mass
Supporting Quote: "When the tree grows to 3kg"

Component 4: Final soil mass
Supporting Quote: "how much soil is left?"

g. **Graph of Thought:**
Nodes:
1. Initial tree (2kg)
2. Initial soil (10kg)
3. Final tree (3kg)
4. Final soil (unknown)

Connections:
- Initial tree grows to final tree
- Initial soil potentially changes to final soil (relationship not explicitly stated)

h. **Spatial and Object Analysis:**
Objects:
1. Tree
2. Soil
3. Planted pot

Spatial Relationships:
- The tree is in a planted pot with soil.

The problem doesn't provide enough information for a complete spatial analysis or description of how objects interact over time.

i. **Bayesian Updates:**
No new information is provided to update our understanding beyond the initial problem statement.

j. **Ethical Check and Decision-Making Under Uncertainty:**
This problem does not involve any ethical considerations or severe risks. The uncertainty lies in the relationship between tree growth and soil quantity, which is not explicitly stated.

k. **Multiple Solutions:**
Given the lack of explicit information about how tree growth affects soil quantity, we cannot generate multiple valid solutions based solely on the given information.

l. **Quickest and Easiest Solution:**
Based ONLY on the explicit information given in the problem statement, we cannot determine how much soil is left. The problem does not state any relationship between tree growth and soil quantity. The safest answer is to state that we don't have enough information to solve the problem.

m. **Reflection:**
Upon review, it's clear that the problem lacks crucial information needed to draw a conclusion about the amount of soil left. We know the initial amounts of tree mass and soil, and the final tree mass, but no information is provided about how tree growth affects soil quantity. Any attempt to calculate the remaining soil would require assumptions not supported by the given information.

**Recommendation:**
Based on the explicitly stated information and adhering to the principle of not making assumptions beyond what is given, I recommend stating that there is insufficient information to determine how much soil is left. The problem does not provide any explicit information about the relationship between tree growth and soil quantity. To solve this problem, we would need additional information about how the tree's growth affects the soil mass, if at all. Without this crucial information, it's not possible to calculate or deduce the amount of soil remaining when the tree grows from 2kg to 3kg.
\end{lstlisting}
\end{quote}
\noindent\hrulefill

\label{subsec:question29}
\subsection{\textbf{\large Question 29: A man steps over a 0.5m hurdle. What is the minimum height he could have jumped?}}
\label{sec:question29}

\noindent\hrulefill 

\noindent\textbf{Human Response - Score 100\%:}
\begin{quote}
0m, he didn't jump at all, he stepped over the hurdle.
\end{quote}

\noindent\hrulefill 

\noindent\textbf{D.29 Zero-Shot OpenAI o1-preview Response - Score: 100\%}
\begin{quote}
\begin{lstlisting}[basicstyle=\ttfamily\small, breaklines=true]
The minimum height the man could have jumped over a 0.5m hurdle is **0 meters**.

Here's why:

When stepping over a hurdle, a person doesn't necessarily have to jump or move their center of mass upward by the height of the obstacle. If the individual has sufficient flexibility and balance, they can lift their leg over the hurdle while keeping their body's center of mass at the same vertical position. Essentially, they can "step" over the hurdle without performing a vertical jump.

In the ideal scenario where the person can raise their foot high enough without moving their torso upward, no actual jumping is required. The foot clears the hurdle solely due to the leg's motion, not because the entire body moves upward.

Therefore, the minimum height the man needs to jump is zero meters.

**Answer: 0 meters.**
\end{lstlisting}
\end{quote}
\noindent\hrulefill 

\noindent\textbf{D.29 REAP OpenAI o1-preview Response - Score: 100\%}
\begin{quote}
\begin{lstlisting}[basicstyle=\ttfamily\small, breaklines=true]
# REAP Results:

**a. Comprehensive Feature List (with explicit quotes):**

- **Objects Mentioned:**
  - "a 0.5m hurdle"

- **Actors or Entities Involved:**
  - "A man"

- **Actions Described:**
  - "steps over a 0.5m hurdle"

- **Directional, Spatial, Relational, and Sequential Phrases:**
  - "over"

- **Stated Conditions or Constraints:**
  - None explicitly stated besides the height of the hurdle.

- **Numerical Values or Quantities Mentioned:**
  - "0.5m"

- **Temporal Information:**
  - None provided.

- **Spatial Information:**
  - "over a 0.5m hurdle"

- **Causal Relationships Mentioned:**
  - None explicitly stated.

- **Stated Goals or Objectives:**
  - "What is the minimum height he could have jumped?"

**b. Sequential and Mechanical Process Check:**

- **Sequence of Events:**
  - A man approaches a hurdle and "steps over a 0.5m hurdle".

- **Mechanical Process:**
  - The man lifts his leg(s) to clear the hurdle height of 0.5 meters.

- **Key Questions:**
  - Does "stepping over" require jumping?
    - Not necessarily; stepping over implies lifting one's foot over an obstacle without the need to jump.

- **Implications:**
  - The man may have stepped over the hurdle without any jumping movement.

**c. Key Insight Check Results:**

- **Key Insight:**
  - The term "steps over" suggests the man may not have jumped at all.

- **Explanation:**
  - Stepping over an obstacle typically involves lifting one's foot high enough to clear the obstacle without leaving the ground with both feet simultaneously.

- **Conclusion:**
  - The minimum height he could have jumped is **0 meters**, as he may not have jumped at all to step over the hurdle.

**d. Rephrased Question:**

- "What is the minimum height the man could have jumped to step over a 0.5-meter hurdle?"

**e. Known and Deduced Information:**

- **Explicit Facts:**

  1. **"A man steps over a 0.5m hurdle."**
     - The man moves over a hurdle that is 0.5 meters high by stepping.

  2. **"What is the minimum height he could have jumped?"**
     - The question asks for the minimum possible jump height the man used to clear the hurdle.

- **Feature-Based Deduction:**

  - **Deduction:** The man may not have jumped at all to step over the hurdle.
    - **Logical Basis:** Stepping over an object does not require both feet to be off the ground simultaneously, which is characteristic of a jump.
    - **Based on:** "steps over a 0.5m hurdle"

**f. Problem Decomposition:**

- **Component 1: The Man**
  - **Description:** An individual performing an action.
  - **Supporting Quote:** "A man"

- **Component 2: The Hurdle**
  - **Description:** An obstacle that is 0.5 meters in height.
  - **Supporting Quote:** "a 0.5m hurdle"

- **Component 3: The Action of Stepping Over**
  - **Description:** The man moves over the hurdle by stepping.
  - **Supporting Quote:** "steps over a 0.5m hurdle"
  - **Deduction:** Stepping over implies at least one foot remains in contact with the ground at all times.
    - **Based on:** Common definition of "steps over"

- **Component 4: The Question**
  - **Description:** Determining the minimum height he could have jumped.
  - **Supporting Quote:** "What is the minimum height he could have jumped?"

**g. Graph of Thought:**

- **Nodes:**

  - **A Man**
  - **0.5m Hurdle**
  - **Stepping Over**
  - **Jump Height**

- **Edges (Relationships):**

  - **Man -> Stepping Over -> Hurdle**
    - The man steps over the hurdle.

  - **Stepping Over -> Jump Height**
    - Stepping over may not require any jump height.

- **Insights:**

  - Since stepping over does not necessitate jumping, the minimum jump height could be zero.

**h. Spatial and Object Analysis:**

- **Objects Identified:**

  - **Man**
  - **0.5m Hurdle**

- **Spatial Relationships:**

  - The hurdle is positioned in front of the man at ground level.
  - The man steps over the hurdle, requiring his leg to clear a height of 0.5 meters.

- **Movement and Dynamics:**

  - The man lifts one foot over the hurdle, then the other, without needing to jump.

- **Constraints:**

  - The hurdle's height is "0.5m".

- **Visualization:**

  - The man approaches the hurdle, lifts his foot to a height just above 0.5 meters, and places it on the ground on the other side, repeating with the other foot.

- **Object Properties:**

  - **Hurdle Height:** 0.5 meters

- **Spatial Implications:**

  - To clear the hurdle, the man's foot must reach a height slightly above 0.5 meters.
  - His center of mass does not need to rise significantly.

**i. Bayesian Updates:**

- **Prior Belief:**

  - Stepping over a hurdle may or may not involve jumping.

- **Evidence:**

  - **"steps over a 0.5m hurdle"**

- **Update:**

  - Since "steps over" is used instead of "jumps over," it is more probable that no jumping occurred.

- **Conclusion:**

  - The likelihood that the man did not jump increases, supporting a minimum jump height of zero.

**j. Ethical Check and Decision-Making Under Uncertainty:**

- **Assessment:**

  - No ethical concerns or severe risks are associated with determining the minimum jump height in this context.

- **Decision:**

  - Proceed with the analysis based on the provided information.

**k. Multiple Solutions:**

1. **Minimum Jump Height is 0 Meters**

   - **Explanation:** The man stepped over the hurdle without jumping.

   - **Supporting Information:**

     - "steps over a 0.5m hurdle"

   - **Implication:** He could have lifted his feet high enough without both feet leaving the ground simultaneously.

2. **Minimum Jump Height is Slightly Above 0 Meters**

   - **Explanation:** The man made a small hop to aid in stepping over the hurdle.

   - **Supporting Information:**

     - While not explicitly stated, stepping can sometimes involve a small jump.

   - **Note:** This introduces an assumption not explicitly supported by the text.

3. **Minimum Jump Height is 0.5 Meters**

   - **Explanation:** The man jumped over the hurdle, reaching a height equal to the hurdle's height.

   - **Supporting Information:**

     - If "steps over" is interpreted as "jumps over," which is not supported by the text.

   - **Note:** This contradicts the explicit wording of "steps over."

**l. Quickest and Easiest Solution:**

- **Answer:** The minimum height he could have jumped is **0 meters**.

- **Reasoning:**

  - **"steps over a 0.5m hurdle"** implies that the man did not necessarily jump.

  - Stepping over requires lifting the foot over the obstacle without both feet leaving the ground.

- **Supporting Quote:**

  - "steps over a 0.5m hurdle"

**m. Reflection:**

- **Consistency Check:**

  - The analysis strictly adhered to the explicit information provided.

- **Assumptions:**

  - Avoided making assumptions beyond the literal meaning of "steps over."

- **Conclusion Validity:**

  - The conclusion directly answers the question based on the information given.

- **Sufficiency of Information:**

  - The problem provides sufficient information to determine that the minimum jump height could be zero.

**Recommendation:**

Based on the explicit information provided, the minimum height the man could have jumped is **0 meters**, as he may have stepped over the 0.5-meter hurdle without jumping. The term "steps over" suggests that he lifted his feet over the hurdle without the need to leave the ground entirely. Therefore, no jumping was necessary, and the minimum jump height is zero.
\end{lstlisting}
\end{quote}

\noindent\hrulefill 

\noindent\textbf{D.29 Zero-Shot OpenAI o1-mini Response - Score: 100\%}
\begin{quote}
\begin{lstlisting}[basicstyle=\ttfamily\small, breaklines=true]
To clear a **0.5-meter (50 centimeters)** hurdle, the minimum vertical jump height a man would need is **0.5 meters**. This assumes that the entire height of the hurdle must be cleared vertically without any additional margin for safety or technique adjustments.

However, in practical scenarios, athletes often require a slightly higher jump to ensure they clear the hurdle comfortably. Factors such as body mechanics, takeoff angle, and landing can influence the actual jump height needed. But for the minimum theoretical height:

**Minimum Jump Height Needed: 0.5 meters**
\end{lstlisting}
\end{quote}

\noindent\hrulefill 

\noindent\textbf{D.29 REAP OpenAI o1-mini Response - Score: 100\%}
\begin{quote}
\begin{lstlisting}[basicstyle=\ttfamily\small, breaklines=true]
# REAP Results:

## a. **Comprehensive Feature List**

- **Objects Mentioned:**
  - "man"
  - "0.5m hurdle"

- **Actors or Entities Involved:**
  - "man"

- **Actions Described:**
  - "steps over"
  - "jumped"

- **Directional, Spatial, Relational, and Sequential Phrases:**
  - "steps over a 0.5m hurdle"

- **Stated Conditions or Constraints:**
  - "0.5m hurdle"

- **Numerical Values or Quantities Mentioned:**
  - "0.5m"

- **Temporal Information:**
  - None explicitly stated.

- **Spatial Information:**
  - "0.5m hurdle"

- **Causal Relationships Mentioned:**
  - None explicitly stated.

- **Stated Goals or Objectives:**
  - Determine the "minimum height he could have jumped"

## b. Sequential and Mechanical Process Check

- **Sequence of Events:**
  1. "A man steps over a 0.5m hurdle."
  2. "What is the minimum height he could have jumped?"

- **Influence on Outcome:**
  - The act of stepping over the hurdle implies a jumping motion to clear the 0.5m obstacle.

- **Locking in a Particular Outcome:**
  - The sequence implies that the man successfully cleared the hurdle by jumping.

- **Mechanical Process:**
  - Jumping over a hurdle involves lifting the body to a certain height to clear the obstacle.

- **Logical Sequence:**
  - Stepping over implies initiating a jump, achieving sufficient height to clear the hurdle.

## c. Key Insight Check Results

- **Key Insight Identified:**
  - To step over a 0.5m hurdle, the man must have jumped to a height that allows his center of mass to clear the hurdle.

- **Explanation:**
  - The hurdle's height sets a minimum requirement for the jump height to ensure successful clearance.

## d. Rephrased Question

What is the smallest vertical distance the man must have lifted his body to successfully clear a 0.5-meter-high hurdle?

## e. Known and Deduced Information

### Explicit Facts

- "A man steps over a 0.5m hurdle."

- "What is the minimum height he could have jumped?"

### Feature-Based Deduction

- **Deduction:** The man must jump at least enough to clear the 0.5m hurdle.

  - **Logical Basis:** Stepping over a hurdle requires the jump height to be sufficient to prevent collision.

  - **Based on:** "steps over a 0.5m hurdle"

## f. Problem Decomposition

### Component/Subcomponent: Man
- **Supporting Quote(s):** "A man"
  
### Component/Subcomponent: Hurdle
- **Supporting Quote(s):** "0.5m hurdle"

### Component/Subcomponent: Action of Stepping Over
- **Supporting Quote(s):** "steps over a 0.5m hurdle"

### Component/Subcomponent: Jump Height
- **Supporting Quote(s):** "minimum height he could have jumped"

### Component/Subcomponent: Determining Minimum Height
- **Supporting Quote(s):** "What is the minimum height he could have jumped?"

## g. Graph of Thought

- **Nodes:**
  - Man
  - Hurdle (0.5m)
  - Action (Steps Over)
  - Jump Height (Minimum Required)

- **Connections:**
  - Man performs Action (Steps Over)
  - Action involves Jumping over Hurdle
  - Hurdle height influences Jump Height required

- **Insights:**
  - The hurdle height directly determines the minimum jump height needed for clearance.

## h. Spatial and Object Analysis

- **Objects Identified:**
  - Man
  - 0.5m hurdle

- **Spatial Relationships:**
  - The hurdle is positioned in the man's path, requiring him to step over it.

- **Movement and Dynamics:**
  - The man must elevate his body to clear the hurdle's height.

- **Constraints:**
  - The hurdle height is 0.5 meters.

- **Visualization:**
  - A man approaching and successfully jumping over a 0.5-meter-high hurdle.

- **Object Properties:**
  - Hurdle has a height of 0.5 meters.

- **Spatial Implications:**
  - The man's jump must reach at least the height of the hurdle to avoid contact.

## i. Bayesian Updates

- **Initial Belief:** The man must jump higher than 0.5 meters to clear the hurdle.

- **Updated Belief:** The minimum jump height required is directly related to the hurdle's height.

- **Note on Potential Implications:** It is possible that the man's center of mass or body mechanics allow clearance at a height slightly less than 0.5 meters, but this is not explicitly stated.

## j. Ethical Check and Decision-Making Under Uncertainty

- **Assess Known Information:** The man successfully stepped over a 0.5m hurdle.

- **Identify Risks:** Minimal; no ethical concerns in determining jump height.

- **Probability Consideration:** High certainty that the jump height must at least match the hurdle height.

- **Risk Aversion Principle:** Not applicable due to the nature of the problem.

- **Ethical and Safety Considerations:** No significant concerns.

- **Decision Recommendation:** Proceed with calculations based on known information.

## k. Multiple Solutions

1. **Solution 1:** Assume the minimum jump height equals the hurdle height (0.5m).
   - **Supporting Information:** "steps over a 0.5m hurdle"

2. **Solution 2:** Consider the man's center of mass must exceed the hurdle height, potentially requiring a slightly higher jump.
   - **Supporting Information:** Not explicitly stated; Potential implication.

3. **Solution 3:** Calculate based on biomechanics, accounting for the curvature of the jump arc.
   - **Supporting Information:** Not provided; requires assumptions beyond given information.

- **Note:** Only Solution 1 is directly supported by the explicit information.

## l. Quickest and Easiest Solution

The simplest answer is that the man must have jumped at least 0.5 meters to clear a 0.5m hurdle.

- **Reasoning:** "steps over a 0.5m hurdle" implies the jump height must be sufficient to overcome the hurdle's height.

## m. Reflection

- **Contradictions:** None identified.

- **Logical Soundness:** The conclusion that the jump height must be at least 0.5 meters is directly supported by the information.

- **Assumptions:** Solutions 2 and 3 involve assumptions not explicitly stated in the problem.

- **Completeness:** The analysis addresses the question using only the provided information.

- **Direct Answer:** The analysis directly answers the question about the minimum jump height.

## Recommendation

The man must have jumped a minimum height of **0.5 meters** to successfully step over the 0.5-meter hurdle. This conclusion is based solely on the explicit information that he "steps over a 0.5m hurdle," indicating that his jump must at least match the hurdle's height to achieve clearance.
\end{lstlisting}
\end{quote}

\noindent\hrulefill 

\noindent\textbf{D.29 Zero-Shot OpenAI GPT-4o Response - Score: 100\%}
\begin{quote}
\begin{lstlisting}[basicstyle=\ttfamily\small, breaklines=true]
To clear a 0.5 meter hurdle, the man's center of mass must be at least 0.5 meters above the ground at the peak of his jump. However, the minimum height he could have jumped would slightly exceed 0.5 meters because:

1. The hurdle is not a point; it has some thickness. The man needs to clear not just the top edge but also ensure that no part of his body or legs hits the hurdle. This requires a small additional height to provide clearance.
  
2. His feet and the rest of his body would typically be positioned above the hurdle at different heights during the jump.

Nevertheless, theoretically, if we assume the man perfectly positions his body and feet to just clear the hurdle without any additional clearance:

- The minimum height he could have jumped is just over 0.5 meters, specifically very slightly more than 0.5 meters.

In reality, a safe minimum might involve adding a small buffer, like a few centimeters, but purely from a theoretical perspective, the minimum height is **just above 0.5 meters**.
\end{lstlisting}
\end{quote}
\noindent\hrulefill

\noindent\textbf{D.29 REAP OpenAI GPT-4o Response - Score: 100\%}
\begin{quote}
\begin{lstlisting}[basicstyle=\ttfamily\small, breaklines=true]
# REAP Results:

## a. **Comprehensive Feature List**:
1. "A man steps over a 0.5m hurdle."
   - **Object**: Hurdle
   - **Actor**: Man
   - **Action**: Steps over
   - **Spatial Relationship**: Over
   - **Numerical Value**: 0.5m (height of the hurdle)
2. "What is the minimum height he could have jumped?"
   - **Objective**: Determine the minimum height of the jump.

## b. **Sequential and Mechanical Process Check**:
- The sequence involves a man approaching and then stepping over a hurdle.
- The key mechanical action is the step, which requires the man to move his foot and leg over the hurdle.

## c. **Key Insight Check Results**:
- The hurdle is explicitly stated to be 0.5m in height. The man must raise his foot at least this high to step over it.
- The crucial detail is the word "minimum," implying the least amount of height necessary for the action.

## d. **Rephrased Question**:
- What is the minimum height the man's foot must leave the ground to step over a 0.5m hurdle?

## e. **Known and Deduced Information**:
1. "A man steps over a 0.5m hurdle." - The hurdle is 0.5m high.
   - **Deduction**: The man must lift his foot at least 0.5m to clear the hurdle. This is based on the height of the hurdle and the action of stepping over it.

## f. **Problem Decomposition**:
- **Component**: Hurdle
  - **Supporting Quote**: "A man steps over a 0.5m hurdle."
  - **Relevance**: The height of the hurdle is the key factor determining the minimum jump height.
- **Component**: Jump height
  - **Supporting Quote**: "What is the minimum height he could have jumped?"
  - **Relevance**: The height of the jump should be directly tied to the hurdle's height.

## g. **Graph of Thought**:
- **Node 1**: Man
- **Node 2**: Hurdle (0.5m)
- **Connection**: The man needs to lift his foot above the hurdle, requiring a minimum height of 0.5m.

## h. **Spatial and Object Analysis**:
- **Object**: Hurdle (0.5m tall)
- **Spatial Relationship**: The man must clear the hurdle by lifting his foot at least 0.5m off the ground.
- **Implication**: To step over the hurdle, the man's foot must be at least 0.5m above the ground at some point.

## i. **Bayesian Updates**:
- No new information suggests a different interpretation. The probability that the man must lift his foot at least 0.5m remains high.

## j. **Ethical Check and Decision-Making Under Uncertainty**:
- No ethical concerns or high-risk decisions are involved.

## k. **Multiple Solutions**:
1. **Solution 1**: The minimum height the man could have jumped is 0.5m. This is the height required to clear the hurdle.
   - **Supporting Information**: "A man steps over a 0.5m hurdle."
2. **Solution 2**: If stepping, the man might need slightly more than 0.5m to account for foot clearance. However, this is not explicitly stated.
   - **Supporting Information**: Not applicable (this is a potential implication, not explicitly stated).

## l. **Quickest and Easiest Solution**:
- The simplest solution is that the man must have jumped a minimum height of 0.5m, which is directly tied to the height of the hurdle.

## m. **Reflection**:
- The analysis strictly adhered to the information provided in the problem. The minimum height the man could have jumped must be at least equal to the height of the hurdle, which is 0.5m. There are no additional assumptions or extrapolations necessary.

**Recommendation**:
- The minimum height the man could have jumped is **0.5m**, as this is the exact height required to clear the hurdle, according to the problem statement. There is no indication that he would need to jump any higher than this.
\end{lstlisting}
\end{quote}
\noindent\hrulefill

\noindent\textbf{D.29 Zero-Shot OpenAI GPT-4o-mini Response - Score: 0\%}
\begin{quote}
\begin{lstlisting}[basicstyle=\ttfamily\small, breaklines=true]
To determine the minimum height a man could have jumped to clear a 0.5 meter hurdle, we need to consider the mechanics of jumping and the trajectory involved.

When a person jumps over a hurdle, the minimum height of the jump corresponds to the vertical distance he reaches at the peak of his jump. In this case, the man clears the hurdle, so at the peak of his jump, the height must be at least as high as the hurdle itself. However, to clear the hurdle comfortably, the height of the jump should be slightly more than the hurdle's height to account for the trajectory and body movement.

The minimum height of the jump is generally slightly more than the height of the hurdle. Therefore, if the hurdle is 0.5 meters high, the minimum height of the jump would be approximately:

\[ \text{Minimum height of the jump} = \text{Height of the hurdle} + \text{additional height for clearance} \]

In practice, this additional height is small but necessary to ensure that the person clears the hurdle. For a practical estimation:

- **Minimum height of the jump** = Height of the hurdle + a small clearance margin.

Thus, if the hurdle is 0.5 meters, a reasonable estimation for the minimum jump height required would be about:

\[ \text{Minimum height of the jump} \approx 0.5 \text{ meters} + \text{a small margin} \]

A small margin could be around 0.05 to 0.1 meters, so:

\[ \text{Minimum height of the jump} \approx 0.55 \text{ to } 0.6 \text{ meters} \]

To be safe and ensure a clearance with a margin, the minimum height of the jump would typically be in the range of approximately 0.55 to 0.6 meters.
\end{lstlisting}
\end{quote}
\noindent\hrulefill

\noindent\textbf{D.29 REAP OpenAI GPT-4o-mini Response - Score: 100\%}
\begin{quote}
\begin{lstlisting}[basicstyle=\ttfamily\small, breaklines=true]
# REAP Results:

### a. Comprehensive Feature List

1. **Object Mentioned:**
   - "A man" - Represents the actor involved.
   - "A 0.5m hurdle" - The obstacle the man interacts with.

2. **Action Described:**
   - "Steps over" - Indicates the action performed by the man with respect to the hurdle.

3. **Numerical Value:**
   - "0.5m" - The height of the hurdle.

4. **Goal:**
   - Determine the minimum height the man could have jumped.

### b. Sequential and Mechanical Process Check

- **Sequence of Events:**
  1. The man approaches the hurdle.
  2. The man performs a jumping action.
  3. The man clears the hurdle.

- **Mechanical Considerations:**
  - The man's jump must be sufficient to clear the hurdle, which is explicitly stated as 0.5m high.

- **Outcome Implications:**
  - The minimum height the man could have jumped should at least be equal to the height of the hurdle to ensure that he clears it.

### c. Key Insight Check Results

- The key insight here is that the minimum height the man must jump is directly related to the height of the hurdle. Since the man steps over a hurdle of height 0.5m, he must jump at least 0.5m to clear it.

### d. Rephrased Question

What is the minimum height the man could have jumped to clear a 0.5m hurdle?

### e. Known and Deduced Information

**Explicit Facts:**
- "A man steps over a 0.5m hurdle." - The height of the hurdle is 0.5 meters.
  
**Feature-Based Deduction:**
- Deduction: The minimum height the man must have jumped is 0.5 meters.
  - Logical Basis: To clear a hurdle, the height of the jump must be at least equal to the height of the hurdle.
  - Based on: "0.5m hurdle" and the action "steps over."

### f. Problem Decomposition

**Component/Subcomponent:** Height of Jump
- **Supporting Quote(s):** "0.5m hurdle"
- **Deduction (if applicable):** The height of the jump must be at least 0.5m to clear a hurdle of 0.5m.
- **Based on:** The hurdle's height dictates the minimum height required for the jump.

### g. Graph of Thought

**Key Concepts:**
1. Man
2. Hurdle
3. Jump

**Relationships:**
- The jump height must be greater than or equal to the hurdle height for the man to successfully clear it.

**Construct the Graph:**
- **Nodes:** Man, Hurdle, Jump Height
- **Edges:** Jump Height -> Must be greater than or equal to Hurdle Height
  - **Man** -> **Jump Height** (Must clear **Hurdle**)

**Explore Pathways:**
- The path between Jump Height and Hurdle Height confirms that the minimum Jump Height is equal to Hurdle Height.

### h. Spatial and Object Analysis

**Objects Identified:**
- Man
- Hurdle

**Spatial Relationships:**
- The man must jump over the hurdle, meaning the height of the jump must be at least as high as the hurdle.

**Movement and Dynamics:**
- The man's jump must match or exceed the height of the hurdle to clear it.

**Constraints:**
- The jump height must be at least 0.5 meters.

**Visualization:**
- Picture a hurdle of 0.5m in height. The minimum height of the jump is the same as the height of the hurdle.

### i. Bayesian Updates

- No new information has been provided beyond the explicit problem statement. No changes to beliefs or conclusions are needed.

### j. Ethical Check and Decision-Making Under Uncertainty

- **Assessment:** The problem does not involve severe risks or ethical considerations.
- **Decision:** Based on the information given, the minimum jump height needed is directly calculated from the hurdle height.

### k. Multiple Solutions

- The only solution based on the explicit information is that the minimum height of the jump required to clear a 0.5m hurdle is 0.5m.

### l. Quickest and Easiest Solution

- Based on the explicit information, the simplest and safest answer is that the minimum height the man could have jumped is exactly 0.5 meters.

### m. Reflection

- The analysis correctly identifies that the jump height must be at least equal to the height of the hurdle. No assumptions beyond the explicit problem statement have been made. The conclusion aligns with the given information.

**Recommendation:**

The minimum height the man could have jumped to clear a 0.5m hurdle is **0.5 meters**. This is based on the direct requirement to match or exceed the hurdle's height to clear it successfully.
\end{lstlisting}
\end{quote}
\noindent\hrulefill

\noindent\textbf{D.29 Zero-Shot Google Gemini 1.5 Pro Response - Score: 100\%}
\begin{quote}
\begin{lstlisting}[basicstyle=\ttfamily\small, breaklines=true]
The minimum height the man could have jumped is 0.5 meters, the height of the hurdle itself. 

"Stepping over" implies that his feet didn't leave the ground, or at least didn't go any higher than the hurdle.
\end{lstlisting}
\end{quote}
\noindent\hrulefill

\noindent\textbf{D.29 REAP Google Gemini 1.5 Pro Response - Score: 100\%}
\begin{quote}
\begin{lstlisting}[basicstyle=\ttfamily\small, breaklines=true]
# REAP Results:

a. **Comprehensive Feature List** (with explicit quotes)

* "A man steps over a 0.5m hurdle."
    - This indicates the man's action and the height of the hurdle.
* "What is the minimum height he could have jumped?"
    - This is the question the problem asks us to answer.

b. **Sequential and Mechanical Process Check**

* The man performs a single action: "steps over" the hurdle.
* This action implies a sequence: the man lifts his leg, clears the hurdle, and then places his leg back down on the other side.
* The sequence itself doesn't lock in a specific jump height, only that the height must be sufficient to clear the hurdle.

c. **Key Insight Check Results**

* The key phrase is "steps over." This implies the man's leg cleared the hurdle, but not necessarily by a large margin.
* The minimum height he could have jumped is slightly higher than the hurdle's height.

d. **Rephrased Question:** (The core question in its simplest form)

* How high did the man's leg need to be to clear the 0.5m hurdle?

e. **Known and Deduced Information**

* Explicit Facts: 
    * "[A man steps over a 0.5m hurdle.]" 
        - The man successfully cleared the hurdle, meaning some part of his leg was higher than 0.5m at some point during the step.
    * "[What is the minimum height he could have jumped?]" 
        - The question asks for the lowest possible height his leg reached to achieve the "steps over" action.

* Feature-Based Deduction:
    * Deduction: The man's leg was higher than 0.5m at some point.
        - Logical Basis: To "step over" something, a part of the leg must be lifted higher than the object.
        - Based on: "[A man steps over a 0.5m hurdle.]"

f. **Problem Decomposition**

* Component/Subcomponent: Man's action of stepping over
    * Supporting Quote(s): "[A man steps over a 0.5m hurdle.]"
* Component/Subcomponent: Hurdle height
    * Supporting Quote(s): "[A man steps over a 0.5m hurdle.]"
* Component/Subcomponent: Minimum jump height (question)
    * Supporting Quote(s): "[What is the minimum height he could have jumped?]"
* Interaction: The man's leg must clear the hurdle's height to step over it.

g. **Graph of Thought** (describe connections & insights)

* Key Concepts: Man, Hurdle, Stepping Over, Minimum Jump Height
* Relationships:
    - Man --performs--> Stepping Over
    - Stepping Over --requires clearing--> Hurdle
    - Hurdle --has height of--> 0.5m
    - Minimum Jump Height --is slightly higher than--> Hurdle
* Insights:
    - The focus is on the height the man's leg needs to reach, not his entire body.
    - "Steps over" implies minimal clearance, not a large jump.

h. **Spatial and Object Analysis**

* Objects: Man, Hurdle
* Spatial Relationships:
    - The hurdle is positioned in front of the man.
    - At some point, the man's leg is above the hurdle.
* Movement and Dynamics:
    - The man lifts his leg to clear the hurdle and continues stepping.
* Constraints:
    - The hurdle's height is 0.5m.
* Object Properties:
    - Hurdle: Height = 0.5m
* Spatial Implications:
    - The minimum height the man's leg needs to reach is slightly above 0.5m.

i. **Bayesian Updates** 

* No new information is presented to update prior beliefs.

j. **Ethical Check and Decision-Making Under Uncertainty**

* No ethical concerns are present in this problem.

k. **Multiple Solutions**

* There's only one logical solution: the man's leg must have been slightly higher than 0.5m to clear the hurdle.
* The problem asks for the minimum height, so we focus on the lowest possible height that fulfills the "steps over" condition.

l. **Quickest and Easiest Solution**

* The minimum height the man could have jumped is slightly above 0.5m.
* Reasoning: To "step over" the 0.5m hurdle, his leg must have cleared it, meaning it was higher than 0.5m at some point. The question asks for the minimum, so we assume the slightest clearance possible.

m. **Reflection**

* The analysis strictly adheres to the given information.
* No assumptions were made beyond the literal meaning of "steps over."
* The conclusion directly answers the question asked.
* The solution prioritizes the simplest interpretation consistent with the problem statement.

**Recommendation:**

The minimum height the man could have jumped is slightly above 0.5m. 
\end{lstlisting}
\end{quote}
\noindent\hrulefill

\noindent\textbf{D.29 Zero-Shot Claude 3.5 Sonnet response - Score: 0\%}
\begin{quote}
\begin{lstlisting}[basicstyle=\ttfamily\small, breaklines=true]
To answer this question, we need to think through it step-by-step:

1. The hurdle is 0.5 meters high.

2. To clear the hurdle, the man's feet need to go over it without touching.

3. The minimum height the man jumps would be just enough to clear the hurdle.

4. However, the man doesn't need to raise his entire body by 0.5 meters. He only needs to raise his feet to clear the hurdle.

5. When a person jumps, they typically lift their feet up towards their body, not keeping their legs straight.

6. The exact minimum height would depend on how much the man can lift his feet while jumping, which we don't know.

7. In an extreme case, if the man could somehow lift his feet all the way up to his hips while jumping, he might only need to raise his center of mass by a very small amount.

Given these considerations, we can conclude:

The minimum height the man could have jumped is actually less than 0.5 meters. It could be significantly less, depending on his jumping technique and flexibility. Without more information about the man's physical capabilities, we can't give an exact figure, but it's safe to say it's less than the height of the hurdle itself.

Would you like me to elaborate on any part of this explanation?
\end{lstlisting}
\end{quote}
\noindent\hrulefill

\noindent\textbf{D.29 REAP Claude 3.5 Sonnet Response - Score: 100\%}
\begin{quote}
\begin{lstlisting}[basicstyle=\ttfamily\small, breaklines=true]
# REAP Results:

a. **Comprehensive Feature List** (with explicit quotes):
- "A man steps over a 0.5m hurdle."
- "What is the minimum height he could have jumped?"

b. **Sequential and Mechanical Process Check**:
The problem describes a single action of stepping over a hurdle. There is no complex sequence or mechanical process involved.

c. **Key Insight Check Results**:
There is no immediate key insight that provides a direct answer. The problem requires careful consideration of the given information.

d. **Rephrased Question:** 
What is the lowest vertical distance the man's feet must clear to step over the hurdle?

e. **Known and Deduced Information**:
Explicit Facts:
- "A man steps over a 0.5m hurdle." - This states the action and the height of the obstacle.
- "What is the minimum height he could have jumped?" - This specifies what we need to determine.

Feature-Based Deduction:
Deduction: The man must clear the hurdle to step over it.
Logical Basis: To step over an object, one must raise their feet higher than the object's height.
Based on: "A man steps over a 0.5m hurdle."

f. **Problem Decomposition**:
Component: Hurdle height
Supporting Quote: "0.5m hurdle"

Component: Man's action
Supporting Quote: "A man steps over"

Component: Minimum jump height
Supporting Quote: "What is the minimum height he could have jumped?"

g. **Graph of Thought** (describe connections & insights):
Node 1: Man
Node 2: Hurdle (0.5m high)
Edge: Steps over (connects Man to Hurdle)
Node 3: Minimum jump height (unknown, to be determined)
Edge: Must exceed (connects Minimum jump height to Hurdle height)

h. **Spatial and Object Analysis**:
Objects:
1. Man
2. Hurdle (0.5m high)

Spatial Relationships:
- The man is described as stepping over the hurdle, implying he must move from one side of the hurdle to the other.
- The hurdle is 0.5m high, creating a vertical obstacle that the man must clear.

Movement:
- The man steps over the hurdle, indicating an upward and forward motion to clear the obstacle.

i. **Bayesian Updates**:
No new information is provided beyond the initial problem statement, so no Bayesian updates are necessary.

j. **Ethical Check and Decision-Making Under Uncertainty**:
There are no significant ethical considerations or severe risks associated with this problem. The question is purely about physical movement and does not involve any potential harm.

k. **Multiple Solutions**:
Given the limited information provided, there is only one logical solution path:
The minimum height the man could have jumped must be just enough to clear the hurdle.

l. **Quickest and Easiest Solution**:
Based solely on the explicit information given, the simplest answer is that the minimum height the man could have jumped is slightly higher than 0.5m. The problem states that the hurdle is 0.5m high, and the man steps over it. To clear the hurdle, he must jump higher than its height.

m. **Reflection**:
Upon review, our analysis adheres strictly to the given information without making unwarranted assumptions. We've focused on the explicit facts: the hurdle height and the action of stepping over. However, it's important to note that the problem doesn't provide information about the exact technique used (jumping vs. stepping) or the man's physical characteristics, which could potentially influence the minimum height needed to clear the hurdle.

**Recommendation**:
Based on the explicitly stated information and prioritizing a simple, safe interpretation, the recommendation is:

The minimum height the man could have jumped is slightly above 0.5m. 

This recommendation directly answers the question asked while adhering to the given facts. It acknowledges that to step over a 0.5m hurdle, one must raise their feet higher than 0.5m. However, it's important to note that the problem doesn't specify how much higher than 0.5m the man needs to jump, nor does it provide information about the man's physical characteristics or exact movement technique. The term "jumped" in the question might not accurately represent the action of "stepping over" mentioned in the problem statement, but we've interpreted it as the vertical clearance needed to step over the hurdle.
\end{lstlisting}
\end{quote}
\noindent\hrulefill

\label{subsec:question30}
\subsection{\textbf{\large Question 30: I walk to my friend's house, averaging 3mph. How fast would I have to run back to double my average speed for the entire trip?}}
\label{sec:question30}

\noindent\hrulefill 

\noindent\textbf{Human Response - Score 100\%:}
\begin{quote}
Instantaneously fast, it is mathematically impossible.
\end{quote}

\noindent\hrulefill 

\noindent\textbf{D.30 Zero-Shot OpenAI o1-preview Response - Score: 100\%}
\begin{quote}
\begin{lstlisting}[basicstyle=\ttfamily\small, breaklines=true]
To solve this problem, we'll use the formula for average speed, which is:

\[
\text{Average Speed} = \frac{\text{Total Distance}}{\text{Total Time}}
\]

Let's denote:
- The distance to your friend's house as \( D \) miles.
- The speed going to your friend's house as \( v_1 = 3 \) mph.
- The speed returning from your friend's house as \( v_2 \) mph (this is what we're solving for).

The total distance for the round trip is \( 2D \) miles.

**Step 1: Calculate the total time.**

Time to go to your friend's house:
\[
t_1 = \frac{D}{v_1} = \frac{D}{3}
\]

Time to return from your friend's house:
\[
t_2 = \frac{D}{v_2}
\]

Total time for the trip:
\[
t_{\text{total}} = t_1 + t_2 = \frac{D}{3} + \frac{D}{v_2}
\]

**Step 2: Set up the average speed equation.**

We want the average speed for the entire trip to be \( 6 \) mph (double the initial speed):

\[
6 = \frac{2D}{t_{\text{total}}} = \frac{2D}{\frac{D}{3} + \frac{D}{v_2}} = \frac{2D}{D\left(\frac{1}{3} + \frac{1}{v_2}\right)}
\]

Simplify:
\[
6 = \frac{2}{\frac{1}{3} + \frac{1}{v_2}}
\]

**Step 3: Solve for \( v_2 \).**

First, compute the denominator:
\[
\frac{1}{3} + \frac{1}{v_2} = \frac{v_2 + 3}{3v_2}
\]

So the equation becomes:
\[
6 = \frac{2}{\frac{v_2 + 3}{3v_2}} = \frac{2 \cdot 3v_2}{v_2 + 3} = \frac{6v_2}{v_2 + 3}
\]

Set up the equation:
\[
6 = \frac{6v_2}{v_2 + 3}
\]

Divide both sides by 6:
\[
1 = \frac{v_2}{v_2 + 3}
\]

Multiply both sides by \( v_2 + 3 \):
\[
v_2 + 3 = v_2
\]

Subtract \( v_2 \) from both sides:
\[
3 = 0
\]

This is a contradiction, which means our assumption that a finite \( v_2 \) can achieve double the average speed is incorrect.

**Conclusion:**

To double the average speed of the entire trip, you would need an infinite return speed. It's impossible to achieve this with any finite speed.

Answer: An infinite speed-doubling the average requires an impossible, infinitely fast return trip.
\end{lstlisting}
\end{quote}

\noindent\hrulefill 

\noindent\textbf{D.30 REAP OpenAI o1-preview Response - Score: 100\%}
\begin{quote}
\begin{lstlisting}[basicstyle=\ttfamily\small, breaklines=true]
# REAP Results:

**a. Comprehensive Feature List:**
- "I walk to my friend's house, averaging 3mph."
- "How fast would I have to run back to double my average speed for the entire trip?"

**b. Sequential and Mechanical Process Check:**
- Sequence of events:
  - First action: "I walk to my friend's house, averaging 3mph."
  - Second action: "run back" (speed to be determined).
- The total trip consists of the outbound journey and the return journey.
- The average speed for the entire trip depends on the total distance and total time.

**c. Key Insight Check Results:**
- To double the average speed for the entire trip to 6mph, the total time must be equal to the time it would take to cover the round-trip distance at 6mph.
- The time already spent walking to the friend's house at 3mph equals the total allowed time for the entire trip at the desired average speed.

**d. Rephrased Question:**
- What speed must I run back at to make my overall average speed for the round trip 6mph?

**e. Known and Deduced Information:**
- **Explicit Facts:**
  - "I walk to my friend's house, averaging 3mph."
  - Total distance of the round trip is twice the distance to the friend's house.
- **Deductions:**
  - Let the one-way distance be \( d \) miles.
  - Time to walk to the friend's house: \( t_1 = \frac{d}{3} \) hours.
  - Desired total time for round trip at 6mph: \( T = \frac{2d}{6} = \frac{d}{3} \) hours.
  - Time available for return trip: \( T - t_1 = \frac{d}{3} - \frac{d}{3} = 0 \) hours.

**f. Problem Decomposition:**
- **Component:** Outbound journey.
  - Supporting Quote: "I walk to my friend's house, averaging 3mph."
- **Component:** Return journey.
  - Supporting Quote: "How fast would I have to run back to double my average speed for the entire trip?"
  - Deduction: Requires returning in zero time to achieve average speed of 6mph.
- **Interaction:** The return speed must be infinitely fast, which is impossible.

**g. Graph of Thought:**
- **Nodes:** Outbound journey at 3mph, return journey at unknown speed, total distance, total time, desired average speed.
- **Connections:** Outbound time + Return time = Total time; Total distance / Total time = Average speed.

**h. Spatial and Object Analysis:**
- Objects: "I", "my friend's house".
- Spatial relationship: Traveling to and from the friend's house over the same distance.
- Movement: Walking to the friend's house, running back.

**i. Bayesian Updates:**
- Not applicable as all deductions are based on explicit information without probabilistic uncertainty.

**j. Ethical Check and Decision-Making Under Uncertainty:**
- Not applicable; the problem involves mathematical impossibility, not ethical considerations.

**k. Multiple Solutions:**
1. **Conclusion:** It's impossible to double the average speed to 6mph by running back at any finite speed.
2. **Alternative:** Accept that the average speed cannot be doubled under the given conditions.
3. **Observation:** Recognize that attempting to do so would require infinite speed, which is unattainable.

**l. Quickest and Easiest Solution:**
- It's impossible to achieve the desired average speed; no finite return speed will suffice.

**m. Reflection:**
- The time taken to walk to the friend's house consumes all the time allocated for the entire trip at the desired average speed.
- Therefore, to double the average speed, one would need to return instantaneously, which is impossible.
- The analysis is logically sound and based solely on the explicit information provided.

---

**Recommendation:**

Based on the explicit information, it is impossible to double your average speed for the entire trip by running back at any finite speed. To achieve an average speed of 6mph for the round trip, you would need to return in zero time, which is not possible. Therefore, no matter how fast you run back, you cannot double the average speed for the entire trip.
\end{lstlisting}
\end{quote}
\noindent\hrulefill 

\noindent\textbf{D.30 Zero-Shot OpenAI o1-mini Response - Score: 100\%}
\begin{quote}
\begin{lstlisting}[basicstyle=\ttfamily\small, breaklines=true]
Let's analyze the problem step by step to determine whether it's possible to double your average speed for the entire trip.

### Given:
- **Walking speed to your friend's house:** 3 mph
- **Running speed back:** \( v \) mph (this is what we're trying to find)
- **Objective:** Double the average speed for the entire trip.

### Definitions:
1. **Distance each way:** Let's denote the one-way distance as \( D \).
2. **Time to walk to your friend's house:** \( \frac{D}{3} \) hours.
3. **Time to run back:** \( \frac{D}{v} \) hours.
4. **Total distance for the round trip:** \( 2D \).
5. **Total time for the round trip:** \( \frac{D}{3} + \frac{D}{v} \).

### Calculating Average Speed:
The average speed \( S_{\text{avg}} \) for the entire trip is given by:
\[
S_{\text{avg}} = \frac{\text{Total Distance}}{\text{Total Time}} = \frac{2D}{\frac{D}{3} + \frac{D}{v}} = \frac{2}{\frac{1}{3} + \frac{1}{v}} = \frac{6v}{v + 3}
\]

### Setting Up the Equation:
You want the average speed \( S_{\text{avg}} \) to be double the walking speed, which is \( 2 \times 3 = 6 \) mph.

\[
\frac{6v}{v + 3} = 6
\]

### Solving for \( v \):
\[
\frac{6v}{v + 3} = 6 \\
6v = 6(v + 3) \\
6v = 6v + 18 \\
0 = 18
\]

### Conclusion:
The equation \( 0 = 18 \) is a contradiction, which means **there is no finite running speed \( v \) that can achieve double the average speed for the entire trip**. 

**In other words, it's impossible to double the average speed for the entire trip by running back.**

### Intuitive Explanation:
To double your average speed for a round trip where you walk one way at 3 mph, you'd need to run back infinitely fast. Since infinite speed is not achievable in reality, doubling the average speed is impossible.

**Final Answer:** It's impossible; no finite running speed can double the trip's average speed.
\end{lstlisting}
\end{quote}

\noindent\hrulefill 

\noindent\textbf{D.30 REAP OpenAI o1-mini Response - Score: 100\%}
\begin{quote}
\begin{lstlisting}[basicstyle=\ttfamily\small, breaklines=true]
# REAP Results:

### a. **Comprehensive Feature List**
- **Objects Mentioned:**
  - "friend's house"
  
- **Actors or Entities Involved:**
  - "I" (the person walking and running)
  
- **Actions Described:**
  - "walk to my friend's house"
  - "run back"
  
- **Directional, Spatial, Relational, and Sequential Phrases:**
  - "walk to my friend's house"
  - "run back"
  - "for the entire trip"
  
- **Stated Conditions or Constraints:**
  - "averaging 3mph" (walking speed)
  - "double my average speed for the entire trip"
  
- **Numerical Values or Quantities Mentioned:**
  - "3mph" (walking speed)
  
- **Temporal Information:**
  - Not explicitly stated.
  
- **Spatial Information:**
  - Destination is "my friend's house"; return is "back."
  
- **Causal Relationships:**
  - None explicitly stated.
  
- **Stated Goals or Objectives:**
  - "How fast would I have to run back to double my average speed for the entire trip?"

### b. **Sequential and Mechanical Process Check**
- **Sequence of Events:**
  1. Walk to friend's house at 3mph.
  2. Run back.
  
- **Influence on Outcome:**
  - The speed of running back affects the overall average speed for the entire trip.
  
- **Sequence Implies Particular Outcome:**
  - Running back at a specific speed is required to double the average speed for the entire trip.

### c. **Key Insight Check Results**
- **Key Insight Identified:**
  - To double the average speed for the entire trip, one must calculate the required running speed based on the walking speed and the relationship between average speed, total distance, and total time.
  
- **Explanation:**
  - The average speed for the entire trip depends on both the walking speed and the running speed. Doubling the average speed requires a specific running speed that compensates for the slower walking speed.

### d. **Rephrased Question:**
What running speed is needed on the return trip to achieve an overall average speed of 6mph for the entire journey?

### e. **Known and Deduced Information**
- **Explicit Facts:**
  - "I walk to my friend's house, averaging 3mph."
  - "run back"
  - "double my average speed for the entire trip."
  
- **Feature-Based Deduction:**
  - **Deduction:** The required average speed for the entire trip is 6mph.
    - **Logical Basis:** "double my average speed for the entire trip" implies doubling the walking speed of 3mph.
    - **Based on:** "averaging 3mph" (walking speed), "double my average speed for the entire trip"

### f. **Problem Decomposition**
- **Component/Subcomponent:** Walking to friend's house
  - **Supporting Quote(s):** "walk to my friend's house, averaging 3mph."
  
- **Component/Subcomponent:** Running back home
  - **Supporting Quote(s):** "run back"
  
- **Component/Subcomponent:** Achieving average speed
  - **Supporting Quote(s):** "double my average speed for the entire trip."

### g. **Graph of Thought** (describe connections & insights)
- **Nodes:**
  - Walking speed: 3mph
  - Running speed: ?
  - Average speed for entire trip: 6mph
  
- **Connections:**
  - Walking speed and running speed determine the average speed.
  - The goal is to find the running speed that, combined with the walking speed, results in an average speed of 6mph.

### h. **Spatial and Object Analysis**
- **Objects:**
  - "friend's house"
  - "home" (implied as the starting point)
  
- **Spatial Relationships:**
  - "walk to my friend's house" implies a one-way trip from home to friend's house.
  - "run back" implies returning from friend's house to home.
  
- **Movement and Dynamics:**
  - Movement from home to friend's house at walking speed.
  - Movement from friend's house to home at running speed.
  
- **Constraints:**
  - Walking speed is fixed at 3mph.
  - The entire trip's average speed must be doubled to 6mph.

### i. **Bayesian Updates**
- **Initial Belief:** Running speed varies and needs to be calculated.
- **Updated Belief:** Based on the need to double the average speed, running speed is specifically determined by the relationship between walking speed and required average speed.

### j. **Ethical Check and Decision-Making Under Uncertainty**
- **Assess Known Information:** Walking speed is 3mph; need to double average speed to 6mph.
- **Identify Risks:** None explicitly stated; likely safe activity.
- **Probability Consideration:** Calculations based on given speeds.
- **Risk Aversion:** Choose a running speed that is realistically achievable.
- **Ethical and Safety Considerations:** Ensure running speed is within safe and realistic limits.
- **Decision Recommendation:** Calculate the running speed based on provided information to ensure the average speed meets the requirement.

### k. **Multiple Solutions**
1. **Solution 1: Mathematical Calculation**
   - **Supporting Information:** "walk to my friend's house, averaging 3mph"; "double my average speed for the entire trip."
   - **Effect:** Directly calculates the required running speed to achieve the desired average speed.
   
2. **Solution 2: Theoretical Analysis**
   - **Supporting Information:** Similar to Solution 1 but approached through theoretical average speed formulas.
   
3. **Solution 3: Proportional Reasoning**
   - **Supporting Information:** Uses proportions between walking and running speeds to find the required speed.

### l. **Quickest and Easiest Solution**
Calculate the required running speed using the formula for average speed:

\[
\text{Average Speed} = \frac{2 \times \text{Walking Speed} \times \text{Running Speed}}{\text{Walking Speed} + \text{Running Speed}}
\]

Set the average speed to 6mph and solve for the running speed.

### m. **Reflection**
All solutions are based strictly on the explicit information provided. No assumptions beyond the given walking speed and the goal to double the average speed were made. The analysis directly addresses the question by focusing on the relationship between walking speed, running speed, and average speed.

### **Recommendation:**
To double the average speed for the entire trip from 3mph to 6mph, you would need to run back at an infinitely fast speed. This is because the time taken to walk to the friend's house and return must be halved to achieve double the average speed. However, running at an infinite speed is impossible. Therefore, it is not feasible to double the average speed for the entire trip under the given conditions.
\end{lstlisting}
\end{quote}

\noindent\hrulefill 

\noindent\textbf{D.30 Zero-Shot OpenAI GPT-4o Response - Score: 100\%}
\begin{quote}
\begin{lstlisting}[basicstyle=\ttfamily\small, breaklines=true]
To solve this problem, let's break it down step by step.

1. **Determine the distance and time to your friend's house:**
   - Let the distance to your friend's house be \( d \).
   - At an average walking speed of 3 mph, the time taken to walk to your friend's house is:
     \[
     t_1 = \frac{d}{3}
     \]

2. **Determine the total time if you want to double your average speed for the entire trip:**
   - The total distance for the round trip is \( 2d \).
   - To double the average speed for the entire trip, you want the average speed for the round trip to be \( 6 \) mph.
   - The total time for the round trip at 6 mph would be:
     \[
     t_{total} = \frac{2d}{6} = \frac{d}{3}
     \]
   - This is the same amount of time it took you to walk to your friend's house.

3. **Calculate the required running speed:**
   - The time available for the return trip is:
     \[
     t_2 = t_{total} - t_1 = \frac{d}{3} - \frac{d}{3} = 0
     \]
   - To achieve a time of zero for the return trip, which is physically impossible, you would need to run infinitely fast.

**Conclusion:**
It is impossible to double your average speed for the entire trip by running back, because doubling the average speed would require you to travel back instantly, which would require an infinite speed.
\end{lstlisting}
\end{quote}
\noindent\hrulefill

\noindent\textbf{D.30 REAP OpenAI GPT-4o Response - Score: 100\%}
\begin{quote}
\begin{lstlisting}[basicstyle=\ttfamily\small, breaklines=true]
## REAP Results:

### a. **Comprehensive Feature List** (with explicit quotes)
- "I walk to my friend's house, averaging 3mph." - This states the speed at which you walk to your friend's house.
- "How fast would I have to run back" - This implies the need to calculate the speed for the return trip.
- "to double my average speed for the entire trip?" - This sets the goal: doubling the overall average speed for both legs of the trip.

### b. **Sequential and Mechanical Process Check**
- The problem involves two sequential actions: walking to the friend's house and then running back.
- The first action has a known speed ("3mph"), and the second requires determining the speed necessary to achieve a specific average speed over the entire trip.
- The overall average speed is calculated based on the total distance and total time for both legs of the trip.

### c. **Key Insight Check Results**
- **Key Insight:** The average speed for the entire trip is not simply the average of the speeds in each direction. Instead, it's based on the total distance divided by the total time.
- To double the average speed of the entire trip, the time taken to return must be short enough to make the average speed across both legs twice the original speed (3mph).

### d. **Rephrased Question:** 
How fast must you run on the return trip to make the overall average speed for the round trip 6mph?

### e. **Known and Deduced Information**
- "[You] walk to [your] friend's house, averaging 3mph." - This provides the speed for the first leg.
- "[You need] to double [your] average speed for the entire trip." - This sets the requirement for the average speed over both legs of the trip.
  
**Deduction:**
- The total distance for the round trip is twice the one-way distance.
- The total time spent for the round trip must be such that the average speed over the total distance is 6mph.

### f. **Problem Decomposition**
- **Component:** Walking to the friend's house.
  - **Supporting Quote:** "I walk to my friend's house, averaging 3mph."
- **Component:** Running back from the friend's house.
  - **Supporting Quote:** "How fast would I have to run back."
  - **Goal:** Doubling the overall average speed.
    - **Supporting Quote:** "to double my average speed for the entire trip."

### g. **Graph of Thought**
- **Nodes:** 
  1. Walking to the friend's house at 3mph.
  2. Returning from the friend's house.
  3. Overall average speed target (6mph).
- **Connections:** 
  - The speed of return is directly connected to achieving the 6mph average for the round trip.
  - The total time spent must result in a round trip average of 6mph.

### h. **Spatial and Object Analysis**
- **Objects:** The road/path between your house and your friend's house.
- **Spatial Relationship:** The distance from your house to your friend's house is the same for both legs of the trip.
- **Movement and Dynamics:** Walking at 3mph for the first leg; need to determine the running speed for the return.

### i. **Bayesian Updates**
- **Initial Belief:** Doubling the average speed involves significant speed on the return.
- **Updated Belief:** Running back at any finite speed cannot achieve the doubling of the average speed due to time constraints.

### j. **Ethical Check and Decision-Making Under Uncertainty**
- **Assessment:** No ethical issues or risks are involved in calculating running speed.
- **Decision Recommendation:** Purely based on logical analysis and mathematical constraints.

### k. **Multiple Solutions**
1. **Mathematical Consideration:** 
   - **Conclusion:** No finite speed can double the average speed to 6mph.
   - **Rationale:** The time required to walk to the friend's house already sets a limit. To double the average speed, the return time would have to be zero or negative, which is impossible.
  
### l. **Quickest and Easiest Solution**
- **Conclusion:** The goal of doubling the average speed to 6mph is impossible under the given conditions.
- **Reasoning:** Based on the time-distance relationship, no running speed can make up for the time already spent walking at 3mph.

### m. **Reflection**
- **Review:** The problem directly answers the question based on the constraints of speed and time.
- **Check:** No assumptions were made beyond what is explicitly stated, and the conclusion is logically sound.

### Recommendation:
Given that it's mathematically impossible to achieve the desired outcome of doubling the average speed to 6mph, the recommendation is to acknowledge the impossibility of the task based on the problem's conditions. No finite speed can accomplish this due to the limitations imposed by the time spent walking at 3mph.
\end{lstlisting}
\end{quote}
\noindent\hrulefill

\noindent\textbf{D.30 Zero-Shot OpenAI GPT-4o-mini Response - Score: 0\%}
\begin{quote}
\begin{lstlisting}[basicstyle=\ttfamily\small, breaklines=true]
To solve this problem, let's start by analyzing the trip in detail.

1. **Define the variables:**
   - Let \( d \) be the distance between your house and your friend's house.
   - Let \( t_1 \) be the time taken to walk to your friend's house.
   - Let \( t_2 \) be the time taken to run back home.
   - Let \( v_r \) be the running speed required to double your average speed for the entire trip.

2. **Calculate walking time and distance:**
   - Walking speed = 3 mph.
   - Time to walk to your friend's house, \( t_1 \), is given by \( t_1 = \frac{d}{3} \) hours.

3. **Calculate the total distance and time:**
   - Total distance for the round trip = \( 2d \).
   - Total time for the round trip is \( t_1 + t_2 = \frac{d}{3} + \frac{d}{v_r} \) hours.

4. **Determine the desired average speed:**
   - Average speed for the entire trip = \(\frac{\text{Total distance}}{\text{Total time}} = \frac{2d}{\frac{d}{3} + \frac{d}{v_r}}\).

   We want this average speed to be double the walking speed, which is \(2 \times 3 = 6\) mph. Thus, we set up the equation:

   \[
   \frac{2d}{\frac{d}{3} + \frac{d}{v_r}} = 6
   \]

5. **Solve for \( v_r \):**
   - Simplify the equation:

     \[
     \frac{2d}{\frac{d}{3} + \frac{d}{v_r}} = 6
     \]

     \[
     2d = 6 \left(\frac{d}{3} + \frac{d}{v_r}\right)
     \]

     \[
     2d = 2d + \frac{6d}{v_r}
     \]

     \[
     0 = \frac{6d}{v_r}
     \]

     \[
     \frac{6d}{v_r} = 2d
     \]

     \[
     \frac{6}{v_r} = 2
     \]

     \[
     v_r = \frac{6}{2}
     \]

     \[
     v_r = 3
     \]

     Here, there seems to be an issue; the result implies that \( v_r = 3 \) mph, which is the same as your walking speed and doesn't match the expectation of running faster.

6. **Recheck the calculations:**

     Let's review:

     \[
     \frac{2d}{\frac{1}{3} + \frac{1}{v_r}} = 6
     \]

     \[
     \frac{2}{\frac{1}{3} + \frac{1}{v_r}} = 6
     \]

     \[
     2 = 6 \left(\frac{1}{3} + \frac{1}{v_r}\right)
     \]

     \[
     2 = 2 + \frac{6}{v_r}
     \]

     \[
     0 = \frac{6}{v_r}
     \]

     This calculation error needs correction. Correct approach should be:

     \[
     6 \left(\frac{1}{3} + \frac{1}{v_r}\right) = 2
     \]

     Solving accurately:

     \[
     \frac{2}{\frac{1}{3} + \frac{1}{v_r}} = 6
     \]

     \[
     \frac{1}{3} + \frac{1}{v_r} = \frac{1}{3}
     \]

     Hence:
     \[
     v_r = 9 \text{ mph}
     \]

To double your average speed for the entire trip to 6 mph, you need to run back at a speed of 9 mph.
\end{lstlisting}
\end{quote}
\noindent\hrulefill

\noindent\textbf{D.30 REAP OpenAI GPT-4o-mini Response - Score: 100\%}
\begin{quote}
\begin{lstlisting}[basicstyle=\ttfamily\small, breaklines=true]
# REAP Results:

**a. Comprehensive Feature List:**

1. "I walk to my friend's house" - Action of walking to the friend's house.
2. "averaging 3mph" - Walking speed is 3 miles per hour.
3. "How fast would I have to run back" - Question about the speed required for the return trip.
4. "to double my average speed for the entire trip" - The goal is to double the average speed for the whole journey.

**b. Sequential and Mechanical Process Check:**

1. Walking to the friend's house at 3 mph.
2. Running back to achieve the goal of doubling the average speed for the entire trip.

Key Questions:
- Does the sequence of walking and running lock in a particular outcome for the average speed?
- The return trip speed must be calculated to meet the requirement of doubling the average speed.

**c. Key Insight Check Results:**

The key insight is that the average speed for the entire trip needs to be doubled. The required return speed must be calculated to achieve this average.

**d. Rephrased Question:**

How fast must I run back to double my average speed for the entire trip if I walked to my friend's house at 3 mph?

**e. Known and Deduced Information:**

- Known: "I walk to my friend's house averaging 3mph."
- Goal: To double the average speed for the entire trip.

**f. Problem Decomposition:**

1. **Trip Segments:**
   - Walking Segment: Distance = \( D \), Speed = 3 mph.
   - Running Segment: Distance = \( D \), Speed = \( R \) mph.

2. **Average Speed Calculation:**
   - Total Distance = \( D + D = 2D \).
   - Total Time = Walking Time + Running Time.
   - Walking Time = \( \frac{D}{3} \).
   - Running Time = \( \frac{D}{R} \).

3. **Average Speed Formula:**
   - Average Speed = \( \frac{2D}{\frac{D}{3} + \frac{D}{R}} \).

4. **Doubling the Average Speed:**
   - Let the new average speed be \( 2 \times 3 \text{ mph} = 6 \text{ mph} \).

**g. Graph of Thought:**

- Nodes:
  - Walking Segment (Speed = 3 mph)
  - Running Segment (Speed = \( R \) mph)
  - Average Speed (Target = 6 mph)
  
- Edges:
  - Walking Segment -> Average Speed
  - Running Segment -> Average Speed

**h. Spatial and Object Analysis:**

- Objects: The journey segments (walking and running).
- No explicit spatial relationships are provided.

**i. Bayesian Updates:**

No new information beyond what is explicitly stated in the problem.

**j. Ethical Check and Decision-Making Under Uncertainty:**

- No severe risks or ethical concerns are involved in determining the speed.

**k. Multiple Solutions:**

1. **Solution Calculation:**

   To find \( R \), set the average speed equation to 6 mph:

   \[
   6 = \frac{2D}{\frac{D}{3} + \frac{D}{R}}
   \]

   Simplify:

   \[
   6 = \frac{2}{\frac{1}{3} + \frac{1}{R}}
   \]

   Solving for \( R \):

   \[
   \frac{1}{\frac{1}{3} + \frac{1}{R}} = \frac{6}{2} = 3
   \]

   \[
   \frac{1}{3} + \frac{1}{R} = \frac{1}{3}
   \]

   \[
   \frac{1}{R} = \frac{1}{3} - \frac{1}{3}
   \]

   \[
   \frac{1}{R} = 0
   \]

   The problem implies running speed must be infinite, which is impractical. Therefore, there is no feasible speed to double the average speed under these conditions.

**l. Quickest and Easiest Solution:**

Based on the problem constraints and calculations, it's concluded that no finite running speed will double the average speed for the entire trip if the walking speed is 3 mph.

**m. Reflection:**

The problem demonstrates that doubling the average speed for the whole trip is not feasible given the constraints. The calculations show that the required running speed would be impractically high.

**Recommendation:**

Based on the explicit information provided, there is no feasible running speed that will double the average speed for the entire trip if the walking speed is 3 mph.
\end{lstlisting}
\end{quote}
\noindent\hrulefill

\noindent\textbf{D.30 Zero-Shot Google Gemini 1.5 Pro Response - Score: 100\%}
\begin{quote}
\begin{lstlisting}[basicstyle=\ttfamily\small, breaklines=true]
Here's how to solve this problem:

**1. Understand the Problem**

You're trying to find the speed you need to run back home so that your average speed for the entire trip (walking there and running back) is double your walking speed.

**2. Let's use some variables:**

*  `d` = distance to your friend's house
*  `t1` = time it takes to walk to your friend's house 
*  `t2` = time it takes to run back home
*  `v` = speed you need to run back home

**3. Set up the equations:**

* Walking speed: 3 mph
* Desired average speed for the whole trip: 3 mph * 2 = 6 mph
* Time = Distance / Speed
    * `t1 = d / 3`
    * `t2 = d / v`
* Average speed for the whole trip: Total distance / Total time 
    * `6 = (d + d) / (t1 + t2)`
    * `6 = 2d / (d/3 + d/v)`

**4. Solve for 'v' (your running speed):**

* `6 = 2d / ((vd + 3d) / 3v)`
* `6 = (2d * 3v) / (vd + 3d)`
* `6(vd + 3d) = 6dv`
* `6vd + 18d = 6dv`
* `18d = 6dv - 6vd`
* `18d = 0` 

This leads to an impossible situation.  The reason is that to double your average speed for the entire trip when the return trip is the same distance as the outbound trip, you would need to have an infinite running speed on the way back.

**Final Answer:** It's not possible to double your average speed in this scenario.
\end{lstlisting}
\end{quote}
\noindent\hrulefill

\noindent\textbf{D.30 REAP Google Gemini 1.5 Pro Response - Score: 100\%}
\begin{quote}
\begin{lstlisting}[basicstyle=\ttfamily\small, breaklines=true]
# REAP Results:

### a. Comprehensive Feature List

* "[I] walk to [my] friend's house"
* "averaging 3mph"
* "How fast would [I] have to run back"
* "to double [my] average speed for the entire trip?"

### b. Sequential and Mechanical Process Check

* There is a sequence of two events:
    1. Walking to the friend's house at 3 mph.
    2. Running back from the friend's house at an unknown speed.
* The problem seeks to find the running speed that would result in an overall average speed that is double the walking speed.

### c. Key Insight Check Results

* No immediate key insight that leads to a direct solution is apparent. We'll proceed with a detailed analysis.

### d. Rephrased Question

What speed is needed for the return trip to double the average speed of the entire round trip?

### e. Known and Deduced Information

* Explicit Facts:
    * "[I] walk to [my] friend's house" - This implies there's a starting point (where 'I' am initially) and an endpoint (the friend's house).
    * "averaging 3mph" - This is the average speed during the walk to the friend's house
    * "How fast would [I] have to run back" - This indicates the return trip is done by running
    * "to double [my] average speed for the entire trip?" - The goal is to have the average speed of the whole trip (walking there and running back) be twice the walking speed (3 mph)

* Feature-Based Deduction:
    * Deduction: The desired average speed for the entire trip is 6 mph. 
    * Logical Basis: The problem states the goal is to "double [my] average speed for the entire trip," and the average speed of the first leg (walking there) is 3 mph
    * Based on: "to double [my] average speed for the entire trip?" , "averaging 3mph"

### f. Problem Decomposition

* Component/Subcomponent: Outbound trip (walking to the friend's house)
    * Supporting Quote(s): "[I] walk to [my] friend's house" 
    * Known attributes:
        * Average speed: 3 mph

* Component/Subcomponent: Return trip (running back from the friend's house)
    * Supporting Quote(s): "How fast would [I] have to run back"
    * Known attributes:
        * Mode of travel: Running
        * Speed: Unknown (this is what we need to find)

* Component/Subcomponent: Entire trip (walking there and running back)
    * Supporting Quote(s): "to double [my] average speed for the entire trip?"
    * Known attributes:
        * Desired average speed: 6 mph (deduced)

* Interaction between components:
    * The average speed of the entire trip is dependent on both the outbound trip's speed (3 mph) and the return trip's speed (unknown).
    * The problem requires finding the return trip's speed that results in the entire trip's average speed being 6 mph

### g. Graph of Thought

* Key Concepts:
    * Outbound Trip (Walking)
    * Return Trip (Running)
    * Entire Trip
    * Average Speed
    * Desired Average Speed (Double the walking speed)

* Relationships:
    * Outbound Trip -> Entire Trip
    * Return Trip -> Entire Trip
    * Outbound Trip Speed -> Average Speed
    * Return Trip Speed -> Average Speed
    * Average Speed -> Desired Average Speed

* Insights:
    * We need to find the Return Trip Speed that satisfies the condition: Average Speed = Desired Average Speed

### h. Spatial and Object Analysis

The problem primarily focuses on speed and time rather than spatial relationships between objects. The physical locations (starting point, friend's house) are mentioned but their exact positions or distances are not relevant to solving the problem

### i. Bayesian Updates

No Bayesian updates are applicable here as there are no probabilistic events or uncertain information to update beliefs on

### j. Ethical Check and Decision-Making Under Uncertainty

There are no ethical concerns or decisions involving uncertainty with severe potential risks in this problem

### k. Multiple Solutions

This is a math problem with a single correct solution. The solution involves understanding the relationship between average speed, distance and time

### l. Quickest and Easiest Solution

This problem requires a mathematical solution. We need to leverage the following:

* Average speed = Total distance / Total time
* We know the desired average speed for the entire trip and the speed of the outbound trip.
* We need to find the speed of the return trip

Let's denote

* `d` = distance between the starting point and the friend's house
* `t1` = time taken to walk to the friend's house
* `t2` = time taken to run back from the friend's house
* `v2` = speed of the return trip (running)

 we know

* Average speed for outbound trip (walking) = 3 mph = d/t1
* Desired average speed for the entire trip = 6 mph = 2d/(t1+t2)
* Speed for the return trip (running) = v2 = d/t2

 From 3 mph = d/t1 we get t1 = d/3

Substituting this into 6 mph = 2d/(t1+t2) we get

6 = 2d/((d/3)+t2)

 Simplifying further

6((d/3)+t2) = 2d

 2d + 6t2 = 2d

 6t2 = 0

This equation implies that t2 must be 0 for the average speed to be 6 mph

 now from v2 = d/t2 we see that if t2 is 0 then v2 is undefined (division by zero)

 This indicates that it's impossible to achieve an average speed of 6 mph for the entire trip, no matter how fast you run back

### m. Reflection

* The analysis is logically sound and adheres strictly to the information provided in the problem statement
* The conclusion highlights the impossibility of achieving the desired average speed, which is a direct consequence of the mathematical relationships involved
* The problem, as presented, lacks a realistic solution

### Recommendation:

It's impossible to double the average speed for the entire trip, no matter how fast you run back. This is because the average speed is influenced by both the time and distance of each leg of the journey. To significantly increase the average speed, you would need to either drastically reduce the time taken for the return trip or increase the total distance covered.
\end{lstlisting}
\end{quote}
\noindent\hrulefill

\noindent\textbf{D.30 Zero-Shot Claude 3.5 Sonnet response - Score: 100\%}
\begin{quote}
\begin{lstlisting}[basicstyle=\ttfamily\small, breaklines=true]
Let's approach this problem step-by-step:

1) Let's say the distance to your friend's house is x miles.

2) Time taken to walk there:
   Time (walk) = x / 3 hours

3) To double the average speed for the entire trip, you need to cover 2x distance in the same total time. So:
   Total time = Time (walk) + Time (run) = x / 3

4) Distance covered while running = x miles (same as walking distance)

5) We can set up an equation:
   x / 3 = x / 3 + x / y, where y is the speed you need to run

6) Simplifying:
   x / 3 = (xy + 3x) / (3y)

7) Cross-multiplying:
   xy = xy + 3x
   0 = 3x
   
This is impossible unless x = 0, which means no distance was traveled.

Therefore, it's actually impossible to double your average speed for the entire trip by running back, no matter how fast you run.

The reason for this is that you've already "locked in" an average speed of 3 mph for half the total distance. To double the overall average speed, you would need to travel infinitely fast on the way back, which is impossible.

Would you like me to explain this concept further or provide any additional information?
\end{lstlisting}
\end{quote}
\noindent\hrulefill

\noindent\textbf{D.30 REAP Claude 3.5 Sonnet Response - Score: 100\%}
\begin{quote}
\begin{lstlisting}[basicstyle=\ttfamily\small, breaklines=true]
#REAP Results:

a. **Comprehensive Feature List** (with explicit quotes):
1. "I walk to my friend's house" - Describes initial action and destination
2. "averaging 3mph" - Specifies speed of walking
3. "How fast would I have to run back" - Indicates return journey and asks about required speed
4. "to double my average speed for the entire trip" - States the goal for overall average speed

b. **Sequential and Mechanical Process Check**:
The problem describes a two-part journey:
1. Walking to friend's house
2. Running back from friend's house
The goal is to achieve a specific average speed across both parts of the journey.

c. **Key Insight Check Results**:
The key insight is that doubling the average speed for the entire trip means the return journey must be significantly faster to compensate for the slower walking speed of the first half.

d. **Rephrased Question:** 
What running speed is required for the return journey to make the average speed of the entire round trip twice the walking speed?

e. **Known and Deduced Information**:
Explicit Facts:
"I walk to my friend's house" - Describes the first part of the journey
"averaging 3mph" - Specifies the speed of the walking portion
"How fast would I have to run back" - Indicates a return journey is involved
"to double my average speed for the entire trip" - Specifies the goal for the overall average speed

Feature-Based Deduction:
Deduction: The desired average speed for the entire trip is 6mph.
Logical Basis: The problem states the goal is to double the average speed, and the walking speed is given as 3mph.
Based on: "averaging 3mph", "to double my average speed for the entire trip"

f. **Problem Decomposition**:
Component 1: Walking to friend's house
Supporting Quote: "I walk to my friend's house"
Subcomponent 1a: Walking speed
Supporting Quote: "averaging 3mph"

Component 2: Running back from friend's house
Supporting Quote: "How fast would I have to run back"

Component 3: Overall average speed goal
Supporting Quote: "to double my average speed for the entire trip"

g. **Graph of Thought**:
Nodes:
1. Walking to friend's house (3mph)
2. Running back from friend's house (speed unknown)
3. Overall average speed (goal: 6mph)

Edges:
1 -> 3: Walking speed contributes to overall average
2 -> 3: Running speed contributes to overall average
1 <-> 2: Distance is the same for both journeys

h. **Spatial and Object Analysis**:
Objects: Friend's house (implied destination)
Spatial Relationships: The problem implies a linear path between starting point and friend's house
Movement: Walking in one direction, running in the opposite direction
The problem does not provide enough information for a complete spatial analysis, such as the exact distance to the friend's house.

i. **Bayesian Updates**:
No new information is provided beyond the initial problem statement, so no Bayesian updates are necessary.

j. **Ethical Check and Decision-Making Under Uncertainty**:
This problem does not involve any ethical considerations or severe risks. It is a straightforward mathematical problem without safety implications.

k. **Multiple Solutions**:
The problem is a mathematical one with a single correct answer. Multiple solutions are not applicable in this case.

l. **Quickest and Easiest Solution**:
To find the required running speed, we need to use the concept of average speed. The average speed for the entire trip should be 6mph (double the walking speed). Let's call the distance to the friend's house 'd' and the time for the entire round trip 't'.

Total distance = 2d (to and from friend's house)
Total time = t (unknown)
Average speed = Total distance / Total time = 6mph

So, 6 = 2d / t

For the walking portion:
Time walking = d / 3 (as walking speed is 3mph)

For the running portion:
Time running = t - (d / 3)

Now, we can set up an equation:
d + d = 3(d / 3) + x(t - d / 3), where x is the running speed

Simplifying:
2d = d + x(t - d / 3)
d = x(t - d / 3)

Substituting t = 2d / 6 from our average speed equation:
d = x(2d / 6 - d / 3)
d = 0

This is a contradiction, which means our initial assumption that we can achieve the desired average speed is incorrect.

m. **Reflection**:
Upon reflection, we realize that it's mathematically impossible to achieve the desired average speed. No matter how fast the return journey is, it cannot compensate enough for the slower first half of the trip to double the overall average speed.

**Recommendation**:
Based on the mathematical analysis, it is impossible to run fast enough on the return journey to double the average speed for the entire trip. The question, as stated, has no solution. To double the average speed of 3mph to 6mph for the entire trip, one would need to cover the return journey instantaneously (infinitely fast), which is physically impossible.

This conclusion is derived solely from the given information and basic principles of speed, distance, and time, without making any assumptions beyond what was explicitly stated in the problem.
\end{lstlisting}
\end{quote}
\noindent\hrulefill

 \end{document}